\documentclass[11pt]{article}

\usepackage[final]{acl}

\usepackage{times}
\usepackage{latexsym}

\usepackage[T1]{fontenc}

\usepackage[utf8]{inputenc}

\usepackage{microtype}

\usepackage{inconsolata}

\usepackage{graphicx}

\usepackage{url}

\usepackage{latexsym}

\usepackage{microtype}

\usepackage{latexsym}
\usepackage{lingmacros}
\usepackage{amsmath}
\usepackage{graphicx}
\usepackage{array, tabularx, caption, boldline}
\usepackage{cellspace}
\usepackage{url}
\usepackage{makecell}
\usepackage{multirow}
\usepackage{xcolor}

\usepackage{subcaption}
\usepackage{booktabs}
\usepackage{comment}
\usepackage{arydshln}
\usepackage{amssymb}

\usepackage{gb4e}
\noautomath

\usepackage{pifont}

\definecolor{CBI1}{HTML}{636EFA}
\definecolor{CBI2}{HTML}{EF553B}
\definecolor{CBI3}{HTML}{00CC96}
\definecolor{CBI4}{HTML}{AB63FA}
\definecolor{CBI5}{HTML}{FFA15A}
\definecolor{CBI6}{HTML}{19D3F3}
\definecolor{CBI7}{HTML}{FF6692}

\newcommand{\secref}[1]{(\S\ref{#1})}
\newcommand{\secreftwo}[2]{(#2 \S\ref{#1})}

\usepackage{amssymb}
\usepackage{mathtools}
\usepackage{mathabx}
\usepackage{wasysym}

\newdimen\supsymwidth
\newdimen\supsymheight
\newdimen\tgtsymwidth
\setbox0=\hbox{$_\bigboxvoid$}
\supsymwidth=\wd0
\supsymheight=\ht0
\setbox0=\hbox{$_\bigovoid$}
\tgtsymwidth=\wd0

\newcommand{\supsym}[1]{%
  \mathrlap{\bigboxvoid}
  \raisebox{0.5pt}{\hbox to \supsymwidth{\hfill{\tiny#1}\hfill}}
}

\newcommand{\tgtsym}[1]{
  \mathrlap{\bigovoid}
  \raisebox{0.5pt}{\hbox to \supsymwidth{\hfill{\tiny#1}\hfill}}
}

\newcommand{\superordinate}[2][\relax]{#2$_{\ifx#1\relax\bigovoid\else\supsym{#1}\fi}$}

\title{Cell-Based Representation of Relational Binding in Language Models}

\author{Qin Dai$^1$, Benjamin Heinzerling$^{2,1}$, Kentaro Inui$^{3,1,2}$ \\ $^1$Tohoku University \qquad $^2$RIKEN AIP \qquad $^3$MBZUAI\\ \tt qin.dai.b8@tohoku.ac.jp, benjamin.heinzerling@riken.jp \\ \tt kentaro.inui@mbzuai.ac.ae}

\begin{document}
\maketitle
\begin{abstract}
Understanding a discourse requires tracking entities and the relations that hold between them. While Large Language Models (LLMs) perform well on relational reasoning, the mechanism by which they bind entities, relations, and attributes remains unclear. We study discourse-level relational binding and show that LLMs encode it via a Cell-based Binding Representation (CBR): a low-dimensional linear subspace in which each ``cell'' corresponds to an entity--relation index pair, and bound attributes are retrieved from the corresponding cell during inference. Using controlled multi-sentence data annotated with entity and relation indices, we identify the CBR subspace by decoding these indices from attribute-token activations with Partial Least Squares regression. Across domains and two model families, the indices are linearly decodable and form a grid-like geometry in the projected space. We further find that context-specific CBR representations are related by translation vectors in activation space, enabling cross-context transfer. Finally, activation patching shows that manipulating this subspace systematically changes relational predictions and that perturbing it disrupts performance, providing causal evidence that LLMs rely on CBR for relational binding. Code and data are available at \url{https://github.com/cl-tohoku/CBR-Subsapce}.
\end{abstract}

\section{Introduction}
\label{sec:introduction}

\begin{figure*}[t]
\centering
\includegraphics[width=15.0cm]{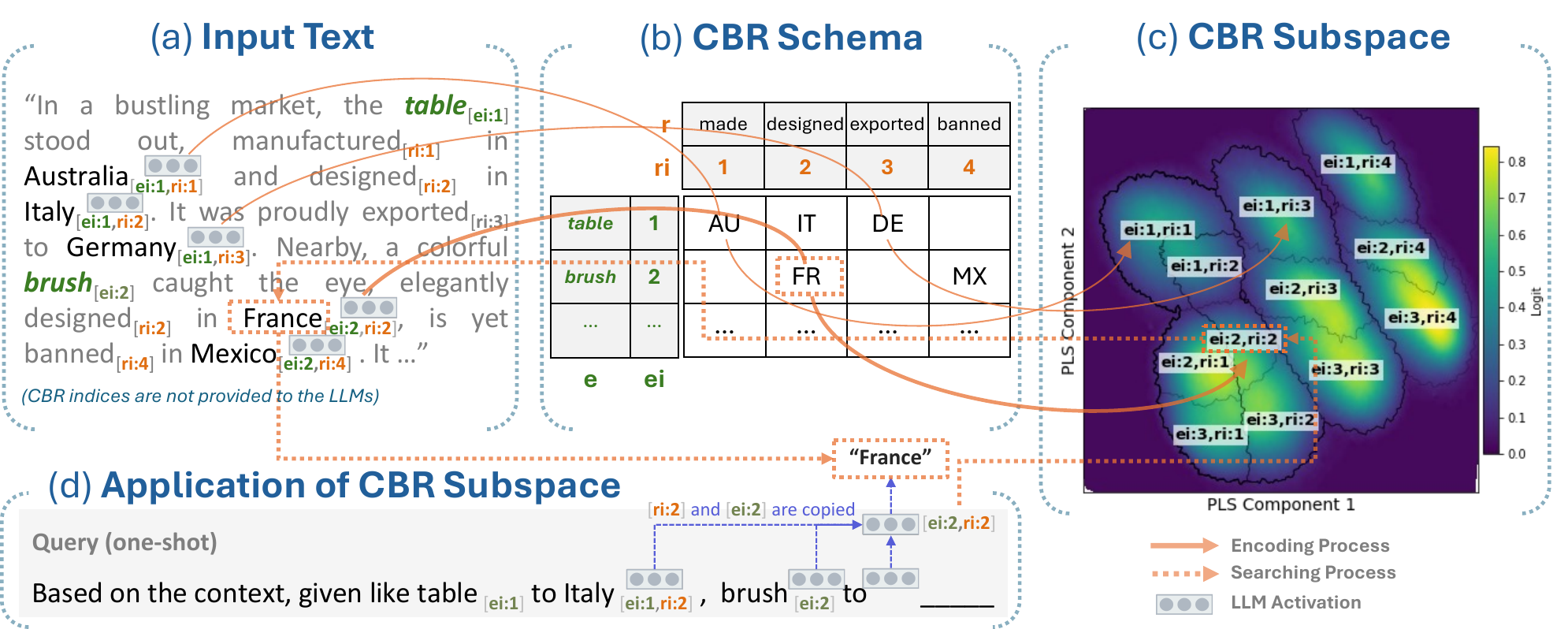}
\caption{
Overview of our Cell-based Binding Representation (CBR):
(a) discourse annotated with entity and relation indices;
(b) CBR schema, called Indexed Relational Schema, that binds entity index $ei$ and relation index $ri$ to their corresponding attributes;
(c) visualization of the cells corresponding to each entity-relation index pair; and
(d) cell–based retrieval, in which the model projects the query \textit{brush} and hidden relation \textit{designed in} onto the cell $[ei:2, ri:2]$ to retrieve the answer \textit{France}.
}
\label{fig:mechanism}
\end{figure*}

A core requirement for language comprehension is to keep track of entities and the relations between them as a discourse unfolds \citep{webber1979formal,van1983strategies,zwaan1998situation}.
It is believed that comprehenders achieve this via a fundamental ``binding'' operation that, on some representational level, ``binds together'' the internal representations of entities among which a discourse relation holds \citep{treisman1996binding}. For example, a reader may bind their internal representation of the \textit{table} in Figure~\ref{fig:mechanism} (a) to that of \textit{Australia} since the \textit{manufactured in} relation holds between these two entities.
Recent work has found evidence that Large Language Models (LLMs) are able to track entities across discourse and has started to uncover mechanisms supporting relational binding \citep{feng2023language,kim2023entity,feng2024monitoring,dai2024representational,gur2025mixing}.
However, this line of research has primarily focused on very short texts involving only a small number of entities \secreftwo{sec:relwork}{see related work in}, leaving discourse-level relational binding in LLMs largely unexplored.
Here, we extend the scope of analysis towards discourse-level relational structures spanning multiple sentences and involving multiple entity–relation bindings, and show that relational binding in LLMs can be understood in terms of what we call \textbf{Cell-based Binding Representation (CBR)}.

A CBR consists of cells that are arranged in a more or less grid-like pattern in a linear subspace of activation space, with each cell corresponding to an entity-relation pair that the model will decode to its bound entity (called attribute) during inference. We assume relational triples of the form $(e, r, a)$, where $e$ denotes an entity, $r$ a relation, and $a$ an attribute. The ordering reflects that the attribute $a$ is contextualized and bound to the entity $e$ under relation $r$ (e.g., (\textit{table}, \textit{manufactured in}, \textit{Australia}) in Figure~\ref{fig:mechanism} (a)), consistent with the autoregressive encoding of LLMs in which later tokens are represented conditioned on the preceding context.
As we will show, a CBR abstracts discourse-level relational structure into discrete entity indices $ei$ and relation indices $ri$, enabling attributes to be represented as bound to specific $[ei,ri]$ pairs. 

Furthermore, these indices are linearly decodable from model activations, revealing a low-dimensional and interpretable relational binding subspace organized along two dominant directions corresponding to entity indices $ei$ and relation indices $ri$.
We further show that context-specific CBR representations are related through translation vectors in activation space, enabling cross-context transfer.
Finally, through causal interventions using activation patching, we demonstrate that manipulating activations within this subspace systematically changes relational predictions, providing evidence that LLMs actively use this cell-based representational mechanism to bind and retrieve relational information over discourse.

\section{Background: Relational Binding in Linguistics and LLMs}
\label{sec:relwork}

In linguistics, relational binding is a core feature of semantic formalisms such as Discourse Representation Theory \citep[DRT][]{heim1982semantics,kamp2013theory}.
For example, the text shown in Figure~\ref{fig:mechanism} (a) can be represented in DRT as follows:
$[x, y ; \textit{table}(x), \textit{brush}(y), \\ \textit{manufactured\_in}(x, \textit{Australia}), ...,\\ \textit{designed\_in}(y, \textit{France}), ...]$.
Here, entity $x$ and \textit{Australia} are bound to form a \textit{manufactured\_in} relation, while $y$ is relationally bound to \textit{France} through the predicate \textit{designed\_in}.

Recent work has found evidence that LLMs are able to track entities across discourse and has started to uncover mechanisms supporting relational binding.
\citet{kim2023entity} provide behavioral evidence that LLMs can learn to track entity states through sequences of state-changing operations, but also show that this capacity is not consistently present in text-only pretrained models and degrades as settings become more complex.
\citet{feng2023language} identify a Binding ID mechanism which binds entity and attribute representations, and \citet{feng2024monitoring} build on this by extracting explicit logical propositions from internal activations using propositional probes in a binding subspace.
\citet{dai2024representational} refine the Binding ID picture by identifying an Ordering ID that causally controls binding, while \citet{gur2025mixing} show that models rely on a mixture of positional, lexical, and reflexive mechanisms.

Despite these advances, prior work has important limitations for understanding \emph{discourse-level relational binding}.
Much of the evidence and analysis is derived from very short contexts, typically involving a small number of entity-attribute pairs.
This leaves open whether the same mechanisms scale to multi-sentence discourse, where relational structure is richer and where models must maintain multiple relations per entity and integrate information across sentences.
Moreover, prior accounts focus on binding as a one-dimensional phenomenon (e.g., entity-to-attribute or order-based binding), whereas discourse-level semantics requires bindings indexed not just by \emph{which entity}, but also by \emph{which relation} holds for that entity.

Our work builds directly on these mechanistic insights, especially the idea that binding information is encoded in a low-dimensional and interpretable subspace and can be tested via causal interventions, but extends them to discourse-level relational structures spanning multiple sentences.
Concretely, we show that relational binding in LLMs can be understood in terms of a Cell-based Binding Representation (CBR): a grid-like organization of activation space in which each ``cell'' corresponds to an entity-relation pair $[ei,ri]$ that can be decoded to its bound attribute during inference.

\begin{figure*}[t]
    \centering 
\begin{subfigure}{0.31\textwidth}
  \includegraphics[width=\linewidth]{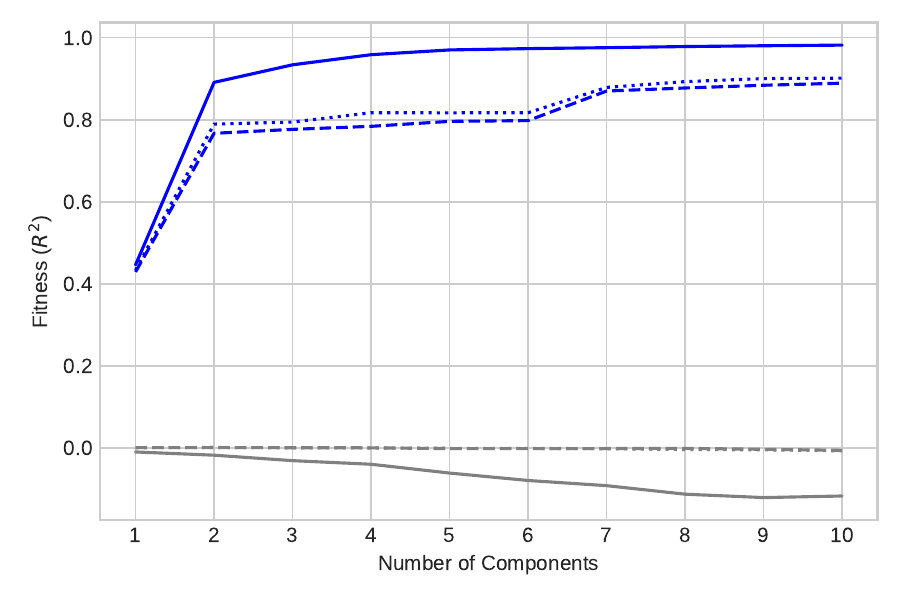}
\end{subfigure}\hfil 
\begin{subfigure}{0.31\textwidth}
  \includegraphics[width=\linewidth]{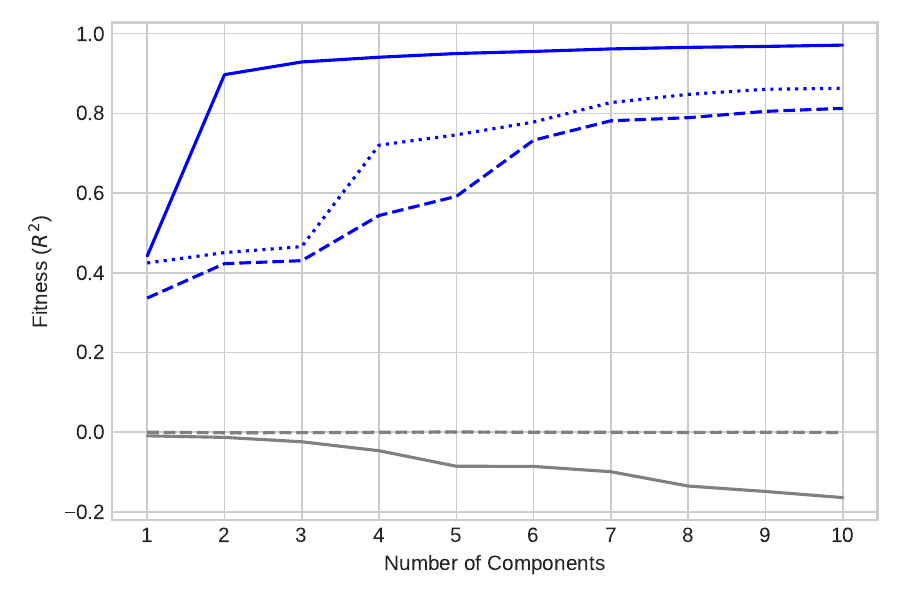}
\end{subfigure}\hfil 
\medskip
\begin{subfigure}{0.31\textwidth}
  \includegraphics[width=\linewidth]{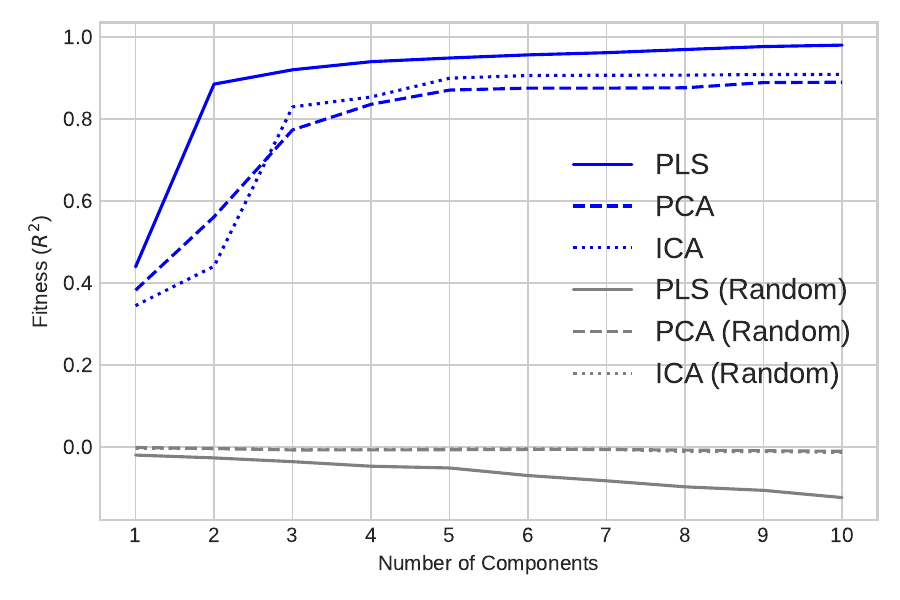}
\end{subfigure}\hfil
\begin{subfigure}{0.31\textwidth}
  \includegraphics[width=\linewidth]{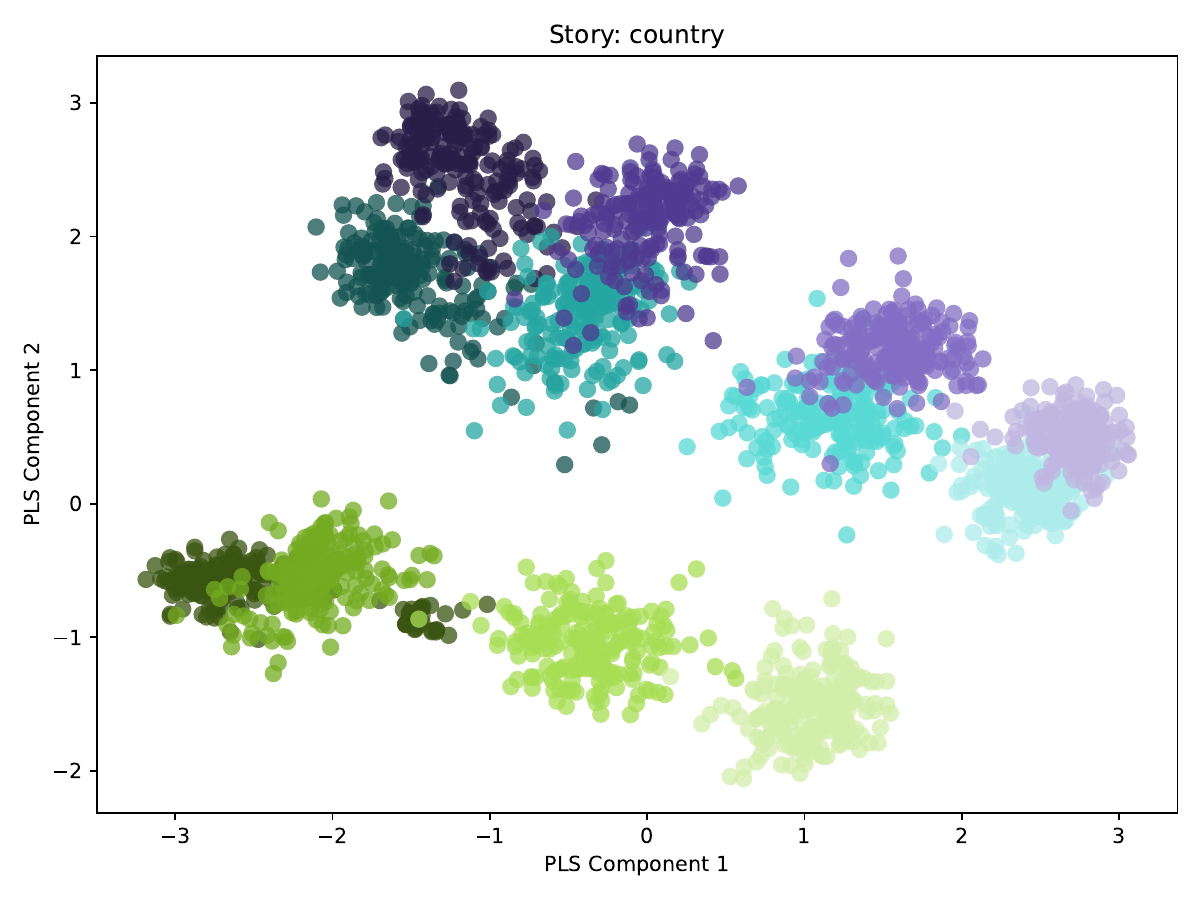}
  \caption{Domain: \textit{country}}
\end{subfigure}\hfil 
\begin{subfigure}{0.31\textwidth}
  \includegraphics[width=\linewidth]{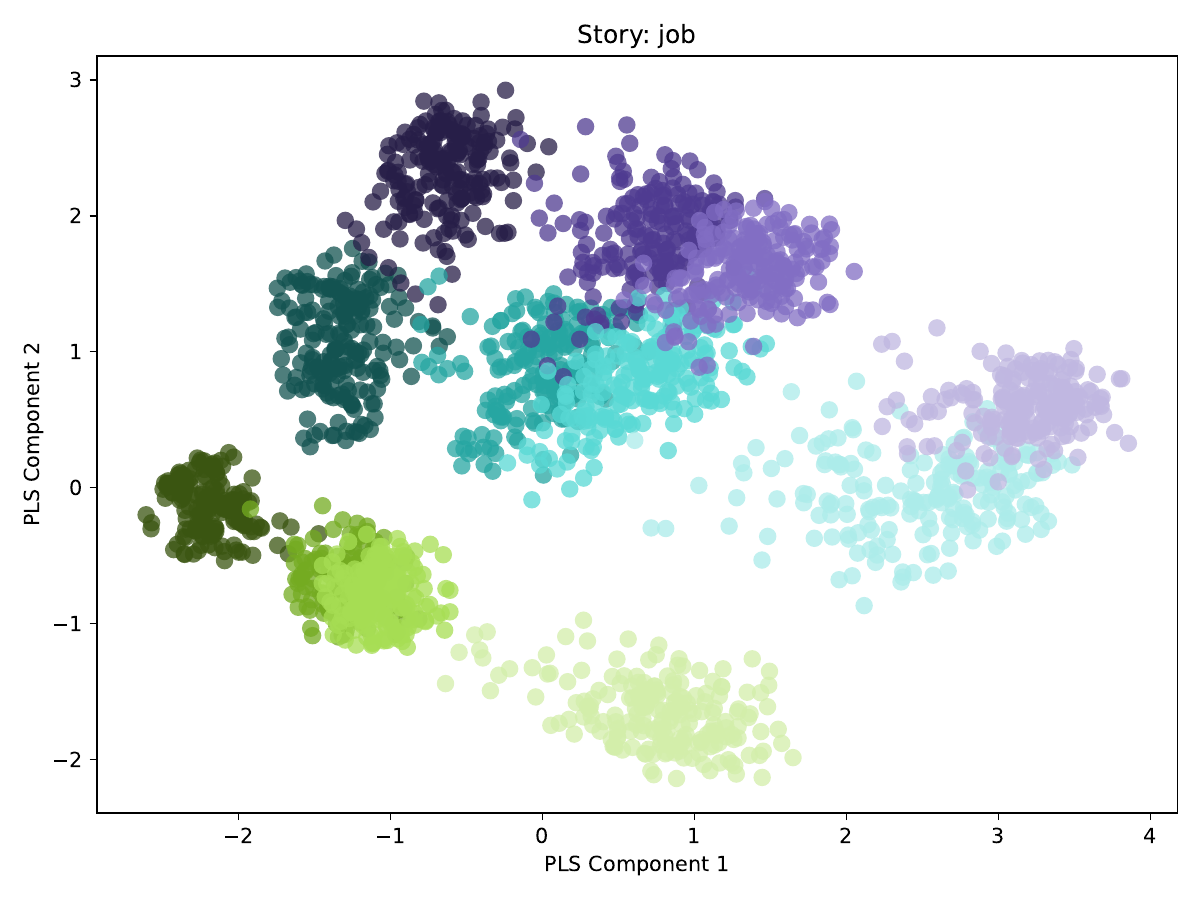}
  \caption{Domain: \textit{job}}
\end{subfigure}\hfil 
\begin{subfigure}{0.31\textwidth}
  \includegraphics[width=\linewidth]{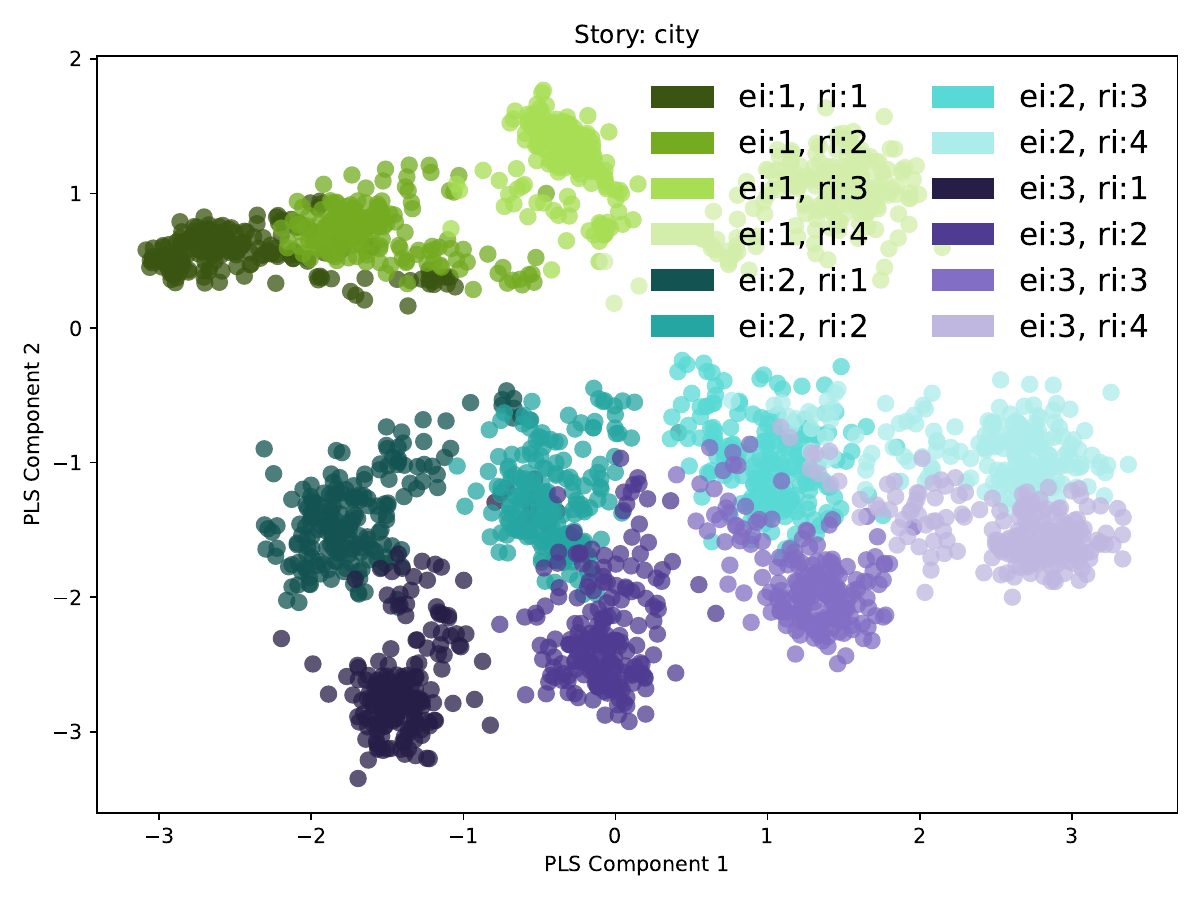}
  \caption{Domain: \textit{city}}
\end{subfigure}\hfil
\caption{
\textbf{Top:} PLS goodness-of-fit when predicting entity and relation indices $[ei,ri]$ from Llama3-8B-Instruct attribute activations across three domains. For comparison, we also fit a Principal Component Analysis regression (PCA), Independent Component Analysis (ICA) regression and include random-label controls.
\textbf{Bottom:} Visualization of attribute activations projected onto the top two PLS components, showing a grid-like distribution organized by entity index ($ei$) and relation index ($ri$).
}
\label{fig:pls_fit_and_vis}
\end{figure*}

\section{Cell-based Binding Representation}

This section presents our main results, proceeding in three steps.
First, we identify and visualize the CBR subspace \secref{sec:irs_subspace}.
Second, we verify its causal relevance \secref{sec:causal_intervention}.
Finally, we analyze the generality and robustness of the CBR subspace across domains, templates, and model families \secref{sec:generality}.

\subsection{Identifying the CBR Subspace}
\label{sec:irs_subspace}

Motivated by prior work on linear representations of relational binding in LLMs~\citep{feng2023language,feng2024monitoring,dai2024representational}, we hypothesize that LLMs encode discourse-level relational binding in a low-dimensional linear subspace of activation space and hence devise a linear probing method for identifying this subspace.
Following \citet{dai2024representational}, we assume that this subspace encodes the indices of entities and relation types according to their order in the discourse.

We formalize entity and relation indices as what we term an Indexed Relational Scheme (IRS).
In contrast to lexicalized discourse representations such as DRT, which encode relations using explicit predicate symbols (e.g., $\textit{manufactured\_in}(x,\textit{Australia})$), the IRS represents relational structure in a \emph{delexicalized} form using discrete indices.
Specifically, an IRS abstracts discourse-level bindings into \emph{entity indices} $ei$ and \emph{relation indices} $ri$, corresponding to the order in which entities and relation types are introduced, and associates each attribute token with a specific index pair $[ei,ri]$.
Formally, an IRS is a set of indexed triples $\{(ei,ri,a_{[ei,ri]})\}$, where $a_{[ei,ri]} \in A$ is an attribute bound to the entity-relation cell $(ei,ri)$.
For example, in Figure~\ref{fig:mechanism} (b), \textit{Australia} is represented as bound to $[ei\!:\!1,\,ri\!:\!1]$ and \textit{France} to $[ei\!:\!2,\,ri\!:\!2]$.
This delexicalized indexing scheme provides a compact and testable target for probing how LLM activations encode relational bindings over discourse.

\paragraph{Data and models.}
We automatically generate multi-sentence discourses containing multiple entities, multiple relation types, and multiple attributes per entity, while ensuring that each attribute is uniquely bound to a specific \textit{(entity, relation)} pair.
Each discourse introduces entities and relation types in a controlled order, yielding a ground-truth IRS annotation for every attribute token, i.e., its entity index $ei$ and relation index $ri$.

To test robustness across semantic domains and surface forms, we construct five discourse contexts that vary in entity, relation, and attribute inventories (e.g., countries, cities, occupations, and objects).
We additionally use multiple templates and naturalistic paraphrased variants for each context to reduce reliance on superficial, repetitive positional cues and to better capture real-world discourse patterns.
Appendix~\S\ref{sec:dataset} provides full details on templates, sampling procedures, and dataset statistics.
Overall, our experiments cover five domains~\footnote{In this paper, the terms ``context'' and ``domain'' are used interchangeably.} and two model families.

\paragraph{Method.}
To identify the CBR subspace, we fit a Partial Least Squares \citep[PLS;][]{wold2001pls} regression model to map activations onto entity and relation indices $ei$ and $ri$.
For each attribute token in a discourse, we collect its activation vector $\mathbf{h}_i \in \mathbb{R}^d$ from a middle layer (layer 15 in our case), where $d$ denotes the dimensionality of the activation.
Each activation vector is paired with its index label $\mathbf{y}_i \in \mathbb{R}^2$ (i.e., $[ei,ri]$).
Stacking all activations yields the matrix $H$ and stacking all labels yields $Y$, and fitting a PLS yields a projection matrix $W_{\text{CBR}}$ that projects the model activations $H$ onto the directions in which $H$ and $Y$ maximally covary.
In other words, PLS identifies the low-dimensional directions in activation space that are most predictive of the entity and relation indices, thereby giving us a candidate CBR subspace.

\paragraph{Results.}
We vary the number of components of PLS models and evaluate using goodness of fit.
As shown for Llama3-8B-Instruct across three domains in Figure~\ref{fig:pls_fit_and_vis} (top) and for all models and domains in \secref{sec:sec:irs_subspace_llama} and \secref{sec:irs_subspace_qwen}, PLS models achieve near-perfect fits with a small number of components, i.e., both entity and relation indices can be linearly decoded from low-dimensional subspace of activation space. 
We further evaluate decoding performance under monotonic transformations of the indices, details are provided in Appendix~\secref{sec:mono}

Projecting the activations of attribute tokens (e.g., ``Australia’’ in Figure~\ref{fig:mechanism} (a)) onto the top two PLS components, we obtain the visualization shown in Figure~\ref{fig:pls_fit_and_vis} (bottom) for Llama3-8B-Instruct and Figure~\ref{fig:irs_visualization_qwen} \secref{sec:irs_visualization_qwen} for Qwen3-8B. 
The plots reveal two dominant and interpretable directions: one that separates the points by $ei$ and another that separates them by $ri$.
Attributes associated with the same entity cluster along the $ei$ direction, while those participating in the same relation align along the $ri$ direction.
These patterns suggest that LLMs encode relational binding structure in a low-dimensional linear subspace, supporting our hypothesis of a CBR subspace that jointly represents entity and relation indices.

\begin{figure}[t]
\centering
\begin{subfigure}{0.485\linewidth}
\centering
\includegraphics[width=\linewidth]{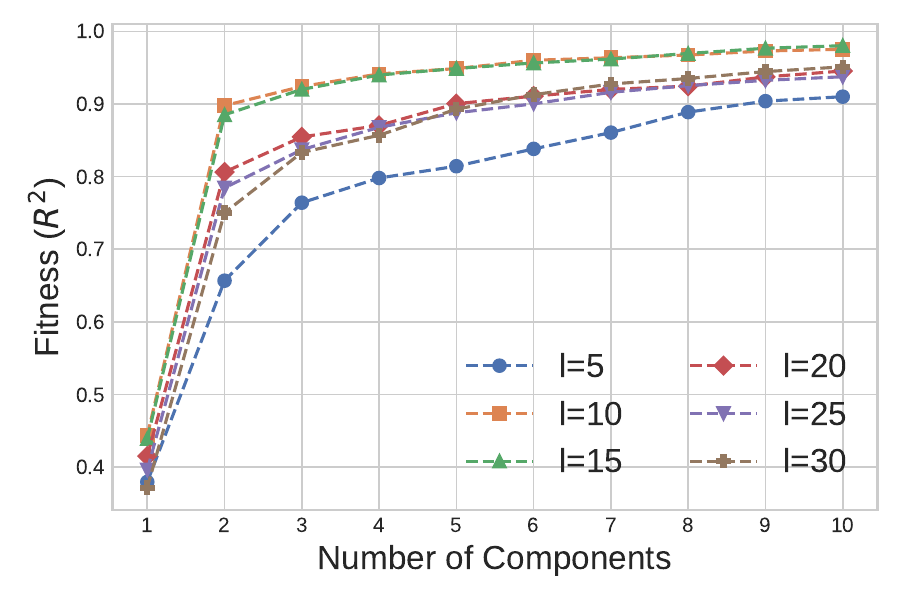}
\caption{PLS}
\end{subfigure}
\begin{subfigure}{0.485\linewidth}
\centering
\includegraphics[width=\linewidth]{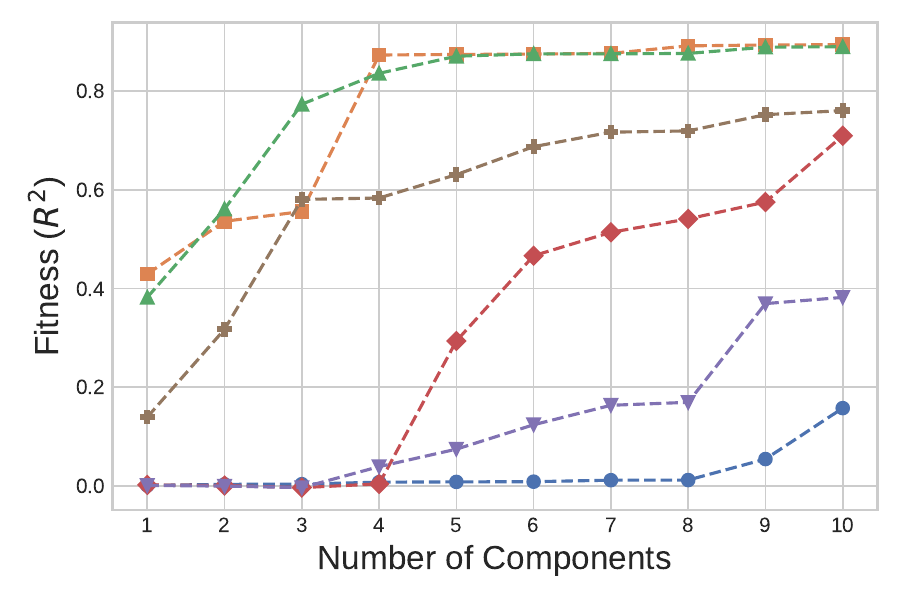}
\caption{PCA}
\end{subfigure}

\caption{Layer-wise and component-wise analysis of CBR signal strength. Predictive performance of CBR indices across layers using PLS and PCA.}
\label{fig:layerwise_llama_city}
\end{figure}

In addition, to identify where the CBR signal is most prominent, we conducted layer-wise and component-wise analyses using both PLS and PCA in \secref{sec:irs_subspace_layerwise}. An example analysis for the \textit{city} domain is shown in Figure~\ref{fig:layerwise_llama_city}. We observe that middle layers (approximately Layers 10–20) exhibit the strongest signal, with Layer 15 achieving peak or near-peak performance across contexts, while earlier and later layers show weaker alignment. In the PLS analysis, performance improves with dimensionality and stabilizes at around 2-5 components; fewer components reduce accuracy, whereas additional components yield only marginal gains. We therefore use Layer 15 and 2-5 components as empirically supported main experimental settings for the following analyses.
\begin{figure}[t]
\centering
\includegraphics[width=7.5cm]{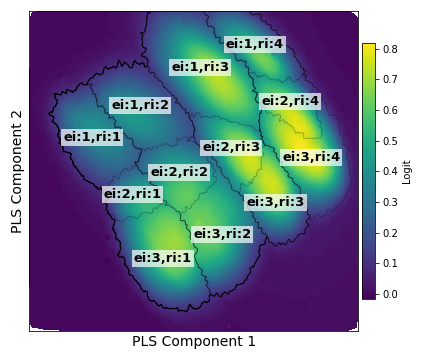}
\caption{Logit landscape of attribute predictions resulting from causal interventions in the CBR subspace. Each ``cell'' corresponds to an entity--relation index pair $[ei,ri]$, with boundaries marking where the predicted attribute switches. See detailed explanation in \S\ref{sec:causal_intervention}.
}
\label{fig:grid_llama_city}
\end{figure}
\subsection{Causal Effect of the CBR Subspace}
\label{sec:causal_intervention}
\begin{figure}[t]
    \centering 
\begin{subfigure}{0.225\textwidth}
  \includegraphics[width=\linewidth]{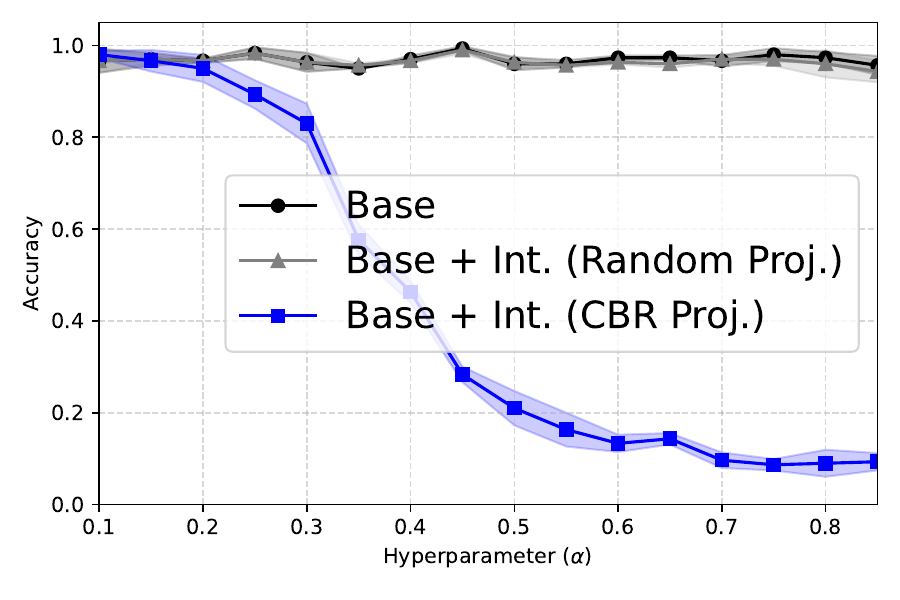}
  \caption{$C_{city}$}
\end{subfigure}\hfil 
\begin{subfigure}{0.225\textwidth}
  \includegraphics[width=\linewidth]{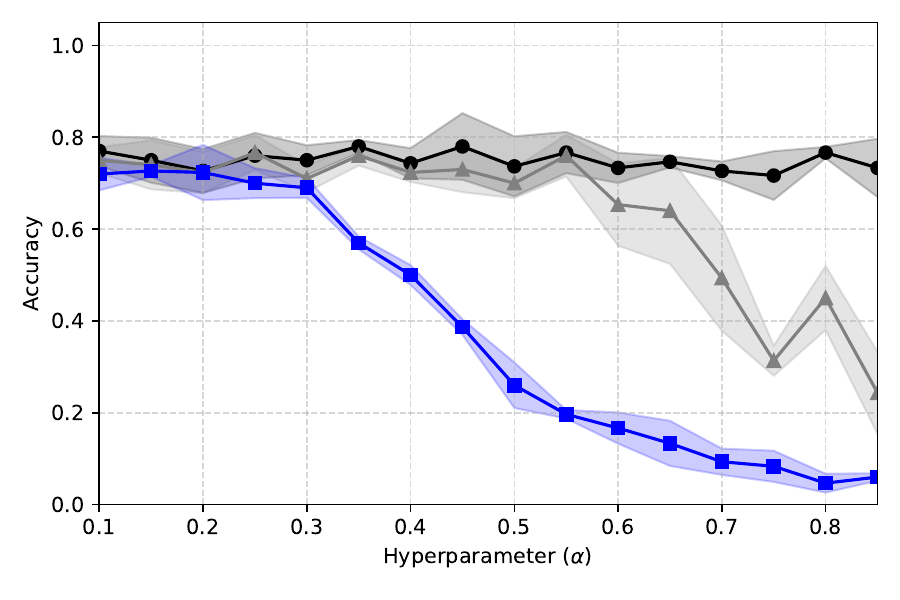}
  \caption{$C_{country}$}
\end{subfigure}\hfil 
\medskip
\begin{subfigure}{0.15\textwidth}
  \includegraphics[width=\linewidth]{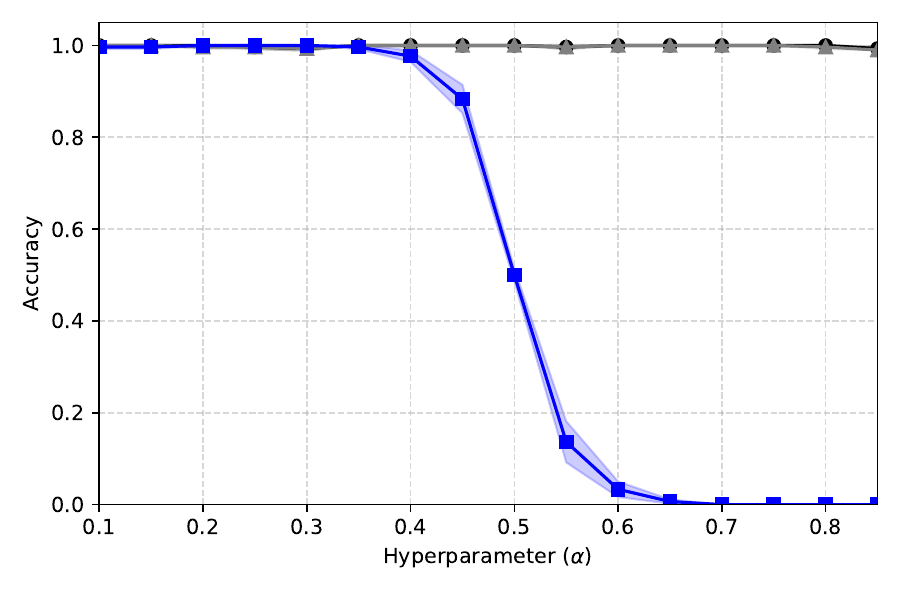}
  \caption{$C_{relation}$}
\end{subfigure}\hfil
\begin{subfigure}{0.15\textwidth}
  \includegraphics[width=\linewidth]{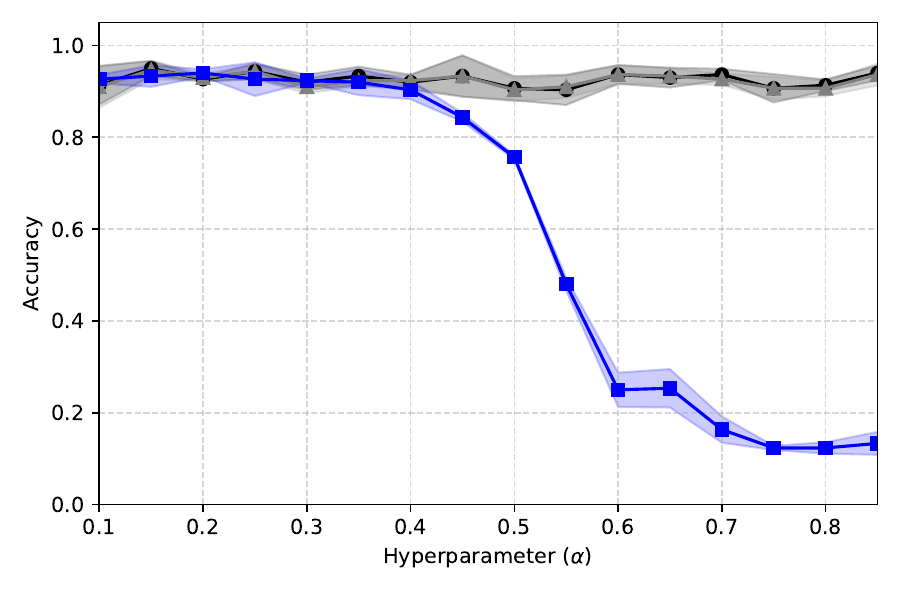}
  \caption{$C_{job}$}
\end{subfigure}\hfil 
\begin{subfigure}{0.15\textwidth}
  \includegraphics[width=\linewidth]{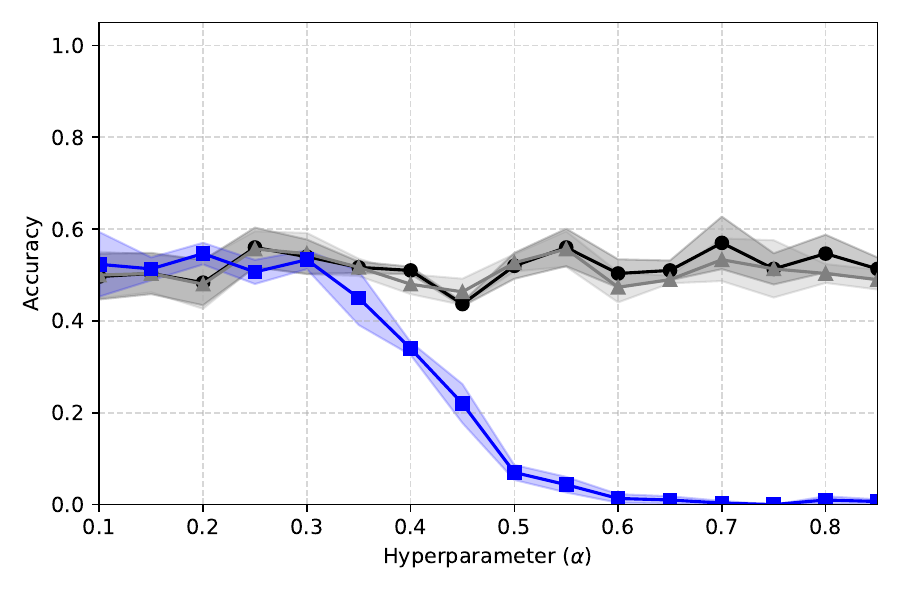}
  \caption{$C_{object}$}
\end{subfigure}\hfil 
\caption{Effect of perturbing activations along the CBR subspace versus a random subspace on Llama3-8B-Instruct. The X-axis shows the perturbation weight $\alpha$ in Equation~\ref{eq:irs_noise1} and \ref{eq:irs_noise2}, and the Y-axis shows the attribute prediction accuracy. Perturbations along the CBR subspace (i.e., blue line) lead to a significant drop in accuracy, while perturbations along a random subspace (i.e., grey line) have minimal effect. This indicates that LLMs rely on the CBR subspace to make relationally bound predictions.}
\label{fig:noise_llama}
\end{figure}

\paragraph{Effect on attribute prediction.} To understand if and how LLMs use the CBR subspace for relational binding, we perform causal interventions via activation patching \citep{vig2020investigating,geiger2020neural,geiger2021causal,wang2022interpretability,stolfo2023mechanistic,heinzerling2024monotonic,hanna2024does}, using a one-shot query setting as shown below (\ref{ex:onshot_que}).
We opt for one-shot queries since alternatives such as providing preceding context (\ref{ex:context}) and using a direct query (\ref{ex:simple_que}) would allow the model to rely on superficial cues and alternative strategies such as context matching via induction heads~\cite{olsson2022context}.
In contrast, the one-shot query format requires the model to rely on relational structure rather than surface form, as the correct answer must be inferred from the underlying entity–relation bindings.
\begin{exe}
   \ex\label{ex:onshot_que}\textbf{Query} (one-shot): Based on the context, given like Sean to Perm, Jose to?
    \ex\label{ex:context}\textbf{Context}: Sean, who hails from Phoenix, currently resides in Perm. ... Meanwhile, Jose was born in Austin and is now living in Berlin. ...
    \ex\label{ex:simple_que}\textbf{Query} (direct): Based on the context, Jose is now living in?
\end{exe}
We now causally intervene on the CBR subspace by patching activations along the top two PLS directions, which we hypothesize to encode entity and relation indices.
Concretely, we uniformly sample $10^4$ points (denoted $p_j$) from the range defined by the minimum and maximum values of the learned CBR subspace obtained through the projection matrix $W_{\text{CBR}}$ \secreftwo{sec:irs_subspace}{see}.
The embedding of each point is denoted as $\mathbf{h}_{p_j}$. For each attribute instance in $50$ randomly selected samples (e.g., ``Berlin'' in Sample~\ref{ex:context}), we gradually move its activation (denoted as $\mathbf{h}_{ei,ri}$) towards one of the sampled target points according to Equation~\ref{eq:irs_sampling2}, where $\alpha$ is a hyperparameter and $\mathbf{h}_{ei,ri}^*$ denotes the updated activation of an attribute, effectively sweeping the activation of attribute across the CBR plane.
At each step, we compute the logit score of the corresponding attribute predicted by the LLM.
For instance, given a Context~\ref{ex:context} and a Query~\ref{ex:onshot_que} for ``Berlin'', we patch its activation in the Context and observe the logit score of the corresponding attribute ``Berlin''.
The resulting logit landscape is shown in Figure~\ref{fig:grid_llama_city}. Additional results for other datasets and corresponding results for Qwen3-8B are reported in \S\ref{sec:irs_sampling_across}.
\begin{align}
\mathbf{s}_{ei,ri \rightarrow p_j} &= \mathbf{h}_{p_j} - W_{\text{CBR}}\mathbf{h}_{ei,ri}, \label{eq:irs_sampling1} \\
\mathbf{h}_{ei,ri}^* &= \mathbf{h}_{ei,ri} + \alpha W_{\text{CBR}}^T \mathbf{s}_{ei,ri \rightarrow p_j} \label{eq:irs_sampling2} 
\end{align}
The logit landscape shows that the CBR subspace is partitioned into ``cells'' arranged in a grid-like pattern, with each cell corresponding to a specific entity–relation index pair $[ei,ri]$.
Within each cell, the attribute bound to that particular index achieves the highest logit score, and the logit value decreases smoothly as the patched activation moves away from center of the cell.
Also see the cross-section plots along $ei$ and $ri$ directions in Appendix~\ref{sec:grid_logit_analysis}.
Taken together, these results show that the identified CBR subspace is causally relevant for relation binding: Attributes are predicted based on the corresponding cell.
In other words, relational binding in LLMs is causally mediated by the geometry of the CBR subspace.
\paragraph{Control: Intervention along random directions.}
\label{sec:perturb_CBR}
To test whether LLMs rely on the CBR subspace for relational binding, we also perturb attribute activations along the CBR directions using Equation~\ref{eq:irs_noise2} and compare the results with perturbations along a random subspace defined by a randomly generated projection matrix in Equation~\ref{eq:irs_noise1}. An example of the CBR subspace distribution after the perturbation (i.e., $\mathbf{h}_{ei,ri}^*$ ) is visualized in Figure~\ref{fig:pertrub_irs_rand} \secref{sec:perturb_irs_qwen}. We use the same one-shot query setting to query each attribute in the context.
\begin{align}
\mathbf{h}_{ei,ri}^* &= \mathbf{h}_{ei,ri} + \alpha W_{\text{rand}}^T (W_\text{CBR} \mathbf{h}_{ei,ri}), \label{eq:irs_noise1} \\
\mathbf{h}_{ei,ri}^* &= \mathbf{h}_{ei,ri} + \alpha W_{\text{CBR}}^T (W_\text{CBR} \mathbf{h}_{ei,ri}) \label{eq:irs_noise2} 
\end{align}
If the LLM indeed uses the CBR subspace to make predictions, perturbing activations along this subspace should degrade performance. Figure~\ref{fig:noise_llama} shows the results as a function of perturbation strength. The corresponding results for Qwen3-8B are reported in Appendix~\secref{sec:perturb_irs_qwen}. We observe that as the perturbation weight increases, the accuracy of attribute predictions decreases significantly. In contrast, perturbation along the random subspace has little or no effect on accuracy.
These results provide strong evidence that LLMs utilize the CBR subspace when predicting attributes: disruptions in this subspace directly impair model performance, whereas unrelated directions do not.
\paragraph{CBR Subspace based Mechanism}
\label{sec:irs_steering}
To understand how LLMs use the CBR subspace to retrieve the correct attribute given a context (e.g., Sample~\ref{ex:context}) and a query (e.g., Query~\ref{ex:onshot_que}), we propose a high-level mechanism illustrated in Figure~\ref{fig:mechanism} (d). When answering a one-shot relational query, the model appears to perform two parallel operations: (i) it extracts relation index information (e.g., $ri:2$) from the attribute exemplar provided in the one-shot part, and (ii) it extracts entity index information (e.g., $ei:2$) from the query entity itself. These two indices together define a point corresponding to a cell in the CBR subspace, which is encoded into the activation of the last token. The model then uses this representation to locate the answer attribute in the context whose indices match this combination.

To test this mechanism, we perform several Activation Patching (AP) interventions that steer model activations along specific CBR subspace directions: (a) Relation-index steering that shifts $ri$ in the one-shot attribute from $ri:j$ to $ri:j+1$, as shown in Figure~\ref{fig:mechanism_att} \secref{sec:irs_steering_other}, (b) Entity-index steering that shifts $ei$ in the query entity from $ei:j$ to $ei:j+1$, as shown in Figure~\ref{fig:mechanism_ent} \secref{sec:irs_steering_other}, (c) Last-token steering that modifies $ei$ or $ri$ at the final token before prediction, as shown in Figure~\ref{fig:mechanism_last} \secref{sec:irs_steering_other}. The AP is applied using Equation~\ref{eq:irs_steer1} and \ref{eq:irs_steer2}, where $\mathbf{s}_{j \rightarrow j+1}$ is the steering vector that shifts $ri$ (or $ei$) from $j$ to $j+1$, $\mathbf{h}_{j}$ denotes the activation of a target token from middle layers~\footnote{We select Layers 10-20, and set $\mathbf{s}_{j \rightarrow j+1} \in \mathbb{R}^5$.} whose $ri$ (or $ei$) is $j$, $\alpha$ is a hyperparameter, and $W_{\text{CBR}}$ is learned from the corresponding tokens, following the method described in Section~\ref{sec:irs_subspace}.
\begin{align}
\mathbf{s}_{j \rightarrow j+1} &= \frac{1}{n}\sum_{k=1}^{n} \left( W_{\text{CBR}}\mathbf{h}_{j+1}^k - W_{\text{CBR}}\mathbf{h}_{j}^k \right), \label{eq:irs_steer1} \\
\mathbf{h}_j^* &= \mathbf{h}_j + \alpha W_{\text{CBR}}^T \mathbf{s}_{j \rightarrow j+1} \label{eq:irs_steer2}
\end{align}

\begin{figure}[t]
\centering
\includegraphics[width=7.0cm]{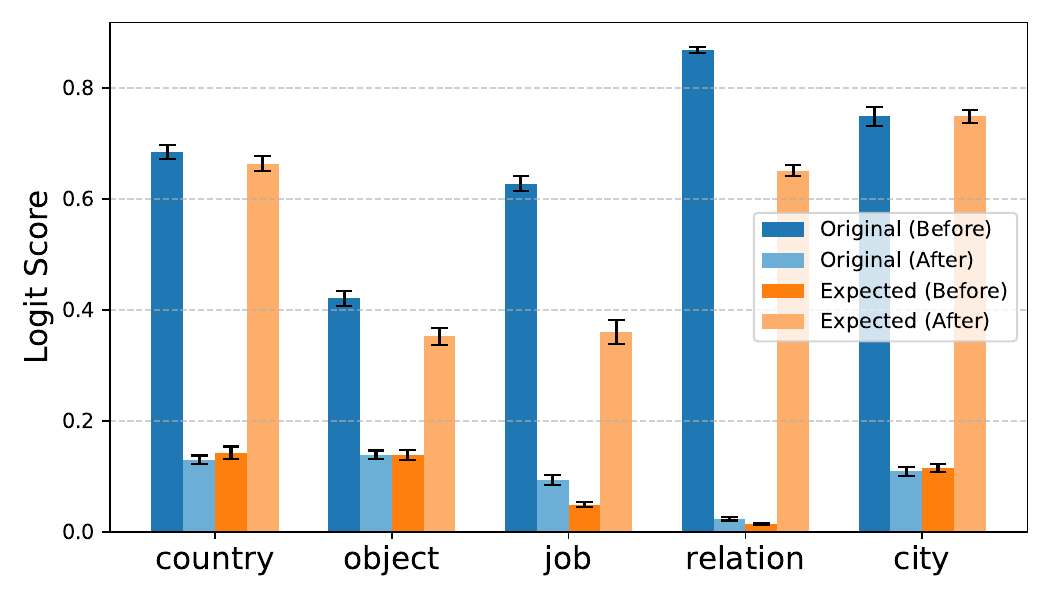}
\caption{Activation patching via Relation-index (i.e., $ri$) steering on the activation of the attribute token (e.g., ``Italy'') in query part across five contexts on Llama3-8B-Instruct. We show the change in logit scores for the original answer and the expected answer before and after activation patching, which are denoted as ``Original (Before)'', ``Original (After)'', ``Expected (Before)'' and ``Expected (After)'' respectively.}
\label{fig:steer_llama_att}
\end{figure}

Following \citet{wang2022interpretability}, we evaluate the effect of steering by measuring the change in logit scores for both the original correct answer and the expected answer after intervention. As shown in Figure~\ref{fig:steer_llama_att}, steering along the CBR subspace direction consistently suppresses the logit of the original answer and increases the logit of the expected answer, precisely in line with the intended index manipulation. The results for (b) Entity-index steering and (c) Last-token steering are reported in Appendix~\secref{sec:irs_steering_other}. The corresponding results for Qwen3-8B are presented in Appendix~\secref{sec:irs_steering_qwen}. These results demonstrate that LLMs rely on CBR-subspace representations to retrieve attributes, and that targeted activation patching along CBR directions can systematically alter the model’s predictions. 

%
\subsection{Generality of the CBR Subspace}
\label{sec:generality}

\paragraph{Does the CBR subspace encode semantics or is it merely positional?}
\begin{figure}[t]
    \centering 
\begin{subfigure}{0.15\textwidth}
  \includegraphics[width=\linewidth]{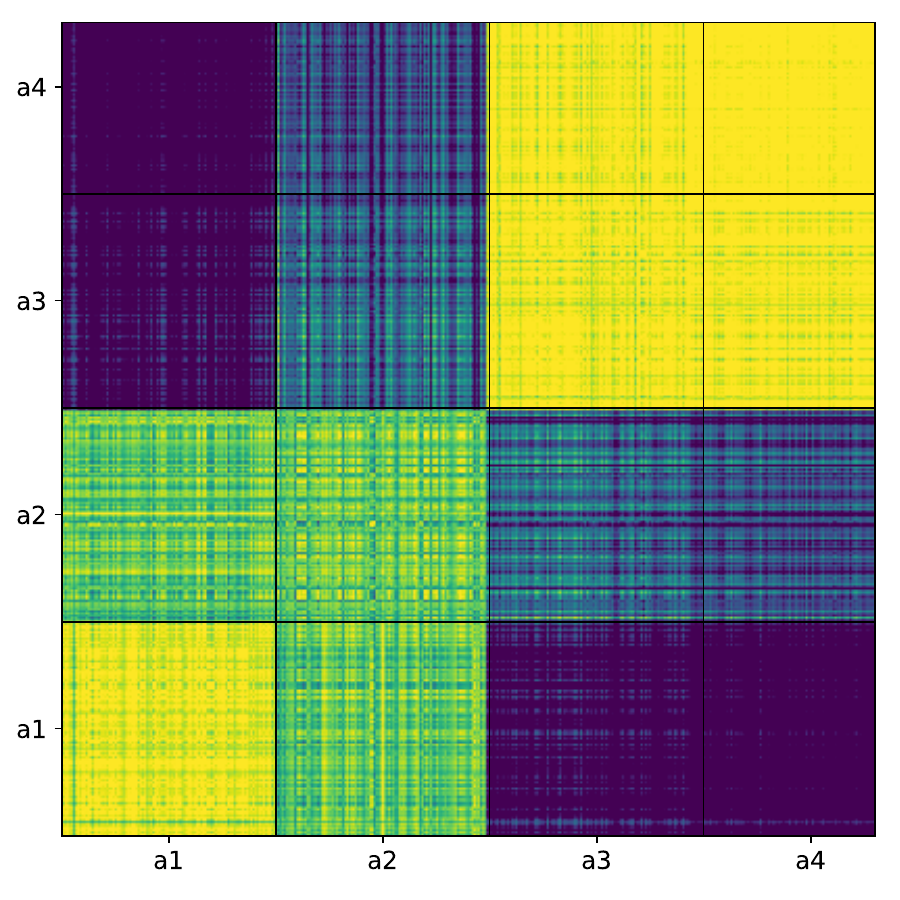}
  \caption{2to2}
\end{subfigure}\hfil 
\begin{subfigure}{0.15\textwidth}
  \includegraphics[width=\linewidth]{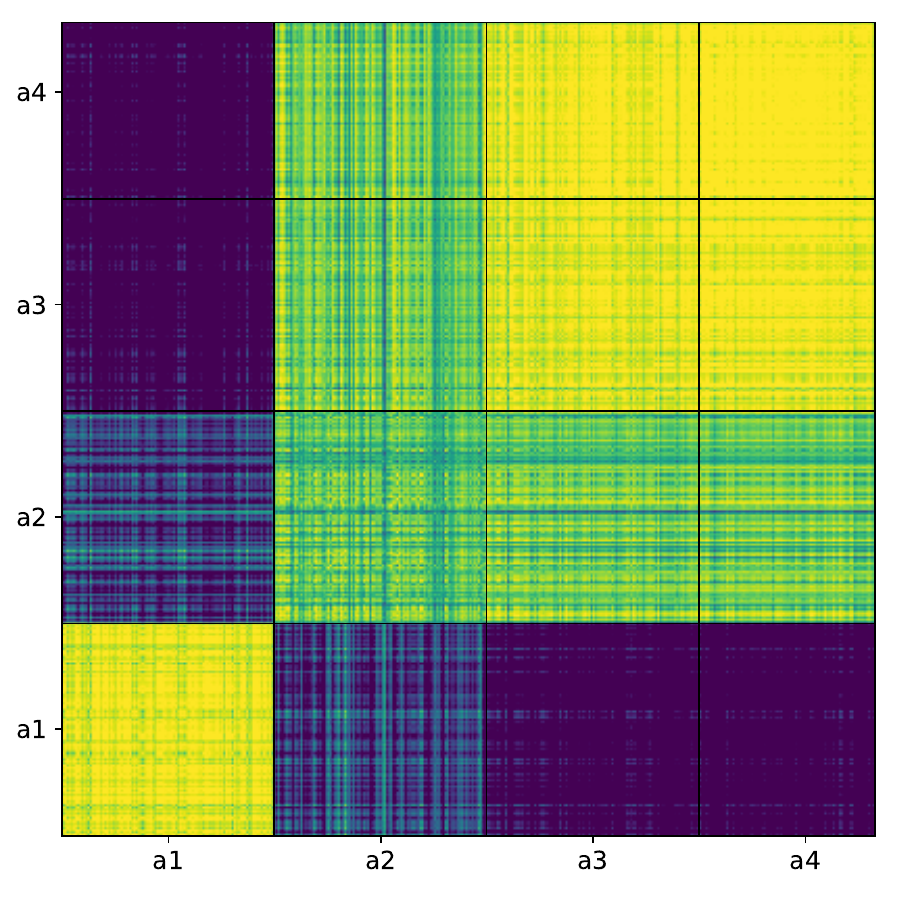}
  \caption{1to3}
\end{subfigure}\hfil 
\begin{subfigure}{0.15\textwidth}
  \includegraphics[width=\linewidth]{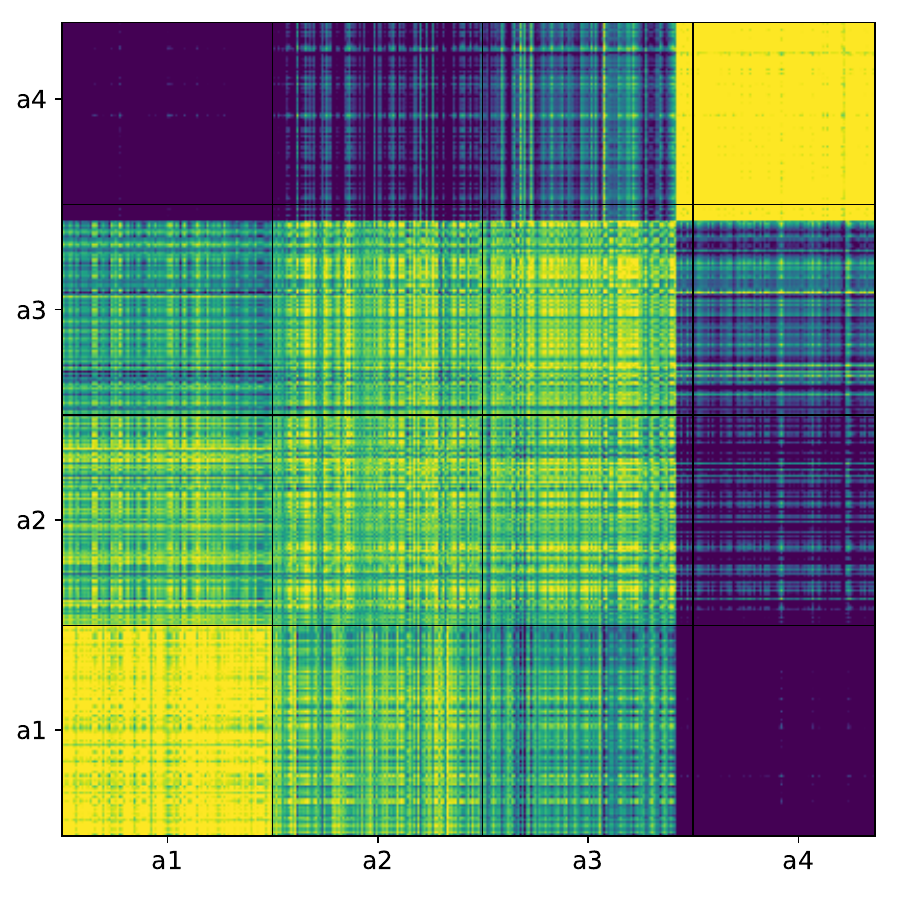}
  \caption{3to1}
\end{subfigure}

\medskip
\begin{subfigure}{0.15\textwidth}
  \includegraphics[width=\linewidth]{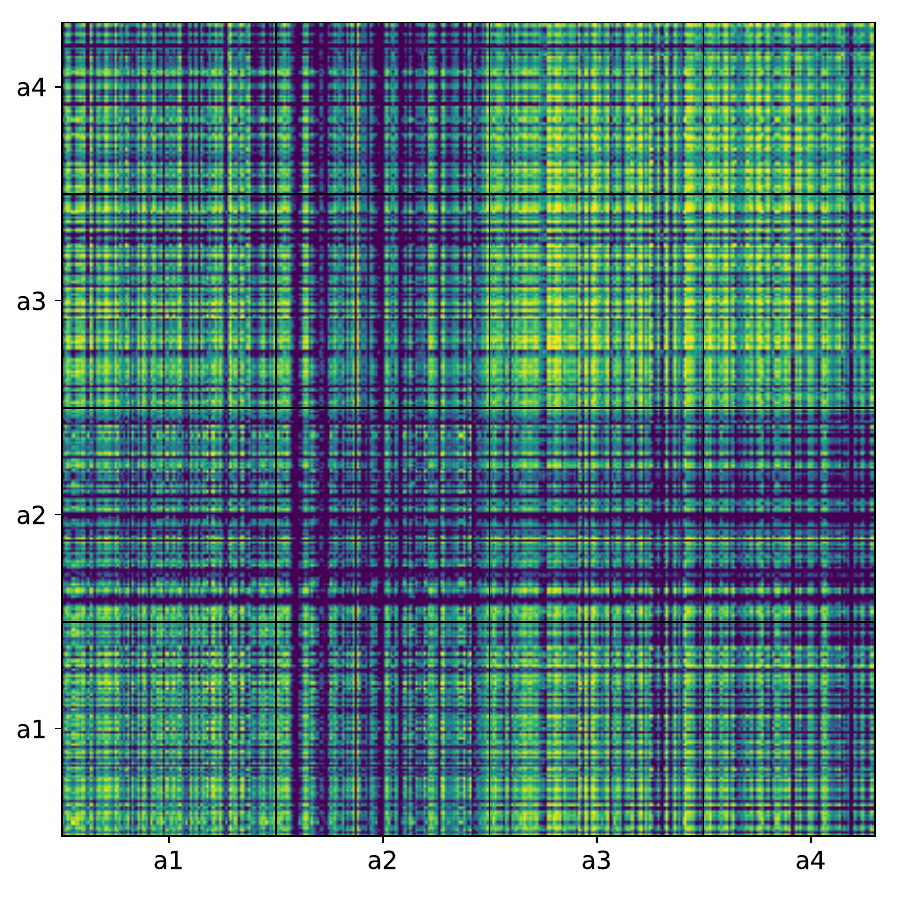}
  \caption{2to2 (Rand.)}
\end{subfigure}\hfil 
\begin{subfigure}{0.15\textwidth}
  \includegraphics[width=\linewidth]{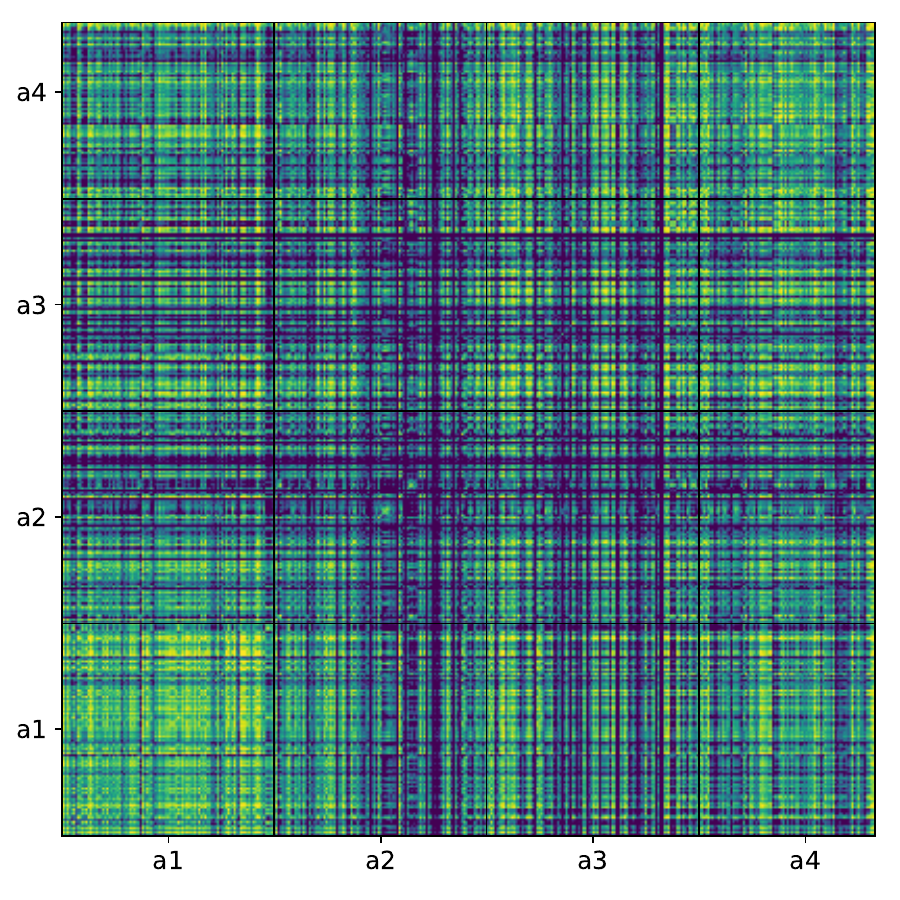}
  \caption{1to3 (Rand.)}
\end{subfigure}\hfil 
\begin{subfigure}{0.15\textwidth}
  \includegraphics[width=\linewidth]{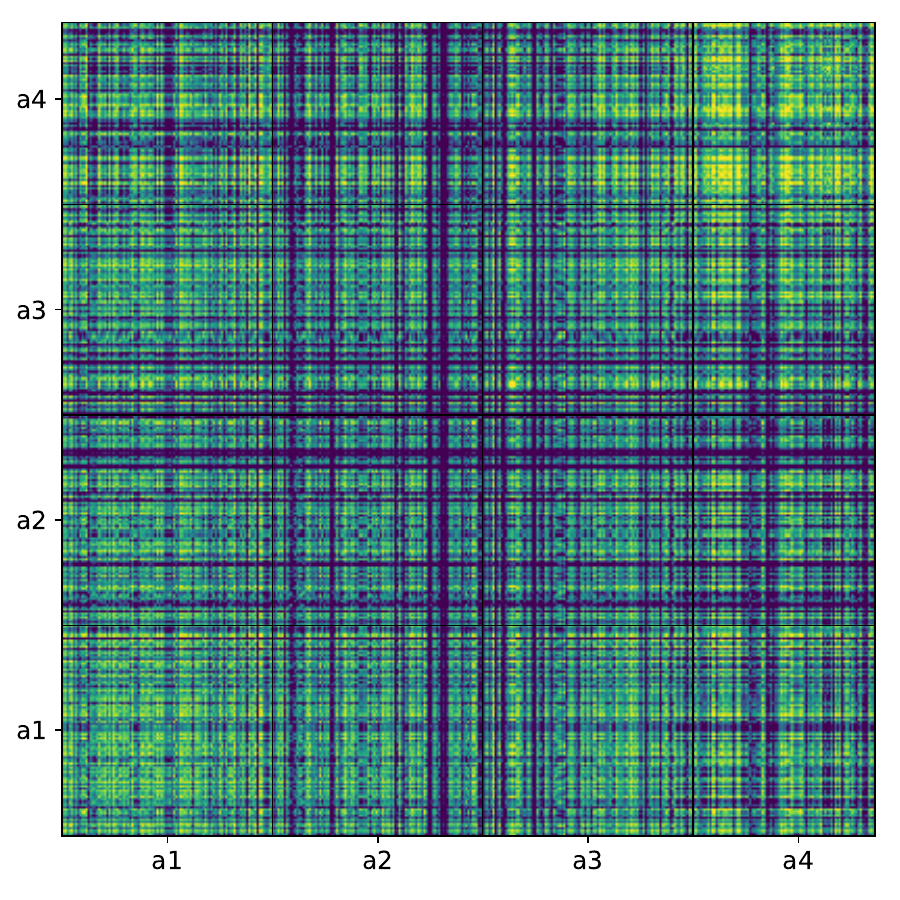}
  \caption{3to1 (Rand.)}
\end{subfigure}
\caption{Cosine similarity heatmaps of attribute representations projected into the CBR subspace (above) and a random subspace (below). 
}
\label{fig:semantic_pattern}
\end{figure}
To examine whether the CBR subspace captures semantic information in addition to indices, we construct an additional dataset in which relations exhibit controlled patterns of semantic similarity. As shown in Table~\ref{tab:datat_semantic}, the four relations are grouped so that: (i) the first and last two relations share similar meaning (denoted as ``2to2''), (ii) the last three relations share similar meaning (denoted as ``1to3''), and (iii) the first three relations share similar meaning (denoted as ``3to1''). For instance, in the ``1to3'' pattern, the last three attributes of ``Bob'': ``editor'', ``guard'' and ``student'' are bound to semantically similar relations. This design allows us to test whether semantic similarity among relations is reflected in the geometry of the CBR subspace.

Using the projection matrix $W_{\text{CBR}}$ learned from the main dataset~\secref{sec:dataset}, we project attribute (e.g., ``$a1$'' in Table~\ref{tab:datat_semantic}) activations into the CBR subspace and compute pairwise cosine similarities between projected representations. We compare these results with a control condition where activations are projected using a random matrix $W_{\text{rand}}$ of the same dimensionality. The resulting cosine similarity heatmaps for Llama3-8B-Instruct are shown in Figure~\ref{fig:semantic_pattern}. Additional results for other datasets are reported in Appendix~\secref{sec:irs_subspace_semantic_various}, and corresponding results for Qwen3-8B are presented in Appendix~\secref{sec:irs_subspace_semantic_qwen}.

The CBR-projected representations clearly reproduce the intended semantic similarity structure: relations designed to be semantically close form coherent similarity blocks, while semantically distant relations remain well separated. In contrast, the random projection produces a comparatively less clear and meaningful structure. These results suggest that the CBR subspace embeds semantic information such that attributes bound to semantically similar relations yield similar representations in the CBR subspace, and thus similar indices.

\begin{table}[t]
\centering
\small
\setlength{\tabcolsep}{4pt}
\begin{tabularx}{\columnwidth}{lX}
\toprule
Pattern & \textbf{Discourse} \\
\midrule
2to2 &
Paul currently works as a writer and serves as a doctor \dots\ Meanwhile, Bob currently works as a \underline{coach}$_{a1}$ and serves as an \underline{editor}$_{a2}$, thriving in his creative endeavors. Yet, he hates being a \underline{guard}$_{a3}$ and dislikes being a \underline{student}$_{a4}$, feeling constrained in those roles. \dots \\
\midrule
1to3 &
Paul currently works as a writer, \dots\ Meanwhile, Bob is a \underline{coach}$_{a1}$, but he too feels the weight of frustration, particularly when he thinks about being an \underline{editor}$_{a2}$. He hates the idea of being a \underline{guard}$_{a3}$ and finds no joy in the thought of being a \underline{student}$_{a4}$. \dots \\
\midrule
3to1 &
Paul currently works as a writer, \dots\ Meanwhile, Bob works as a \underline{coach}$_{a1}$ and serves as an \underline{editor}$_{a2}$. He takes on the role of a \underline{guard}$_{a3}$, but he dislikes being a \underline{student}$_{a4}$. \dots \\
\bottomrule
\end{tabularx}
\caption{Samples of the dataset for semantic information analysis in the CBR subspace, where ``$a1$'' denotes the first attribute for a given entity (e.g., ``Bob''), and so on.}
\label{tab:datat_semantic}
\end{table}

\paragraph{Does the CBR subspace generalize across domains?}
As described in \S\ref{sec:irs_subspace}, we learn an CBR projection matrix $W_{\textbf{CBR}}$ independently for each context. Here, we analyze the generality of these $W_{\textbf{CBR}}$: specifically, whether a $W_{\textbf{CBR}}$ learned from one context can effectively recover index information in another.
Therefore, we further hypothesize that the CBR subspace may exhibit shared structure across contexts. In particular, we consider two possibilities.

\textbf{Hypothesis 1 (Global CBR subspace).}
The CBR subspace is global across contexts. Under this hypothesis, a single context-independent projection matrix $W$ extracts CBR indices from all contexts, such that $Wh_{c_1} \approx Wh_{c_2}$,
where $h_c \in \mathbb{R}^d$ denotes the activation vector under context $c$, and $W \in \mathbb{R}^{k \times d}$ projects activations into the CBR subspace.

\textbf{Hypothesis 2 (Context-dependent subspace with consistent structure).}
The exact alignment of the CBR subspace may vary across contexts, but the mappings between these context-specific subspaces follow a consistent second-order structure. In particular, activations across contexts may be related by a translation in activation space:
\begin{equation}
W_{c_1} h_{c_1}^{(ei,ri)} \approx
W_{c_1}\left(h_{c_2}^{(ei,ri)} + \Delta_{c_2 \rightarrow c_1}^{(ei, ri)}\right), \label{eq:irs_trans_hypothesis}
\end{equation}
where $h_c^{(ei,ri)} \in \mathbb{R}^d$ denotes the activation vector under context $c$ with CBR indices $(ei,ri)$, $W_{c_1} \in \mathbb{R}^{k \times d}$ is the projection matrix learned in context $c_1$, and $\Delta_{c_2 \rightarrow c_1} \in \mathbb{R}^d$ is a context-dependent translation vector aligning activations from context $c_2$ (e.g., \textit{country}) with those from context $c_1$ (e.g., \textit{relation}).

\begin{figure}[t]
\centering
\begin{subfigure}{0.485\linewidth}
\centering
\includegraphics[width=\linewidth]{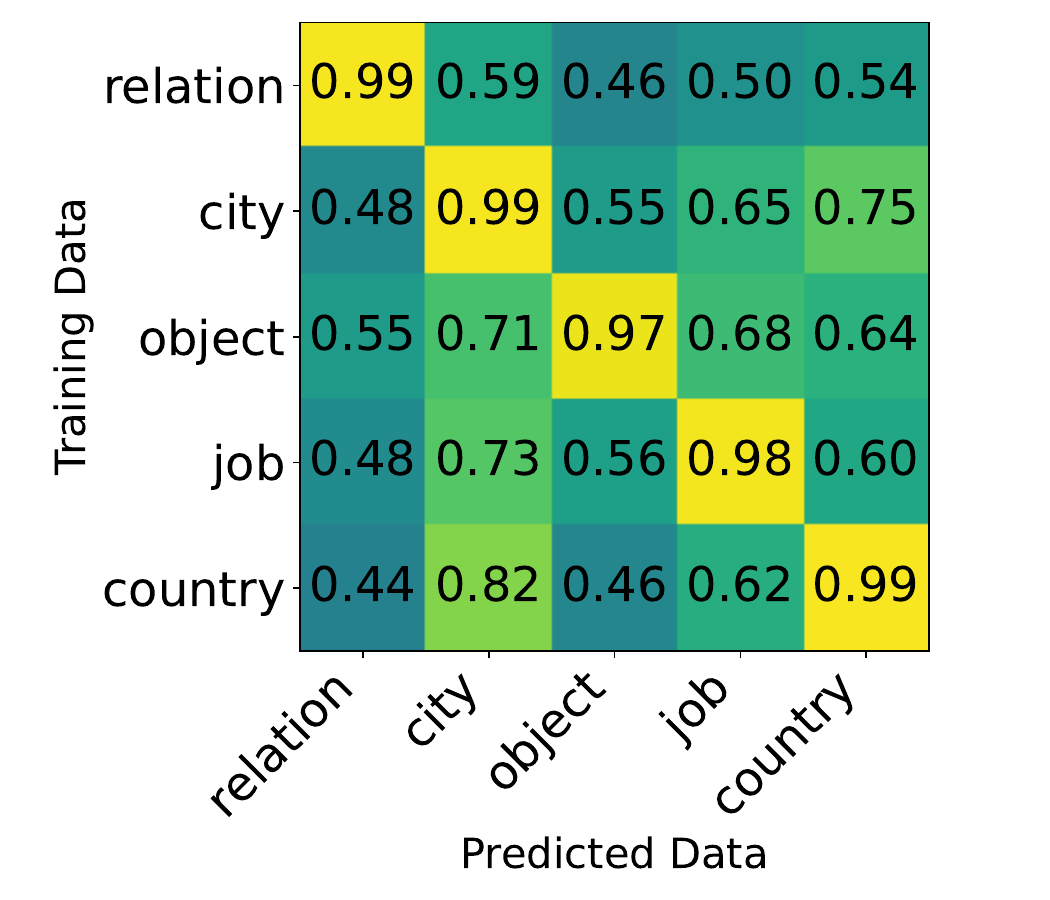}
\caption{w/o $\Delta$}
\label{fig:generality_llama}
\end{subfigure}
\begin{subfigure}{0.485\linewidth}
\centering
\includegraphics[width=\linewidth]{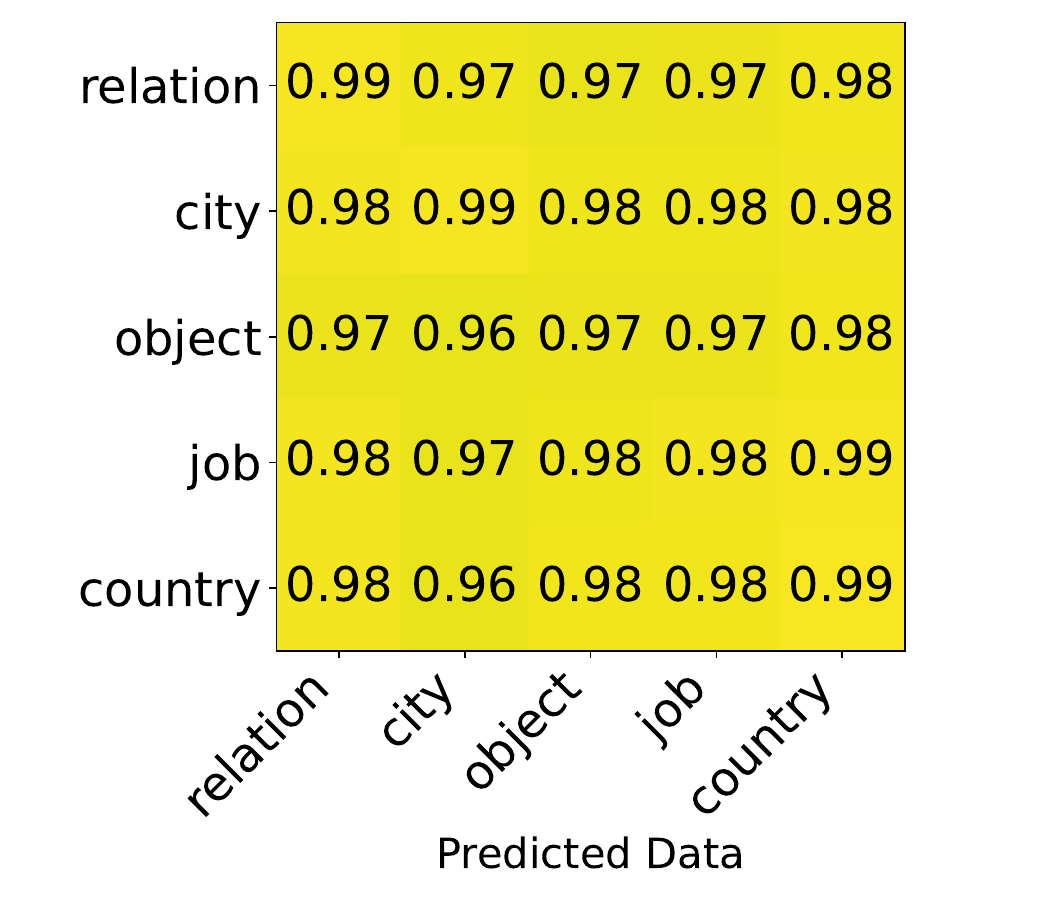}
\caption{w/ $\Delta$}
\label{fig:generality_trans_llama}
\end{subfigure}

\caption{Each cell shows the $R^2$ fitness score obtained from Llama3-8B-Instruct. The projection matrix learned from one context (column) is used to predict the index information of another context (row). 
}
\end{figure}

To evaluate this, we compute cross-context fitness scores. For each source (or trained) context, we apply its $W_{\textbf{CBR}}$ to the activations from a different target context and measure how well the projected representations predict the index information of target context. The resulting scores are shown in Figure~\ref{fig:generality_llama} for Llama3-8B-Instruct and Figure~\ref{fig:generality_qwen} \secref{sec:generality_qwen} for Qwen3-8B.


We observe that the cross-context $R^2$ scores vary across contexts, suggesting that the first hypothesis is less likely, while the second hypothesis is more plausible. To further examine the second hypothesis, we learn a translation vector using Equation~\ref{eq:irs_trans_v}, defined as the mean difference between activation vectors under two contexts for the same $(ei, ri)$ pair. The results after applying this translation via Equation~\ref{eq:irs_trans_hypothesis} are shown in Figure~\ref{fig:generality_trans_llama}. The improved $R^2$ scores support the \textbf{Hypothesis 2}, suggesting that while position of the CBR subspace varies across contexts, the mappings between these context-specific subspaces exhibit a consistent second-order structure. Further ablation analyses on $\Delta_{c_2 \rightarrow c_1}$ are presented in Appendix~\secref{sec:transv_ab}, and a discussion of the underlying reason is provided in Appendix~\secref{sec:transv_theory}.

\begin{equation}
\Delta_{c_2 \rightarrow c_1}^{(ei,ri)} =
\mathbb{E}\left[h_{c_1}^{(ei,ri)} - h_{c_2}^{(ei,ri)}\right]
\label{eq:irs_trans_v}
\end{equation}

%
%
%
%

%
\begin{figure*}[t]
\centering
\includegraphics[width=15.0cm]{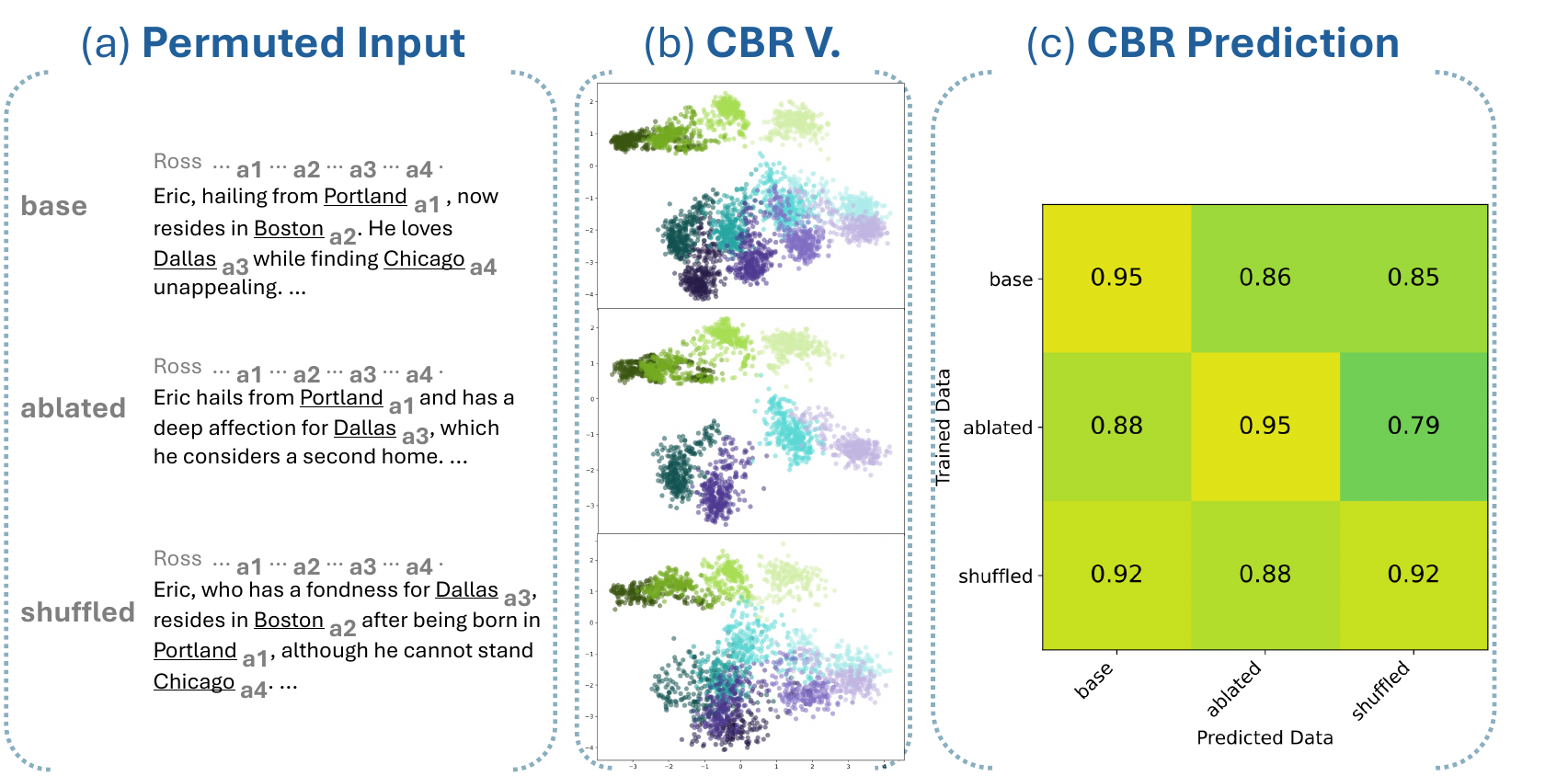}
\caption{(a) Samples of Ablated and Shuffled Dataset from $C_{city}$. The ablated sample removes the attribute ``a2'' (i.e., ``Boston'') and ``a4'' (i.e., ``Chicago''), while the shuffled sample alters the order of ``a1'' (i.e., ``Portland'') and ``a3'' (i.e., ``Dallas'').
(b) Visualization of the corresponding CBR subspace.
(c) Cross-dataset $R^2$ scores for CBR index prediction on Llama3-8B-Instruct. 
}
\label{fig:consistency_ablate_shuffle}
\end{figure*}

\paragraph{How stable is the CBR subspace?}
\label{sec:consistency_of_irs_subspace}
To examine the stability of the CBR subspace, we evaluate its behavior under controlled discourse perturbations. By permuting the order of subsequent mentions while keeping the original introduction fixed, we test two hypotheses: (1) $ri$ encodes introduction order as structural discourse information, or (2) $ri$ merely reflects surface-level token order.

To this end, we introduce two types of perturbations. First, we ablate certain attributes and relations associated with the second and third entities while leaving the relations of the first entity intact, as illustrated in Figure~\ref{fig:consistency_ablate_shuffle} (a). 
This manipulation preserves the original entity and relation indices but alters surface-level content. 
Second, we shuffle the order of attributes for the second and third entities while keeping the original indices unchanged.

\textbf{Representation Stability.} In both cases, the resulting CBR subspace visualizations preserve a similar geometric organization: although individual points exhibit some positional changes relative to the base input, the distribution continues to align with the directions corresponding to $ei$ and $ri$ (Figure~\ref{fig:consistency_ablate_shuffle} (b)). These results indicates that the overall structure of the CBR subspace is robust such permutations. 

\textbf{Stability of CBR Index Prediction.} We assess whether projection matrices learned under these permuted conditions generalize across datasets. Using the CBR projection matrix learned from the ablated or shuffled dataset to predict index information in the original dataset (and vice versa), we observe high predictive performance ($R^2$ is around $0.8$) in Figure~\ref{fig:consistency_ablate_shuffle} (c). This further confirms that the CBR subspace is highly consistent and thus its overall structure is preserved through ablation and shuffling. Additional results for other datasets are reported in Appendix~\secref{sec:consistency_of_irs_subspace_across_contexts}, and corresponding results for Qwen3-8B are presented in Appendix~\ref{sec:irs_consistency_qwen}. Overall, these results reconfirm our conclusions about the stability of CBR subspace. 

\textbf{Stability of the CBR-Based Mechanism.} Our activation steering analysis, introduced in \secref{sec:irs_steering}, continues to causally manipulate model outputs under the perturbed conditions in the same manner as in the base setting (Appendix~\ref{sec:irs_steering_shuffle_ablate}). This provides complementary evidence that the index-based mechanism remains functionally intact despite perturbation of the surface geometry.

\textbf{Prevalence of CBR Across Diverse Patterns.}
To rule out the possibility that the observed CBR structure is a pattern-specific artifact, we construct 13 distinct discourse patterns with progressively increasing structural complexity and variation, such as pattern 5: ($e_1r_1$, $e_1r_2$, $e_2r_1$, $e_2r_3$, $e_3r_2$, $e_3r_4$) and pattern 8: ($e_1r_1$, $e_1r_2$, $e_2r_1$, $e_2r_3$, $e_2r_4$, $e_3r_1$, $e_3r_2$, $e_3r_4$), where $e_ir_j$ denotes the attribute binds the $i$-th entity through the $j$-th relation , as detailed in the Appendix~\secref{sec:pattern}. Across all patterns, the CBR index prediction scores (measured by $R^2$) remain stable and consistent at around $0.95$ across contexts. The observed CBR index–based structure is unlikely to be an artifact of a specific repetitive pattern.
In addition, Appendix~\ref{sec:irs_jump} shows that the CBR structure persists under reference-distance manipulations, indicating that the indexing scheme captures more than superficial cues.

\paragraph{How to apply the CBR Subspace?}
Our analysis suggests that the CBR subspace provides a useful representation for monitoring relational bindings in LLMs. We compare our method with the Hessian-based approach of \citet{feng2024monitoring}, which proposes propositional probes for monitoring latent world states in LLMs and identifying binding entities in discourses such as: ``\textit{The nurse lives in Singapore. The \underline{CEO} lives in Canada. The person living in Singapore is male. The person living in Canada is \underline{female}.}'' In this setting, the correct binding is \textit{CEO–female} rather than the stereotypical association \textit{CEO–male}. Using the same setup, our CBR-based method predicts the correct bound entity index $ei$ from the activation of the attribute token (e.g., ``\textit{female}'') and achieves accuracies of $0.95$ (pro) and $0.94$ (anti), which are higher than the reported mean accuracies of the Hessian-based baseline. 
These results indicate that CBR could effectively capture binding information and monitor internal world states in LLMs. Details of the evaluation are provided in the Appendix~\secref{sec:genderbias}. 

The CBR structure may also benefit downstream NLP tasks such as Relation Extraction (RE). 
The CBR representation could support more stable RE and enable new approaches for interpreting and controlling LLM reasoning. An analysis of CBR indices on a commonly used real-world document-level RE dataset is provided in the Appendix~\secref{sec:docred}. The results indicate that the CBR signal could be identified in the real-world dataset when the target samples contextually resemble training examples.

\section{Conclusions}
In this work, we investigated how LLMs internally represent relational binding in discourse. To this end, we proposed the Cell-based Binding Representation (CBR) framework and applied PLS to identify a low-dimensional subspace that linearly encodes entity and relation indices. Visualizations reveal two interpretable directions corresponding to entity and relation indices.
Causal interventions show that manipulating the CBR subspace reliably alters LLMs predictions, supporting a CBR subspace based mechanism in which models retrieve attributes by combining the index information embedded in activations of LLMs. These findings suggest that discourse-level relational binding in LLMs is supported by a structural, compositional, and low-dimensional representation, which extends the Linear Representation Hypothesis.

\section*{Limitations}
Several limitations remain in this work. First, our analysis operates at the level of model activations, linear subspaces and CBR related head detection\footnote{We discuss this in Appendix~\secref{sec:irs_head}. }, and we do not perform fine-grained, head-level or neuron-level circuit analysis to localize the exact transformer components that implement the CBR-based mechanism. Second, while we empirically demonstrate the existence of a CBR subspace, we do not investigate how this subspace is generated during the model’s forward computation. In particular, this work does not address how LLMs select entities that carry index information, how the model resolves situations where multiple entities are present but only a subset participates in relational structures, or how to assign index information when the role of an entity shifts (e.g., from entity to attribute) as discourse unfolds.

In addition, although we use GPT-4o-mini to generate more naturalistic narrative discourse and apply various permutations to simulate the feature of real world discourse, our datasets are still synthetic and controlled in nature. They represent human-like but machine-generated text, and thus may not fully capture the complexity and variability of the real-world discourse. As a result, we do not investigate the extent to which the CBR schema generalizes to or breaks down in naturally occurring human text. 

Finally, although we propose the CBR subspace based mechanism for explaining relational binding, LLMs may also rely on other mechanisms, such as surface-level cues or statistical regularities, to encode binding information. This work does not analyze the potential interactions between these mechanisms or their relative contributions to binding behavior.

\section*{Ethical Statement}
LLMs (i.e., Llama3-8B-Instruct, Qwen3-8B, GPT-4o-mini) are applied according their intended research purposes and licenses. In addition, we construct synthetic datasets by following the framework of existing benchmarks, with entities and attributes sampled from diverse sets of single-token names and other concepts. Because our data are automatically generated and do not involve human annotation, this work does not introduce risks associated with annotation bias or the disclosure of personal or sensitive information. The datasets and code are publicly available to ensure the reproducibility of our experiments.

\section*{Acknowledgements}
This work was supported by JST CREST Grant Number JPMJCR20D2, AMED Grant Number JP25wm0625405 and Japan Science and Technology Agency under Grant No. JST BOOST JPMJBY24F9.

\bibliography{acl_latex}



\clearpage
\onecolumn
\appendix

\section{Appendix}

\subsection{Related Work}
\label{sec:related_work}

\paragraph{Linear Representation} Recent research on the Linear Representation Hypothesis (LRH) shows that language models encode a wide range of semantic and structural concepts as linear directions or low-dimensional subspaces. Prior work has found linear representations for Othello board states~\cite{li2022emergent,nanda2023emergent}, truth values of propositions~\cite{marks2023geometry}, sentiment~\cite{tigges2023linear}, semantic concepts~\cite{zhao2024beyond,saglam2025large}, spatial information~\cite{tehenan2025linear}, safe and jailbroken states~\cite{chia2025probing} and numeric attributes such as elevation, population, and dates~\cite{gurnee2023language,heinzerling2024monotonic,el2025geometry}. 

In addition, several studies have demonstrated that various forms of binding exhibit linear structure in LLM representations~\cite{feng2023language,dai2024representational,feng2024monitoring,gur2025mixing}. For example, given an input such as ``Alice lives in Laos.'', \citet{feng2024monitoring} typically analyzes the binding subspace between the entity ``Alice'' and the attribute ``Laos''. \citet{gur2025mixing} reveal several mixing mechanisms such as positional, lexical and reflexive mechanism. Such analyses are limited to pairwise bindings between an entity and an attribute, and therefore fail to capture the full binding structure among entities, relations, and attributes. Moreover, these approaches primarily focus on entity–attribute pair binding within individual sentences and do not examine discourse-level relational structures, which are fundamental to discourse understanding for LLMs.
To address these gaps, we show that LLMs encode discourse-level relational structure via a Cell-based Binding Representation (CBR). We analyze relational binding at the level of entire discourses, capturing structures that involve three or more interacting elements. In addition, we introduce IRS as a systematic annotation scheme for probing how LLMs bind entities, relations, and attributes into coherent discourse-level representations.

\paragraph{Learned Knowledge} Prior work has attempted to localize and edit factual relations, such as “capital of”, that language models acquire during pretraining and store in their parameters~\citep{geva2021transformer,dai2022knowledge,meng2022locating,geva2023dissecting,hernandez2023linearity}. These studies primarily examine how static knowledge is encoded in model weights. In contrast, our work focuses on in-context representations of relations and investigates how relational structure is encoded in model activations during discourse processing.

\paragraph{Mechanistic Analysis} Substantial progress has been made in uncovering internal circuits that implement specific computations in language models~\cite{elhage2021mathematical,wang2022interpretability,wu2024interpretability}. Recent work has begun to analyze mechanisms for entity tracking and binding, including the identification of circuits for entity tracking~\citep{prakash2024fine}, the proposal of a Binding ID~\citep{feng2023language} and binding subspace based high-level mechanism~\cite{feng2024monitoring} to explain binding. While these studies provide valuable insights into sentence-level or pairwise binding, they do not capture the structured index information over extended discourse, especially entity–relation bindings. Our work complements this line of research by introducing CBR subspace based mechanism and empirically analyzing how both entity and relation indices are represented and combined in model activations to support discourse-level relational binding.

\clearpage

\subsection{Datasets and Language Models for CBR Subspace Analysis}
\label{sec:dataset}
\paragraph{Datasets} To analyze the CBR subspace, we construct a controlled dataset that systematically varies entity and relation indices while keeping semantic content simple and interpretable. Following the data curation method~\cite{feng2024monitoring}, we begin by defining a closed world consisting of five contexts: $C_{relation}$, $C_{object}$, $C_{city}$, $C_{job}$ and $C_{country}$, and each context (e.g., $C_{country}$) includes four relation types (e.g., ``manufactured in'', ``designed in'', ``exported to'' and ``banned in''), reflecting distinct event roles, as shown in the first row of Table~\ref{tab:datat_temp_other_table} \secref{sec:data_sample}. Entity and attribute for a relation type are sampled from five sets of one-token words: $S_{name}$, $S_{object}$, $S_{city}$, $S_{job}$ and $S_{country}$, as shown in Table~\ref{tab:datat_one_token} \secref{sec:data_sample}. Here, $S_{name}$, for instance, denotes a set of $name$ tokens such as ``Eric''.

Taking $C_{country}$ as an example, we first sample three entities (e.g., ``boot'') from $S_{object}$ and twelve attributes (e.g., ``Mexico'') from $S_{country}$ as shown in Table~\ref{tab:datat_ent} and organize them into a structured table (i.e., Table Template Input in Table~\ref{tab:datat_temp}) that specifies which attribute is bound to which entity and relation. These tables are then instantiated into discourse templates to generate concise descriptions of the relational structure (i.e., the Discourse Template Input in Table~\ref{tab:datat_temp}). The CBR subspace visualization and causal intervention results for this template input are shown in Appendix~\ref{sec:irs_visualization_tabtemp} and Appendix~\ref{sec:irs_steering_tabtemp}, respectively. Finally, to produce naturalistic inputs suitable for analyzing LLM activations, we prompt GPT-4o-mini to rewrite each structured description into a short, coherent narrative (i.e., Story Input in Table~\ref{tab:datat_temp}) while preserving the underlying relational bindings. This pipeline ensures that surface variation does not obscure the latent entity–relation indices, enabling precise analysis of how these indices are encoded in model representations. Each context contains $1,000$ naturalistic narratives. Additional samples for Table Template Input, Discourse Template Input and Story Input are shown in Table~\ref{tab:datat_temp_other_table}, \ref{tab:datat_temp_other_temp} and \ref{tab:datat_temp_other_story} in Appendix~\secref{sec:data_sample} respectively.

\begin{table}[!htbp]
\centering
\scalebox{0.8}{
\begin{tabular}{lll}
\hline
  & \textbf{Entity} & \textbf{Attribute} \\\cmidrule(l){1-3}
 $C_{country}$ & \makecell[l]{boot, radio, chair (from $S_{object}$)} & \makecell[l]{Mexico, Jordan, Turkey, India, \\ Japan, Brazil, France, Canada, \\ Sweden, Argentina, Australia, Spain \\ (from $S_{country}$)} \\\cmidrule(l){1-3}
 $C_{relation}$ & \makecell[l]{Eric, Ian, Dan (from $S_{name}$)} & \makecell[l]{Tara, Brad, Jack, Nick, ... (from $S_{name}$)} \\\cmidrule(l){1-3}
 $C_{...}$ & \makecell[l]{...} & \makecell[l]{...} \\\cmidrule(l){1-3}
\hline
\end{tabular}
}
\caption{Samples of Entity and Attribute.}
\label{tab:datat_ent}
\end{table}
\begin{table*}[!htbp]
\centering
\scalebox{0.8}{
\begin{tabular}{ll}
\hline
  & \textbf{Table Template Input} \\\cmidrule(l){1-2}
 $C_{country}$ & \makecell[l]{| Product | Manufactured in | Designed in | Exported to | Banned in |\\| boot | Mexico | Jordan | Turkey | India |\\| radio | Japan | Brazil | France | Canada |\\| chair | Sweden | Argentina | Australia | Spain |} \\\cmidrule(l){1-2}
 $C_{relation}$ & \makecell[l]{| Name | Spouse | Child | Teacher | Boss |\\ | Eric | Tara | Brad | Jack | Nick |...} \\\cmidrule(l){1-2}
 $C_{...}$ & \makecell[l]{...} \\\cmidrule(l){1-2}
  & \textbf{Discourse Template Input} \\\cmidrule(l){1-2}
 $C_{country}$ & \makecell[l]{The boot is manufactured in Mexico and designed in Jordan, and it is exported to Turkey, \\ but it is banned in India . The radio is manufactured in Japan and designed in Brazil, \\ and it is exported to France, but it is banned in Canada . The chair is manufactured in Sweden \\ and designed in Argentina, and it is exported to Australia, but it is banned in Spain .} \\\cmidrule(l){1-2}
 $C_{relation}$ & \makecell[l]{Eric is married to Tara and has a child named Brad, he was taught by Jack and works under Nick . ...} \\\cmidrule(l){1-2}
 $C_{...}$ & \makecell[l]{...} \\\cmidrule(l){1-2}
  & \textbf{Story Input} \\\cmidrule(l){1-2}
 $C_{country}$ & \makecell[l]{In a bustling market, a unique boot caught everyone's attention,  as it was manufactured in \textbf{Mexico} \\ and designed in Jordan. It had made its way to Turkey for export, but unfortunately, it faced a ban \\ in India. Meanwhile, a sleek radio, manufactured in Japan and designed in Brazil, was shipped \\ to France, yet it couldn't be sold in Canada due to restrictions. Lastly, a stylish chair, crafted in \\ Sweden and designed in Argentina, found its way to Australia, but it was banned in Spain. Each \\ item told a tale of international trade and the complexities of global regulations.} \\\cmidrule(l){1-2}
 $C_{relation}$ & \makecell[l]{Eric lives happily with his spouse Tara and they have a child named Brad. He fondly remembers \\ being taught by Jack and appreciates working under Nick. Meanwhile, ...} \\\cmidrule(l){1-2}
 $C_{...}$ & \makecell[l]{...} \\\cmidrule(l){1-2}
 \hline
\end{tabular}
}
\caption{Samples of Dataset.}
\label{tab:datat_temp}
\end{table*}
\begin{table}[!htbp]
\centering
\scalebox{0.8}{
\begin{tabular}{lll}
\hline
  & $|S|$ & \textbf{Sample} \\\cmidrule(l){1-3}
 $S_{name}$ & 47 & \makecell[l]{``Ray'', ``Eric'', ``Leo'', ``Ross'', ``James'', ``Matt'', ``Brad'', ``Jeff'', ``Todd'', ...} \\\cmidrule(l){1-3}
 $S_{object}$ & 51 & \makecell[l]{``window'', ``glass'', ``door'', ``paper'', ``book'', ``toy'', ``mirror'', ``ball'', ``clock'', ...} \\\cmidrule(l){1-3}
 $S_{city}$ & 20 & \makecell[l]{``Atlanta'', ``Seattle'', ``Phoenix'', ``London'', ``Hamilton'', ``Boston'', ``Kansas'', ``Toronto'', ``Miami'', ...} \\\cmidrule(l){1-3}
 $S_{job}$ & 16 & \makecell[l]{``writer'', ``student'', ``driver'', ``artist'', ``editor'', ``actor'', ``athlete'', ``guard'', ``chef'', ...} \\\cmidrule(l){1-3}
 $S_{country}$ & 23 & \makecell[l]{``Georgia'', ``India'', ``Japan'', ``Spain'', ``Italy'', ``Australia'', ``China'', ``Russia'', ``Egypt'', ...} \\\cmidrule(l){1-3}
\hline
\end{tabular}
}
\caption{Samples of One-token Words.}
\label{tab:datat_one_token}
\end{table}

\paragraph{Language Models} We adopt Llama3-8B-Instruct~\cite{meta2024llama3} and Qwen3-8B~\cite{yang2025qwen3} for CBR subspace analysis. Llama3-8B-Instruct is a 32-layer Transformer with a hidden dimension of 4096, while Qwen3-8B consists of 36 Transformer layers with a hidden size of 4096. 

\clearpage

\subsection{Additional Data Samples}
\label{sec:data_sample}

\begin{table*}[!htbp]
\centering
\scalebox{0.8}{
\begin{tabular}{ll}
\hline
  & \textbf{Table Template Input} \\\cmidrule(l){1-2}
 $C_{country}$ & \makecell[l]{| Product | Manufactured in | Designed in | Exported to | Banned in |\\| ring | Russia | France | Iraq | Singapore |\\| plant | China | India | Australia | Jordan |\\| jar | Sweden | Iran | Pakistan | Georgia |} \\\cmidrule(l){1-2}
 $C_{city}$ & \makecell[l]{| Name | Birthplace | Lived City | Loved City| Disliked City |\\| Brad | Paris | Houston | London | Detroit |\\| Gary | Austin | Berlin | Chicago | Portland |\\| James | Hamilton | Split | Atlanta | Dallas |} \\\cmidrule(l){1-2}
 $C_{relation}$ & \makecell[l]{| Name | Spouse | Child | Teacher | Boss |\\| Brad | Ava | Jack | Paul | Rob |\\| Gary | Kim | Joe | Fred | Leo |\\| Mike | Tara | Tom | Lee | Jake |} \\\cmidrule(l){1-2}
  $C_{job}$ & \makecell[l]{| Name | Current Job | Dream Job | Previous Job | Disliked Job |\\| Sean | guard | builder | student | chef |\\| Luke | coach | actor | judge | artist |\\| Sam | teacher | writer | driver | manager |} \\\cmidrule(l){1-2}
  $C_{object}$ & \makecell[l]{| Name | Created Object | Bought Object | Sold Object| Favorite Object |\\| Nick | lamp | brush | mat | table |\\| Jay | stamp | jar | shirt | belt |\\| Gary | ball | book | phone | fork |} \\\cmidrule(l){1-2}
 \hline
\end{tabular}
}
\caption{Other Samples for Table Template Input.}
\label{tab:datat_temp_other_table}
\end{table*}
\begin{table*}[htbp]
\centering
\scalebox{0.8}{
\begin{tabular}{ll}
\hline
  & \textbf{Discourse Template Input} \\\cmidrule(l){1-2}
 $C_{country}$ & \makecell[l]{The ring is manufactured in Russia and designed in France, and it is exported to Iraq, \\ but it is banned in Singapore .  The plant is manufactured in China and designed in India, and \\ it is exported to Australia, but it is banned in Jordan . The jar is manufactured in Sweden and \\ designed in Iran, and it is exported to Pakistan, but it is banned in Georgia .} \\\cmidrule(l){1-2}
 $C_{city}$ & \makecell[l]{Brad was born in Paris and currently lives in Houston, he loves London and dislike Detroit .\\  Gary was born in Austin and currently lives in Berlin, he loves Chicago and dislike Portland .\\ James was born in Hamilton and currently lives in Split, he loves Atlanta and dislike Dallas .} \\\cmidrule(l){1-2}
 $C_{relation}$ & \makecell[l]{Brad is married to Ava and has a child named Jack, he was taught by Paul and works under Rob . \\  Gary is married to Kim and has a child named Joe, he was taught by Fred and works under Leo . \\ Mike is married to Tara and has a child named Tom, he was taught by Lee and works under Jake .} \\\cmidrule(l){1-2}
 $C_{job}$ & \makecell[l]{Sean currently works as a guard and dreams of becoming a builder, he previously worked as a student, \\ and he dislikes being a chef . Luke currently works as a coach and dreams of becoming a actor, \\ he previously worked as a judge, and he dislikes being a artist . Sam currently works as a teacher \\ and dreams of becoming a writer, he previously worked as a driver, and he dislike
s being a manager .} \\\cmidrule(l){1-2}
 $C_{object}$ & \makecell[l]{Nick created the lamp and also bought the brush, he sold the mat, and his favorite object is the table . \\ Jay created the stamp and also bought the jar, he sold the shirt, and his favorite object is the belt . \\ Gary created the ball and also bought the book, he sold the phone, and his favorite object is the fork .} \\\cmidrule(l){1-2}
 \hline
\end{tabular}
}
\caption{Other Samples for Discourse Template Input.}
\label{tab:datat_temp_other_temp}
\end{table*}
\begin{table*}[!htbp]
\centering
\scalebox{0.8}{
\begin{tabular}{ll}
\hline
  & \textbf{Story Input} \\\cmidrule(l){1-2}
 $C_{country}$ & \makecell[l]{In a bustling marketplace, a unique ring caught the eye of many. This exquisite piece was manufactured \\ in Russia and designed in France. It found its way to Iraq, although it faced a ban in Singapore. Nearby, \\ a vibrant plant stood tall, manufactured in China and designed in India. It was exported to Australia, \\yet it was banned in Jordan. Lastly, a beautifully crafted jar sparkled under the sun, manufactured in \\ Sweden and designed in Iran. It was destined for Pakistan but faced a ban in Georgia. Each product\\ had its own journey, marked by both success and restriction.} \\\cmidrule(l){1-2}
 $C_{city}$ & \makecell[l]{Brad, born in Paris, now lives in Houston. He has a fondness for London but harbors a dislike for Detroit. \\ Meanwhile, Gary hails from Austin and currently resides in Berlin. He adores Chicago, although he has \\ an aversion to Portland. Lastly, James was born in Hamilton and makes his home in Split. He is passionate \\ about Atlanta, yet he dislikes Dallas. Each of these men carries a unique relationship with \\ the cities that have shaped their lives.} \\\cmidrule(l){1-2}
 $C_{relation}$ & \makecell[l]{Brad is happily married to Ava, and together they have a son named Jack. Throughout his education, \\ he learned from Paul and currently works under the supervision of Rob. Similarly, Gary shares a life \\ with Kim, and they are proud parents of a boy named Joe. His guiding mentor was Fred, and he reports \\ to Leo at work. Lastly, Mike enjoys a loving marriage with Tara, and their child is Tom. He was instructed \\ by Lee and is employed under Jake's leadership. Each family thrives, supported by their mentors and bosses.} \\\cmidrule(l){1-2}
 $C_{job}$ & \makecell[l]{Sean is currently a guard, aspiring to be a builder someday. He once held the position of a student but found \\ no joy in being a chef. Luke, on the other hand, works as a coach and dreams of becoming an actor. His \\ previous role was as a judge, yet he has no fondness for being an artist. Lastly, Sam is a teacher who \\ hopes to transition into a writer. Before this, he worked as a driver, and he particularly disliked being \\ a manager. Each of them navigates their careers, longing for something more fulfilling.} \\\cmidrule(l){1-2}
 $C_{object}$ & \makecell[l]{Nick was an inventor who created a lamp. He found a great deal and bought a brush. In his entrepreneurial spirit, \\ he sold a mat. His favorite object, however, was the table. Jay was an artist who created a stamp and decided to \\ buy a jar for his projects. He sold a shirt that he no longer needed, but his favorite object remained the belt. Lastly, \\ Gary was a playful spirit who created a ball. He bought a book to inspire his creativity, sold a phone he no longer \\ used, and cherished the fork as his favorite object.} \\\cmidrule(l){1-2}
 \hline
\end{tabular}
}
\caption{Other Samples for Story Input.}
\label{tab:datat_temp_other_story}
\end{table*}

\clearpage

\subsection{Identification of CBR Subspace on Llama3-8B-Instruct}
\label{sec:sec:irs_subspace_llama}

\begin{figure*}[htb]
    \centering 
\begin{subfigure}{0.25\textwidth}
  \includegraphics[width=\linewidth]{graph/pls_llama_country.pdf}
  \caption{$C_{country}$}
  \label{fig:l1}
\end{subfigure}\hfil 
\begin{subfigure}{0.25\textwidth}
  \includegraphics[width=\linewidth]{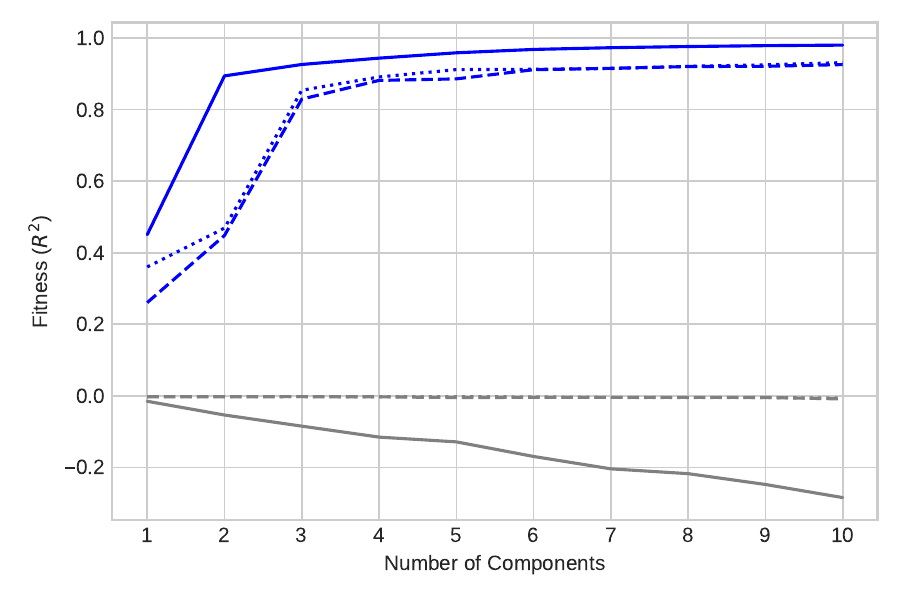}
  \caption{$C_{relation}$}
  \label{fig:l2}
\end{subfigure}\hfil 
\begin{subfigure}{0.25\textwidth}
  \includegraphics[width=\linewidth]{graph/pls_llama_job.pdf}
  \caption{$C_{job}$}
  \label{fig:l3}
\end{subfigure}\hfil
\begin{subfigure}{0.25\textwidth}
  \includegraphics[width=\linewidth]{graph/pls_llama_city.pdf}
  \caption{$C_{city}$}
  \label{fig:l4}
\end{subfigure}\hfil 
\begin{subfigure}{0.25\textwidth}
  \includegraphics[width=\linewidth]{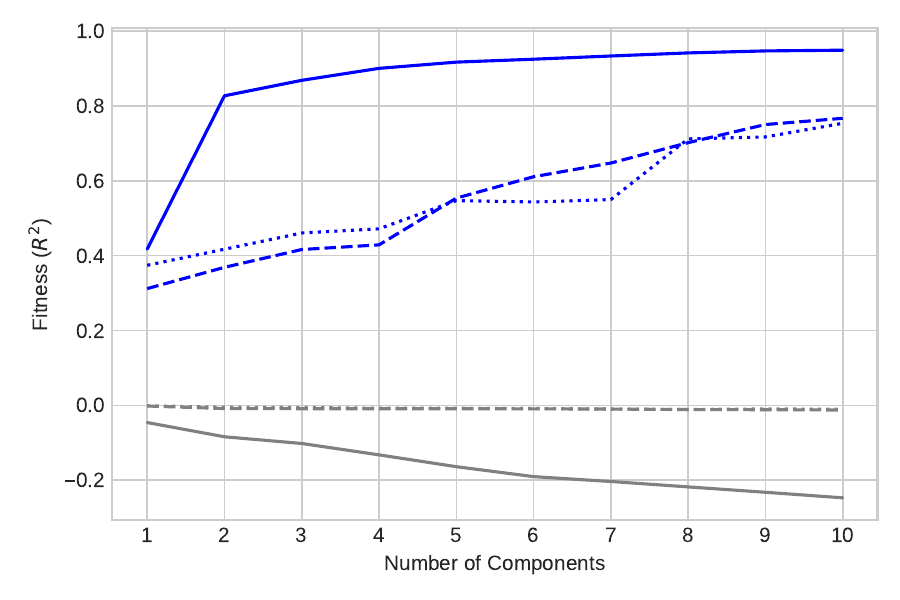}
  \caption{$C_{object}$}
  \label{fig:l5}
\end{subfigure}\hfil 
\caption{Decoding performance of $[ei,ri]$ from activations of Llama3-8B-Instruct using linear subspace methods. Each subplot shows how accurately entity and relation indices can be predicted as the dimensionality of the projected subspace increases for a given discourse context. The Y-axis indicates the fitness score ($R^2$), and the X-axis shows the number of components used for PLS, ICA and PCA projections. ``(Random)'' refers to model trained on random indices, serving as baseline controls.}
\label{fig:pls_fitness_llama}
\end{figure*}

\subsection{Identification of CBR Subspace on Qwen3-8B}
\label{sec:irs_subspace_qwen}
\begin{figure*}[htbp]
    \centering 
\begin{subfigure}{0.25\textwidth}
  \includegraphics[width=\linewidth]{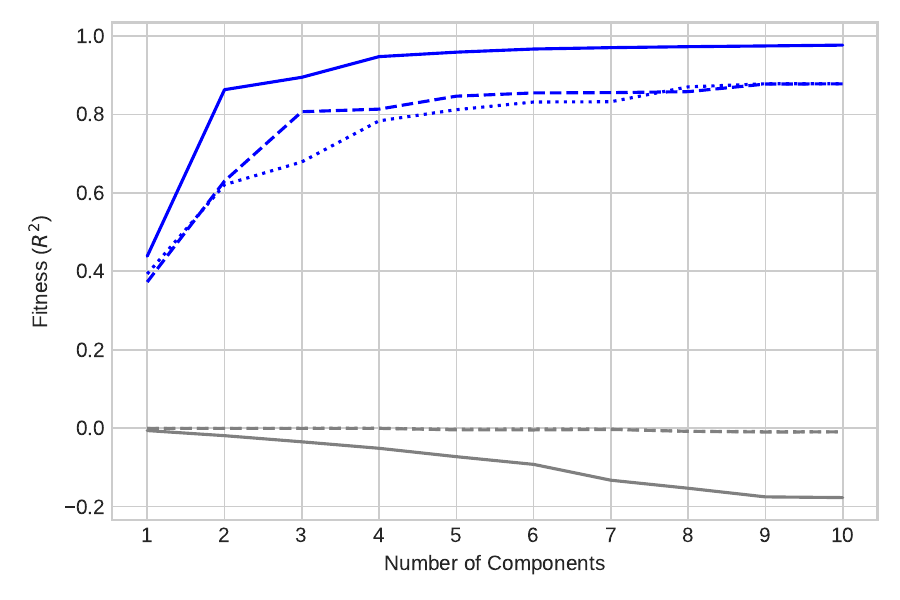}
  \caption{$C_{country}$}
  \label{fig:l1q}
\end{subfigure}\hfil 
\begin{subfigure}{0.25\textwidth}
  \includegraphics[width=\linewidth]{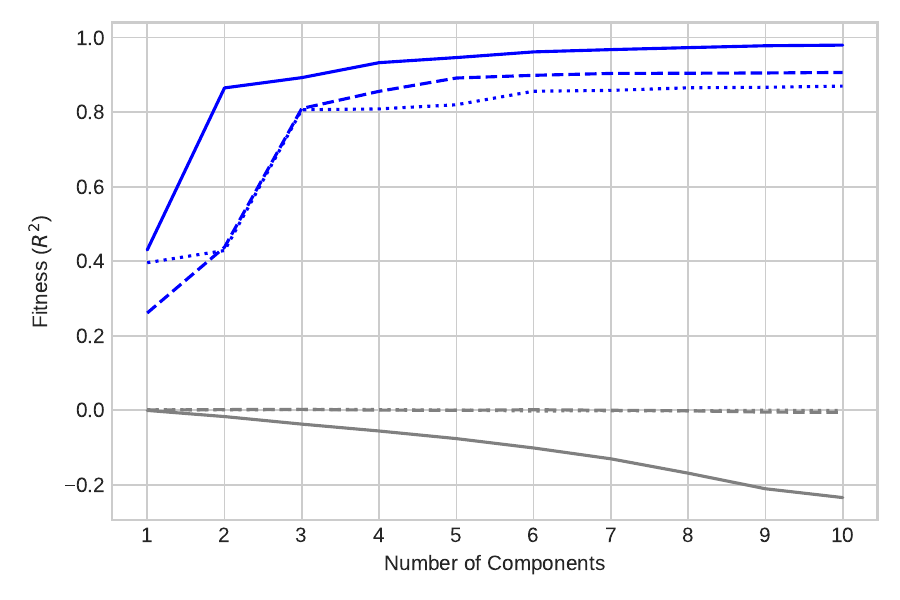}
  \caption{$C_{relation}$}
  \label{fig:l2q}
\end{subfigure}\hfil 
\begin{subfigure}{0.25\textwidth}
  \includegraphics[width=\linewidth]{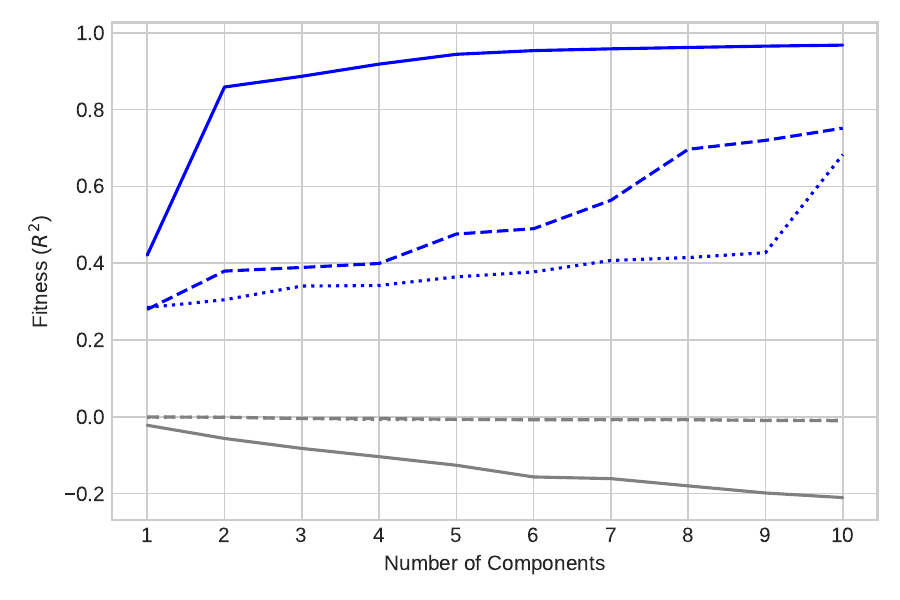}
  \caption{$C_{job}$}
  \label{fig:l3q}
\end{subfigure}\hfil
\begin{subfigure}{0.25\textwidth}\hfil
  \includegraphics[width=\linewidth]{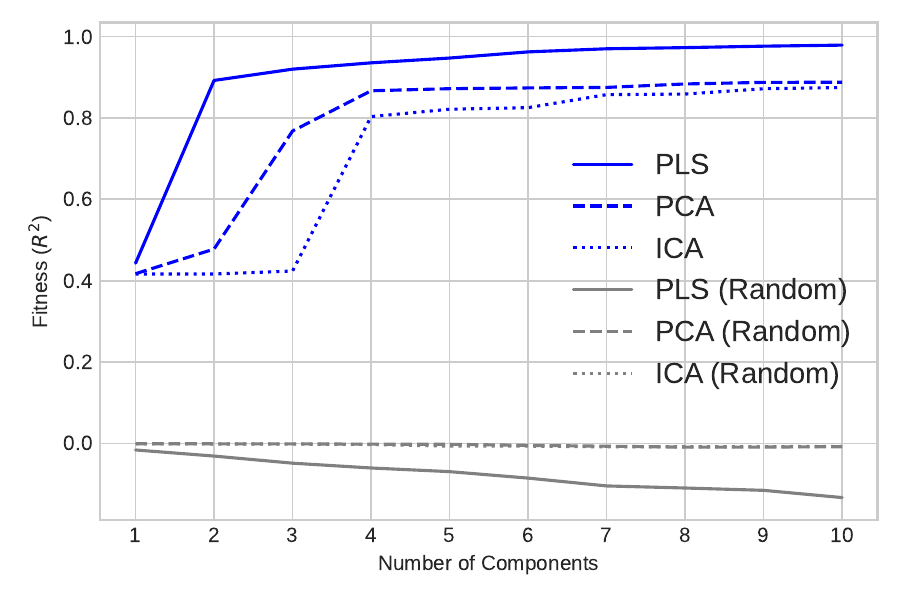}
  \caption{$C_{city}$}
  \label{fig:l4q}
\end{subfigure}\hfil 
\begin{subfigure}{0.25\textwidth}
  \includegraphics[width=\linewidth]{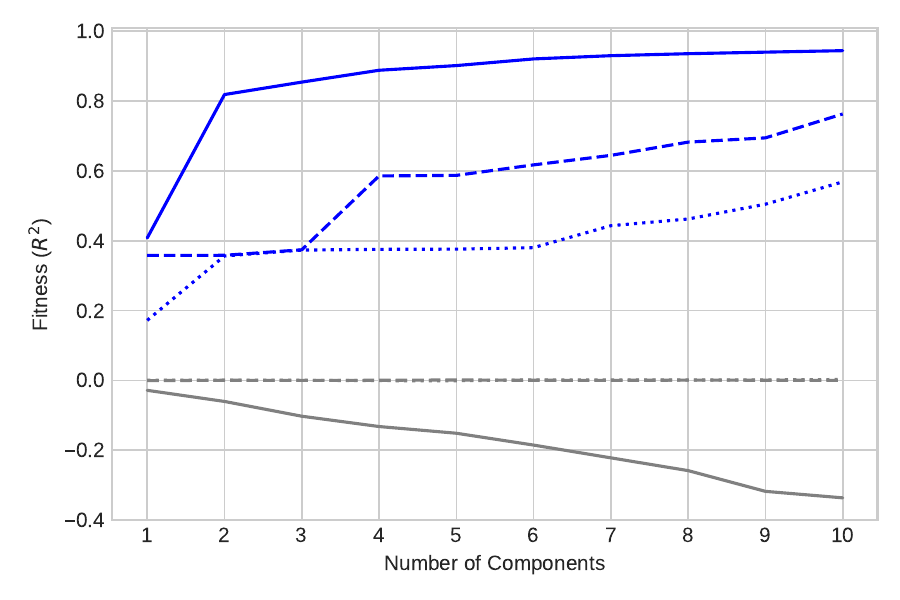}
  \caption{$C_{object}$}
  \label{fig:l5q}
\end{subfigure}\hfil 
\caption{Decoding performance of $[ei,ri]$ from activations of Qwen3-8B using linear subspace methods. Qwen3-8B shows consistency with Llama3-8B-Instruct, indicating that the CBR subspace emerges across model families and may reflect a general property of activation in LLMs.}
\label{fig:pls_fitness_qwen}
\end{figure*}

Figure~\ref{fig:pls_fitness_qwen} shows the decoding performance to predict $[ei,ri]$ from the activation of Qwen3-8B. The results are consistent with Llama3-8B-Instruct, indicating that the CBR subspace is a general phenomenon shared across LLMs from different families. In addition, detailed layer-wise prediction results are provided in the Appendix.~\ref{sec:irs_subspace_layerwise}.

\clearpage

\subsection{Visualization of CBR Subspace on Llama3-8B-Instruct}
\label{sec:irs_visualization_llama}

\begin{figure*}[htb]
    \centering 
\begin{subfigure}{0.3\textwidth}
  \includegraphics[width=\linewidth]{graph/vis_story_llama_country.pdf}
  \caption{$C_{country}$}
\end{subfigure}\hfil 
\begin{subfigure}{0.3\textwidth}
  \includegraphics[width=\linewidth]{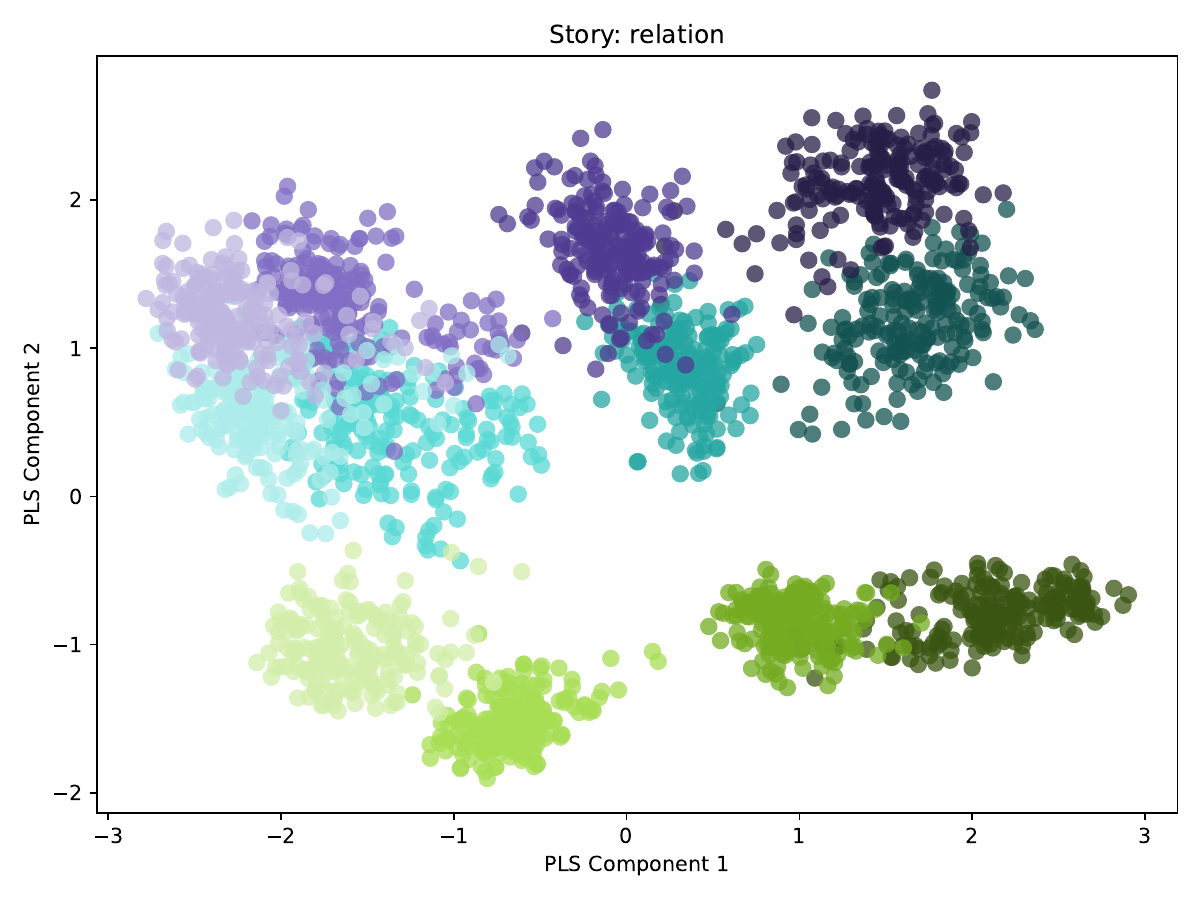}
  \caption{$C_{relation}$}
\end{subfigure}\hfil 
\begin{subfigure}{0.3\textwidth}
  \includegraphics[width=\linewidth]{graph/vis_story_llama_city.pdf}
  \caption{$C_{city}$}
\end{subfigure}\hfil
\begin{subfigure}{0.3\textwidth}
  \includegraphics[width=\linewidth]{graph/vis_story_llama_job.pdf}
  \caption{$C_{job}$}
\end{subfigure}\hfil 
\begin{subfigure}{0.3\textwidth}
  \includegraphics[width=\linewidth]{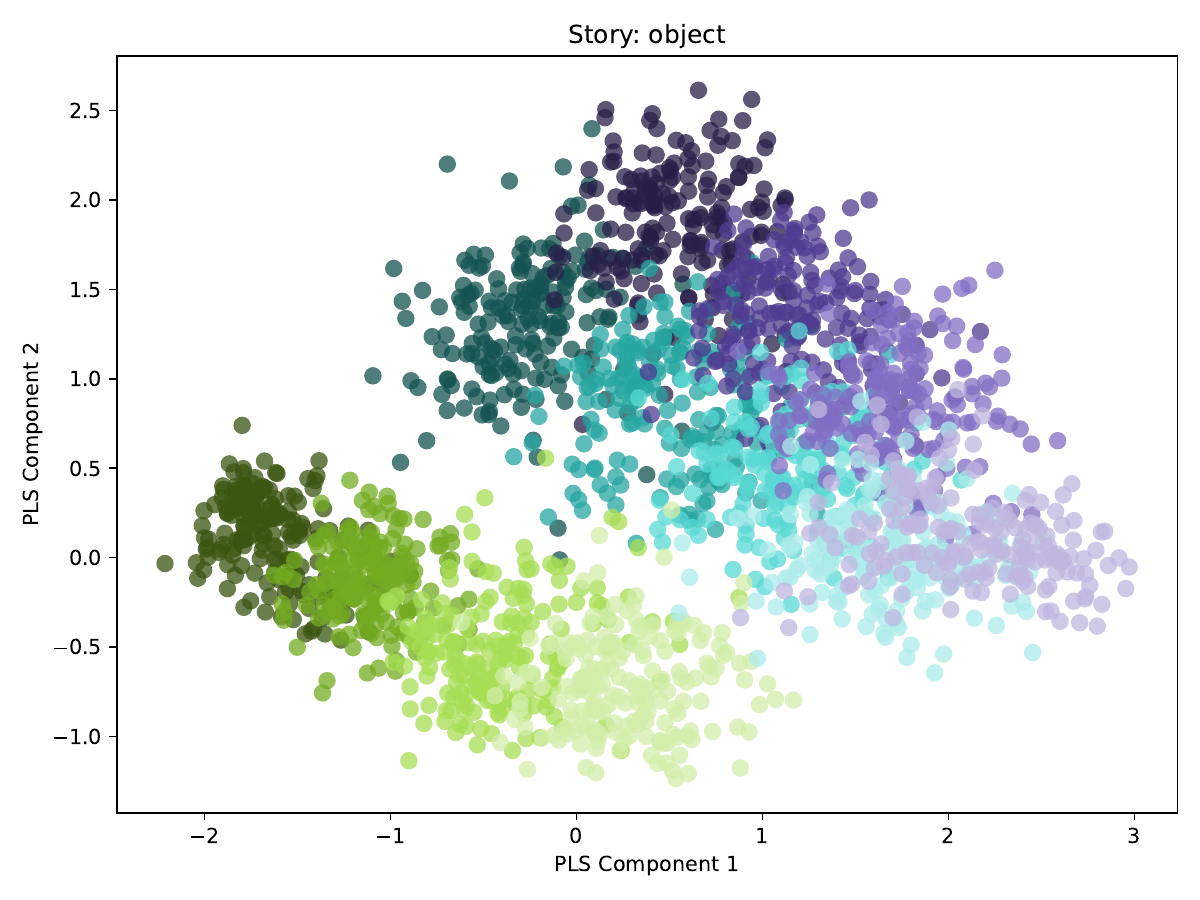}
  \caption{$C_{object}$}
\end{subfigure}\hfil 
\caption{Visualization of the CBR subspace from Llama3-8B-Instruct. Each point represents the projected activation of an attribute token. Colors indicate groups of attributes sharing the same $[ei,ri]$. The structure reveals clear distribution along both the $ei$ and $ri$ increasing (or decreasing) directions.}
\label{fig:irs_visualization_llama}
\end{figure*}

\subsection{Visualization of CBR Subspace on Qwen3-8B}
\label{sec:irs_visualization_qwen}
\begin{figure*}[!htbp]
    \centering 
\begin{subfigure}{0.3\textwidth}
  \includegraphics[width=\linewidth]{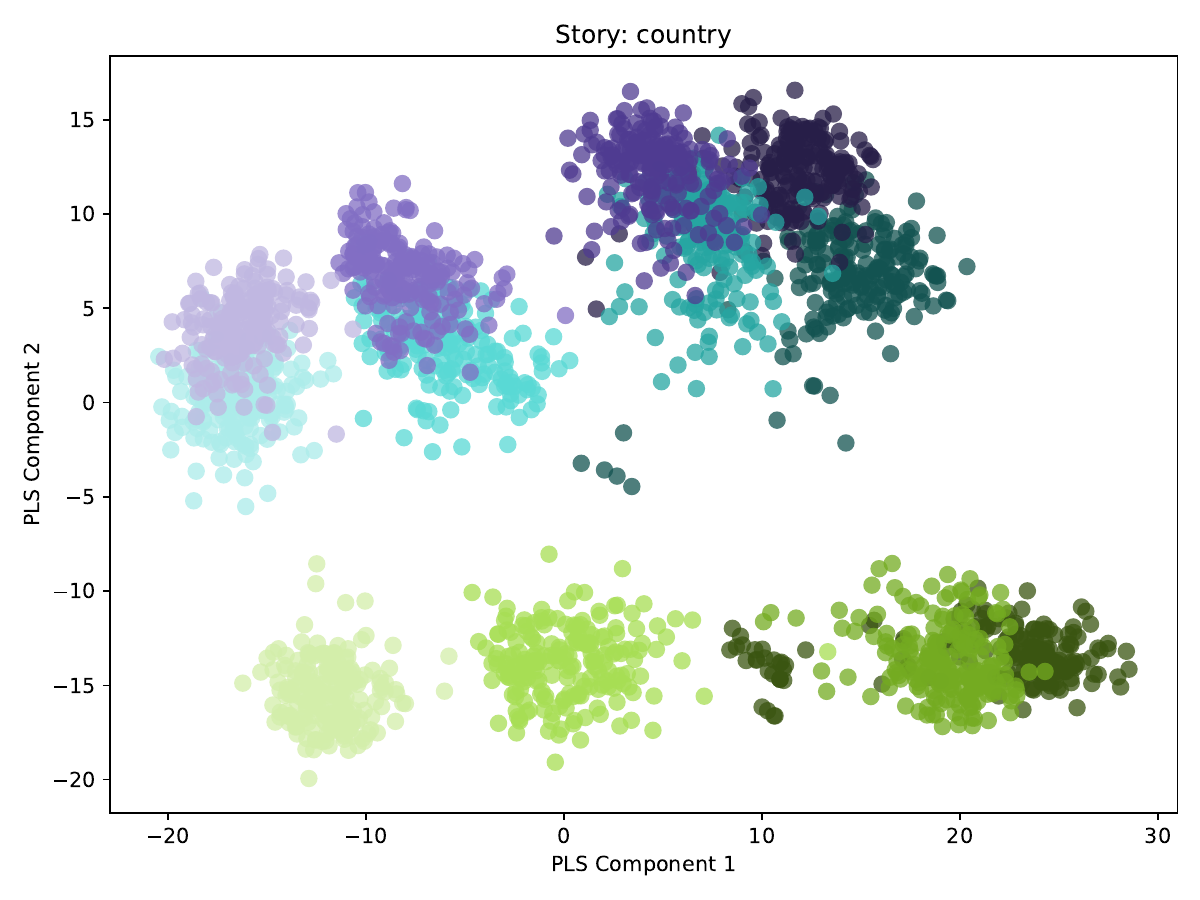}
  \caption{$C_{country}$}
\end{subfigure}\hfil 
\begin{subfigure}{0.3\textwidth}
  \includegraphics[width=\linewidth]{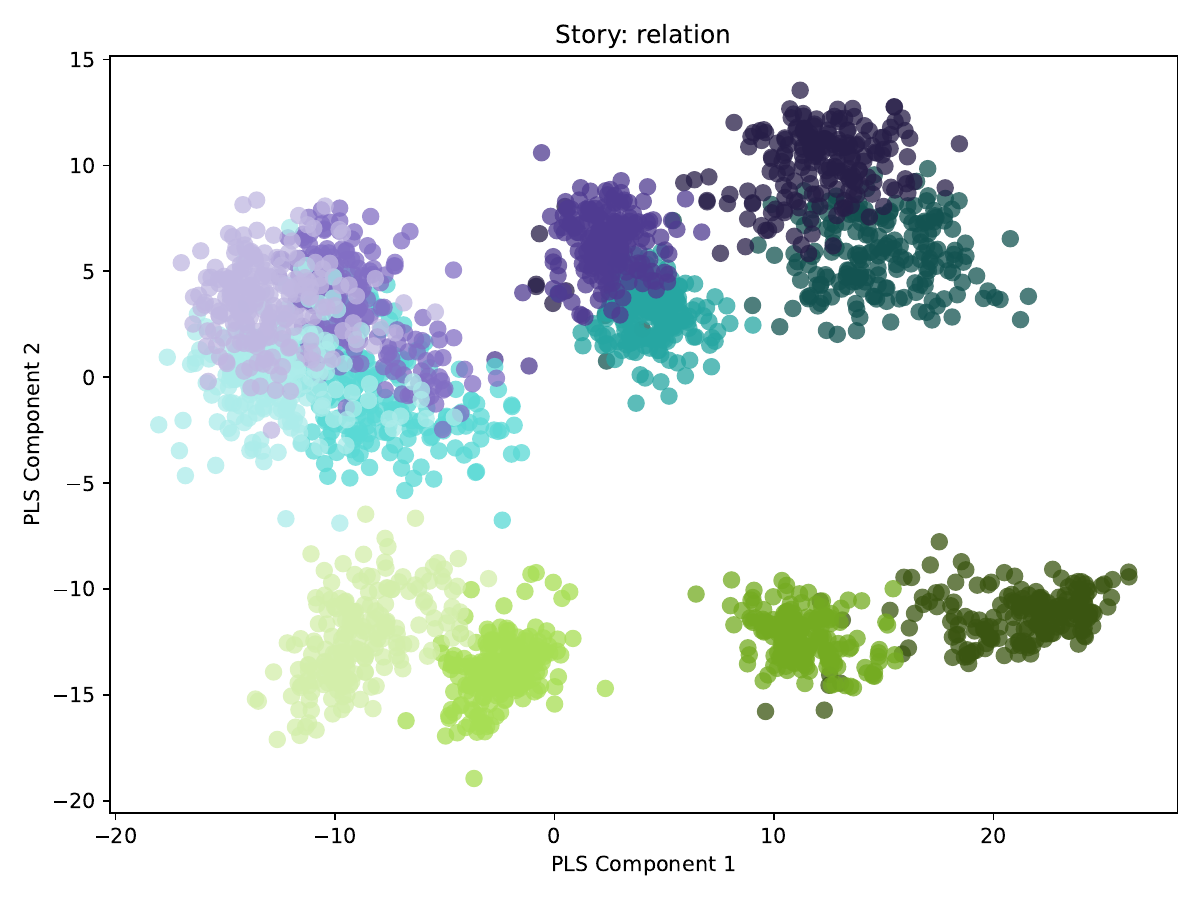}
  \caption{$C_{relation}$}
\end{subfigure}\hfil 
\begin{subfigure}{0.3\textwidth}
  \includegraphics[width=\linewidth]{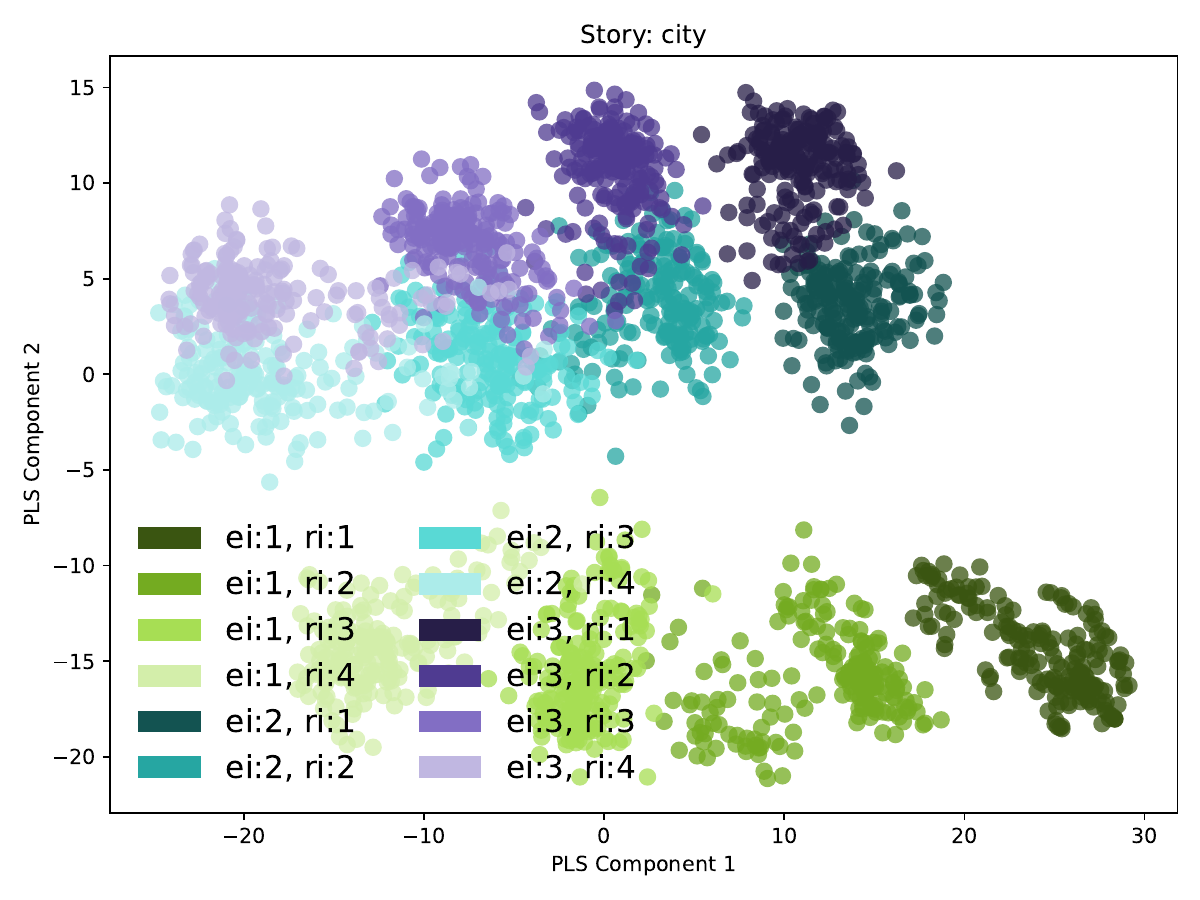}
  \caption{$C_{city}$}
\end{subfigure}\hfil
\begin{subfigure}{0.3\textwidth}
  \includegraphics[width=\linewidth]{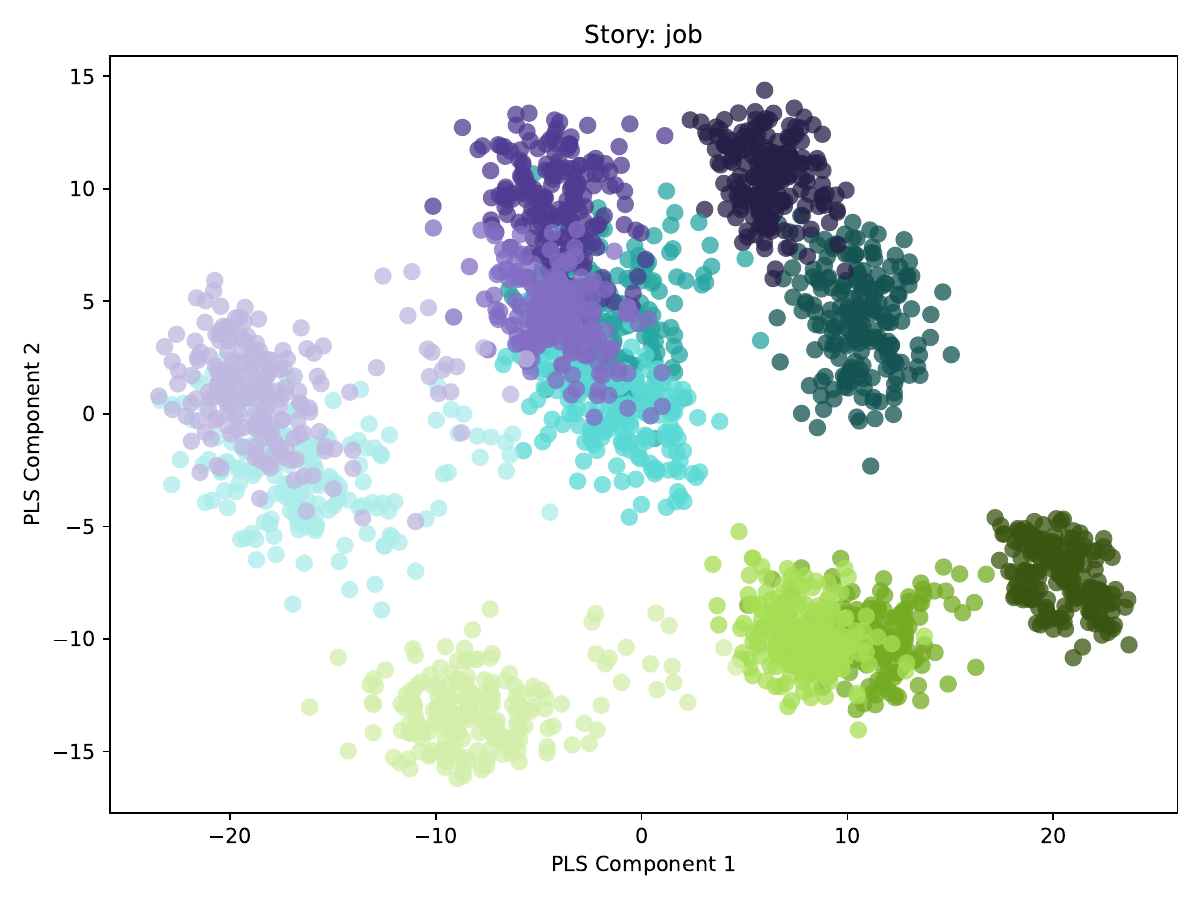}
  \caption{$C_{job}$}
\end{subfigure}\hfil 
\begin{subfigure}{0.3\textwidth}
  \includegraphics[width=\linewidth]{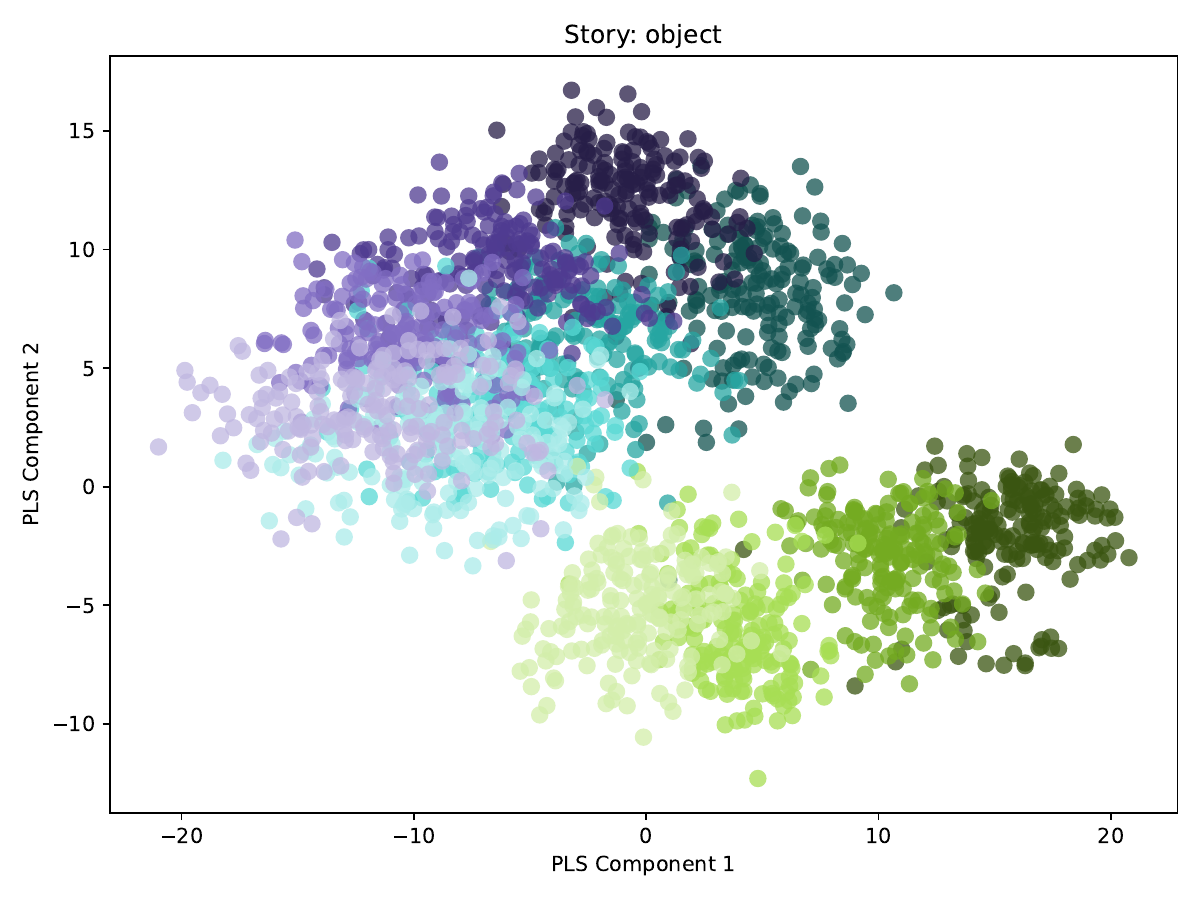}
  \caption{$C_{object}$}
\end{subfigure}\hfil 
\caption{Visualization of the CBR subspace from Qwen3-8B.}
\label{fig:irs_visualization_qwen}
\end{figure*}
The CBR subspace visualization for Qwen3-8B, shown in Figure~\ref{fig:irs_visualization_qwen}, is similarly organized along the entity and relation index, indicating that the CBR subspace is a general geometric property common to both LLM families.

\clearpage

\subsection{Layer-wise Identification of CBR Subspace}
\label{sec:irs_subspace_layerwise}

\begin{figure*}[!htbp]
    \centering 
\begin{subfigure}{0.3\textwidth}
  \includegraphics[width=\linewidth]{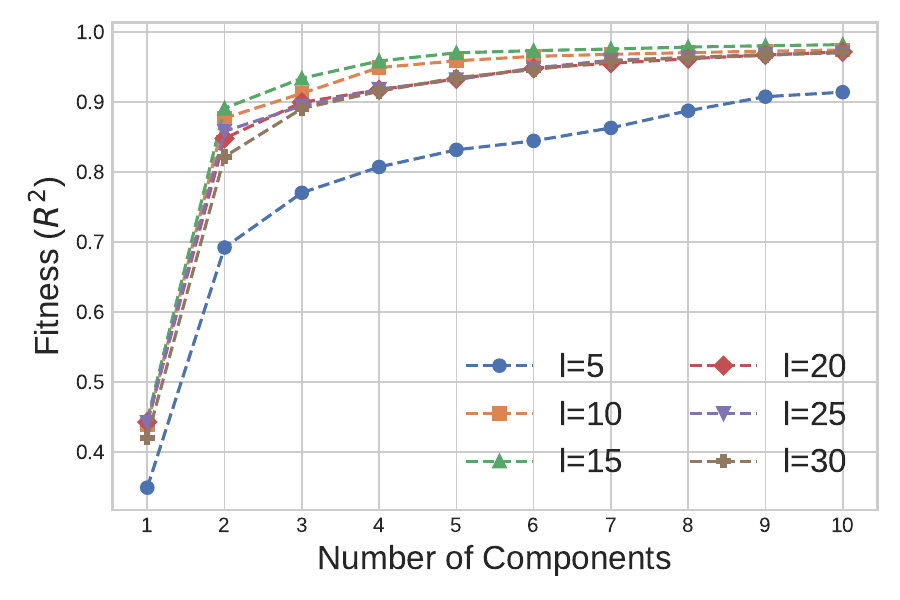}
  \caption{$C_{country}$}
  \label{fig:ll1q}
\end{subfigure}\hfil 
\begin{subfigure}{0.3\textwidth}
  \includegraphics[width=\linewidth]{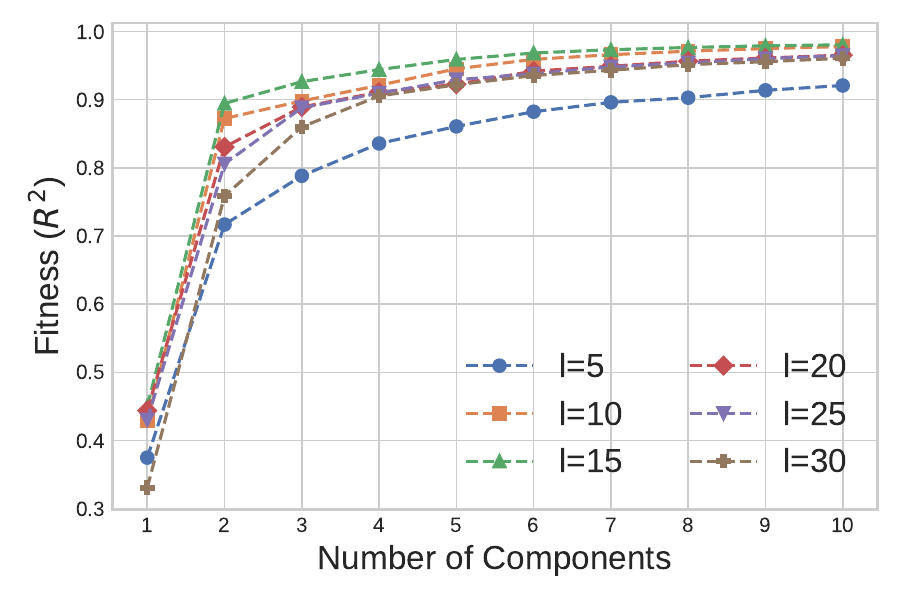}
  \caption{$C_{relation}$}
  \label{fig:ll2q}
\end{subfigure}\hfil 
\medskip
\begin{subfigure}{0.3\textwidth}
  \includegraphics[width=\linewidth]{graph/pls_layer_llama_city.pdf}
  \caption{$C_{city}$}
  \label{fig:ll3q}
\end{subfigure}\hfil
\begin{subfigure}{0.3\textwidth}\hfil
  \includegraphics[width=\linewidth]{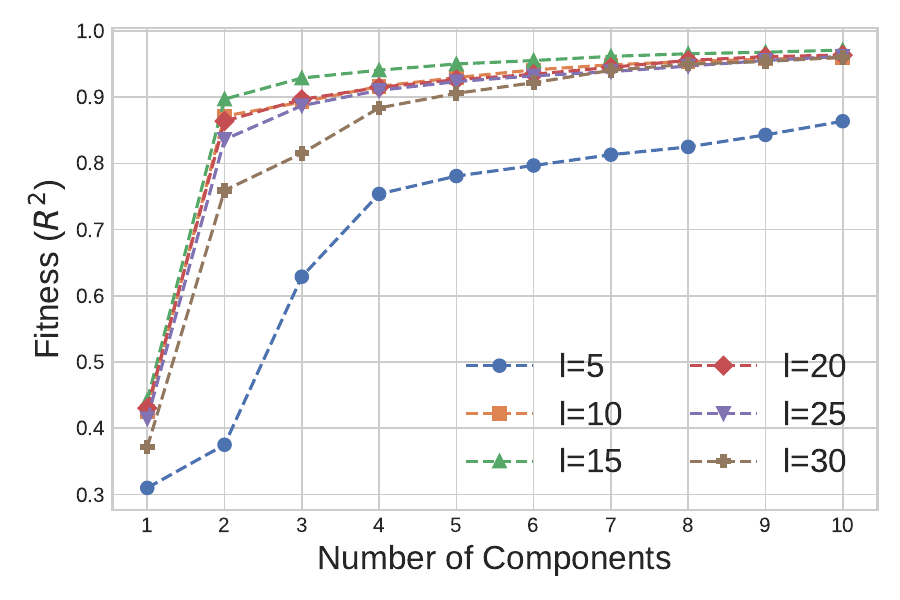}
  \caption{$C_{job}$}
  \label{fig:ll4q}
\end{subfigure}\hfil 
\begin{subfigure}{0.3\textwidth}
  \includegraphics[width=\linewidth]{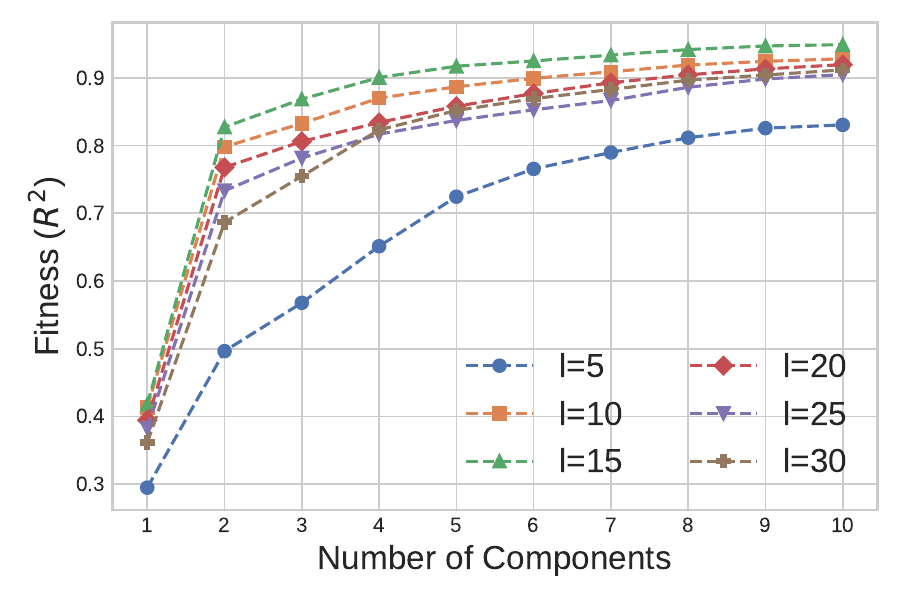}
  \caption{$C_{object}$}
  \label{fig:ll5q}
\end{subfigure}\hfil 
\caption{Decoding layer-wise performance of $[ei,ri]$ from activations of Llama3-8B-Instruct using PLS. Performance peaks in the middle layers, while both lower and higher layers show reduced fitness.}
\label{fig:pls_layer_fitness_llama}
\end{figure*}

\begin{figure*}[!htbp]
    \centering 
\begin{subfigure}{0.3\textwidth}
  \includegraphics[width=\linewidth]{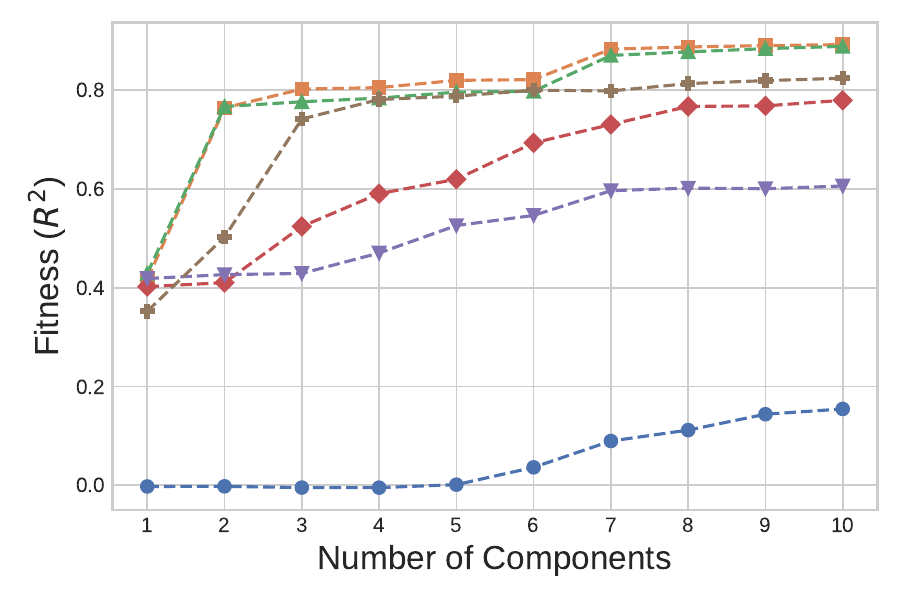}
  \caption{$C_{country}$}
  \label{fig:ll1q_}
\end{subfigure}\hfil 
\begin{subfigure}{0.3\textwidth}
  \includegraphics[width=\linewidth]{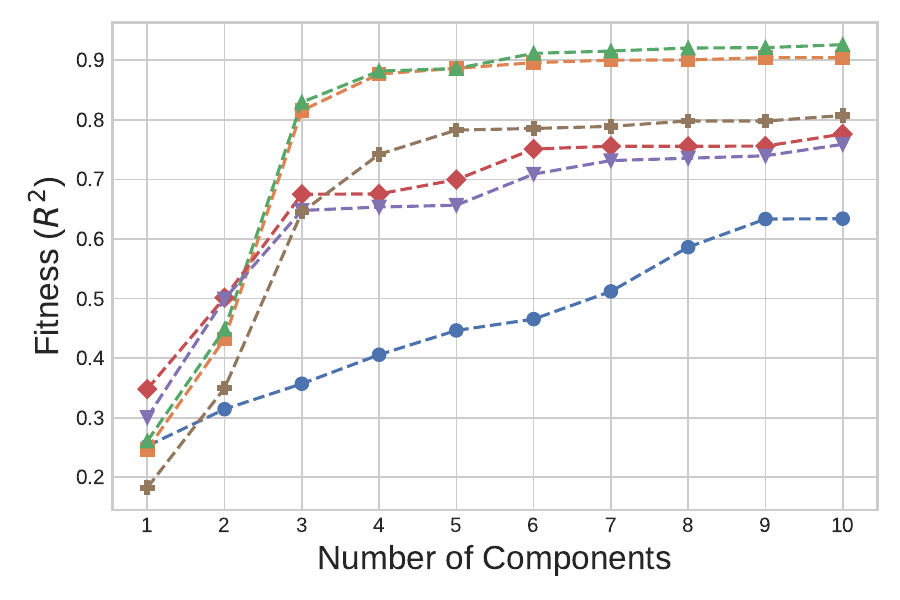}
  \caption{$C_{relation}$}
  \label{fig:ll2q_}
\end{subfigure}\hfil 
\medskip
\begin{subfigure}{0.3\textwidth}
  \includegraphics[width=\linewidth]{graph/pca_layer_llama_city.pdf}
  \caption{$C_{city}$}
  \label{fig:ll3q_}
\end{subfigure}\hfil
\begin{subfigure}{0.3\textwidth}\hfil
  \includegraphics[width=\linewidth]{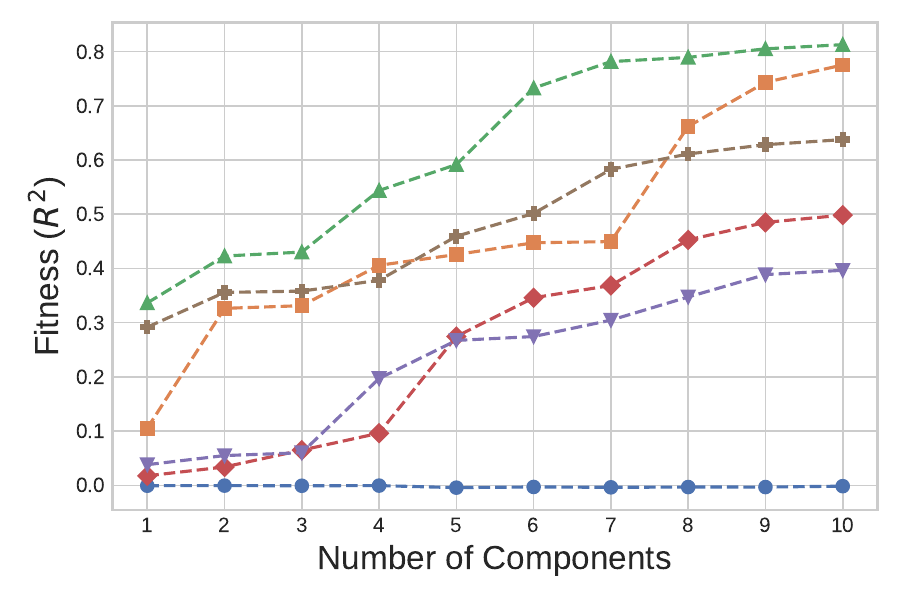}
  \caption{$C_{job}$}
  \label{fig:ll4q_}
\end{subfigure}\hfil 
\begin{subfigure}{0.3\textwidth}
  \includegraphics[width=\linewidth]{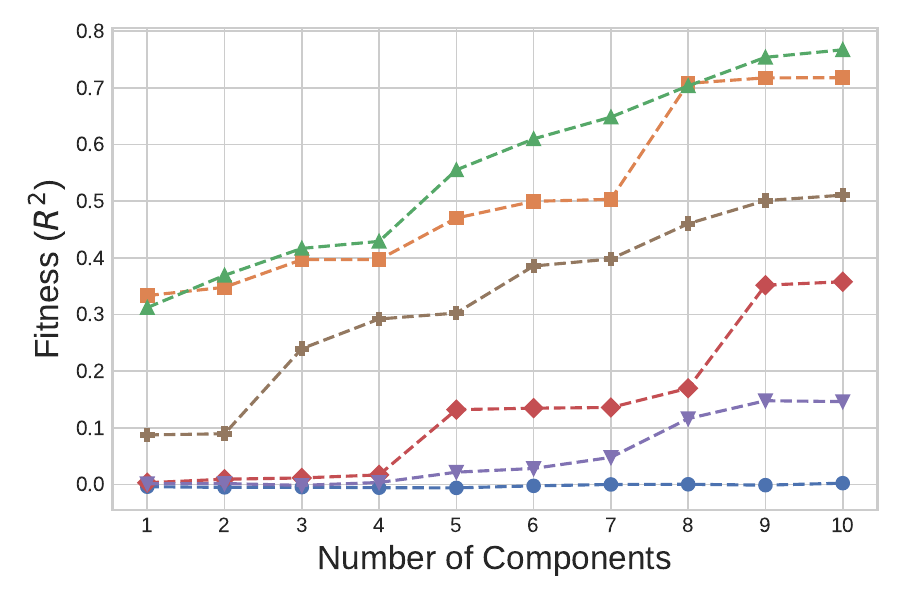}
  \caption{$C_{object}$}
  \label{fig:ll5q_}
\end{subfigure}\hfil 
\caption{Decoding layer-wise performance of $[ei,ri]$ from activations of Llama3-8B-Instruct using PCA regression.}
\label{fig:pca_layer_fitness_llama}
\end{figure*}

\begin{figure*}[!htbp]
    \centering 
\begin{subfigure}{0.3\textwidth}
  \includegraphics[width=\linewidth]{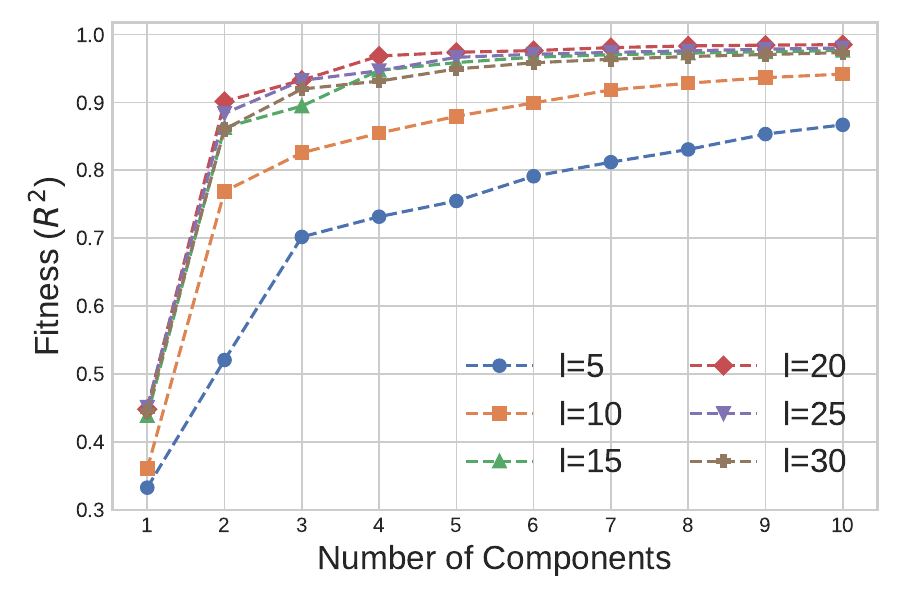}
  \caption{$C_{country}$}
  \label{fig:lll1q}
\end{subfigure}\hfil 
\begin{subfigure}{0.3\textwidth}
  \includegraphics[width=\linewidth]{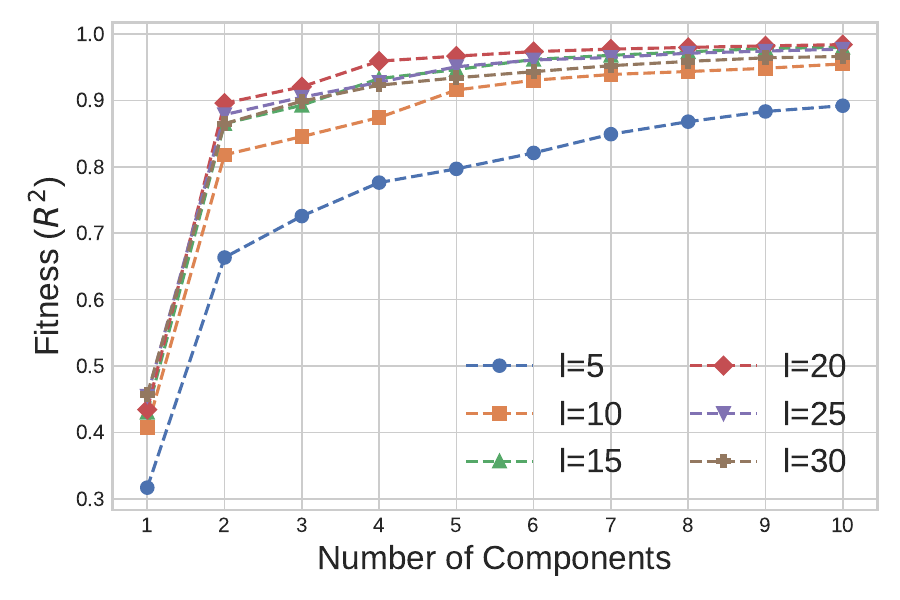}
  \caption{$C_{relation}$}
  \label{fig:lll2q}
\end{subfigure}\hfil 
\medskip
\begin{subfigure}{0.3\textwidth}
  \includegraphics[width=\linewidth]{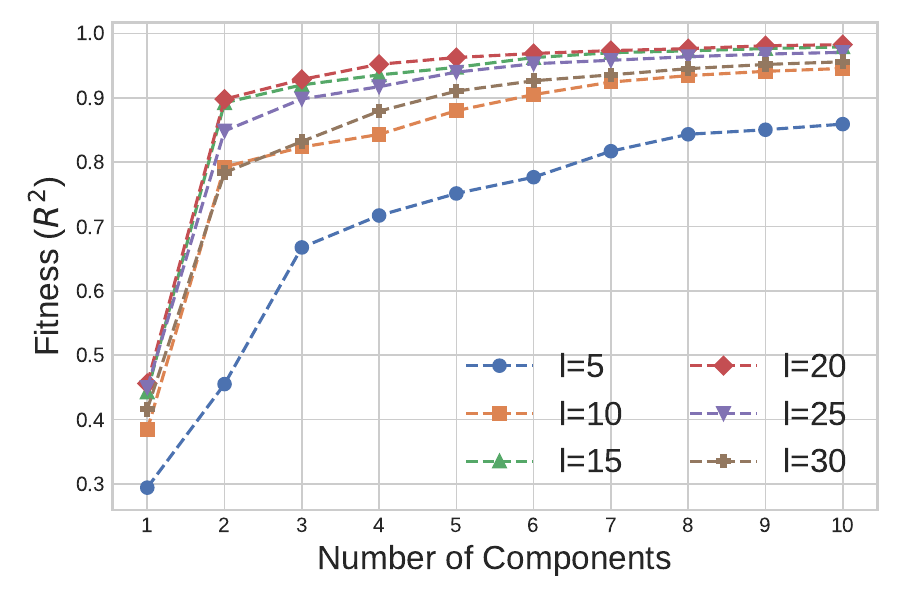}
  \caption{$C_{city}$}
  \label{fig:lll3q}
\end{subfigure}\hfil
\begin{subfigure}{0.3\textwidth}\hfil
  \includegraphics[width=\linewidth]{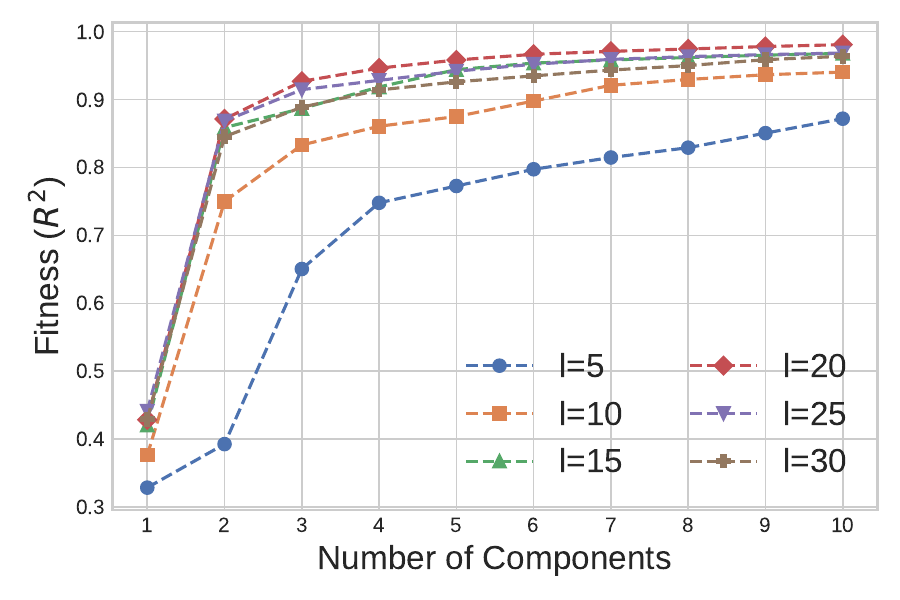}
  \caption{$C_{job}$}
  \label{fig:lll4q}
\end{subfigure}\hfil 
\begin{subfigure}{0.3\textwidth}
  \includegraphics[width=\linewidth]{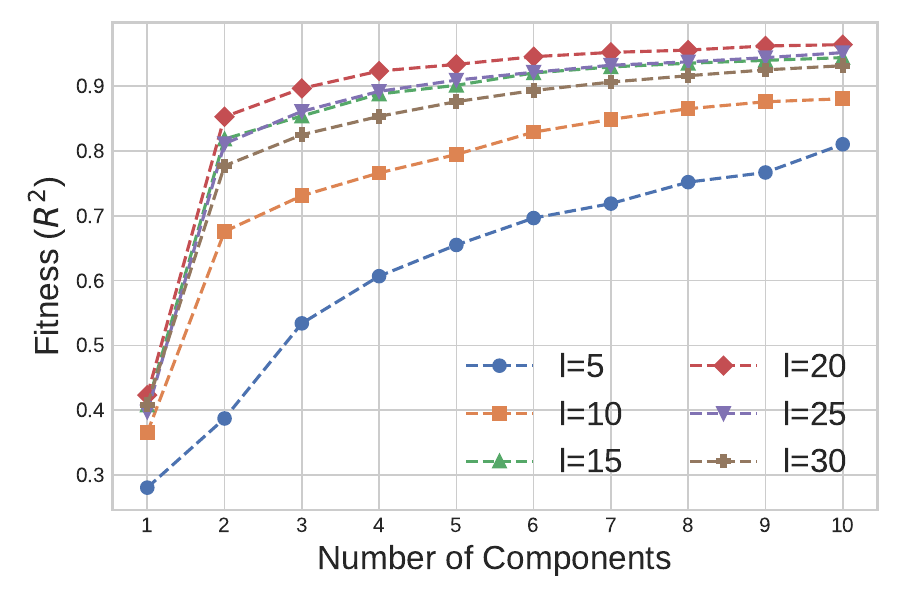}
  \caption{$C_{object}$}
  \label{fig:lll5q}
\end{subfigure}\hfil 
\caption{Decoding layer-wise performance of $[ei,ri]$ from activations of Qwen3-8B using PLS.}
\label{fig:pls_layer_fitness_qwen}
\end{figure*}

\begin{figure*}[!htbp]
    \centering 
\begin{subfigure}{0.3\textwidth}
  \includegraphics[width=\linewidth]{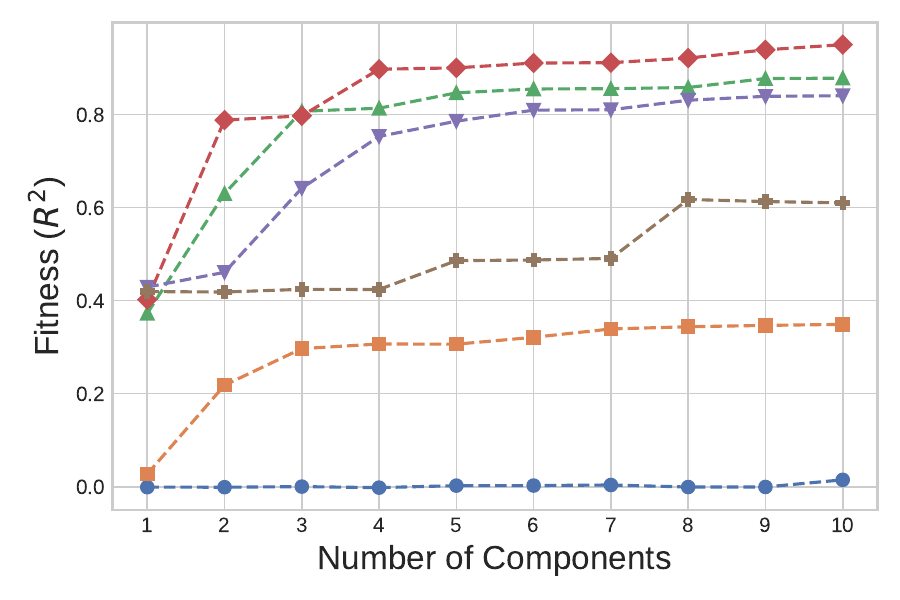}
  \caption{$C_{country}$}
  \label{fig:lll1q_}
\end{subfigure}\hfil 
\begin{subfigure}{0.3\textwidth}
  \includegraphics[width=\linewidth]{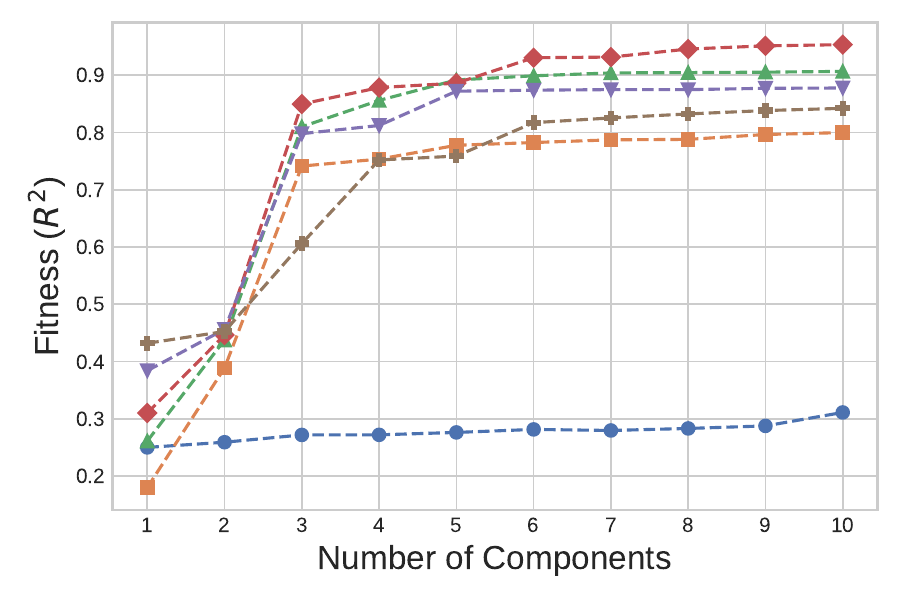}
  \caption{$C_{relation}$}
  \label{fig:lll2q_}
\end{subfigure}\hfil 
\medskip
\begin{subfigure}{0.3\textwidth}
  \includegraphics[width=\linewidth]{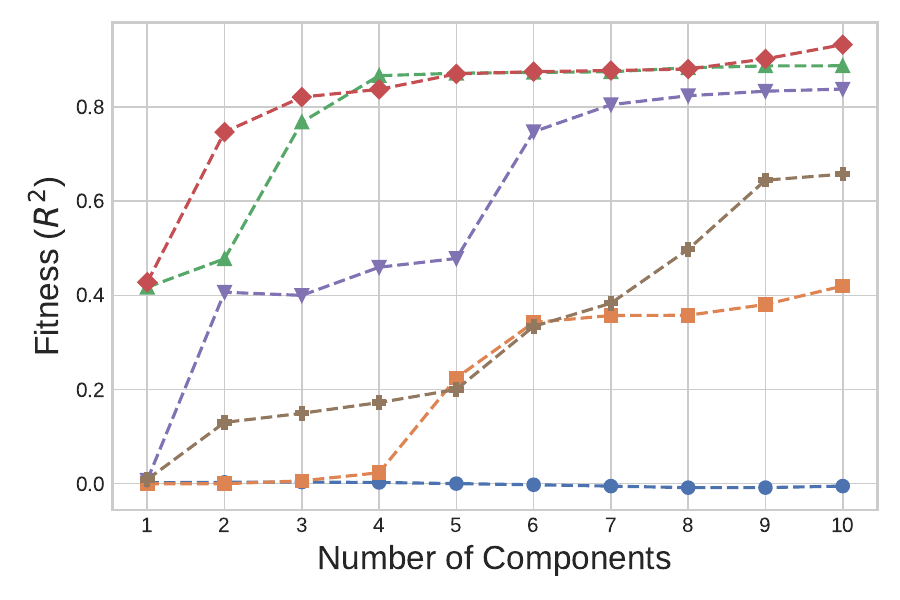}
  \caption{$C_{city}$}
  \label{fig:lll3q_}
\end{subfigure}\hfil
\begin{subfigure}{0.3\textwidth}\hfil
  \includegraphics[width=\linewidth]{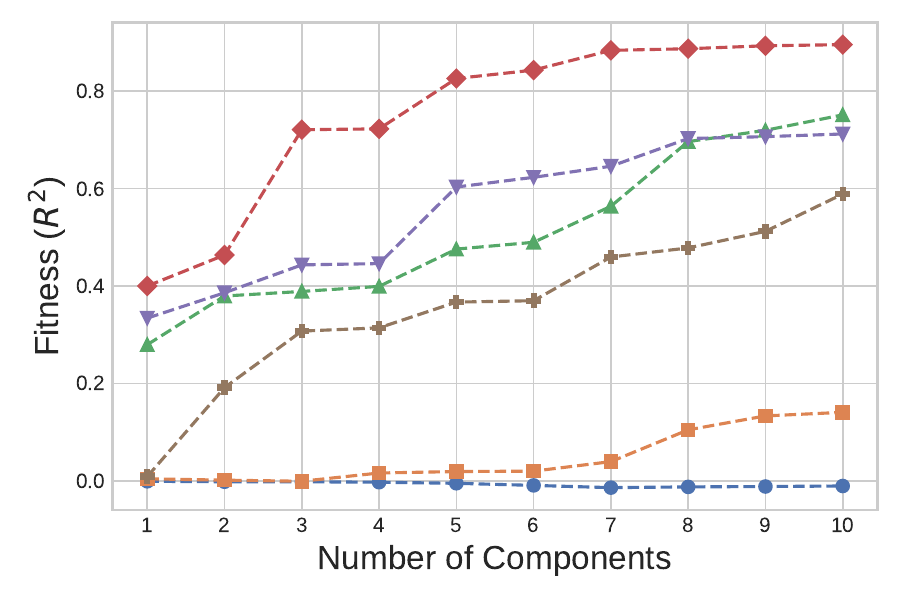}
  \caption{$C_{job}$}
  \label{fig:lll4q_}
\end{subfigure}\hfil 
\begin{subfigure}{0.3\textwidth}
  \includegraphics[width=\linewidth]{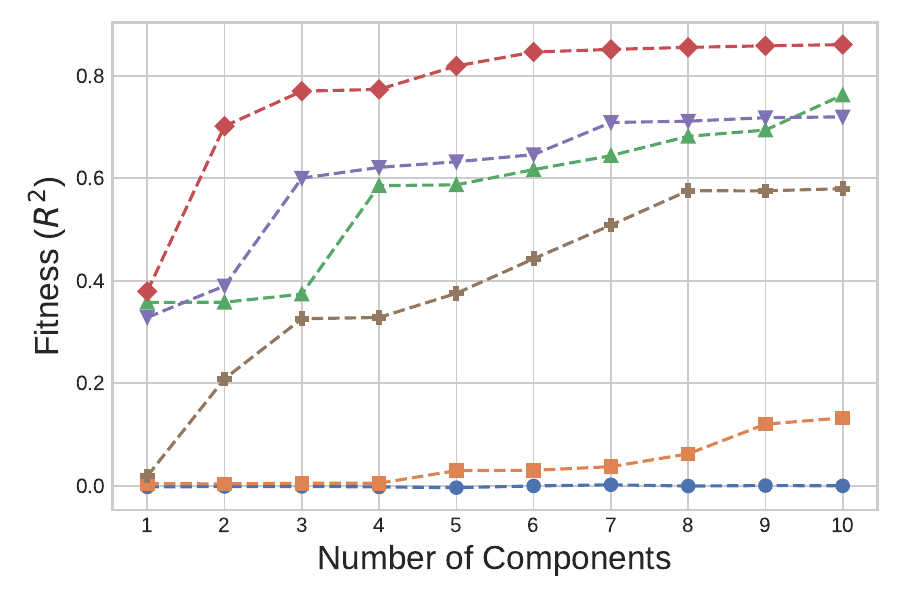}
  \caption{$C_{object}$}
  \label{fig:lll5q_}
\end{subfigure}\hfil 
\caption{Decoding layer-wise performance of $[ei,ri]$ from activations of Qwen3-8B using PCA regression.}
\label{fig:pca_layer_fitness_qwen}
\end{figure*}

We further apply linear subspace based methods (i.e., PLS and PCA regression) to predict entity and relation indices in a layer-wise setting. The results, shown in Figure~\ref{fig:pls_layer_fitness_llama}, \ref{fig:pca_layer_fitness_llama}, \ref{fig:pls_layer_fitness_qwen} and \ref{fig:pca_layer_fitness_qwen}, indicate that the $R^2$ score peaks in the middle layers (e.g., $l=15$) and is lower in both earlier layers (e.g., $l=5$) and later layers (e.g., $l=25$). This pattern is consistent with the ``stages of inference hypothesis''~\cite{lad2024remarkable}, which states that intermediate layers are primarily responsible for feature engineering (e.g., the feature of the discourse relational structure in our case). In addition, the same trend is observed for both Llama3-8B-Instruct and Qwen3-8B, suggesting that this layer-wise organization of the CBR subspace is consistent across model families.


\clearpage

\subsection{Monotonic Encoding of CBR Indices}
\label{sec:mono}
To analyze the monotonic property of the CBR index, we consider not only the standard integer indices (i.e., $[1, 2, 3, 4]$), but also several alternative monotonic sequences. Specifically, we construct exponential sequences (i.e., $[1, 3, 9, 27]$), logarithmic sequences (i.e., $[1, 4.64, 21.54, 100]$), and manually selected monotonic values (e.g., $[1, 5, 30, 100]$) to represent the indices of $ei$ and $ri$. For example, under the manually selected sequence, the CBR index of the second entity or relation is assigned the value $5$ instead of $2$.

\begin{figure*}[!htbp]
    \centering 
\begin{subfigure}{0.25\textwidth}
  \includegraphics[width=\linewidth]{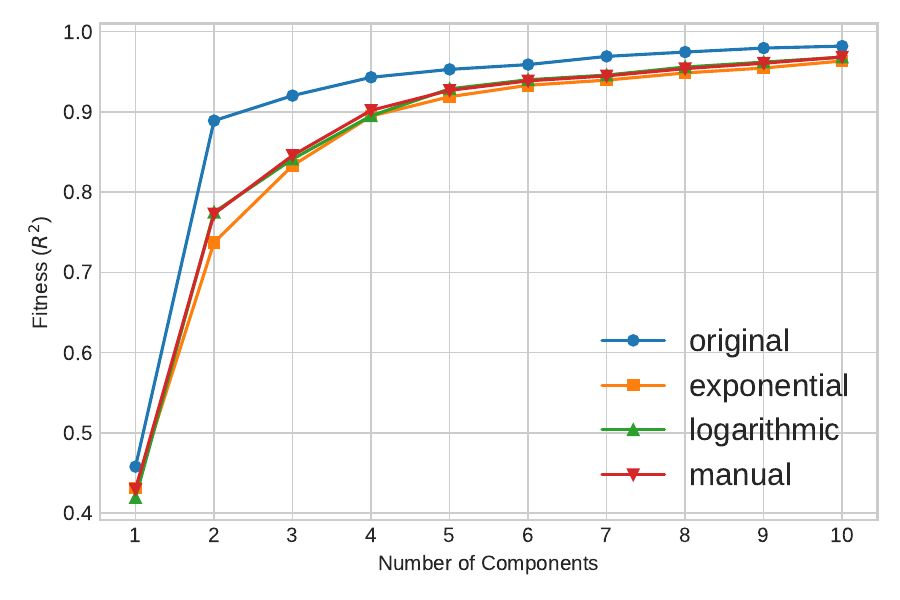}
  \caption{$C_{city}$}
\end{subfigure}\hfil 
\begin{subfigure}{0.25\textwidth}
  \includegraphics[width=\linewidth]{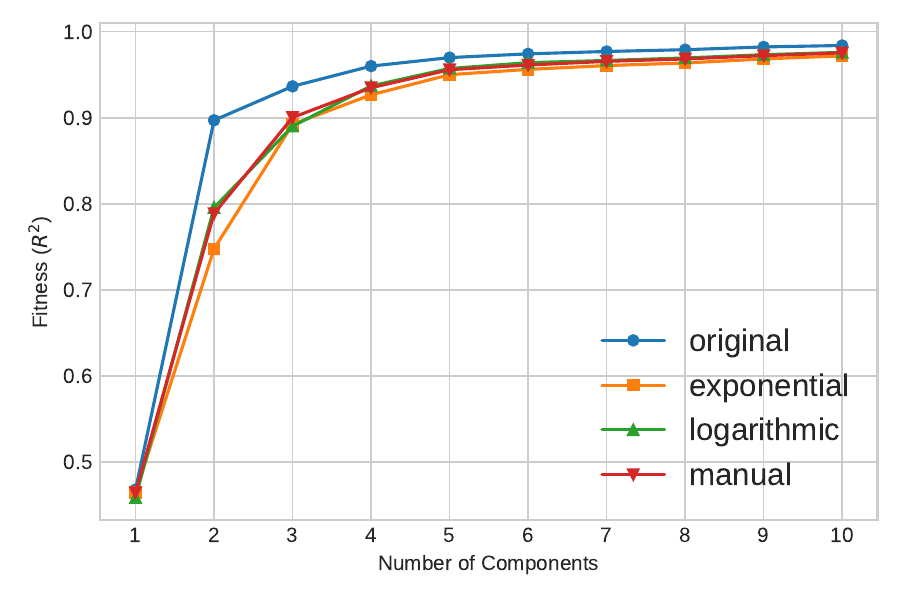}
  \caption{$C_{country}$}
\end{subfigure}\hfil 
\begin{subfigure}{0.25\textwidth}
  \includegraphics[width=\linewidth]{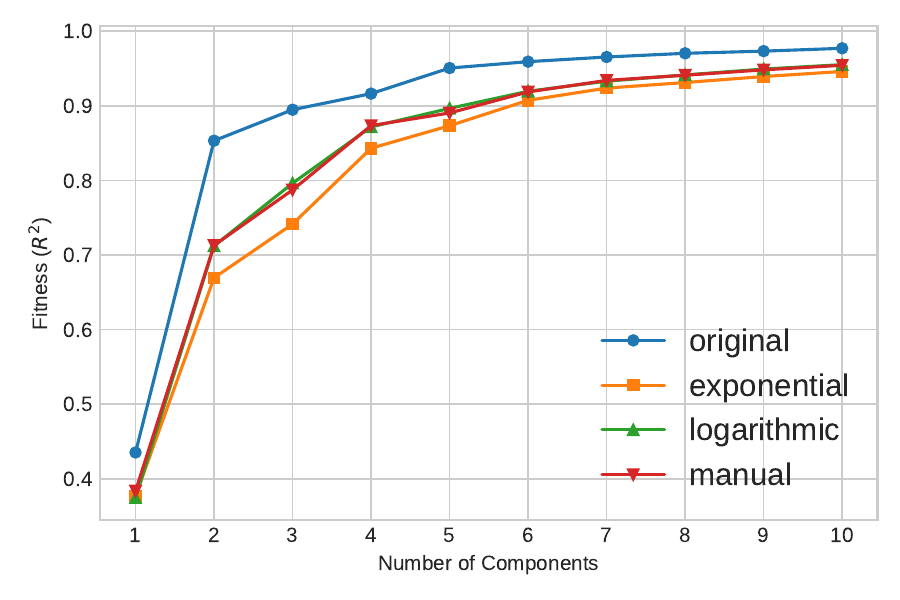}
  \caption{$C_{relation}$}
\end{subfigure}\hfil
\begin{subfigure}{0.25\textwidth}
  \includegraphics[width=\linewidth]{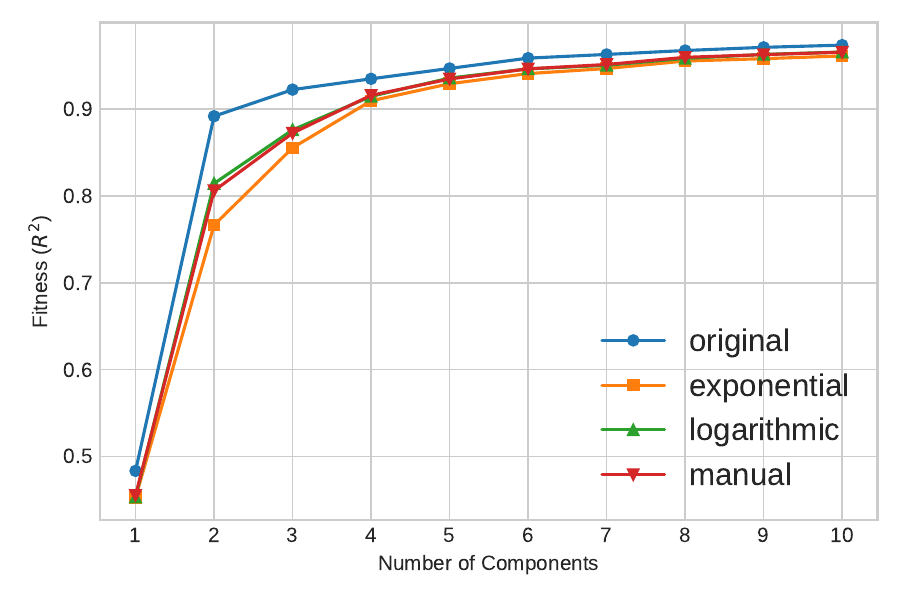}
  \caption{$C_{job}$}
\end{subfigure}\hfil 
\begin{subfigure}{0.25\textwidth}
  \includegraphics[width=\linewidth]{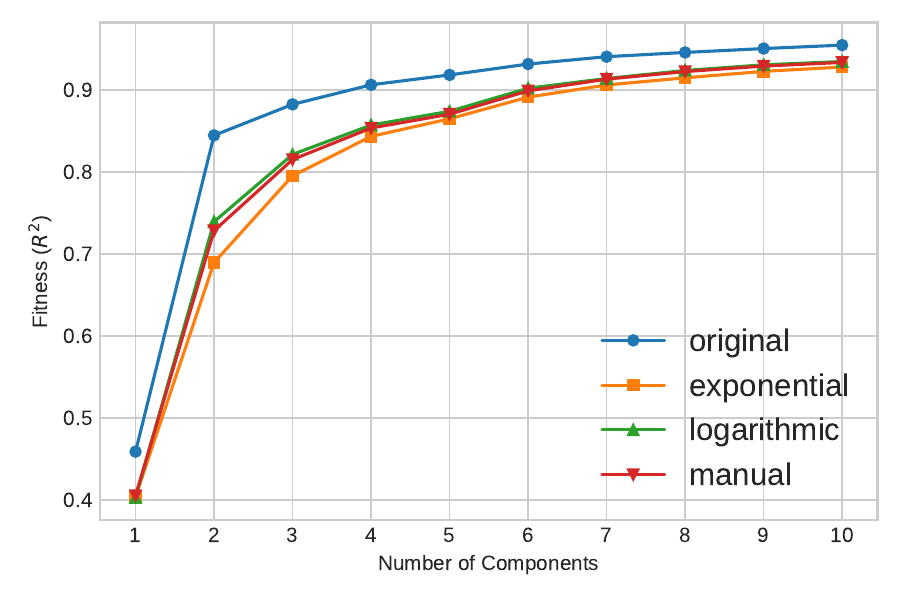}
  \caption{$C_{object}$}
\end{subfigure}\hfil 
\caption{Decoding performance of $[ei,ri]$ from activations of Llama3-8B-Instruct under different monotonic index sequences. ``original'' denotes the original integer index.}
\label{fig:mono_llama}
\end{figure*}

\begin{figure*}[!htbp]
    \centering 
\begin{subfigure}{0.25\textwidth}
  \includegraphics[width=\linewidth]{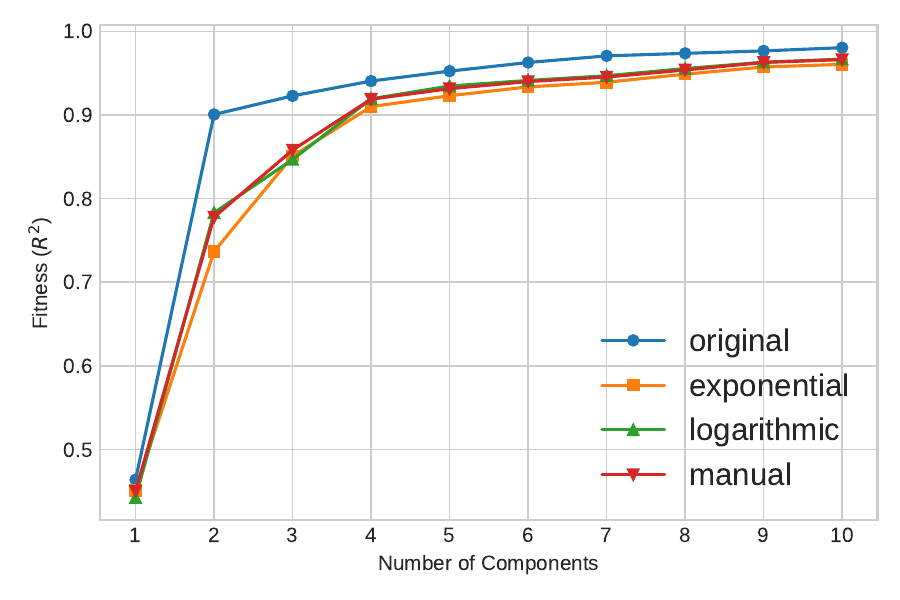}
  \caption{$C_{city}$}
\end{subfigure}\hfil 
\begin{subfigure}{0.25\textwidth}
  \includegraphics[width=\linewidth]{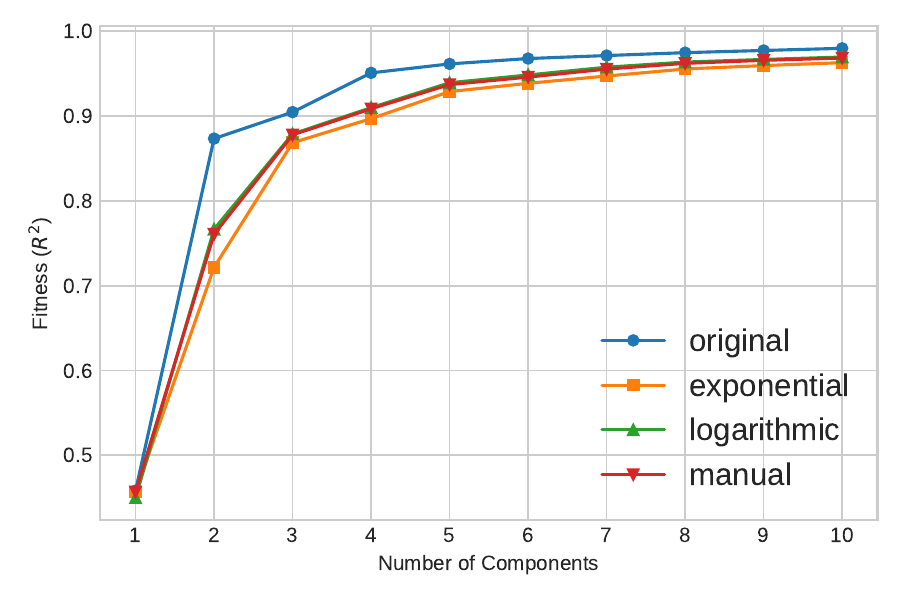}
  \caption{$C_{country}$}
\end{subfigure}\hfil 
\begin{subfigure}{0.25\textwidth}
  \includegraphics[width=\linewidth]{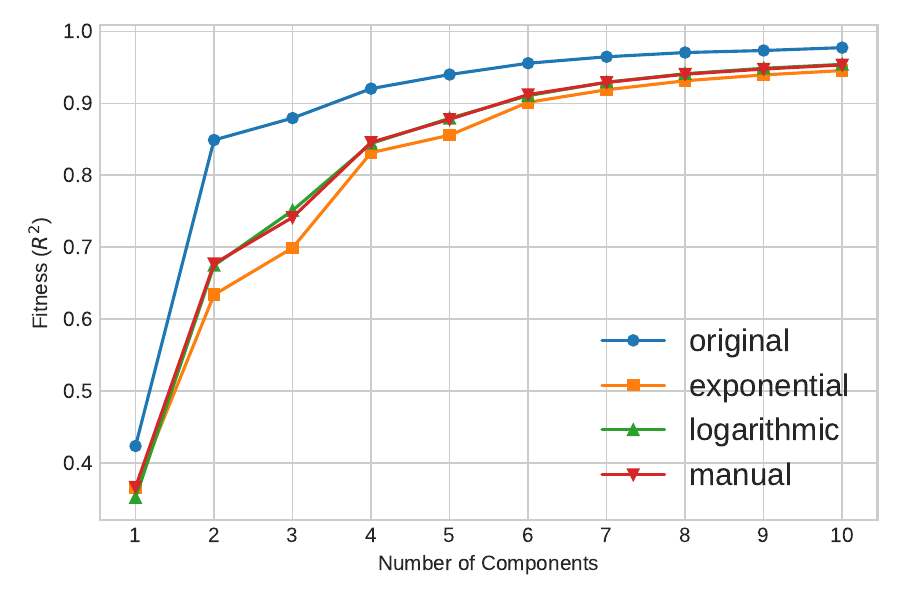}
  \caption{$C_{relation}$}
\end{subfigure}\hfil
\begin{subfigure}{0.25\textwidth}
  \includegraphics[width=\linewidth]{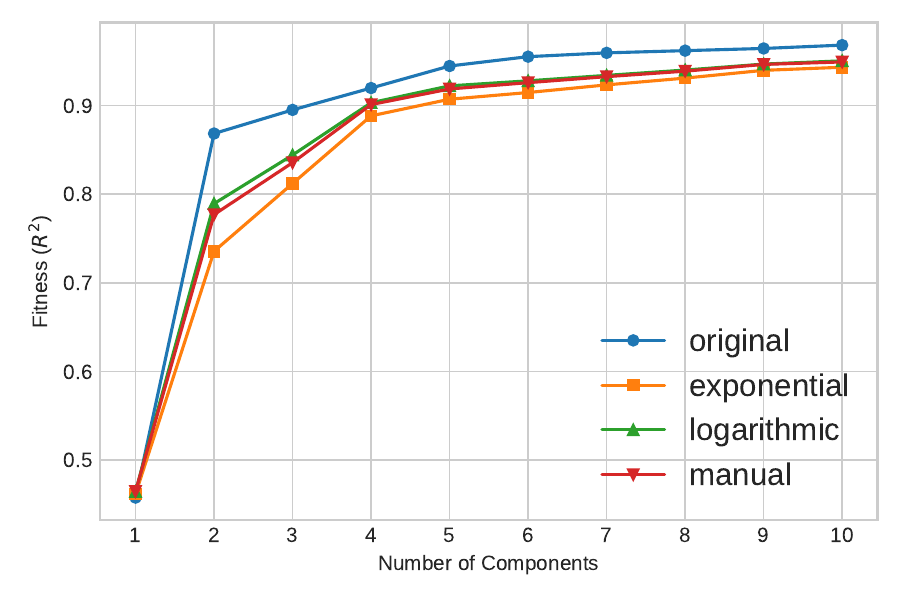}
  \caption{$C_{job}$}
\end{subfigure}\hfil 
\begin{subfigure}{0.25\textwidth}
  \includegraphics[width=\linewidth]{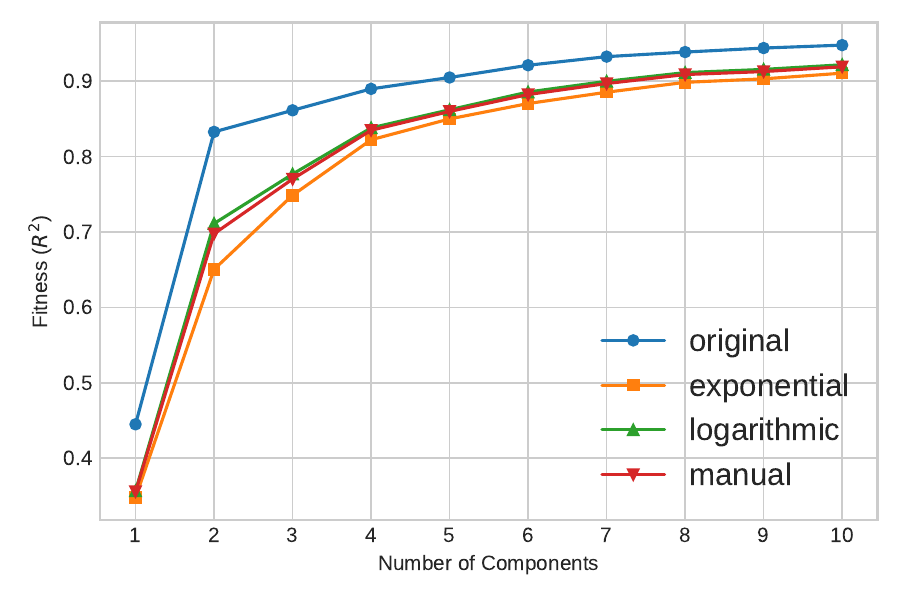}
  \caption{$C_{object}$}
\end{subfigure}\hfil 
\caption{Decoding performance of different monotonic $[ei,ri]$ from activations of Qwen3-8B.}
\label{fig:mono_qwen}
\end{figure*}

We then apply PLS to predict these indices from the hidden activations. The decoding performance is shown in Figure~\ref{fig:mono_llama} and \ref{fig:mono_qwen}. The results show that PLS can successfully predict not only the original CBR indices $[1,2,3,\ldots]$, but also arbitrary monotonic transformations of these indices, including exponential sequences and irregularly spaced monotonic values. Performance is slightly higher when using the original integer indices.

This observation suggests that the activations encode the \emph{ordinal structure} of CBR cells rather than their exact numerical indices. In particular, the representations appear to organize CBR cells along a latent direction in activation space that preserves their ordering, allowing linear probes to recover any monotonic transformation of the indices. These findings indicate that CBR indices are encoded through an \emph{order-preserving embedding} in the activation space of LLMs.


\clearpage

\subsection{CBR Subspace and Semantic Information on Qwen3-8B}
\label{sec:irs_subspace_semantic_qwen}
\begin{figure}[!htbp]
    \centering 
\begin{subfigure}{0.2\textwidth}
  \includegraphics[width=\linewidth]{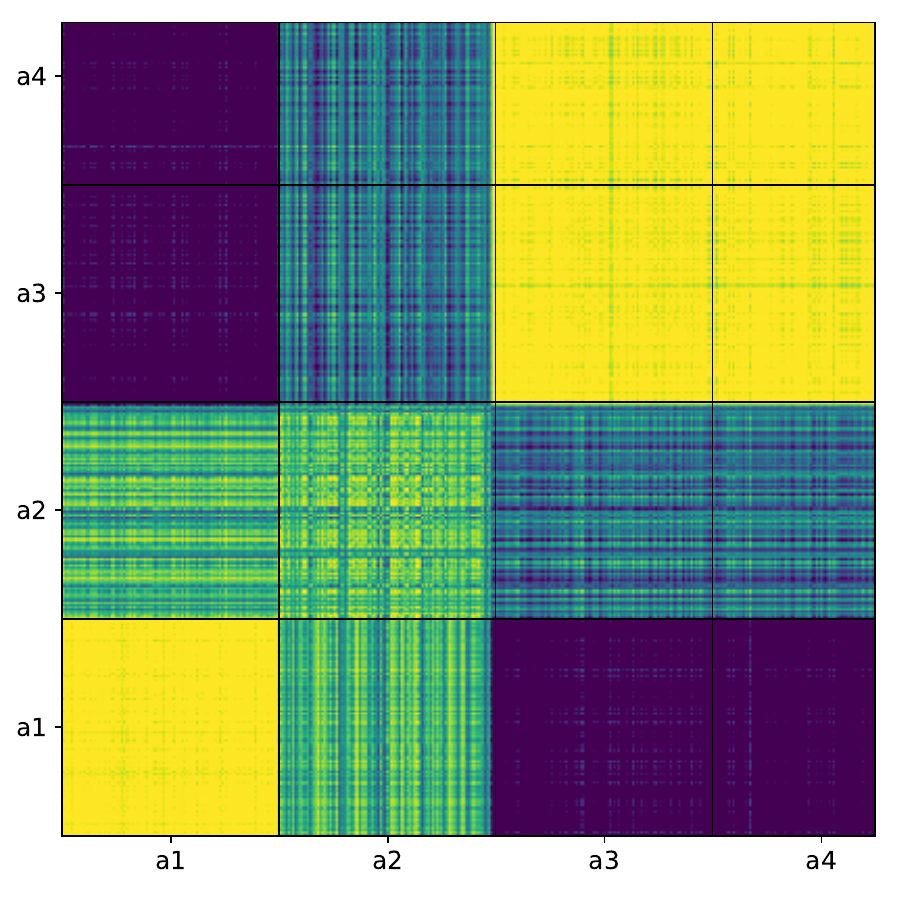}
  \caption{2to2}
\end{subfigure}\hfil 
\begin{subfigure}{0.2\textwidth}
  \includegraphics[width=\linewidth]{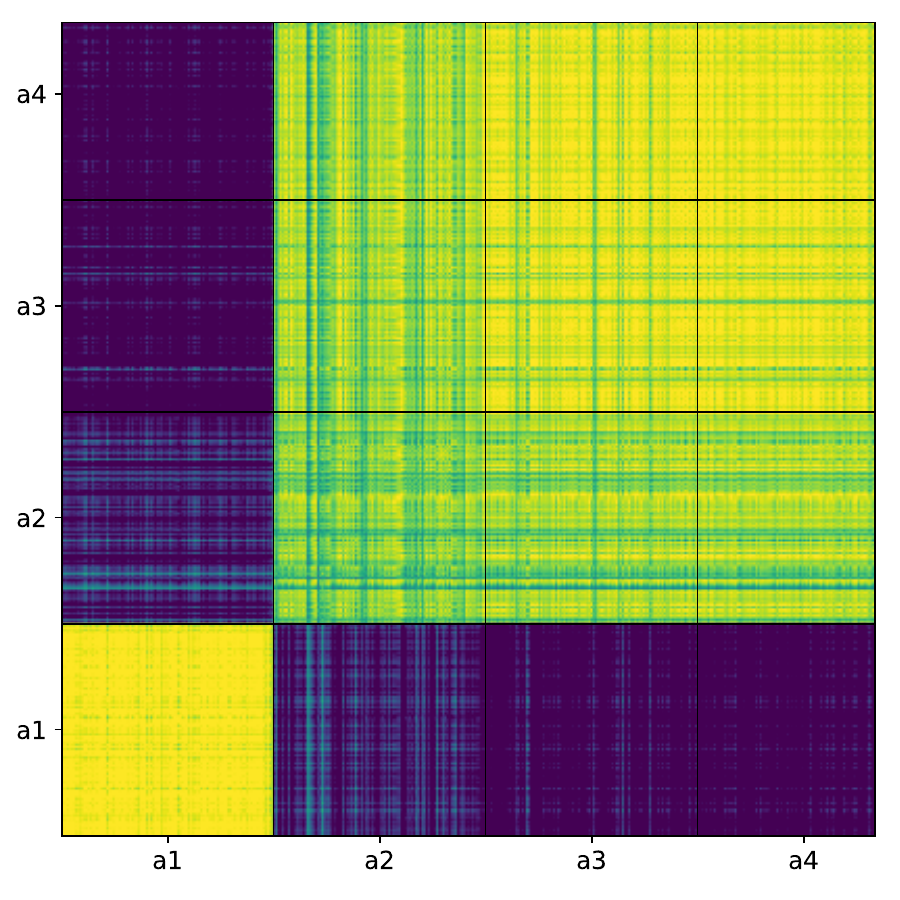}
  \caption{1to3}
\end{subfigure}\hfil 
\begin{subfigure}{0.2\textwidth}
  \includegraphics[width=\linewidth]{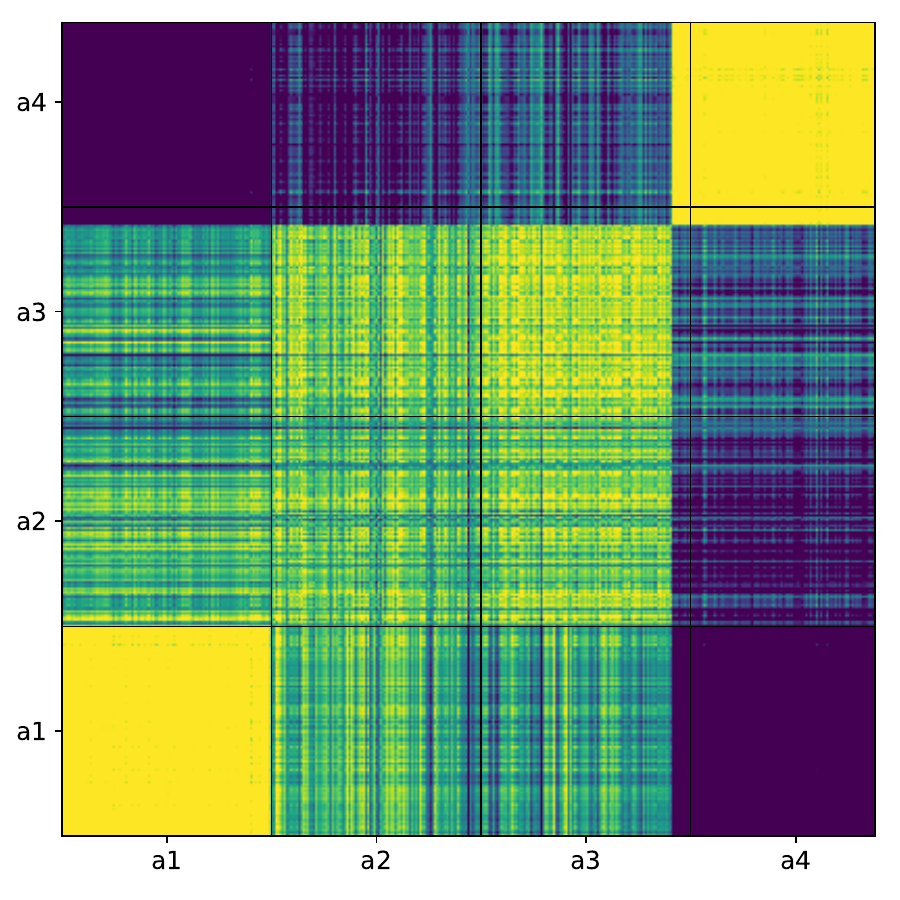}
  \caption{3to1}
\end{subfigure}

\medskip
\begin{subfigure}{0.2\textwidth}
  \includegraphics[width=\linewidth]{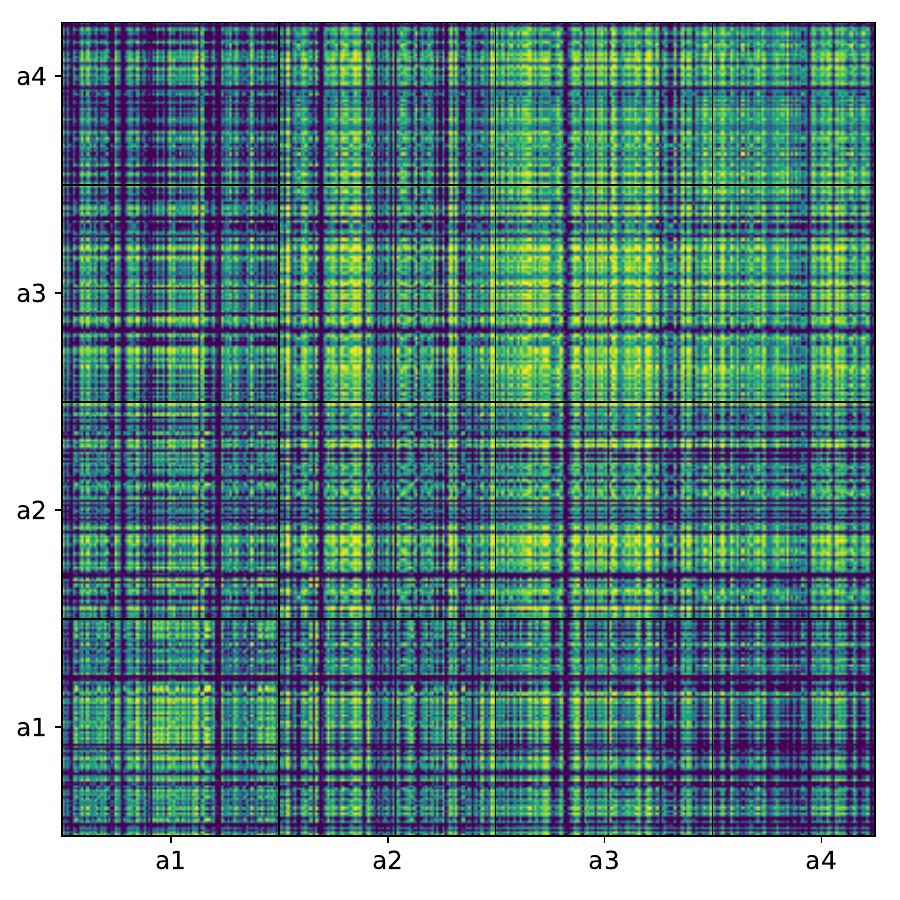}
  \caption{2to2 (Rand.)}
\end{subfigure}\hfil 
\begin{subfigure}{0.2\textwidth}
  \includegraphics[width=\linewidth]{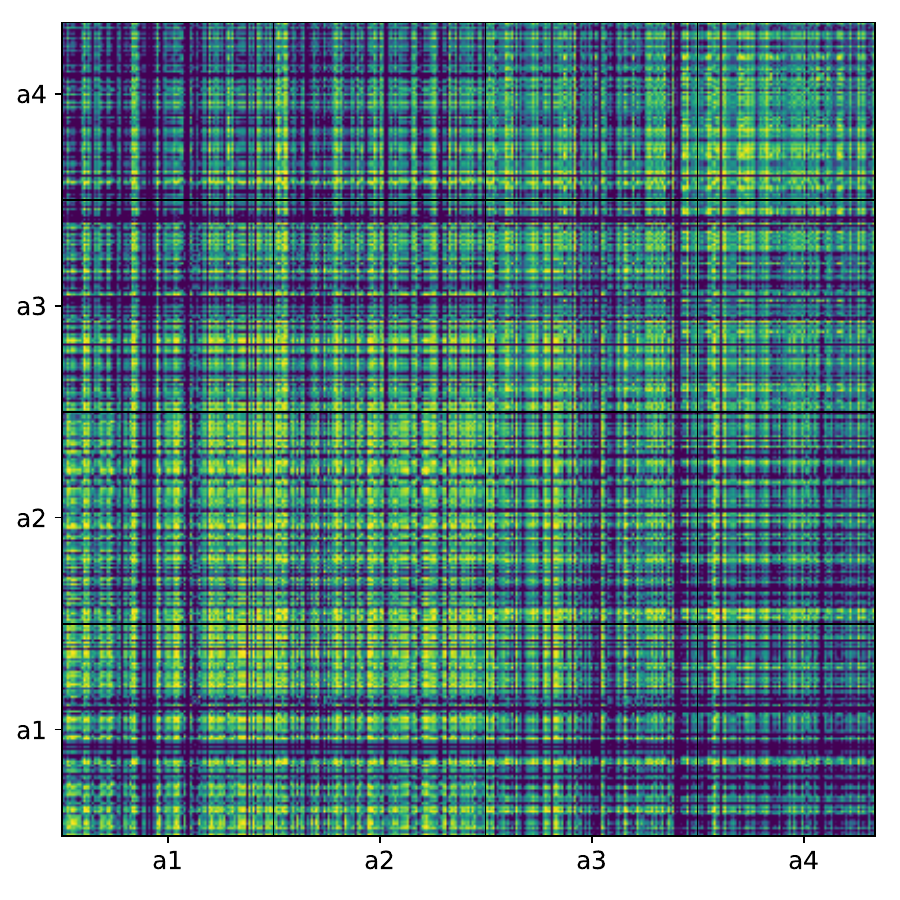}
  \caption{1to3 (Rand.)}
\end{subfigure}\hfil 
\begin{subfigure}{0.2\textwidth}
  \includegraphics[width=\linewidth]{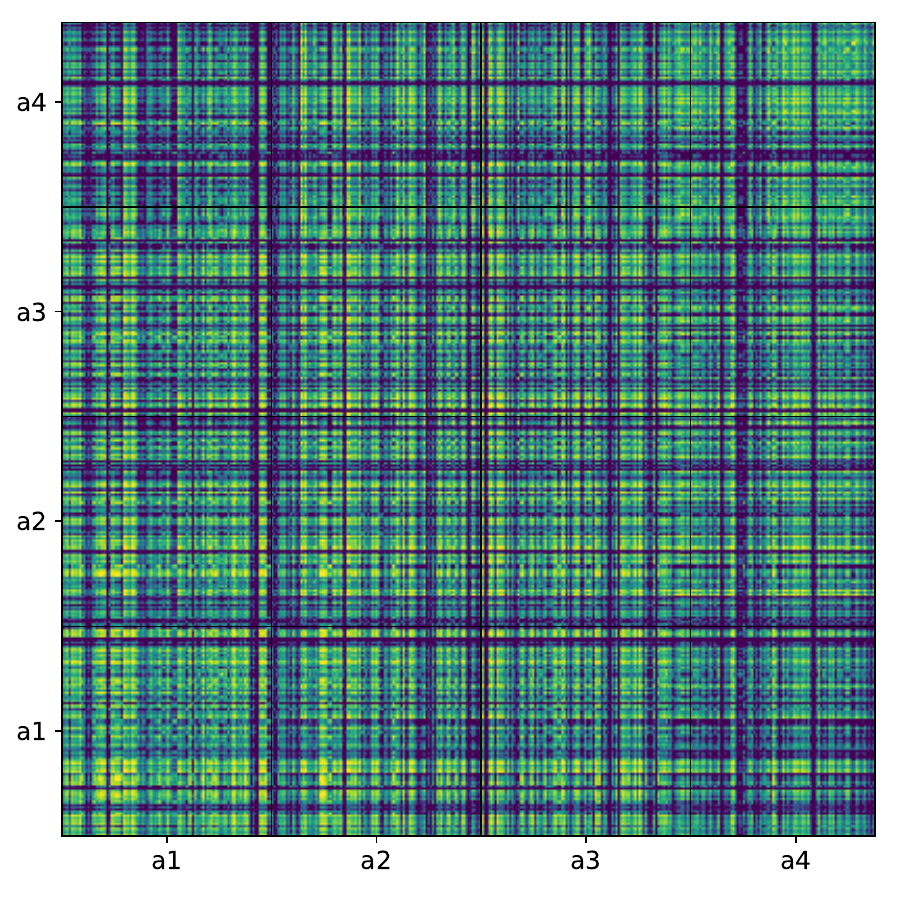}
  \caption{3to1 (Rand.)}
\end{subfigure}
\caption{Cosine similarity heatmaps of attribute representations projected into the CBR subspace (above) and a random subspace (below) from Qwen3-8B on $C_{job}$. The CBR projection recovers the intended semantic similarity structure among relations, whereas the random projection does not clearly capture it.}
\label{fig:semantic_pattern_job_qwen}
\end{figure}

\begin{figure}[!htbp]
    \centering 
\begin{subfigure}{0.2\textwidth}
  \includegraphics[width=\linewidth]{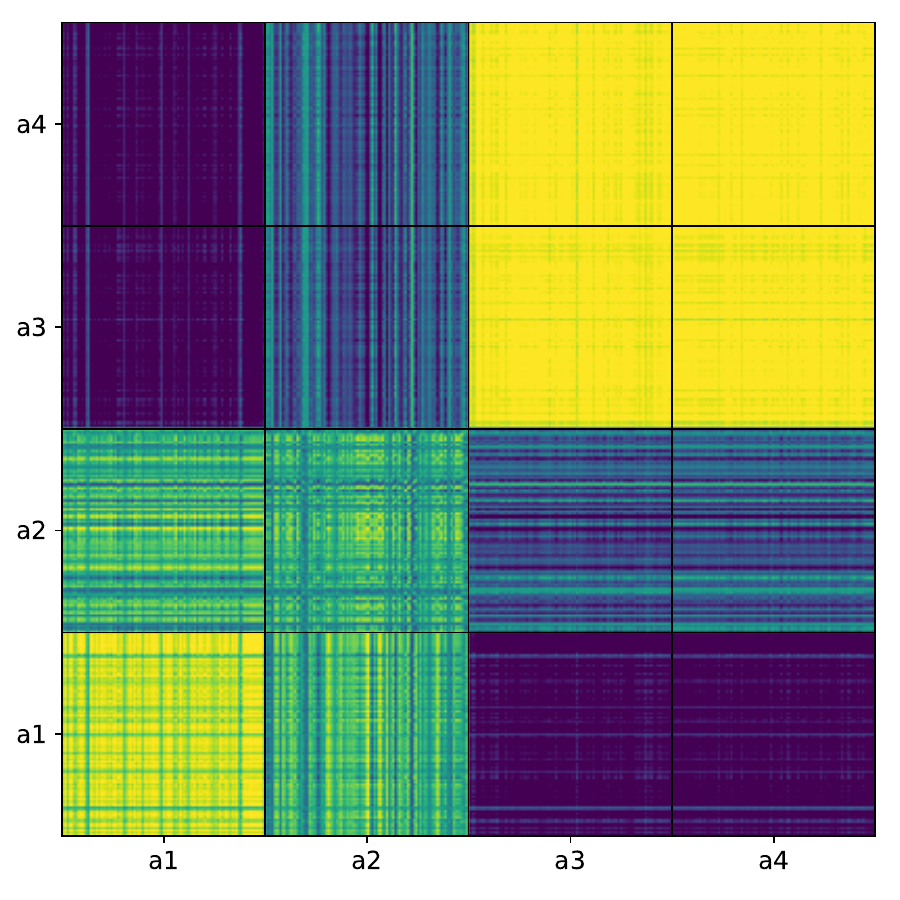}
  \caption{2to2}
\end{subfigure}\hfil 
\begin{subfigure}{0.2\textwidth}
  \includegraphics[width=\linewidth]{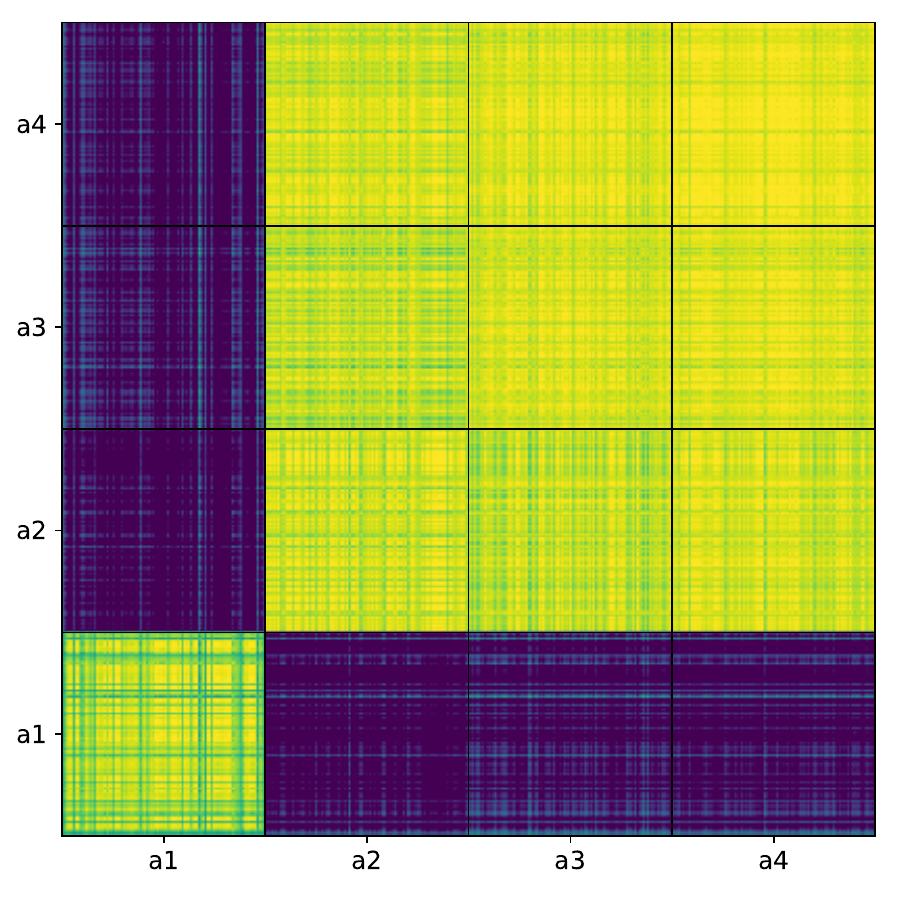}
  \caption{1to3}
\end{subfigure}\hfil 
\begin{subfigure}{0.2\textwidth}
  \includegraphics[width=\linewidth]{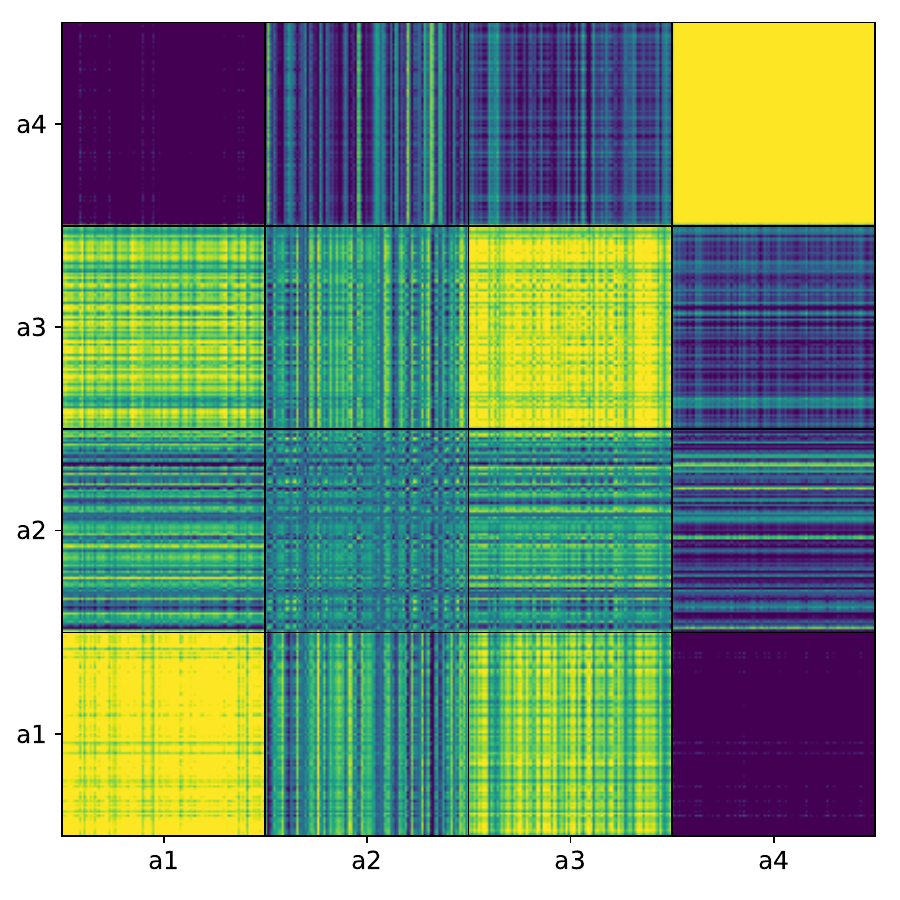}
  \caption{3to1}
\end{subfigure}

\medskip
\begin{subfigure}{0.2\textwidth}
  \includegraphics[width=\linewidth]{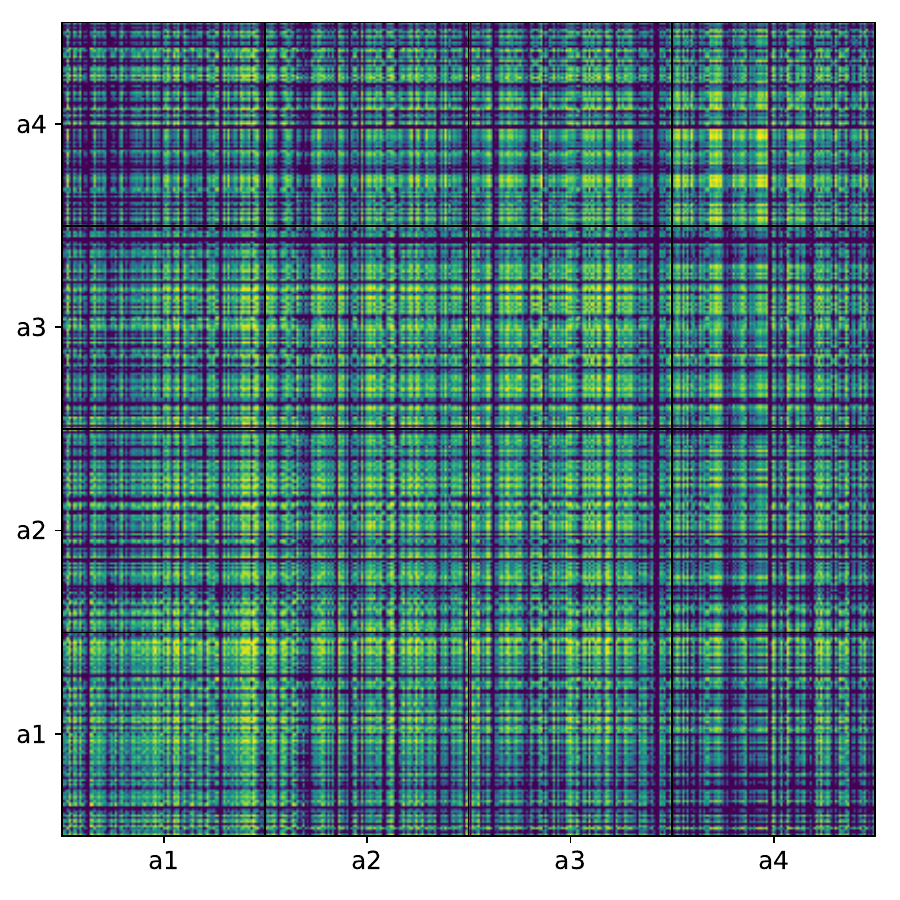}
  \caption{2to2 (Rand.)}
\end{subfigure}\hfil 
\begin{subfigure}{0.2\textwidth}
  \includegraphics[width=\linewidth]{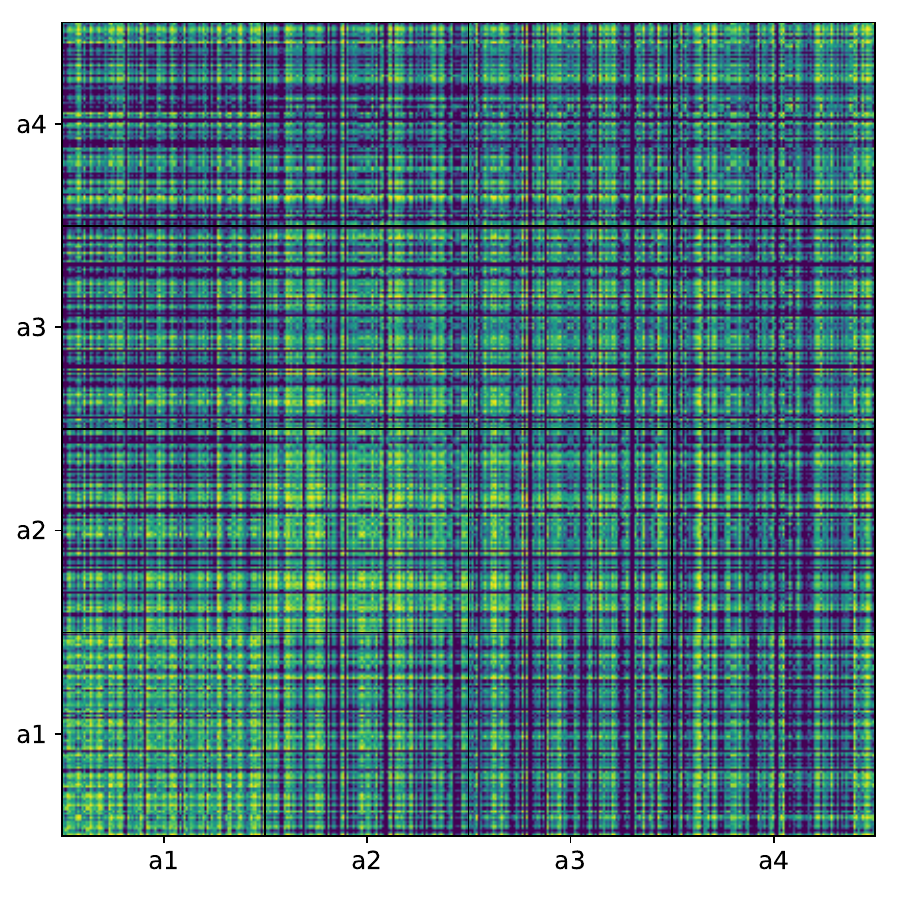}
  \caption{1to3 (Rand.)}
\end{subfigure}\hfil 
\begin{subfigure}{0.2\textwidth}
  \includegraphics[width=\linewidth]{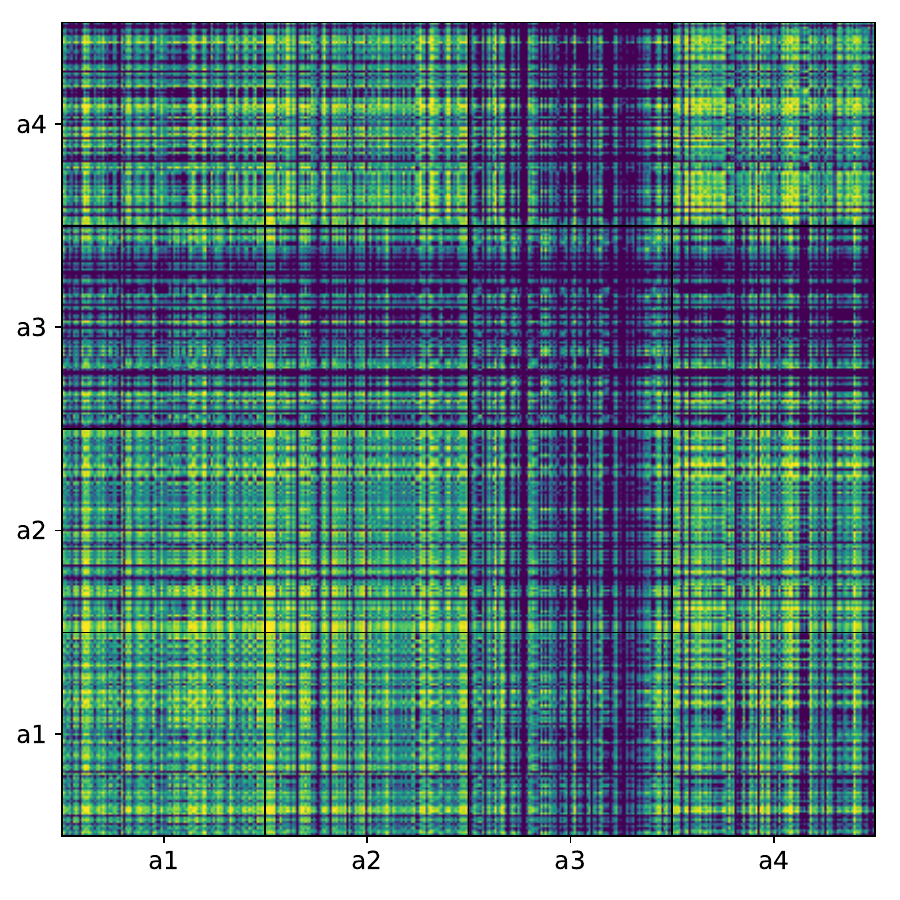}
  \caption{3to1 (Rand.)}
\end{subfigure}
\caption{Cosine similarity heatmaps of attribute representations from Qwen3-8B on $C_{city}$.}
\label{fig:semantic_pattern_city_qwen}
\end{figure}

\begin{figure}[!htbp]
    \centering 
\begin{subfigure}{0.2\textwidth}
  \includegraphics[width=\linewidth]{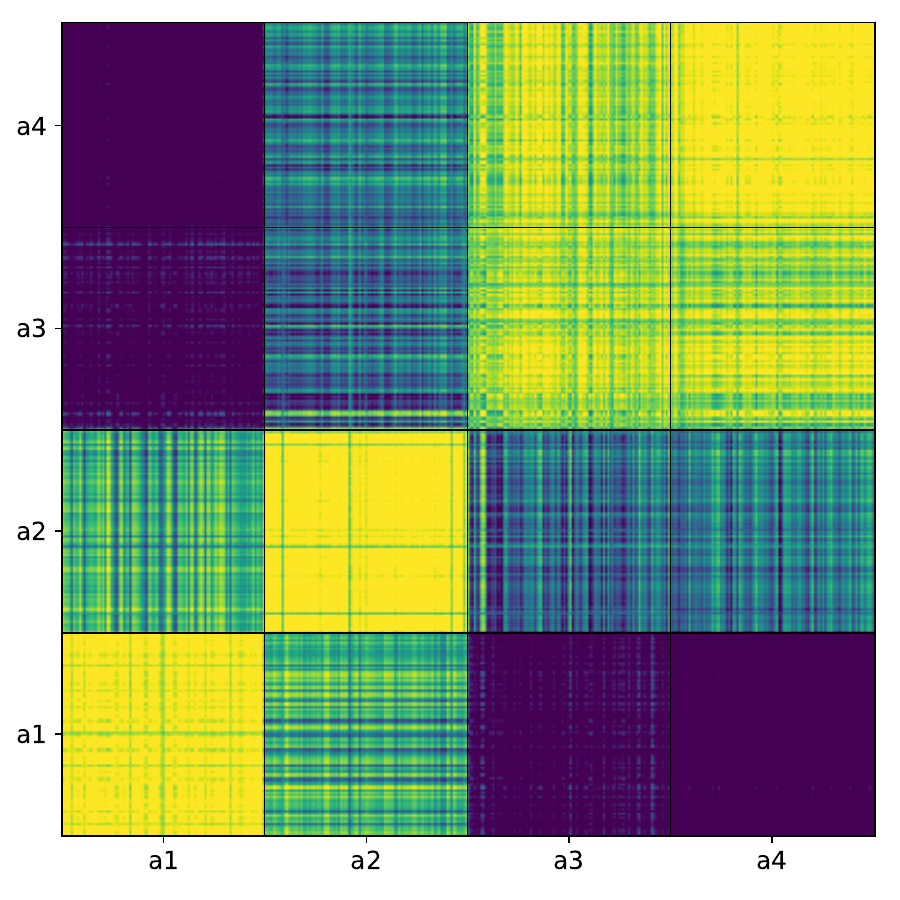}
  \caption{2to2}
\end{subfigure}\hfil 
\begin{subfigure}{0.2\textwidth}
  \includegraphics[width=\linewidth]{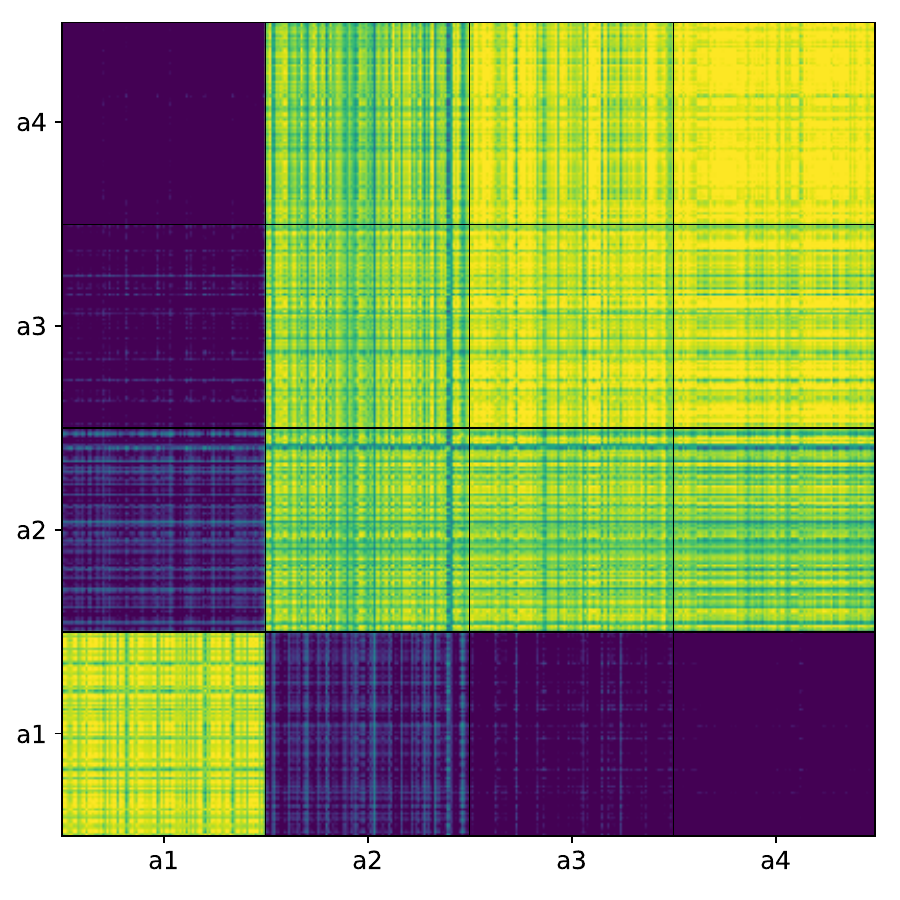}
  \caption{1to3}
\end{subfigure}\hfil 
\begin{subfigure}{0.2\textwidth}
  \includegraphics[width=\linewidth]{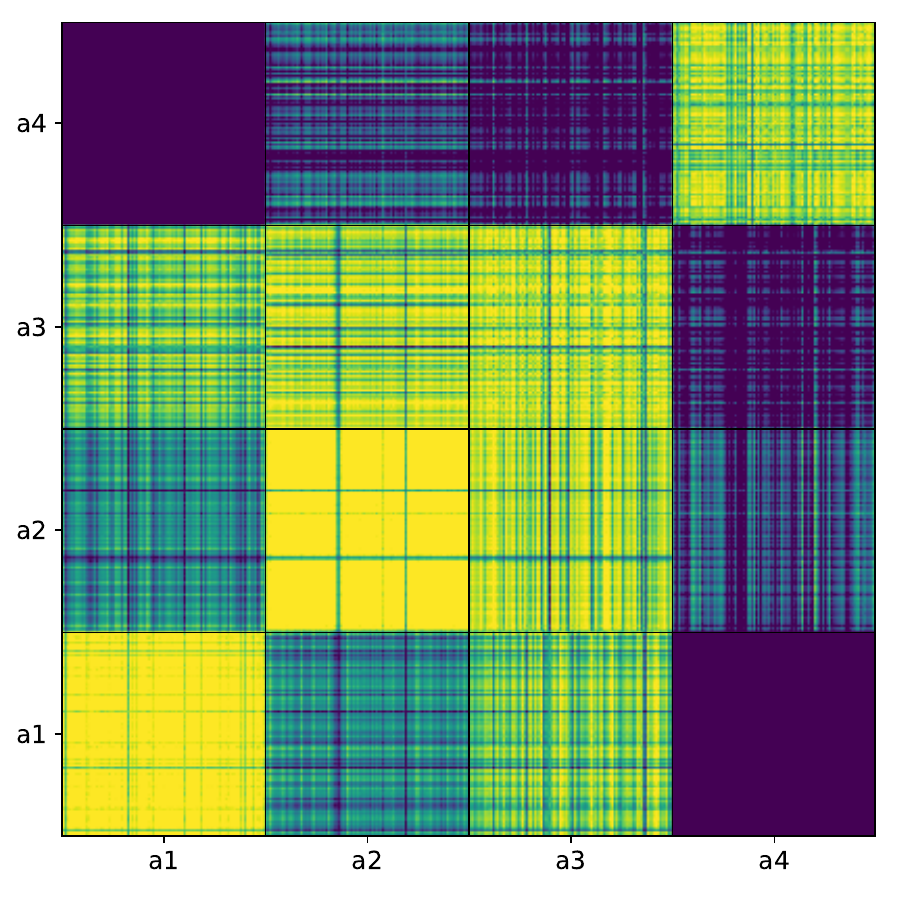}
  \caption{3to1}
\end{subfigure}

\medskip
\begin{subfigure}{0.2\textwidth}
  \includegraphics[width=\linewidth]{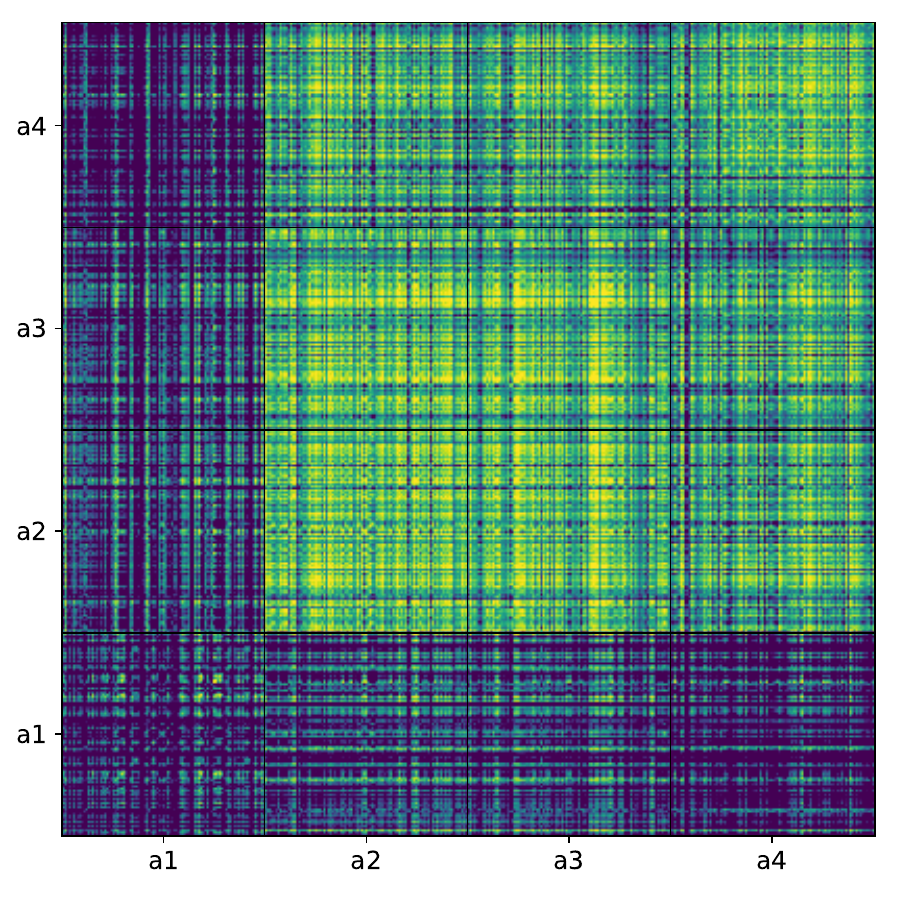}
  \caption{2to2 (Rand.)}
\end{subfigure}\hfil 
\begin{subfigure}{0.2\textwidth}
  \includegraphics[width=\linewidth]{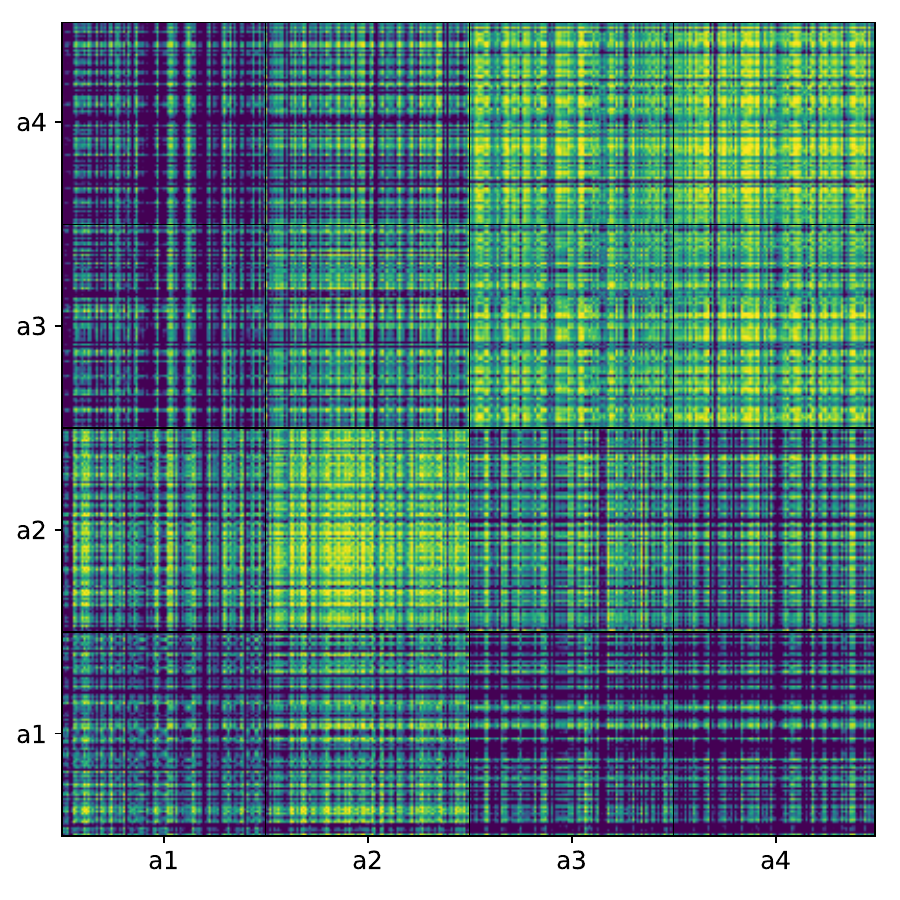}
  \caption{1to3 (Rand.)}
\end{subfigure}\hfil 
\begin{subfigure}{0.2\textwidth}
  \includegraphics[width=\linewidth]{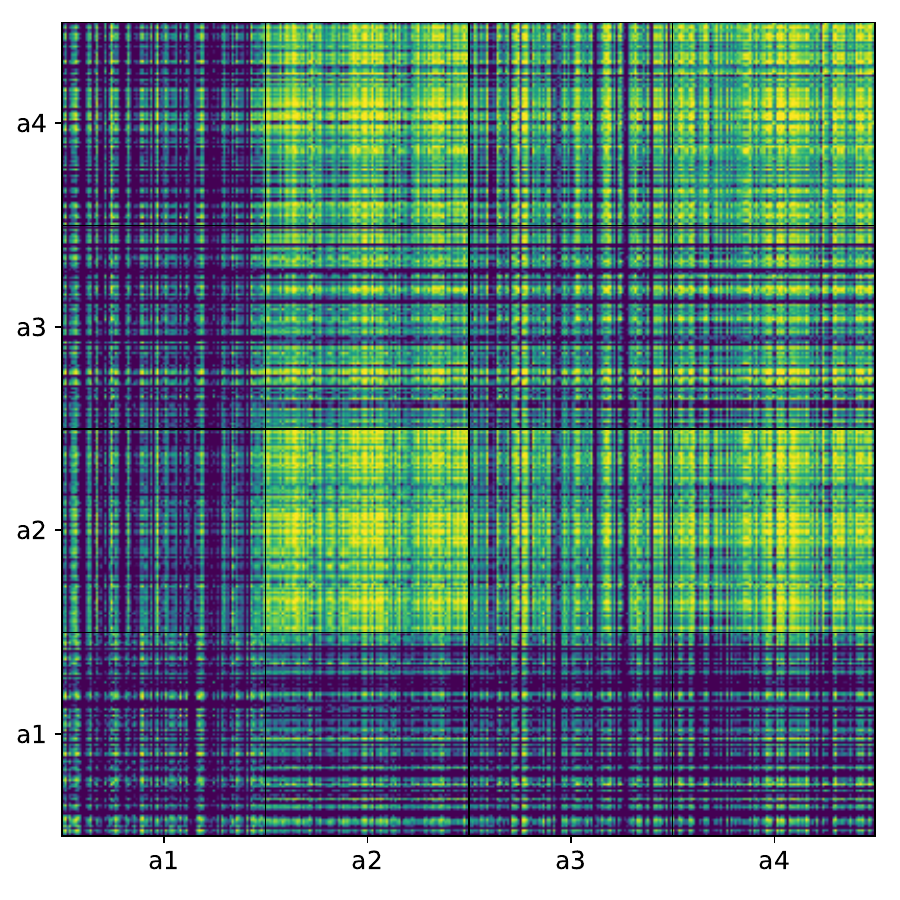}
  \caption{3to1 (Rand.)}
\end{subfigure}
\caption{Cosine similarity heatmaps of attribute representations from Qwen3-8B on $C_{country}$.}
\label{fig:semantic_pattern_country_qwen}
\end{figure}

The heatmap in Figure~\ref{fig:semantic_pattern_job_qwen}, \ref{fig:semantic_pattern_city_qwen} and \ref{fig:semantic_pattern_country_qwen} illustrates CBR subspace embedding similarity among attributes, showing that the CBR subspace in Qwen3-8B also captures semantic similarity patterns.

\subsection{CBR Subspace and Semantic Information across Contexts}
\label{sec:irs_subspace_semantic_various}
\begin{figure}[!htbp]
    \centering 
\begin{subfigure}{0.2\textwidth}
  \includegraphics[width=\linewidth]{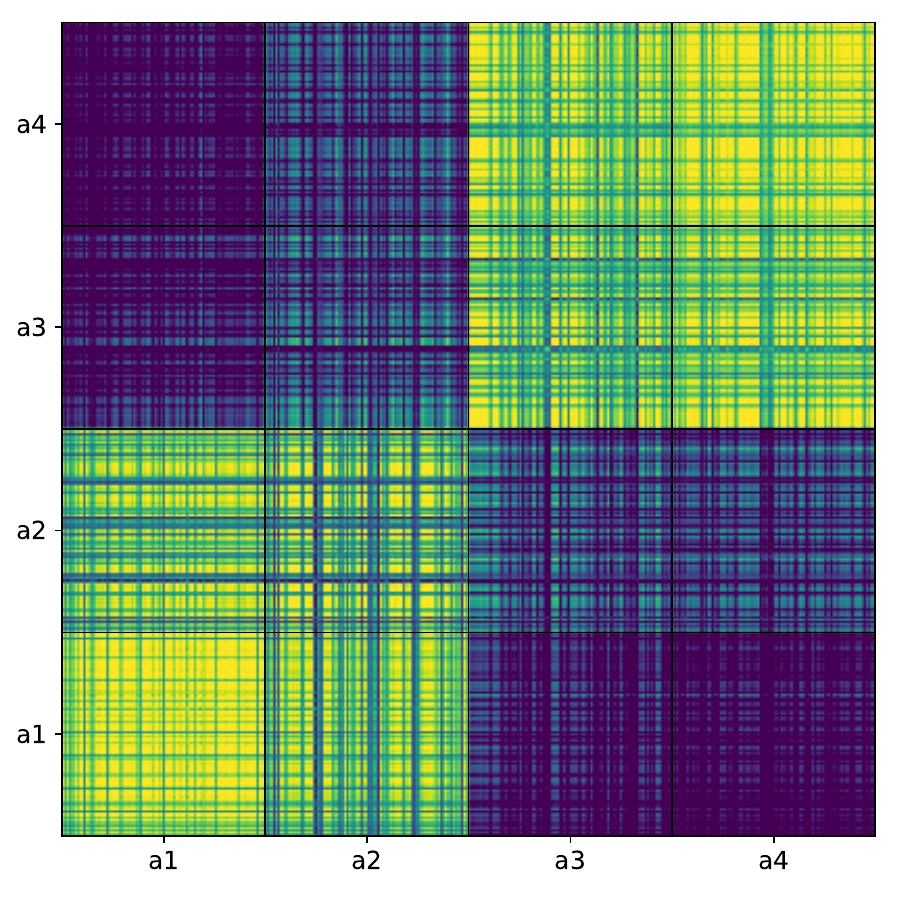}
  \caption{2to2}
\end{subfigure}\hfil 
\begin{subfigure}{0.2\textwidth}
  \includegraphics[width=\linewidth]{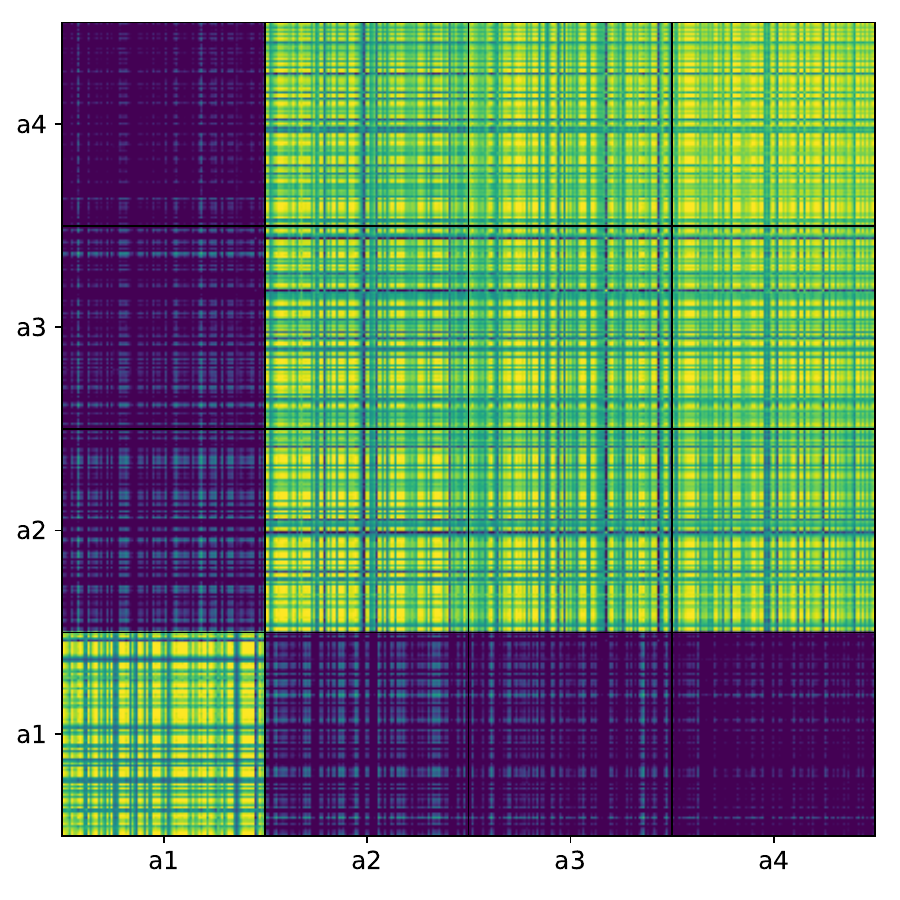}
  \caption{1to3}
\end{subfigure}\hfil 
\begin{subfigure}{0.2\textwidth}
  \includegraphics[width=\linewidth]{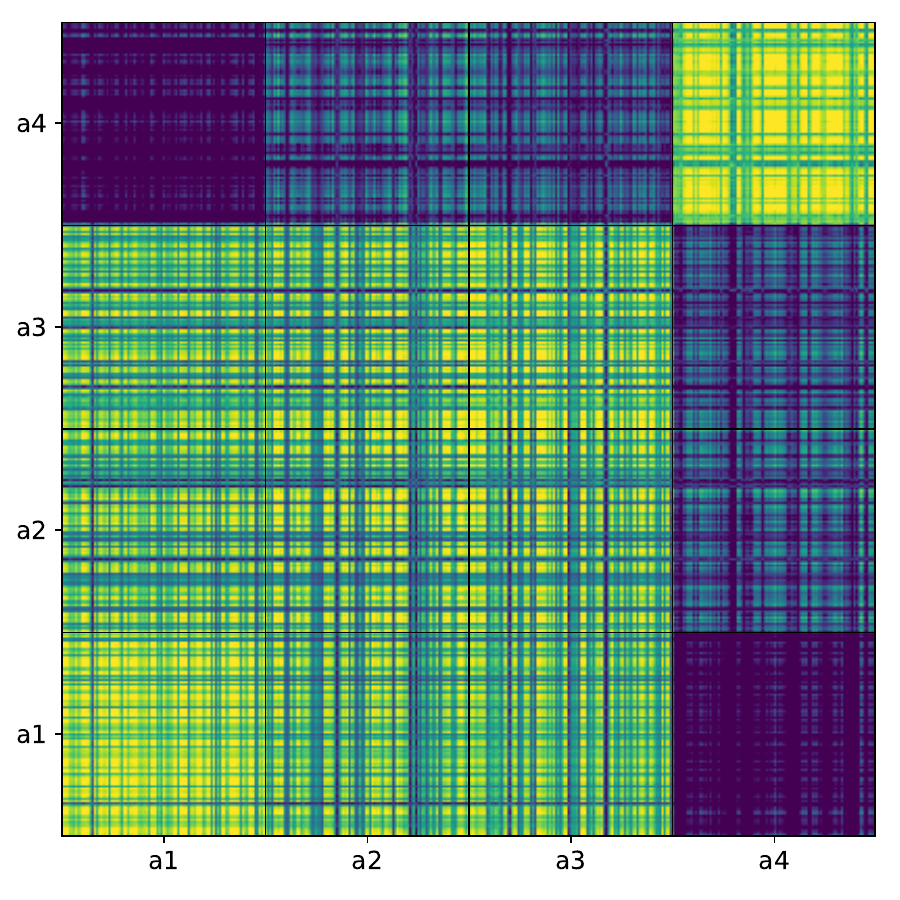}
  \caption{3to1}
\end{subfigure}

\medskip
\begin{subfigure}{0.2\textwidth}
  \includegraphics[width=\linewidth]{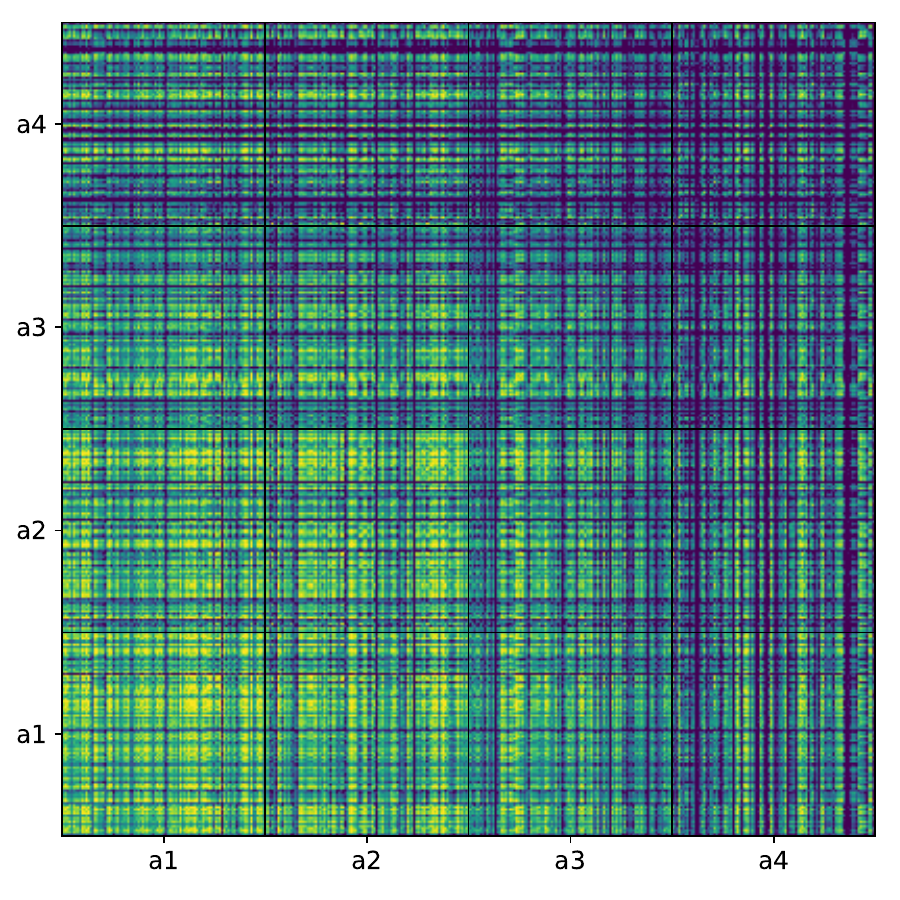}
  \caption{2to2 (Rand.)}
\end{subfigure}\hfil 
\begin{subfigure}{0.2\textwidth}
  \includegraphics[width=\linewidth]{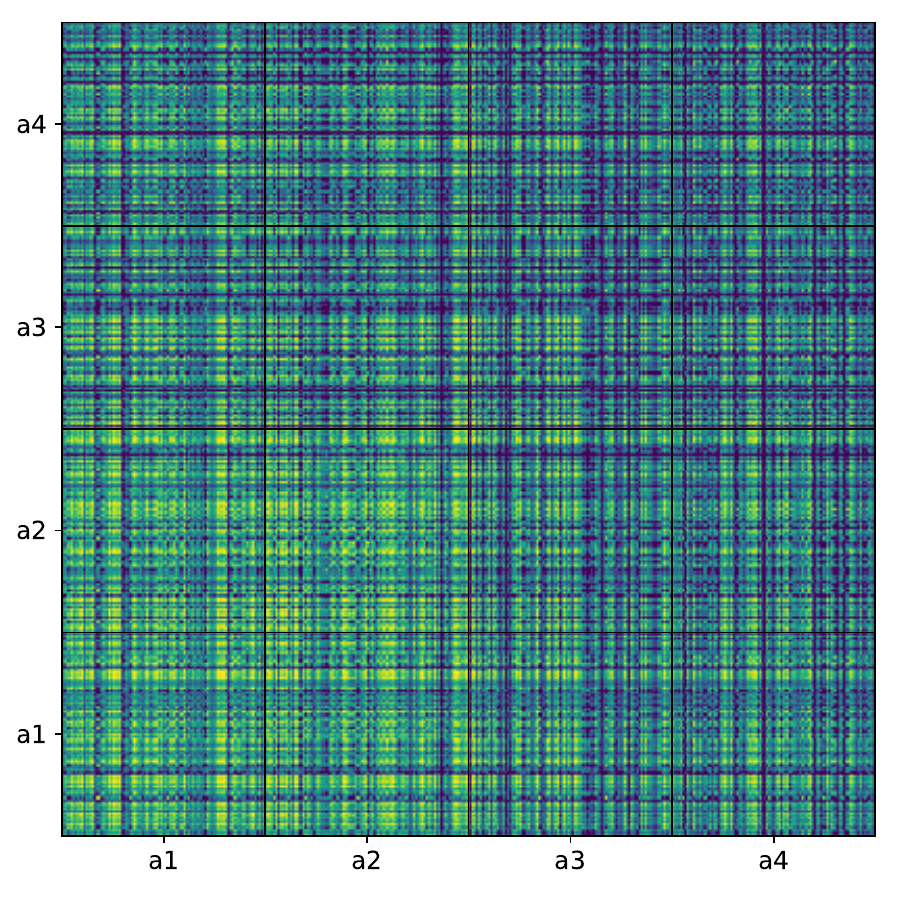}
  \caption{1to3 (Rand.)}
\end{subfigure}\hfil 
\begin{subfigure}{0.2\textwidth}
  \includegraphics[width=\linewidth]{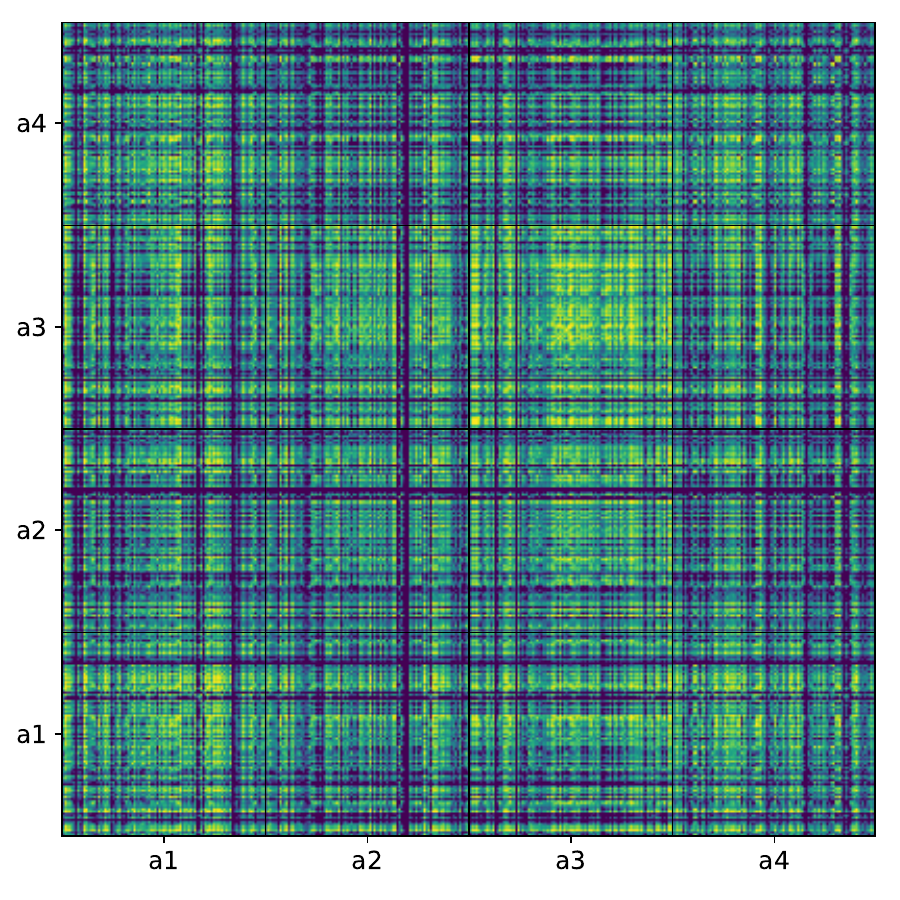}
  \caption{3to1 (Rand.)}
\end{subfigure}
\caption{Cosine similarity heatmaps of attribute representations from Llama3-8B-Instruct on $C_{city}$.}
\label{fig:semantic_pattern_city}
\end{figure}

\begin{figure}[!htbp]
    \centering 
\begin{subfigure}{0.2\textwidth}
  \includegraphics[width=\linewidth]{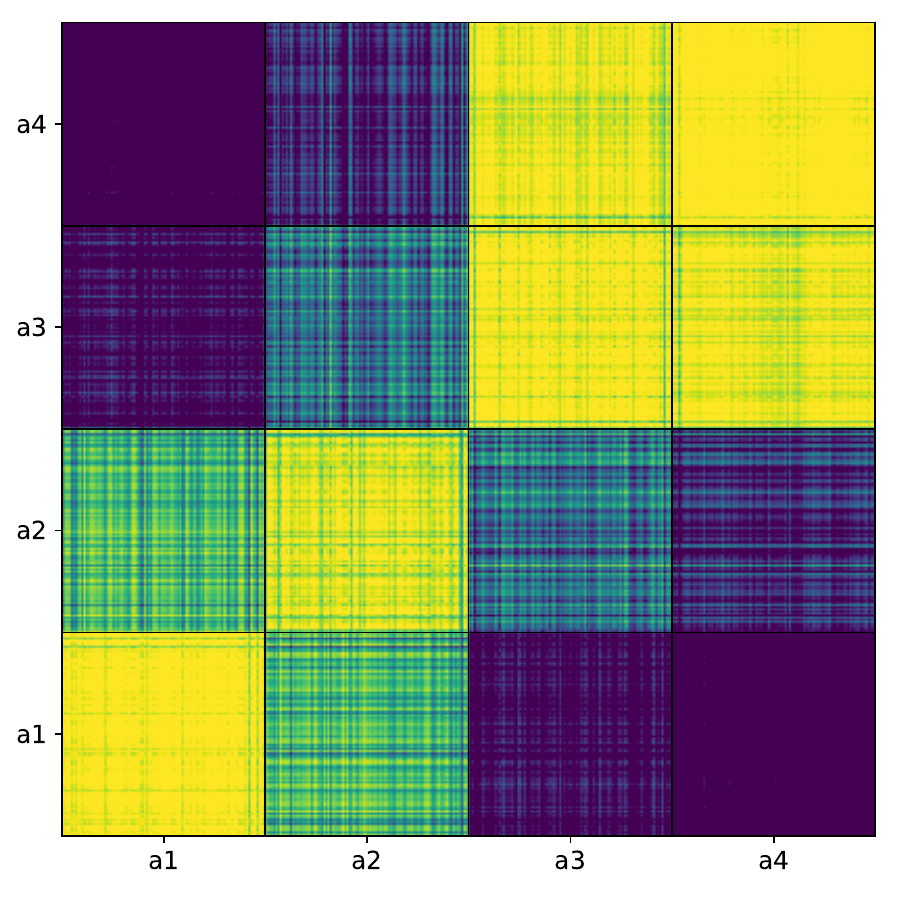}
  \caption{2to2}
\end{subfigure}\hfil 
\begin{subfigure}{0.2\textwidth}
  \includegraphics[width=\linewidth]{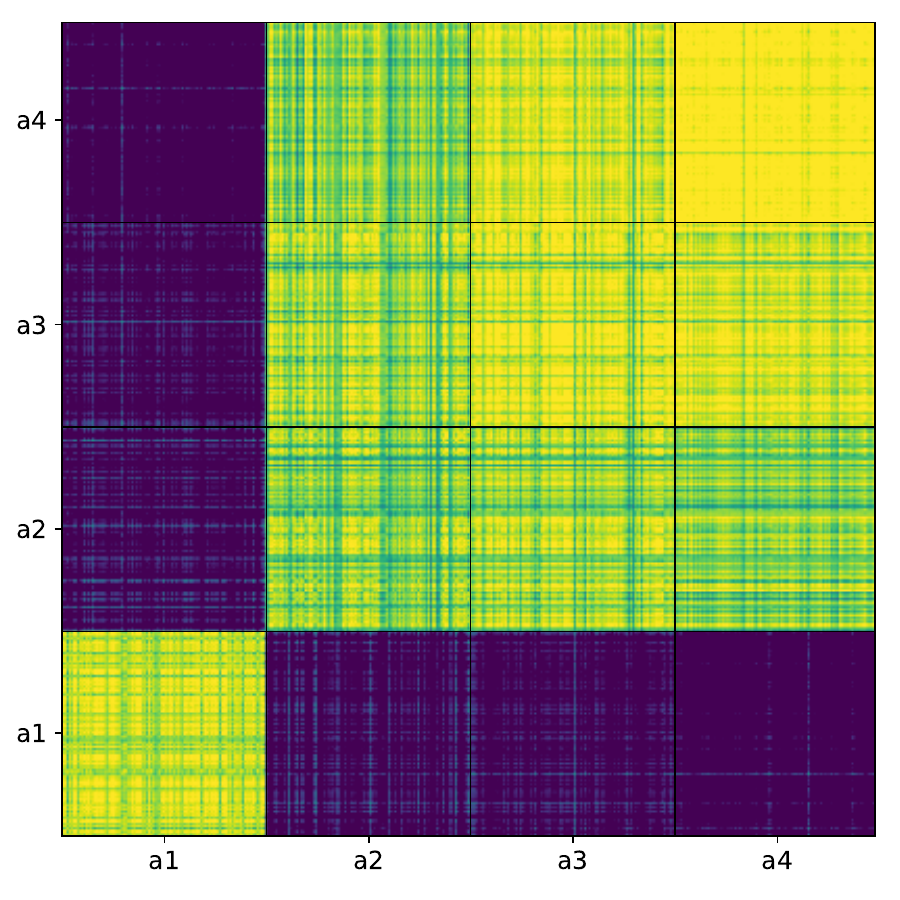}
  \caption{1to3}
\end{subfigure}\hfil 
\begin{subfigure}{0.2\textwidth}
  \includegraphics[width=\linewidth]{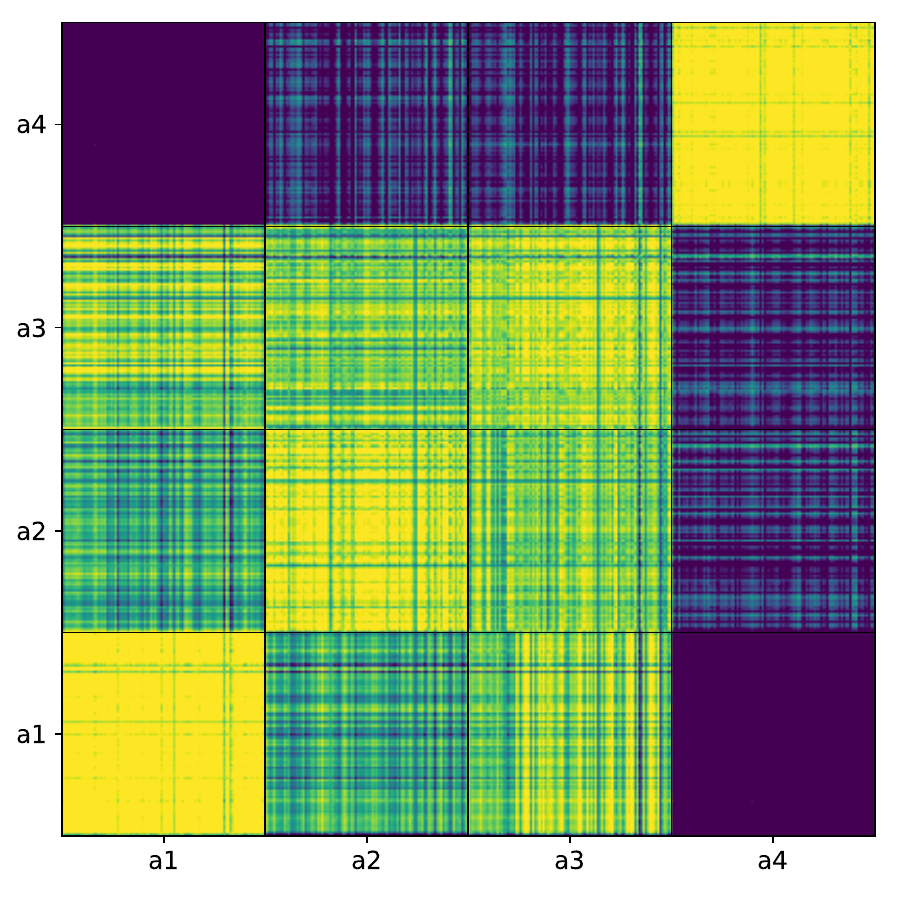}
  \caption{3to1}
\end{subfigure}

\medskip
\begin{subfigure}{0.2\textwidth}
  \includegraphics[width=\linewidth]{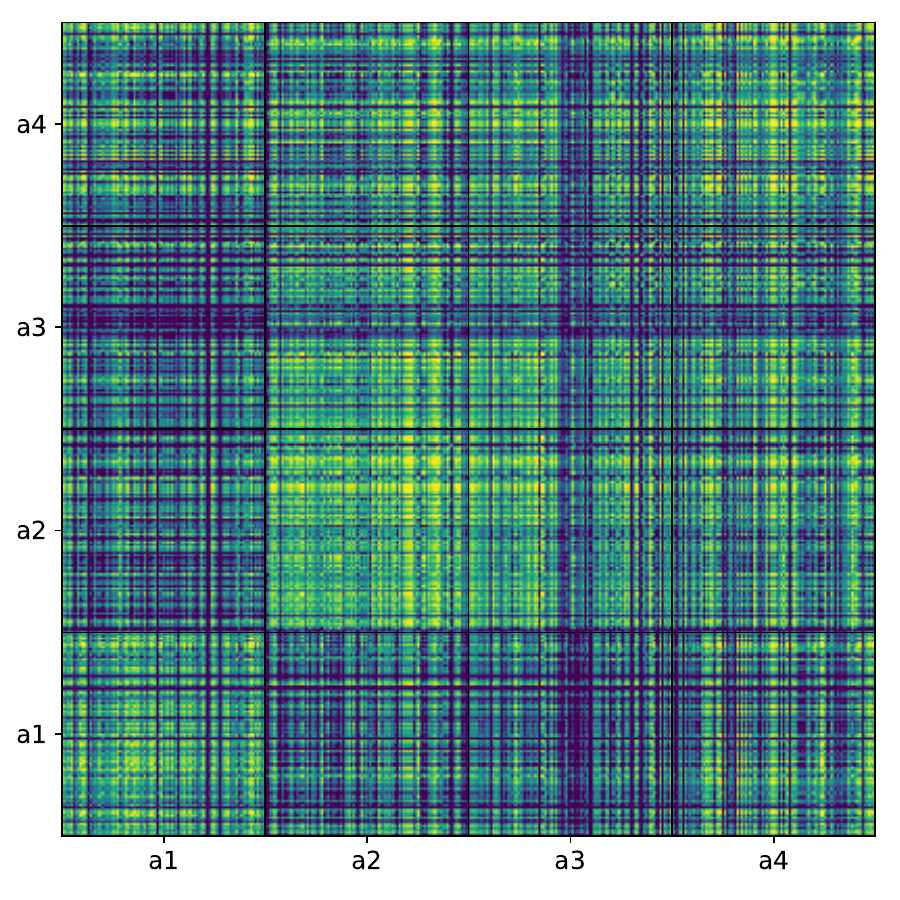}
  \caption{2to2 (Rand.)}
\end{subfigure}\hfil 
\begin{subfigure}{0.2\textwidth}
  \includegraphics[width=\linewidth]{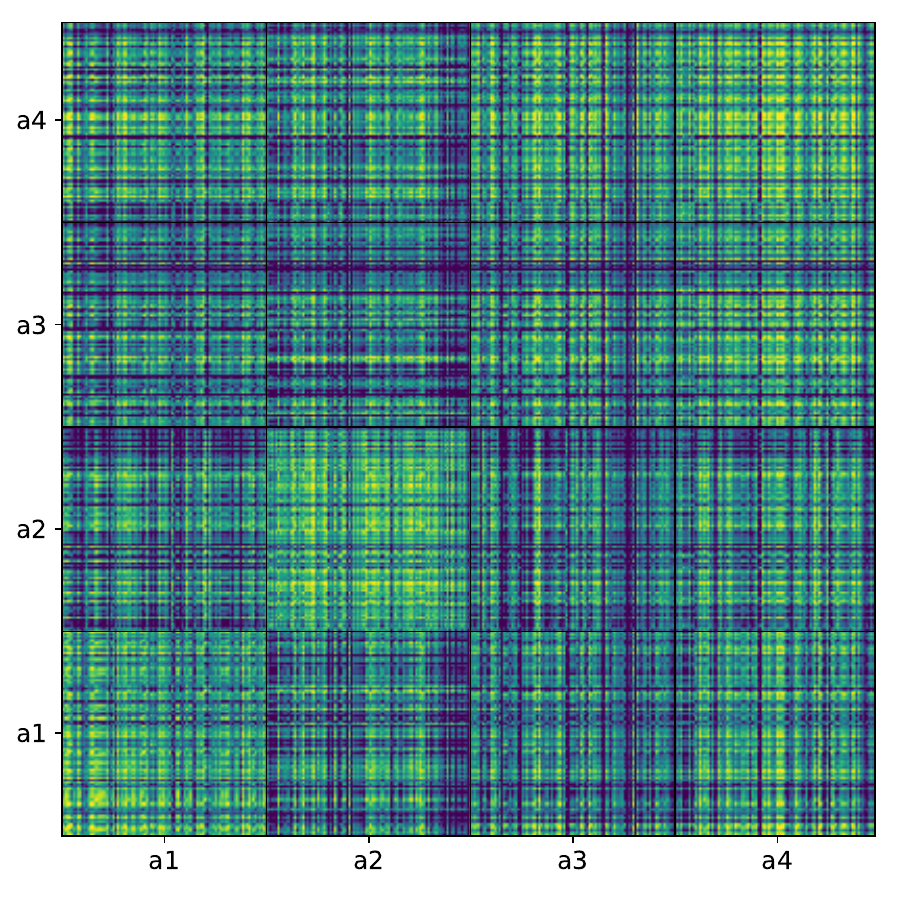}
  \caption{1to3 (Rand.)}
\end{subfigure}\hfil 
\begin{subfigure}{0.2\textwidth}
  \includegraphics[width=\linewidth]{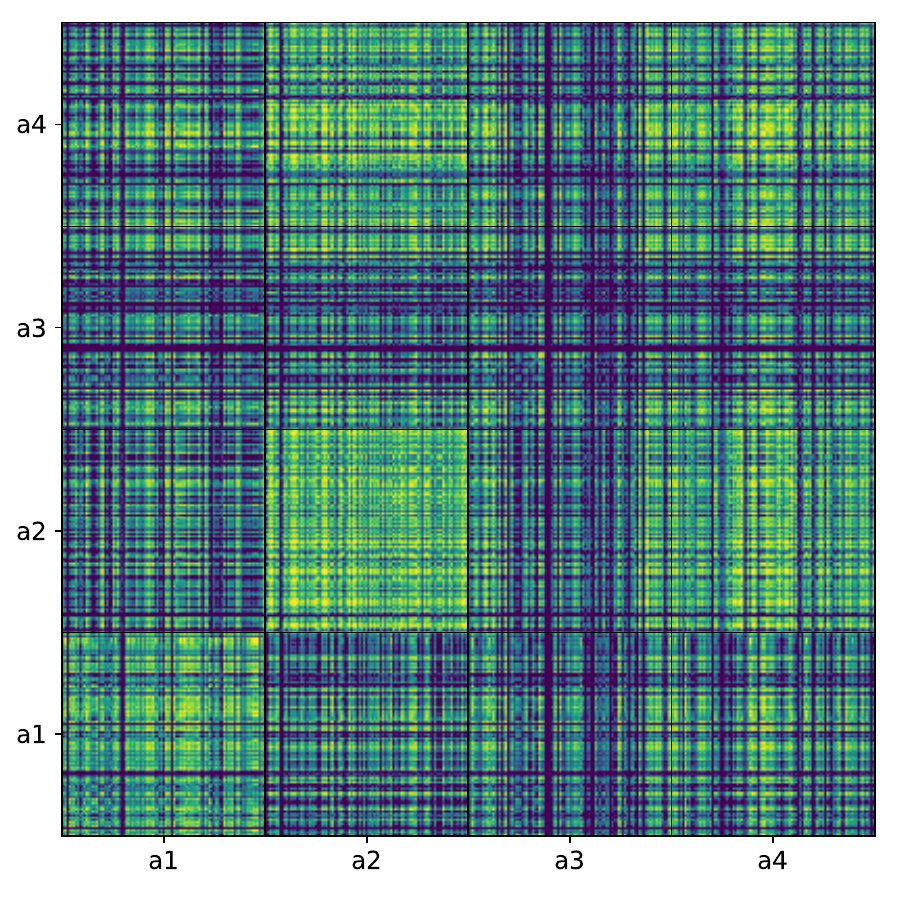}
  \caption{3to1 (Rand.)}
\end{subfigure}
\caption{Cosine similarity heatmaps of attribute representations from Llama3-8B-Instruct on $C_{country}$.}
\label{fig:semantic_pattern_country}
\end{figure}

The heatmap in Figure~\ref{fig:semantic_pattern_city} and \ref{fig:semantic_pattern_country} illustrates CBR subspace embedding similarity among attributes in $C_{city}$ and $C_{country}$ respectively. The consistency of these pattern across different contexts indicates that the CBR subspace in Llama3-8B-Instruct can capture semantic similarity pattern.

\subsection{Generality of CBR Subspace on Qwen3-8B}
\label{sec:generality_qwen}
\begin{figure*}[htb]
    \centering 
\begin{subfigure}{0.485\textwidth}
  \includegraphics[width=\linewidth]{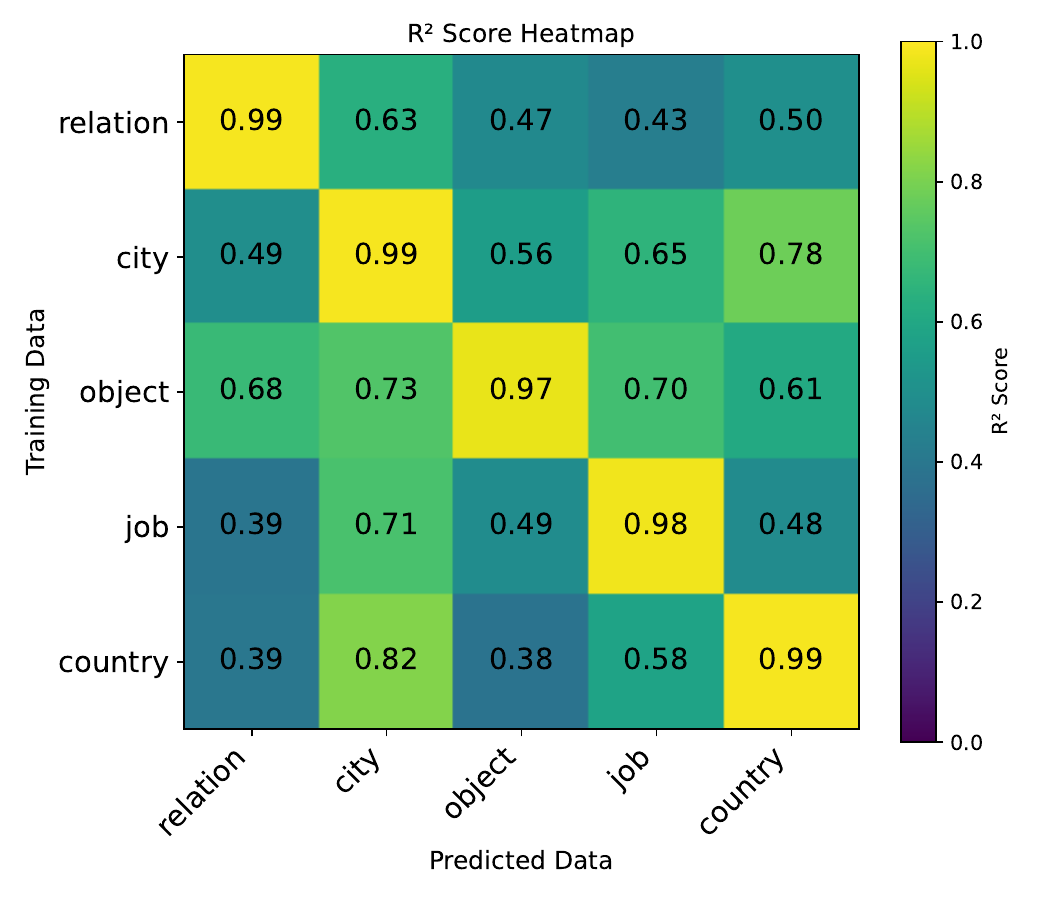}
  \caption{w/o $\Delta$}
\end{subfigure}\hfil 
\begin{subfigure}{0.485\textwidth}
  \includegraphics[width=\linewidth]{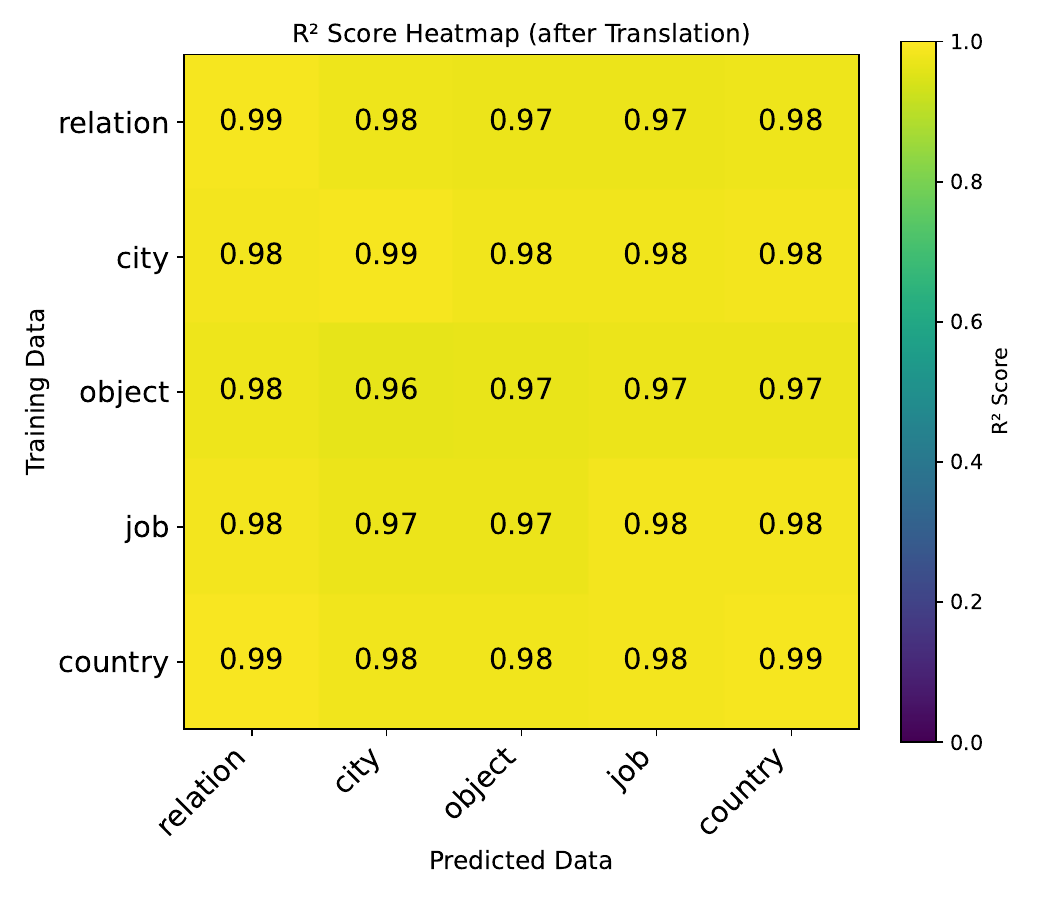}
  \caption{w/ $\Delta$}
\end{subfigure}\hfil 
\caption{Each cell shows the $R^2$ fitness score obtained from Qwen3-8B. The projection matrix learned from one context (column) is used to predict the index information of another context (row). Higher values indicate better cross-context generality.}
\label{fig:generality_qwen}
\end{figure*}

As shown in Figure~\ref{fig:generality_qwen}, Qwen3-8B exhibits a similar tendency to Llama3-8B-Instruct, where the projection matrix is influenced by contextual variations. However, by applying the translation vectors, the projection matrix trained on one context can correctly predict the CBR indices in another context.

\subsection{Ablation Study on Translation Vectors}
\label{sec:transv_ab}
\begin{figure*}[t]
\centering
\includegraphics[width=16.0cm]{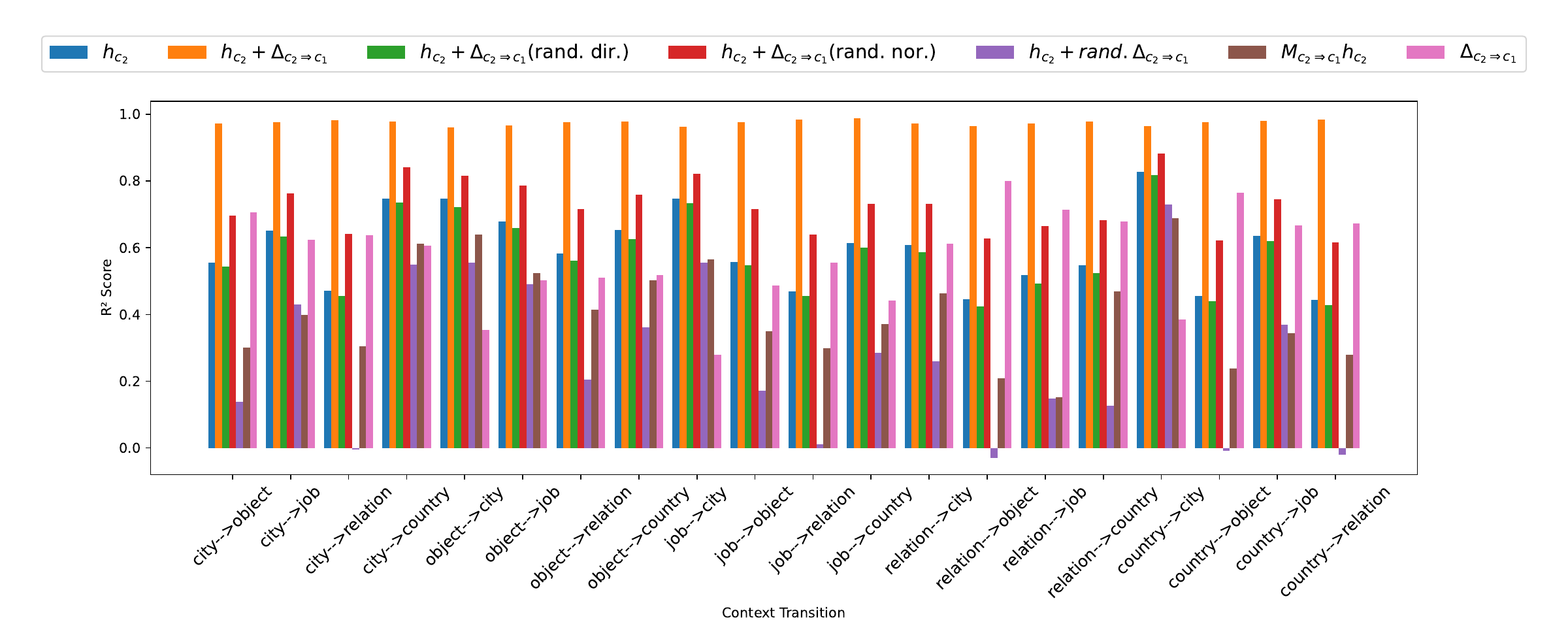}
\caption{Performance comparison of the translation vector $\Delta_{c_1 \rightarrow c_2}$ and baseline methods for cross context CBR decoding from Llama3-8B-Instruct. The translation vector outperforms alternatives including random vectors (denoted as $rand\Delta_{c_1 \rightarrow c_2}$), randomized direction ($\Delta_{c_1 \rightarrow c_2}(rand. dir.)$) and norm ($\Delta_{c_1 \rightarrow c_2}(rand. nor.)$), a learned transformation matrix ($M_{c_1 \rightarrow c_2}$) and the $\Delta_{c_1 \rightarrow c_2}$ alone).}
\label{fig:vari_trans_llama}
\end{figure*}

\begin{figure*}[t]
\centering
\includegraphics[width=16.0cm]{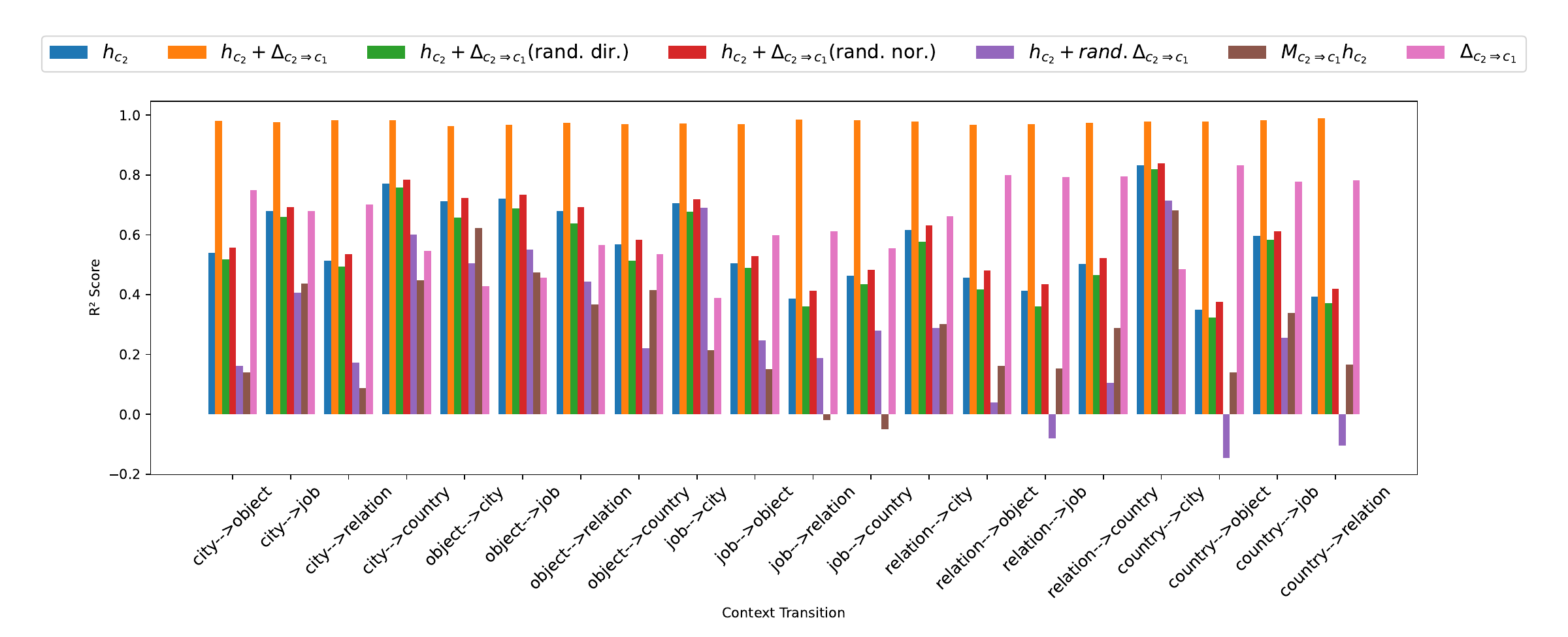}
\caption{Performance comparison from Qwen3-8B.}
\label{fig:vari_trans_qwen}
\end{figure*}

To further verify the effectiveness of the translation vector, we evaluate several variants that modify the translation vector, including randomizing its norm or direction. We also compare it with baseline methods, such as using a random vector or a transformation matrix $M_{c_1 \rightarrow c_2}$ learned by minimizing the mean squared error between $h_{c_1}$ and $M_{c_1 \rightarrow c_2}h_{c_2}$. The results are shown in Figure~\ref{fig:vari_trans_llama} and Figure~\ref{fig:vari_trans_qwen}. We observe that these alternative methods perform worse than the translation vector, further demonstrating its effectiveness. 

One possible explanation of the effectiveness is that the translation vector partially contains CBR index information. However, this explanation is only partially supported by the results. As shown in Figures~\ref{fig:vari_trans_llama} and \ref{fig:vari_trans_qwen}, in some context transformations, such as ``object-->city'' and ``job-->city'', the performance of using only the translation vector $\Delta_{c_2 \rightarrow c_1}$ is worse than directly using $h_{c_2}$. 

The fact that $\Delta_{c_2 \rightarrow c_1}$ sometimes outperforms $h_{c_2}$ suggests that the translation vector may contain index-related information, which could be explained by superposition theory \citep{elhage2022toy}. The theory suggests that different features may be encoded in overlapping neuron directions. Under this view, the translation vector may capture a mixture of contextual and index-related components in the shared activation space.

\subsection{Translation Vector under Superposition}
\label{sec:transv_theory}
To explain the effectiveness of translation vectors, we interpret the CBR representation through the lens of superposition theory \citep{elhage2022toy}. In this framework, a hidden activation $h$ can be expressed as a superposition of feature directions:
\[
h = \sum_i f_i v_i,
\]
where $f_i$ denotes the activation strength of feature $i$, and $v_i$ represents the direction (or subspace) encoding that feature. Importantly, multiple features may share neurons through superposition.

Assume that a hidden state in context $c_1$ contains both a CBR index feature and context-specific features:

\begin{equation}
h_{c_1} = f_{\text{index}} v_{\text{index}} + f_{c_1} v_c
\end{equation}
where $v_{\text{index}}$ denotes the direction encoding the CBR index and $v_c$ represents the direction corresponding to contextual features. These directions are not necessarily orthogonal. Let $M_{c_1}$ denote the projection matrix trained under context $c_1$ to decode the CBR indices. Applying this projection yields:

\begin{equation}
M_{c_1} h_{c_1}
=
M_{c_1} f_{\text{index}} v_{\text{index}}
+
M_{c_1} f_{c_1} v_c
\approx \text{indices}_{\text{CBR}} .
\label{eq:trans_v_c1}
\end{equation}
Since $v_{\text{index}}$ and $v_c$ are generally not orthogonal, the contextual component satisfies $M_{c_1} f_{c_1} v_c \neq 0$. 
However, if the activation comes from another context $c_2$,

\begin{equation}
h_{c_2} =
f_{\text{index}} v_{\text{index}} + f_{c_2} v_c ,
\end{equation}
then applying the same projection gives:

\begin{equation}
M_{c_1} h_{c_2}
=
M_{c_1} f_{\text{index}} v_{\text{index}}
+
M_{c_1} f_{c_2} v_c .
\end{equation}
Because $f_{c_1} v_c \neq f_{c_2} v_c$, the contextual component changes. Compared to Equation~\ref{eq:trans_v_c1}, this leads to

\[
M_{c_1} h_{c_2} \not\approx \text{indices}_{\text{CBR}} .
\]
Nevertheless, if contexts $c_1$ and $c_2$ are similar, $M_{c_1}$ tends to remain effective for context $c_2$. This is supported by the observations in Appendix~\secref{sec:docred}. Alternatively, their contextual features may be related by a translation vector $\Delta$ in the activation space:
\[
f_{c_1} v_c \approx f_{c_2} v_c + \Delta_{c_2 \rightarrow c_1}.
\]
Consequently, applying this translation before decoding yields:

\[
M_{c_1}(h_{c_2} + \Delta_{c_2 \rightarrow c_1})
=
M_{c_1}(f_{\text{index}} v_{\text{index}} +
f_{c_2} v_c + \Delta_{c_2 \rightarrow c_1})
\approx
M_{c_1}(f_{\text{index}} v_{\text{index}} +
f_{c_1} v_c)
\approx
\text{indices}_{\text{CBR}} .
\]
This may explain why translation vectors can restore decoding performance across contexts: they compensate for context-dependent components in the superposed representation while preserving the shared CBR index direction.

\clearpage

\subsection{Consistency of CBR Subspace across Contexts}
\label{sec:consistency_of_irs_subspace_across_contexts}
\begin{figure*}[htb]
    \centering 
\begin{subfigure}{0.325\textwidth}
  \includegraphics[width=\linewidth]{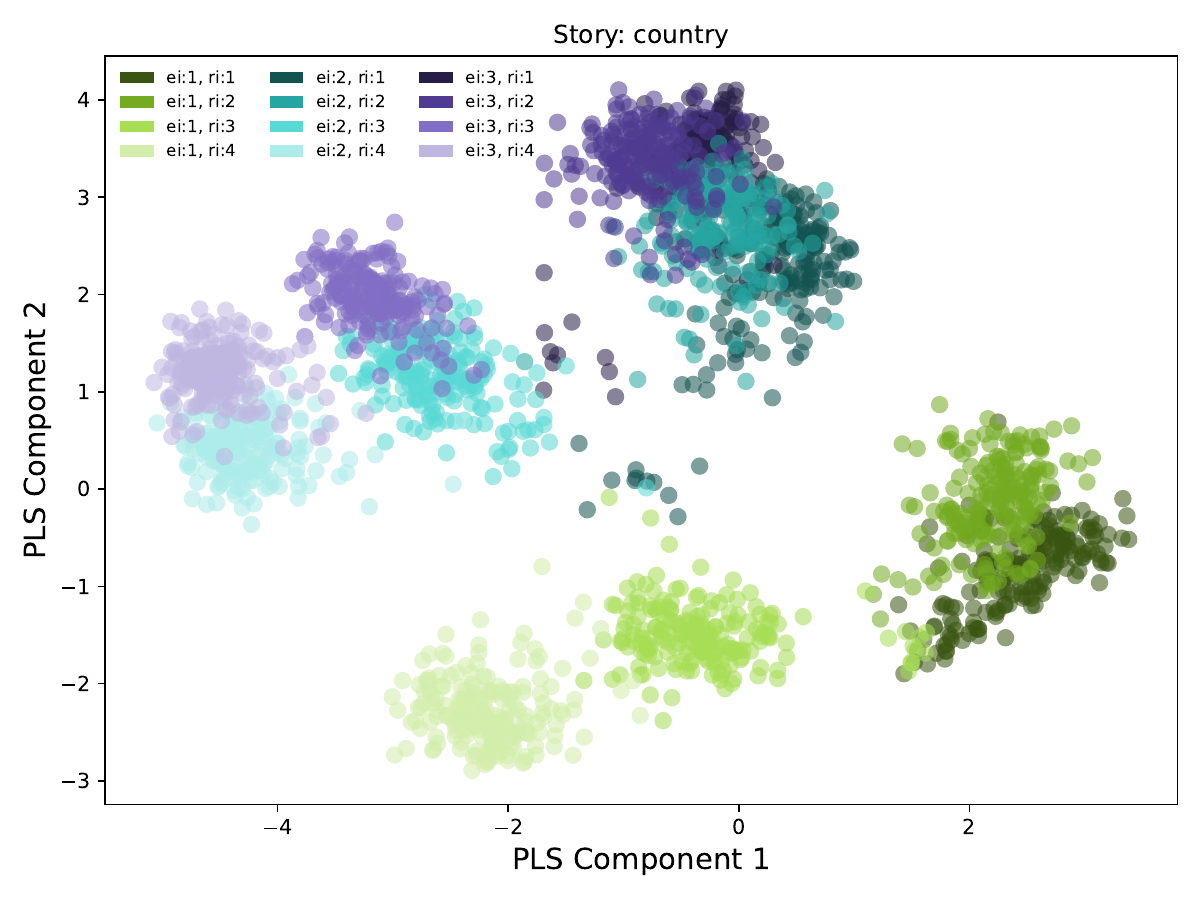}
  \caption{base}
\end{subfigure}\hfil 
\begin{subfigure}{0.325\textwidth}
  \includegraphics[width=\linewidth]{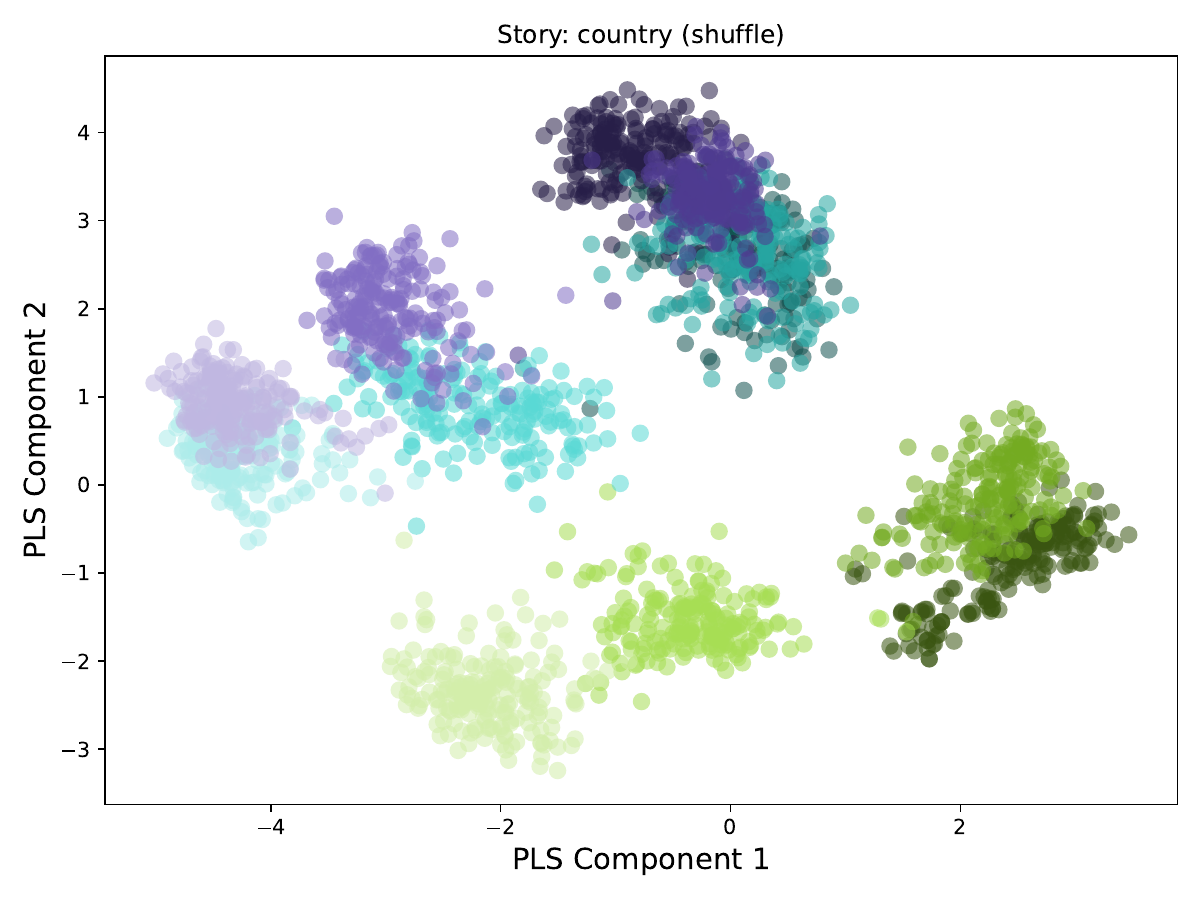}
  \caption{shuffled}
\end{subfigure}\hfil 
\caption{Visualization of the CBR subspace before and after \textbf{shuffling} the relation order of the second and third entities from Llama3-8B-Instruct on $C_{country}$.}
\label{fig:consistency_shuffle_country_llama}
\end{figure*}

\begin{figure*}[htb]
    \centering 
\begin{subfigure}{0.325\textwidth}
  \includegraphics[width=\linewidth]{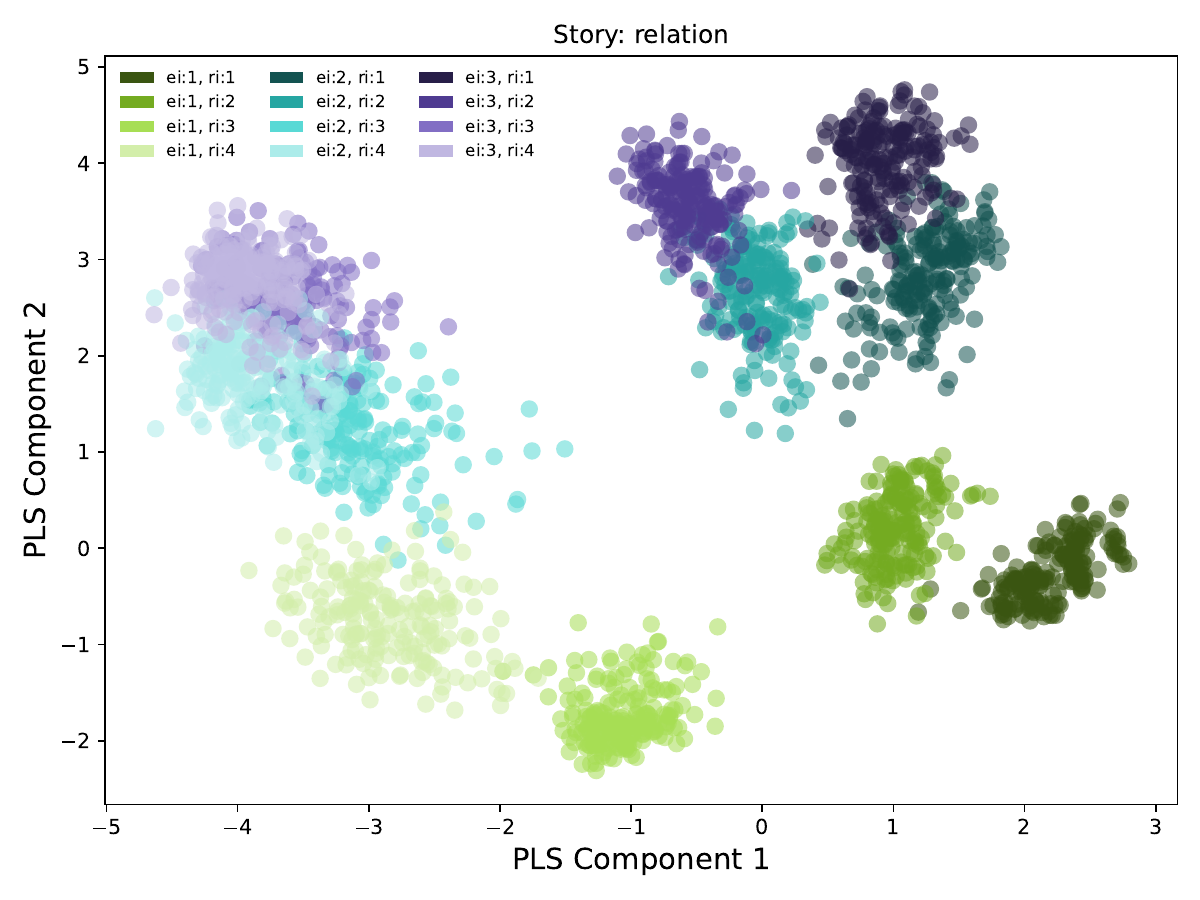}
  \caption{base}
\end{subfigure}\hfil 
\begin{subfigure}{0.325\textwidth}
  \includegraphics[width=\linewidth]{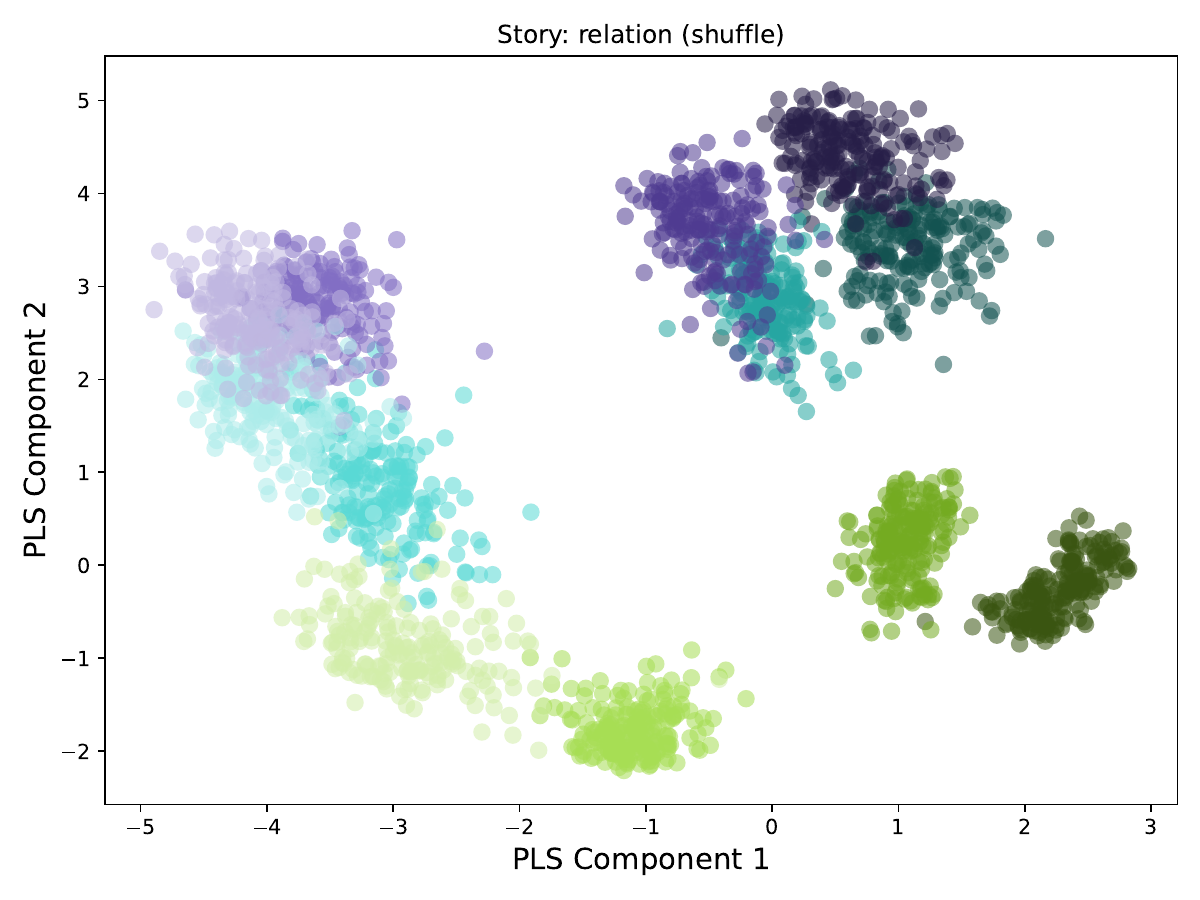}
  \caption{shuffled}
\end{subfigure}\hfil 
\caption{Visualization of the CBR subspace before and after \textbf{shuffling} the relation order of the second and third entities from Llama3-8B-Instruct on $C_{relation}$.}
\label{fig:consistency_shuffle_relation_llama}
\end{figure*}

\begin{figure*}[htb]
    \centering 
\begin{subfigure}{0.325\textwidth}
  \includegraphics[width=\linewidth]{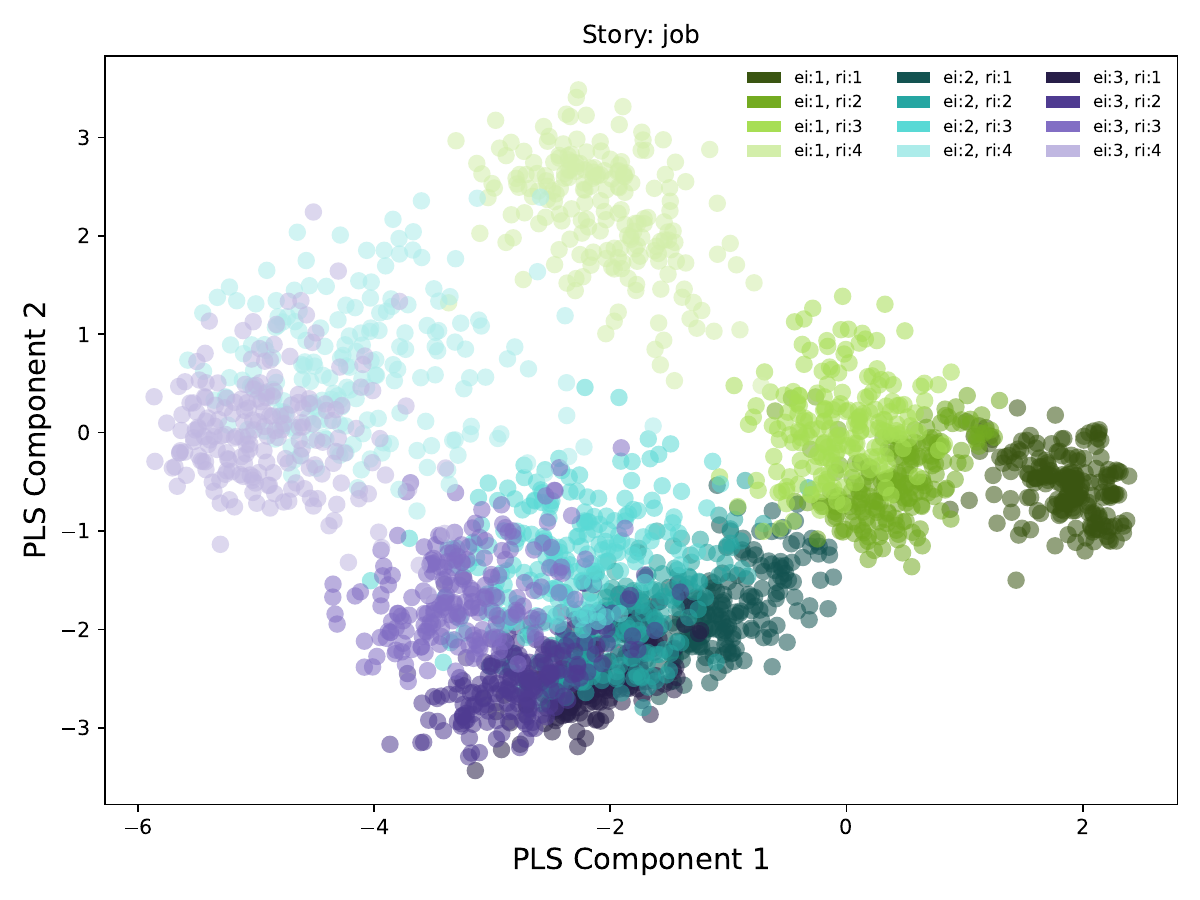}
  \caption{base}
\end{subfigure}\hfil 
\begin{subfigure}{0.325\textwidth}
  \includegraphics[width=\linewidth]{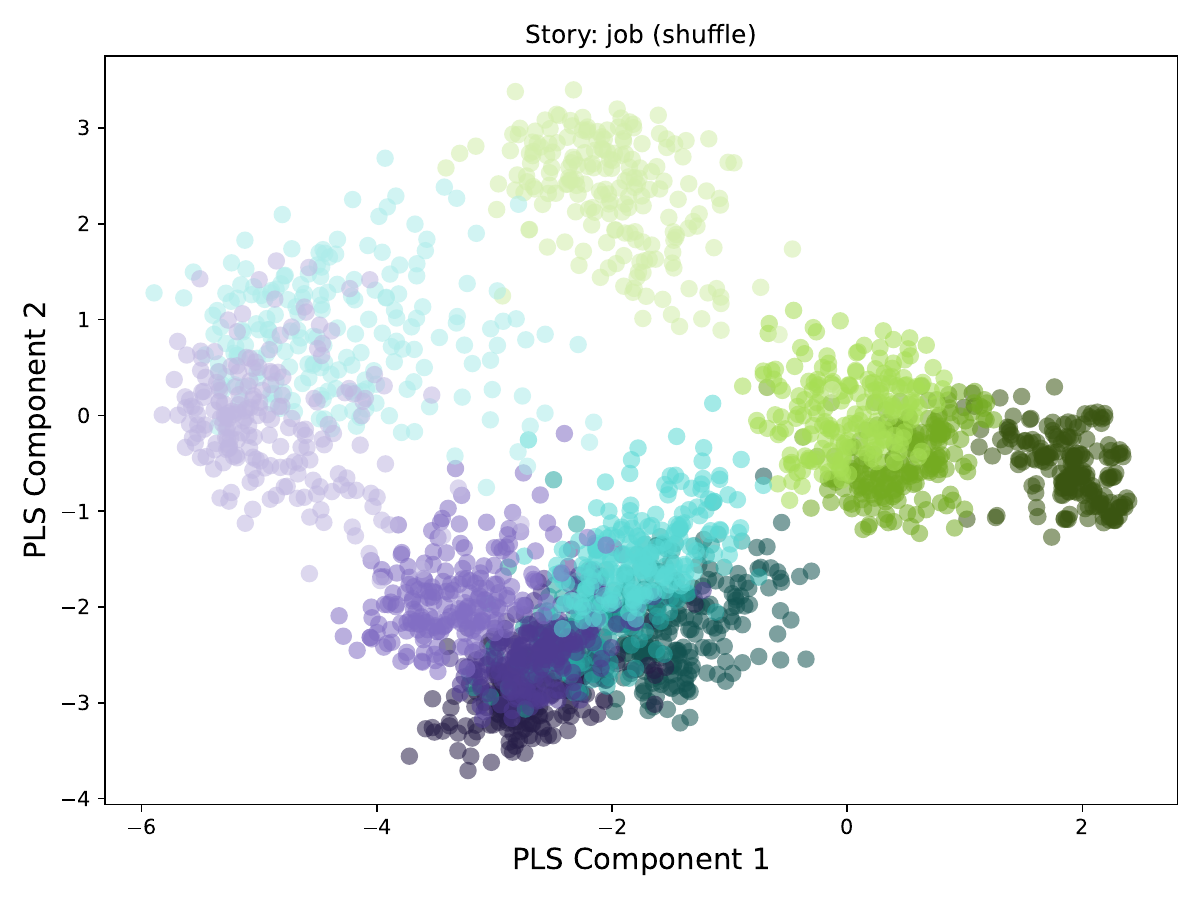}
  \caption{shuffled}
\end{subfigure}\hfil 
\caption{Visualization of the CBR subspace before and after \textbf{shuffling} from Llama3-8B-Instruct on $C_{job}$.}
\label{fig:consistency_shuffle_job_llama}
\end{figure*}

\begin{figure*}[htb]
    \centering 
\begin{subfigure}{0.325\textwidth}
  \includegraphics[width=\linewidth]{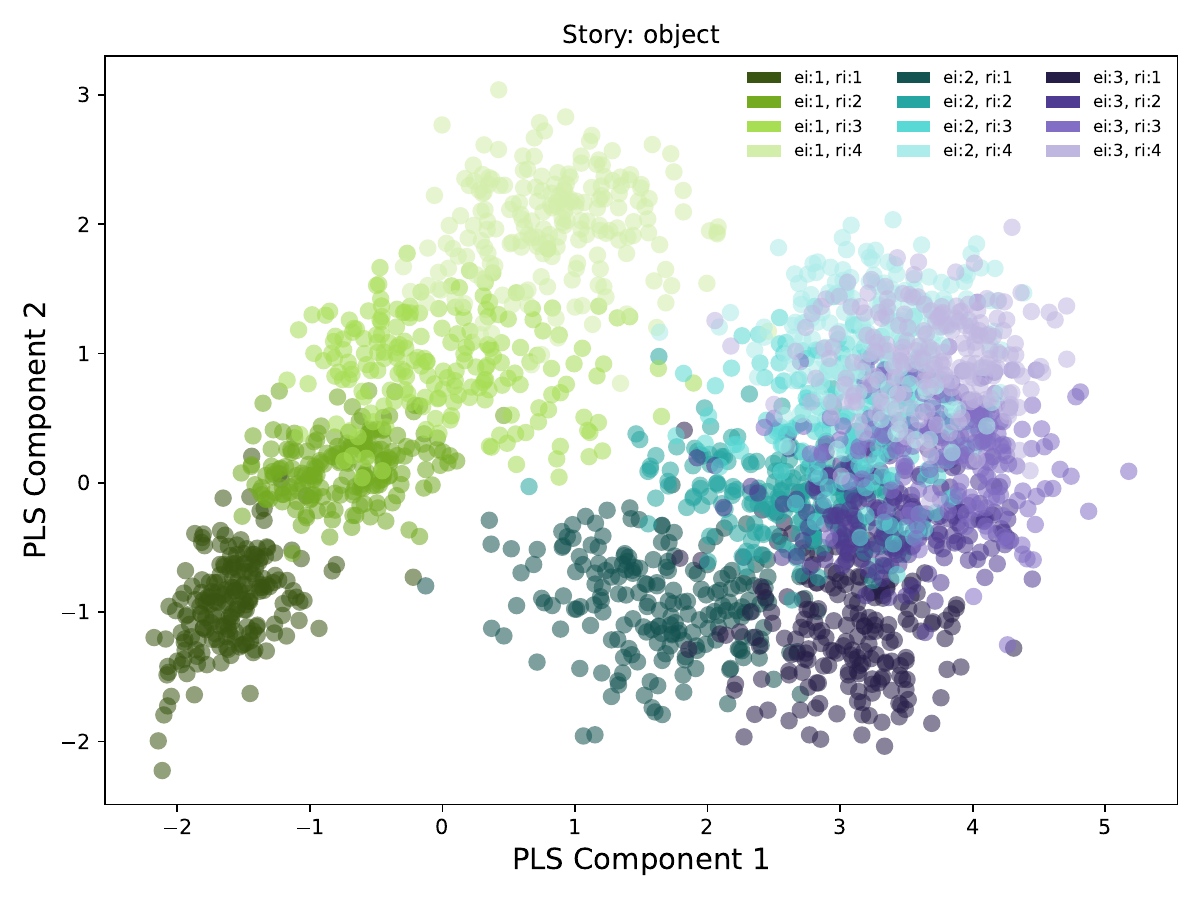}
  \caption{base}
\end{subfigure}\hfil 
\begin{subfigure}{0.325\textwidth}
  \includegraphics[width=\linewidth]{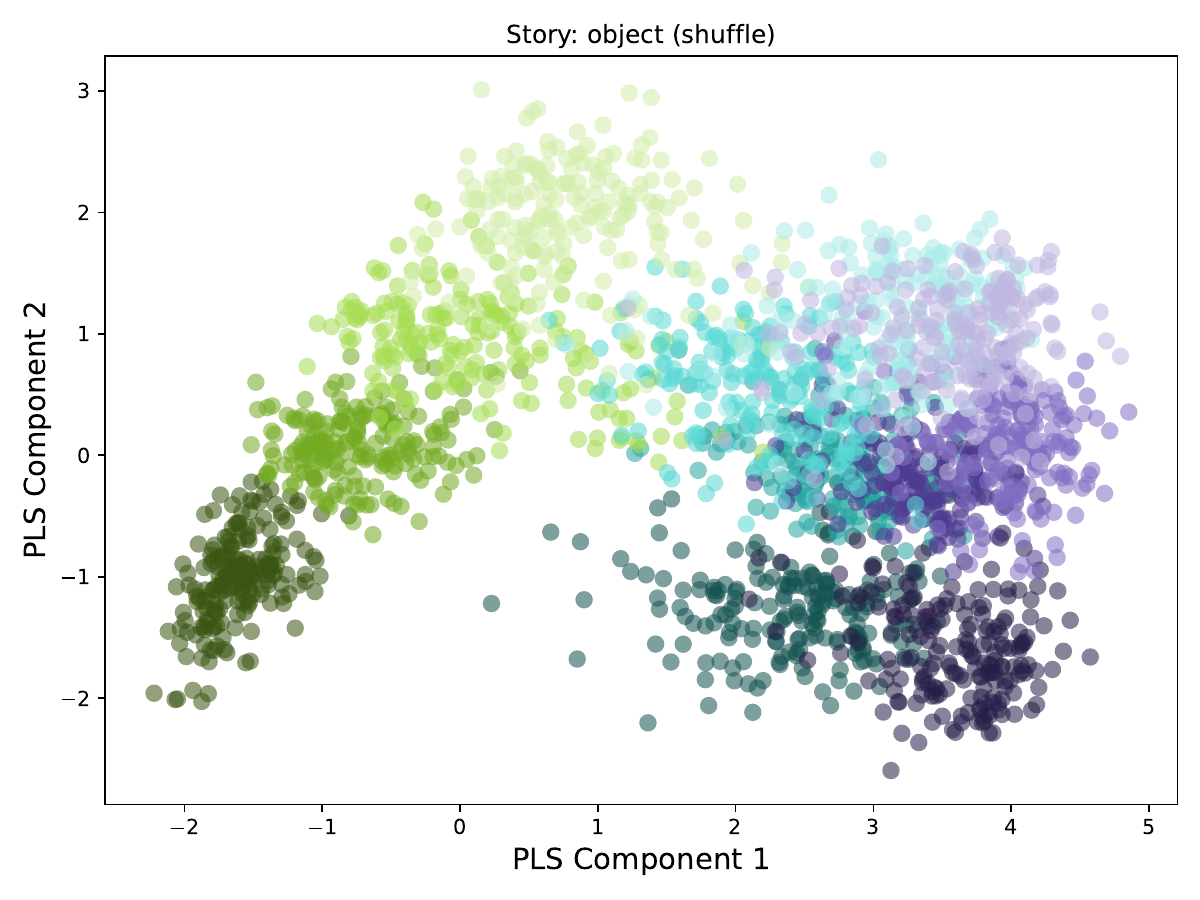}
  \caption{shuffled}
\end{subfigure}\hfil 
\caption{Visualization of the CBR subspace before and after \textbf{shuffling} from Llama3-8B-Instruct on $C_{object}$.}
\label{fig:consistency_shuffle_object_llama}
\end{figure*}

\begin{figure*}[htb]
    \centering 
\begin{subfigure}{0.325\textwidth}
  \includegraphics[width=\linewidth]{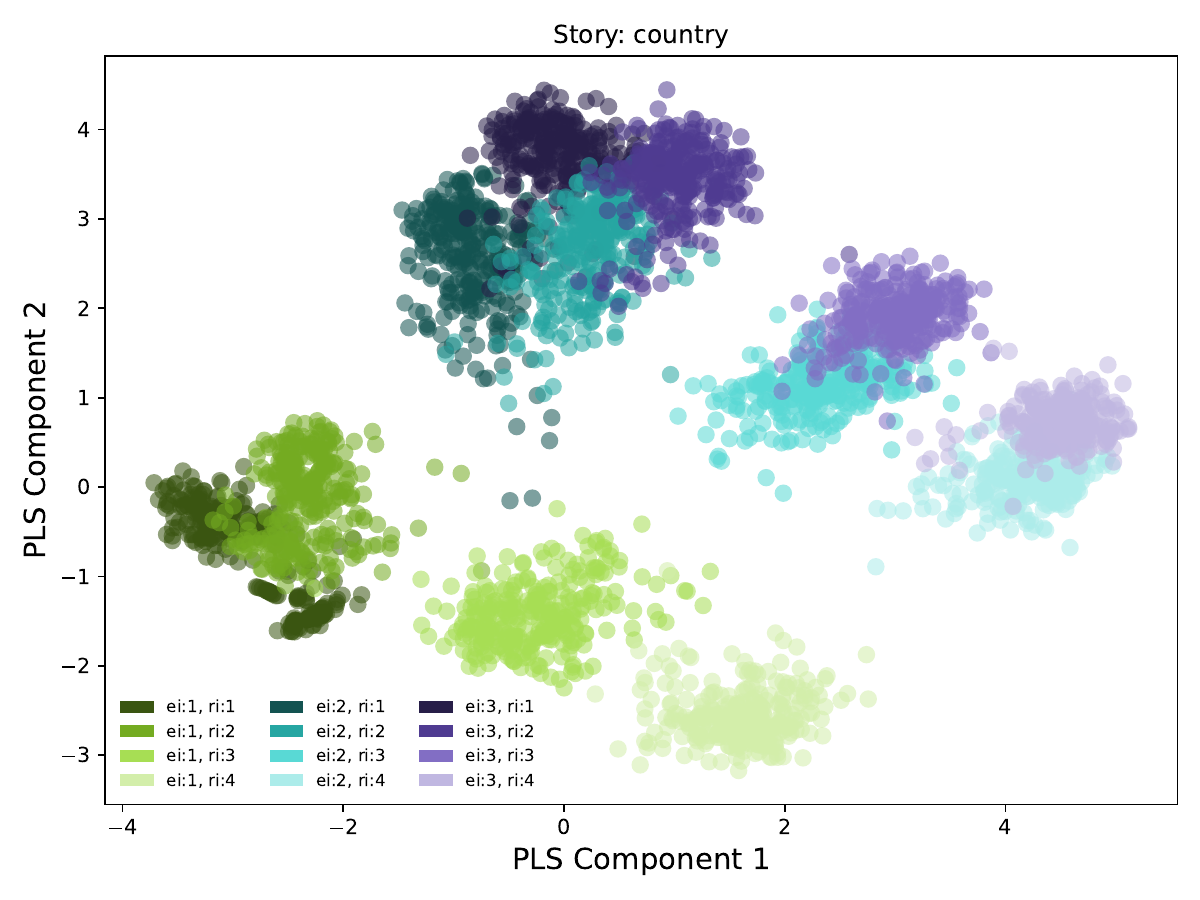}
  \caption{base}
\end{subfigure}\hfil 
\begin{subfigure}{0.325\textwidth}
  \includegraphics[width=\linewidth]{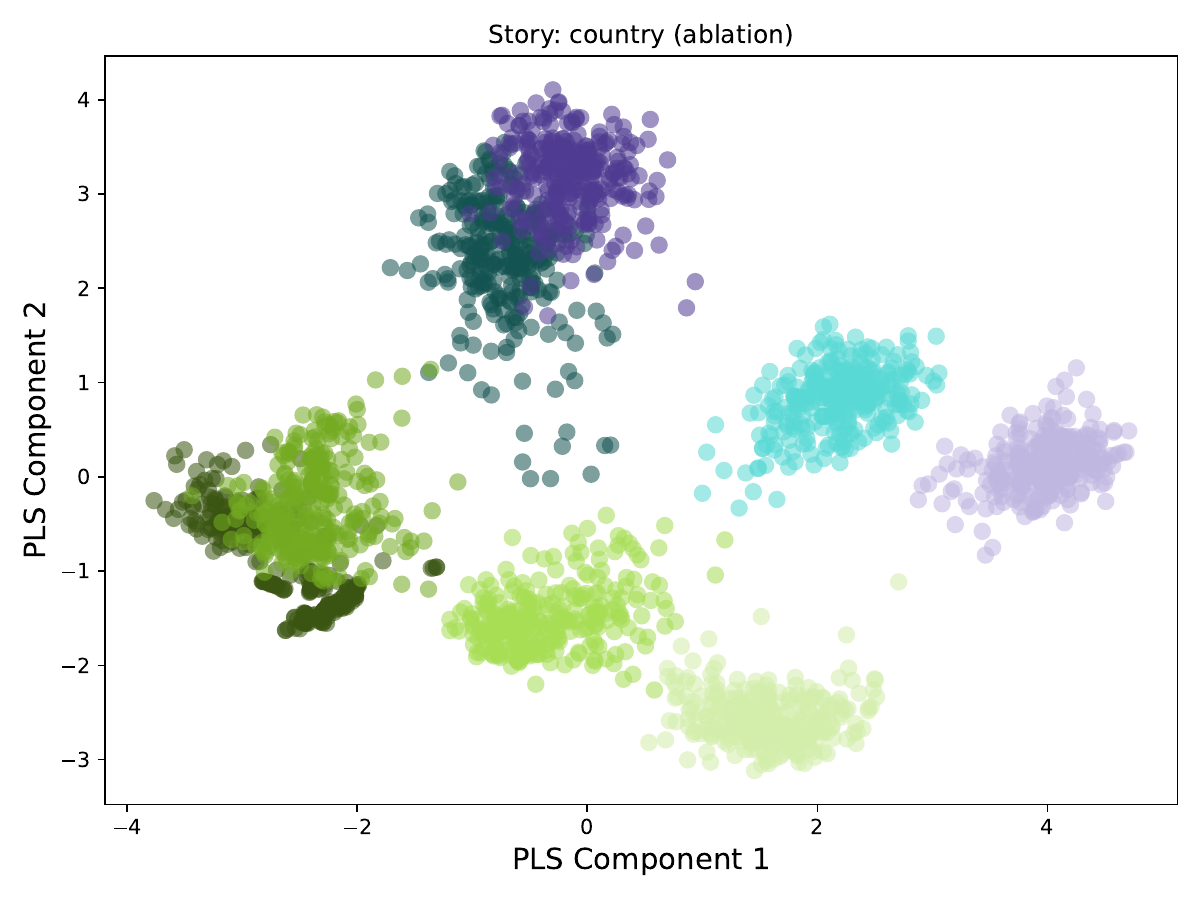}
  \caption{ablated}
\end{subfigure}\hfil 
\caption{Visualization of the CBR subspace before and after relation \textbf{ablation} from Llama3-8B-Instruct on $C_{country}$.}
\label{fig:consistency_ablate_country_llama}
\end{figure*}

\begin{figure*}[!htb]
    \centering 
\begin{subfigure}{0.325\textwidth}
  \includegraphics[width=\linewidth]{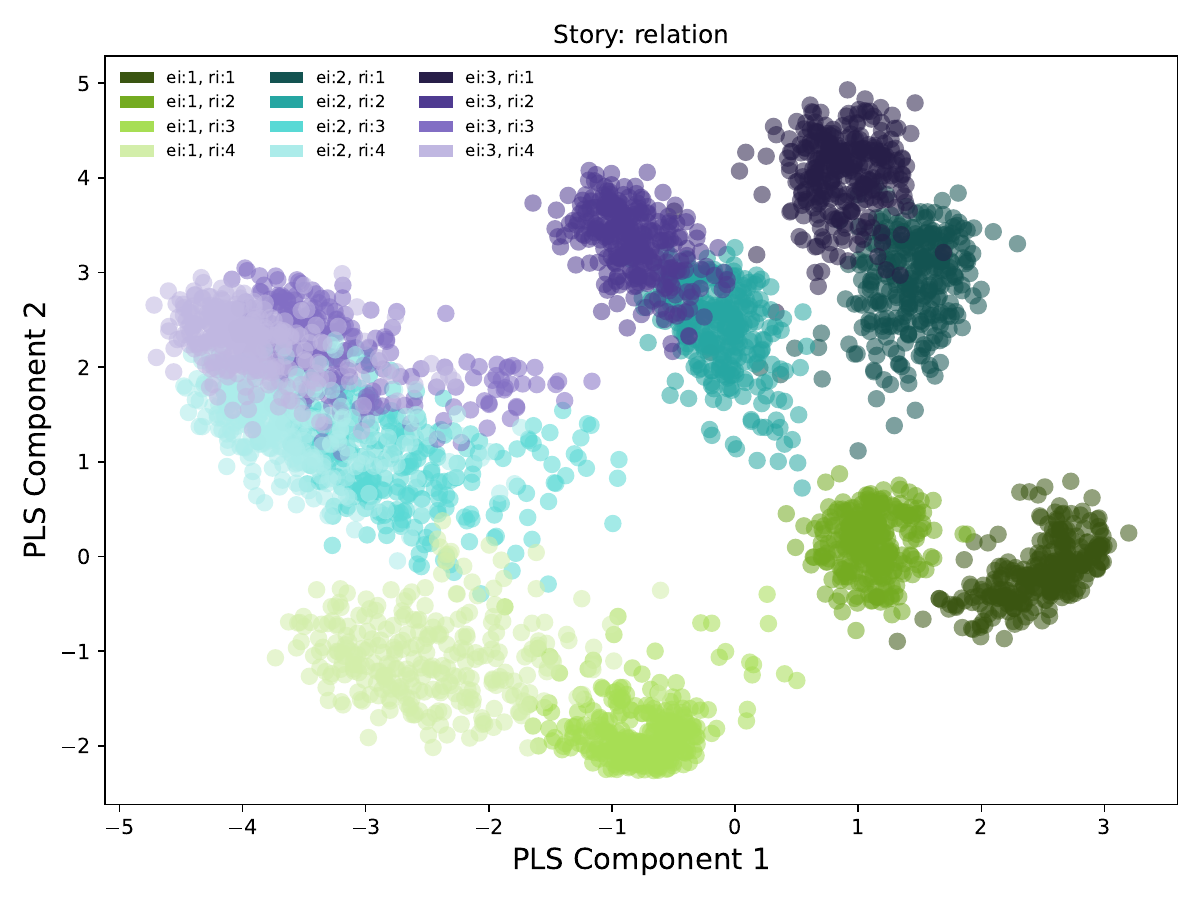}
  \caption{base}
\end{subfigure}\hfil 
\begin{subfigure}{0.325\textwidth}
  \includegraphics[width=\linewidth]{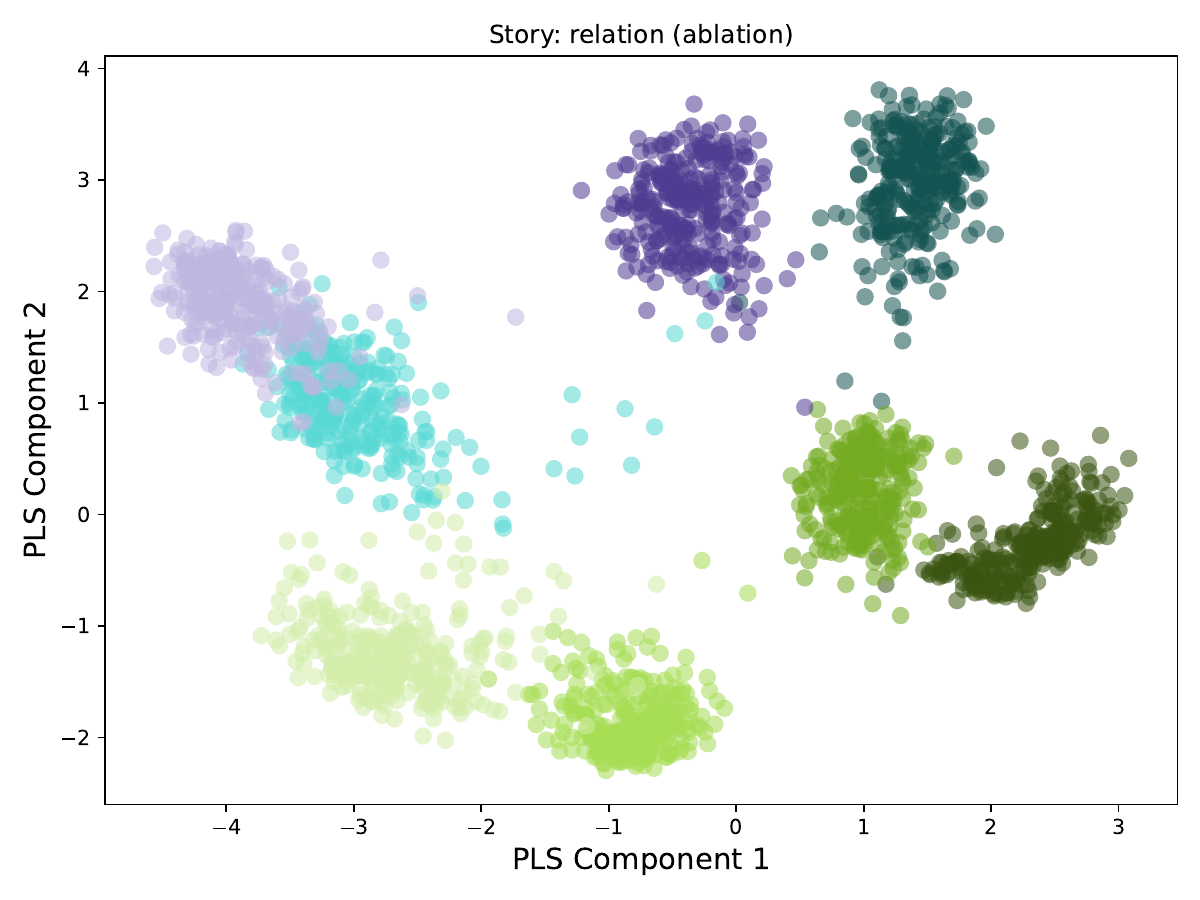}
  \caption{ablated}
\end{subfigure}\hfil 
\caption{Visualization of the CBR subspace before and after relation \textbf{ablation} from Llama3-8B-Instruct on $C_{relation}$.}
\label{fig:consistency_ablate_relation_llama}
\end{figure*}

\begin{figure*}[!htb]
    \centering 
\begin{subfigure}{0.325\textwidth}
  \includegraphics[width=\linewidth]{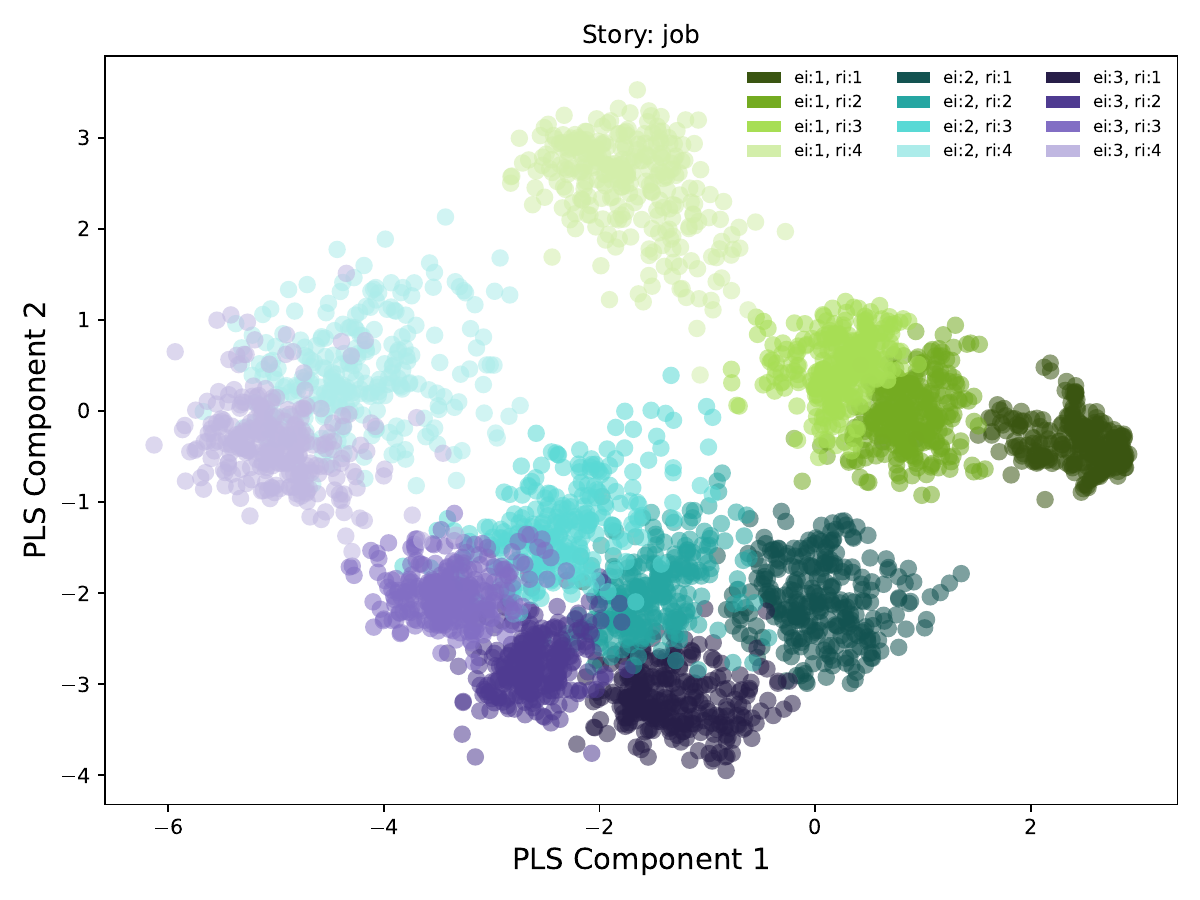}
  \caption{base}
\end{subfigure}\hfil 
\begin{subfigure}{0.325\textwidth}
  \includegraphics[width=\linewidth]{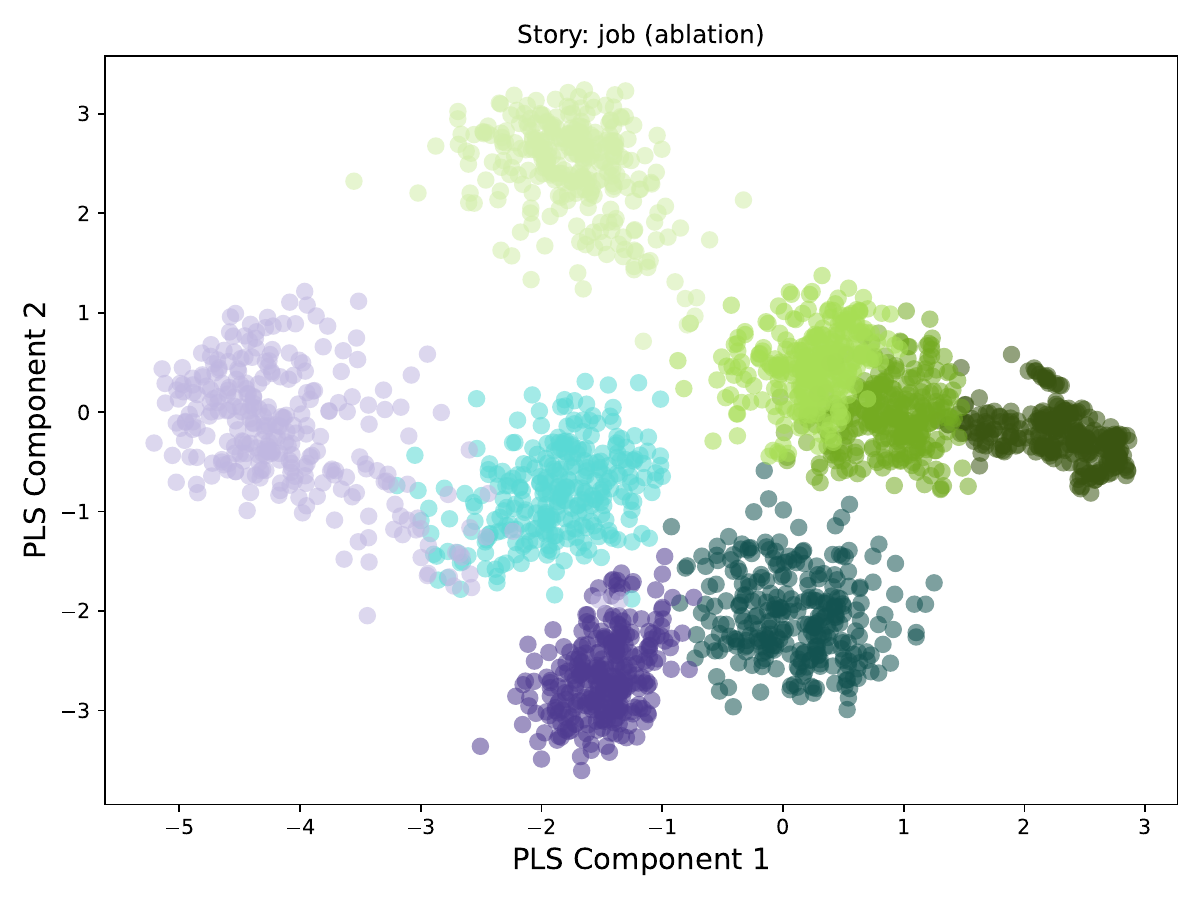}
  \caption{ablated}
\end{subfigure}\hfil 
\caption{Visualization of the CBR subspace before and after relation \textbf{ablation} from Llama3-8B-Instruct on $C_{job}$.}
\label{fig:consistency_ablate_job_llama}
\end{figure*}

\begin{figure*}[!htb]
    \centering 
\begin{subfigure}{0.325\textwidth}
  \includegraphics[width=\linewidth]{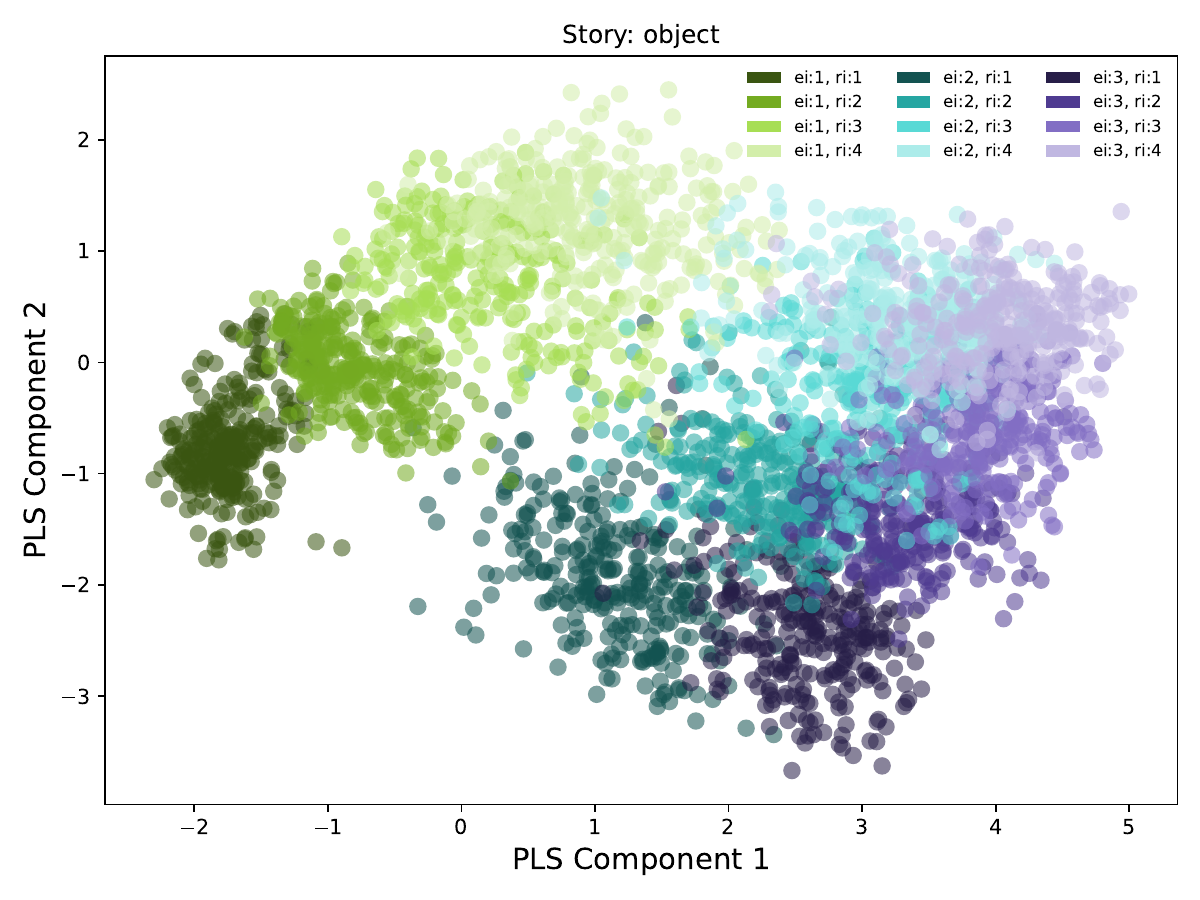}
  \caption{base}
\end{subfigure}\hfil 
\begin{subfigure}{0.325\textwidth}
  \includegraphics[width=\linewidth]{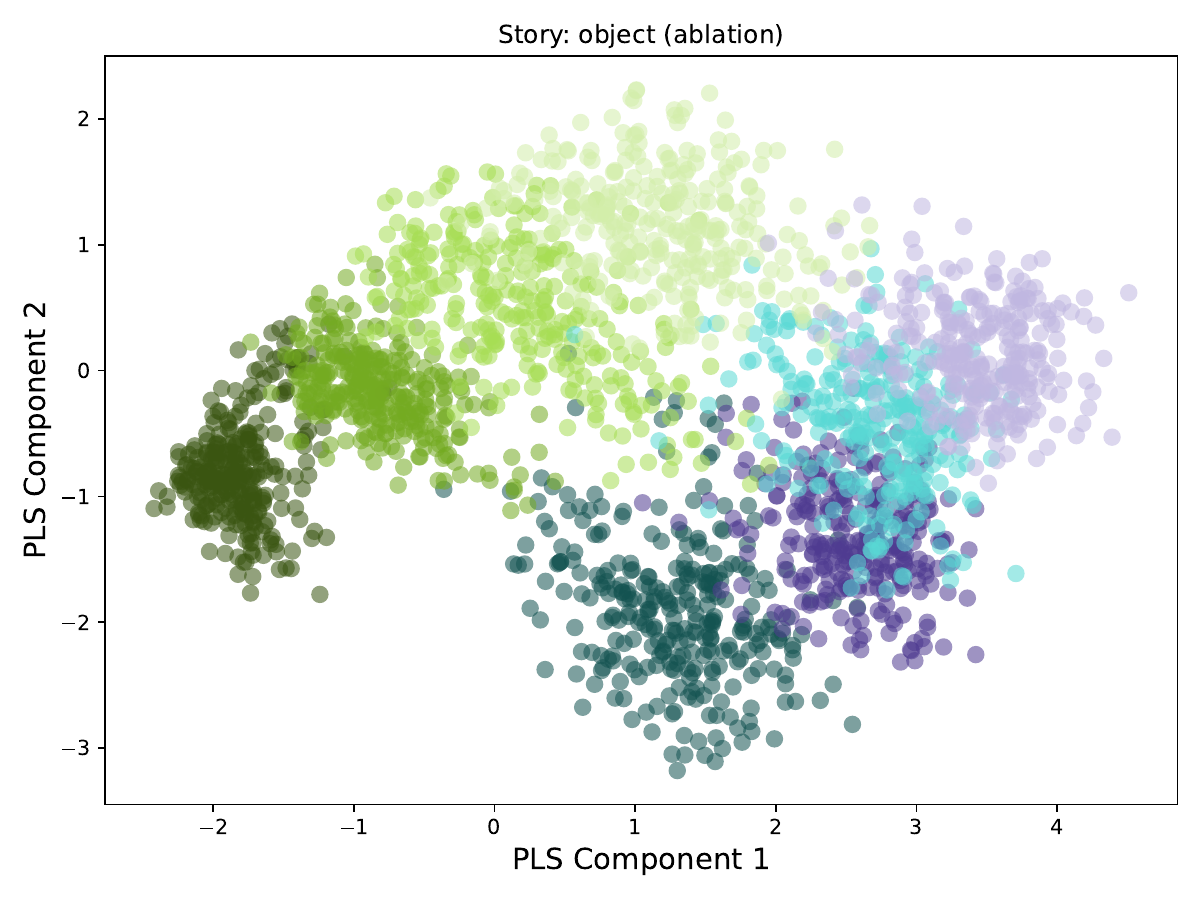}
  \caption{ablated}
\end{subfigure}\hfil 
\caption{Visualization of the CBR subspace before and after relation \textbf{ablation} from Llama3-8B-Instruct on $C_{object}$.}
\label{fig:consistency_ablate_object_llama}
\end{figure*}

Section~\secref{sec:consistency_of_irs_subspace} mentions that the overall structure of the CBR subspace remains stable under the permutation of shuffling and ablation, because the projected representations preserving their original geometric arrangement. We apply shuffling across all other discourse contexts in Figures~\ref{fig:consistency_shuffle_country_llama}, \ref{fig:consistency_shuffle_relation_llama}, \ref{fig:consistency_shuffle_job_llama}, and \ref{fig:consistency_shuffle_object_llama}. The visualizations after ablation are shown in Figures~\ref{fig:consistency_ablate_country_llama}, \ref{fig:consistency_ablate_relation_llama}, \ref{fig:consistency_ablate_job_llama}, and \ref{fig:consistency_ablate_object_llama}, where representations are visualized using the projection matrix learned from the permuted dataset. In all cases, the same pattern holds: the global geometry of the CBR subspace aligns with the $ei$ and $ri$ directions despite surface level variations. This cross-context stability confirms our earlier observations and further supports the conclusion that the CBR geometry primarily depends on the index structure rather than superficial variations.

\subsection{Consistency of CBR Subspace on Qwen3-8B}
\label{sec:irs_consistency_qwen}
\begin{figure*}[htb]
    \centering 
\begin{subfigure}{0.325\textwidth}
  \includegraphics[width=\linewidth]{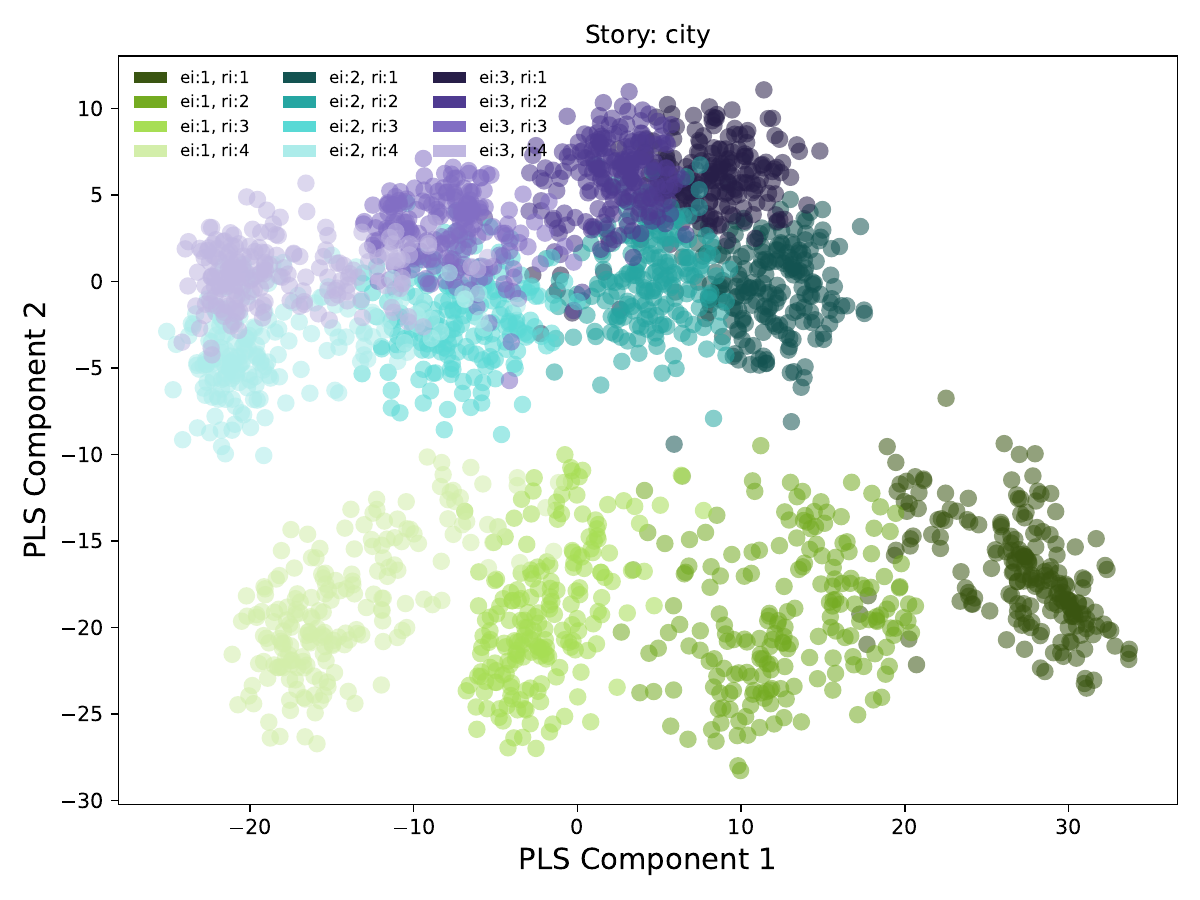}
  \caption{base}
\end{subfigure}\hfil 
\begin{subfigure}{0.325\textwidth}
  \includegraphics[width=\linewidth]{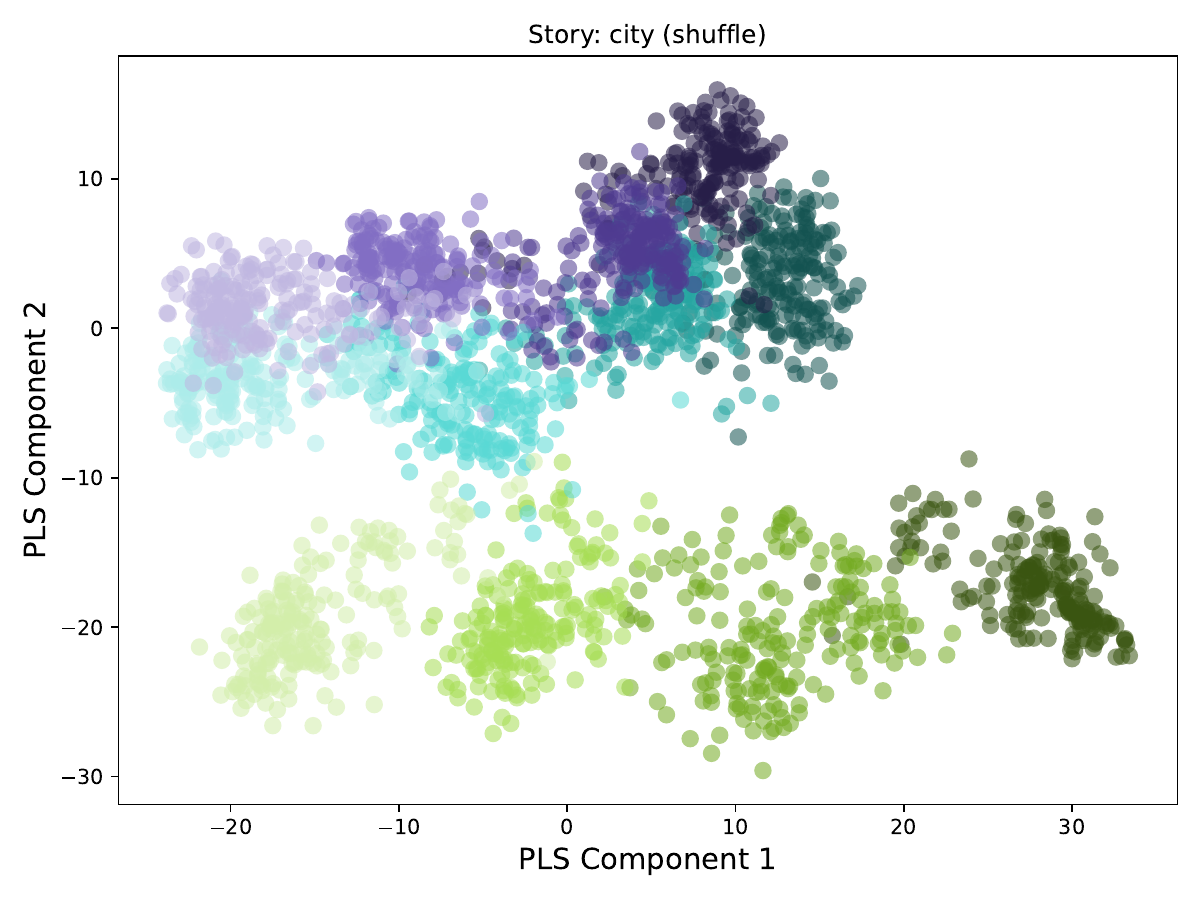}
  \caption{shuffled}
\end{subfigure}\hfil 
\caption{Visualization of the CBR subspace before and after \textbf{shuffling} from Qwen3-8B on $C_{city}$.}
\label{fig:consistency_shuffle_city_qwen}
\end{figure*}

\begin{figure*}[htb]
    \centering 
\begin{subfigure}{0.325\textwidth}
  \includegraphics[width=\linewidth]{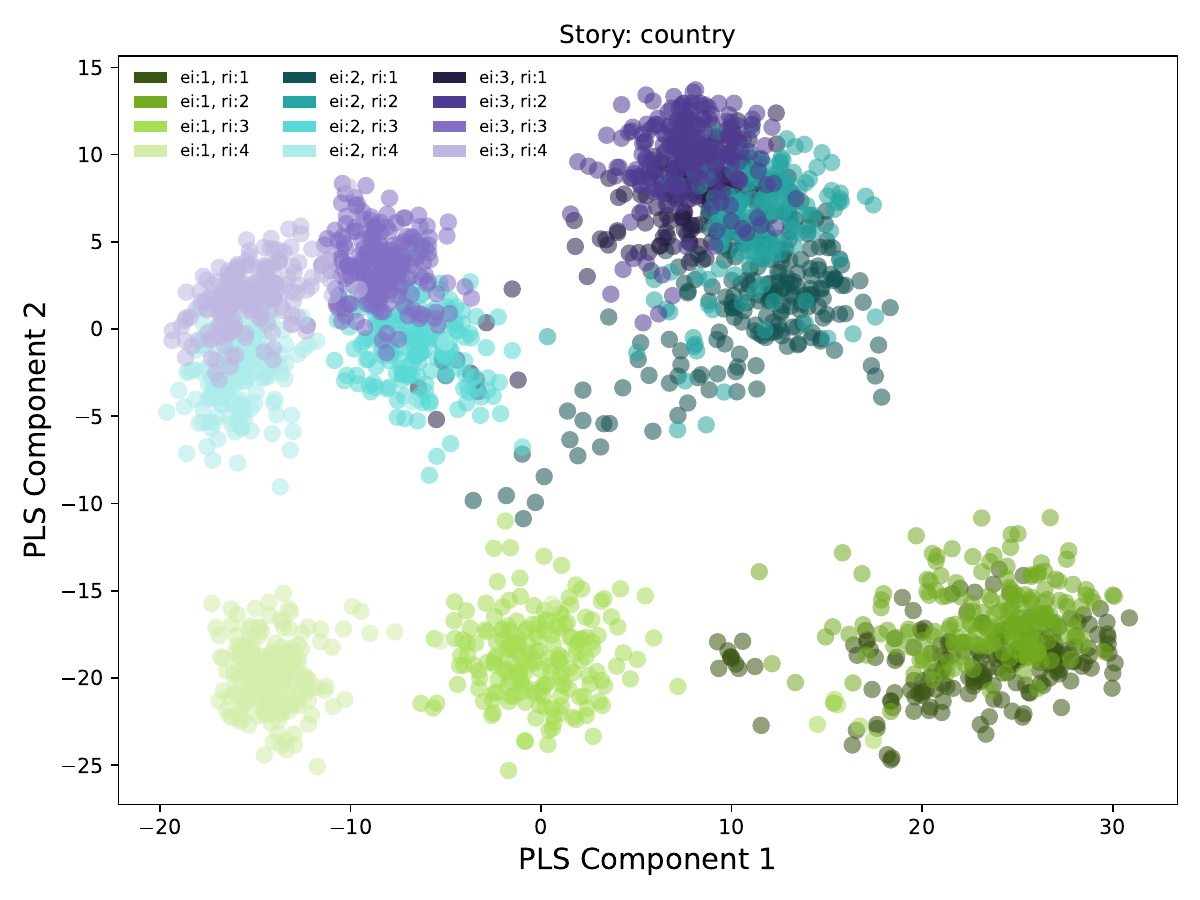}
  \caption{base}
\end{subfigure}\hfil 
\begin{subfigure}{0.325\textwidth}
  \includegraphics[width=\linewidth]{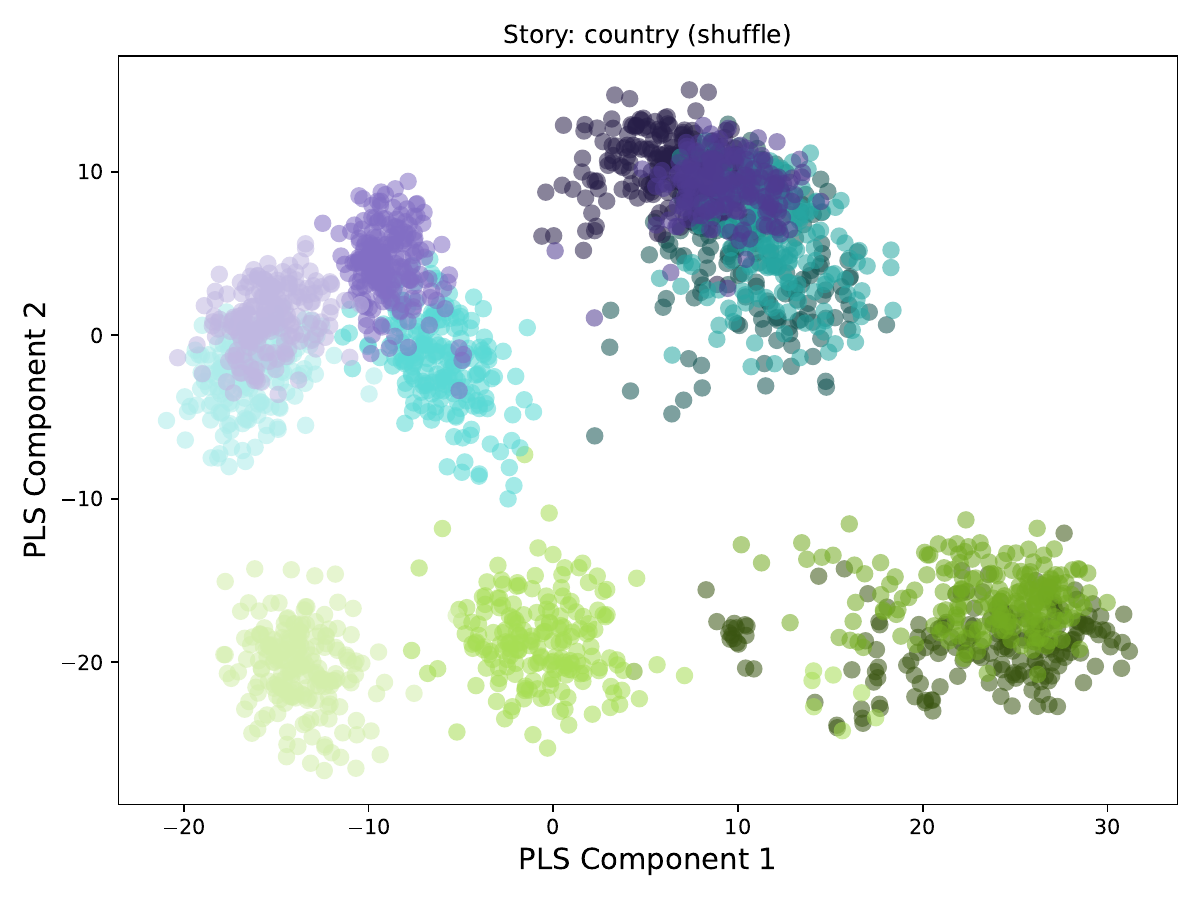}
  \caption{shuffled}
\end{subfigure}\hfil 
\caption{Visualization of the CBR subspace before and after \textbf{shuffling} from Qwen3-8B on $C_{country}$.}
\label{fig:consistency_shuffle_country_qwen}
\end{figure*}

\begin{figure*}[htb]
    \centering 
\begin{subfigure}{0.325\textwidth}
  \includegraphics[width=\linewidth]{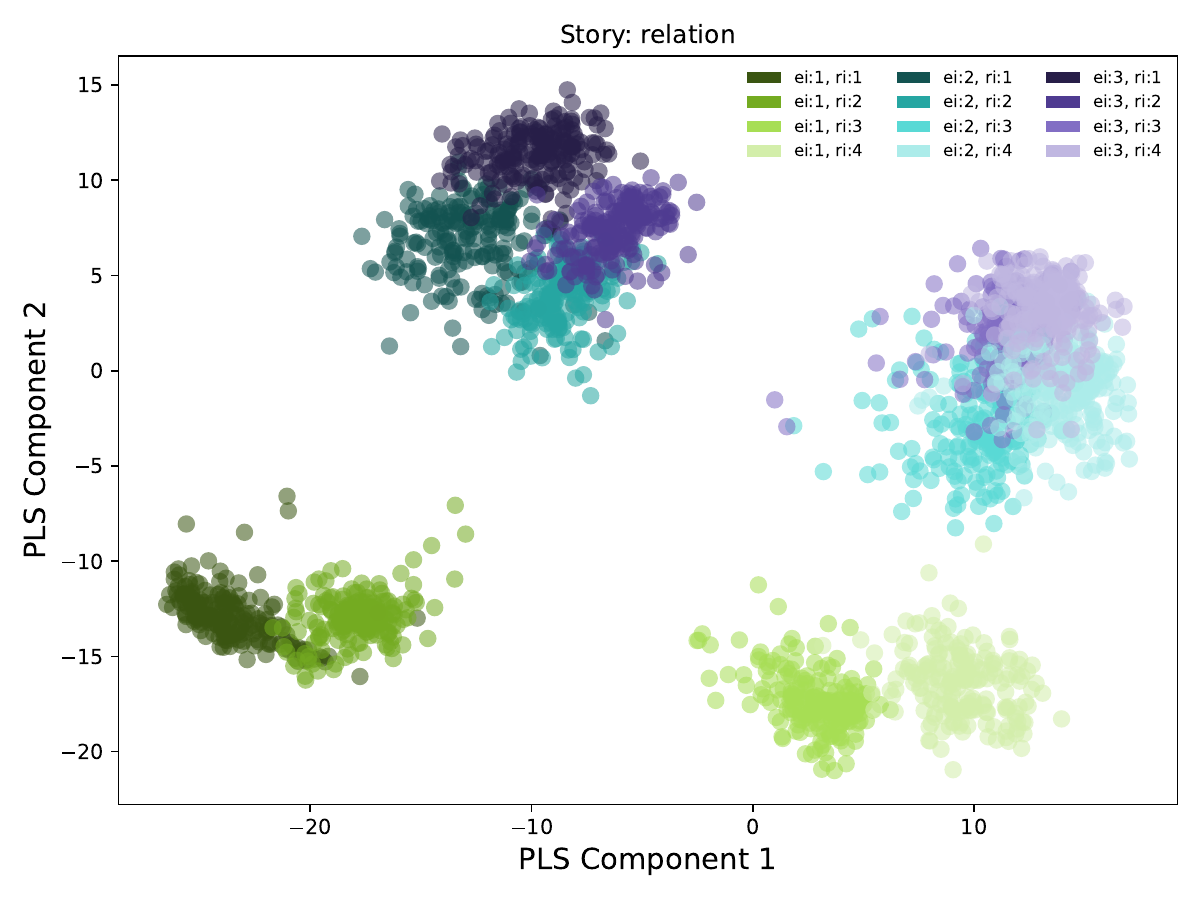}
  \caption{base}
\end{subfigure}\hfil 
\begin{subfigure}{0.325\textwidth}
  \includegraphics[width=\linewidth]{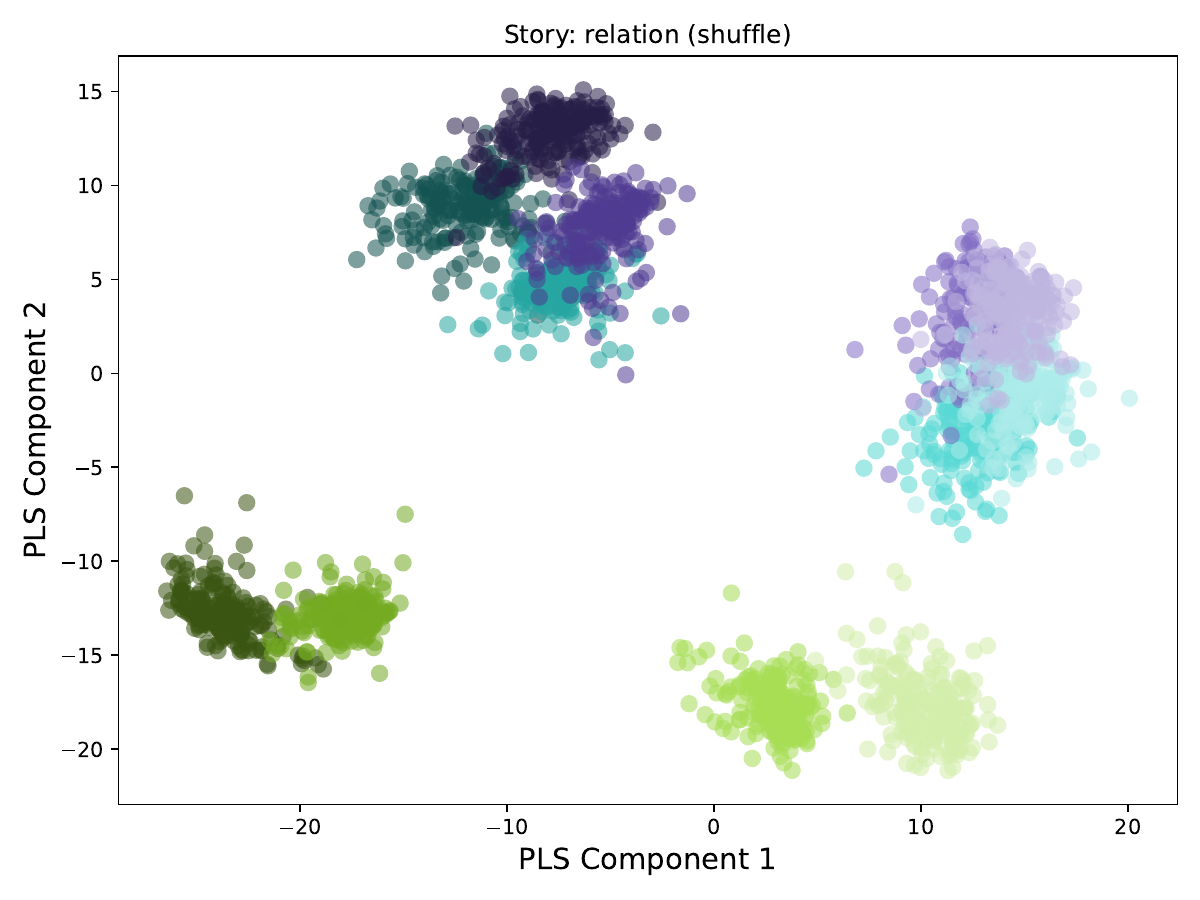}
  \caption{shuffled}
\end{subfigure}\hfil 
\caption{Visualization of the CBR subspace before and after \textbf{shuffling} from Qwen3-8B on $C_{relation}$.}
\label{fig:consistency_shuffle_relation_qwen}
\end{figure*}

\begin{figure*}[htb]
    \centering 
\begin{subfigure}{0.325\textwidth}
  \includegraphics[width=\linewidth]{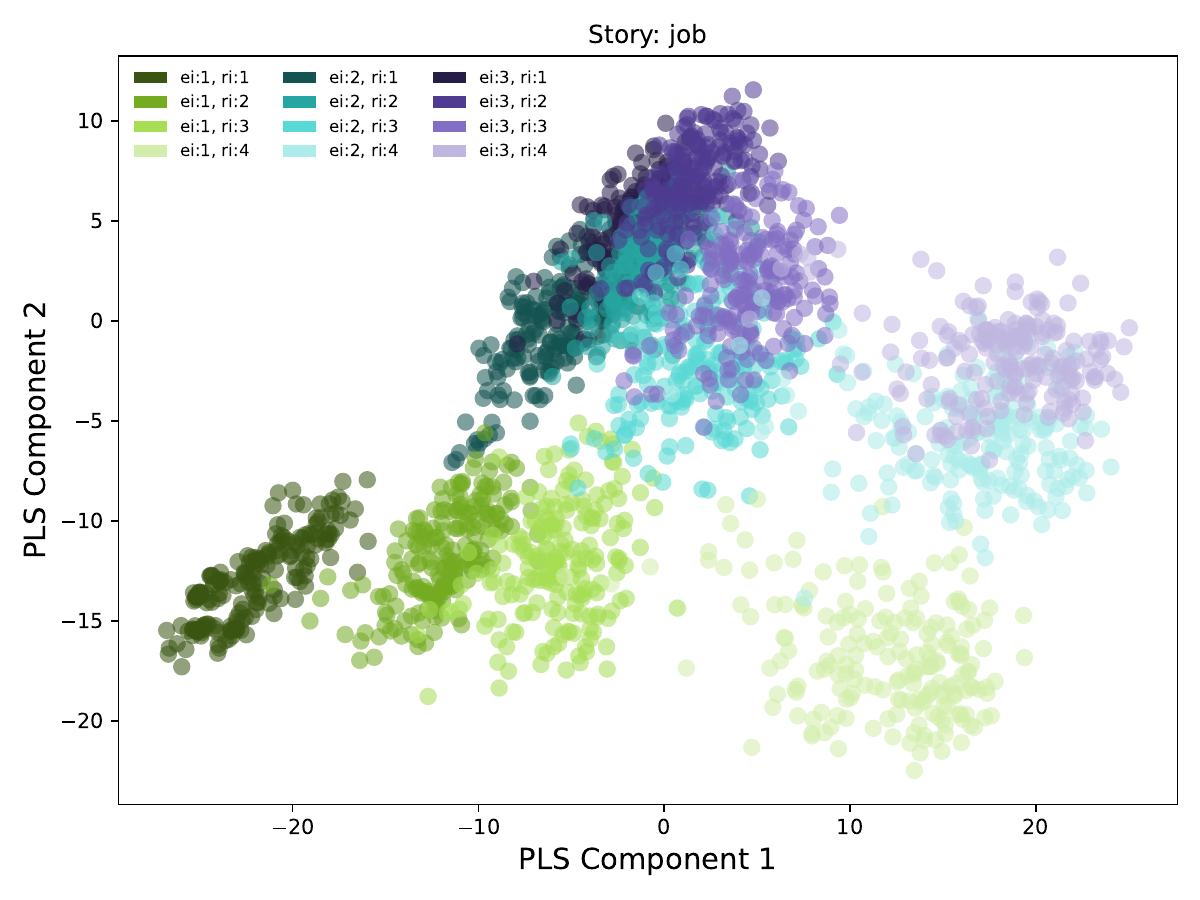}
  \caption{base}
\end{subfigure}\hfil 
\begin{subfigure}{0.325\textwidth}
  \includegraphics[width=\linewidth]{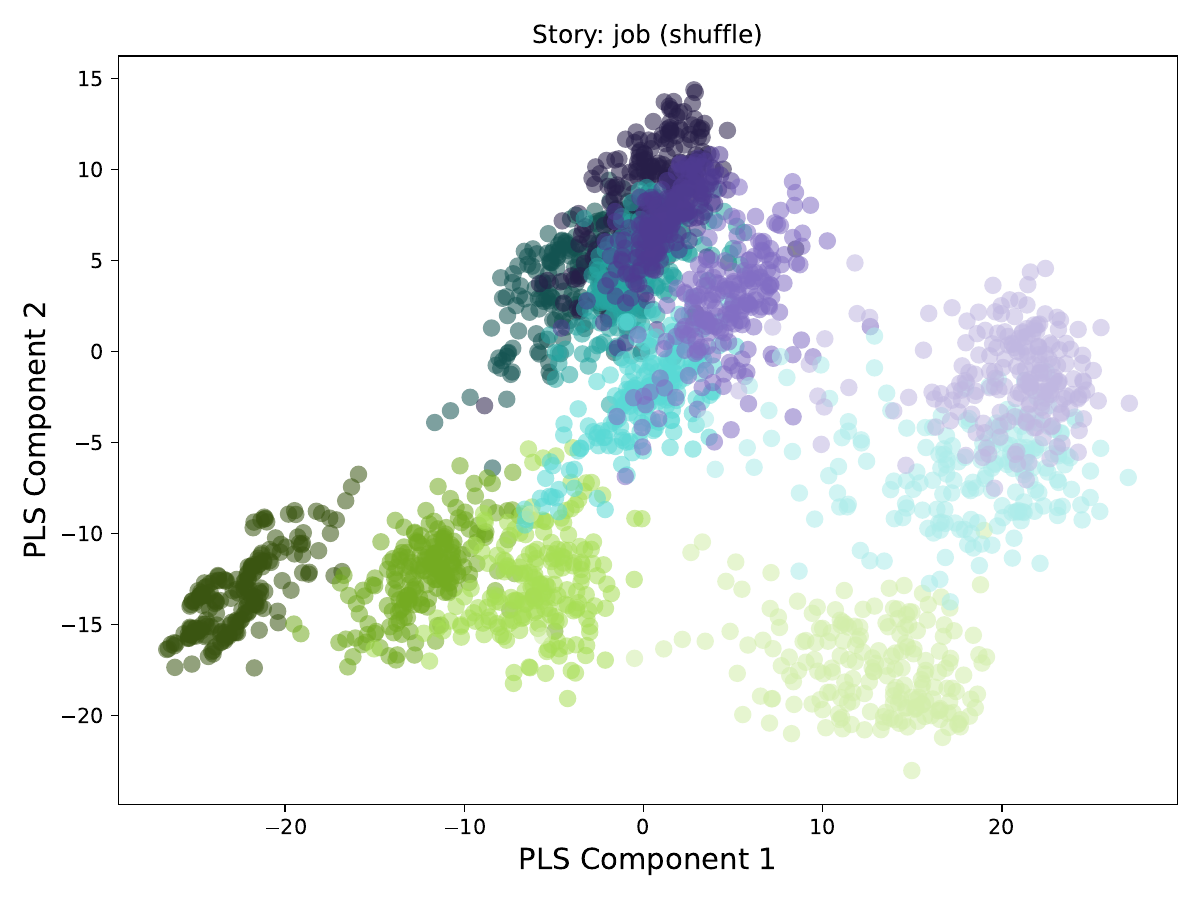}
  \caption{shuffled}
\end{subfigure}\hfil 
\caption{Visualization of the CBR subspace before and after \textbf{shuffling} from Qwen3-8B on $C_{job}$.}
\label{fig:consistency_shuffle_job_qwen}
\end{figure*}

\begin{figure*}[htb]
    \centering 
\begin{subfigure}{0.325\textwidth}
  \includegraphics[width=\linewidth]{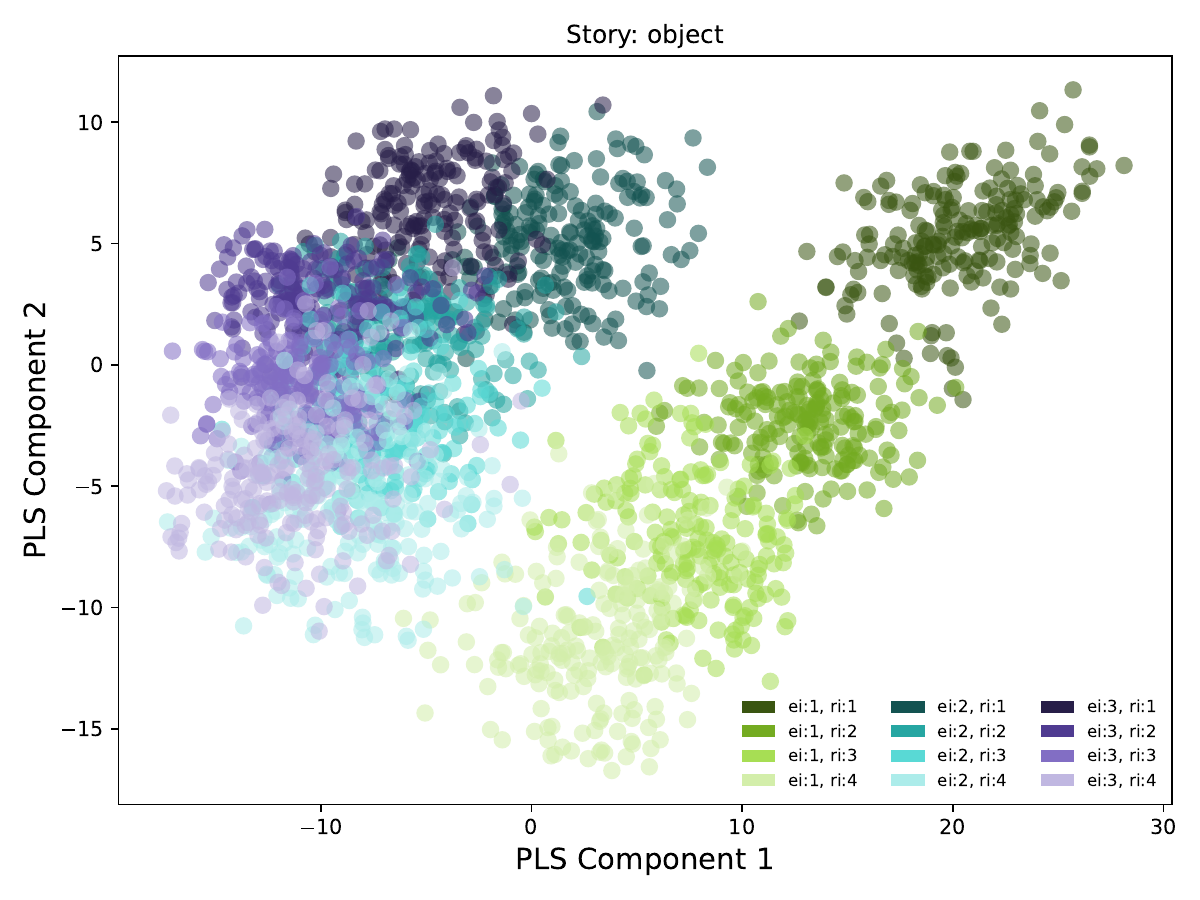}
  \caption{base}
\end{subfigure}\hfil 
\begin{subfigure}{0.325\textwidth}
  \includegraphics[width=\linewidth]{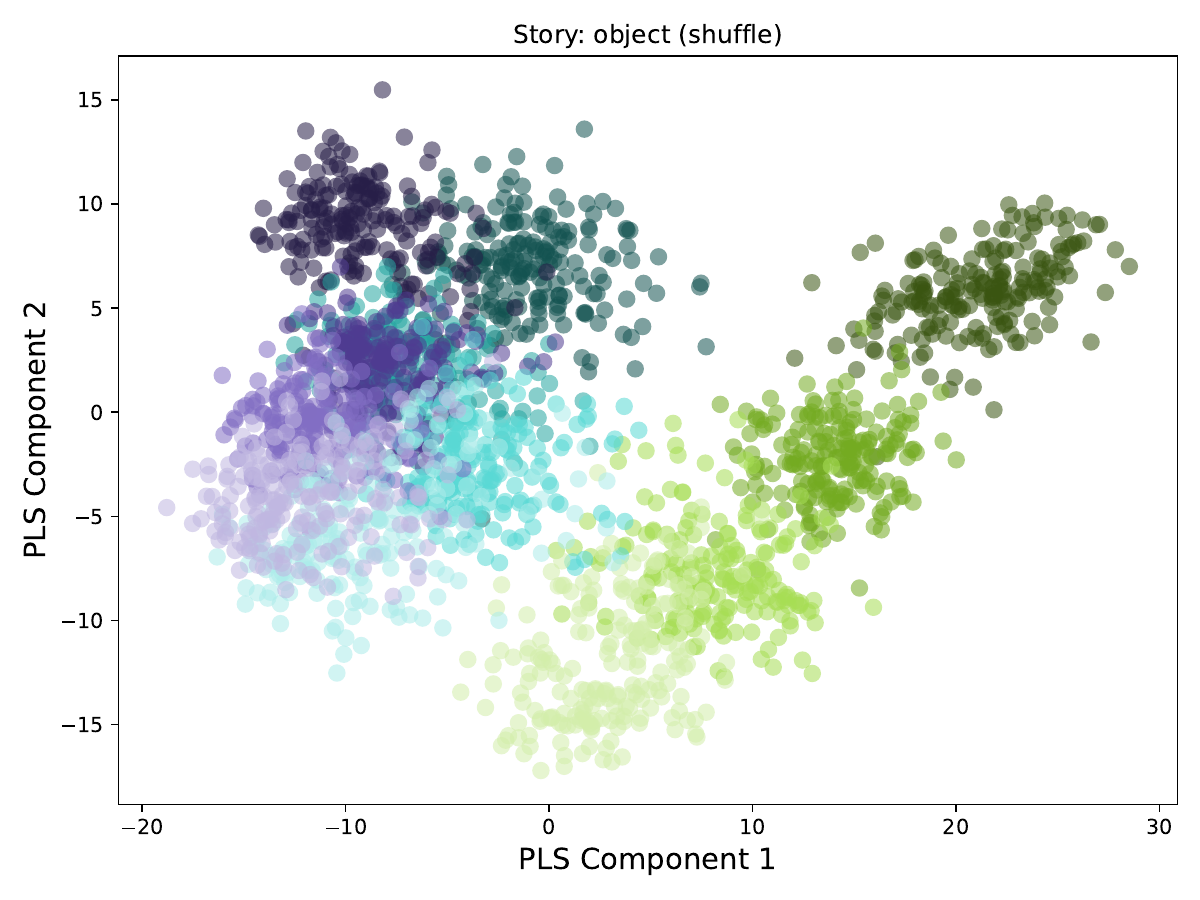}
  \caption{shuffled}
\end{subfigure}\hfil 
\caption{Visualization of the CBR subspace before and after \textbf{shuffling} from Qwen3-8B on $C_{object}$.}
\label{fig:consistency_shuffle_object_qwen}
\end{figure*}

Figure~\ref{fig:consistency_shuffle_city_qwen}, \ref{fig:consistency_shuffle_country_qwen}, \ref{fig:consistency_shuffle_relation_qwen}, \ref{fig:consistency_shuffle_job_qwen} and \ref{fig:consistency_shuffle_object_qwen} visualize the CBR subspace before and after shuffling the relation order of the second and third entities in all the contexts. The stability of the distribution shows that the CBR subspace in Qwen3-8B primarily depends on index structure rather than superficial permutation.

\begin{figure*}[htb]
    \centering 
\begin{subfigure}{0.325\textwidth}
  \includegraphics[width=\linewidth]{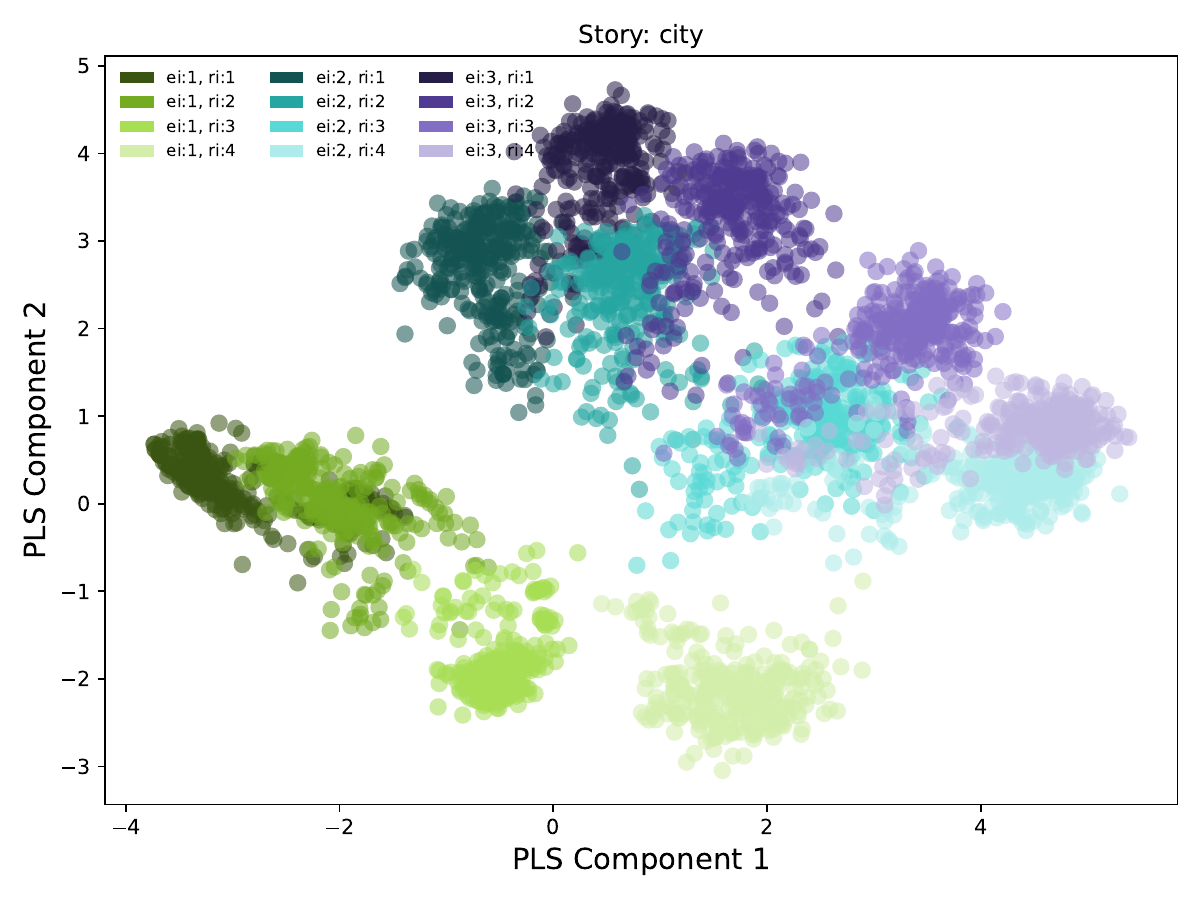}
  \caption{base}
\end{subfigure}\hfil 
\begin{subfigure}{0.325\textwidth}
  \includegraphics[width=\linewidth]{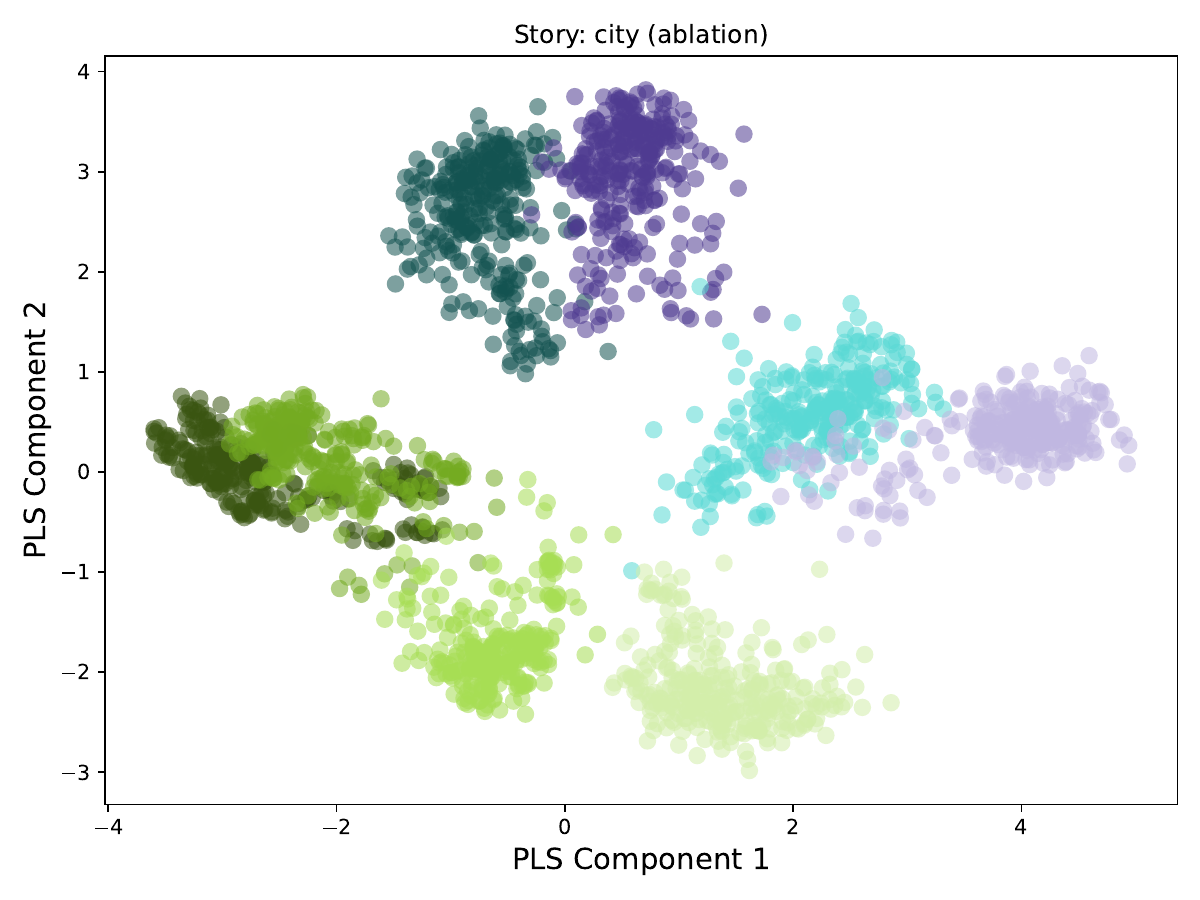}
  \caption{ablated}
\end{subfigure}\hfil 
\caption{Visualization of the CBR subspace before and after relation \textbf{ablation} from Qwen3-8B on $C_{city}$.}
\label{fig:consistency_ablate_city_qwen}
\end{figure*}

\begin{figure*}[htb]
    \centering 
\begin{subfigure}{0.325\textwidth}
  \includegraphics[width=\linewidth]{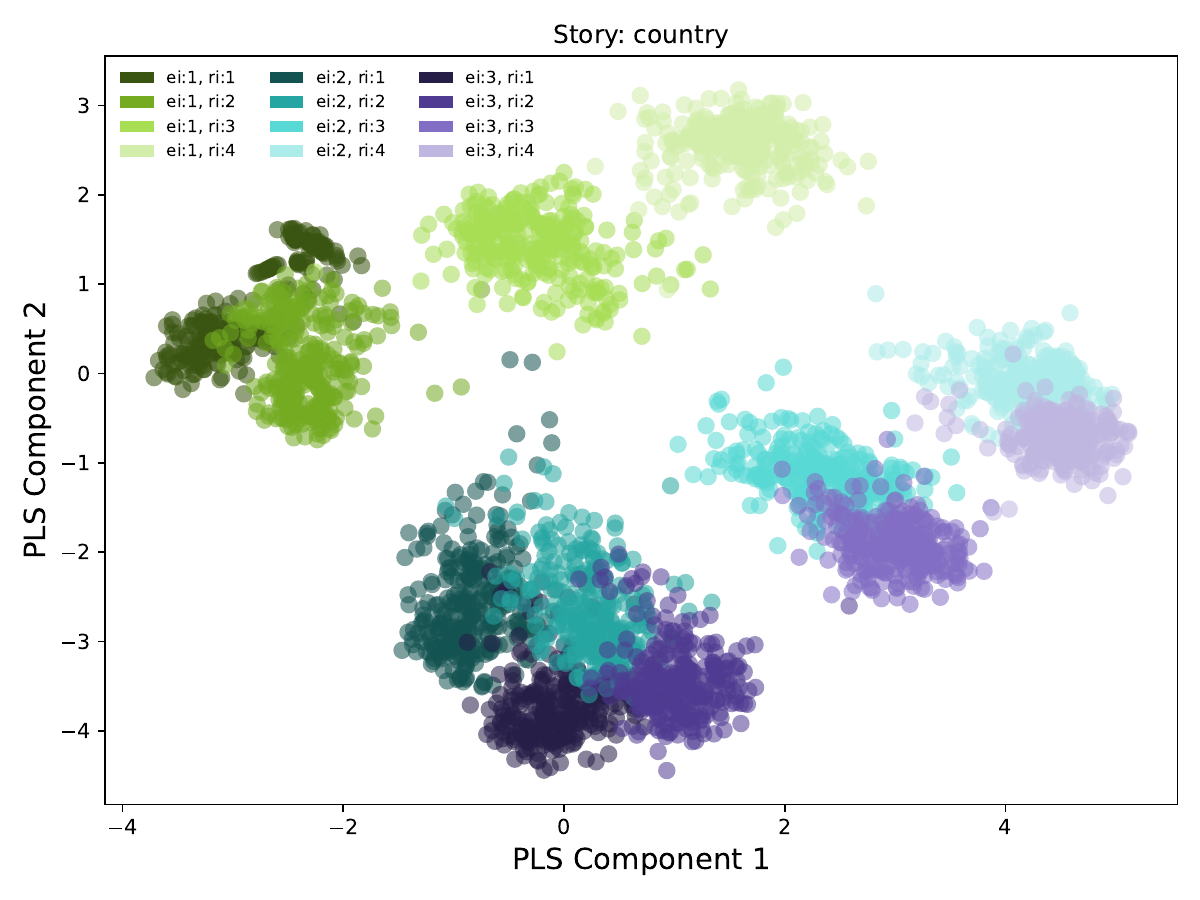}
  \caption{base}
\end{subfigure}\hfil 
\begin{subfigure}{0.325\textwidth}
  \includegraphics[width=\linewidth]{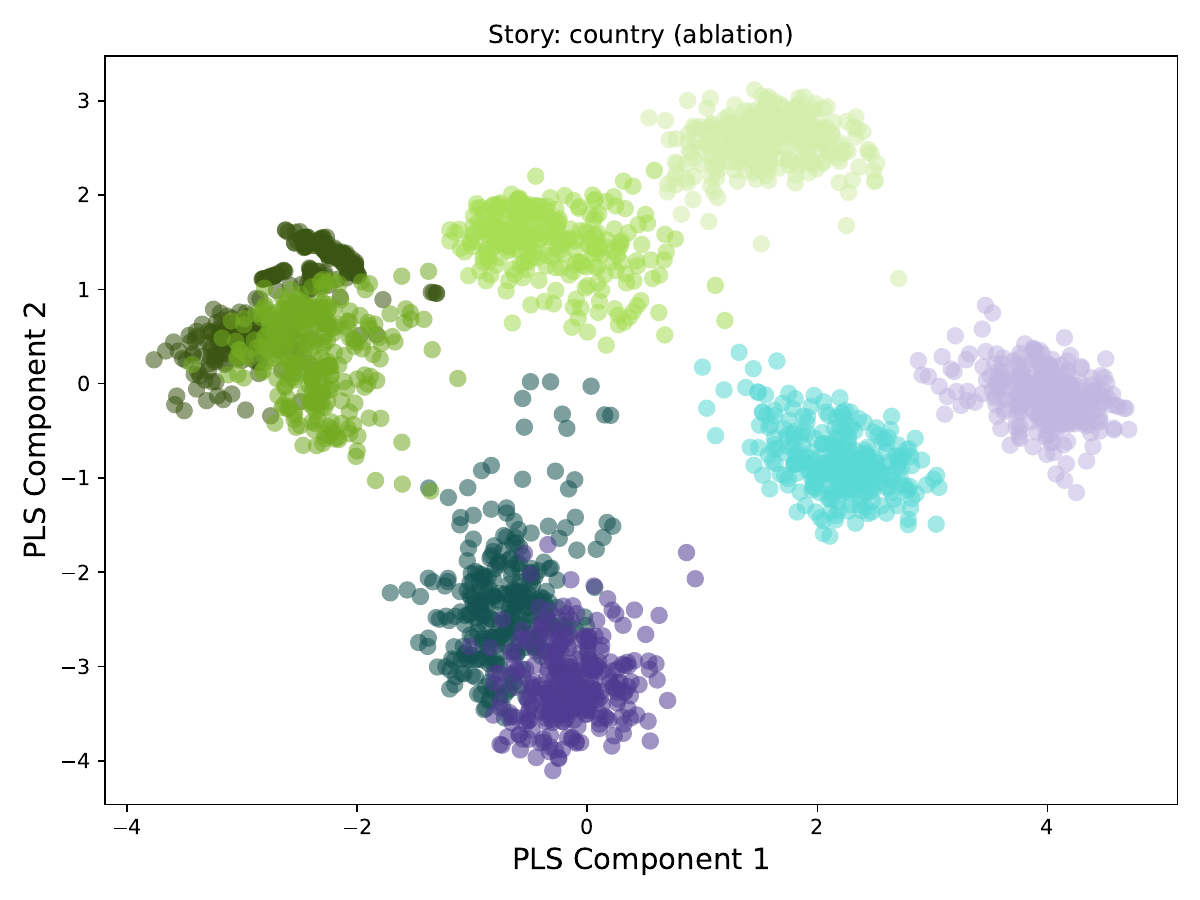}
  \caption{ablated}
\end{subfigure}\hfil 
\caption{Visualization of the CBR subspace before and after relation \textbf{ablation} from Qwen3-8B on $C_{country}$.}
\label{fig:consistency_ablate_country_qwen}
\end{figure*}

\begin{figure*}[htb]
    \centering 
\begin{subfigure}{0.325\textwidth}
  \includegraphics[width=\linewidth]{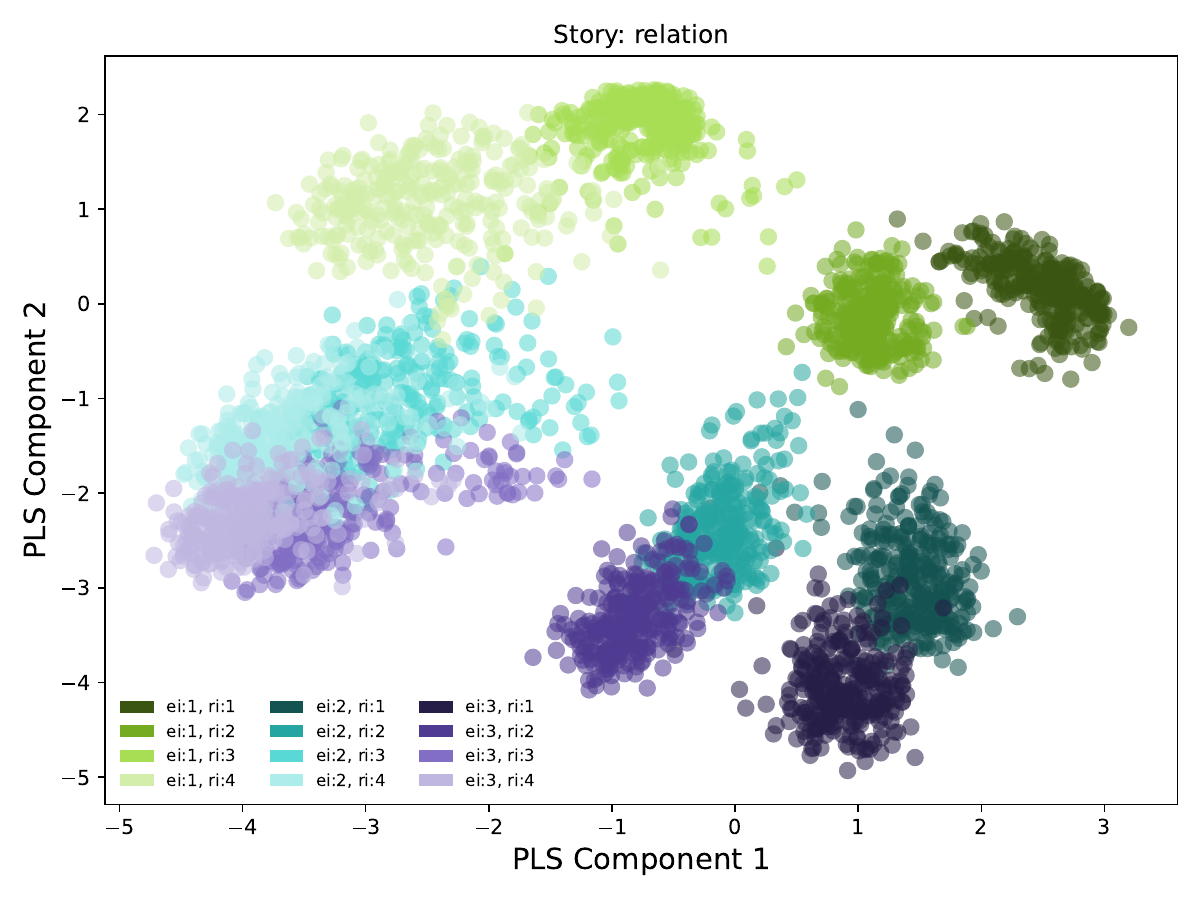}
  \caption{base}
\end{subfigure}\hfil 
\begin{subfigure}{0.325\textwidth}
  \includegraphics[width=\linewidth]{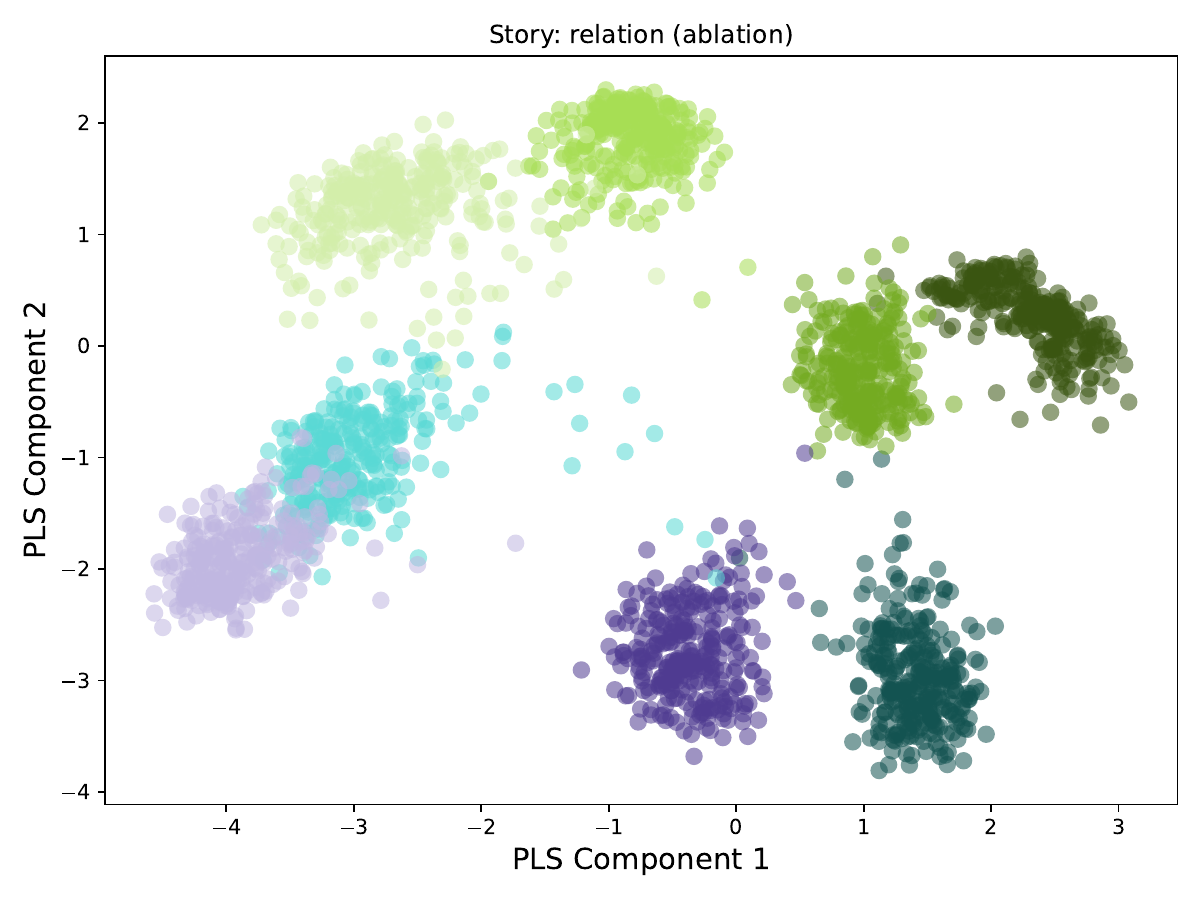}
  \caption{ablated}
\end{subfigure}\hfil 
\caption{Visualization of the CBR subspace before and after relation \textbf{ablation} from Qwen3-8B on $C_{relation}$.}
\label{fig:consistency_ablate_relation_qwen}
\end{figure*}

\begin{figure*}[htb]
    \centering 
\begin{subfigure}{0.325\textwidth}
  \includegraphics[width=\linewidth]{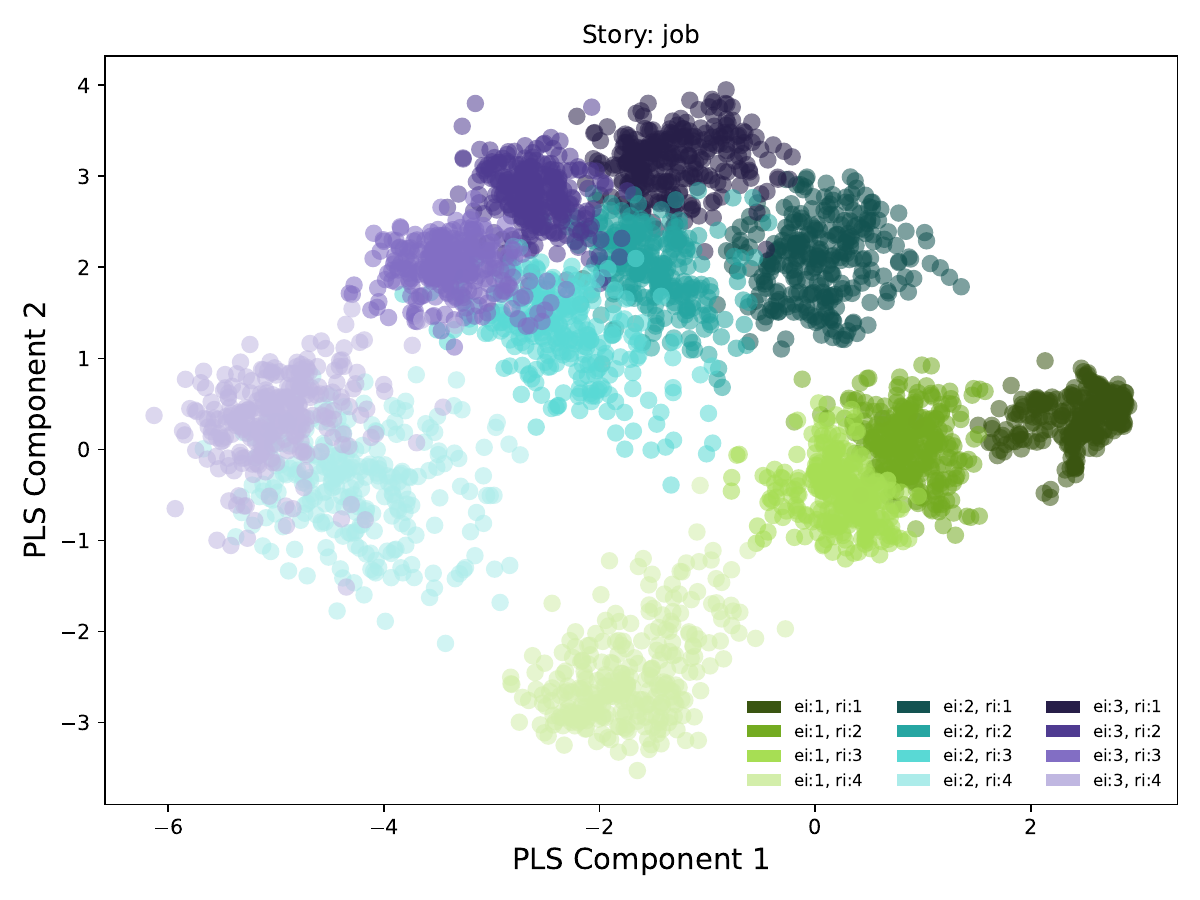}
  \caption{base}
\end{subfigure}\hfil 
\begin{subfigure}{0.325\textwidth}
  \includegraphics[width=\linewidth]{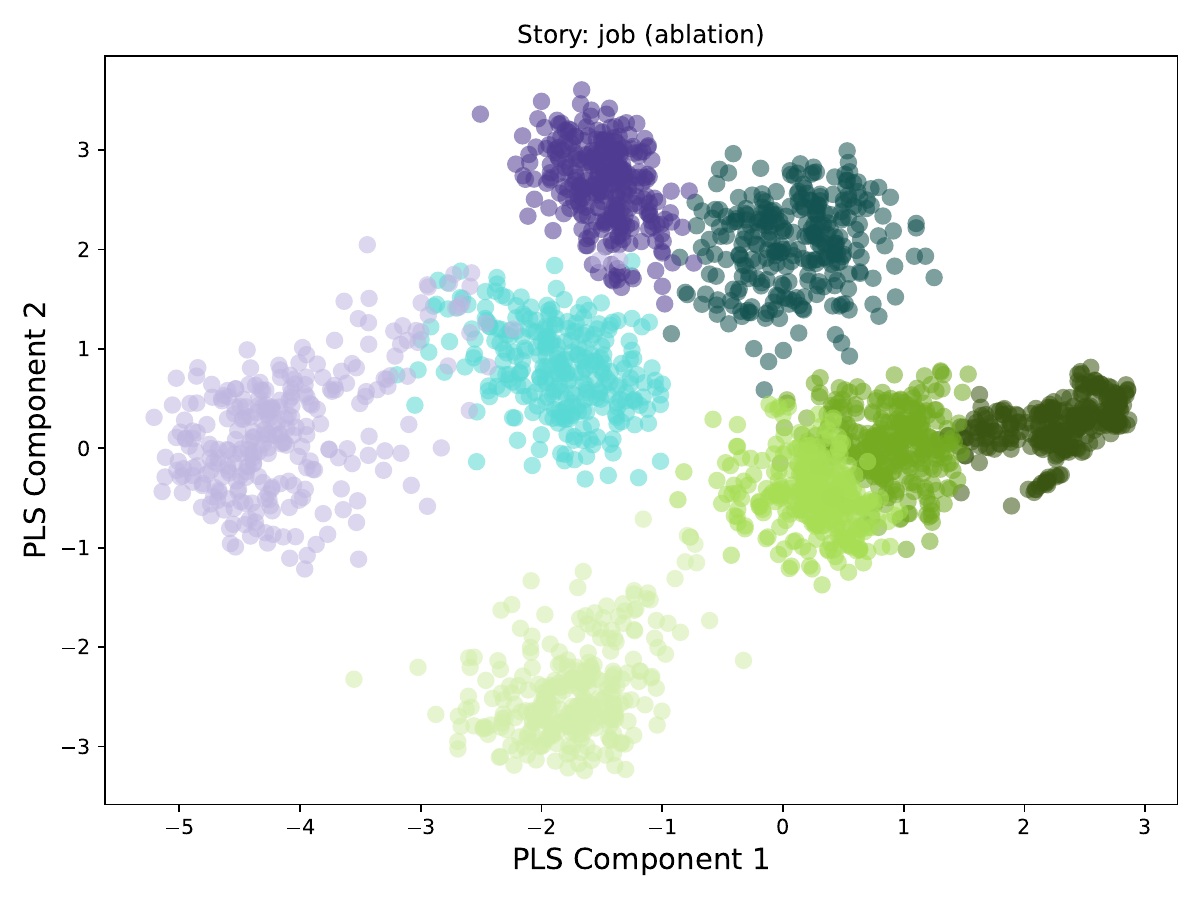}
  \caption{ablated}
\end{subfigure}\hfil 
\caption{Visualization of the CBR subspace before and after relation \textbf{ablation} from Qwen3-8B on $C_{job}$.}
\label{fig:consistency_ablate_job_qwen}
\end{figure*}

\begin{figure*}[htb]
    \centering 
\begin{subfigure}{0.325\textwidth}
  \includegraphics[width=\linewidth]{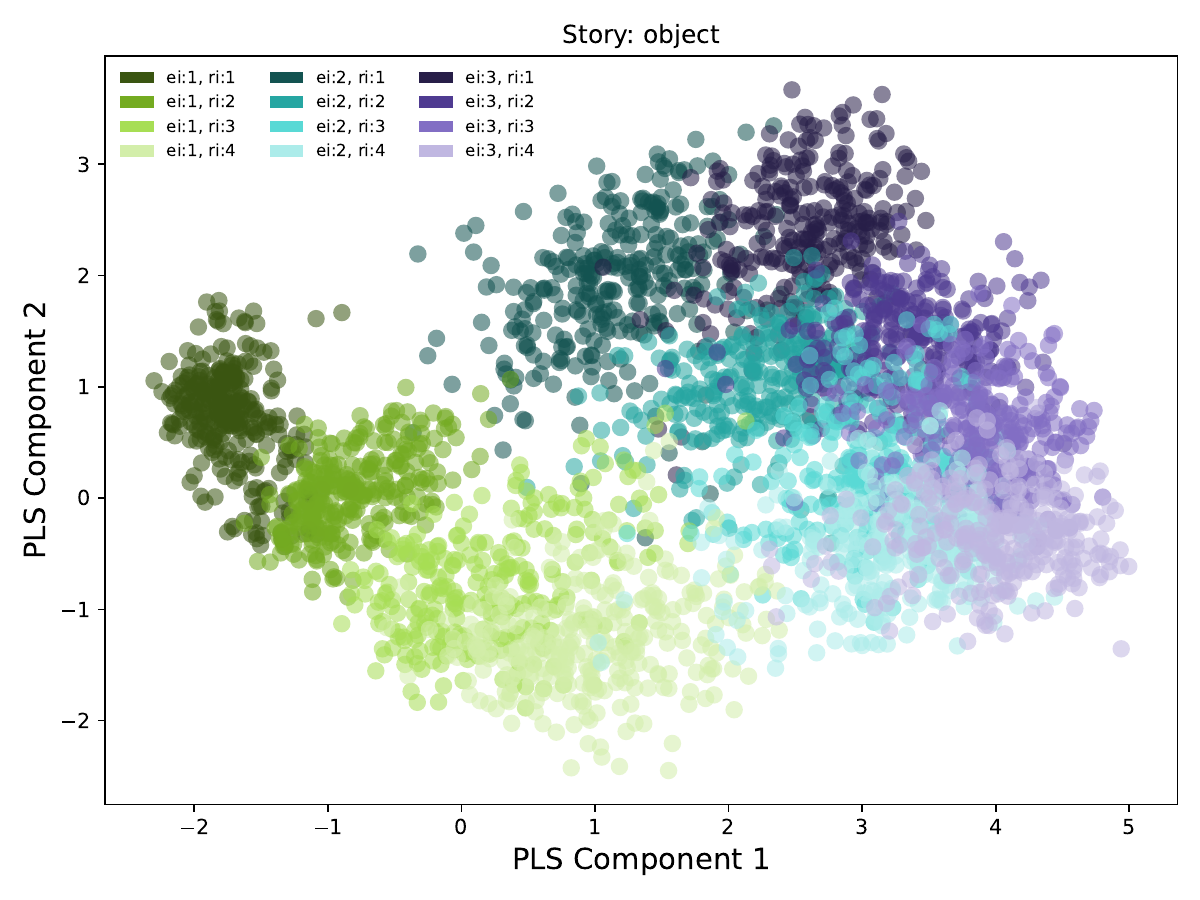}
  \caption{base}
\end{subfigure}\hfil 
\begin{subfigure}{0.325\textwidth}
  \includegraphics[width=\linewidth]{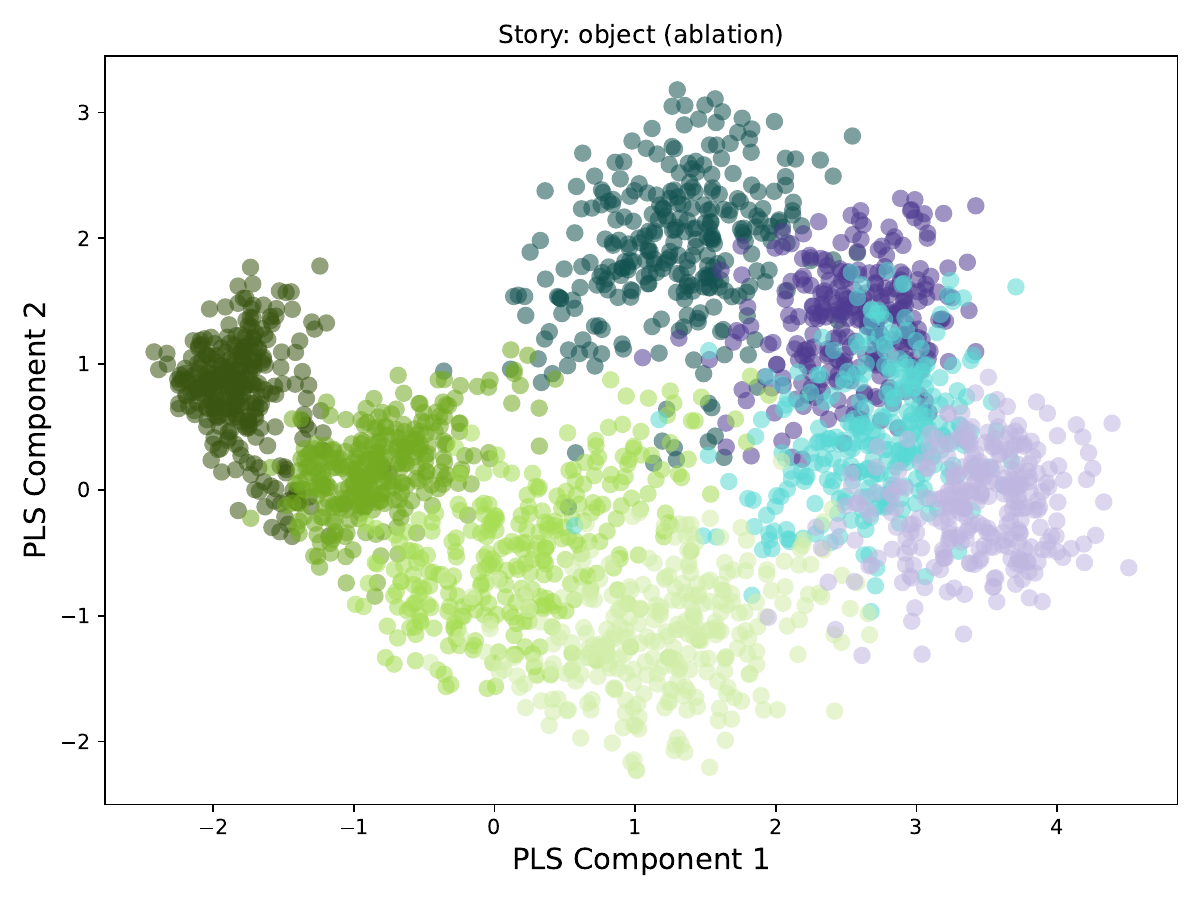}
  \caption{ablated}
\end{subfigure}\hfil 
\caption{Visualization of the CBR subspace before and after relation \textbf{ablation} from Qwen3-8B on $C_{object}$.}
\label{fig:consistency_ablate_object_qwen}
\end{figure*}

Figure~\ref{fig:consistency_ablate_city_qwen}, \ref{fig:consistency_ablate_country_qwen}, \ref{fig:consistency_ablate_relation_qwen}, \ref{fig:consistency_ablate_job_qwen} and \ref{fig:consistency_ablate_object_qwen} visualize the CBR subspace before and after relation ablation in all the contexts. The overall geometric structure aligns with the $ei$ and $ri$ directions despite surface level variations, indicating that removing relational content that does not affect the overall structure of CBR subspace in Qwen3-8B.


\clearpage

\subsection{Consistency Analysis via Index Prediction}
\label{sec:consistency_analyisis_via_index_prediction}

\begin{figure*}[!htbp]
    \centering 
\begin{subfigure}{0.35\textwidth}
  \includegraphics[width=\linewidth]{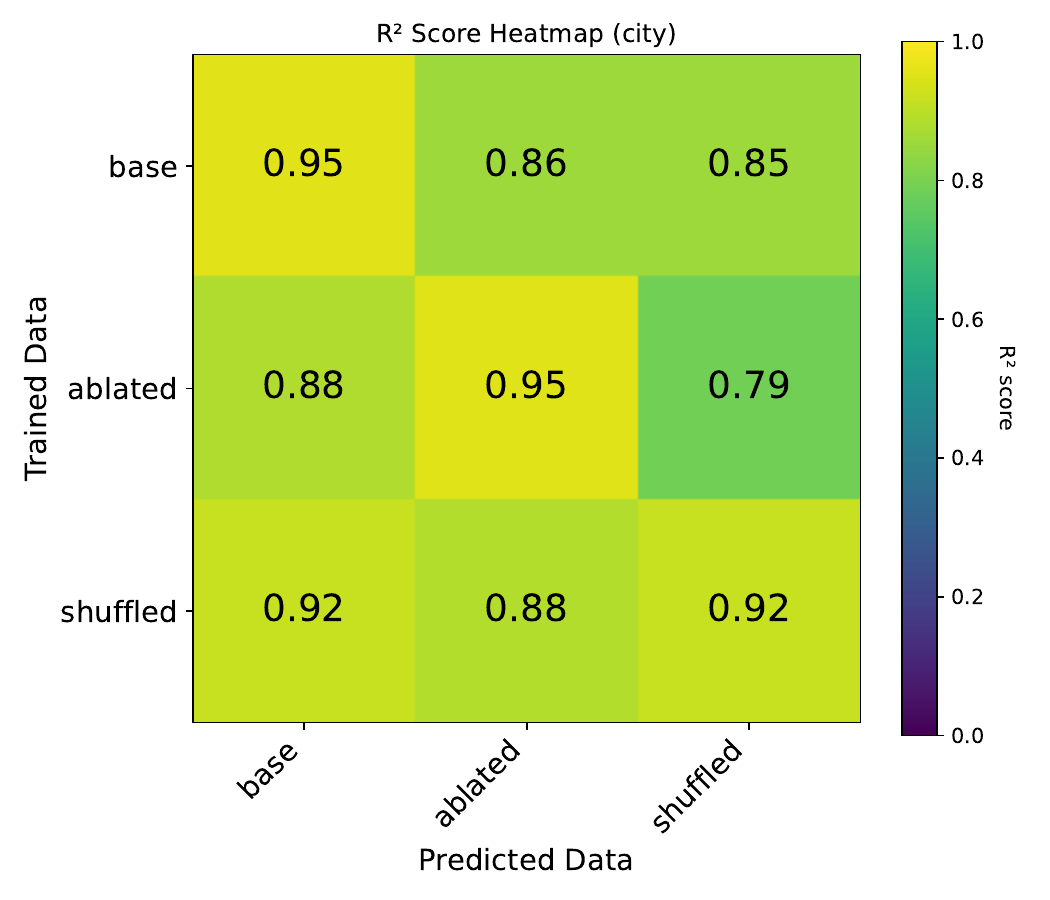}
  \caption{$C_{city}$}
\end{subfigure}\hfil 
\begin{subfigure}{0.35\textwidth}
  \includegraphics[width=\linewidth]{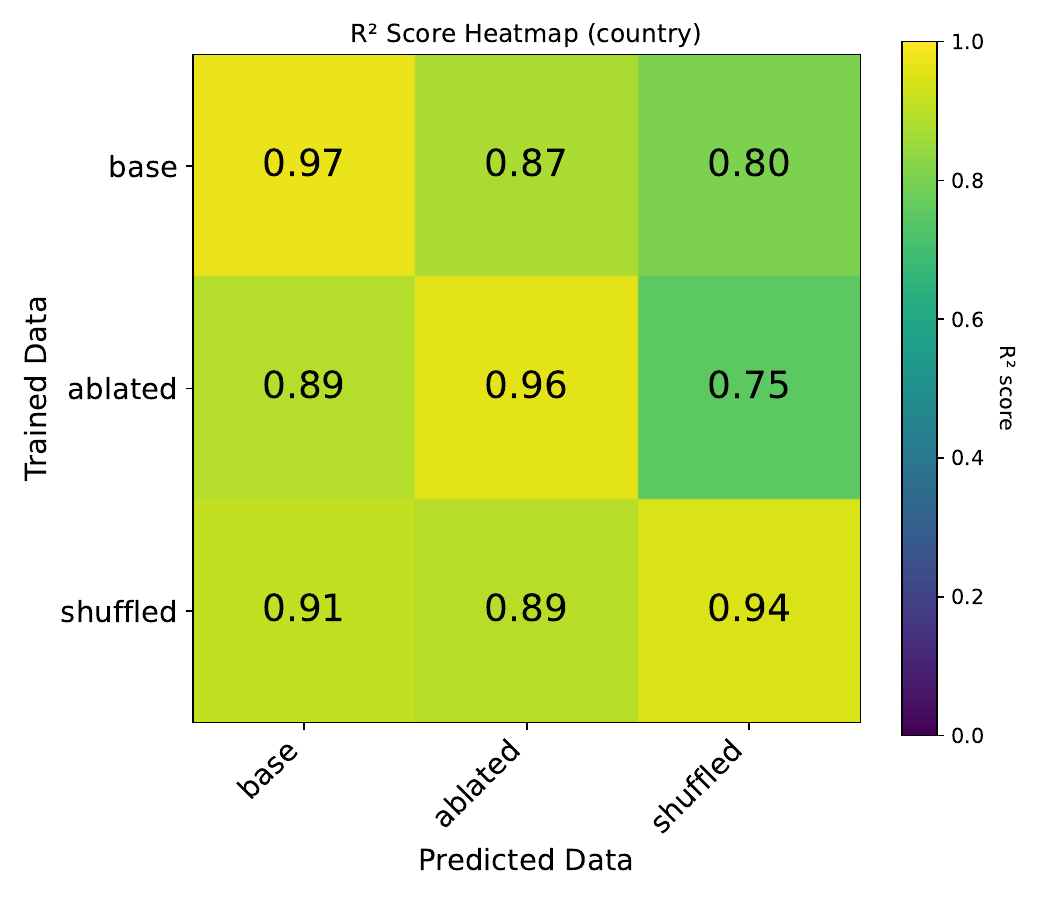}
  \caption{$C_{country}$}
\end{subfigure}\hfil 
\begin{subfigure}{0.3\textwidth}
  \includegraphics[width=\linewidth]{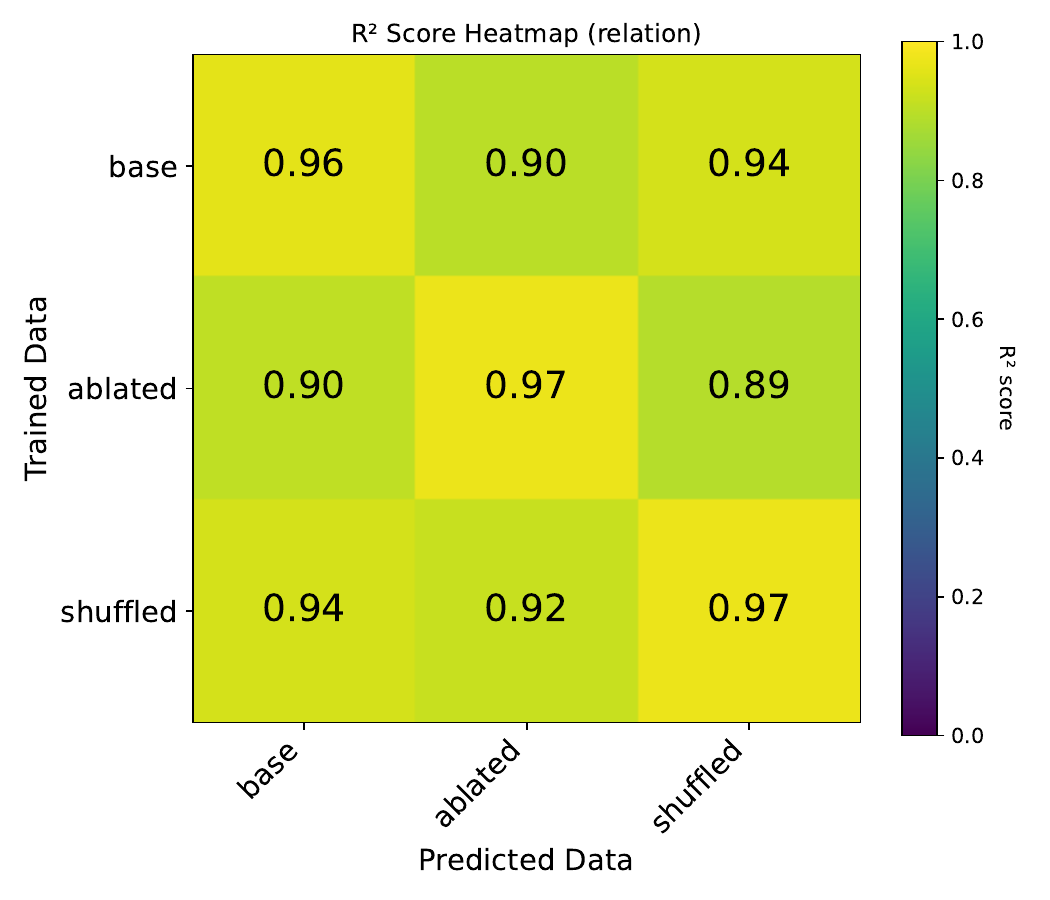}
  \caption{$C_{relation}$}
\end{subfigure}\hfil
\begin{subfigure}{0.3\textwidth}
  \includegraphics[width=\linewidth]{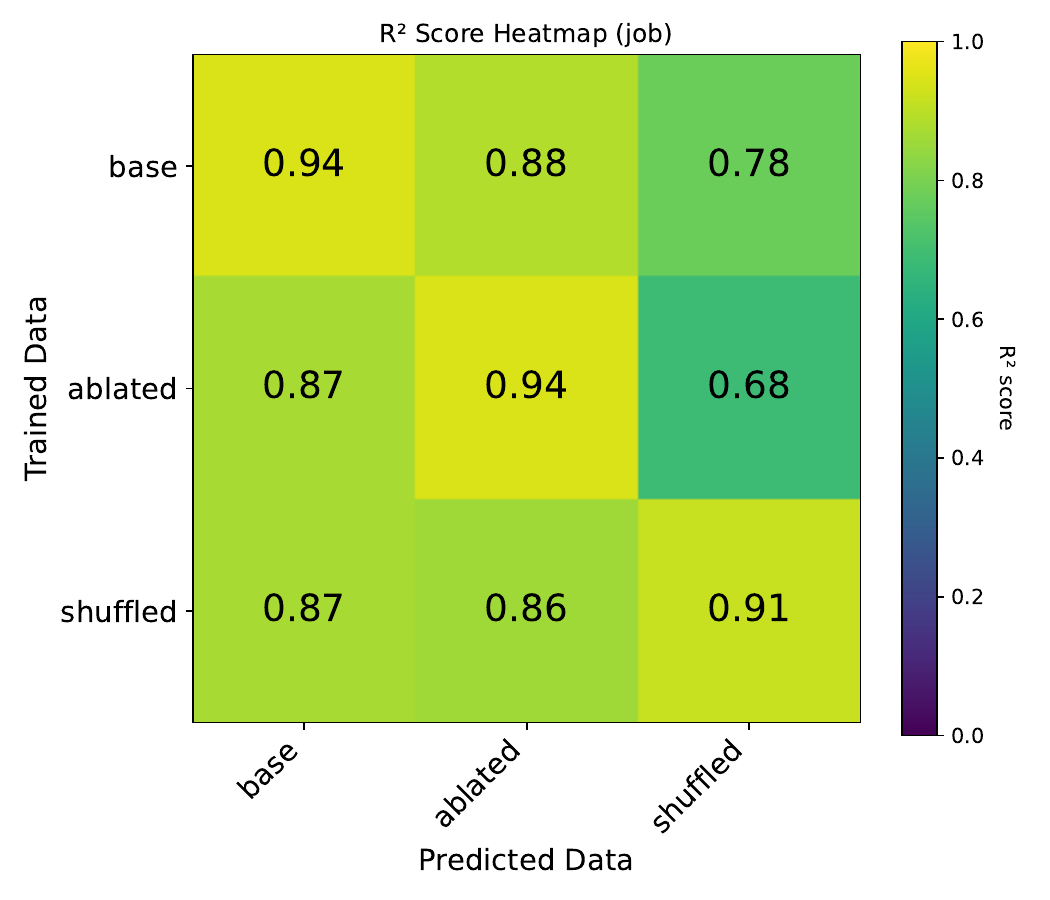}
  \caption{$C_{job}$}
\end{subfigure}\hfil 
\begin{subfigure}{0.3\textwidth}
  \includegraphics[width=\linewidth]{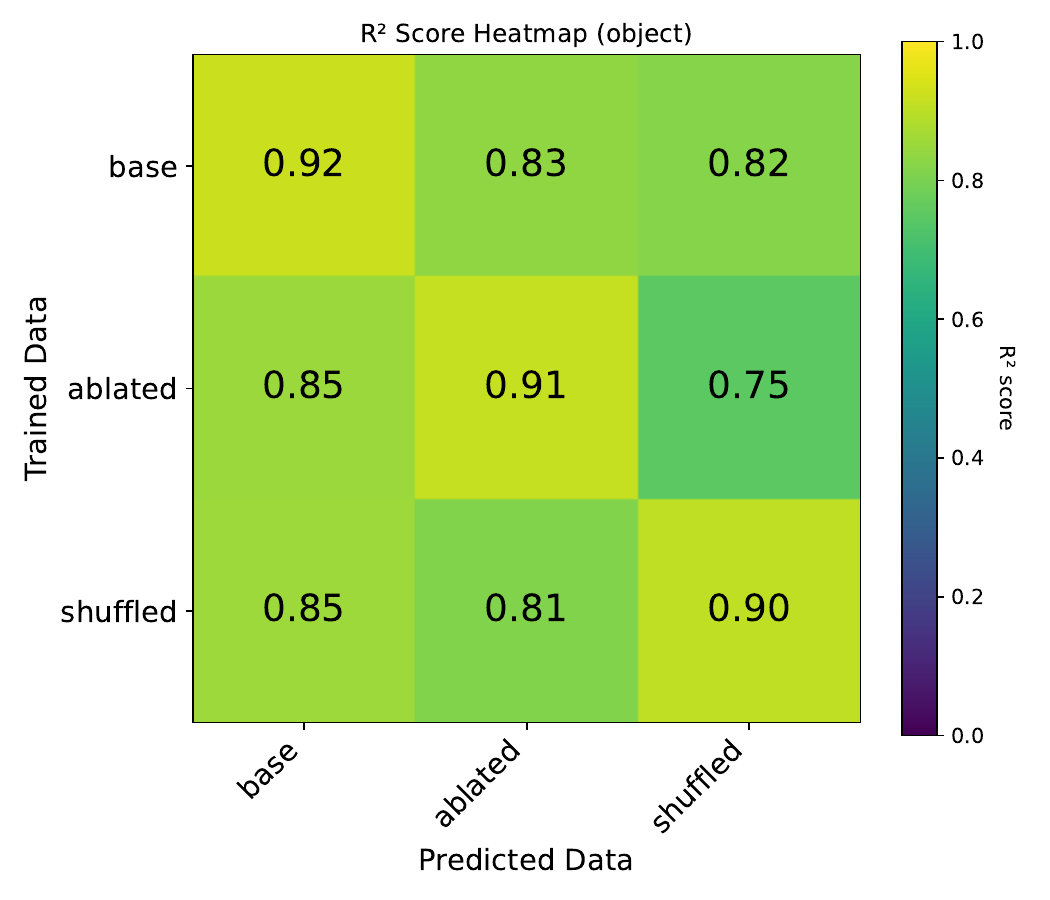}
  \caption{$C_{object}$}
\end{subfigure}\hfil 
\caption{Cross permutation $R^2$ scores for index prediction on Llama3-8B-Instruct.}
\label{fig:consistency_score_llama}
\end{figure*}
\begin{figure*}[!htbp]
    \centering 
\begin{subfigure}{0.35\textwidth}
  \includegraphics[width=\linewidth]{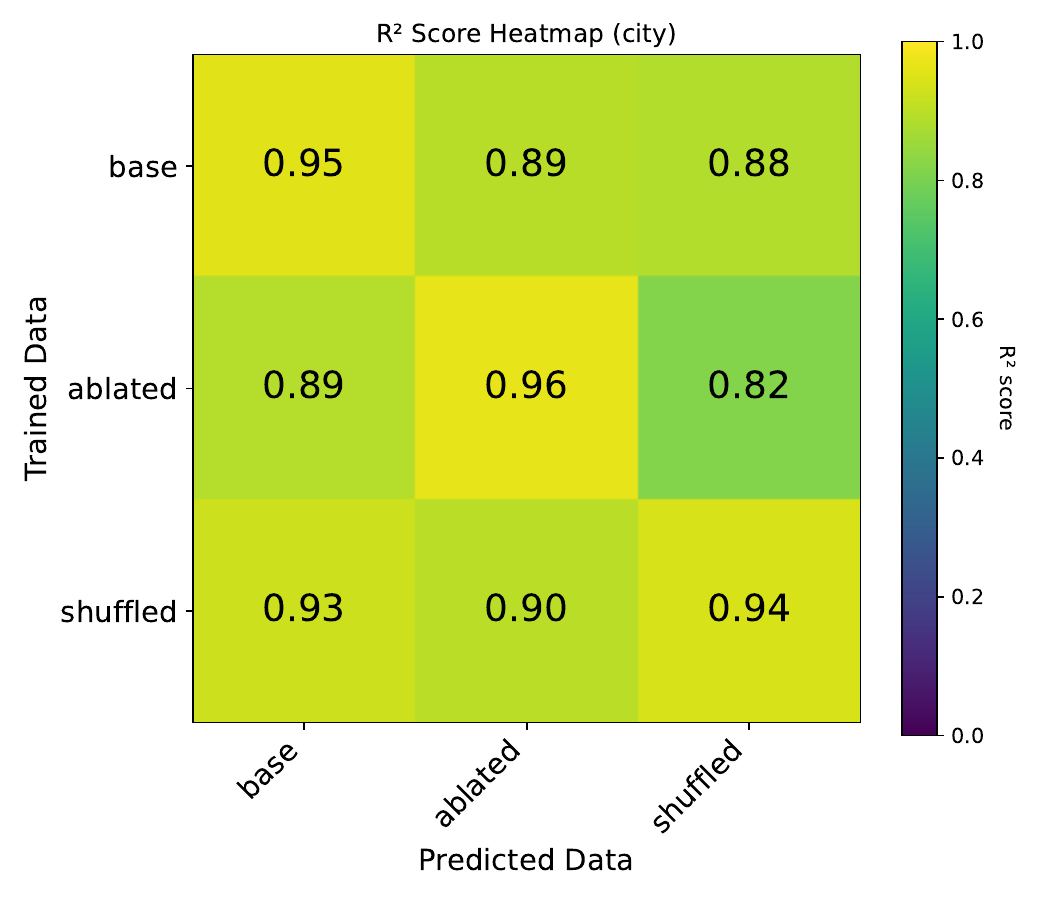}
  \caption{$C_{city}$}
\end{subfigure}\hfil 
\begin{subfigure}{0.35\textwidth}
  \includegraphics[width=\linewidth]{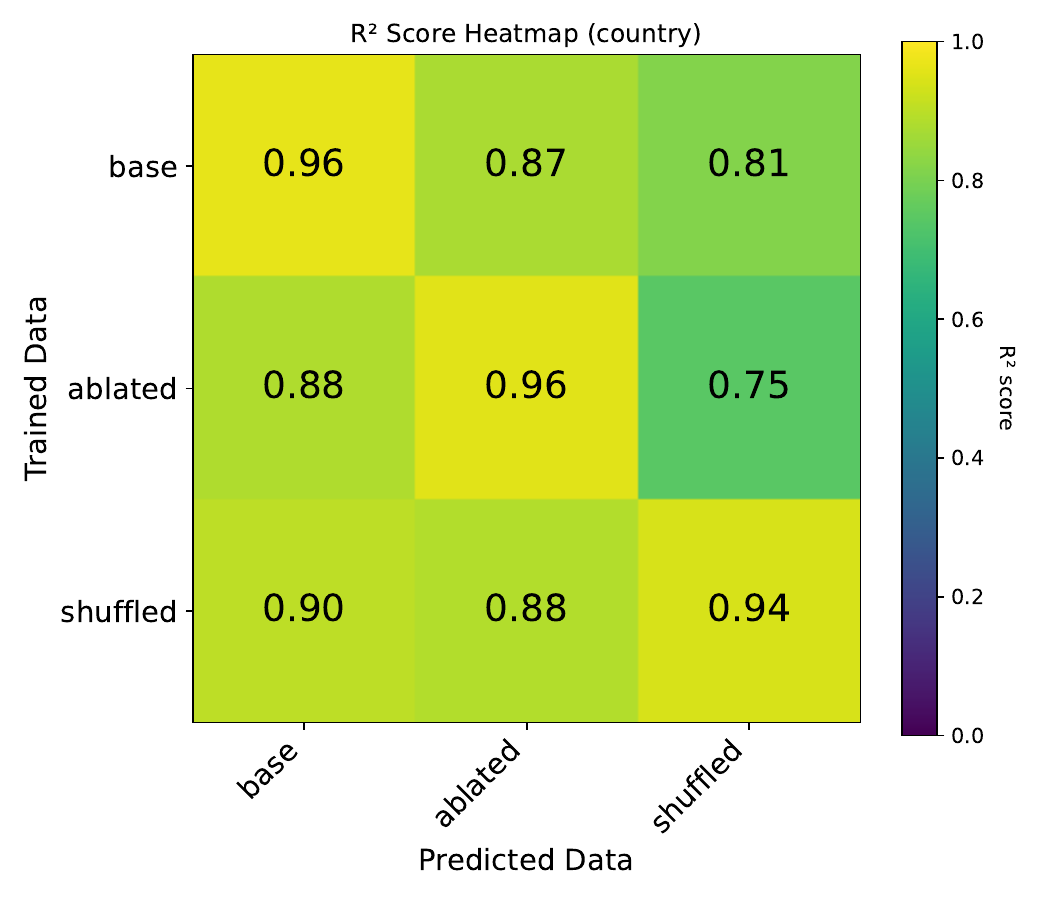}
  \caption{$C_{country}$}
\end{subfigure}\hfil 
\begin{subfigure}{0.3\textwidth}
  \includegraphics[width=\linewidth]{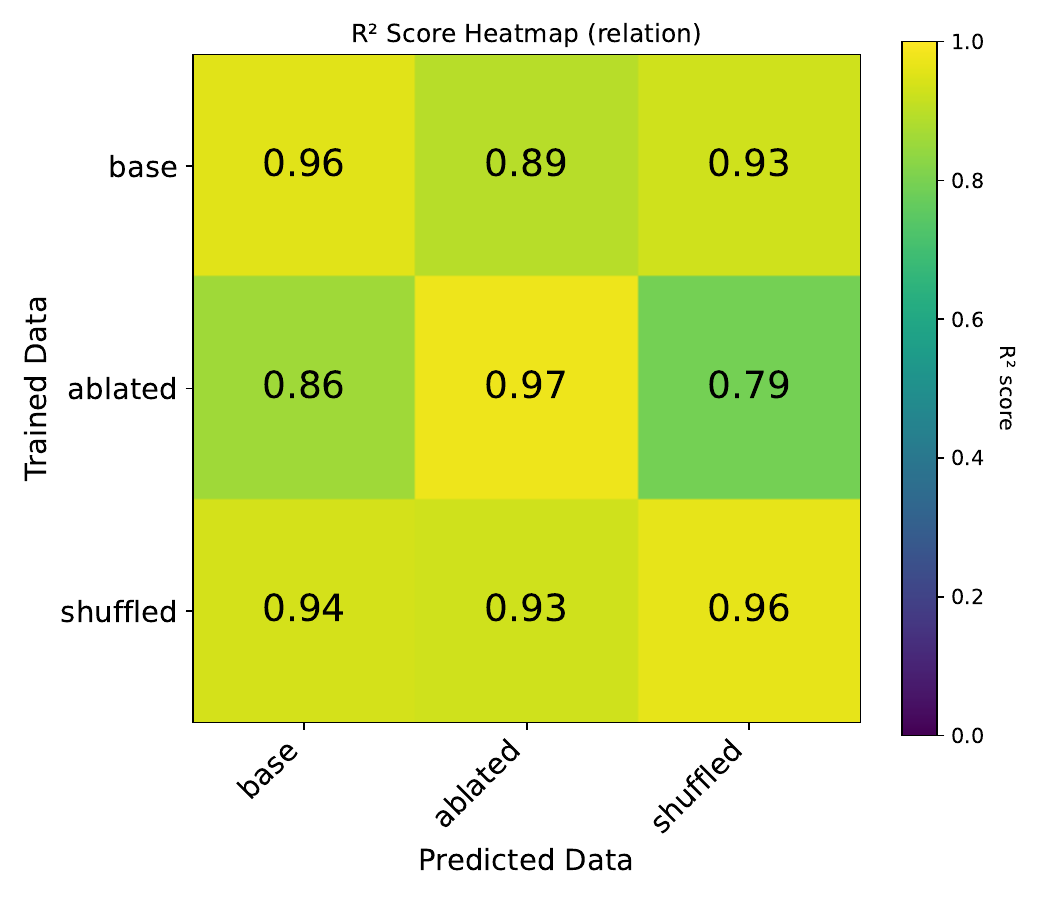}
  \caption{$C_{relation}$}
\end{subfigure}\hfil
\begin{subfigure}{0.3\textwidth}
  \includegraphics[width=\linewidth]{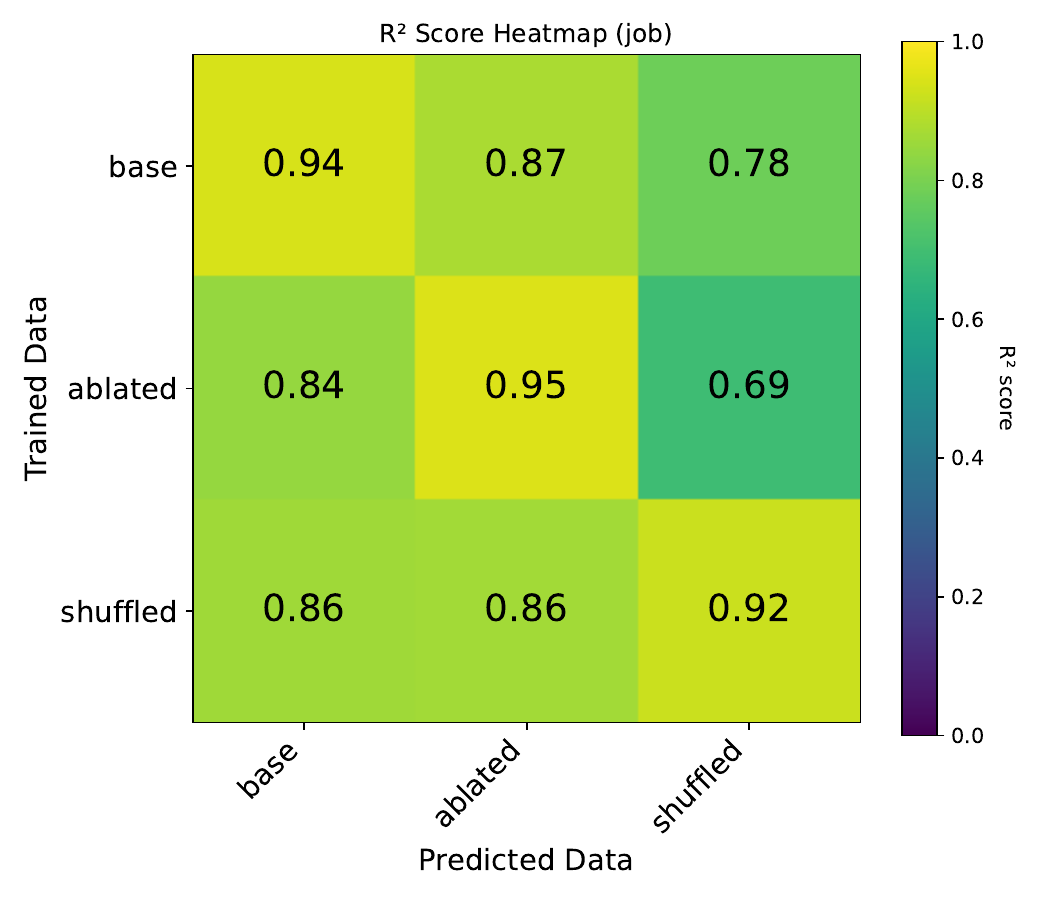}
  \caption{$C_{job}$}
\end{subfigure}\hfil 
\begin{subfigure}{0.3\textwidth}
  \includegraphics[width=\linewidth]{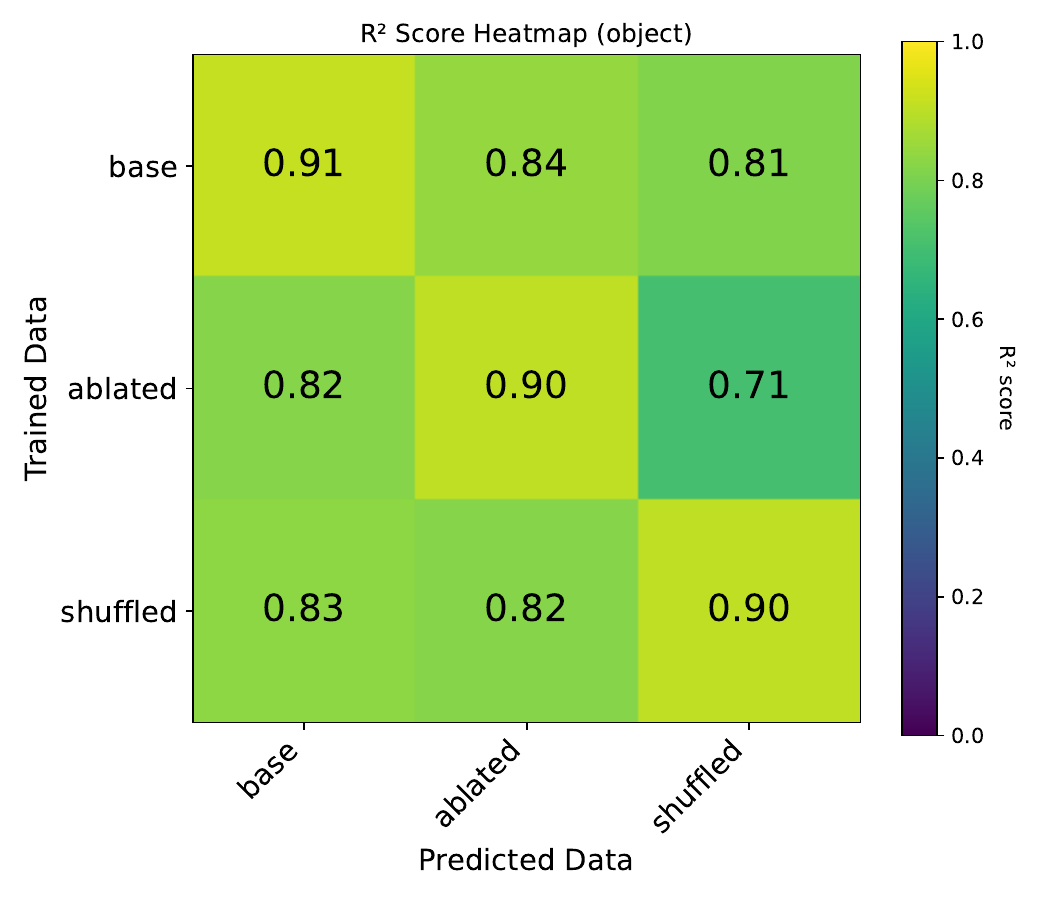}
  \caption{$C_{object}$}
\end{subfigure}\hfil 
\caption{Cross permutation $R^2$ scores for index prediction on Qwen3-8B.}
\label{fig:consistency_score_qwen}
\end{figure*}

We evaluate whether projection matrices learned under permuted conditions generalize to original setting. Specifically, we use CBR projection matrices learned from the ablated and shuffled datasets to predict index information in the original datasets, and vice versa, for all discourse contexts. As shown in Figure~\ref{fig:consistency_score_llama}, we observe consistently high predictive performance across contexts, with $R^2$ scores typically around $0.8$.

In addition, this strong performance is not limited to a single LLM family: Qwen3-8B also exhibit similarly high cross permutation prediction accuracy shown in Figure~\ref{fig:consistency_score_qwen}. These results demonstrate that the overall geometry of CBR subspace remains stable under both relation ablation and relation shuffling across Llama3-8B-Instruct and Qwen3-8B. This provides further evidence that the CBR subspace primarily reflects relational structure rather than a surface artifact of a particular dataset configuration.

\subsection{Effect of Entity Separation on the CBR Subspace}
\label{sec:irs_jump}

To further analyze the relationship between the CBR subspace and input format variations that reflect diverse discourse patterns observed in real-world text, we also construct additional modified datasets. In the setting (denoted as \textbf{separation}), we vary the distance between entity mentions without altering the CBR structure. For example, in Sample~\ref{ex:jump_1}, the entity ``window'' is re-mentioned only after the discourse introduces one attribute (denoted as \textbf{\# separation=1}) for other entities such as ``glass'' and ``keyboard'' in contrast to earlier Sample in Figure~\ref{fig:mechanism} (a), where ``table'' is referenced immediately. This modification captures variation in reference distance, a common phenomenon in natural discourse.

\begin{exe}
    \ex\label{ex:jump_1}\textbf{\# separation=1}: \textit{The \underline{window} is crafted in Australia, while the glass is produced in Mexico. The keyboard finds its origins in China. Meanwhile, \textbf{the \underline{window} is designed in Germany and exported to Brazil, yet it faces a ban in Spain. The glass is designed in Argentina, exported to Italy, but is also banned in Georgia. The keyboard is designed in France, sent to Russia, but prohibited in Sweden.} Together, these products tell a tale of global manufacturing, design, and the complexities of international trade regulations.}
    \ex\label{ex:jump_2}\textbf{\# separation=2}: \textit{The \underline{book}, crafted in Israel and designed in Jordan, captivates readers with its unique story. The paper, produced in Argentina and designed in Canada, adds a special touch to the pages. Meanwhile, the ball, made in France and designed in Brazil, brings joy to children everywhere. \textbf{The \underline{book} finds its way to Sweden, but it faces a ban in Spain. The paper travels to India, yet it is prohibited in Japan. The ball is exported to Germany, but it is banned in Italy,} creating a web of intrigue around these beloved items.}
    \ex\label{ex:jump_3}\textbf{\# separation=3}: \textit{The \underline{basket}, crafted in Germany and designed in Iraq, finds its way to Brazil. Meanwhile, the flower, produced in Iran and conceptualized in France, is sent to Turkey. The monitor, made in Japan and styled in Italy, is dispatched to Egypt. However, \textbf{the journey of the \underline{basket} comes to an abrupt halt as it is banned in Jordan. The flower faces a similar fate, being banned in Pakistan, while the monitor is restricted and banned in Canada,} leaving them all unable to reach certain markets.}
\end{exe}

\begin{figure*}[!htbp]
    \centering 
\begin{subfigure}{0.25\textwidth}
  \includegraphics[width=\linewidth]{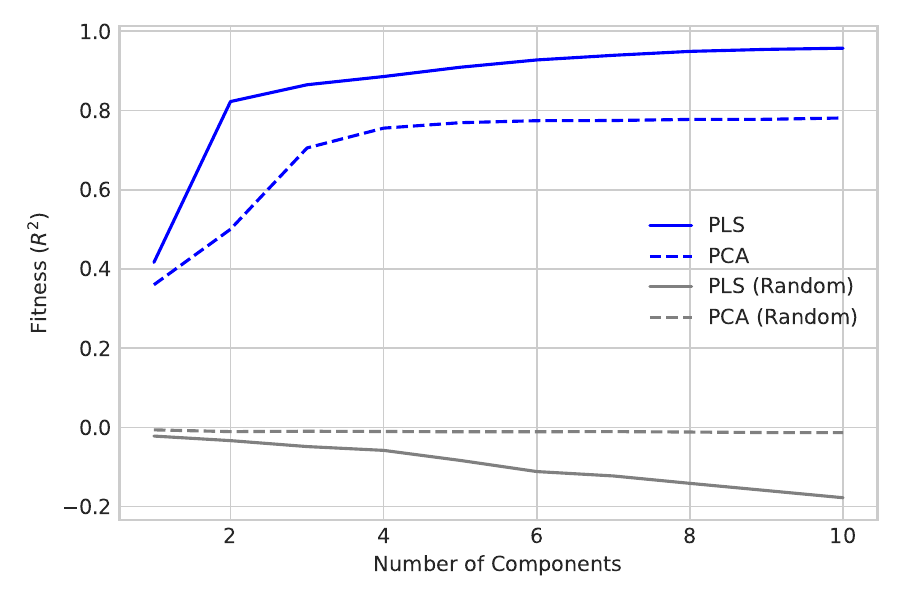}
  \caption{$C_{city}$}
\end{subfigure}\hfil 
\begin{subfigure}{0.25\textwidth}
  \includegraphics[width=\linewidth]{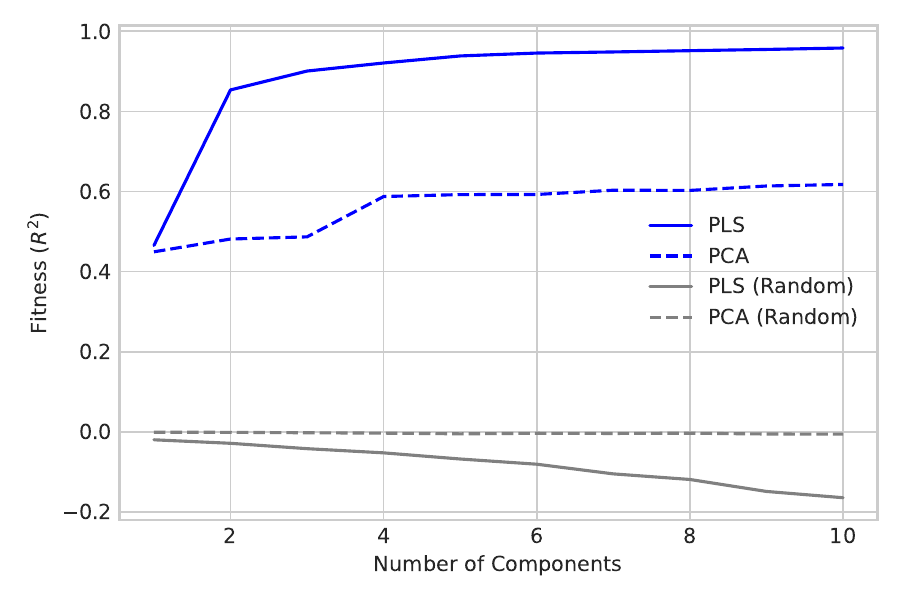}
  \caption{$C_{country}$}
\end{subfigure}\hfil 
\begin{subfigure}{0.25\textwidth}
  \includegraphics[width=\linewidth]{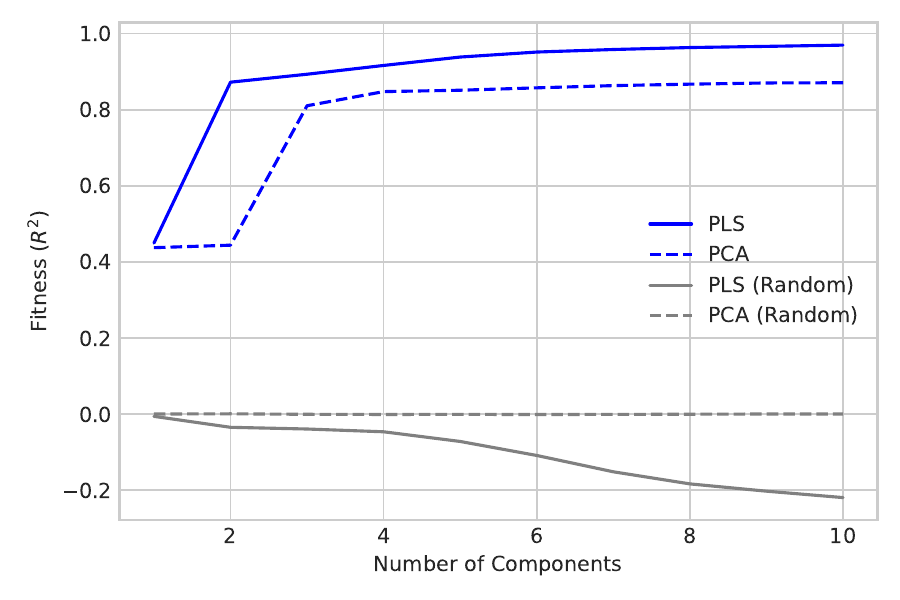}
  \caption{$C_{relation}$}
\end{subfigure}\hfil
\begin{subfigure}{0.25\textwidth}
  \includegraphics[width=\linewidth]{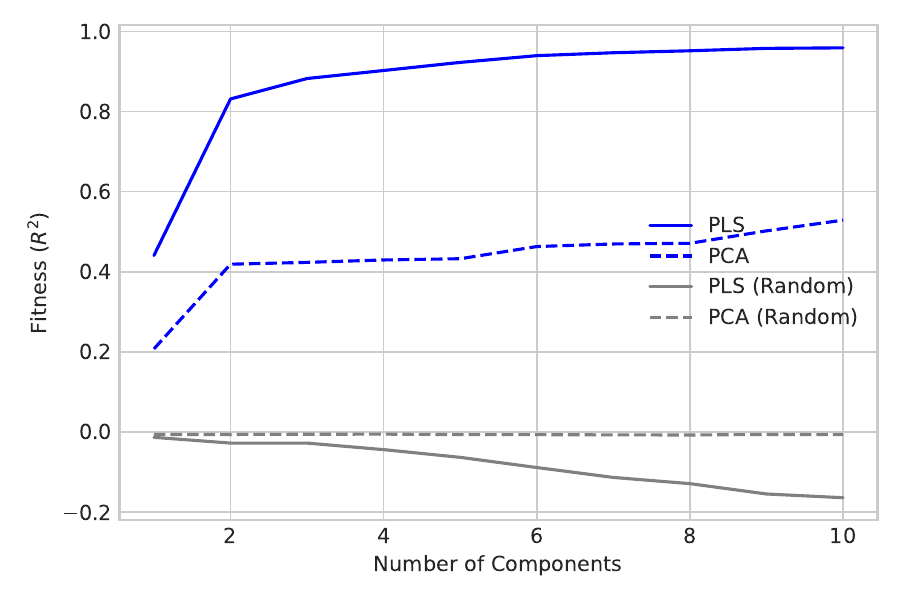}
  \caption{$C_{job}$}
\end{subfigure}\hfil 
\begin{subfigure}{0.25\textwidth}
  \includegraphics[width=\linewidth]{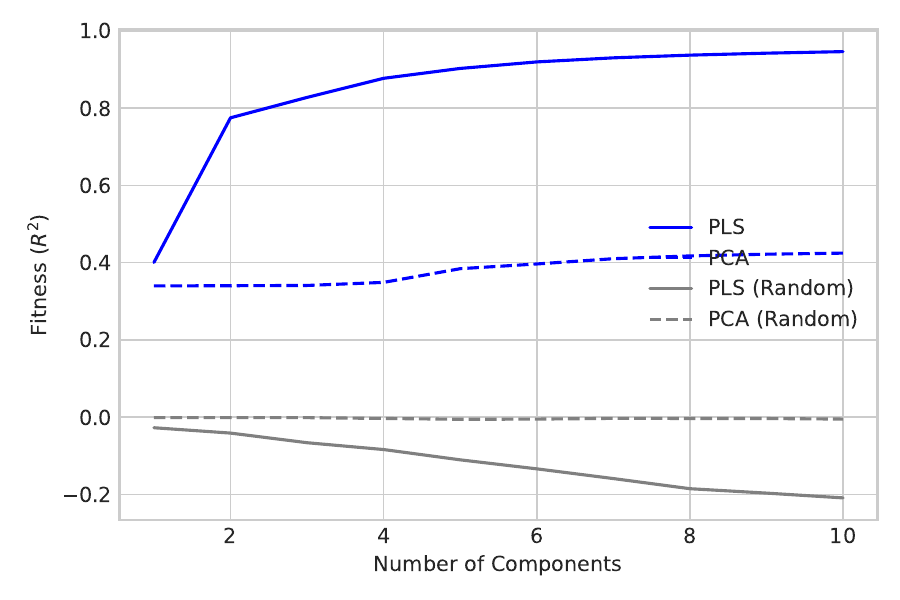}
  \caption{$C_{object}$}
\end{subfigure}\hfil 
\caption{Decoding performance of $[ei,ri]$ from activations of Llama3-8B-Instruct on the \textbf{\# separation=1} setting.}
\label{fig:jump_score_llama}
\end{figure*}

\begin{figure*}[!htbp]
    \centering 
\begin{subfigure}{0.25\textwidth}
  \includegraphics[width=\linewidth]{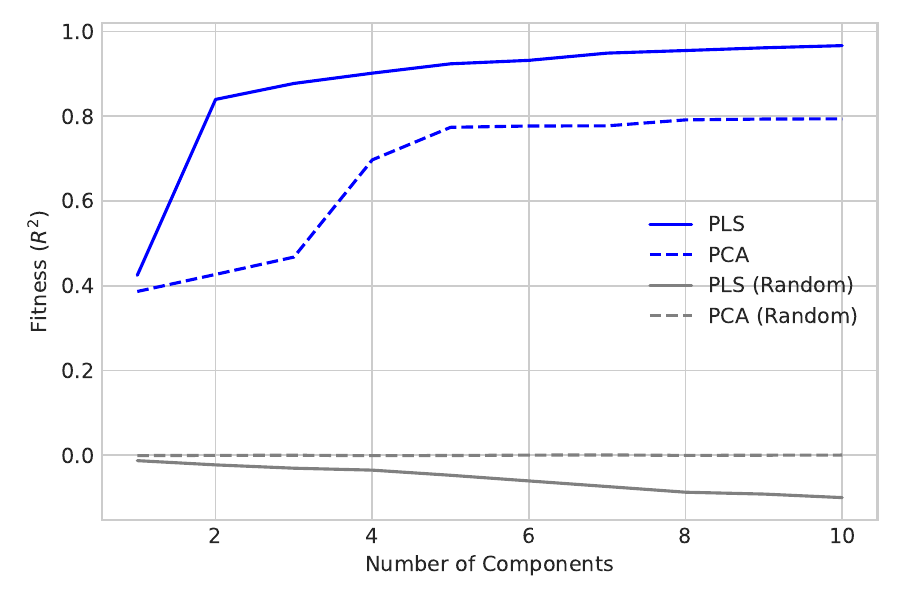}
  \caption{$C_{city}$}
\end{subfigure}\hfil 
\begin{subfigure}{0.25\textwidth}
  \includegraphics[width=\linewidth]{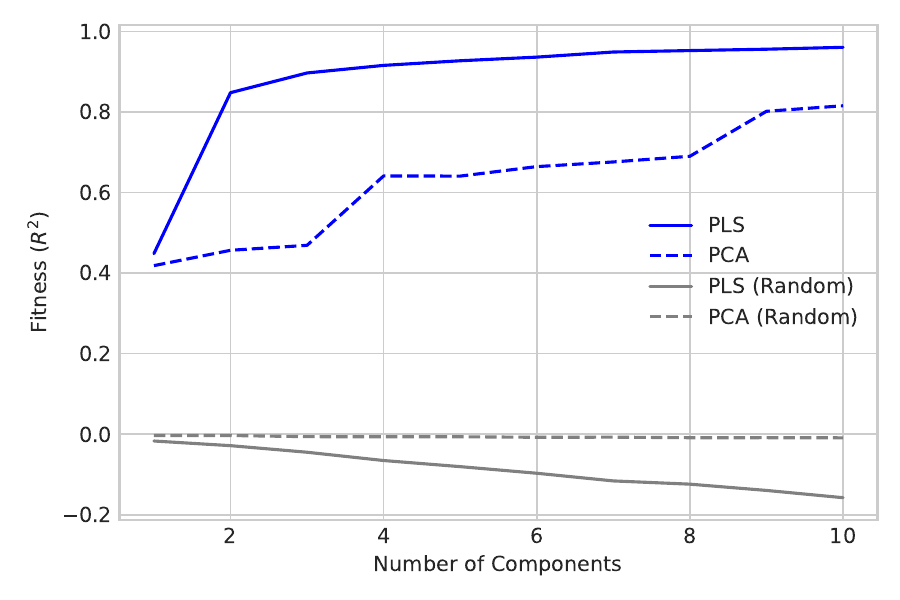}
  \caption{$C_{country}$}
\end{subfigure}\hfil 
\begin{subfigure}{0.25\textwidth}
  \includegraphics[width=\linewidth]{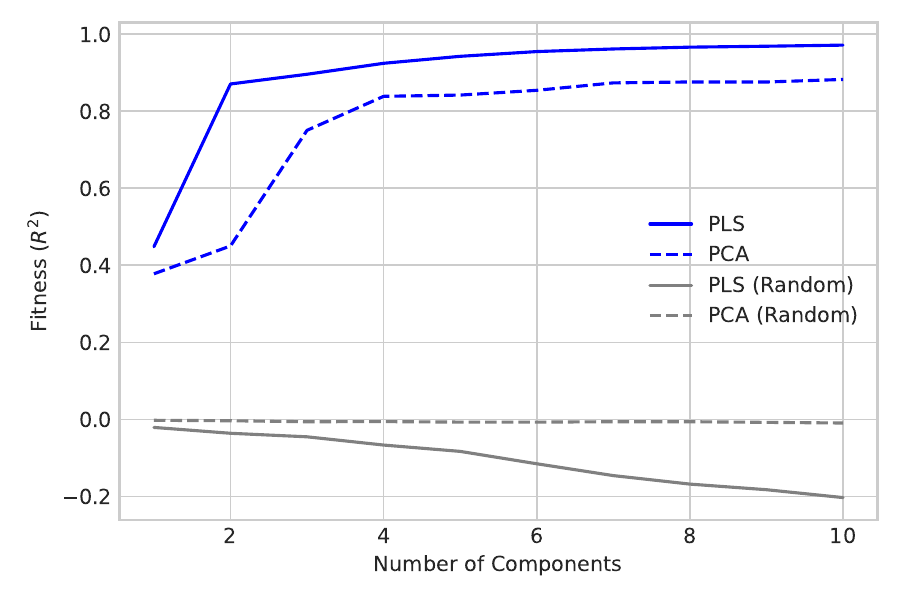}
  \caption{$C_{relation}$}
\end{subfigure}\hfil
\begin{subfigure}{0.25\textwidth}
  \includegraphics[width=\linewidth]{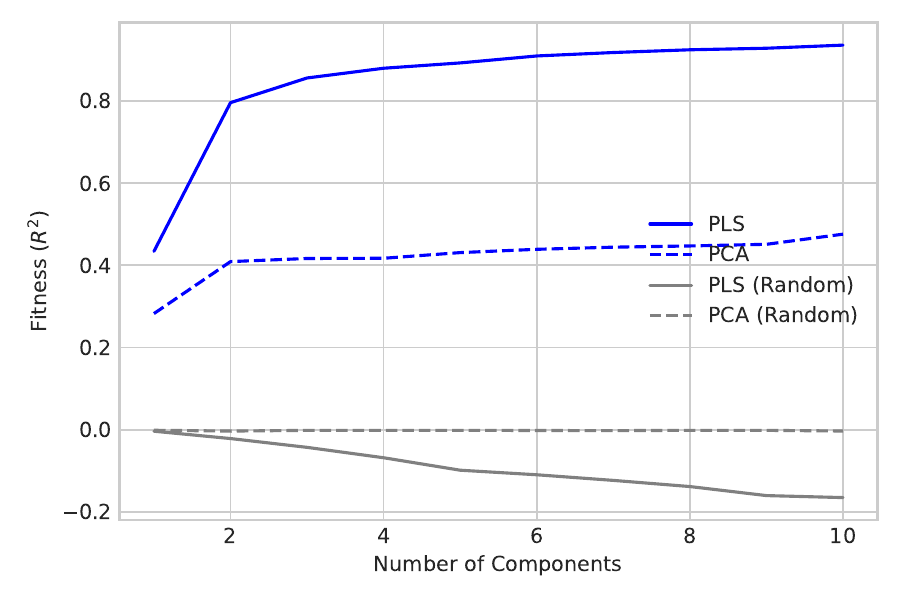}
  \caption{$C_{job}$}
\end{subfigure}\hfil 
\begin{subfigure}{0.25\textwidth}
  \includegraphics[width=\linewidth]{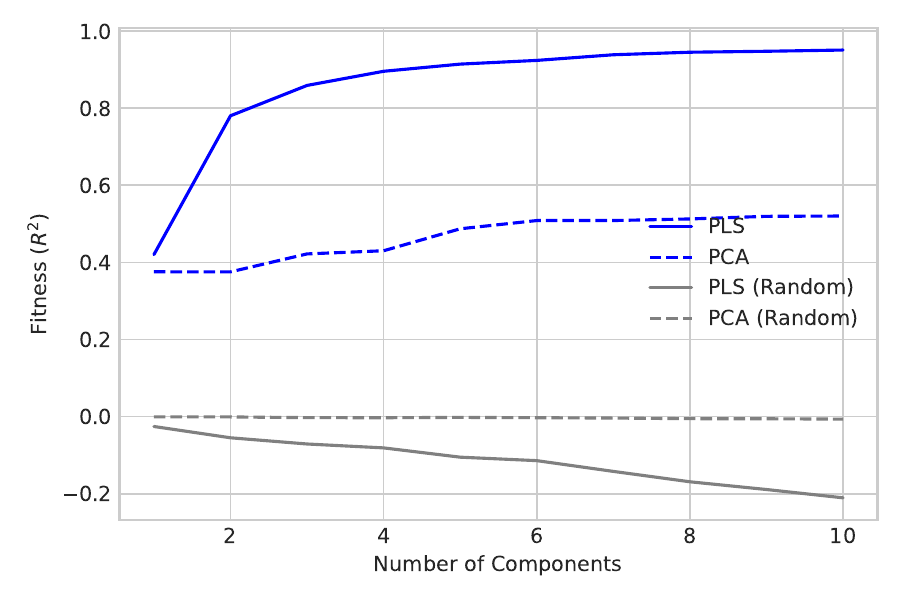}
  \caption{$C_{object}$}
\end{subfigure}\hfil 
\caption{Decoding performance of $[ei,ri]$ from activations of Llama3-8B-Instruct on the \textbf{\# separation=2} setting.}
\label{fig:jump2_score_llama}
\end{figure*}

\begin{figure*}[!htbp]
    \centering 
\begin{subfigure}{0.25\textwidth}
  \includegraphics[width=\linewidth]{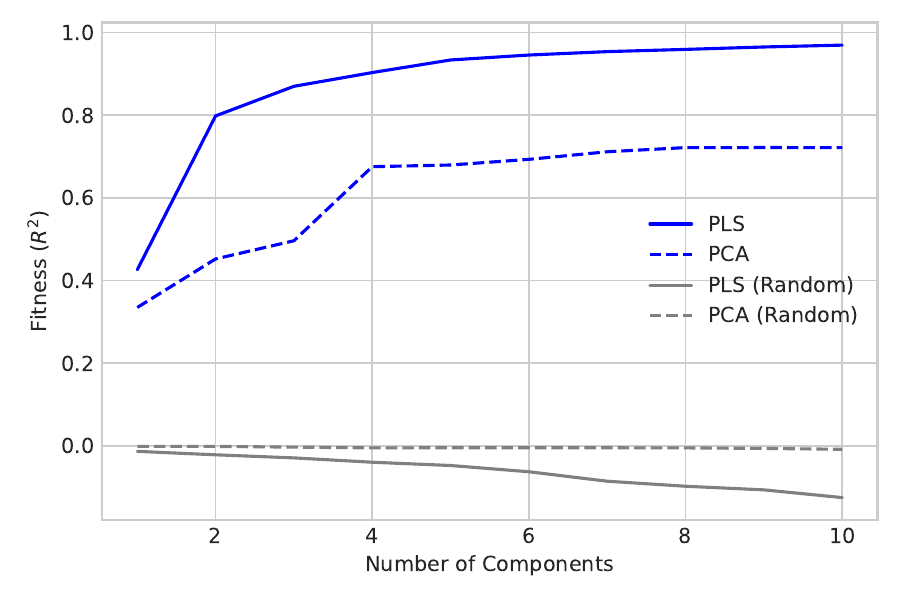}
  \caption{$C_{city}$}
\end{subfigure}\hfil 
\begin{subfigure}{0.25\textwidth}
  \includegraphics[width=\linewidth]{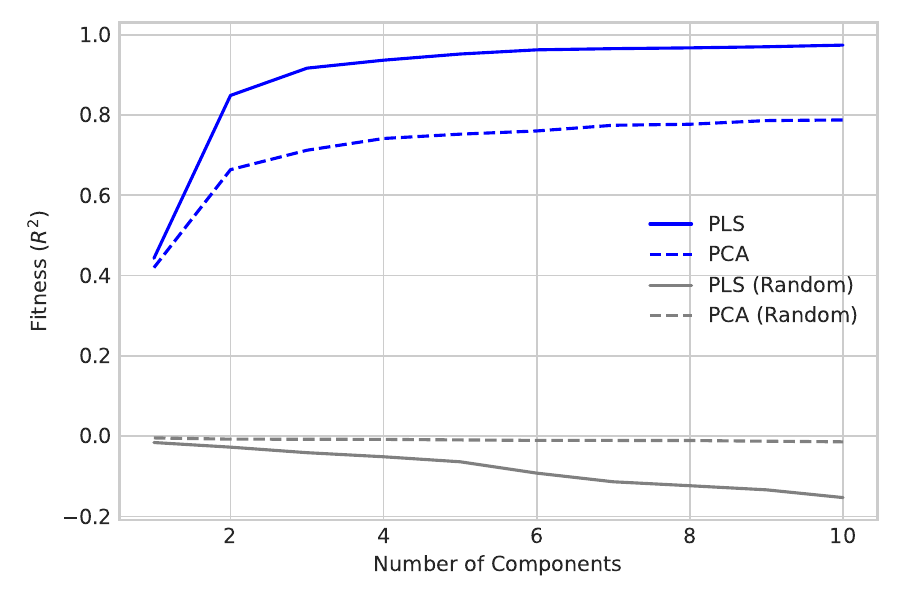}
  \caption{$C_{country}$}
\end{subfigure}\hfil 
\begin{subfigure}{0.25\textwidth}
  \includegraphics[width=\linewidth]{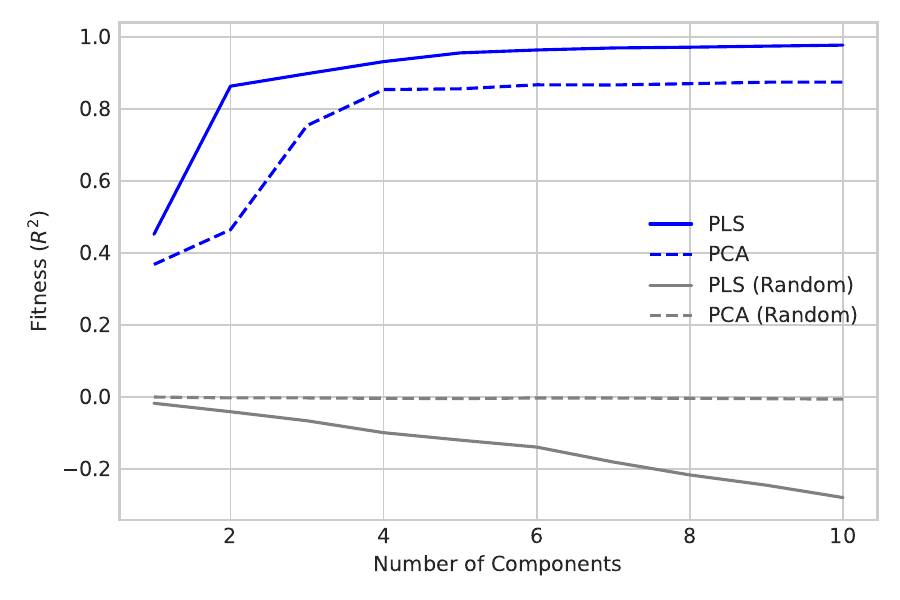}
  \caption{$C_{relation}$}
\end{subfigure}\hfil
\begin{subfigure}{0.25\textwidth}
  \includegraphics[width=\linewidth]{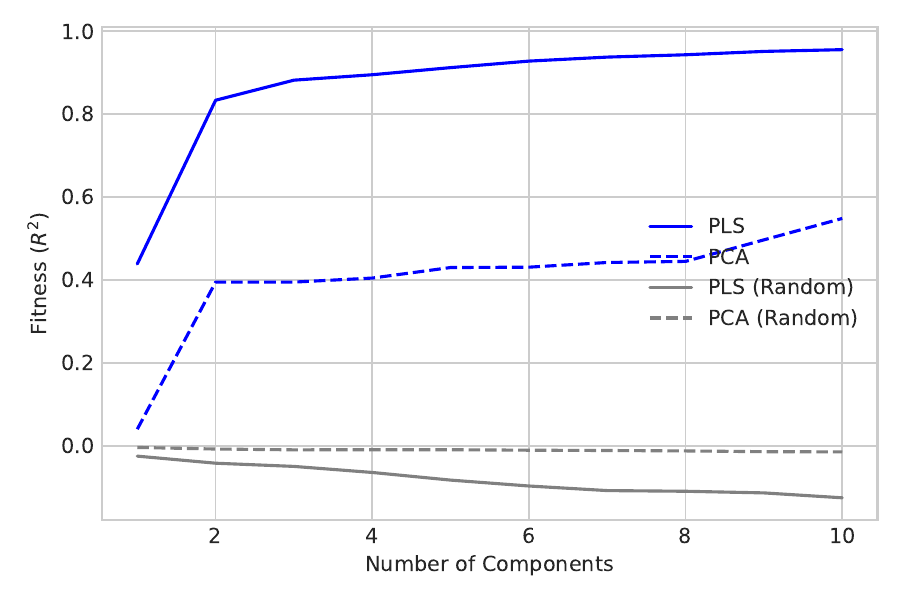}
  \caption{$C_{job}$}
\end{subfigure}\hfil 
\begin{subfigure}{0.25\textwidth}
  \includegraphics[width=\linewidth]{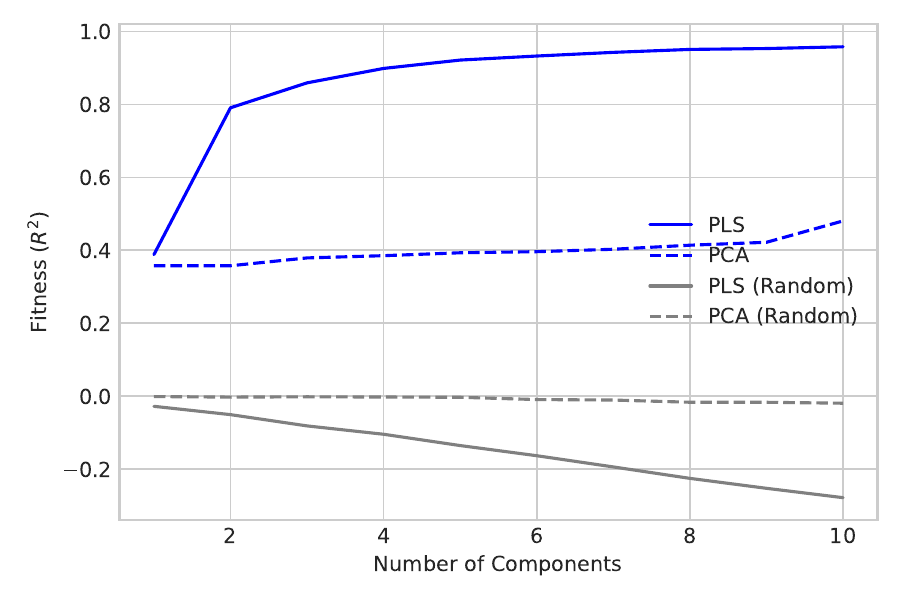}
  \caption{$C_{object}$}
\end{subfigure}\hfil 
\caption{Decoding performance of $[ei,ri]$ from activations of Llama3-8B-Instruct on the \textbf{\# separation=3} setting.}
\label{fig:jump3_score_llama}
\end{figure*}

\begin{figure*}[!htbp]
    \centering 
\begin{subfigure}{0.25\textwidth}
  \includegraphics[width=\linewidth]{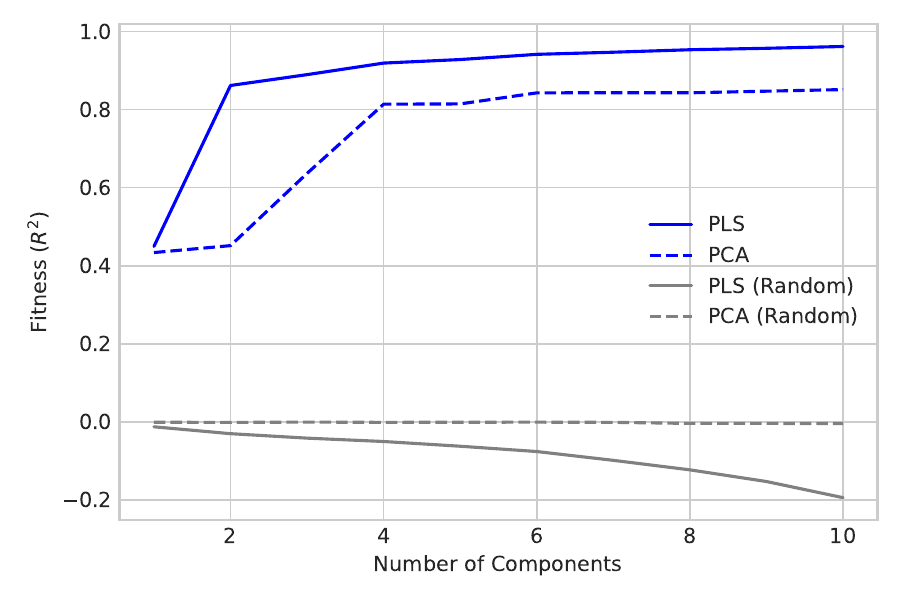}
  \caption{$C_{city}$}
\end{subfigure}\hfil 
\begin{subfigure}{0.25\textwidth}
  \includegraphics[width=\linewidth]{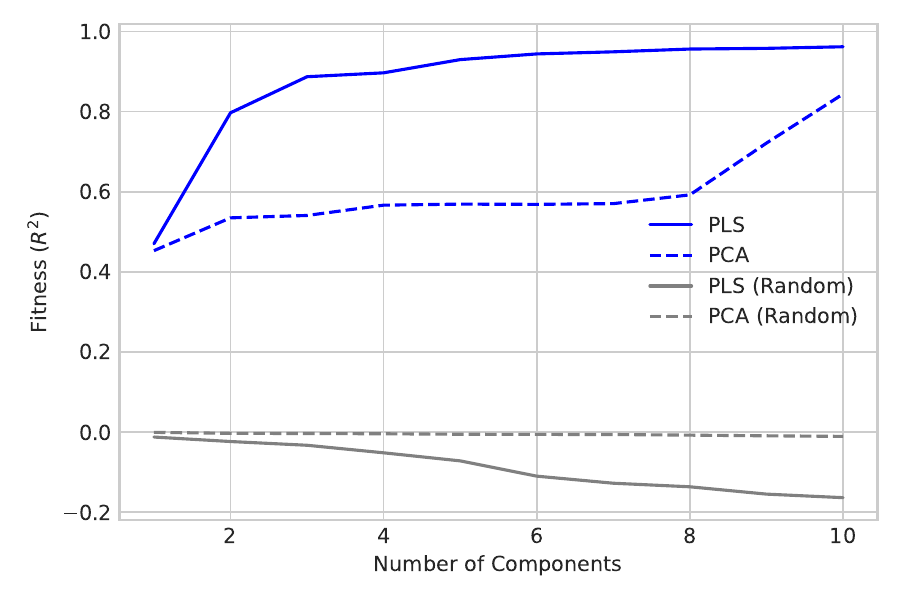}
  \caption{$C_{country}$}
\end{subfigure}\hfil 
\begin{subfigure}{0.25\textwidth}
  \includegraphics[width=\linewidth]{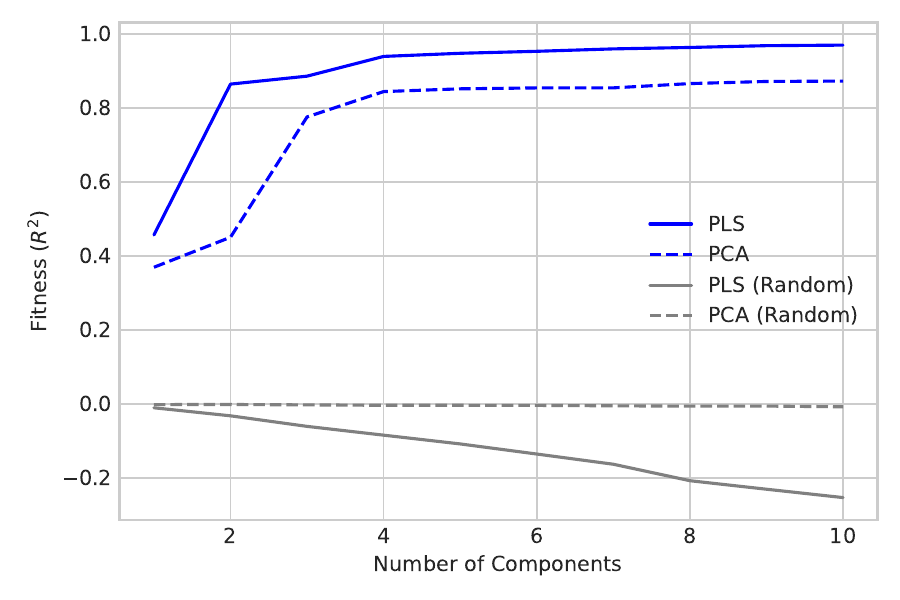}
  \caption{$C_{relation}$}
\end{subfigure}\hfil
\begin{subfigure}{0.25\textwidth}
  \includegraphics[width=\linewidth]{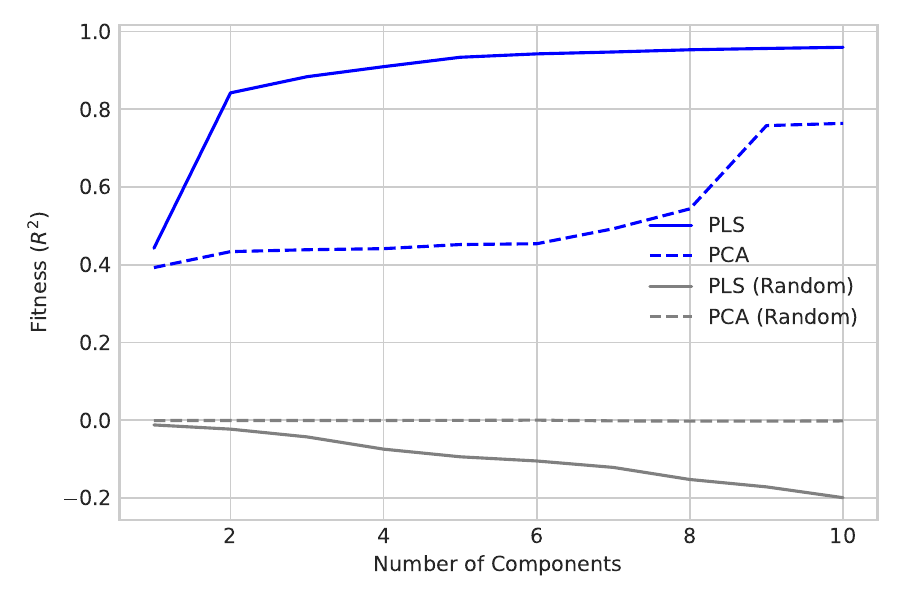}
  \caption{$C_{job}$}
\end{subfigure}\hfil 
\begin{subfigure}{0.25\textwidth}
  \includegraphics[width=\linewidth]{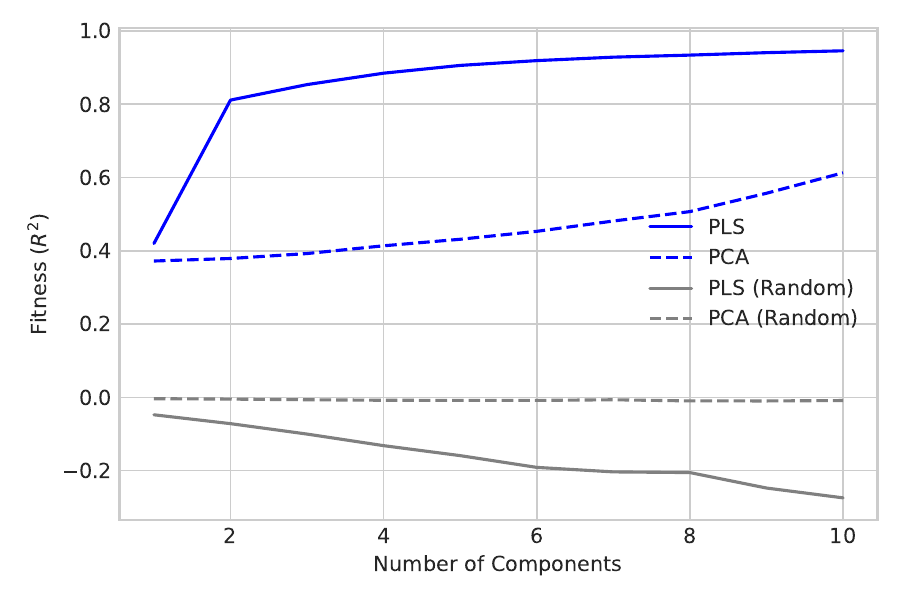}
  \caption{$C_{object}$}
\end{subfigure}\hfil 
\caption{Decoding performance of $[ei,ri]$ from activations of Qwen3-8B on the \textbf{\# separation=1} setting.}
\label{fig:jump_score_qwen}
\end{figure*}

\begin{figure*}[!htbp]
    \centering 
\begin{subfigure}{0.25\textwidth}
  \includegraphics[width=\linewidth]{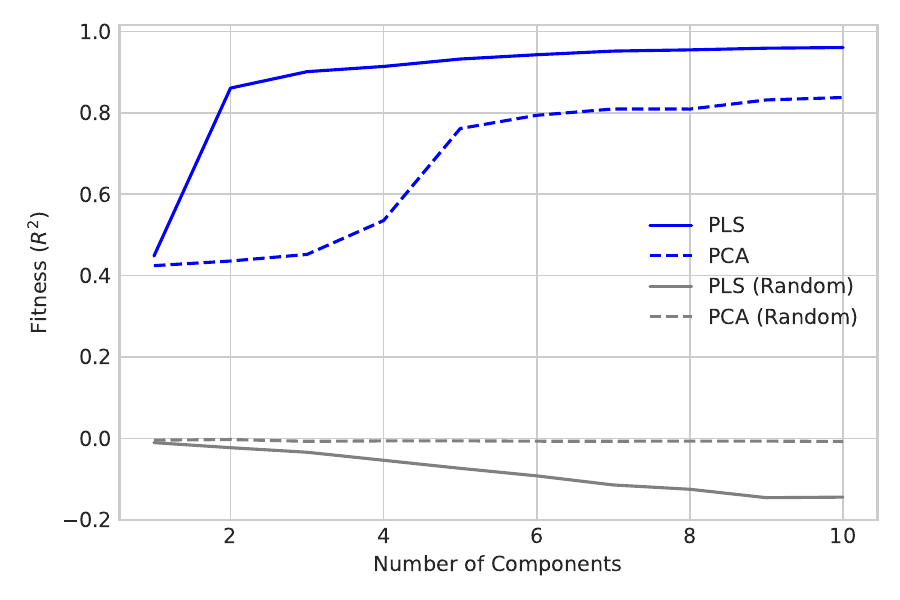}
  \caption{$C_{city}$}
\end{subfigure}\hfil 
\begin{subfigure}{0.25\textwidth}
  \includegraphics[width=\linewidth]{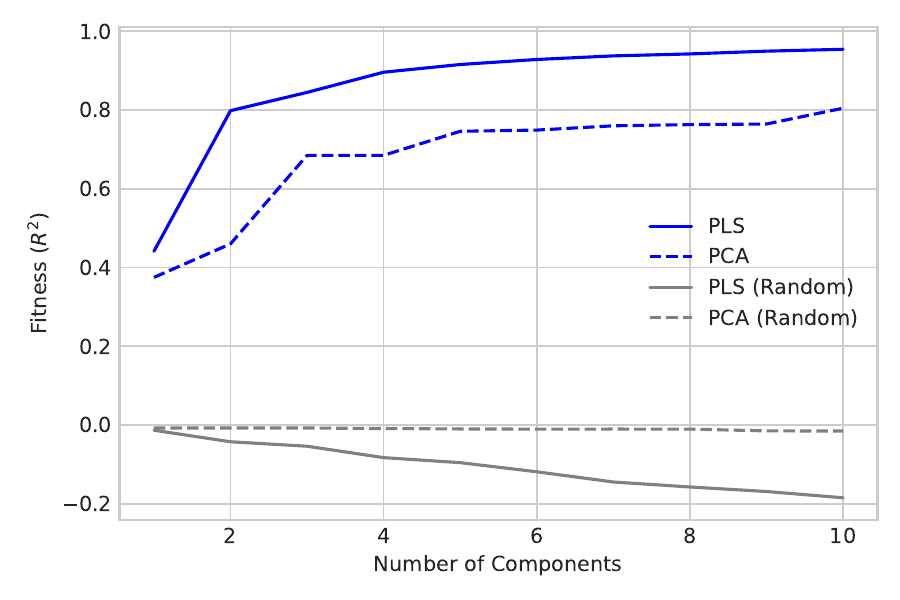}
  \caption{$C_{country}$}
\end{subfigure}\hfil 
\begin{subfigure}{0.25\textwidth}
  \includegraphics[width=\linewidth]{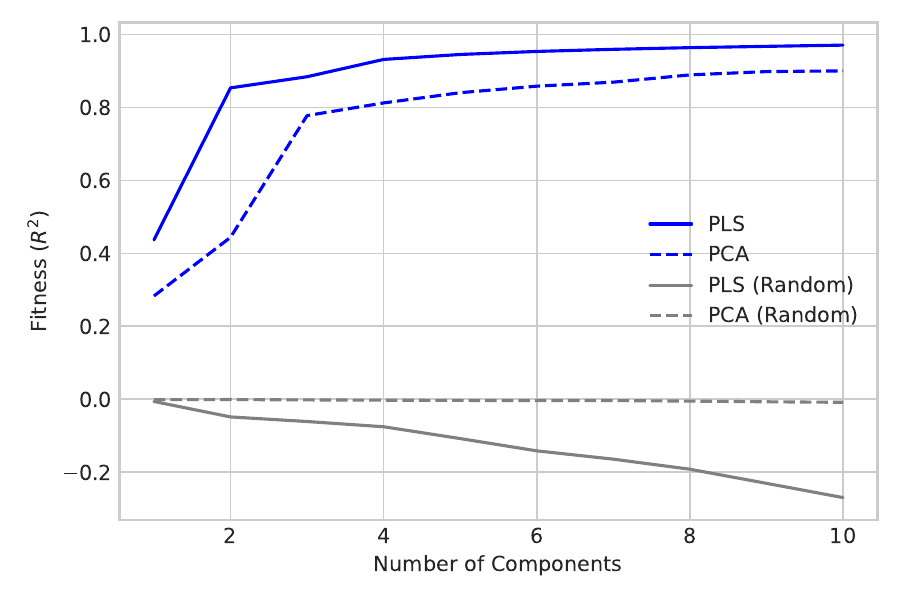}
  \caption{$C_{relation}$}
\end{subfigure}\hfil
\begin{subfigure}{0.25\textwidth}
  \includegraphics[width=\linewidth]{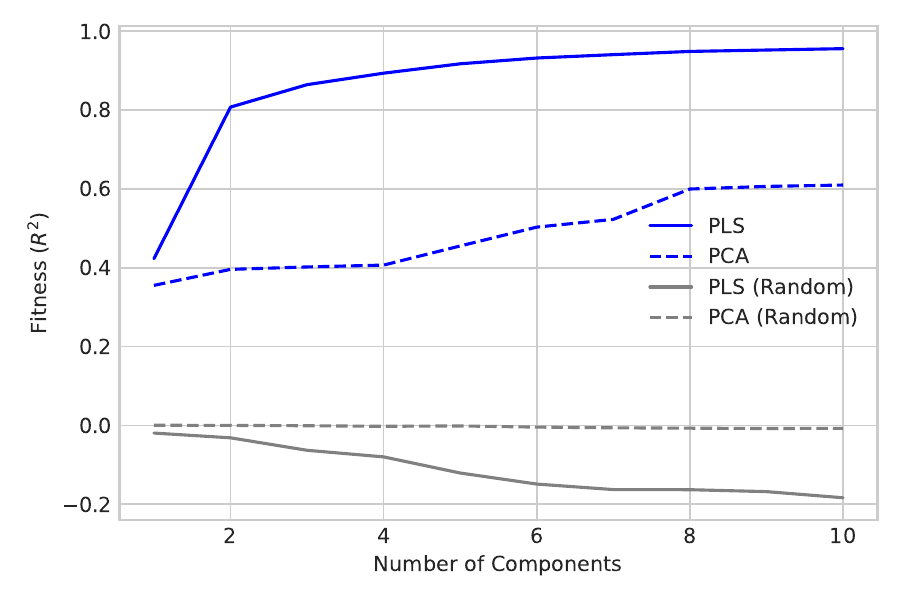}
  \caption{$C_{job}$}
\end{subfigure}\hfil 
\begin{subfigure}{0.25\textwidth}
  \includegraphics[width=\linewidth]{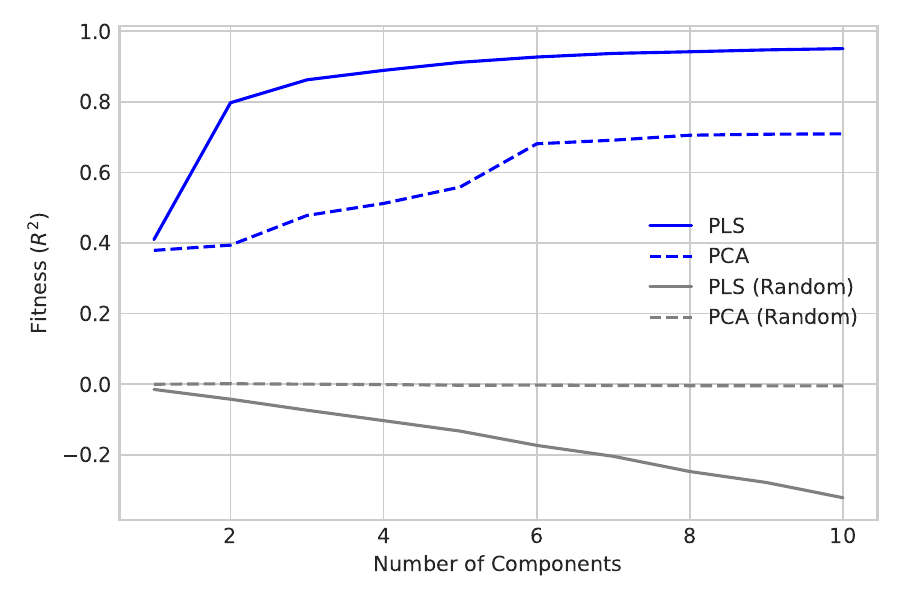}
  \caption{$C_{object}$}
\end{subfigure}\hfil 
\caption{Decoding performance of $[ei,ri]$ from activations of Qwen3-8B on the \textbf{\# separation=2} setting.}
\label{fig:jump2_score_qwen}
\end{figure*}

\begin{figure*}[!htbp]
    \centering 
\begin{subfigure}{0.25\textwidth}
  \includegraphics[width=\linewidth]{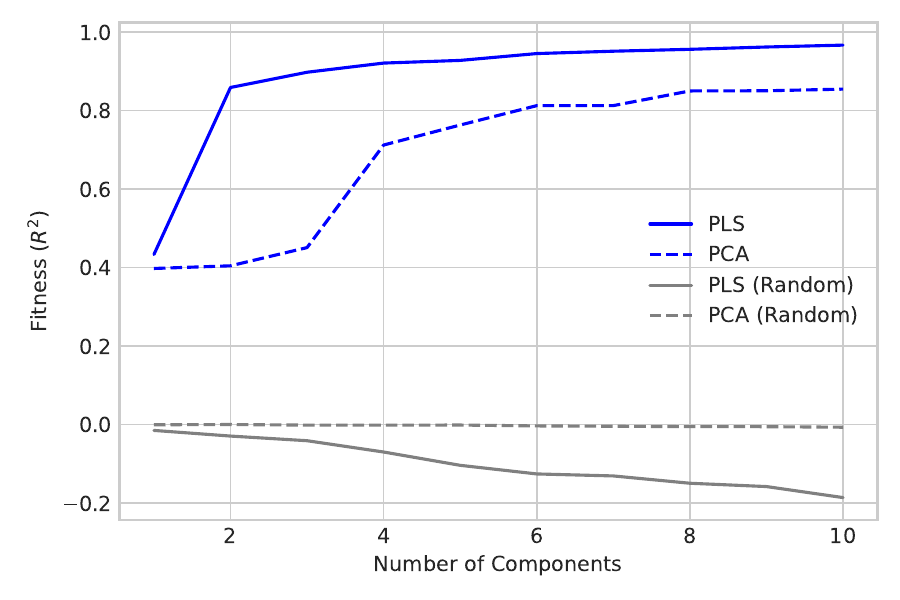}
  \caption{$C_{city}$}
\end{subfigure}\hfil 
\begin{subfigure}{0.25\textwidth}
  \includegraphics[width=\linewidth]{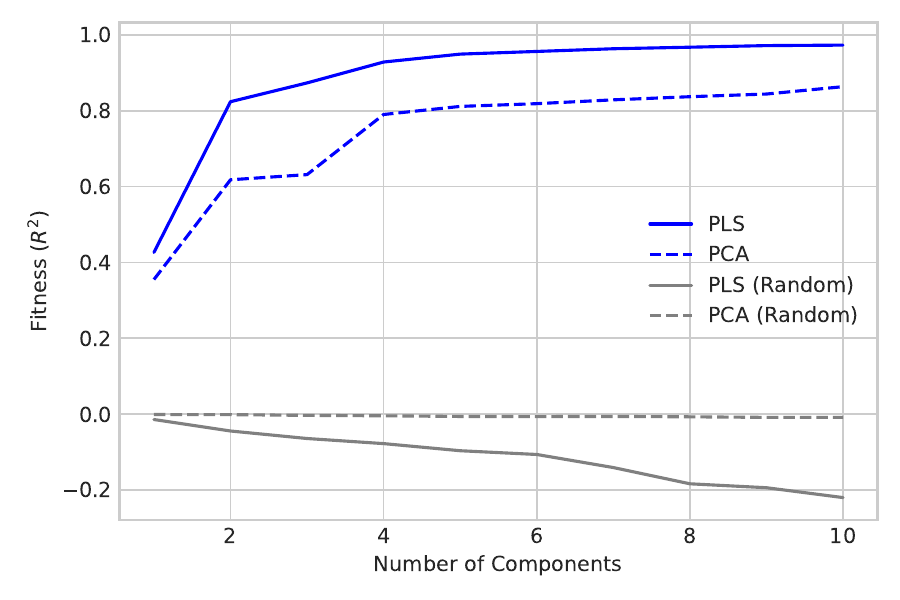}
  \caption{$C_{country}$}
\end{subfigure}\hfil 
\begin{subfigure}{0.25\textwidth}
  \includegraphics[width=\linewidth]{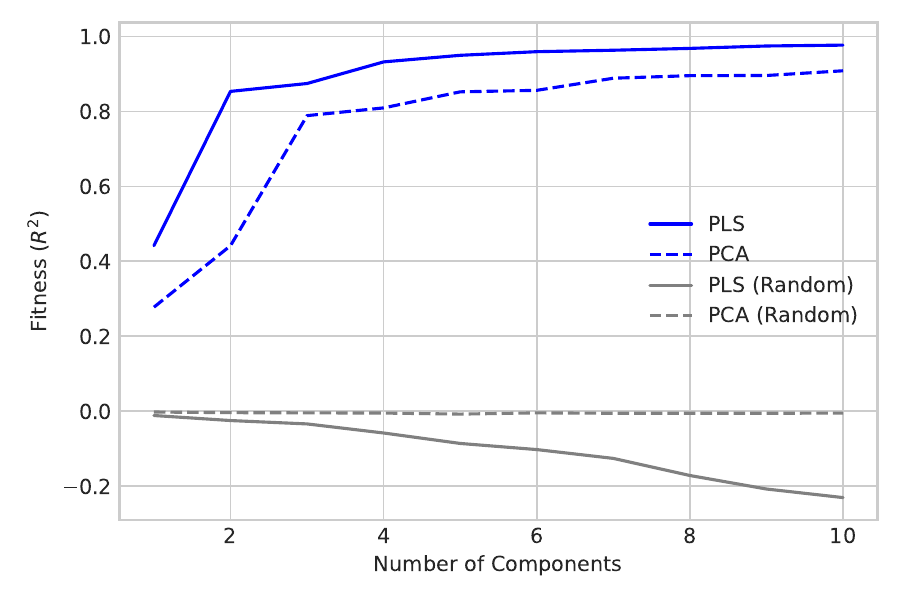}
  \caption{$C_{relation}$}
\end{subfigure}\hfil
\begin{subfigure}{0.25\textwidth}
  \includegraphics[width=\linewidth]{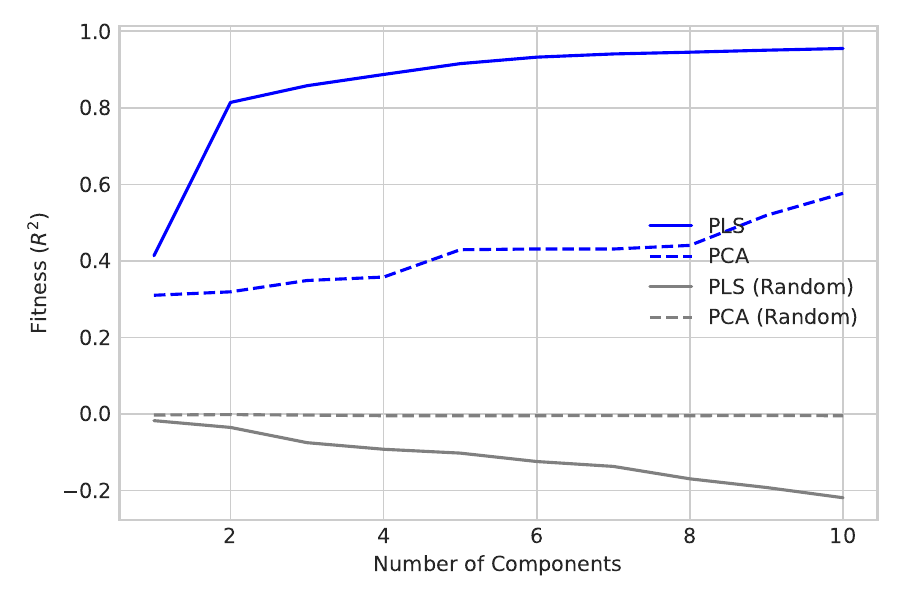}
  \caption{$C_{job}$}
\end{subfigure}\hfil 
\begin{subfigure}{0.25\textwidth}
  \includegraphics[width=\linewidth]{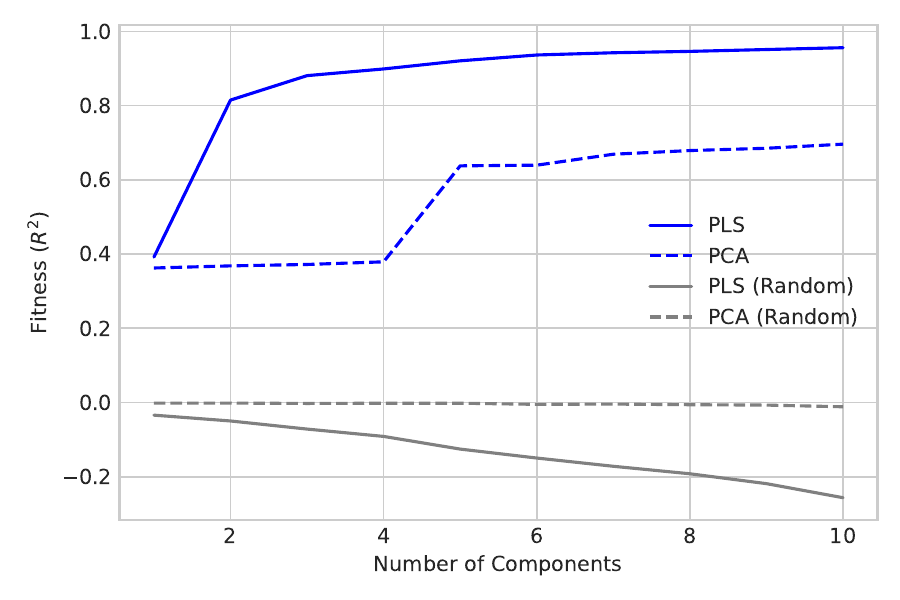}
  \caption{$C_{object}$}
\end{subfigure}\hfil 
\caption{Decoding performance of $[ei,ri]$ from activations of Qwen3-8B on the \textbf{\# separation=3} setting.}
\label{fig:jump3_score_qwen}
\end{figure*}

\begin{figure*}[!htbp]
    \centering 
\begin{subfigure}{0.35\textwidth}
  \includegraphics[width=\linewidth]{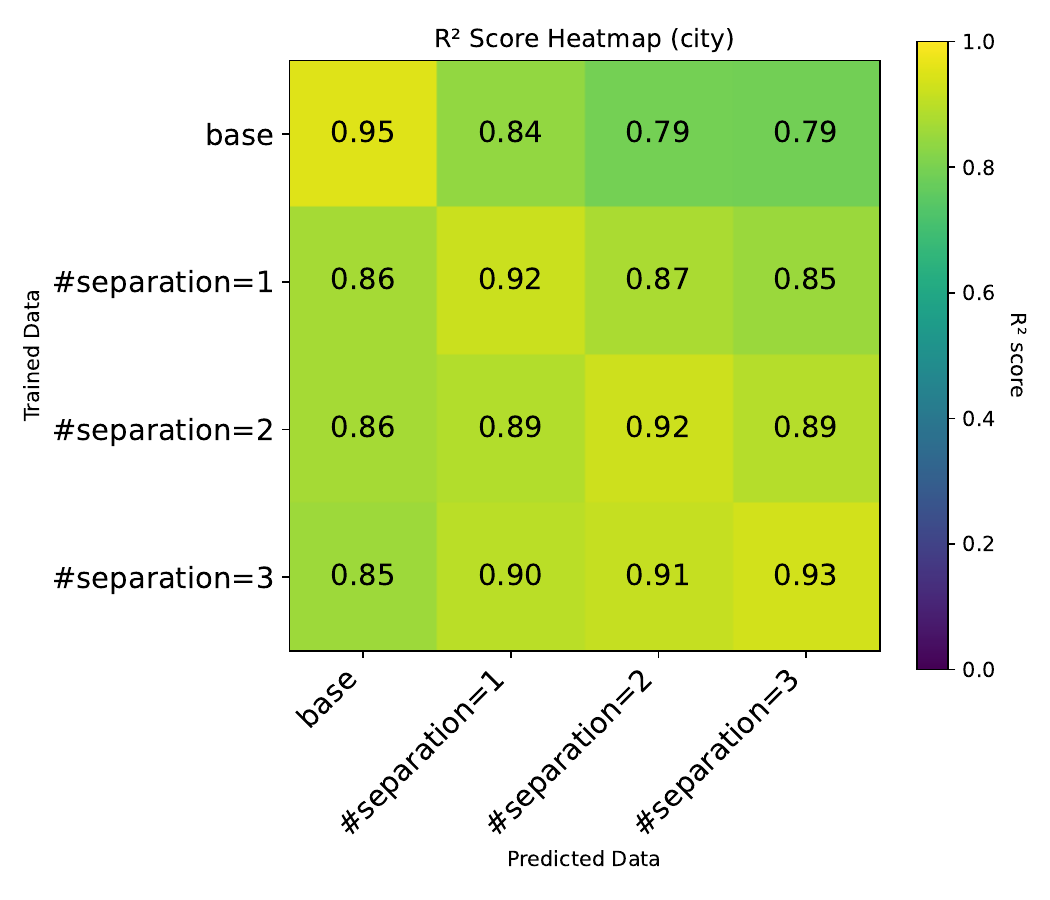}
  \caption{$C_{city}$}
\end{subfigure}\hfil 
\begin{subfigure}{0.35\textwidth}
  \includegraphics[width=\linewidth]{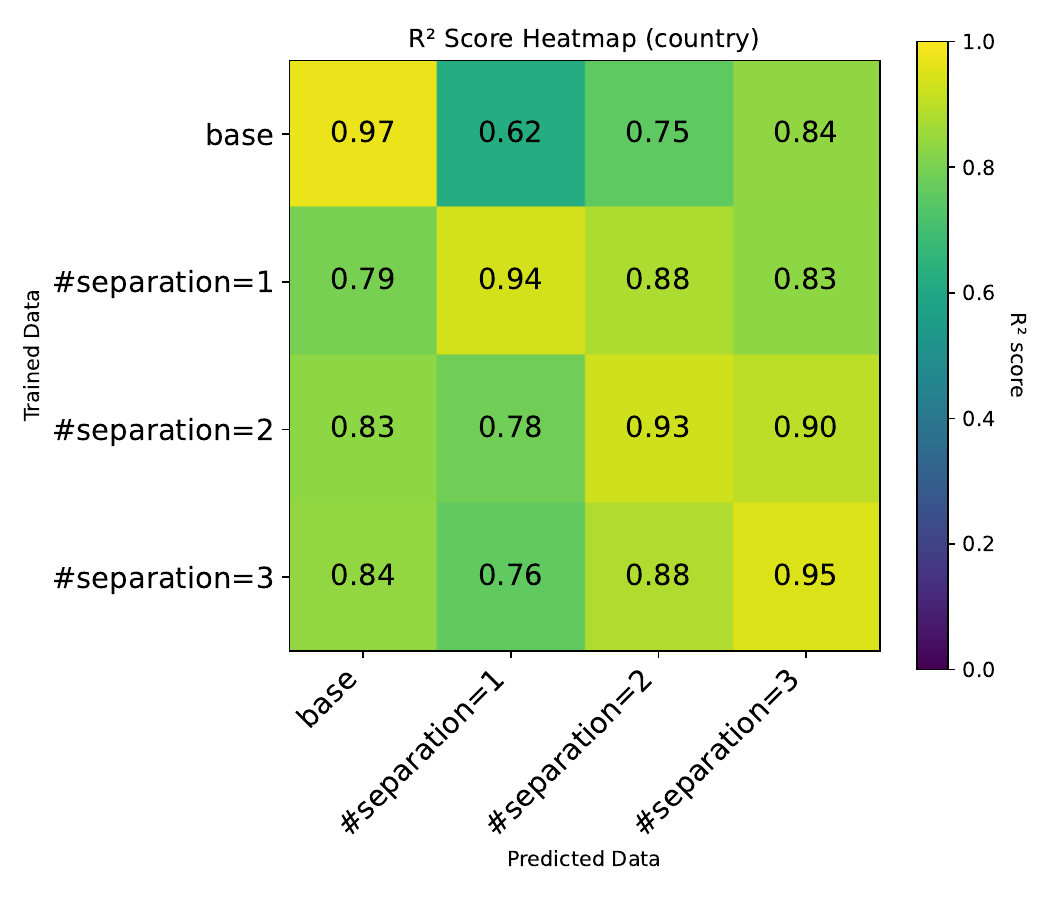}
  \caption{$C_{country}$}
\end{subfigure}\hfil 
\begin{subfigure}{0.3\textwidth}
  \includegraphics[width=\linewidth]{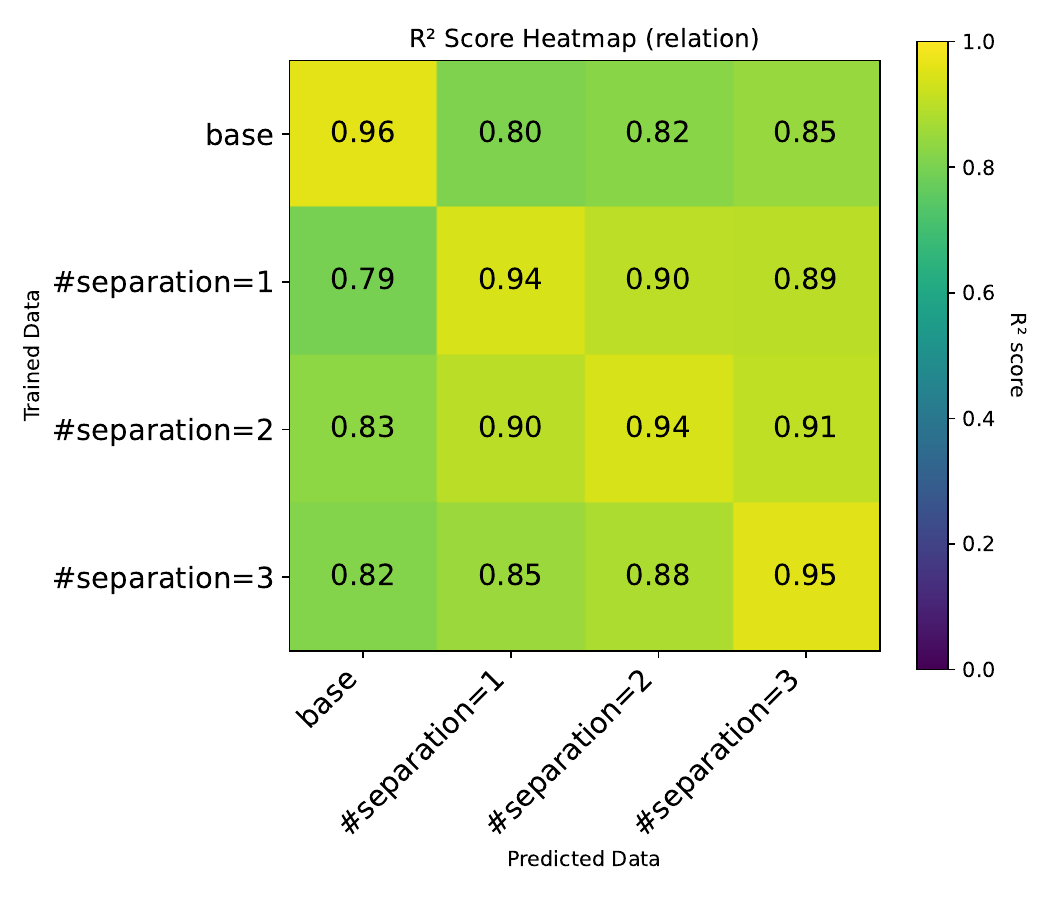}
  \caption{$C_{relation}$}
\end{subfigure}\hfil
\begin{subfigure}{0.3\textwidth}
  \includegraphics[width=\linewidth]{graph/heat_jump_llama_city.pdf}
  \caption{$C_{job}$}
\end{subfigure}\hfil 
\begin{subfigure}{0.3\textwidth}
  \includegraphics[width=\linewidth]{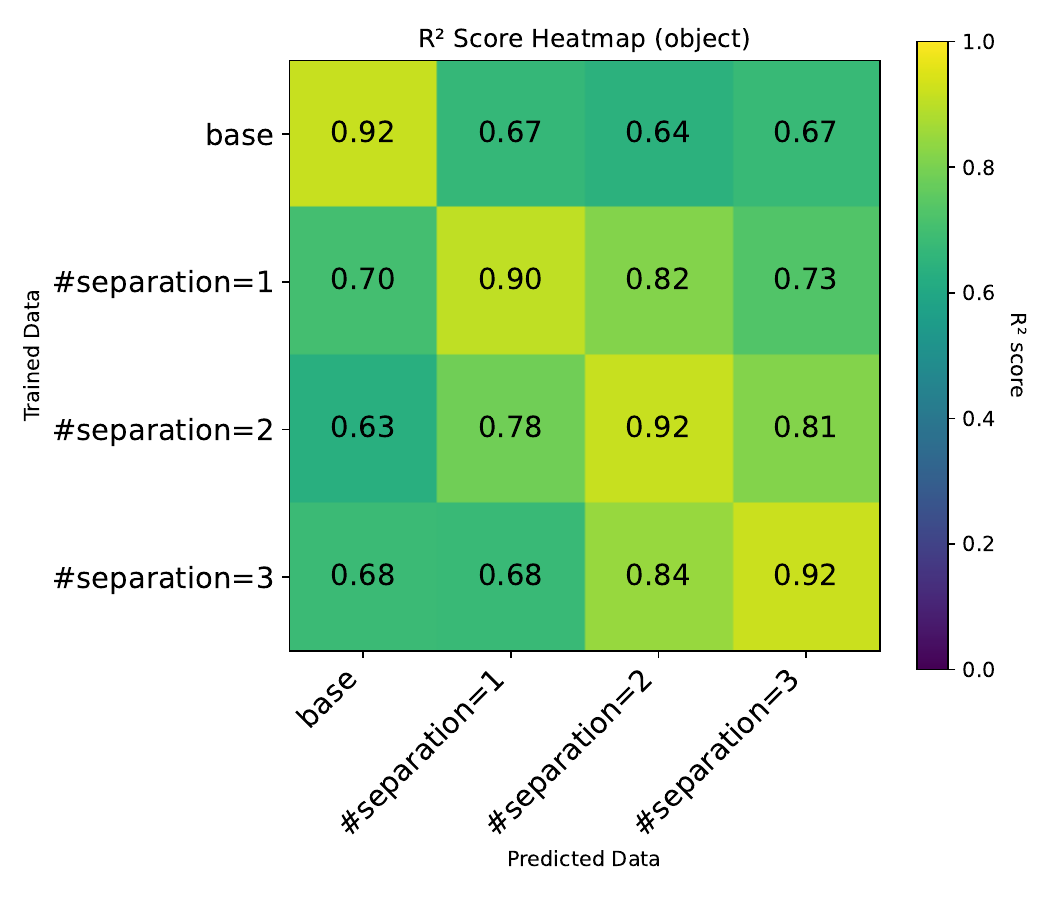}
  \caption{$C_{object}$}
\end{subfigure}\hfil 
\caption{Cross permutation (separation) $R^2$ scores for index prediction on Llama3-8B-Instruct.}
\label{fig:jump_heat_llama}
\end{figure*}

\begin{figure*}[!htbp]
    \centering 
\begin{subfigure}{0.35\textwidth}
  \includegraphics[width=\linewidth]{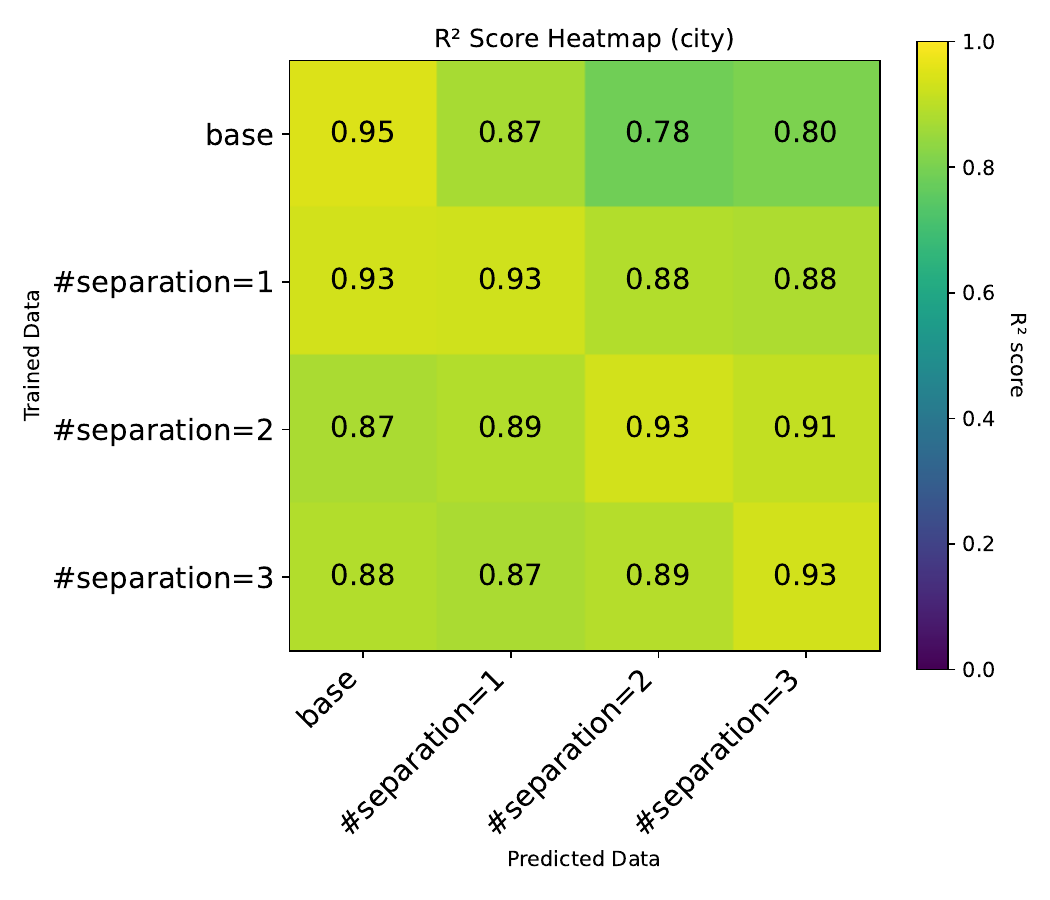}
  \caption{$C_{city}$}
\end{subfigure}\hfil 
\begin{subfigure}{0.35\textwidth}
  \includegraphics[width=\linewidth]{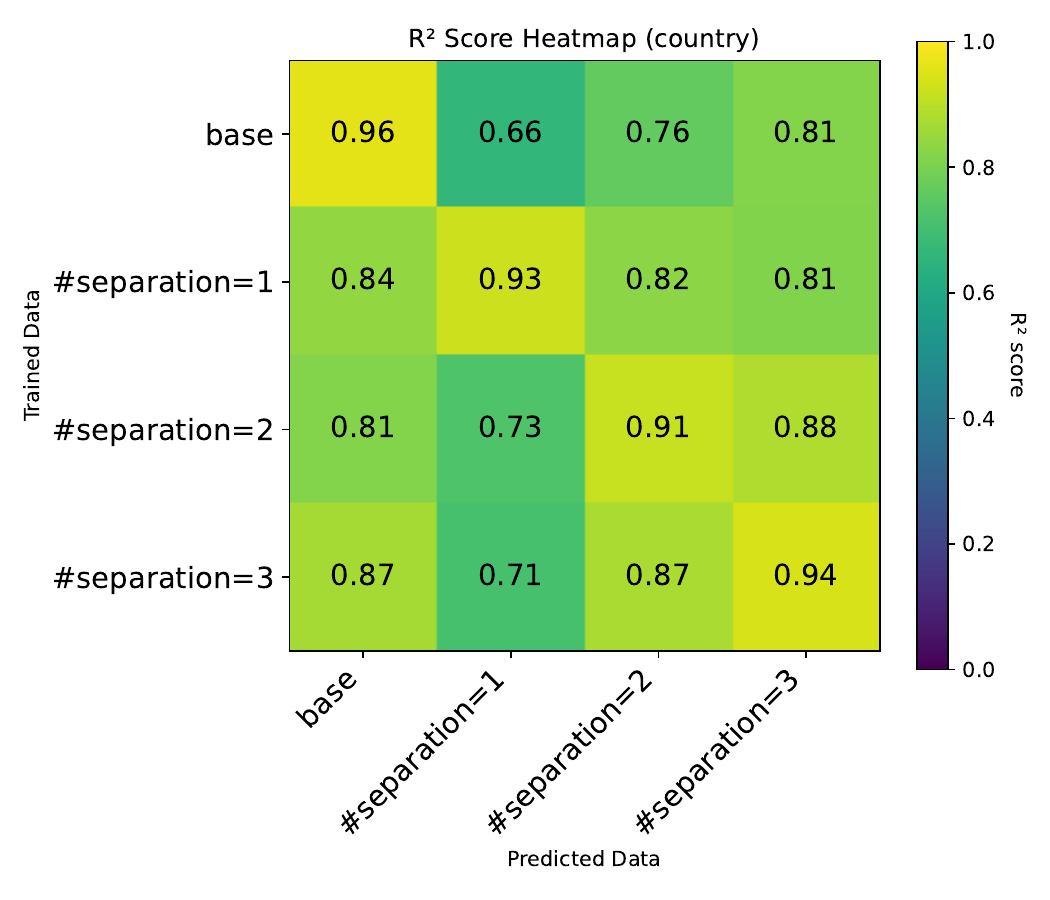}
  \caption{$C_{country}$}
\end{subfigure}\hfil 
\begin{subfigure}{0.3\textwidth}
  \includegraphics[width=\linewidth]{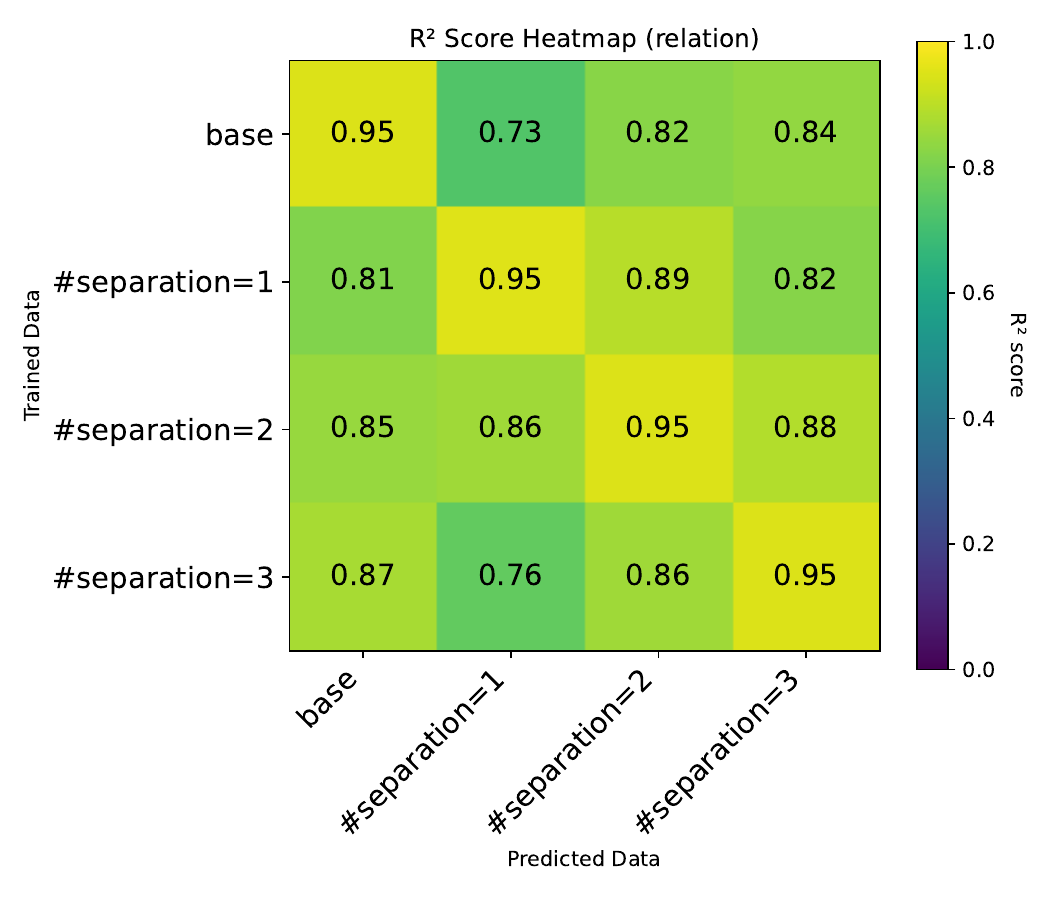}
  \caption{$C_{relation}$}
\end{subfigure}\hfil
\begin{subfigure}{0.3\textwidth}
  \includegraphics[width=\linewidth]{graph/heat_jump_qwen_city.pdf}
  \caption{$C_{job}$}
\end{subfigure}\hfil 
\begin{subfigure}{0.3\textwidth}
  \includegraphics[width=\linewidth]{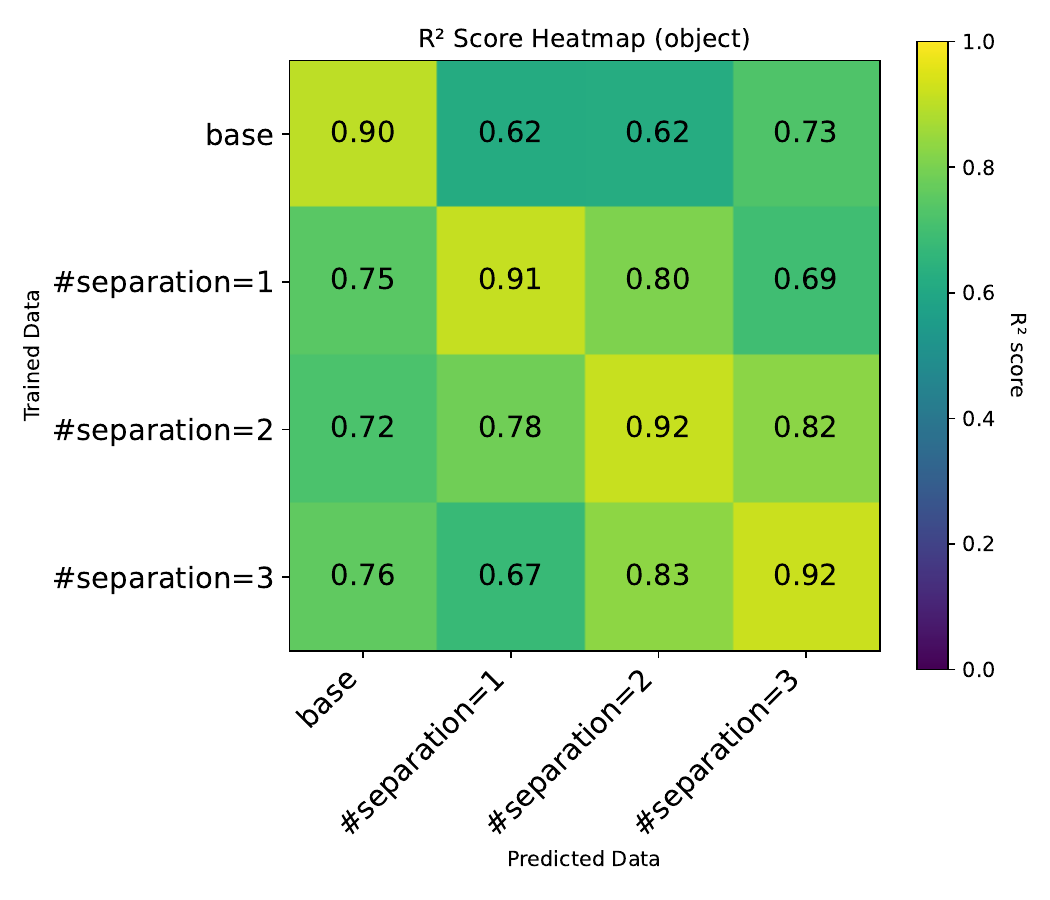}
  \caption{$C_{object}$}
\end{subfigure}\hfil 
\caption{Cross permutation (separation) $R^2$ scores for index prediction on Qwen3-8B.}
\label{fig:jump_heat_qwen}
\end{figure*}

We apply the same PLS-based approach to predict entity and relation indices under each modified setting. The resulting fitness scores are shown in Figure~\ref{fig:jump_score_llama}, \ref{fig:jump2_score_llama}, \ref{fig:jump3_score_llama}, \ref{fig:jump_score_qwen}, \ref{fig:jump2_score_qwen} and \ref{fig:jump3_score_qwen}. Across all settings, we observe consistently high $R^2$ scores, indicating that binding information can still be recovered from a low-dimensional subspace despite these input-level modifications. Furthermore, we evaluate the generality of the learned CBR projection matrices across these settings by using a projection matrix learned in one setting to predict entity and relation indices in another. As shown in Figure~\ref{fig:jump_heat_llama} and \ref{fig:jump_heat_qwen}, predictive performance remains largely stable under the separation such as for index prediction on the original input. Together, these findings suggest that the overall structure of the CBR subspace is robust to reference distance, supporting the stability of CBR representations across diverse discourse formats.


\clearpage

\subsection{Prevalence of CBR Across Diverse Discourse Patterns}
\label{sec:pattern}
To ensure that the observed grid-like CBR representation is not an artifact of a repetitive pattern, we construct 13 discourse patterns with progressively increasing structural complexity (Patt.1–Patt.13, shown below). The patterns vary in attribute count, relational overlap, and the distribution of relations across entities, and each template contains 1,000 naturalistic samples.

\begin{itemize}
\item Patt.~1: $(e_1r_1, e_2r_2, e_3r_3, e_3r_4)$
\item Patt.~2: $(e_1r_1, e_2r_2, e_2r_3, e_3r_4)$
\item Patt.~3: $(e_1r_1, e_1r_2, e_2r_3, e_3r_4)$
\item Patt.~4: $(e_1r_1, e_1r_2, e_2r_2, e_2r_3, e_3r_3, e_3r_4)$
\item Patt.~5: $(e_1r_1, e_1r_2, e_2r_1, e_2r_3, e_3r_2, e_3r_4)$
\item Patt.~6: $(e_1r_1, e_1r_2, e_2r_3, e_2r_4, e_3r_1, e_3r_4)$
\item Patt.~7: $(e_1r_1, e_1r_2, e_2r_2, e_2r_3, e_2r_4, e_3r_2, e_3r_3, e_3r_4)$
\item Patt.~8: $(e_1r_1, e_1r_2, e_2r_1, e_2r_3, e_2r_4, e_3r_1, e_3r_2, e_3r_4)$
\item Patt.~9: $(e_1r_1, e_1r_2, e_2r_1, e_2r_3, e_2r_4, e_3r_1, e_3r_3, e_3r_4)$
\item Patt.~10: $(e_1r_1, e_1r_2, e_1r_3, e_2r_1, e_2r_3, e_2r_4, e_3r_2, e_3r_3, e_3r_4)$
\item Patt.~11: $(e_1r_1, e_1r_2, e_1r_3, e_2r_1, e_2r_3, e_2r_4, e_3r_1, e_3r_2, e_3r_4)$
\item Patt.~12: $(e_1r_1, e_1r_2, e_1r_3, e_2r_1, e_2r_3, e_2r_4, e_3r_1, e_3r_3, e_3r_4)$
\item Patt.~13: $(e_1r_1, e_1r_2, e_1r_3, e_2r_1, e_2r_3, e_2r_4, e_3r_1, e_3r_2, e_3r_3)$
\end{itemize}

Here, $e_ir_j$ denotes an attribute bound to entity $e_i$ through relation $r_j$. 
For example, in the example of Patt.~6 (Sample~\ref{ex:patt6}), $e_2r_3$ corresponds 
to the attribute ``\textit{Perm}'', which is connected to entity $e_2$ (i.e., 
``\textit{Rick}'') via relation $r_3$ (i.e., \textit{fondness for}). Importantly, these patterns are not simple repetitions of a single structure. Instead, they progressively increase structural density and relational entanglement, covering a wide range of possible entity–relation configurations for a fixed entity set.
To illustrate these patterns, we provide examples from the Context: \textit{city}.

\begin{exe}
    \ex \label{ex:patt1}\textbf{Patt.~1 example.} \textit{Lee, who was born in \underline{Split}$_{e_1r_1}$, often dreams of traveling to new places. Meanwhile, Rick enjoys his life in \underline{Detroit}$_{e_2r_2}$, where he explores the city's vibrant culture. Ian, a passionate traveler, expresses his love for \underline{Paris}$_{e_3r_3}$ but openly shares his dislike for \underline{Austin}$_{e_3r_4}$, preferring the charm of the French capital instead. The three friends often discuss their different experiences and preferences, bringing their unique perspectives together.}
    \ex \label{ex:patt6}\textbf{Patt.~6 example.}
    \textit{Lee, born in \underline{Split}$_{e_1r_1}$, enjoys his life in \underline{Hamilton}$_{e_1r_2}$. Meanwhile, Rick has a fondness for \underline{Perm}$_{e_2r_3}$ but holds a strong dislike for \underline{Houston}$_{e_2r_4}$. Ian, originally from \underline{Portland}$_{e_3r_1}$, also expresses his aversion to \underline{Austin}$_{e_3r_4}$. Each of them navigates their preferences and experiences, shaping their unique perspectives on the places they call home.}
    \ex \label{ex:patt10}\textbf{Patt.~10 example.}
    \textit{Lee, born in \underline{Split}$_{e_1r_1}$, now lives in \underline{Hamilton}$_{e_1r_2}$ and has a deep affection for \underline{Toronto}$_{e_1r_3}$. Rick, hailing from \underline{Boston}$_{e_2r_1}$, holds a fondness for \underline{Perm}$_{e_2r_3}$ while harboring a dislike for \underline{Houston}$_{e_2r_4}$. Ian, who currently resides in \underline{Phoenix}$_{e_3r_2}$, adores \underline{Paris}$_{e_3r_3}$ but has negative feelings towards \underline{Austin}$_{e_3r_4}$. Each of them carries their unique experiences and preferences, shaping the places they love and those they do not.}
\end{exe}

\begin{figure*}[!htbp]
    \centering 
\begin{subfigure}{0.3\textwidth}
  \includegraphics[width=\linewidth]{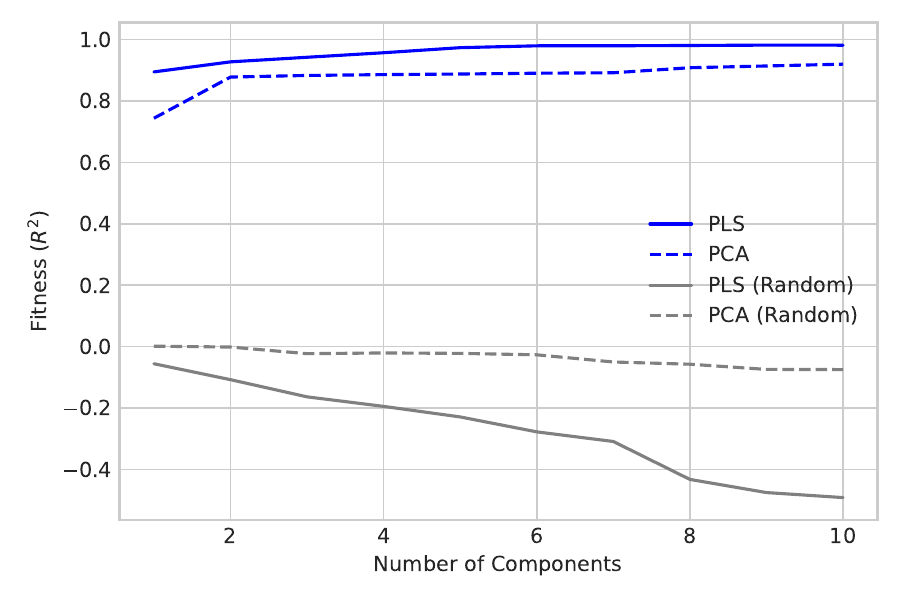}
  \caption{Patt.1}
\end{subfigure}\hfil
\begin{subfigure}{0.3\textwidth}
  \includegraphics[width=\linewidth]{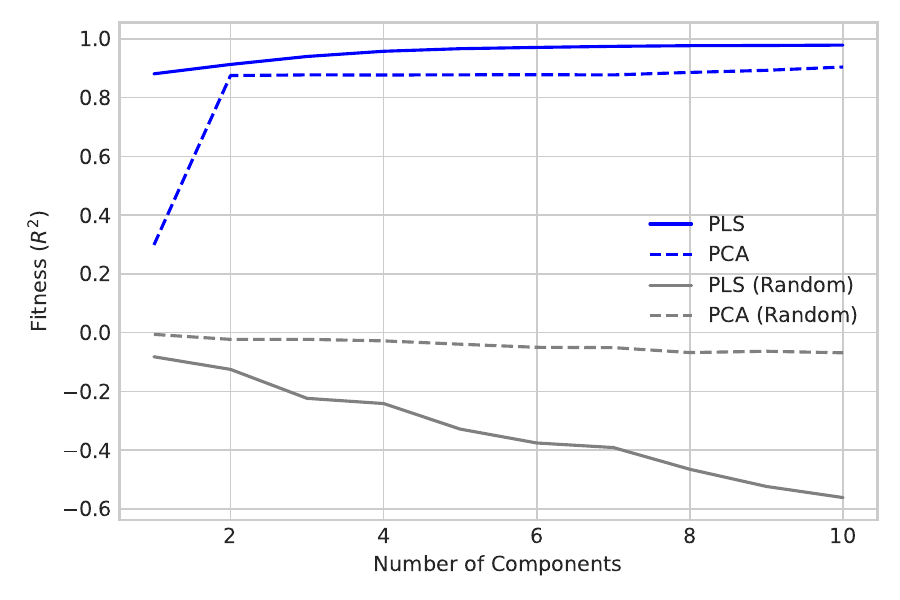}
  \caption{Patt.2}
\end{subfigure}\hfil 
\begin{subfigure}{0.3\textwidth}
  \includegraphics[width=\linewidth]{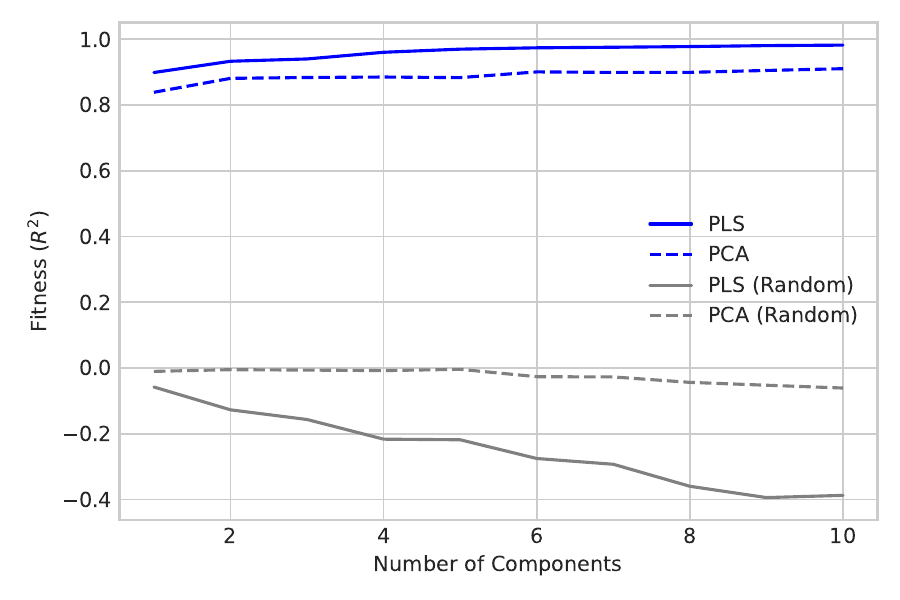}
  \caption{Patt.3}
\end{subfigure}\hfil 
\begin{subfigure}{0.3\textwidth}
  \includegraphics[width=\linewidth]{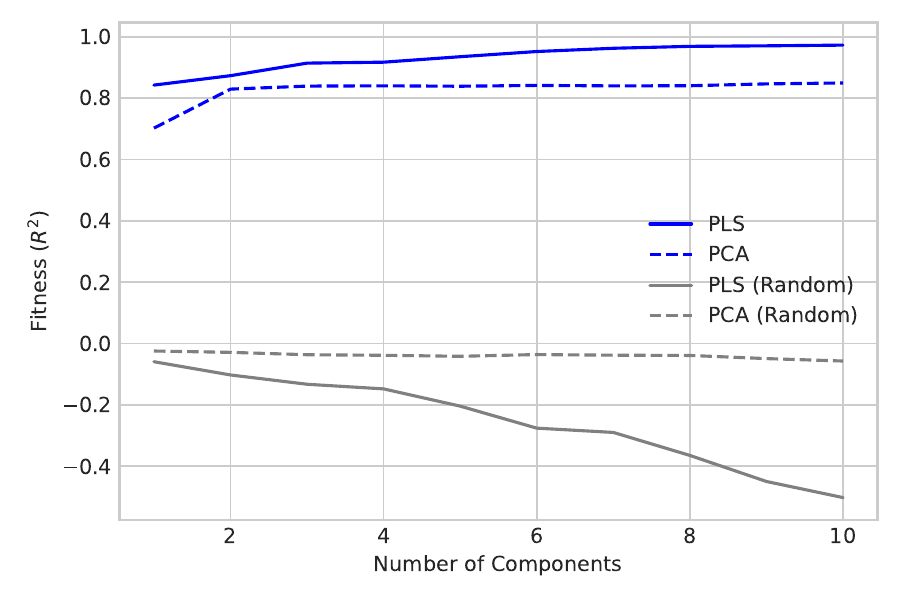}
  \caption{Patt.4}
\end{subfigure}\hfil
\begin{subfigure}{0.3\textwidth}
  \includegraphics[width=\linewidth]{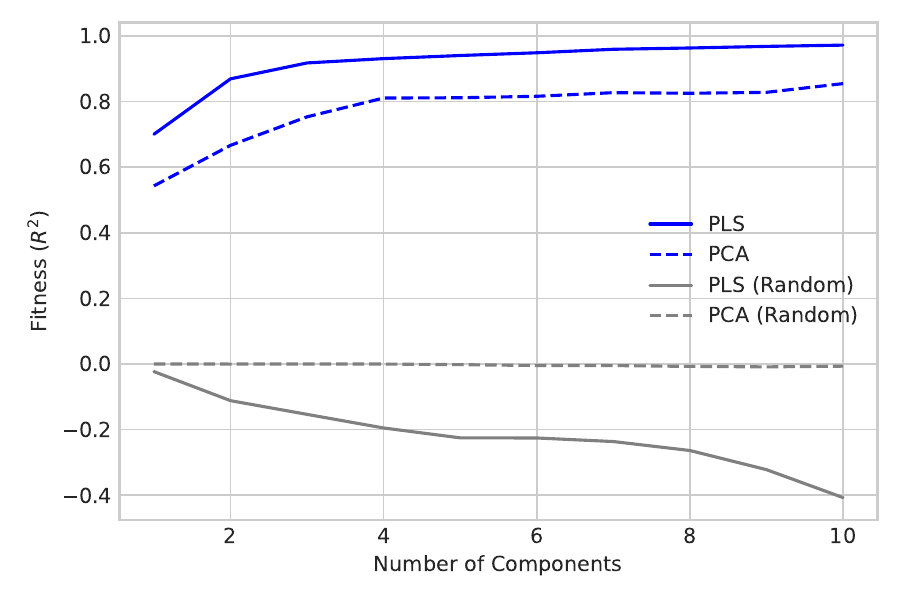}
  \caption{Patt.5}
\end{subfigure}\hfil 
\begin{subfigure}{0.3\textwidth}
  \includegraphics[width=\linewidth]{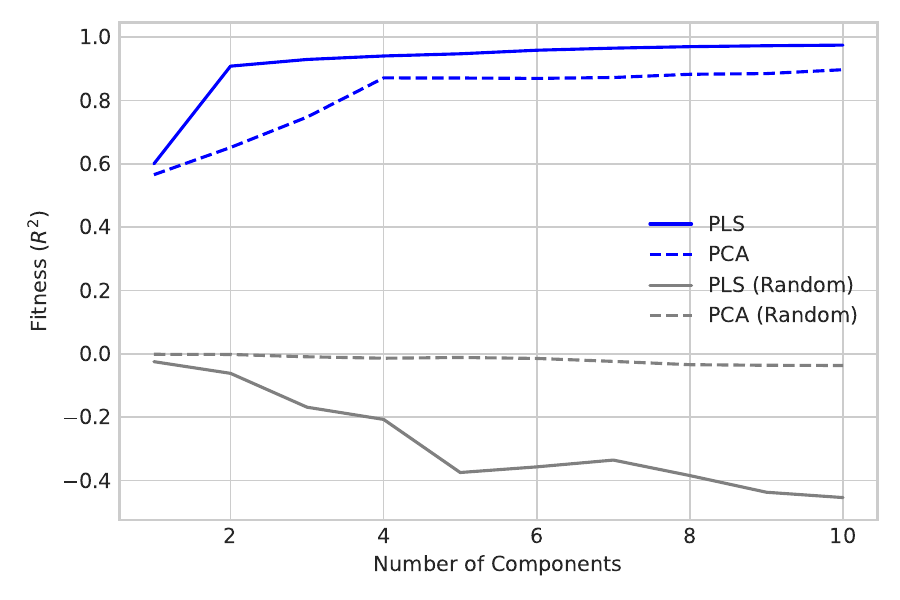}
  \caption{Patt.6}
\end{subfigure}\hfil 
\begin{subfigure}{0.3\textwidth}
  \includegraphics[width=\linewidth]{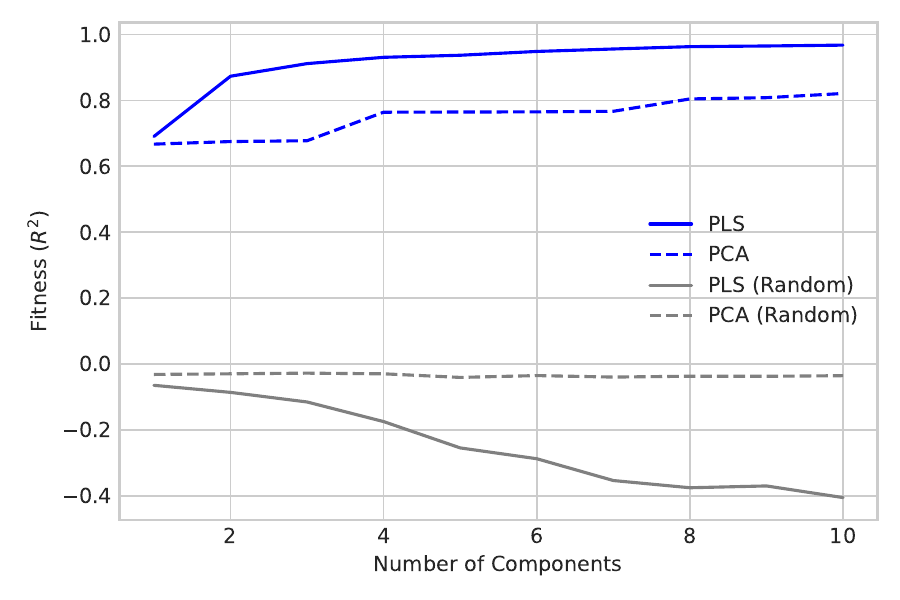}
  \caption{Patt.7}
\end{subfigure}\hfil
\begin{subfigure}{0.3\textwidth}
  \includegraphics[width=\linewidth]{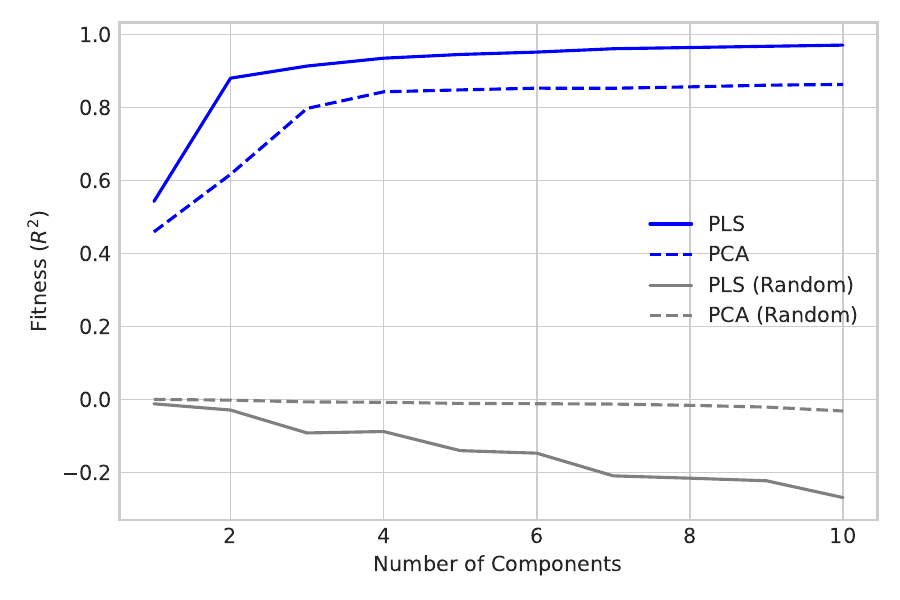}
  \caption{Patt.8}
\end{subfigure}\hfil 
\begin{subfigure}{0.3\textwidth}
  \includegraphics[width=\linewidth]{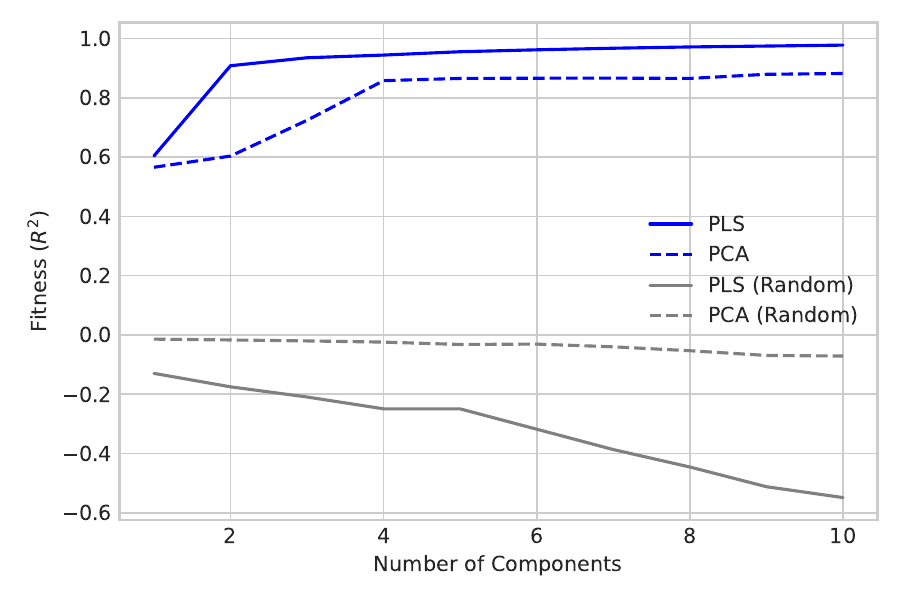}
  \caption{Patt.9}
\end{subfigure}\hfil 
\begin{subfigure}{0.3\textwidth}
  \includegraphics[width=\linewidth]{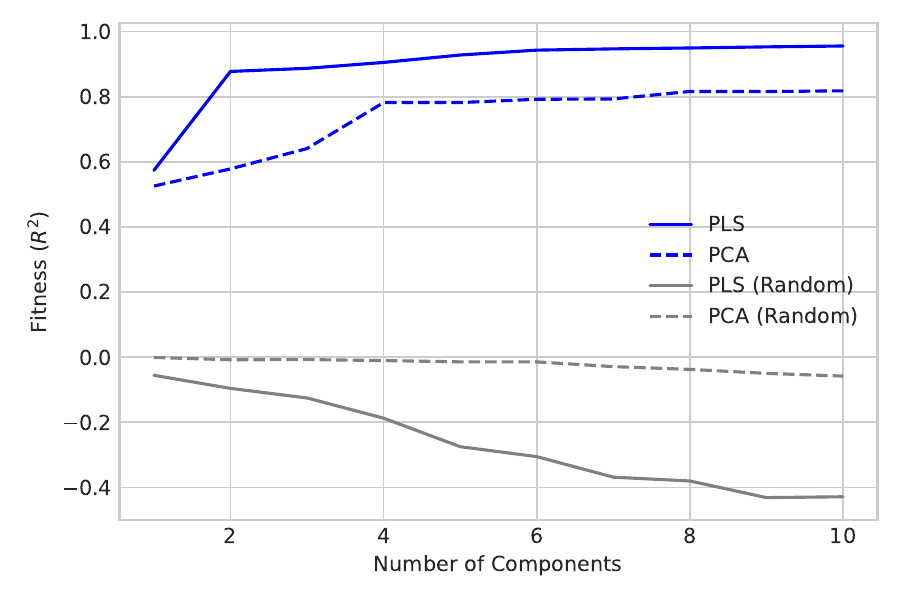}
  \caption{Patt.10}
\end{subfigure}\hfil
\begin{subfigure}{0.3\textwidth}
  \includegraphics[width=\linewidth]{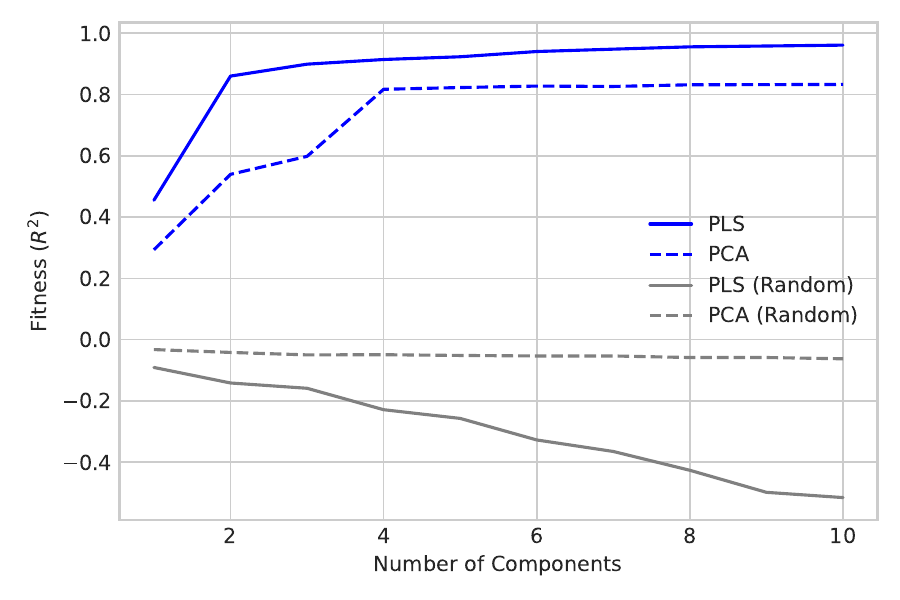}
  \caption{Patt.11}
\end{subfigure}\hfil 
\begin{subfigure}{0.3\textwidth}
  \includegraphics[width=\linewidth]{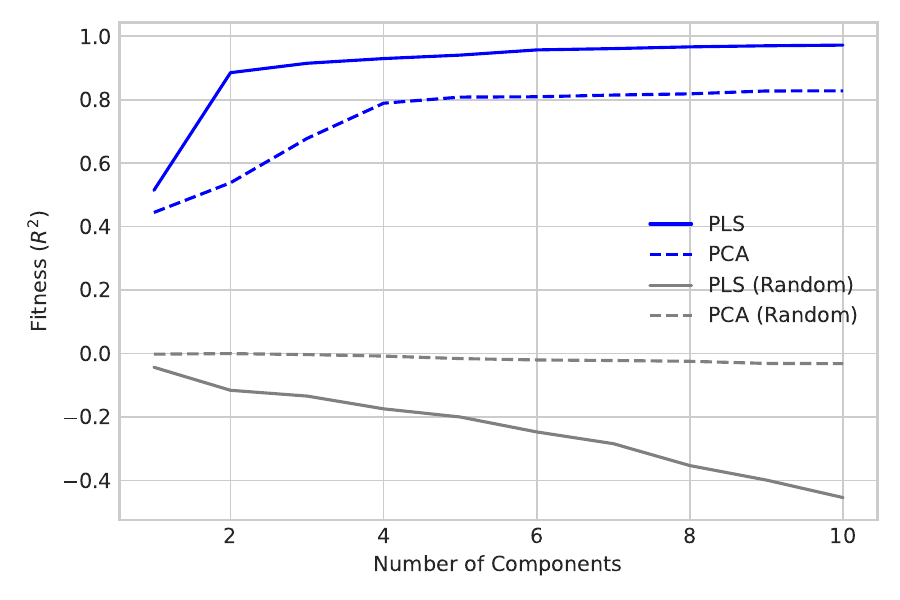}
  \caption{Patt.12}
\end{subfigure}\hfil 
\begin{subfigure}{0.3\textwidth}
  \includegraphics[width=\linewidth]{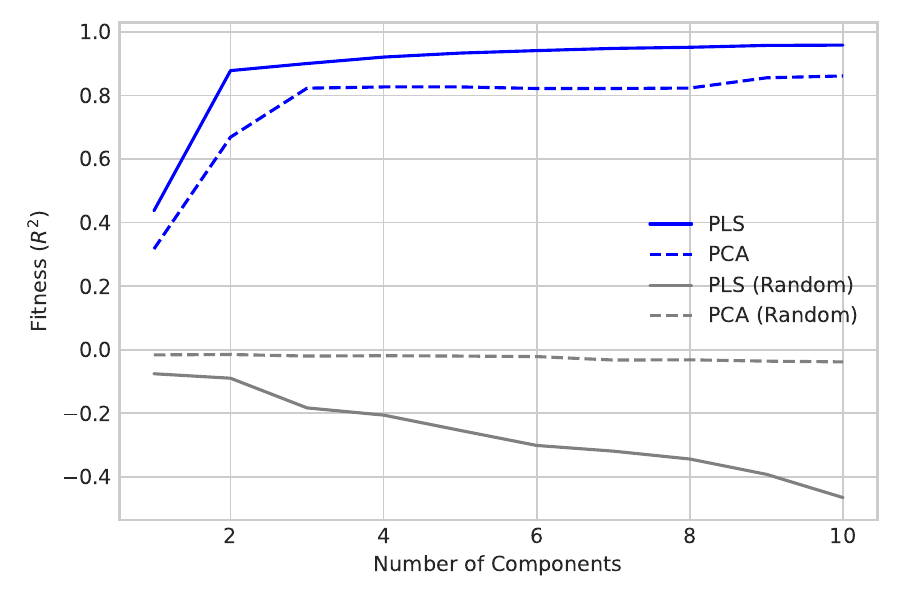}
  \caption{Patt.13}
\end{subfigure}\hfil 
\caption{Decoding performance of $[ei,ri]$ from activations of Llama3-8B-Instruct on $C_{city}$.}
\label{fig:pls_pattern_llama_city}
\end{figure*}

\begin{figure*}[!htbp]
    \centering 
\begin{subfigure}{0.3\textwidth}
  \includegraphics[width=\linewidth]{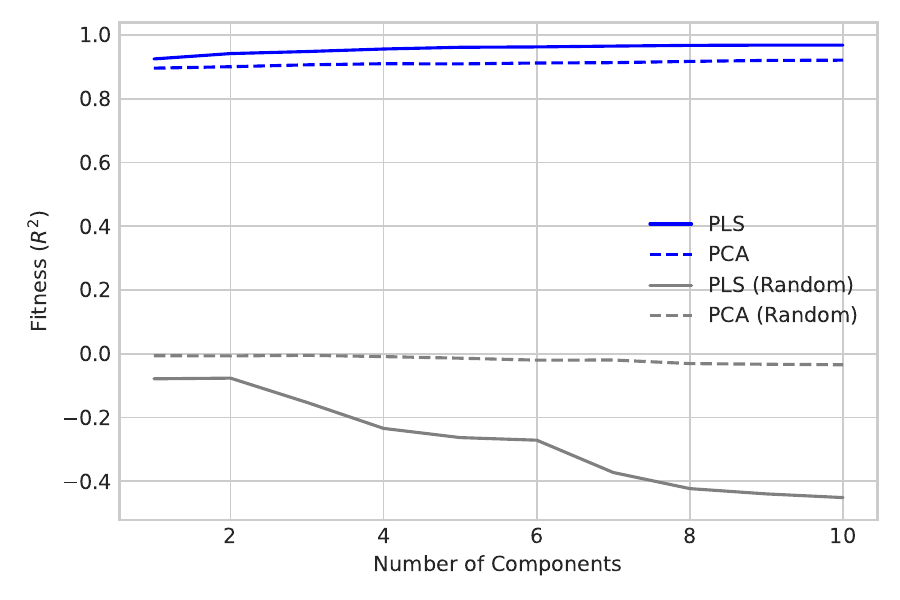}
  \caption{Patt.1}
\end{subfigure}\hfil
\begin{subfigure}{0.3\textwidth}
  \includegraphics[width=\linewidth]{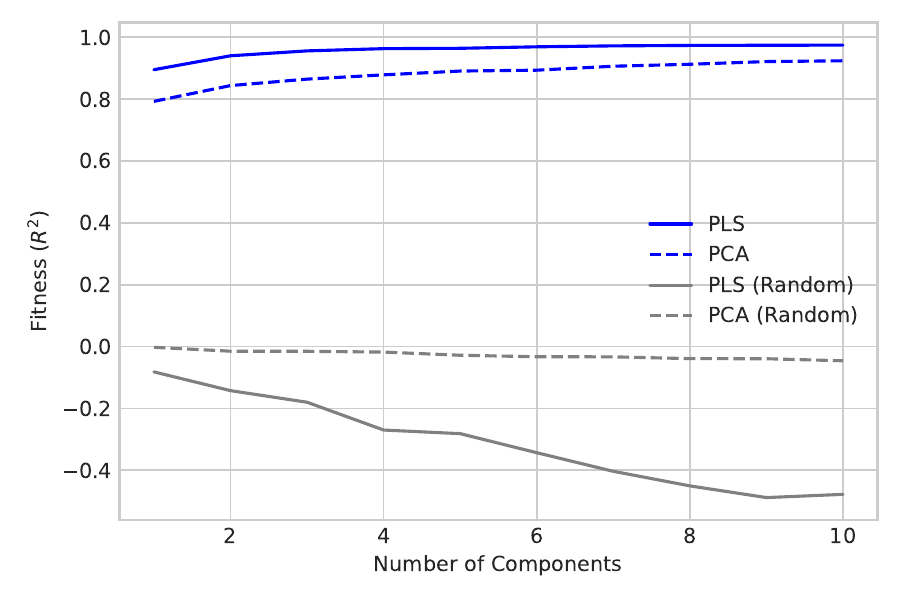}
  \caption{Patt.2}
\end{subfigure}\hfil 
\begin{subfigure}{0.3\textwidth}
  \includegraphics[width=\linewidth]{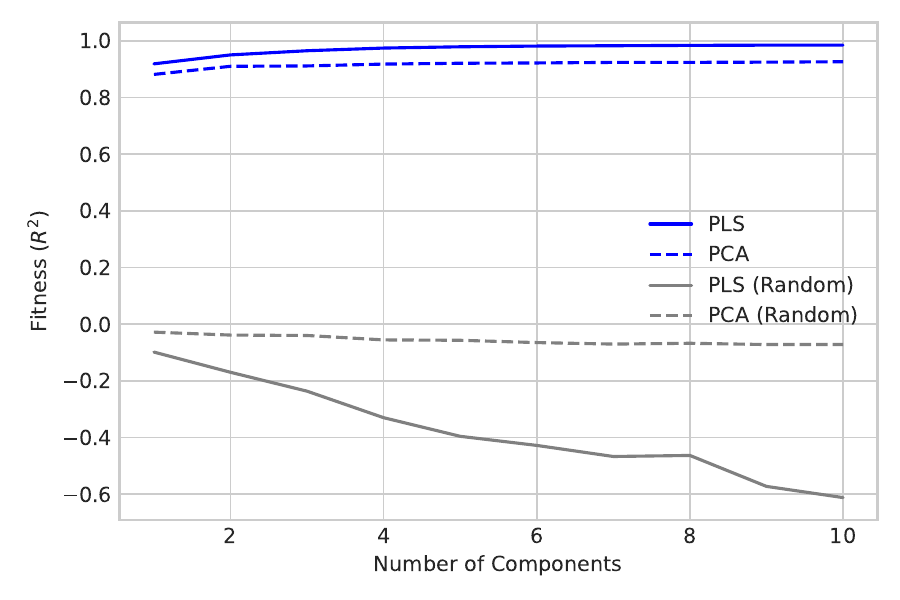}
  \caption{Patt.3}
\end{subfigure}\hfil 
\begin{subfigure}{0.3\textwidth}
  \includegraphics[width=\linewidth]{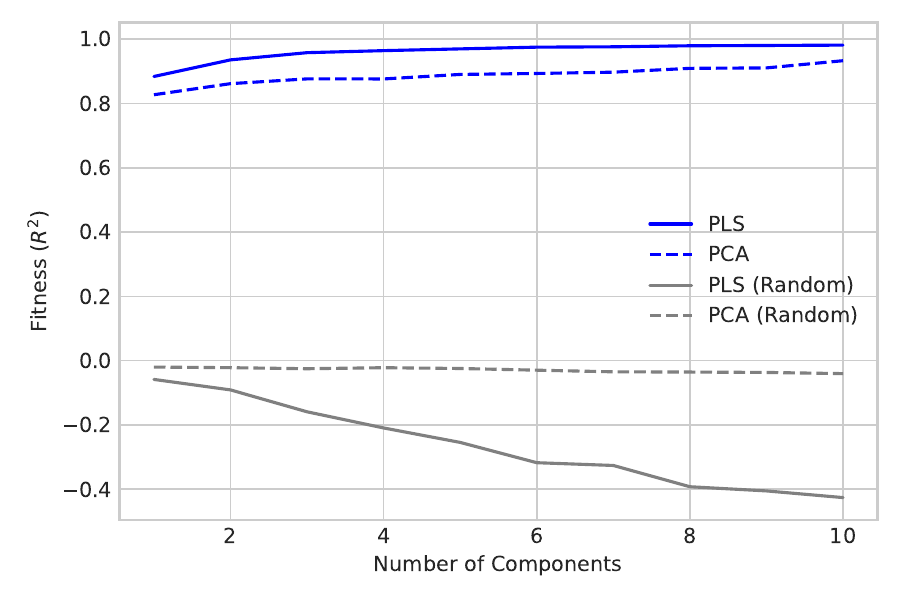}
  \caption{Patt.4}
\end{subfigure}\hfil
\begin{subfigure}{0.3\textwidth}
  \includegraphics[width=\linewidth]{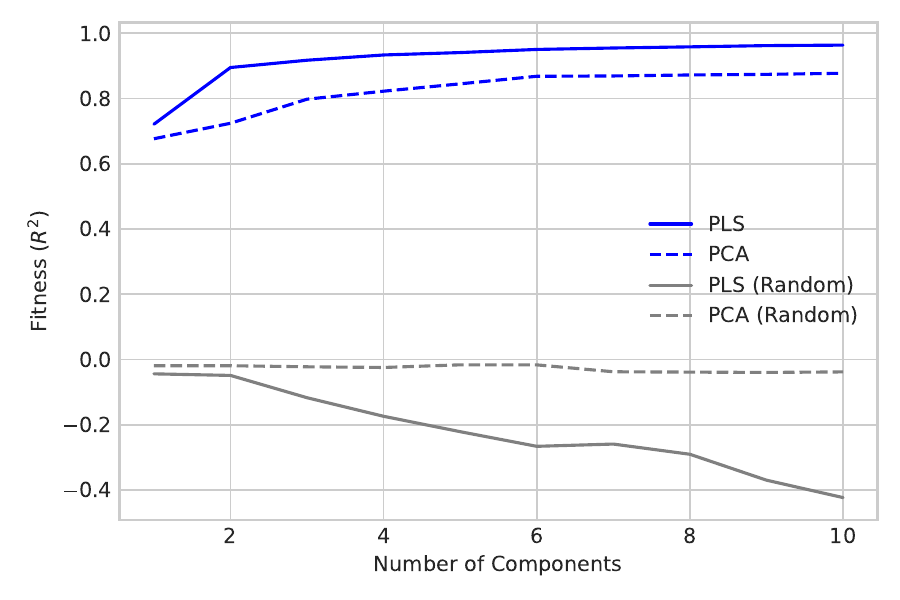}
  \caption{Patt.5}
\end{subfigure}\hfil 
\begin{subfigure}{0.3\textwidth}
  \includegraphics[width=\linewidth]{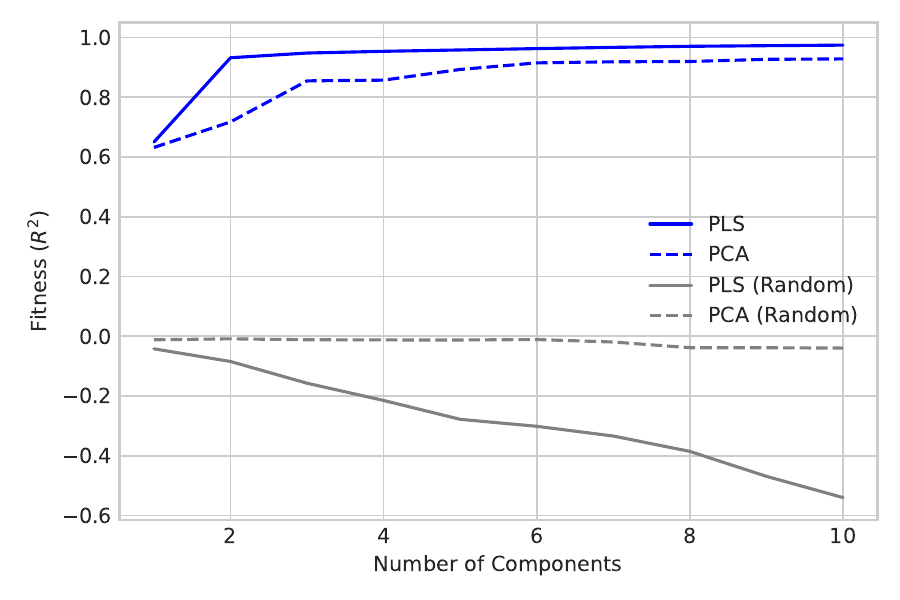}
  \caption{Patt.6}
\end{subfigure}\hfil 
\begin{subfigure}{0.3\textwidth}
  \includegraphics[width=\linewidth]{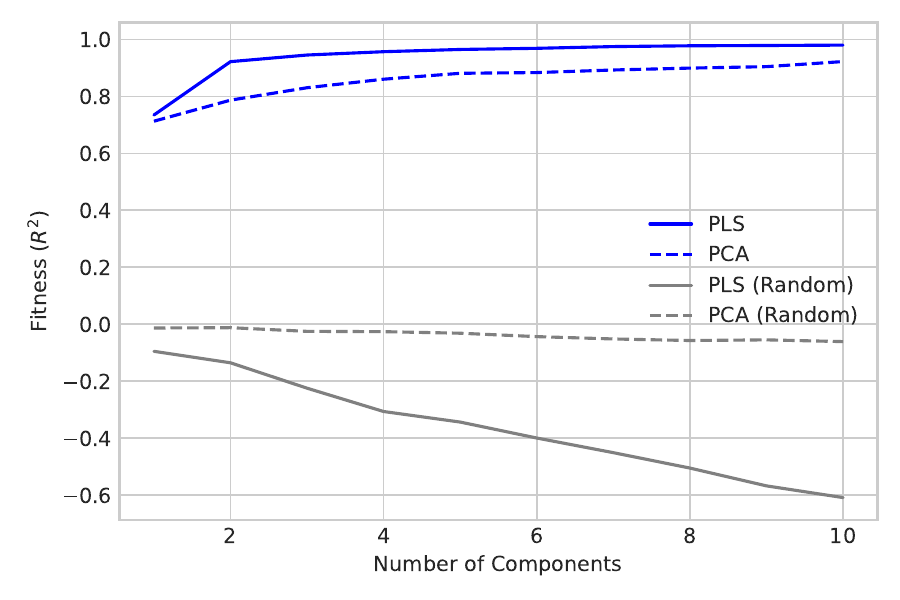}
  \caption{Patt.7}
\end{subfigure}\hfil
\begin{subfigure}{0.3\textwidth}
  \includegraphics[width=\linewidth]{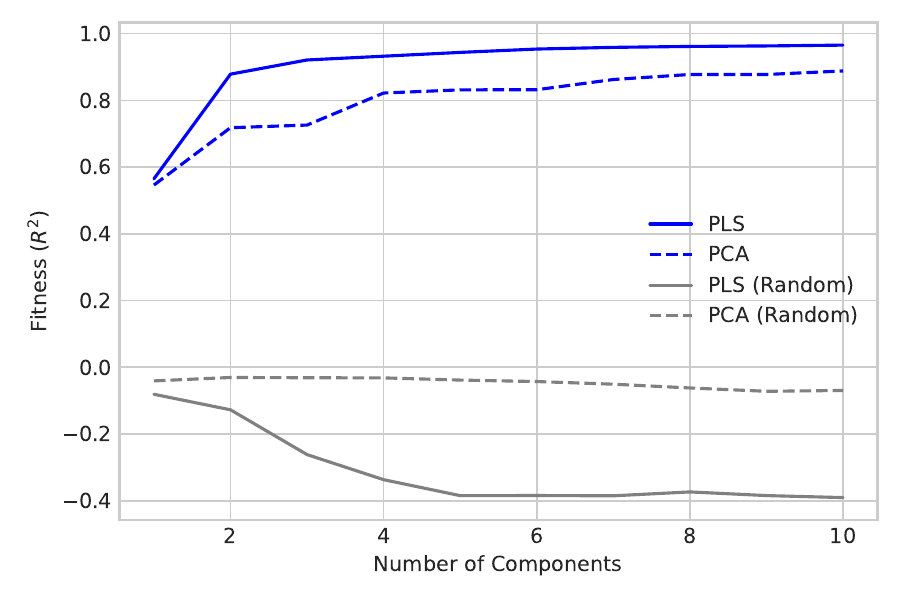}
  \caption{Patt.8}
\end{subfigure}\hfil 
\begin{subfigure}{0.3\textwidth}
  \includegraphics[width=\linewidth]{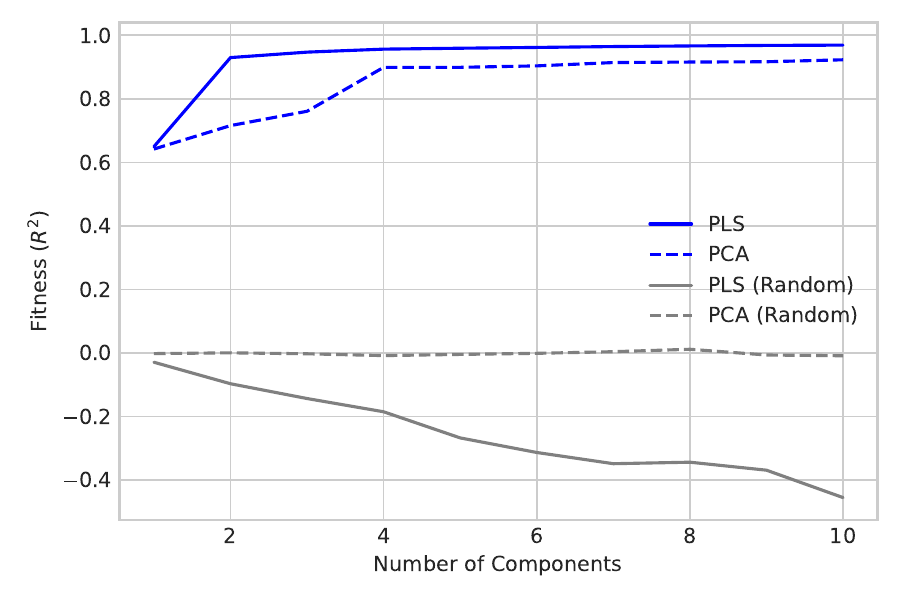}
  \caption{Patt.9}
\end{subfigure}\hfil 
\begin{subfigure}{0.3\textwidth}
  \includegraphics[width=\linewidth]{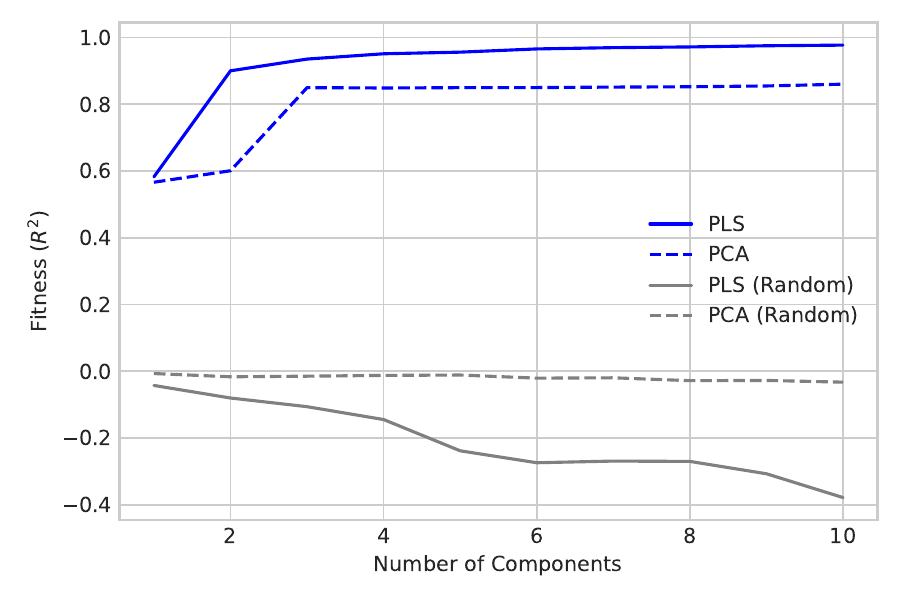}
  \caption{Patt.10}
\end{subfigure}\hfil
\begin{subfigure}{0.3\textwidth}
  \includegraphics[width=\linewidth]{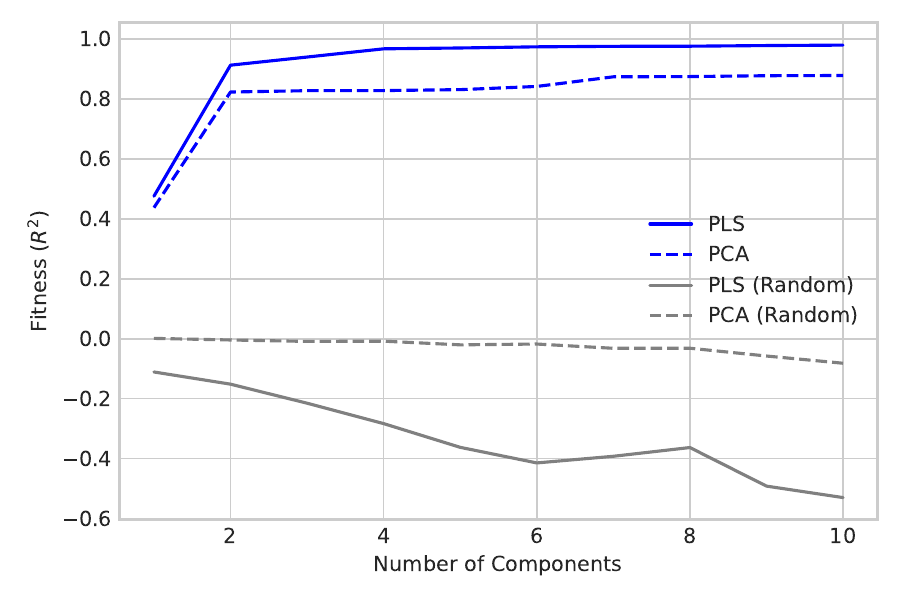}
  \caption{Patt.11}
\end{subfigure}\hfil 
\begin{subfigure}{0.3\textwidth}
  \includegraphics[width=\linewidth]{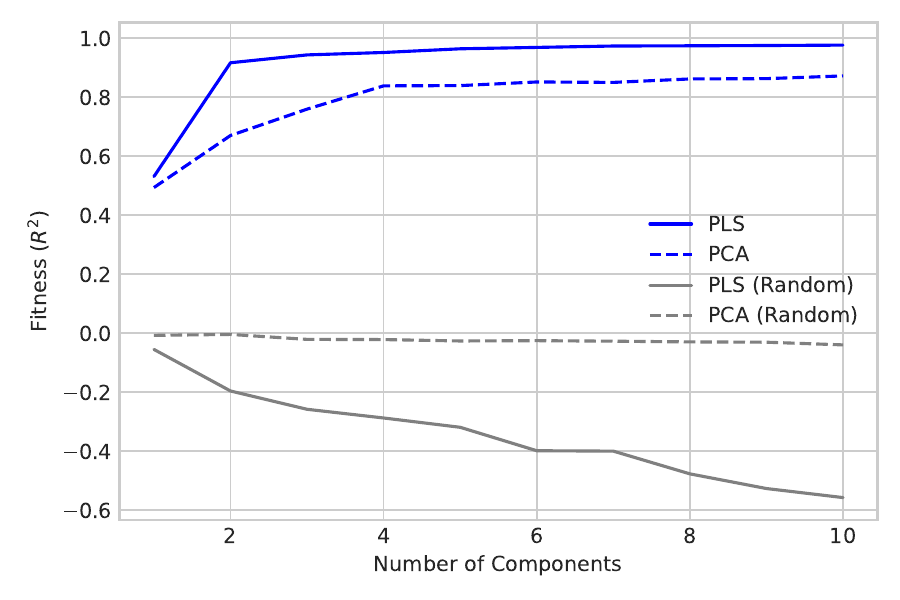}
  \caption{Patt.12}
\end{subfigure}\hfil 
\begin{subfigure}{0.3\textwidth}
  \includegraphics[width=\linewidth]{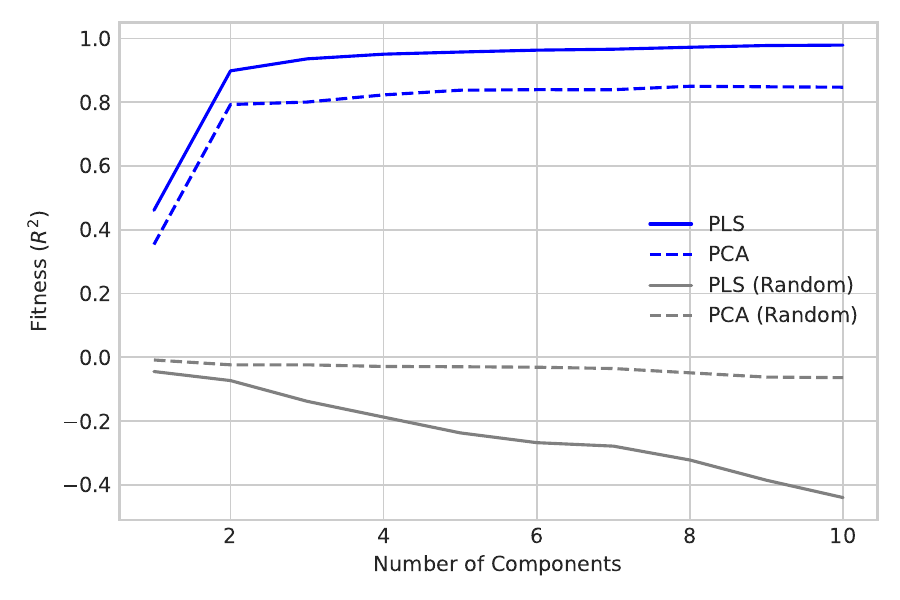}
  \caption{Patt.13}
\end{subfigure}\hfil 
\caption{Decoding performance of $[ei,ri]$ from activations of Llama3-8B-Instruct on $C_{country}$.}
\label{fig:pls_pattern_llama_country}
\end{figure*}

We apply the same PLS-based framework to estimate entity and relation indices across the 13 discourse patterns. The resulting performance scores are shown in Figure~\ref{fig:pls_pattern_llama_city} and Figure~\ref{fig:pls_pattern_llama_country}. Across all patterns, prediction accuracy remains consistently high even when using a small number of PLS components. This observation suggests that the information required to reconstruct entity–relation bindings is preserved in a low-dimensional subspace, even when the surface structure of the discourse varies substantially.

\begin{figure*}[!htbp]
    \centering 
\begin{subfigure}{0.3\textwidth}
  \includegraphics[width=\linewidth]{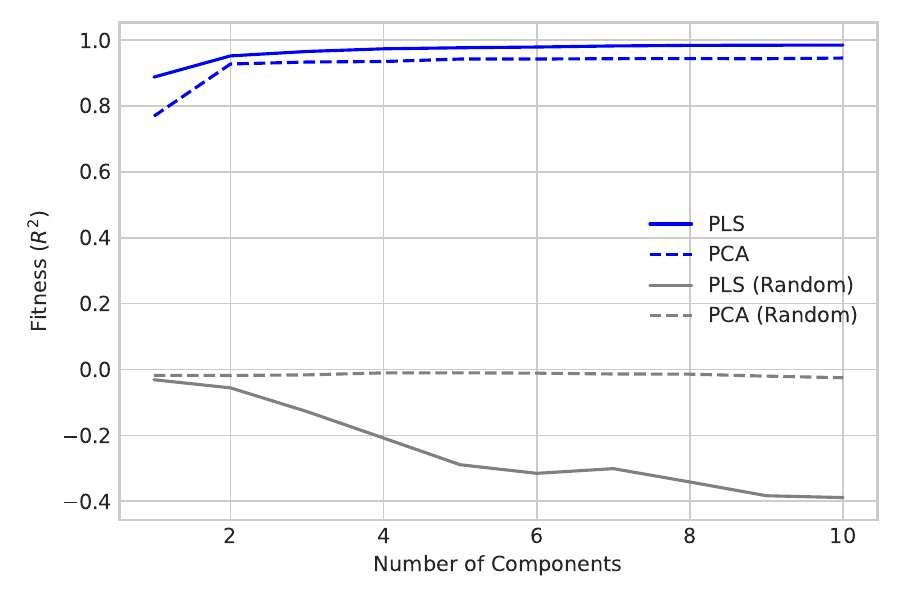}
  \caption{Patt.1}
\end{subfigure}\hfil
\begin{subfigure}{0.3\textwidth}
  \includegraphics[width=\linewidth]{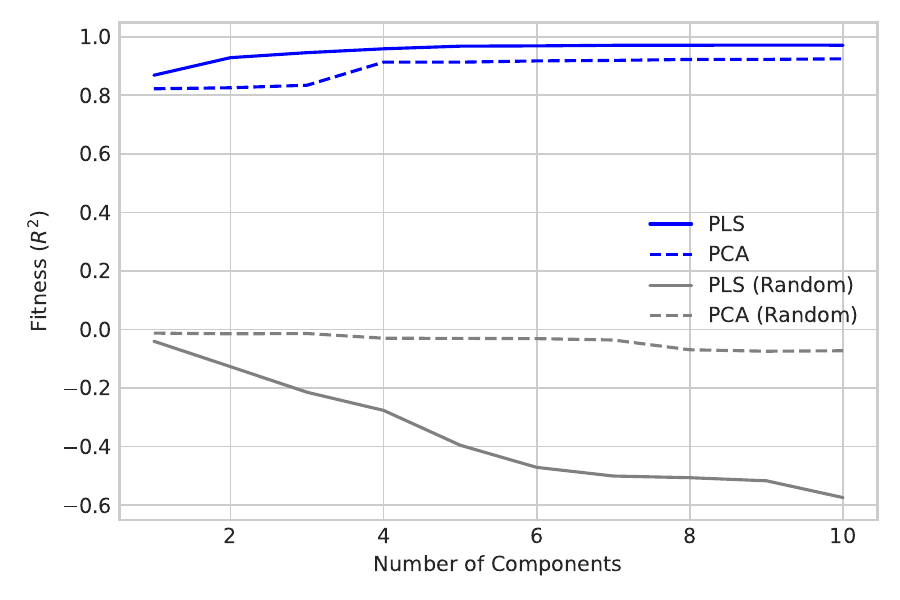}
  \caption{Patt.2}
\end{subfigure}\hfil 
\begin{subfigure}{0.3\textwidth}
  \includegraphics[width=\linewidth]{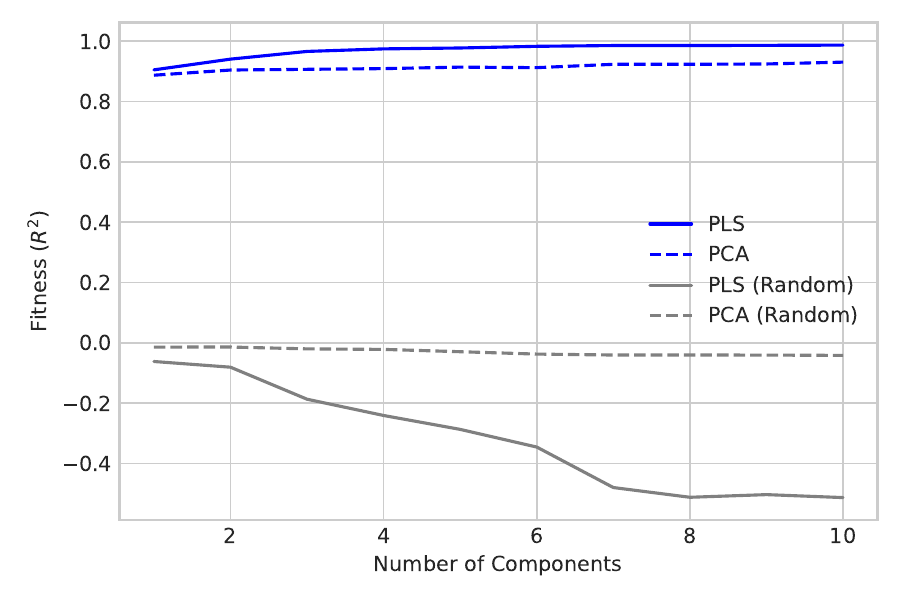}
  \caption{Patt.3}
\end{subfigure}\hfil 
\begin{subfigure}{0.3\textwidth}
  \includegraphics[width=\linewidth]{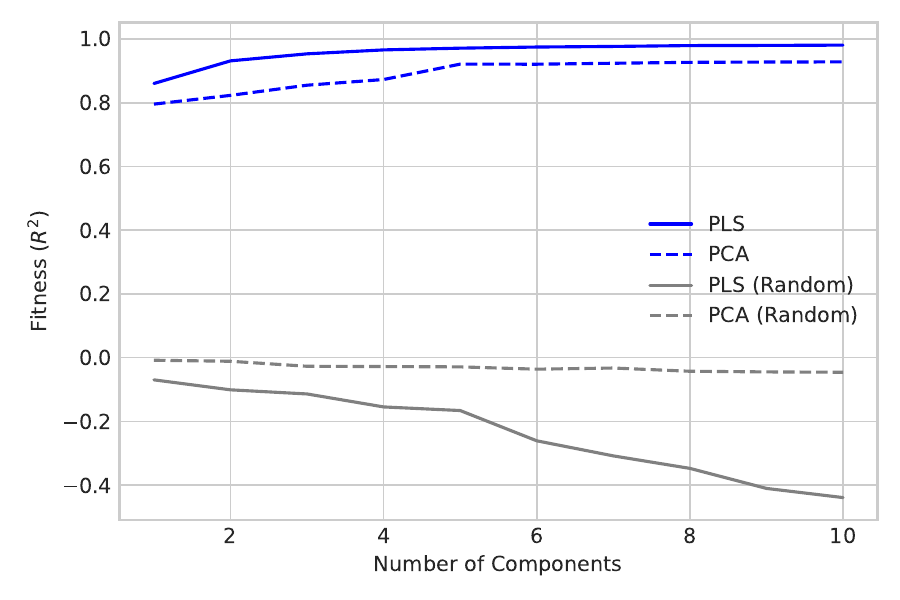}
  \caption{Patt.4}
\end{subfigure}\hfil
\begin{subfigure}{0.3\textwidth}
  \includegraphics[width=\linewidth]{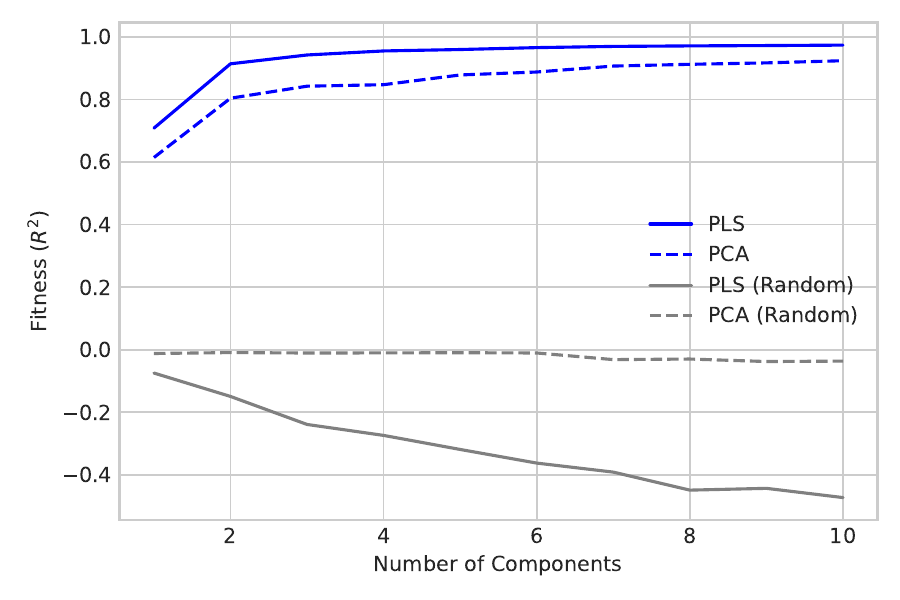}
  \caption{Patt.5}
\end{subfigure}\hfil 
\begin{subfigure}{0.3\textwidth}
  \includegraphics[width=\linewidth]{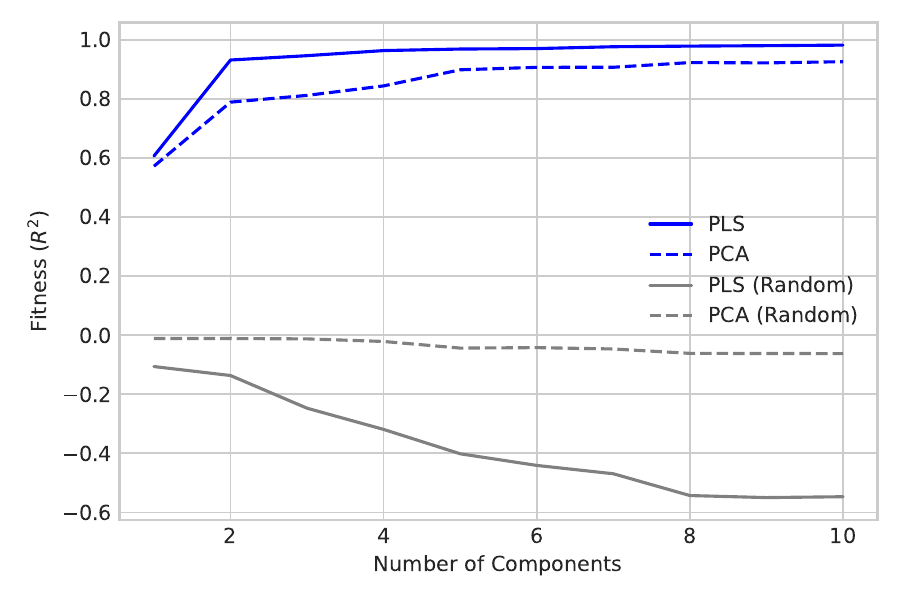}
  \caption{Patt.6}
\end{subfigure}\hfil 
\begin{subfigure}{0.3\textwidth}
  \includegraphics[width=\linewidth]{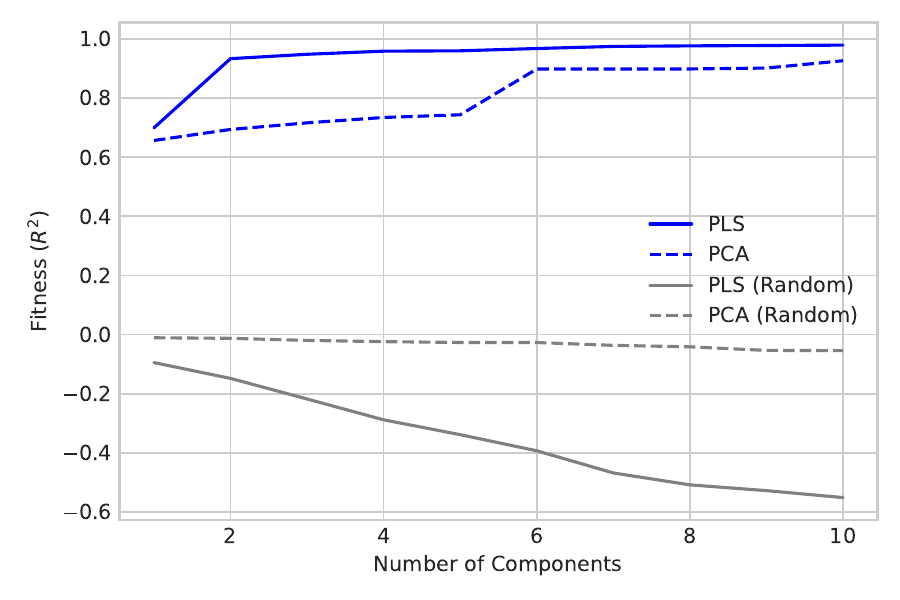}
  \caption{Patt.7}
\end{subfigure}\hfil
\begin{subfigure}{0.3\textwidth}
  \includegraphics[width=\linewidth]{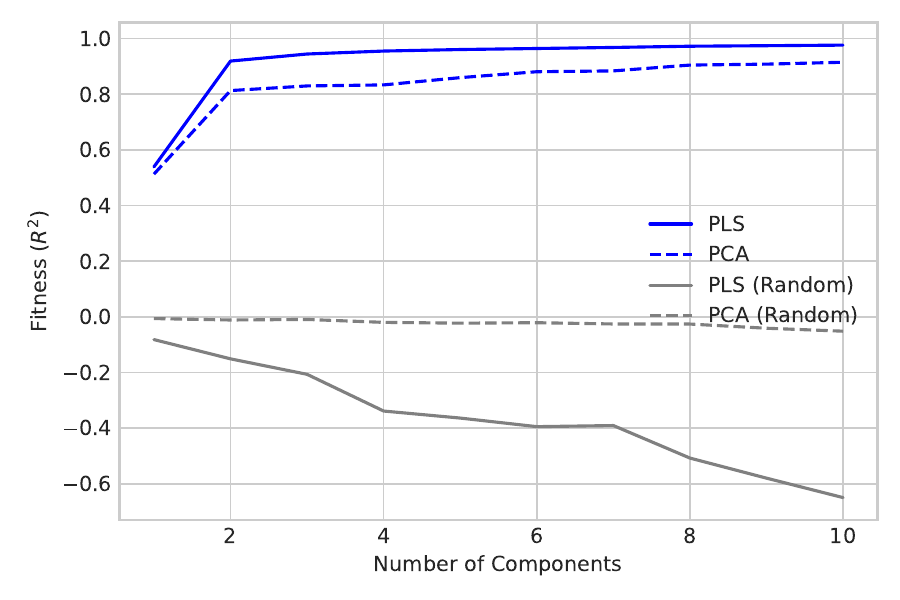}
  \caption{Patt.8}
\end{subfigure}\hfil 
\begin{subfigure}{0.3\textwidth}
  \includegraphics[width=\linewidth]{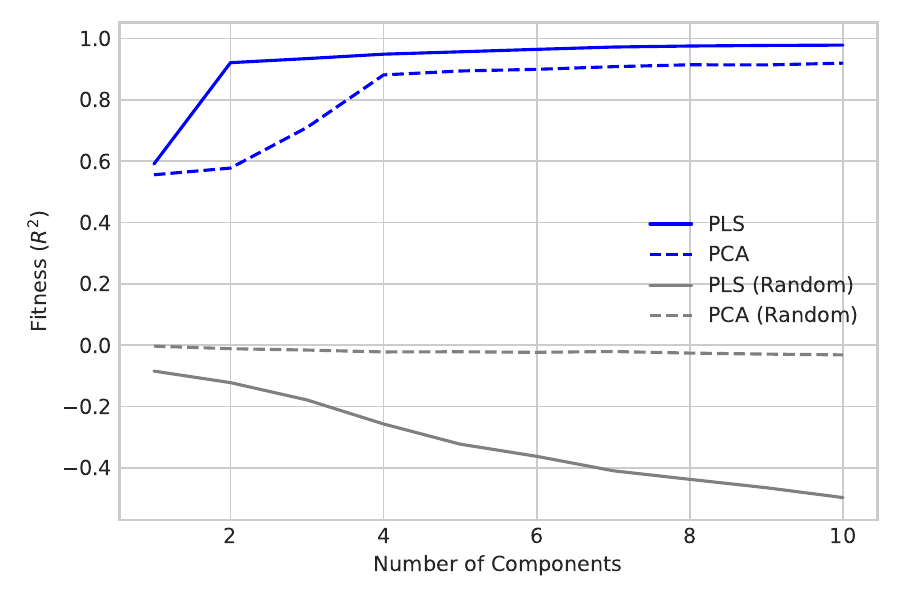}
  \caption{Patt.9}
\end{subfigure}\hfil 
\begin{subfigure}{0.3\textwidth}
  \includegraphics[width=\linewidth]{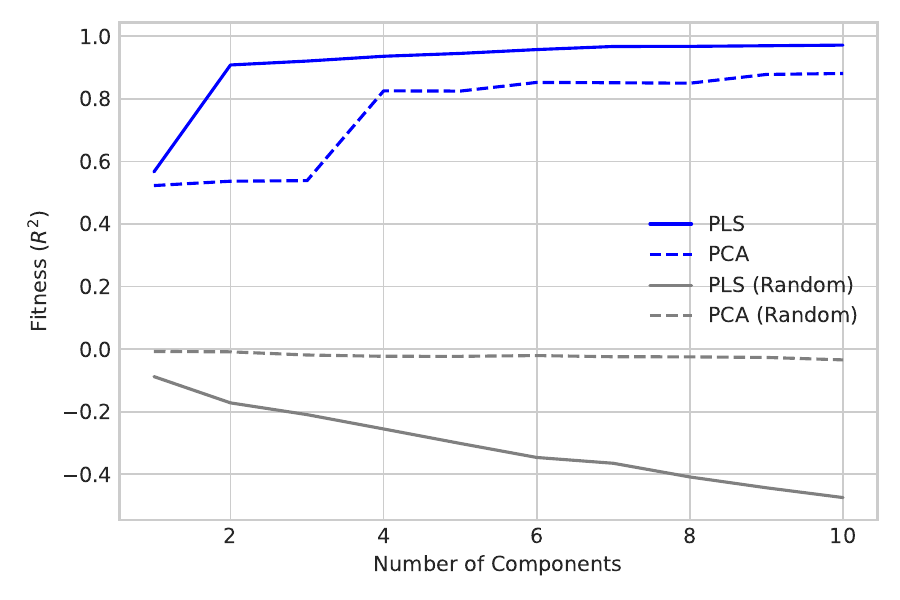}
  \caption{Patt.10}
\end{subfigure}\hfil
\begin{subfigure}{0.3\textwidth}
  \includegraphics[width=\linewidth]{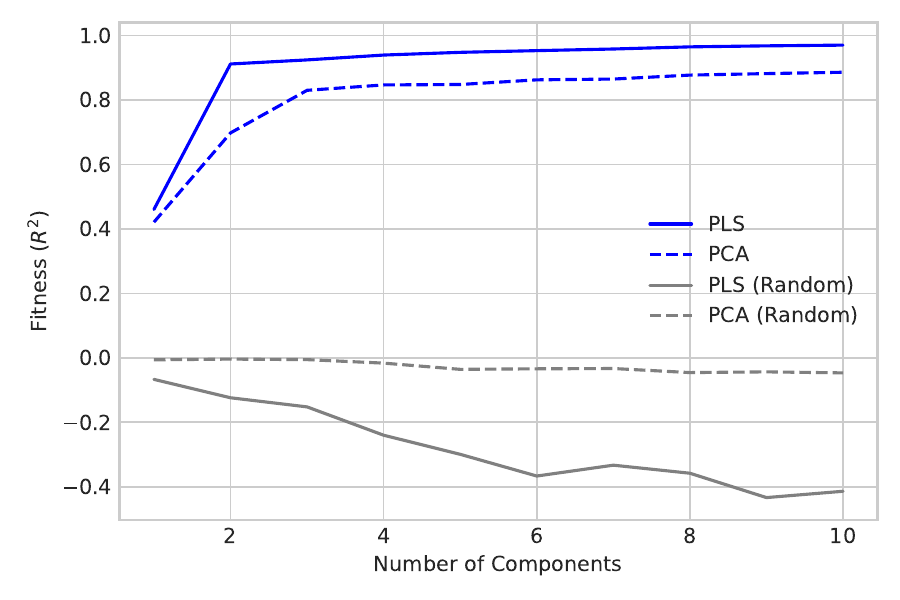}
  \caption{Patt.11}
\end{subfigure}\hfil 
\begin{subfigure}{0.3\textwidth}
  \includegraphics[width=\linewidth]{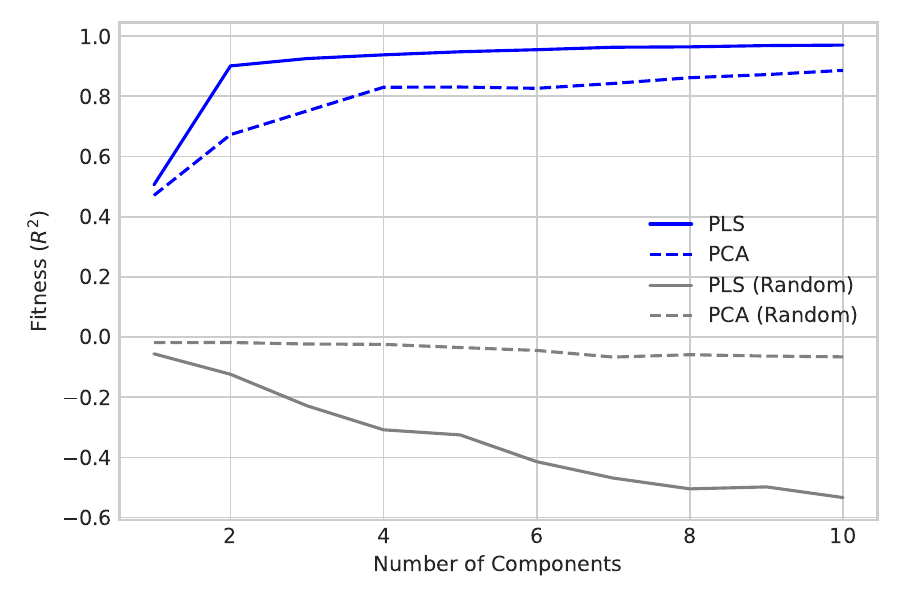}
  \caption{Patt.12}
\end{subfigure}\hfil 
\begin{subfigure}{0.3\textwidth}
  \includegraphics[width=\linewidth]{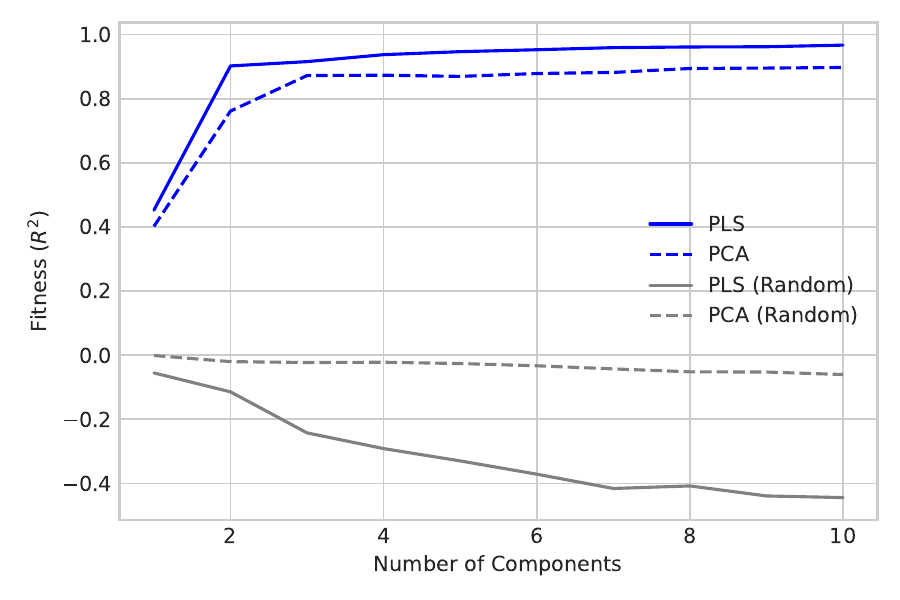}
  \caption{Patt.13}
\end{subfigure}\hfil 
\caption{Decoding performance of $[ei,ri]$ from activations of Qwen3-8B on $C_{city}$.}
\label{fig:pls_pattern_qwen_city}
\end{figure*}
\begin{figure*}[!htbp]
    \centering 
\begin{subfigure}{0.3\textwidth}
  \includegraphics[width=\linewidth]{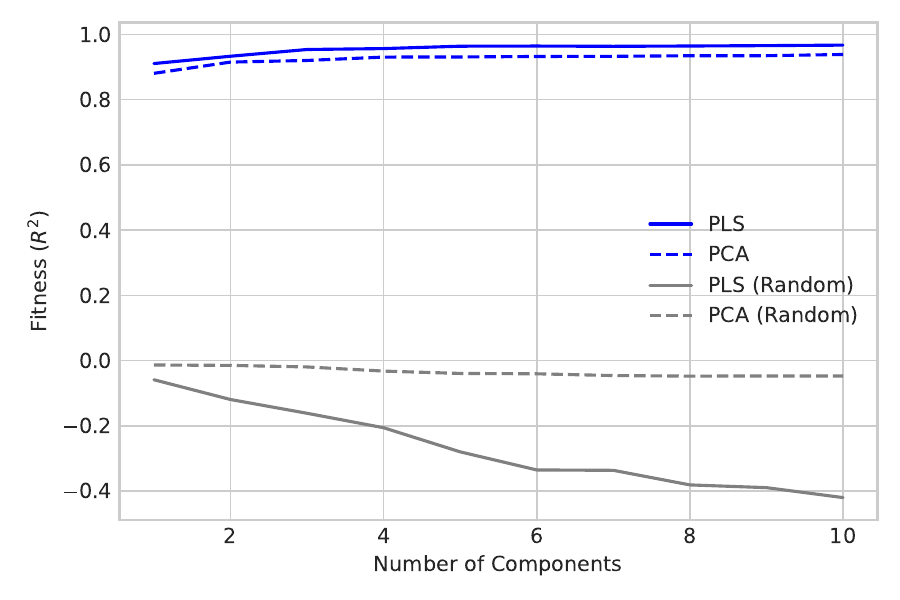}
  \caption{Patt.1}
\end{subfigure}\hfil
\begin{subfigure}{0.3\textwidth}
  \includegraphics[width=\linewidth]{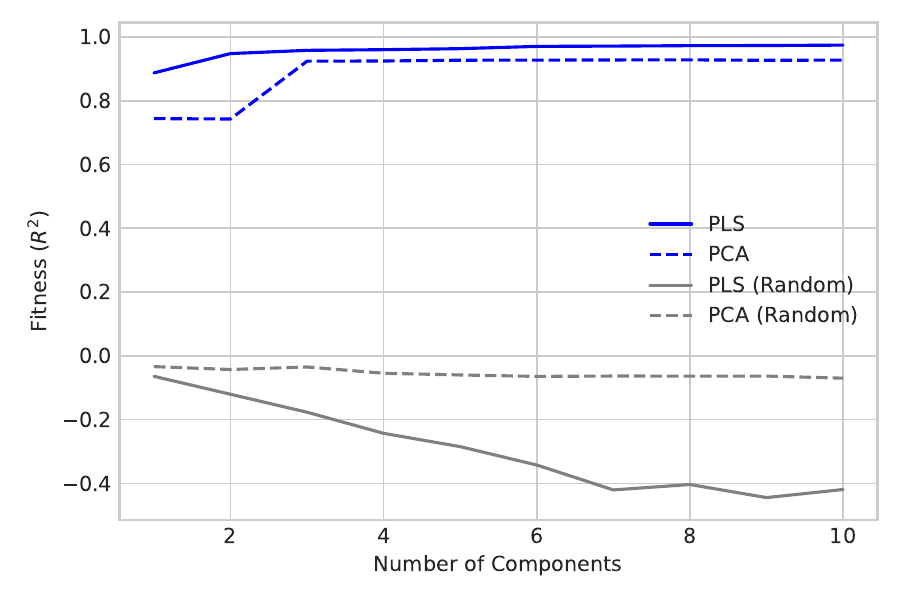}
  \caption{Patt.2}
\end{subfigure}\hfil 
\begin{subfigure}{0.3\textwidth}
  \includegraphics[width=\linewidth]{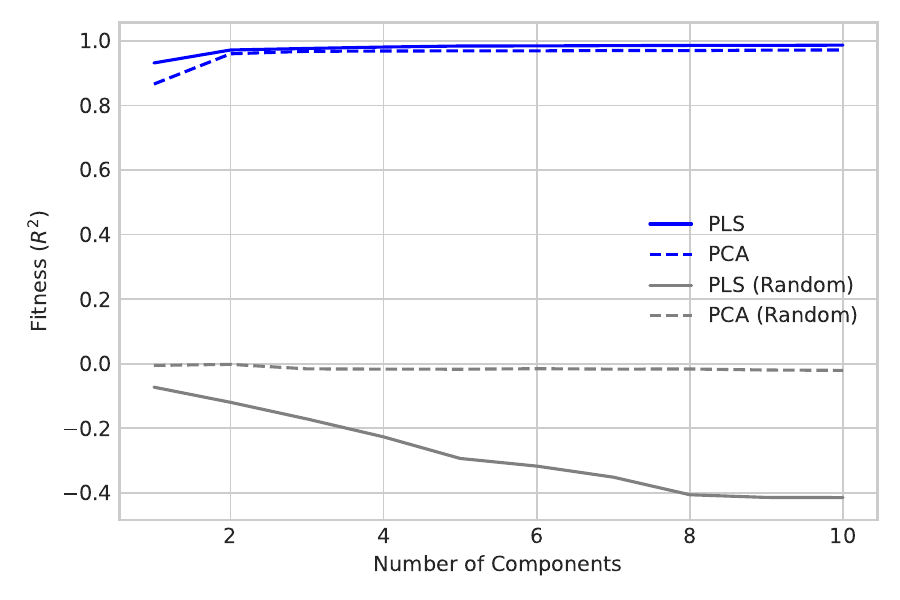}
  \caption{Patt.3}
\end{subfigure}\hfil 
\begin{subfigure}{0.3\textwidth}
  \includegraphics[width=\linewidth]{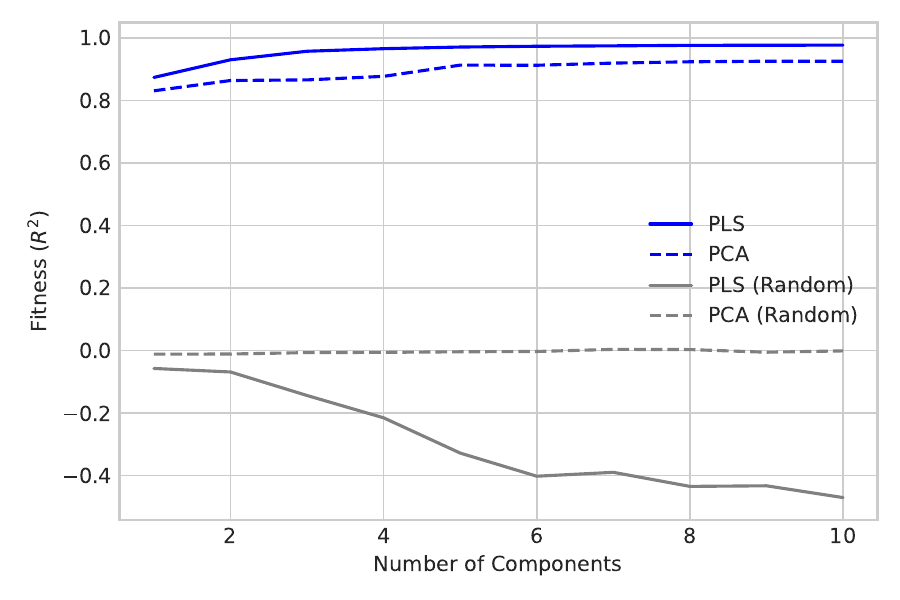}
  \caption{Patt.4}
\end{subfigure}\hfil
\begin{subfigure}{0.3\textwidth}
  \includegraphics[width=\linewidth]{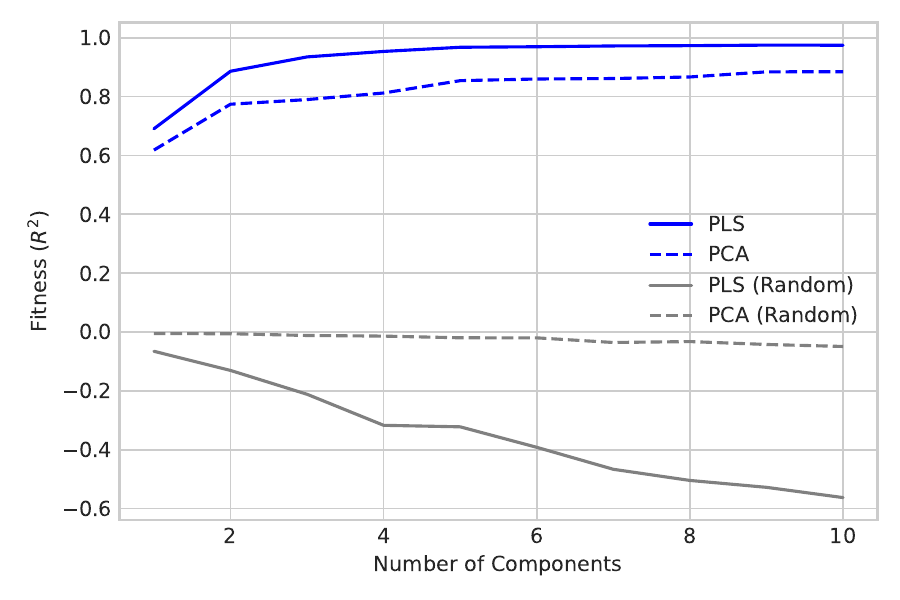}
  \caption{Patt.5}
\end{subfigure}\hfil 
\begin{subfigure}{0.3\textwidth}
  \includegraphics[width=\linewidth]{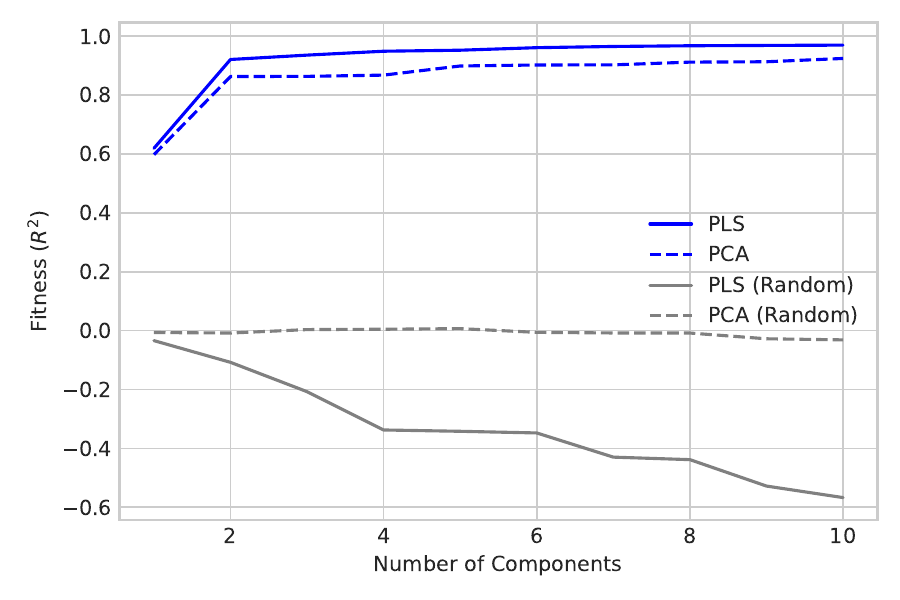}
  \caption{Patt.6}
\end{subfigure}\hfil 
\begin{subfigure}{0.3\textwidth}
  \includegraphics[width=\linewidth]{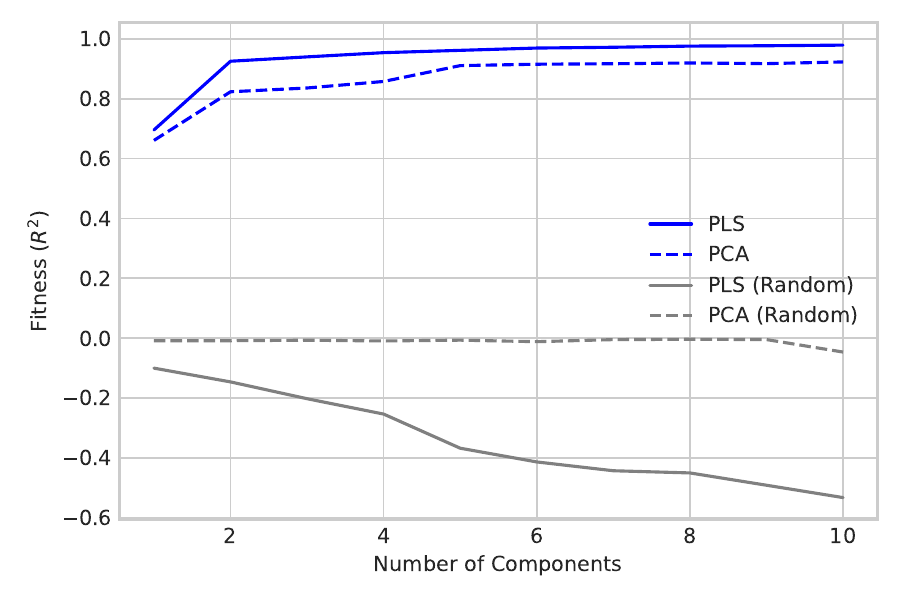}
  \caption{Patt.7}
\end{subfigure}\hfil
\begin{subfigure}{0.3\textwidth}
  \includegraphics[width=\linewidth]{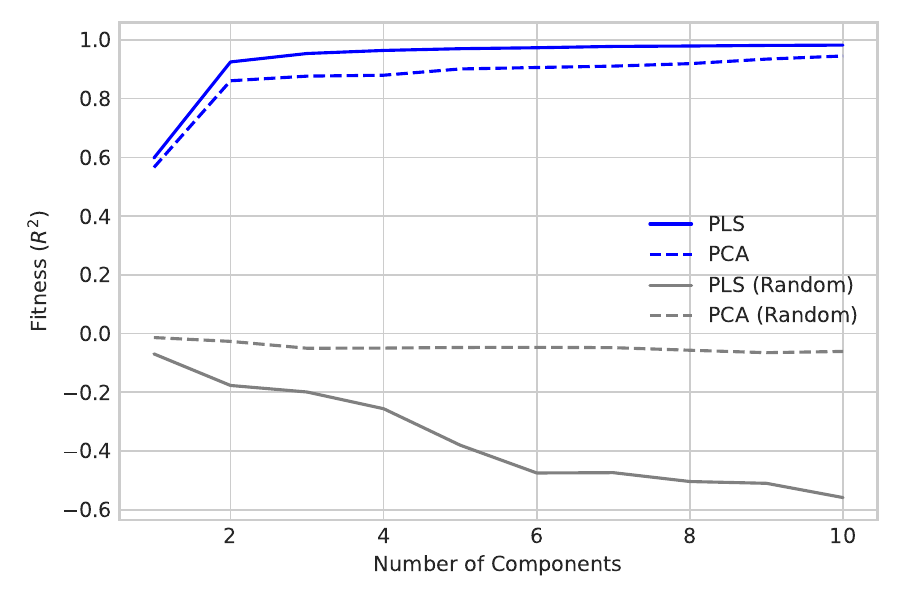}
  \caption{Patt.8}
\end{subfigure}\hfil 
\begin{subfigure}{0.3\textwidth}
  \includegraphics[width=\linewidth]{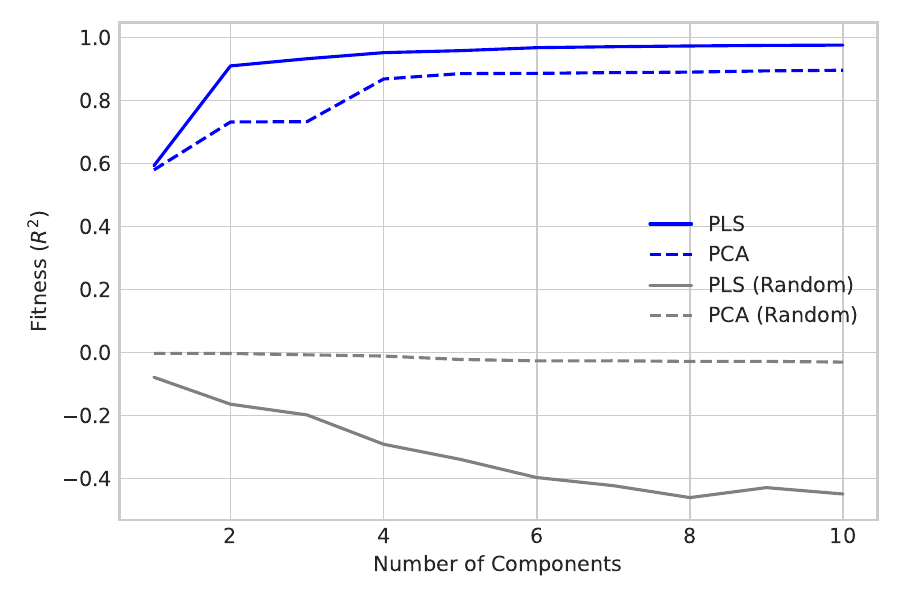}
  \caption{Patt.9}
\end{subfigure}\hfil 
\begin{subfigure}{0.3\textwidth}
  \includegraphics[width=\linewidth]{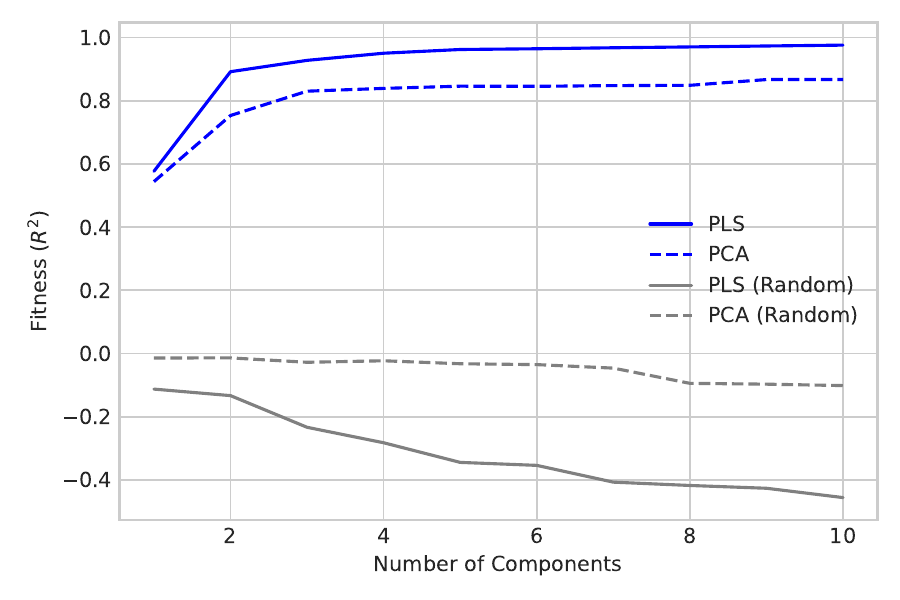}
  \caption{Patt.10}
\end{subfigure}\hfil
\begin{subfigure}{0.3\textwidth}
  \includegraphics[width=\linewidth]{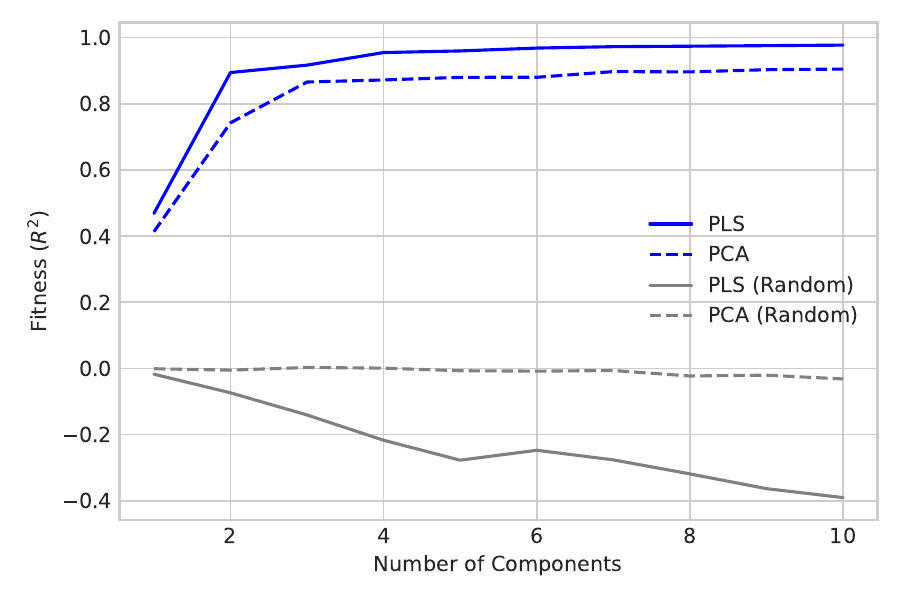}
  \caption{Patt.11}
\end{subfigure}\hfil 
\begin{subfigure}{0.3\textwidth}
  \includegraphics[width=\linewidth]{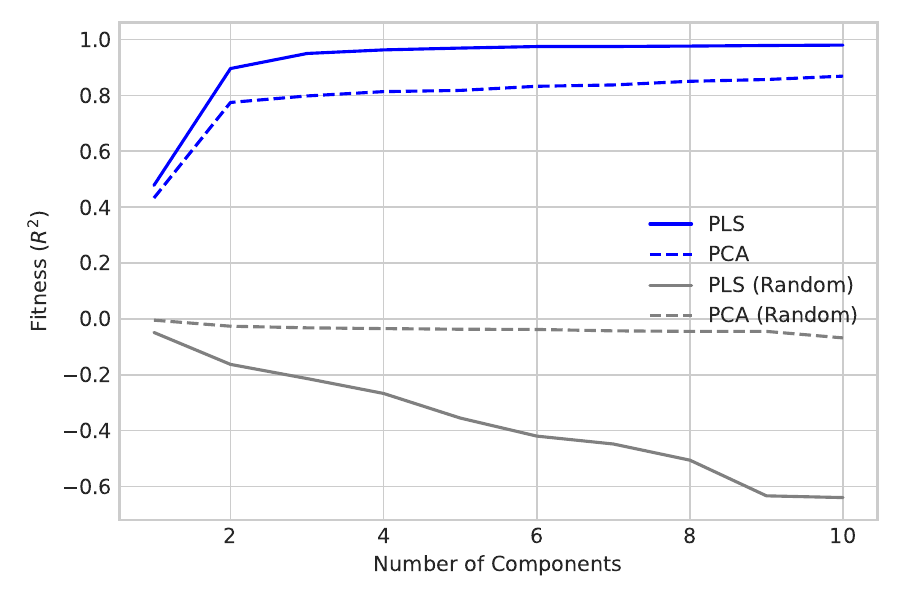}
  \caption{Patt.12}
\end{subfigure}\hfil 
\begin{subfigure}{0.3\textwidth}
  \includegraphics[width=\linewidth]{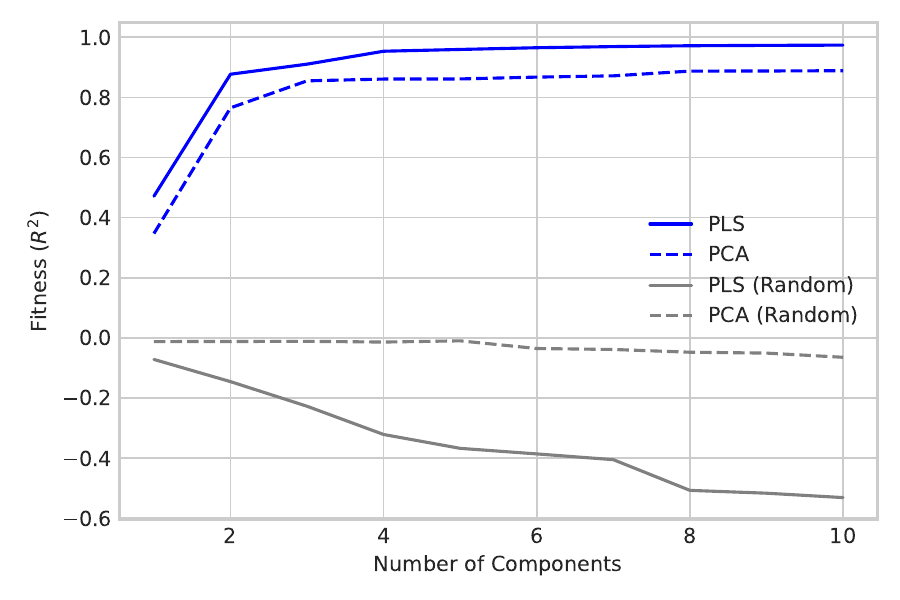}
  \caption{Patt.13}
\end{subfigure}\hfil 
\caption{Decoding performance of $[ei,ri]$ from activations of Qwen3-8B on $C_{country}$}
\label{fig:pls_pattern_qwen_country}
\end{figure*}
\clearpage

\begin{figure*}[htb]
    \centering 
\begin{subfigure}{0.485\textwidth}
  \includegraphics[width=\linewidth]{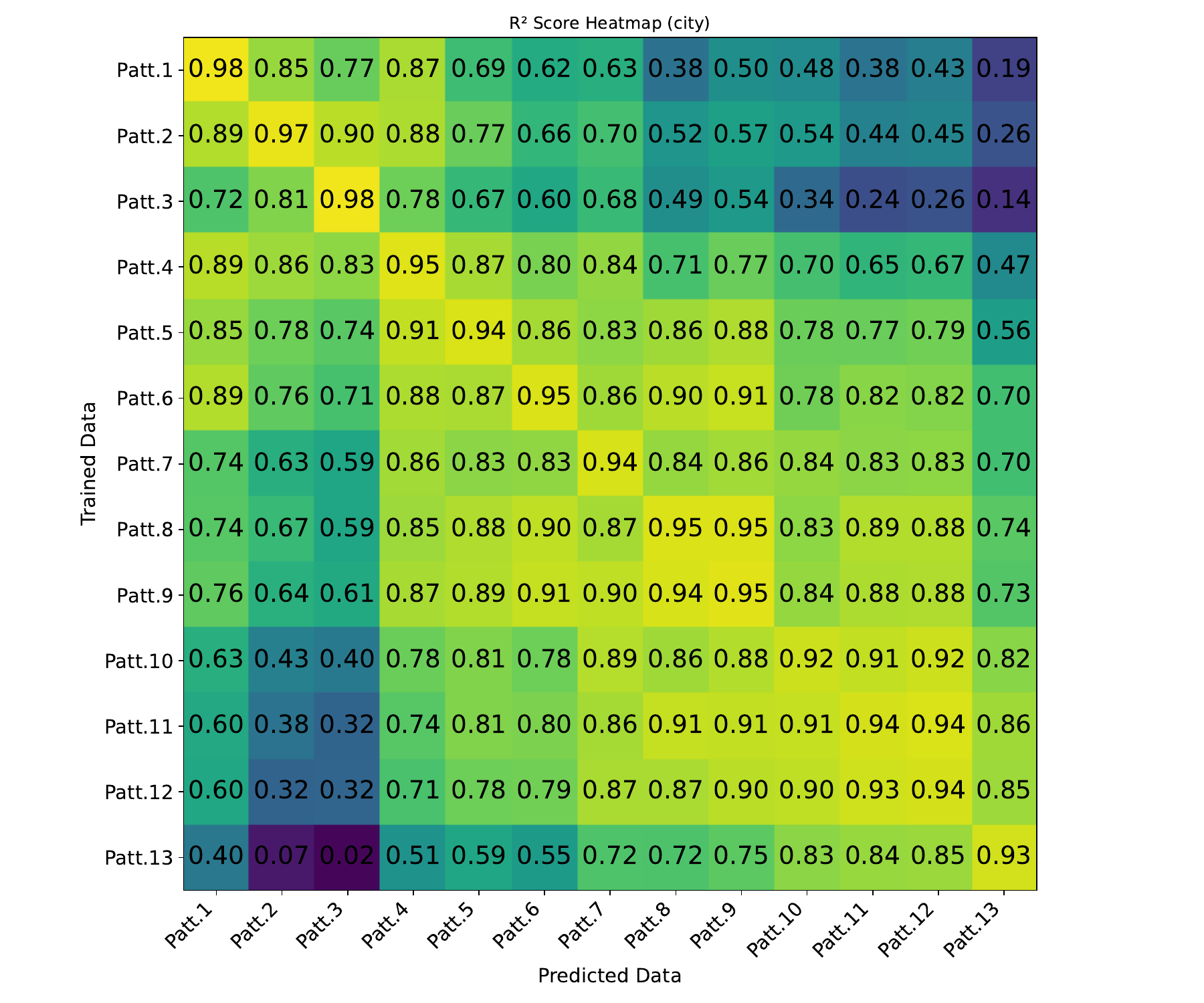}
  \caption{Decoding performance of $[ei,ri]$}
\end{subfigure}\hfil 
\begin{subfigure}{0.485\textwidth}
  \includegraphics[width=\linewidth]{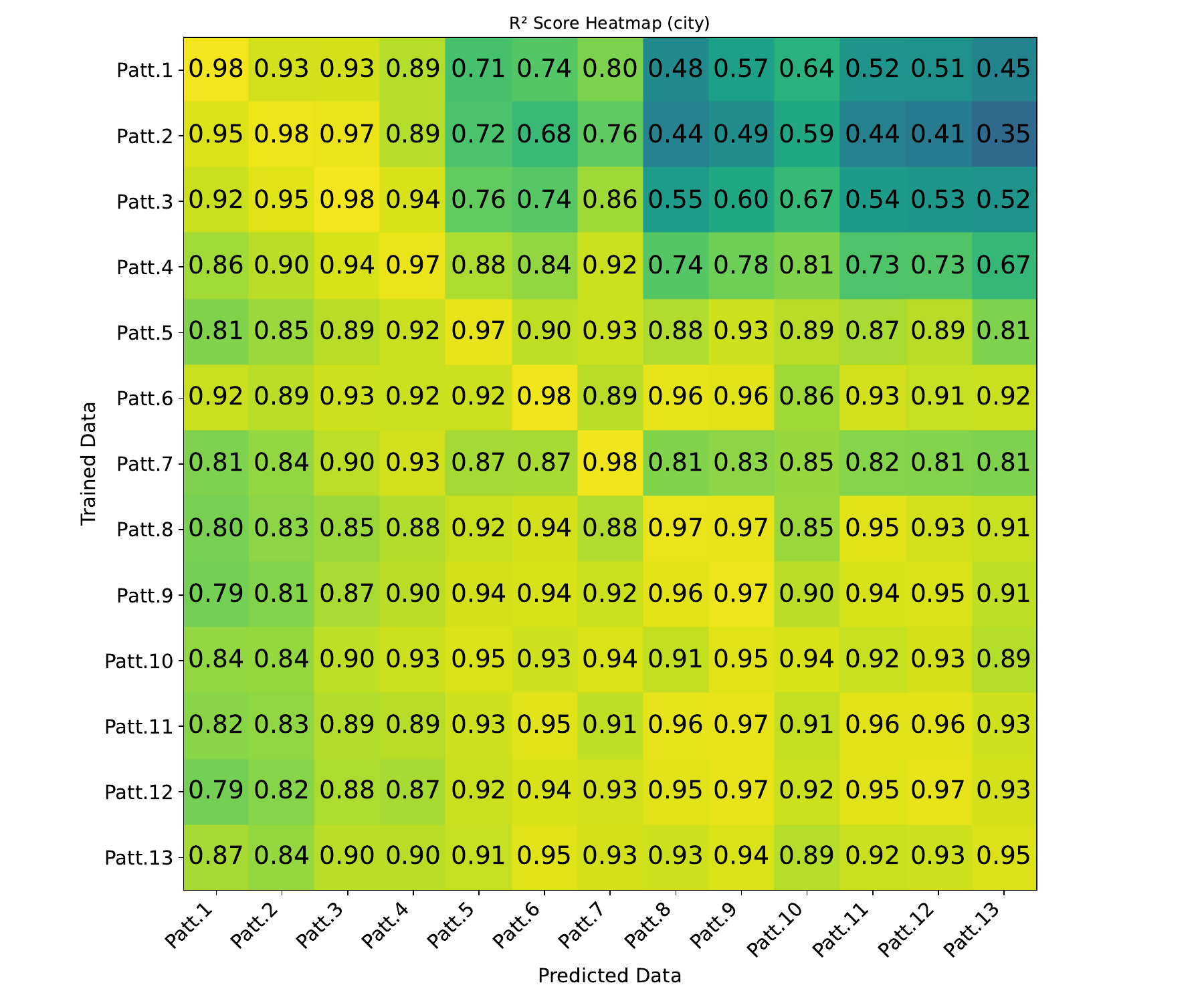}
  \caption{Decoding performance of $[ri]$}
\end{subfigure}\hfil 
\caption{Cross patterns $R^2$ scores for index prediction from Llama-8B-Instruct on $C_{city}$.}
\label{fig:consistency_pattern_llama_city}
\end{figure*}
\begin{figure*}[htb]
    \centering 
\begin{subfigure}{0.485\textwidth}
  \includegraphics[width=\linewidth]{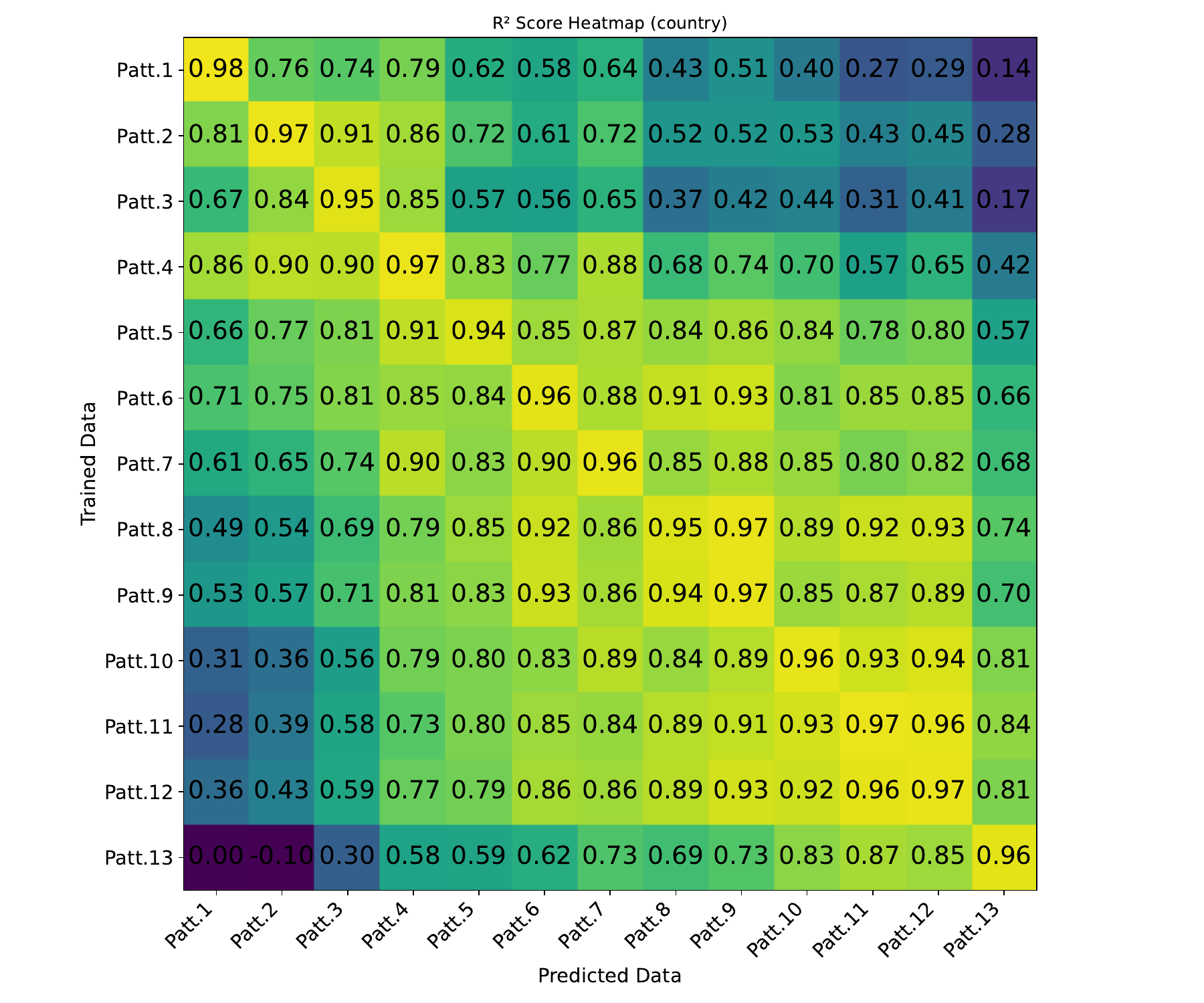}
  \caption{Decoding performance of $[ei,ri]$}
\end{subfigure}\hfil 
\begin{subfigure}{0.485\textwidth}
  \includegraphics[width=\linewidth]{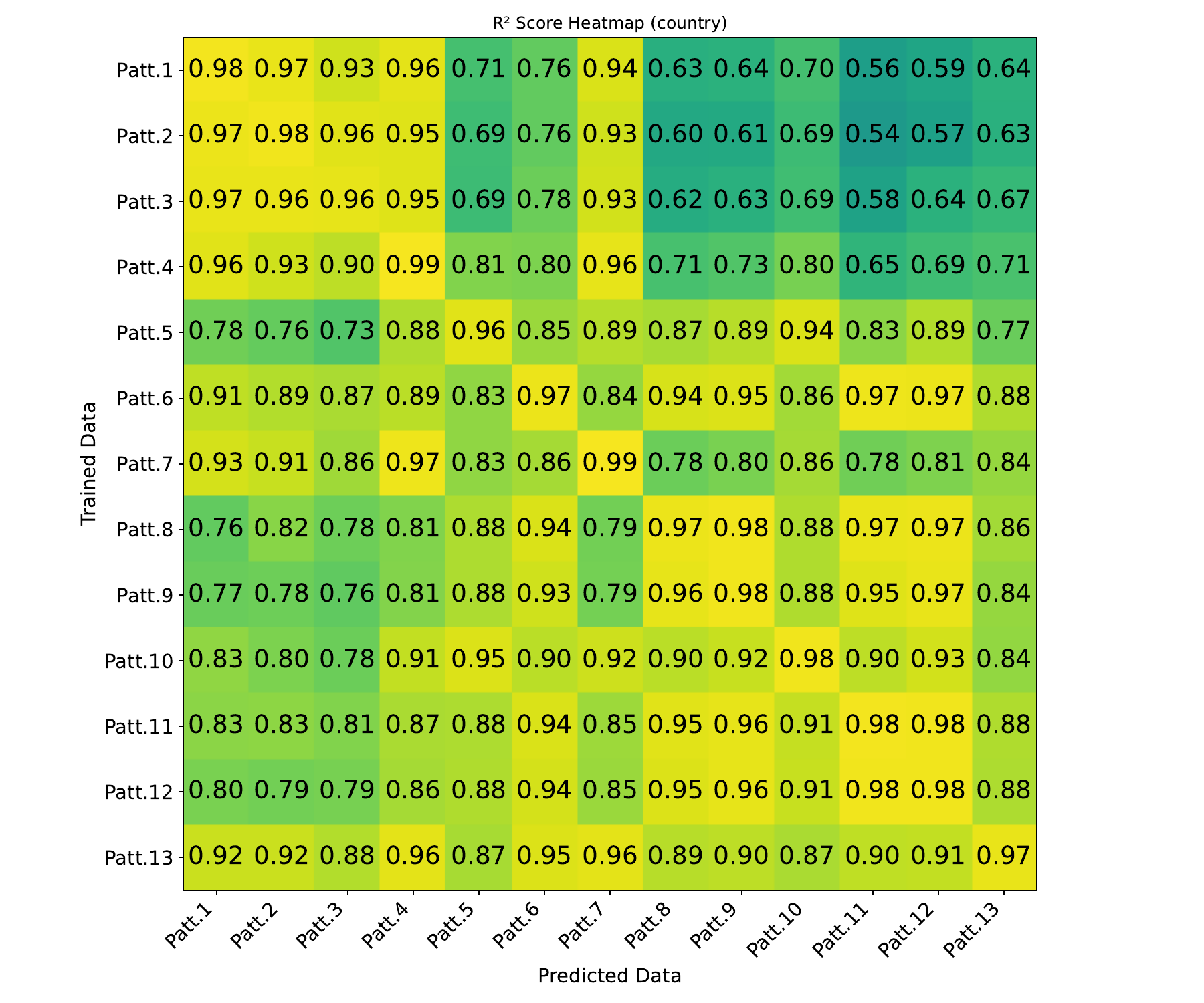}
  \caption{Decoding performance of $[ri]$}
\end{subfigure}\hfil 
\caption{Cross patterns $R^2$ scores for index prediction from Llama-8B-Instruct on $C_{country}$.}
\label{fig:consistency_pattern_llama_country}
\end{figure*}
\begin{figure*}[htb]
    \centering 
\begin{subfigure}{0.485\textwidth}
  \includegraphics[width=\linewidth]{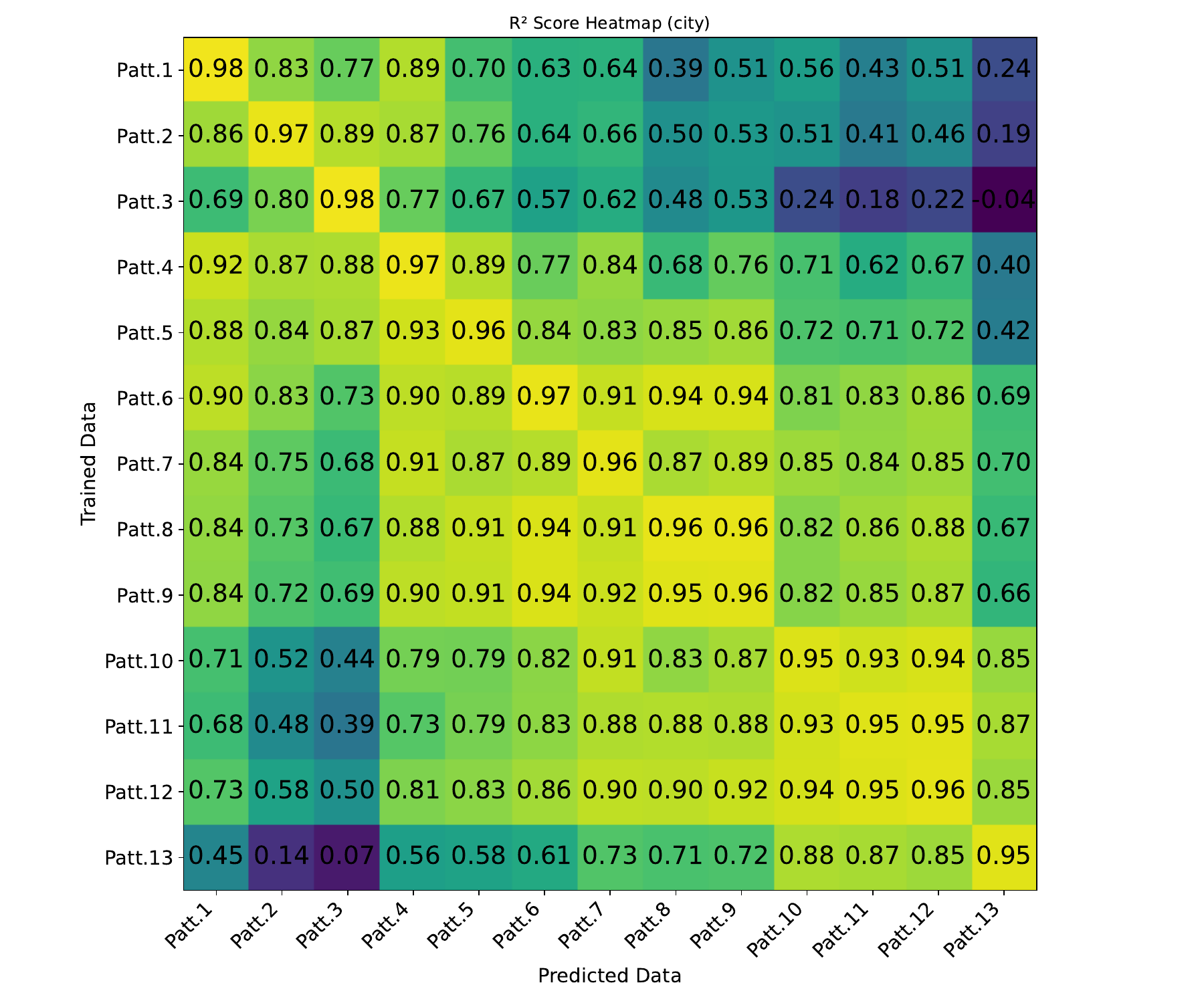}
  \caption{Decoding performance of $[ei,ri]$}
\end{subfigure}\hfil 
\begin{subfigure}{0.485\textwidth}
  \includegraphics[width=\linewidth]{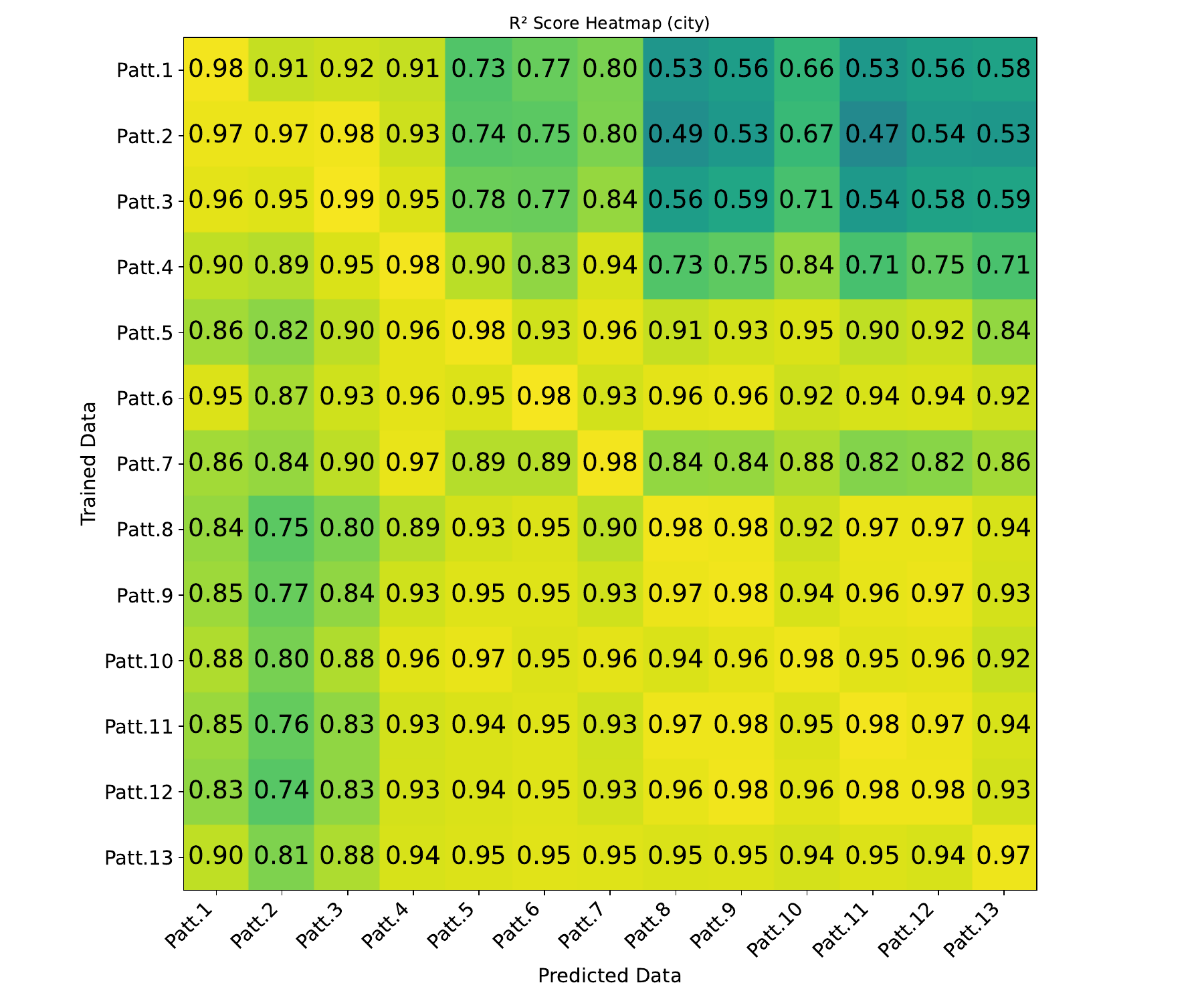}
  \caption{Decoding performance of $[ri]$}
\end{subfigure}\hfil 
\caption{Cross patterns $R^2$ scores for index prediction from Qwen-8B on $C_{city}$}
\label{fig:consistency_pattern_qwen_city}
\end{figure*}

\begin{figure*}[htb]
    \centering 
\begin{subfigure}{0.485\textwidth}
  \includegraphics[width=\linewidth]{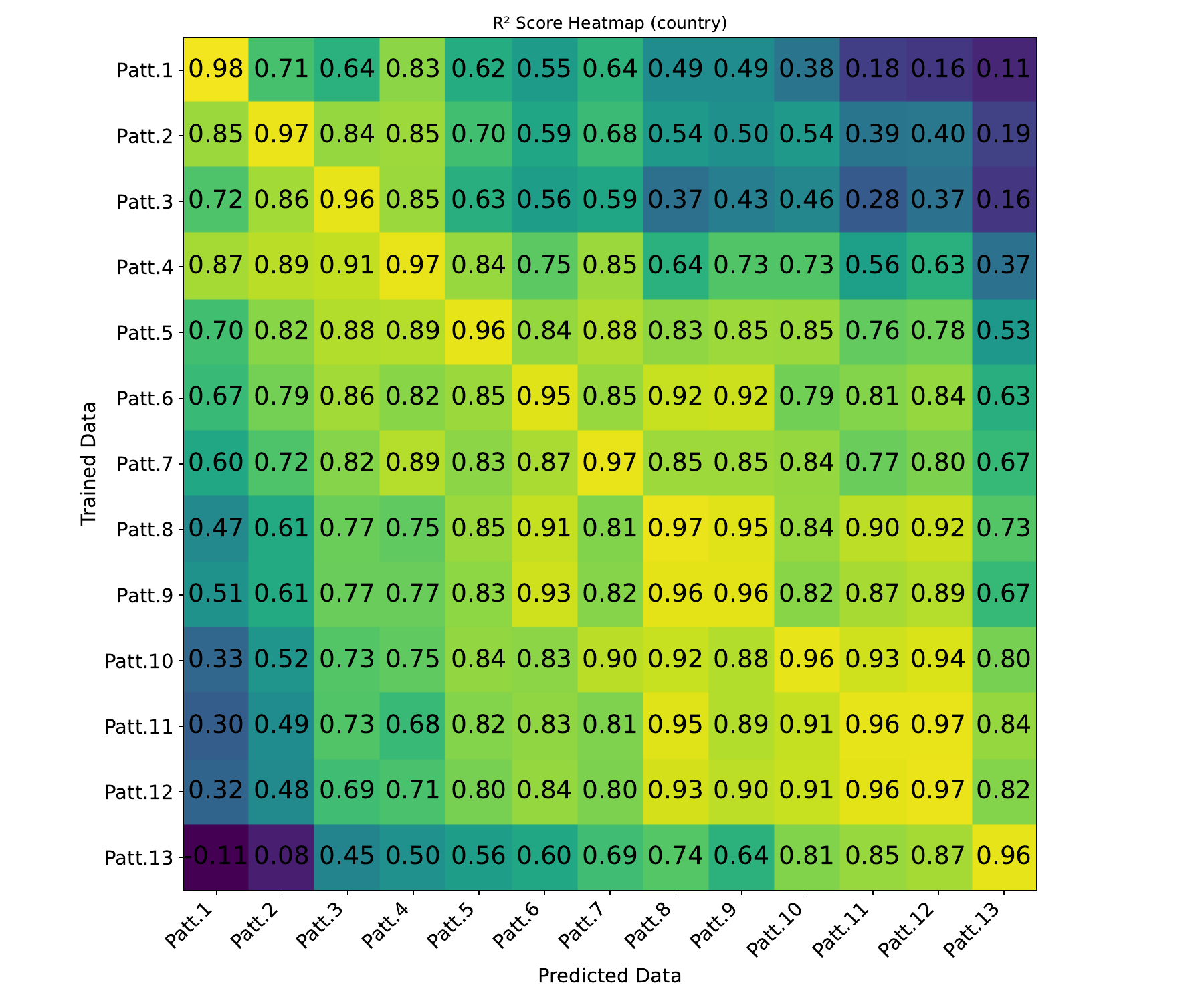}
  \caption{Decoding performance of $[ei,ri]$}
\end{subfigure}\hfil 
\begin{subfigure}{0.485\textwidth}
  \includegraphics[width=\linewidth]{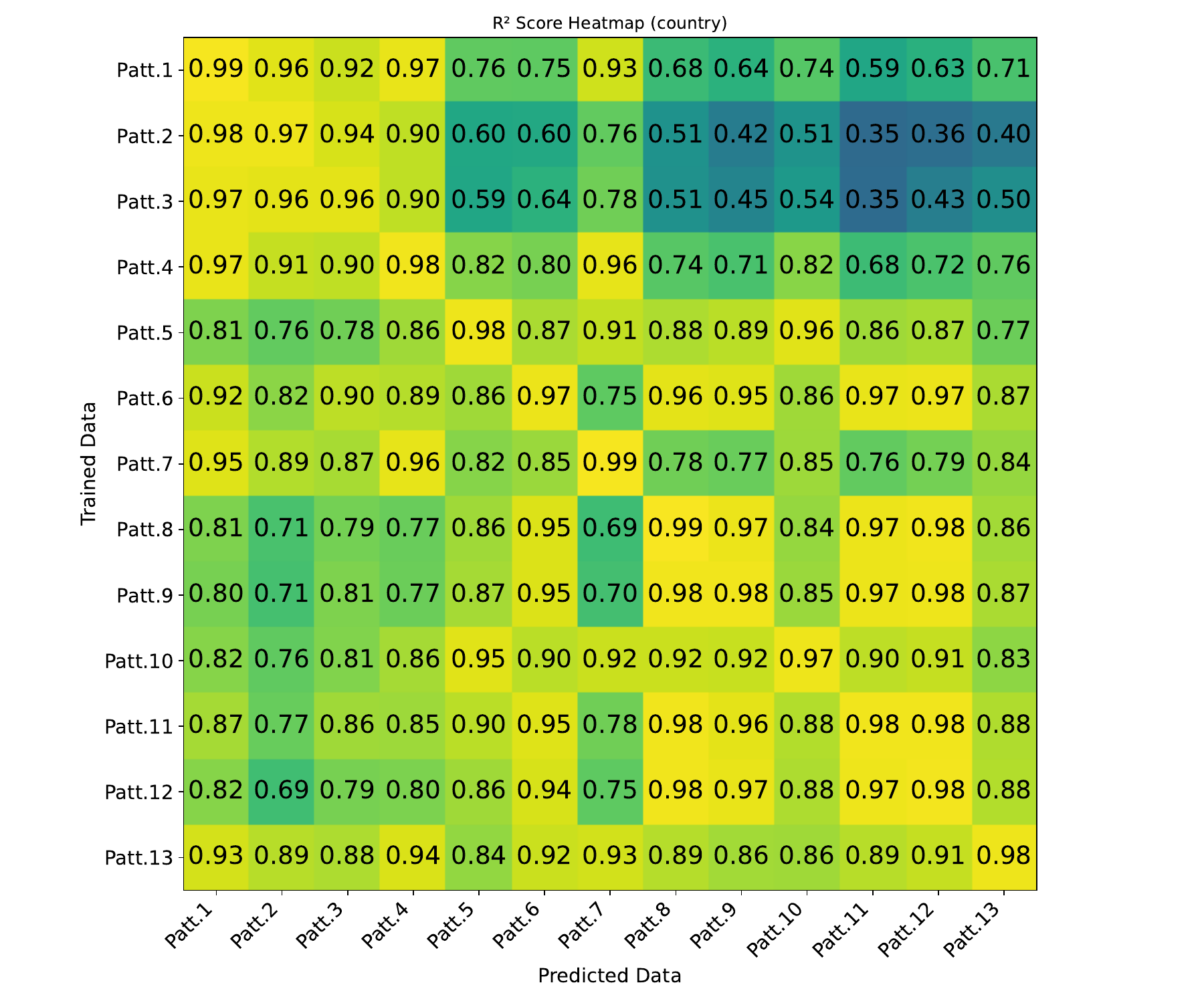}
  \caption{Decoding performance of $[ri]$}
\end{subfigure}\hfil 
\caption{Cross patterns $R^2$ scores for index prediction from Qwen-8B on $C_{country}$.}
\label{fig:consistency_pattern_qwen_country}
\end{figure*}

To further examine whether the learned representations generalize across structural variations, we conduct cross-pattern evaluations by applying a projection matrix trained on one pattern to predict indices in another. The results, shown in Figure~\ref{fig:consistency_pattern_llama_city}, \ref{fig:consistency_pattern_llama_country}, \ref{fig:consistency_pattern_qwen_city}, and \ref{fig:consistency_pattern_qwen_country}, indicate that predictive performance transfers well across patterns despite differences in relational selection, repetition, attribute density, and overall structural complexity.

One exception arises for the simplest templates (Patt.1–3), where the entity and relation indices partially overlap. In these cases, the attribute can be uniquely identified using the relation index $ri$ alone, making the entity index $ei$ redundant. This is supported by the PLS results in Figure~\ref{fig:pls_pattern_llama_city}, which show that a single component achieves a high $R^2$ score for simple patterns such as Patt.~1–4. As a result, a projection matrix trained on more complex patterns (e.g., Patt.13), which learns to jointly decode $[ei, ri]$, does not directly transfer when evaluated on Patt.1–3. However, when decoding is restricted to $ri$ only, the projection matrix again shows strong cross-pattern performance. Therefore, the lower scores observed when transferring from Patt.13 to Patt.1–3 do not contradict our main observation that the CBR representation generalizes across structural variations. Conversely, the lower performance for Patt.1–3 to Patt.8–13 may result from the limited structural complexity of the simpler patterns. Because Patt.1–3 lack the richer relational structures present in patterns such as Patt.10, the learned projection may not generalize to more complex templates.

Taken together, these observations suggest that the overall structure of CBR subspace is stable across a wide range of discourse configurations. The persistence of accurate index prediction under these heterogeneous patterns indicates that the discovered structure does not arise from a particular template design, but instead reflects a broader organizational property of how LLMs encode entity–relation bindings.

\clearpage

\subsection{Sampling CBR Subspace across Contexts and LLMs}
\label{sec:irs_sampling_across}

\begin{figure*}[!htbp]
    \centering 
\begin{subfigure}{0.4\textwidth}
  \includegraphics[width=\linewidth]{graph/grid_llama_city.png}
  \caption{$C_{city}$}
\end{subfigure}\hfil 
\begin{subfigure}{0.4\textwidth}
  \includegraphics[width=\linewidth]{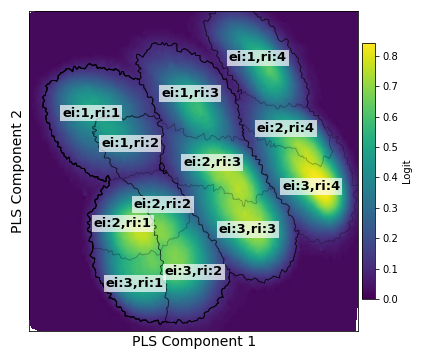}
  \caption{$C_{country}$}
\end{subfigure}\hfil 
\begin{subfigure}{0.3\textwidth}
  \includegraphics[width=\linewidth]{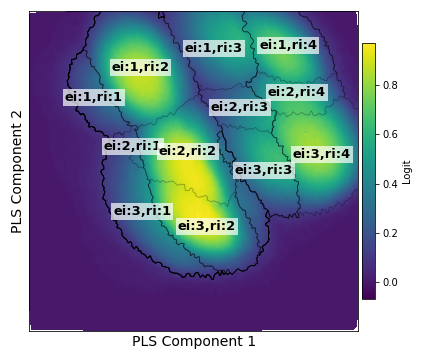}
  \caption{$C_{relation}$}
\end{subfigure}\hfil
\begin{subfigure}{0.3\textwidth}
  \includegraphics[width=\linewidth]{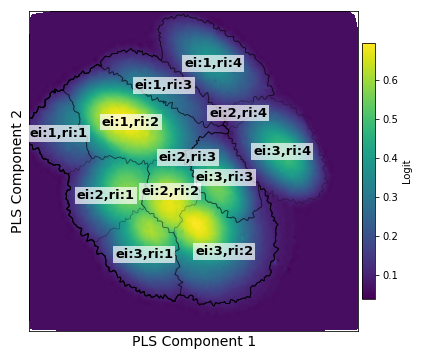}
  \caption{$C_{job}$}
\end{subfigure}\hfil 
\begin{subfigure}{0.3\textwidth}
  \includegraphics[width=\linewidth]{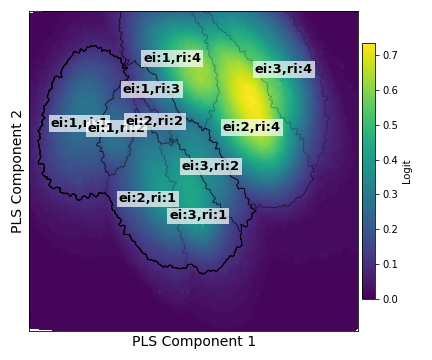}
  \caption{$C_{object}$}
\end{subfigure}\hfil 
\caption{Logit landscape of attribute predictions in
the CBR subspace across context on Llama3-8B-Instruct.}
\label{fig:grid_llama}
\end{figure*}

\begin{figure*}[!htbp]
    \centering 
\begin{subfigure}{0.4\textwidth}
  \includegraphics[width=\linewidth]{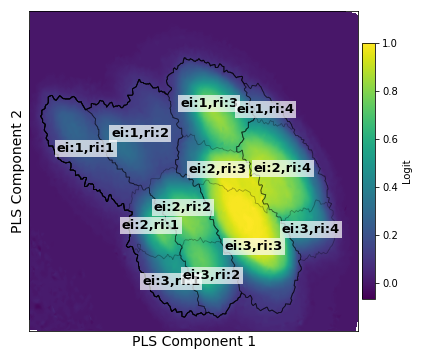}
  \caption{$C_{city}$}
\end{subfigure}\hfil 
\begin{subfigure}{0.4\textwidth}
  \includegraphics[width=\linewidth]{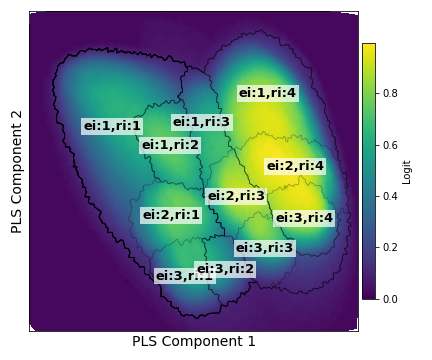}
  \caption{$C_{country}$}
\end{subfigure}\hfil 
\begin{subfigure}{0.3\textwidth}
  \includegraphics[width=\linewidth]{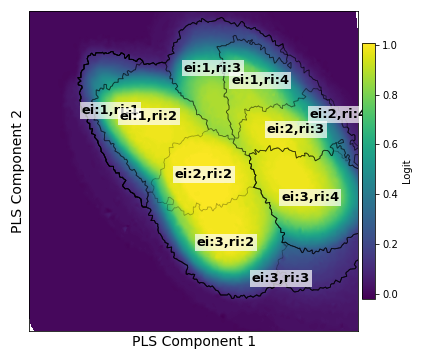}
  \caption{$C_{relation}$}
\end{subfigure}\hfil
\begin{subfigure}{0.3\textwidth}
  \includegraphics[width=\linewidth]{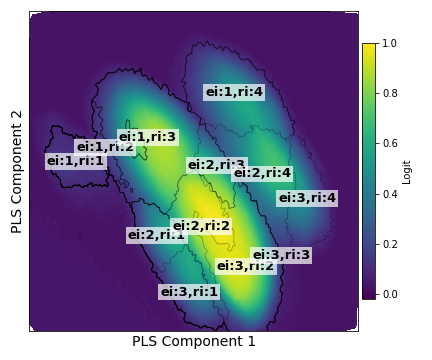}
  \caption{$C_{job}$}
\end{subfigure}\hfil 
\begin{subfigure}{0.3\textwidth}
  \includegraphics[width=\linewidth]{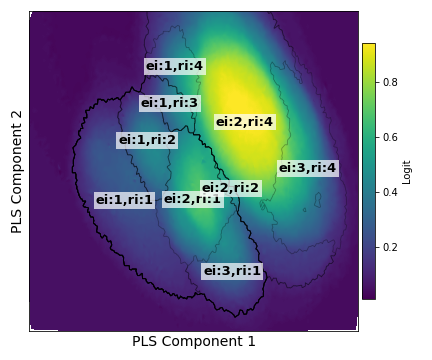}
  \caption{$C_{object}$}
\end{subfigure}\hfil 
\caption{Logit landscape of attribute predictions in
the CBR subspace across context on Qwen3-8B.}
\label{fig:grid_qwen}
\end{figure*}

Section~\secref{sec:causal_intervention} reveals that the CBR subspace is partitioned into regions representing a Voronoi diagram (or cells), where each region corresponds to a specific entity–relation index pair $[ei,ri]$.
To evaluate the generality of this Voronoi-like structure, we perform the same CBR-based sampling and visualization procedure across contexts, shown in Figure~\ref{fig:grid_llama}, and on a different model family, Qwen3-8B, shown in Figure~\ref{fig:grid_qwen}. In all tested settings, we observe the same characteristic partitioning of the CBR subspace into index-specific regions, indicating that this binding geometry is not context or model specific. This suggests that the Voronoi-like organization of the CBR subspace is a general property of LLMs.

In addition, we also observe that in some contexts, certain entity–relation regions appear to be missing. This likely results from limited sampling density, the low dimensionality of the sampled subspace and improper hyperparameter selection, which may prevent full coverage of all potential regions in those cases. This is further analyzed in the next section.


\clearpage

\subsection{Analysis of Logit Score along Entity and Relation Index Directions}
\label{sec:grid_logit_analysis}

\begin{figure*}[!htbp]
    \centering 
\begin{subfigure}{0.475\textwidth}
  \includegraphics[width=\linewidth]{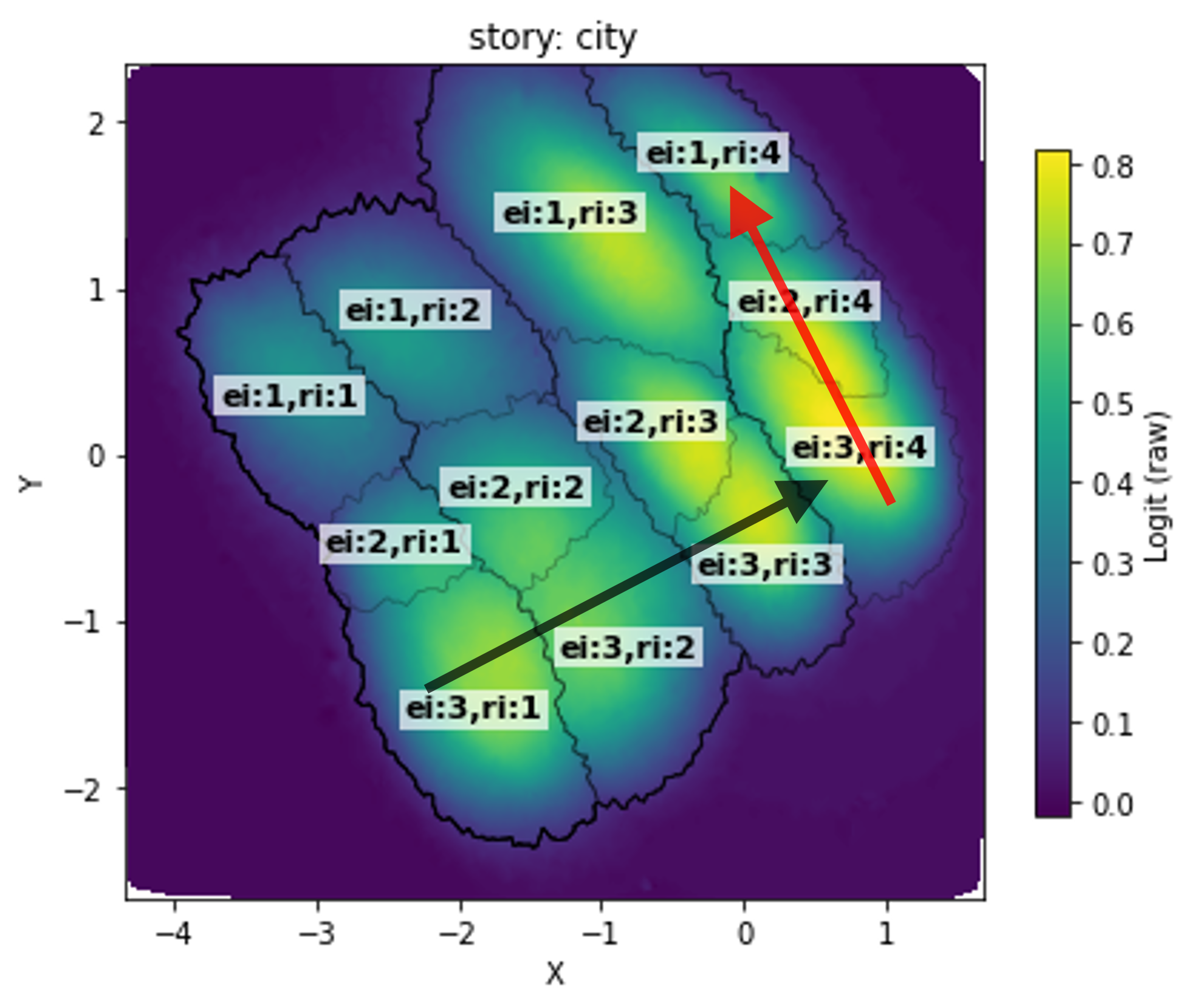}
  \caption{Llama3-8B-Instruct.}
  \label{fig:grid_direction_llama_city}
\end{subfigure}\hfil 
\begin{subfigure}{0.475\textwidth}
  \includegraphics[width=\linewidth]{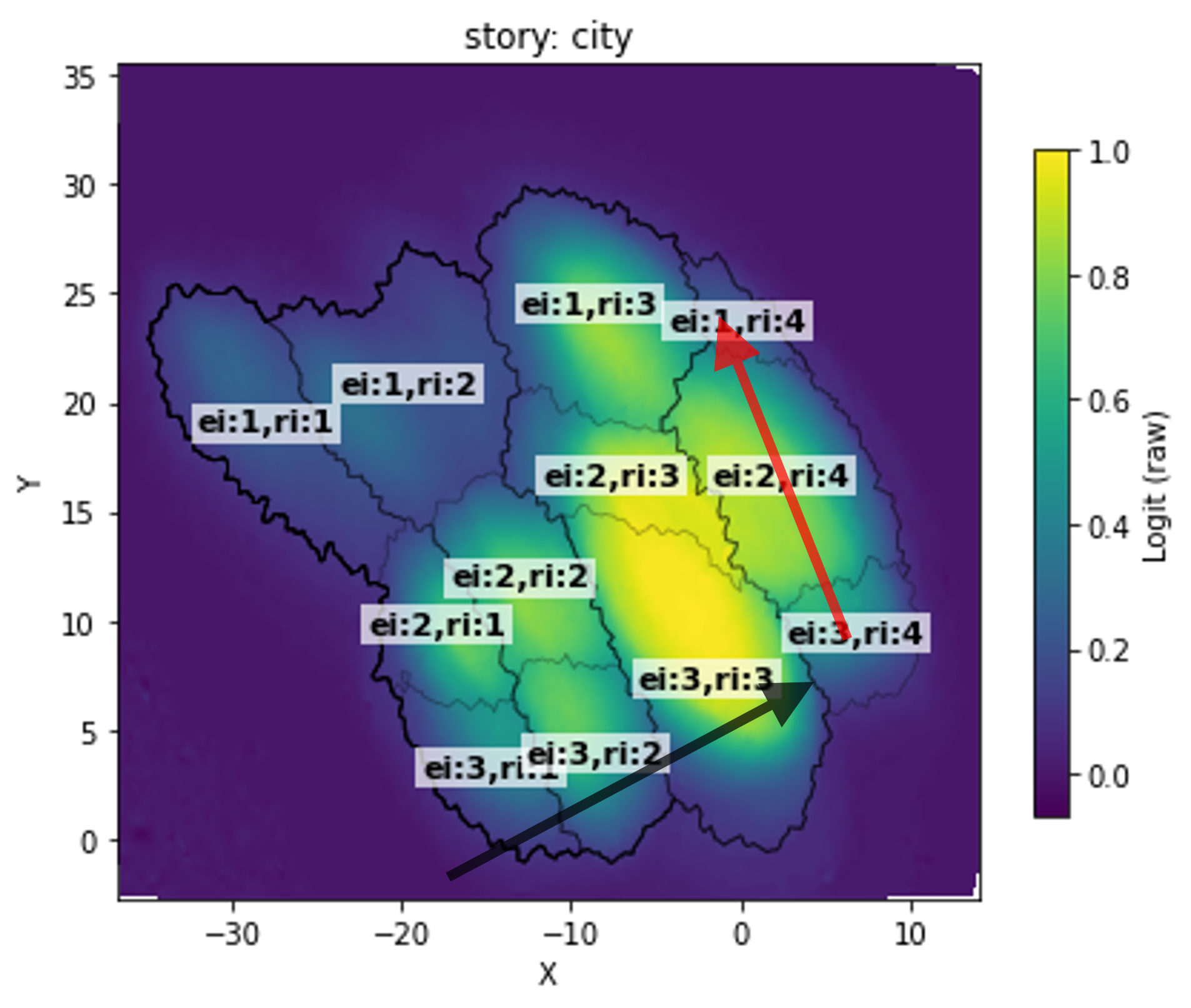}
  \caption{Qwen3-8B.}
  \label{fig:grid_direction_qwen_city}
\end{subfigure}\hfil 
\caption{$ei$ (red arrow) and $ri$ (black arrow) directions on $C_{city}$}
\label{fig:grid_dicreciton}
\end{figure*}
\begin{figure*}[!htbp]
    \centering 
\begin{subfigure}{0.23\textwidth}
  \includegraphics[width=\linewidth]{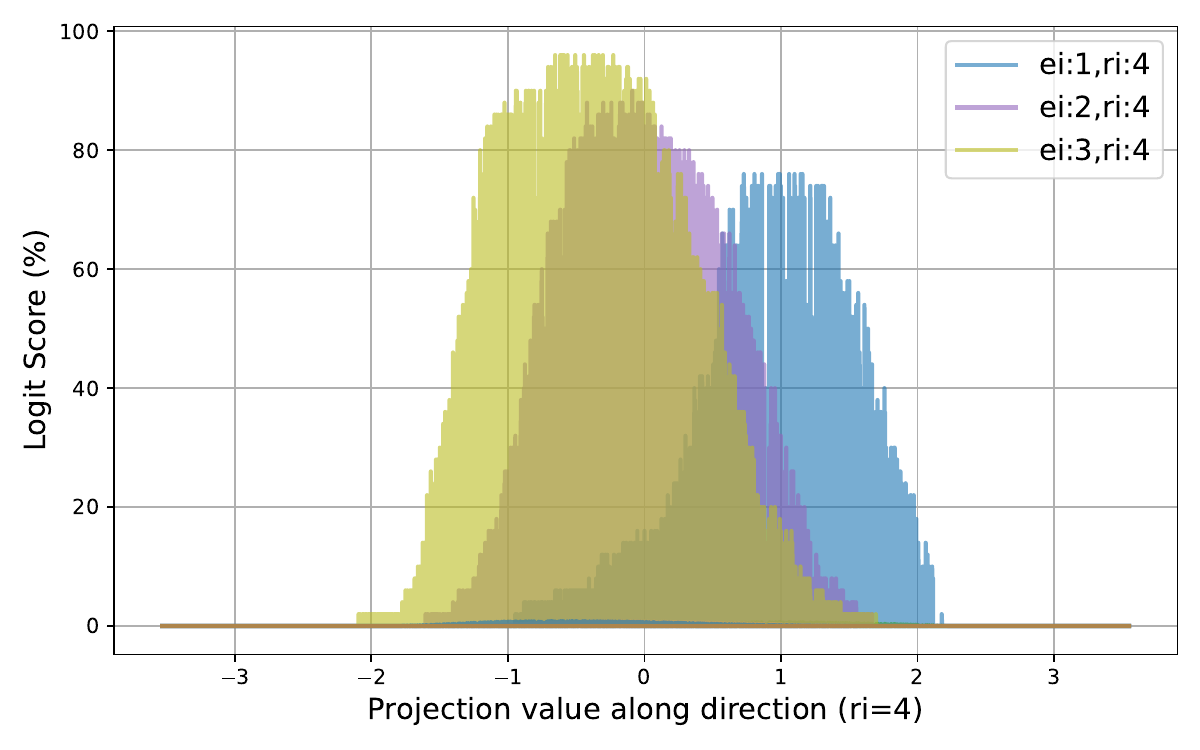}
  \caption{ $ei$ direction with $ri=4$.}
\end{subfigure}\hfil 
\begin{subfigure}{0.23\textwidth}
  \includegraphics[width=\linewidth]{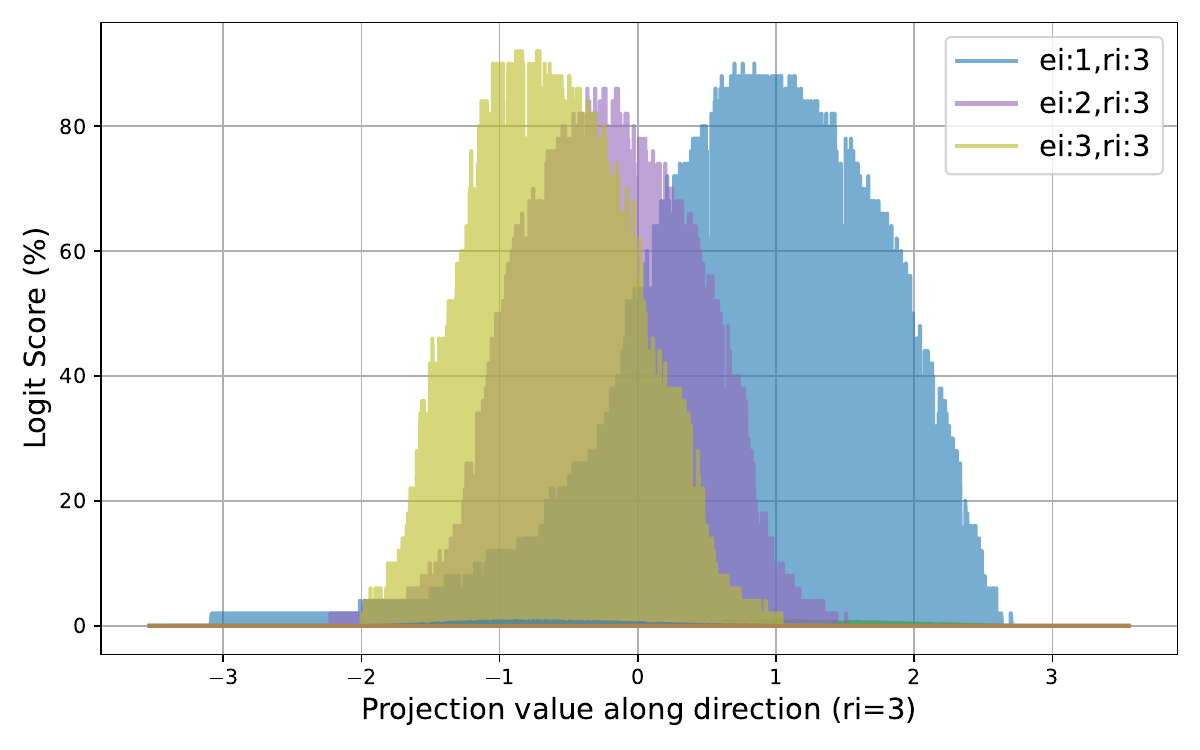}
  \caption{$ei$ direction with $ri=3$.}
\end{subfigure}\hfil 
\begin{subfigure}{0.23\textwidth}
  \includegraphics[width=\linewidth]{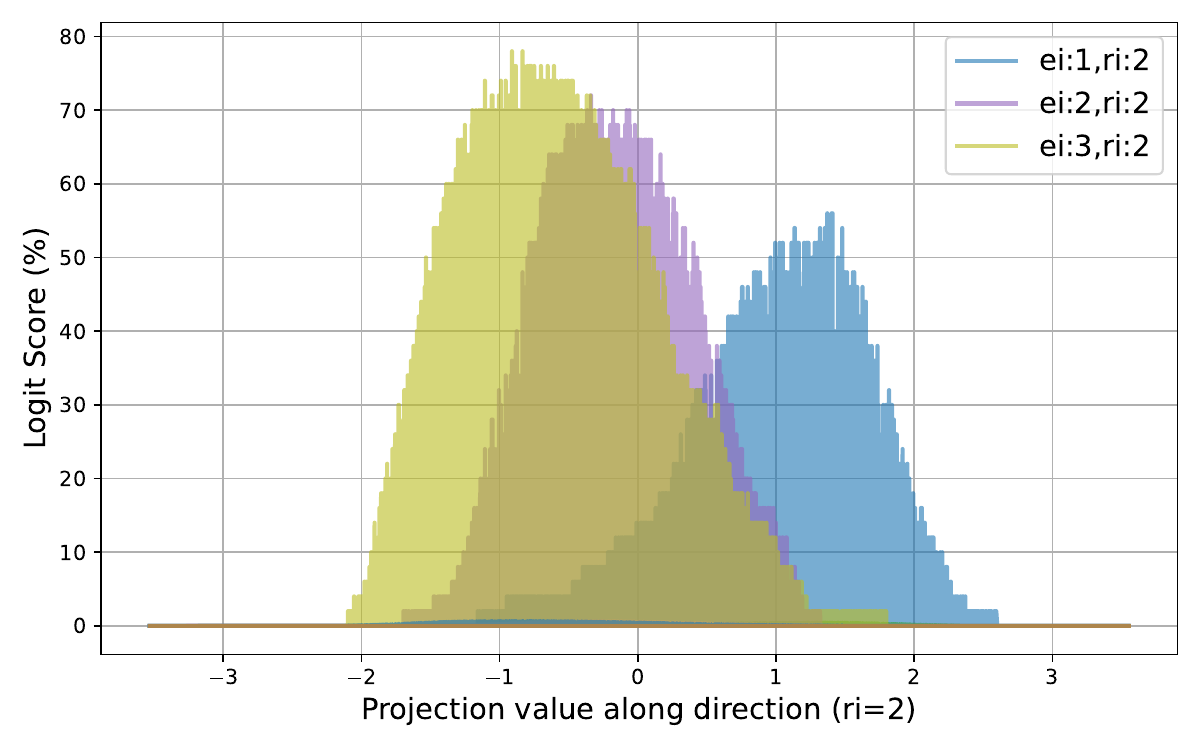}
  \caption{$ei$ direction with $ri=2$.}
\end{subfigure}\hfil 
\begin{subfigure}{0.23\textwidth}
  \includegraphics[width=\linewidth]{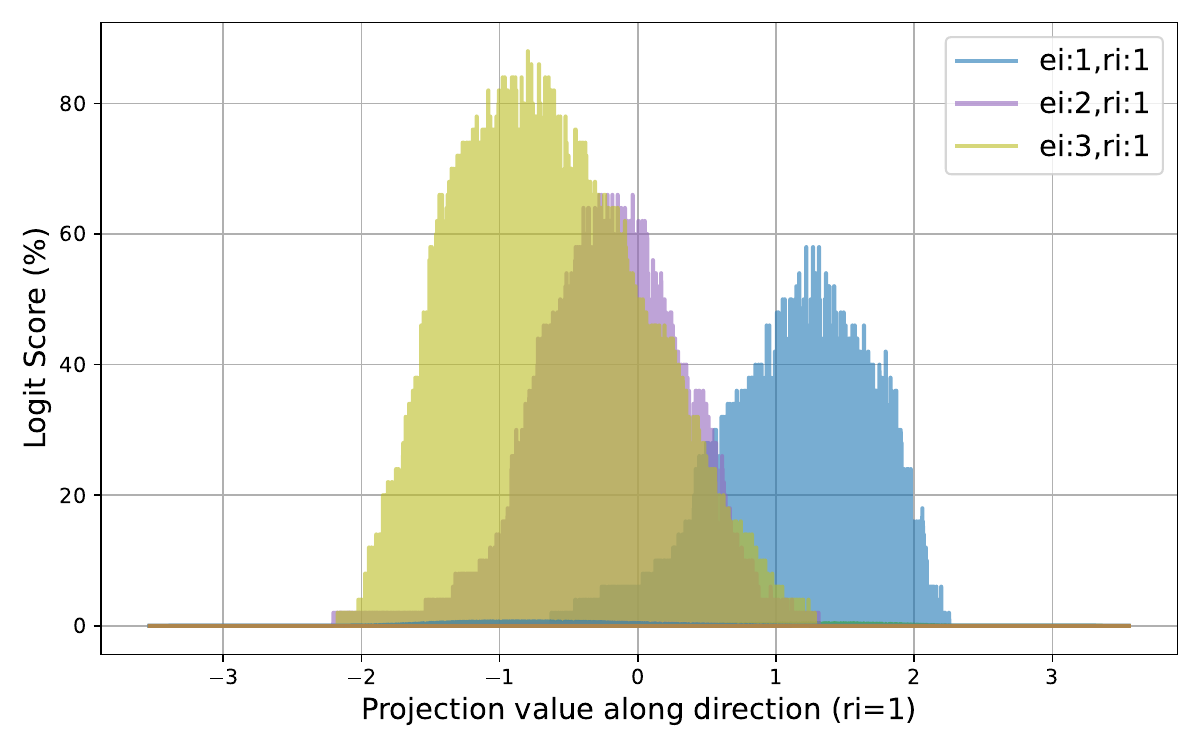}
  \caption{$ei$ direction with $ri=1$.}
\end{subfigure}\hfil 
\begin{subfigure}{0.3\textwidth}
  \includegraphics[width=\linewidth]{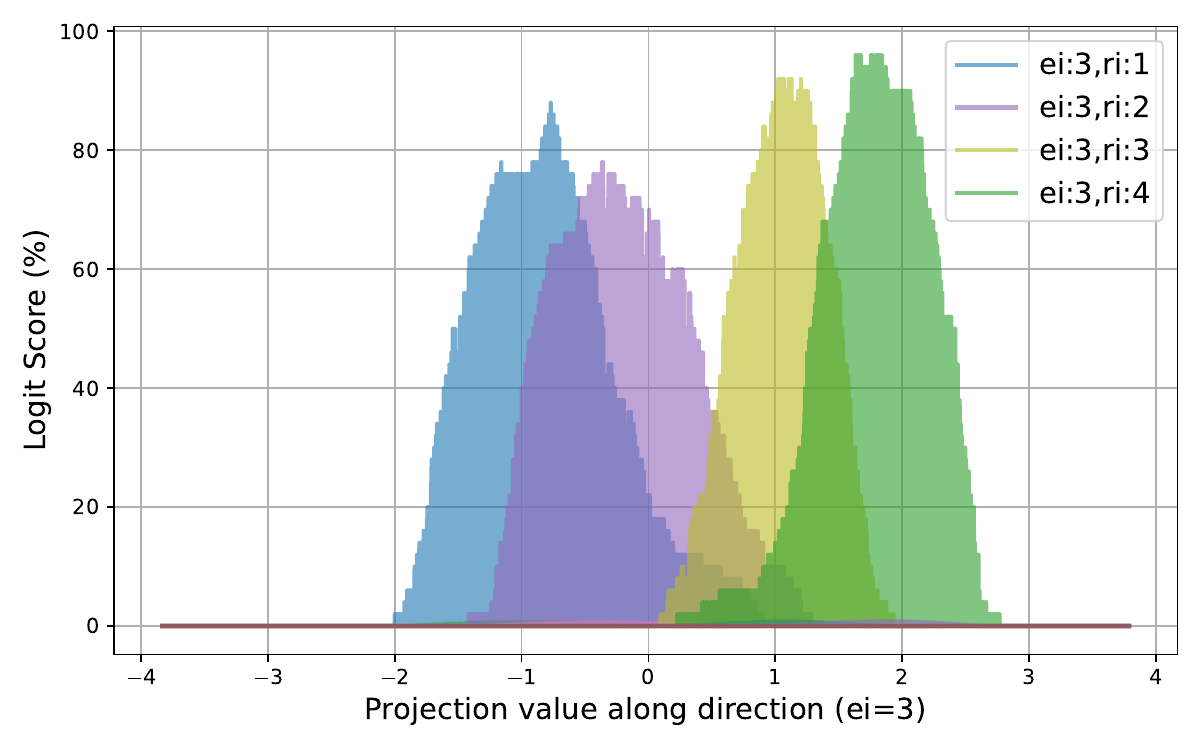}
  \caption{$ri$ direction with $ei=3$.}
\end{subfigure}\hfil 
\begin{subfigure}{0.3\textwidth}
  \includegraphics[width=\linewidth]{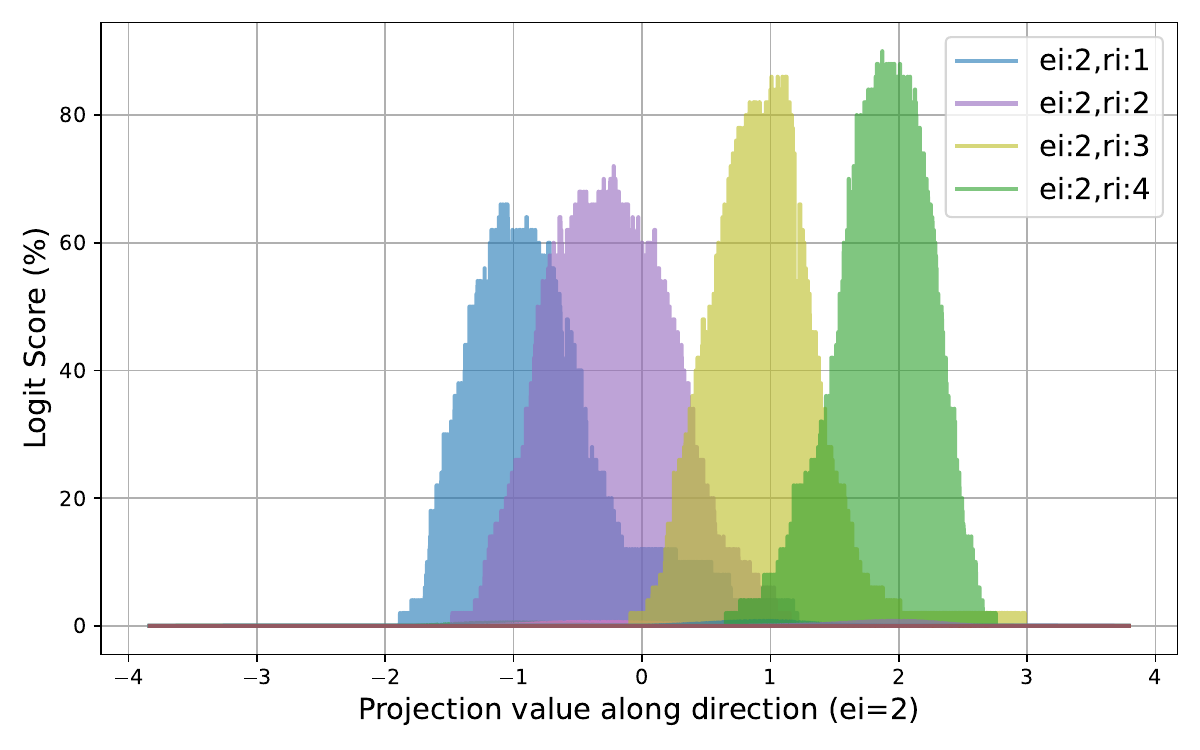}
  \caption{$ri$ direction with $ei=2$.}
\end{subfigure}\hfil 
\begin{subfigure}{0.3\textwidth}
  \includegraphics[width=\linewidth]{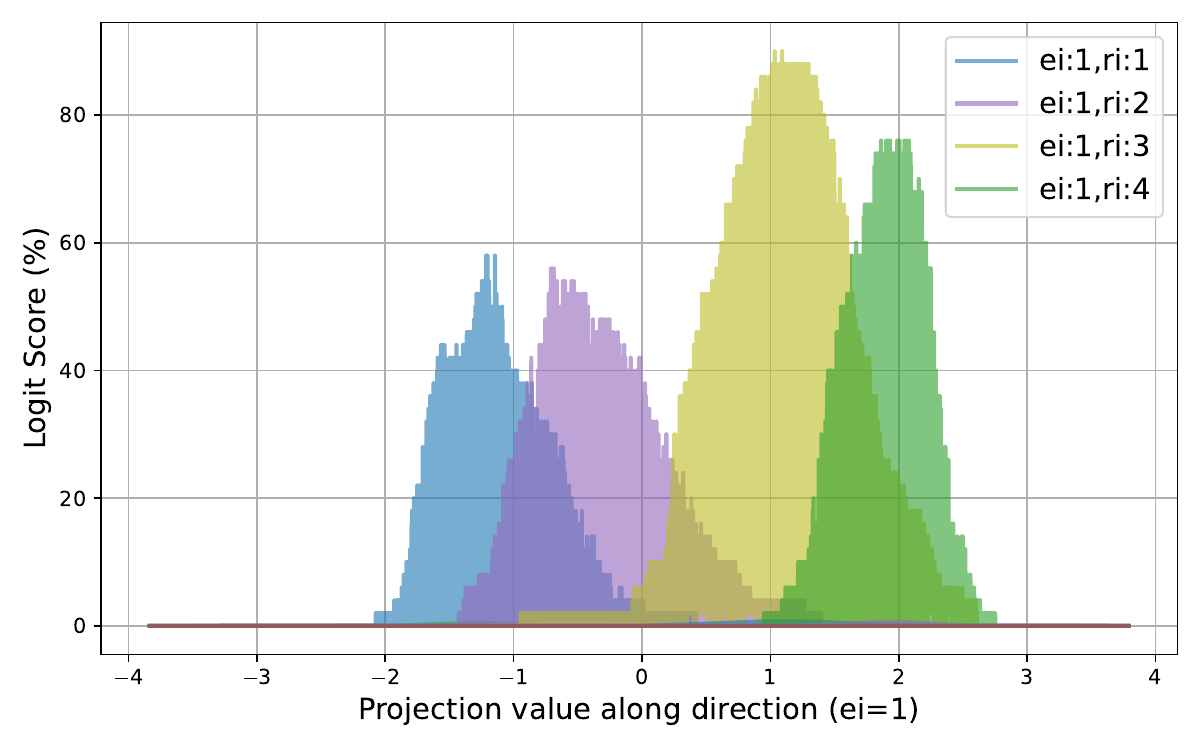}
  \caption{$ri$ direction with $ei=1$.}
\end{subfigure}\hfil 
\caption{Logit score curves on $C_{city}$ for Llama3-8B-Instruct.}
\label{fig:grid_logit_llama_city}
\end{figure*}
\begin{figure*}[!htbp]
    \centering 
\begin{subfigure}{0.23\textwidth}
  \includegraphics[width=\linewidth]{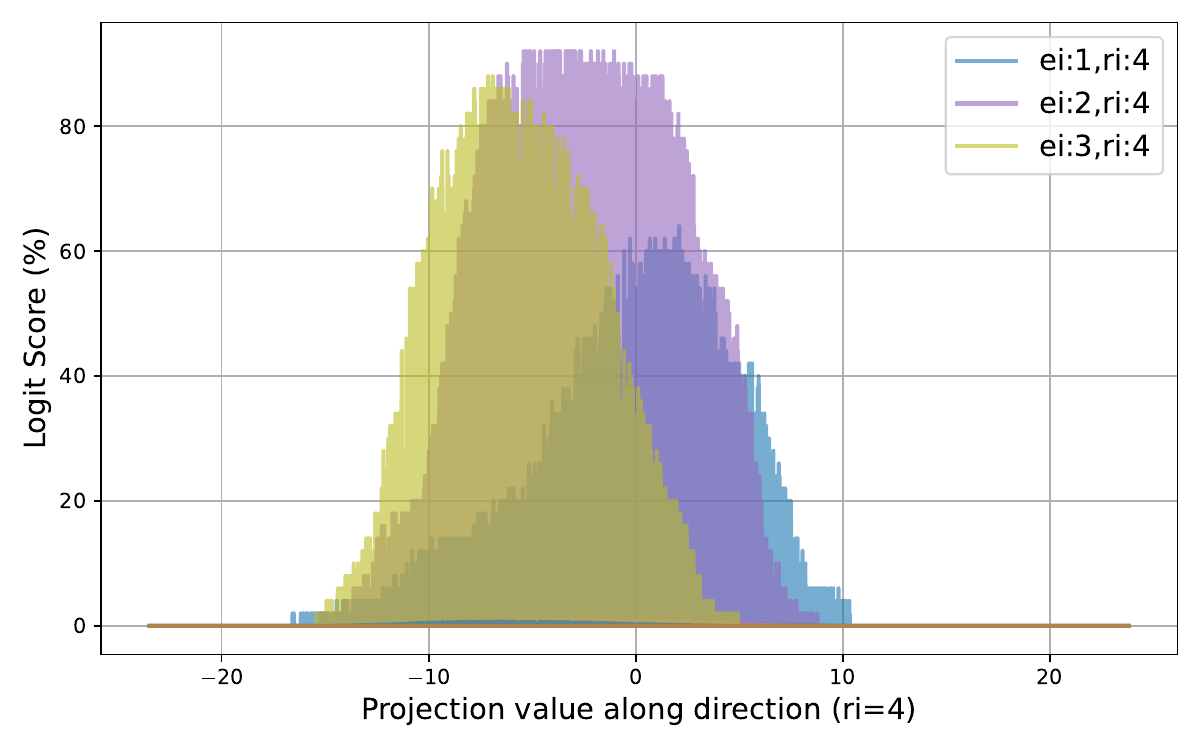}
  \caption{$ei$ direction with $ri=4$.}
\end{subfigure}\hfil 
\begin{subfigure}{0.23\textwidth}
  \includegraphics[width=\linewidth]{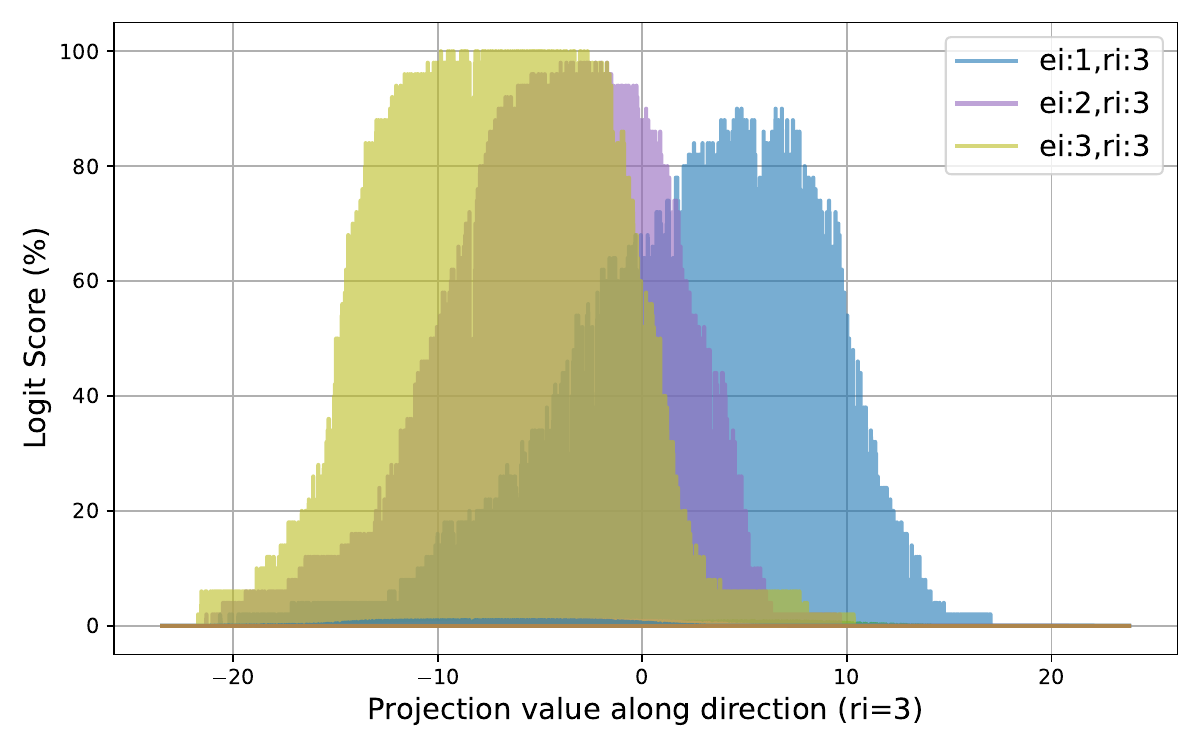}
  \caption{$ei$ direction with $ri=3$.}
\end{subfigure}\hfil 
\begin{subfigure}{0.23\textwidth}
  \includegraphics[width=\linewidth]{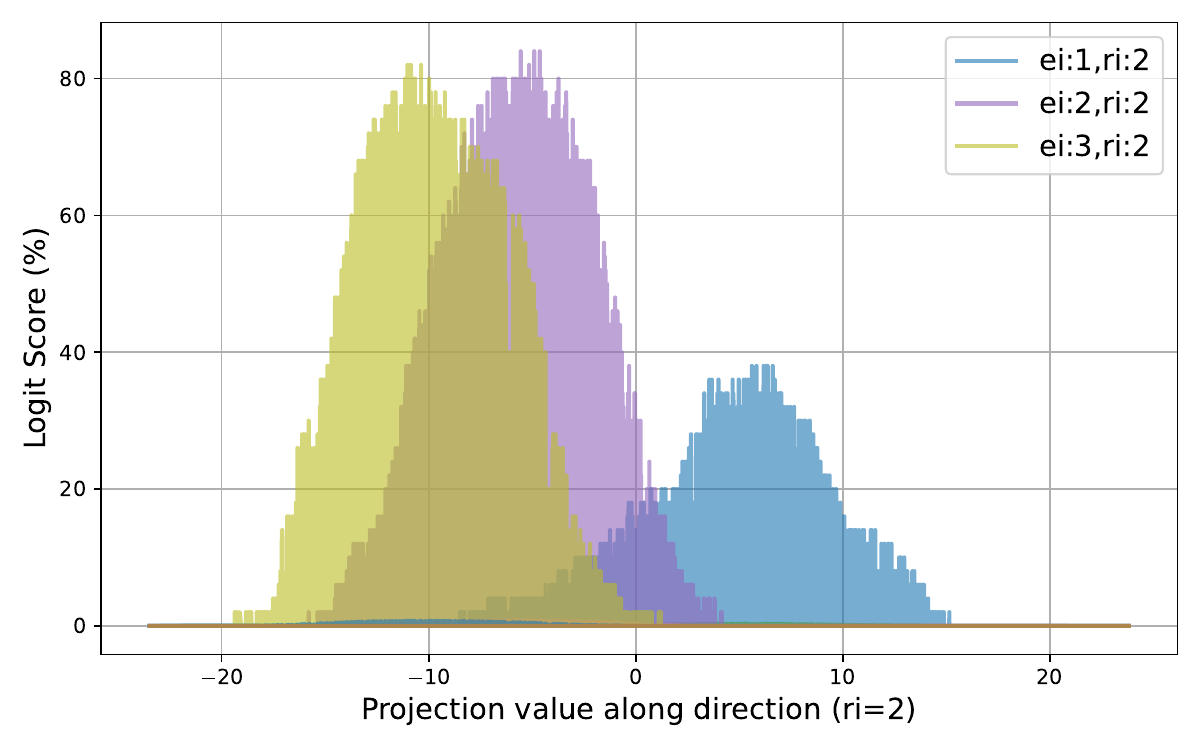}
  \caption{$ei$ direction with $ri=2$.}
\end{subfigure}\hfil 
\begin{subfigure}{0.23\textwidth}
  \includegraphics[width=\linewidth]{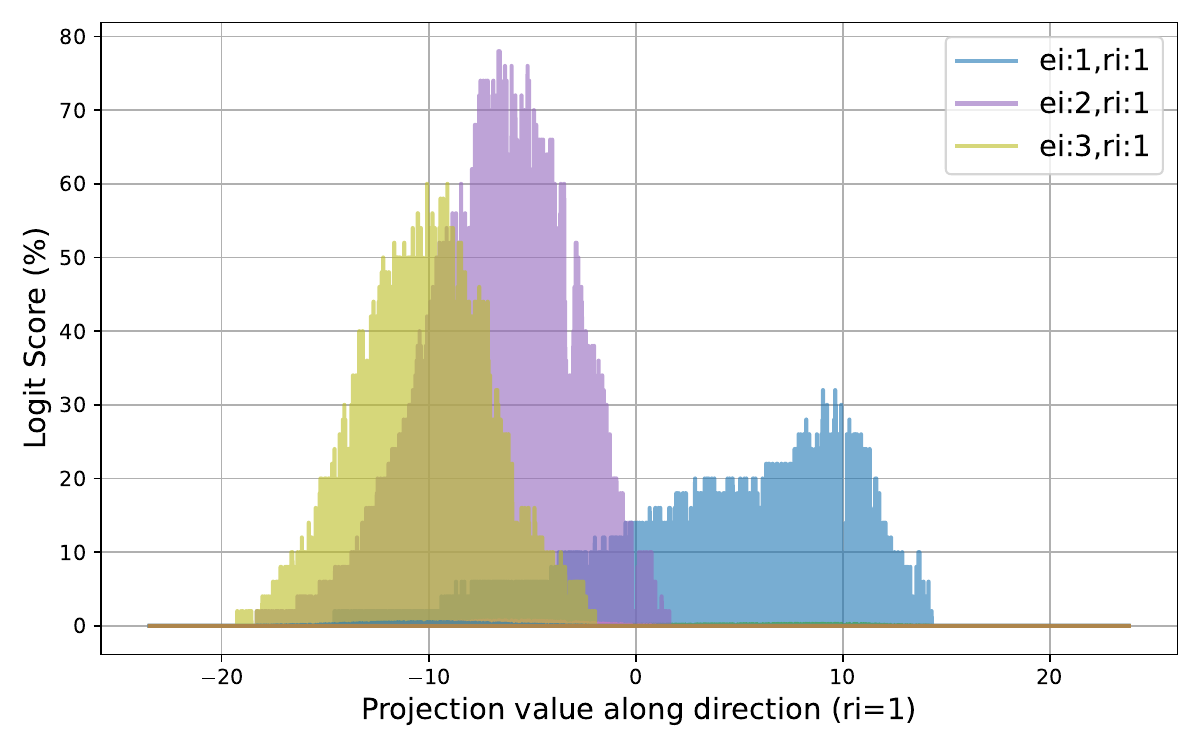}
  \caption{$ei$ direction with $ri=1$.}
\end{subfigure}\hfil 
\begin{subfigure}{0.3\textwidth}
  \includegraphics[width=\linewidth]{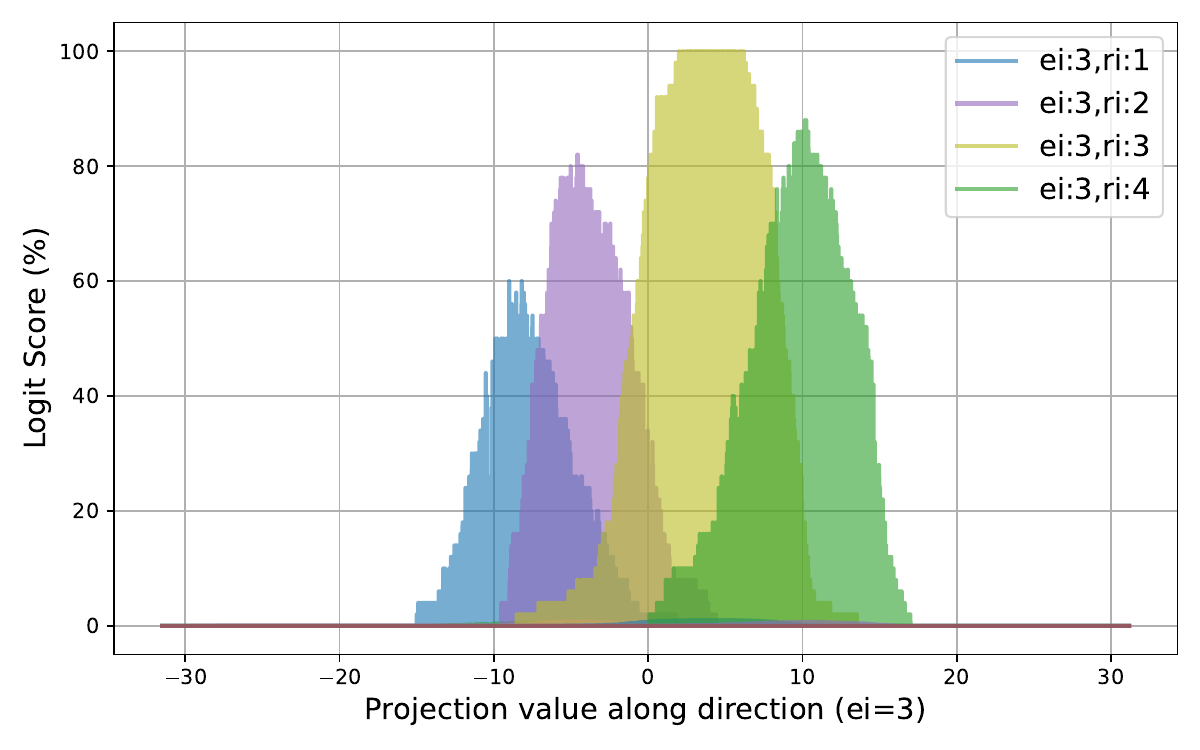}
  \caption{$ri$ direction with $ei=3$.}
\end{subfigure}\hfil 
\begin{subfigure}{0.3\textwidth}
  \includegraphics[width=\linewidth]{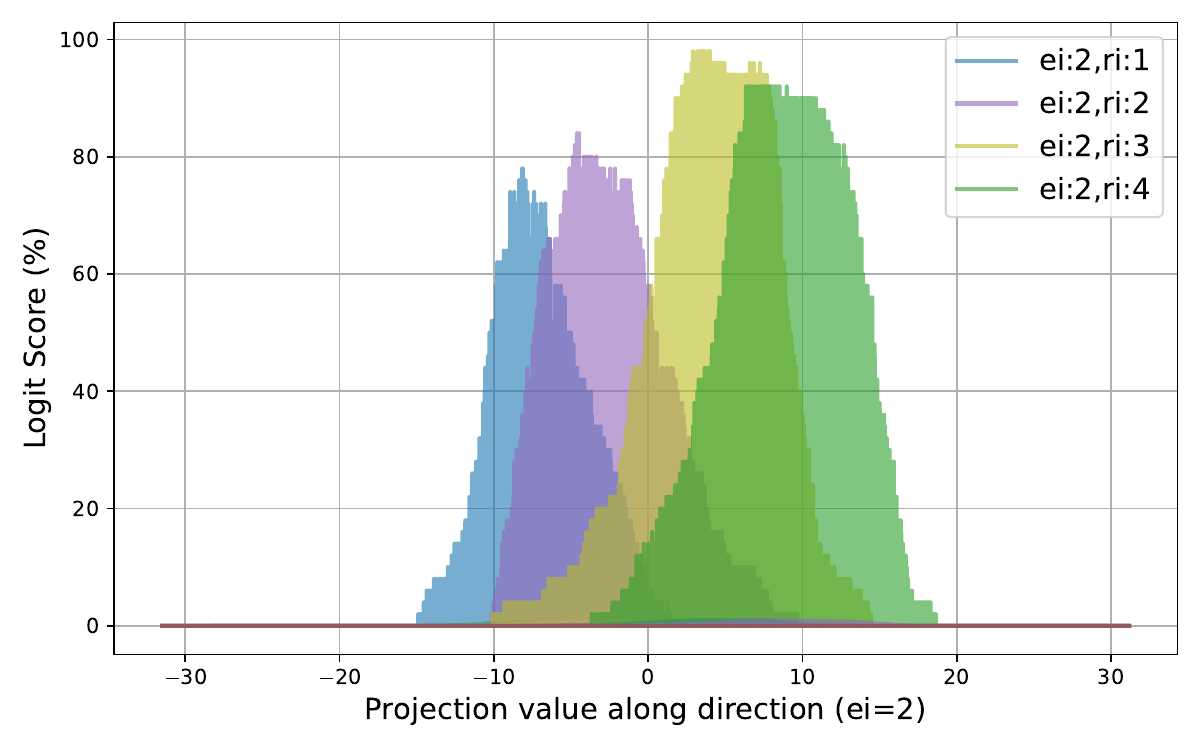}
  \caption{$ri$ direction with $ei=2$.}
\end{subfigure}\hfil 
\begin{subfigure}{0.3\textwidth}
  \includegraphics[width=\linewidth]{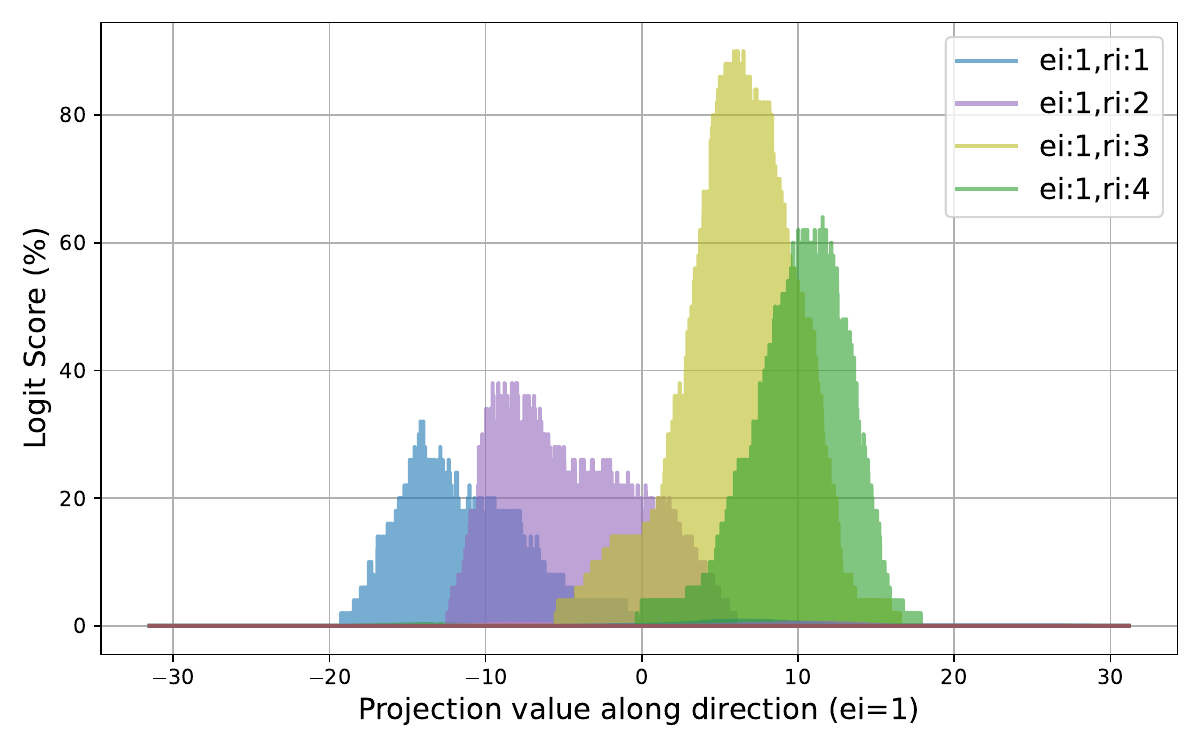}
  \caption{$ri$ direction with $ei=1$.}
\end{subfigure}\hfil 
\caption{Logit score curves on $C_{city}$ for Qwen3-8B.}
\label{fig:grid_logit_qwen_city}
\end{figure*}
\begin{figure*}[!htbp]
    \centering 
\begin{subfigure}{0.35\textwidth}
  \includegraphics[width=\linewidth]{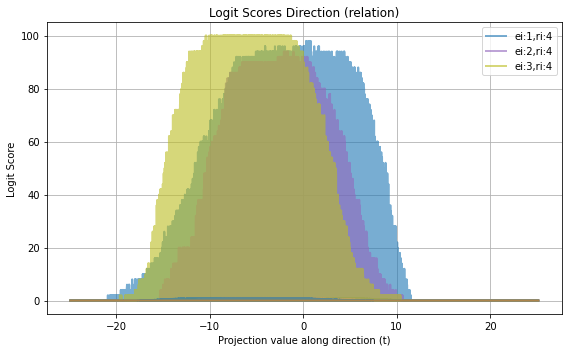}
  \caption{Logit score curves along $ei$ direction, where ($ri=4$).}
\end{subfigure}\hfil 
\begin{subfigure}{0.35\textwidth}
  \includegraphics[width=\linewidth]{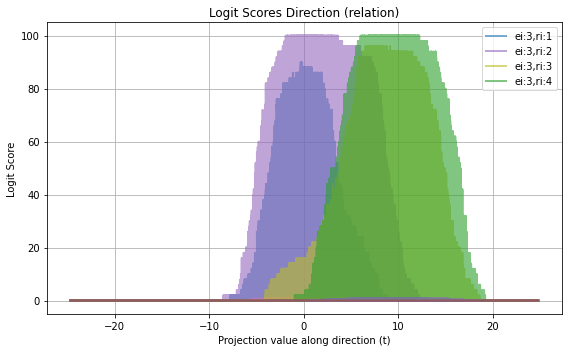}
  \caption{Logit score curves along $ri$ direction, where ($ei=3$).}
\end{subfigure}\hfil 
\begin{subfigure}{0.35\textwidth}
  \includegraphics[width=\linewidth]{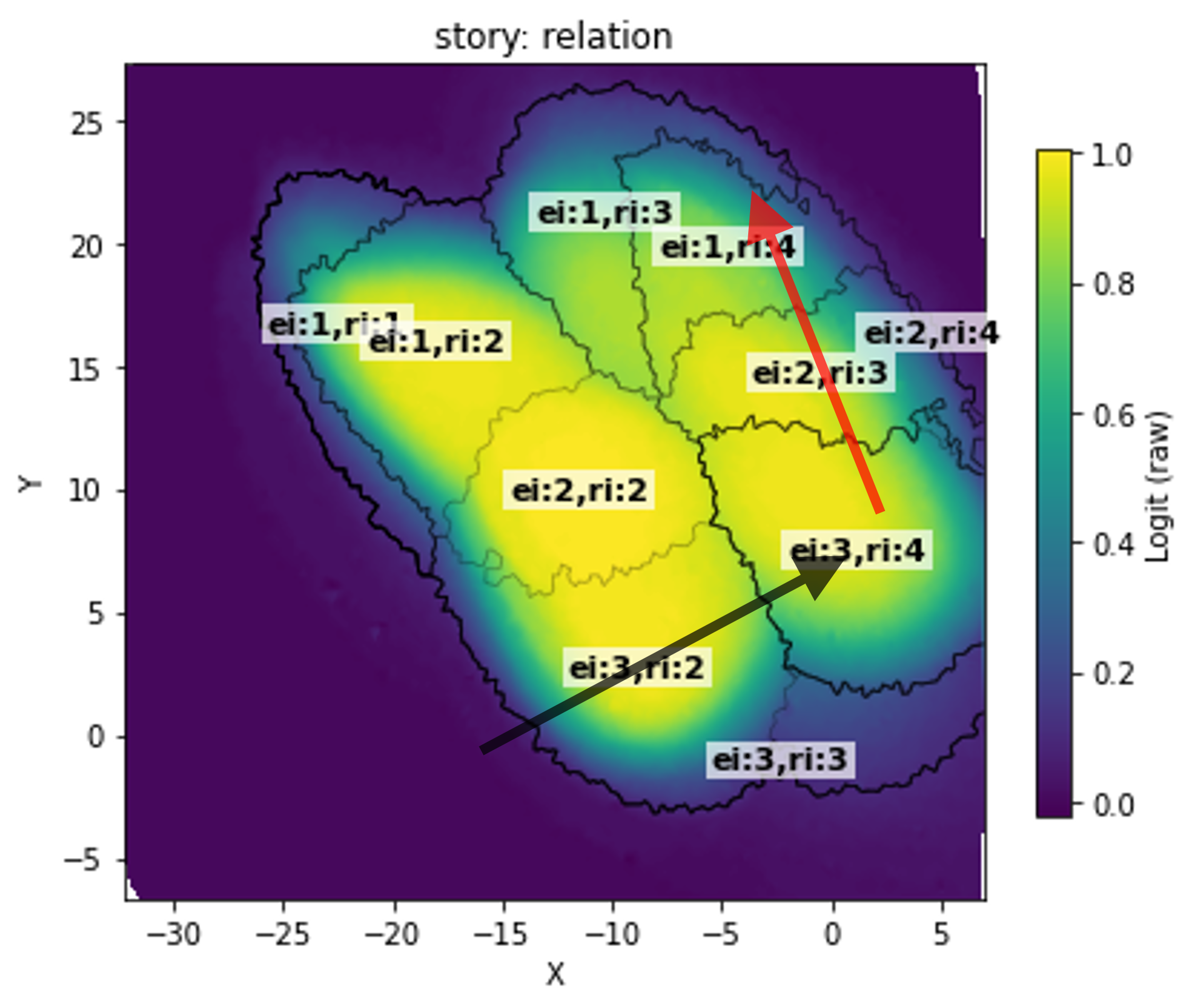}
  \caption{$ei$ (red arrow) and $ri$ (black arrow) directions.}
\end{subfigure}\hfil
\caption{Logit score curves on $C_{relation}$ for Qwen3-8B.}
\label{fig:grid_logit_qwen_relation}
\end{figure*}

We analyze how the logit score varies as we move along the learned entity-index ($ei$) and relation-index ($ri$) directions illustrated in Figure~\ref{fig:grid_direction_llama_city} and \ref{fig:grid_direction_qwen_city}, and the resulting logit curves are shown in Figure~\ref{fig:grid_llama} and \ref{fig:grid_logit_qwen_city}. The results show a characteristic pattern: as we move along one direction, the logit of a particular attribute rises to a peak near the center of its region and then falls off, as the logit of a neighboring attribute begins rising toward its own peak further along the direction. This behavior reinforces our earlier observation that attributes achieve their highest logit near the center of their respective Voronoi-like regions in the CBR subspace.

As discussed previously, some regions appear missing in the 2D visualization (e.g., $ei:3,ri:1$), as shown in Figure~\ref{fig:grid_logit_qwen_relation}. The logit scanning analysis clarifies this phenomenon: in the reduced 2D projection, the logit curves of multiple attributes overlap substantially, indicating that certain regions cannot be fully separated in only two dimensions. This provides further evidence that for a given context (e.g., $C_{relation})$, the 2D visualization could under-sample the true geometry, and that some Voronoi regions become indistinguishable when compressed into a low-dimensional space. 

\clearpage

\subsection{Perturbing CBR Subspace on Qwen3-8B}
\label{sec:perturb_irs_qwen}

\begin{figure}[htb]
    \centering 
\begin{subfigure}{0.4\textwidth}
  \includegraphics[width=\linewidth]{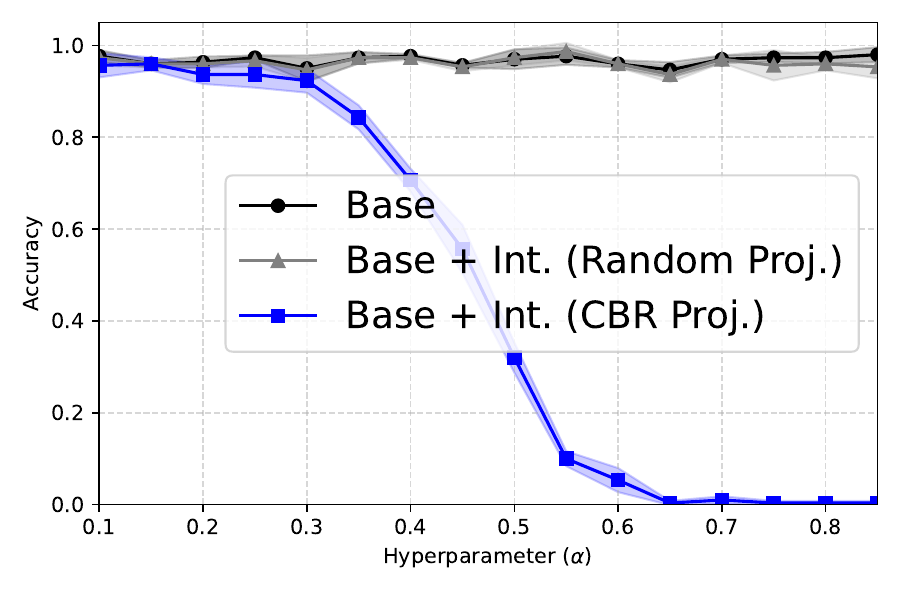}
  \caption{$C_{city}$}
\end{subfigure}\hfil 
\begin{subfigure}{0.4\textwidth}
  \includegraphics[width=\linewidth]{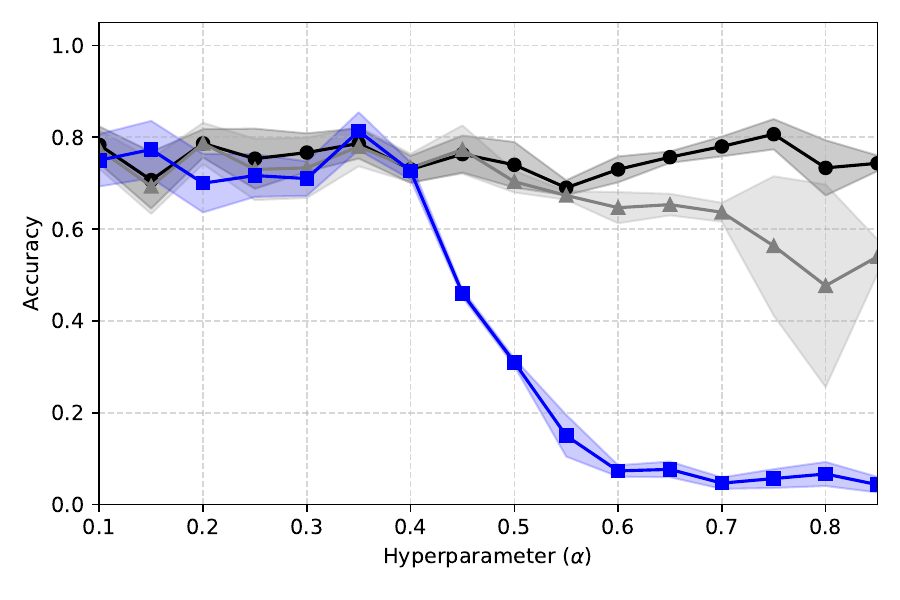}
  \caption{$C_{country}$}
\end{subfigure}\hfil 
\medskip
\begin{subfigure}{0.3\textwidth}
  \includegraphics[width=\linewidth]{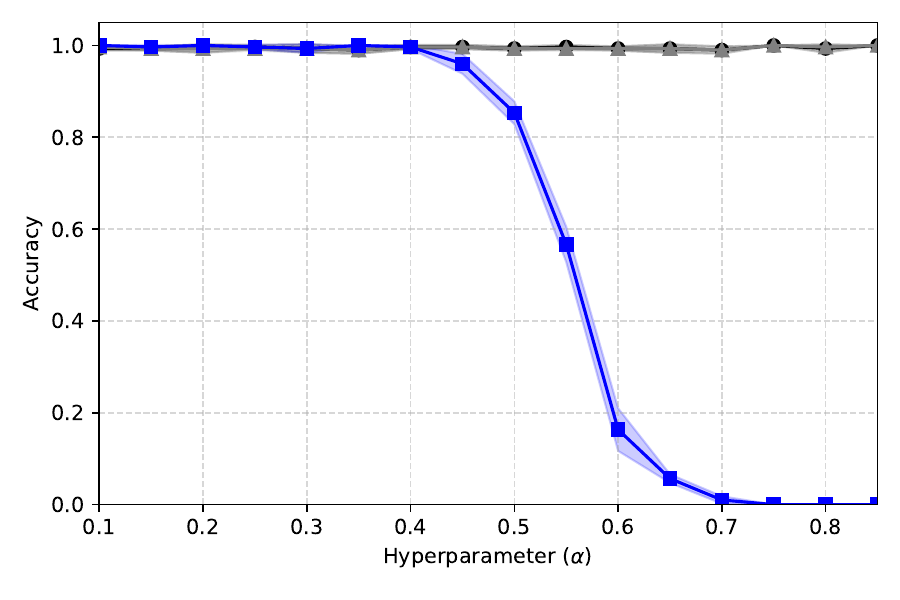}
  \caption{$C_{relation}$}
\end{subfigure}\hfil
\begin{subfigure}{0.3\textwidth}
  \includegraphics[width=\linewidth]{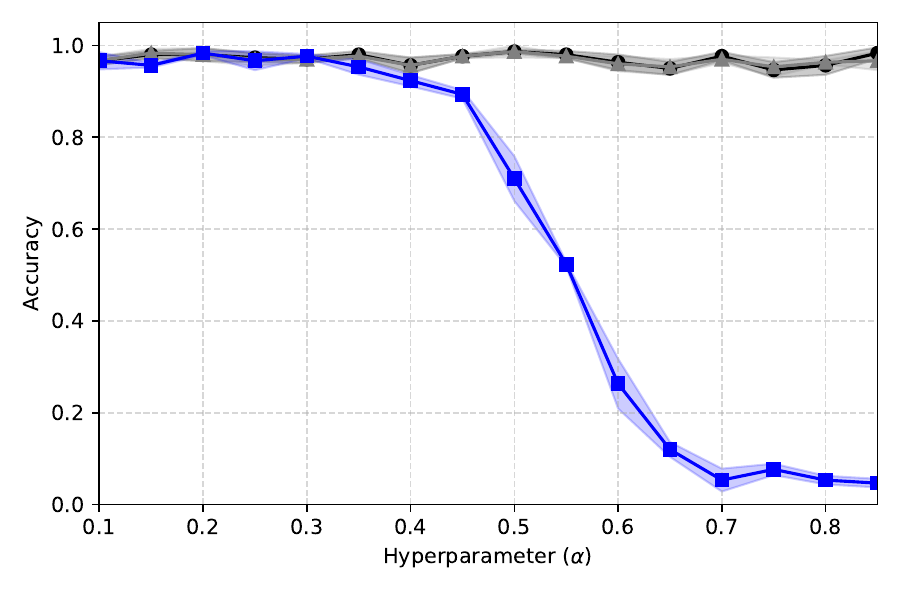}
  \caption{$C_{job}$}
\end{subfigure}\hfil 
\begin{subfigure}{0.3\textwidth}
  \includegraphics[width=\linewidth]{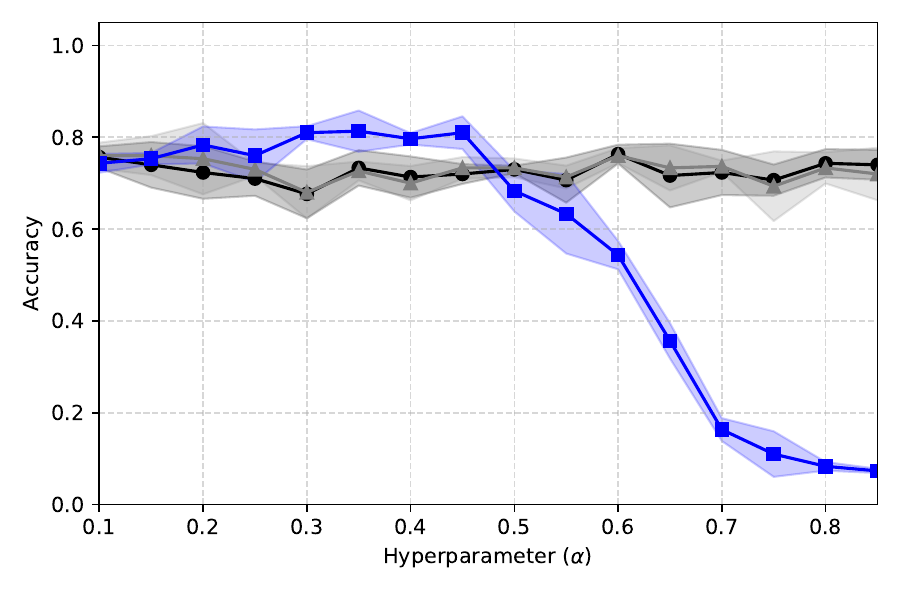}
  \caption{$C_{object}$}
\end{subfigure}\hfil 
\caption{Effect of perturbing activations along the CBR subspace versus a random subspace on Qwen3-8B. The X-axis shows the perturbation weight $\alpha$ in Equation~\ref{eq:irs_noise1} and \ref{eq:irs_noise2}, and the Y-axis shows the attribute prediction accuracy. Perturbations along the CBR subspace (i.e., blue line) lead to a significant drop in accuracy, while perturbations along a random subspace (i.e., grey line) have minimal effect. This indicates that LLMs rely on the CBR subspace to make relationally bound predictions.}
\label{fig:noise_qwen}
\end{figure}

Section~\secref{sec:perturb_CBR} mentions that as the perturbation weight increases, the accuracy of attribute predictions decreases significantly. In contrast, perturbations along a random subspace have little or no effect on accuracy. This pattern is also consistently observed in Qwen3-8B, as shown in Figure~\ref{fig:noise_qwen}, indicating that this behavior is prevalent across different LLM families rather than being model-specific.

\begin{figure*}[!htb]
    \centering 
\begin{subfigure}{0.45\textwidth}
  \includegraphics[width=\linewidth]{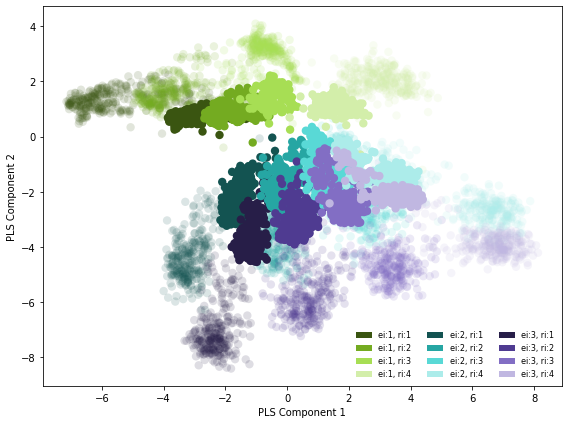}
  \caption{Perturbing via $W_{\text{CBR}}$}
\end{subfigure}\hfil 
\begin{subfigure}{0.45\textwidth}
  \includegraphics[width=\linewidth]{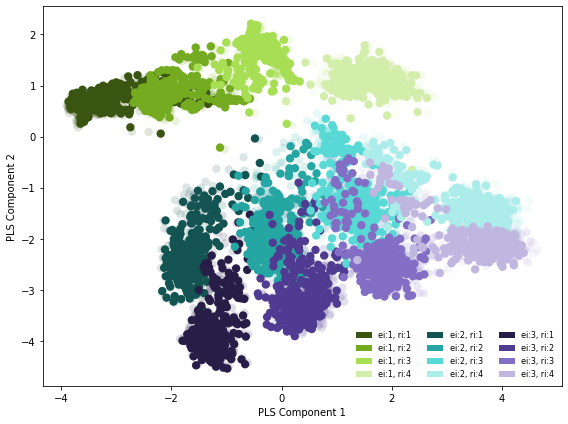}
  \caption{Perturbing via $W_{\text{rand}}$}
\end{subfigure}\hfil 
\caption{Visualization of the CBR subspace under perturbations along CBR directions using $W_{\text{CBR}}$ (Equation~\ref{eq:irs_noise2}),  and along random directions using $W_{\text{rand}}$ (Equation~\ref{eq:irs_noise1}) on Llama3-8B-Instruct, where dark colored points denote the distribution before perturbation, while light colored points represent the distribution after perturbation. This demonstrates that the perturbing method using Equation~\ref{eq:irs_noise2} could change the original distribution, while the method using Equation~\ref{eq:irs_noise1} does not.}
\label{fig:pertrub_irs_rand}
\end{figure*}

\clearpage

\subsection{Activation Steering on other Setting}
\label{sec:irs_steering_other}
\begin{figure}[!t]
\centering
\includegraphics[width=12.0cm]{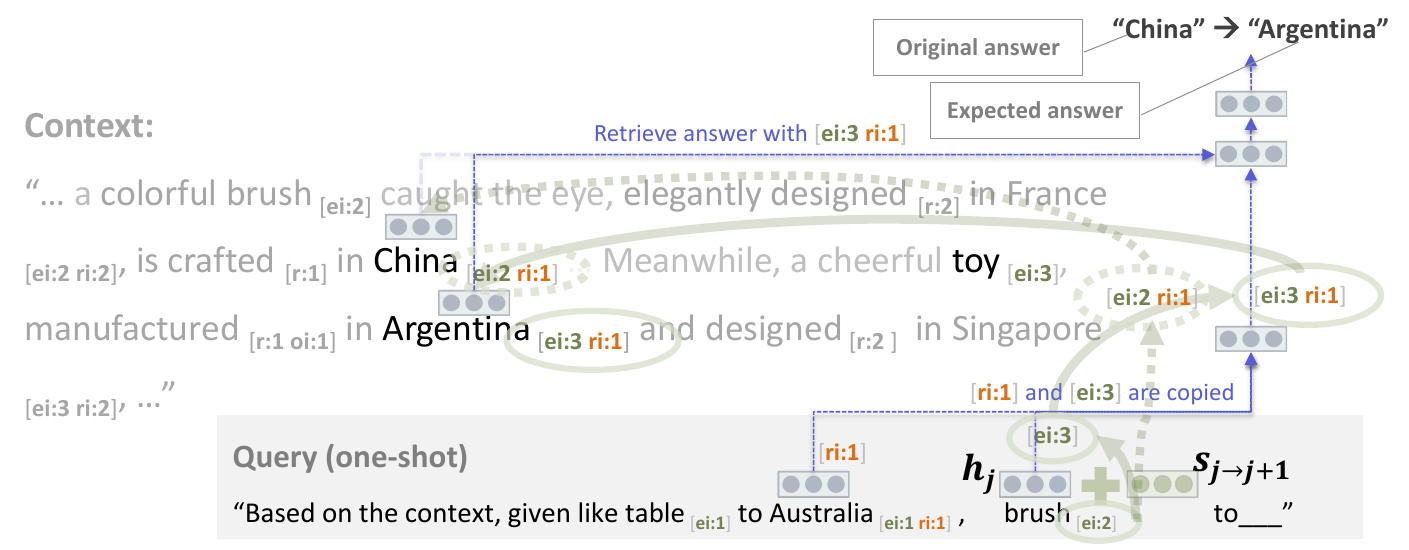}
\caption{Entity-index Steering.}
\label{fig:mechanism_ent}
\end{figure}

\begin{figure}[!t]
\centering
\includegraphics[width=12.0cm]{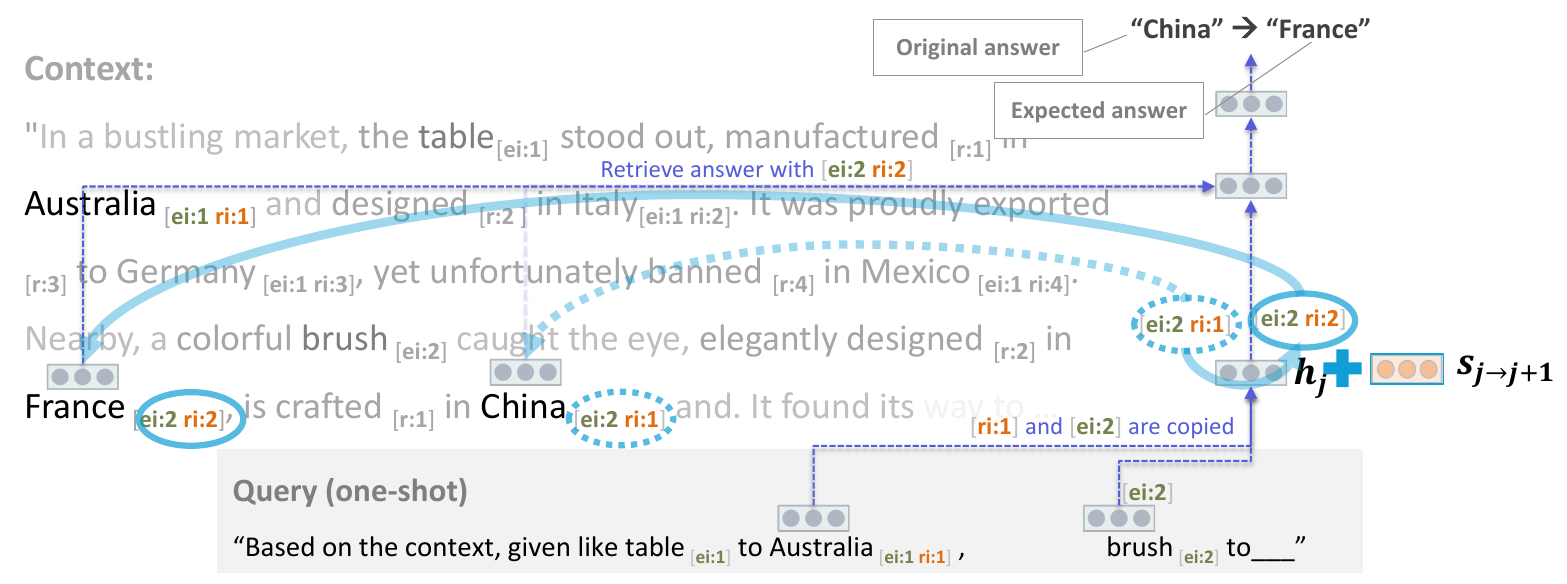}
\caption{Last-token Steering.}
\label{fig:mechanism_last}
\end{figure}

\begin{figure}[!t]
\centering
\includegraphics[width=8.5cm]{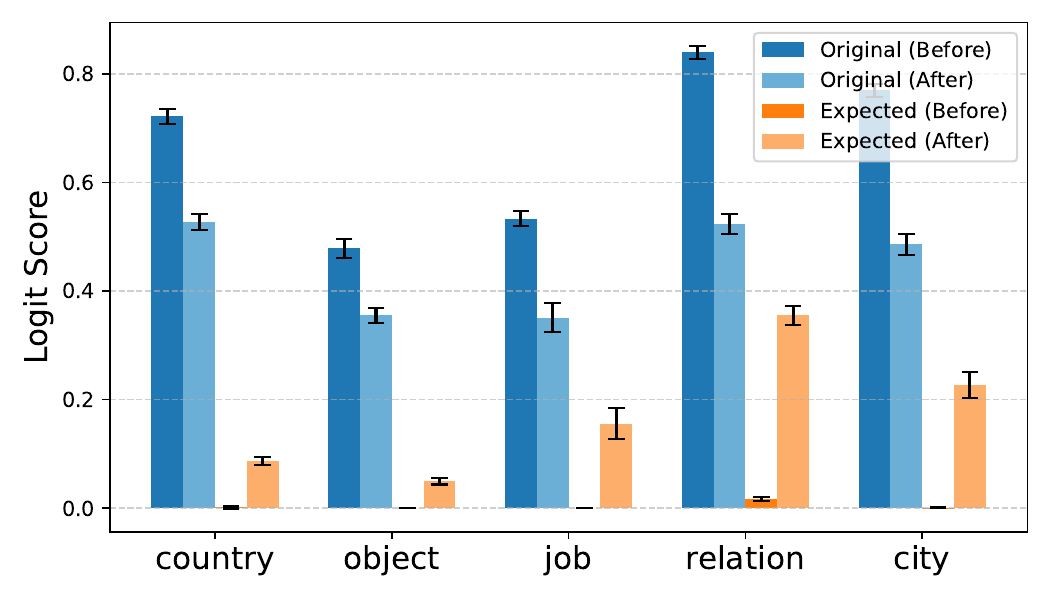}
\caption{Activation patching via Entity-index (i.e., $ei$) steering on the activation of the entity token (e.g., ``brush'') in query part across five contexts on Llama3-8B-Instruct. Each subplot corresponds to one context. We show the change in logit scores for the original answer and the expected answer before and after activation patching, which are denoted as ``Original (Before)'', ``Original (After)'', ``Expected (Before)'' and ``Expected (After)'' respectively.}
\label{fig:steer_llama_ent}
\end{figure}

\begin{figure*}[!htbp]
    \centering 
\begin{subfigure}{0.475\textwidth}
  \includegraphics[width=\linewidth]{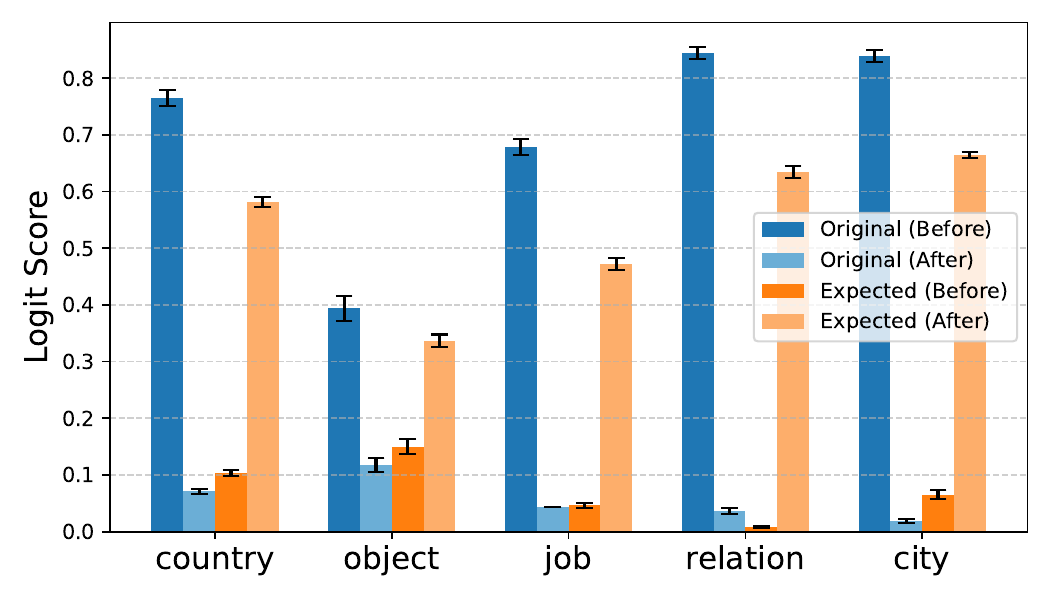}
  \caption{Relation-index (i.e., $ri$) steering.}
  \label{fig:steer_llama_last_att}
\end{subfigure}\hfil 
\begin{subfigure}{0.475\textwidth}
  \includegraphics[width=\linewidth]{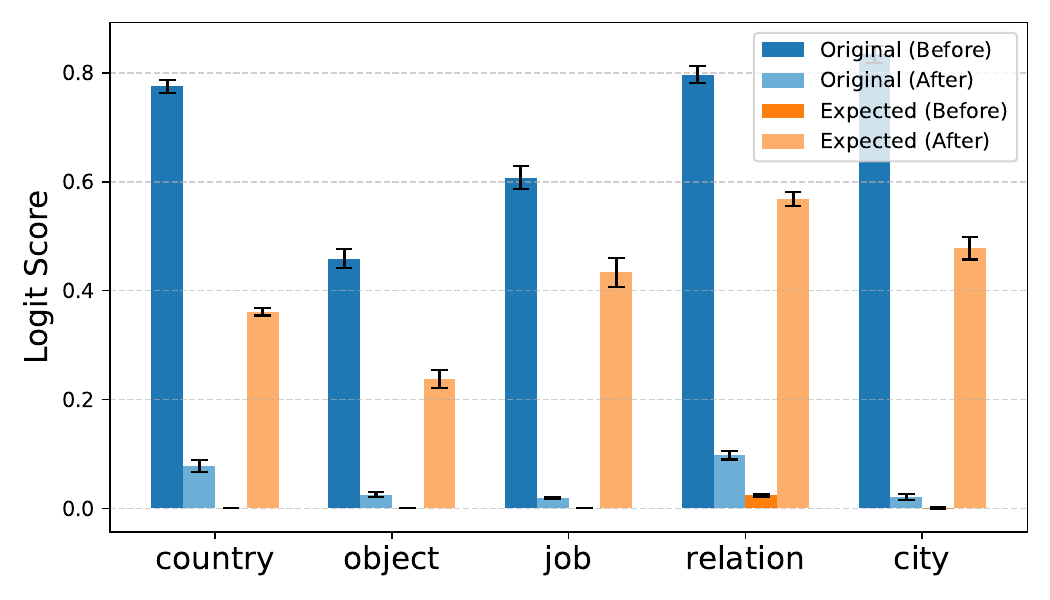}
  \caption{Entity-index (i.e., $ei$) steering.}
  \label{fig:steer_llama_last_ent}
\end{subfigure}\hfil 
\caption{Activation patching on the activation of the last token (i.e., $\mathbf{h}$) across five contexts on Llama3-8B-Instruct.}
\label{fig:steer_llama_last}
\end{figure*}

The results for (b) Entity-index steering illustrated in Figure~\ref{fig:mechanism_ent} and (c) Last-token steering illustrated in Figure~\ref{fig:mechanism_last} are shown in Figure~\ref{fig:steer_llama_ent}, \ref{fig:steer_llama_last_att} and \ref{fig:steer_llama_last_ent}, demonstrating that the logit change pattern is consistent across different steering configurations.

\begin{figure*}[t]
\centering
\includegraphics[width=12.0cm]{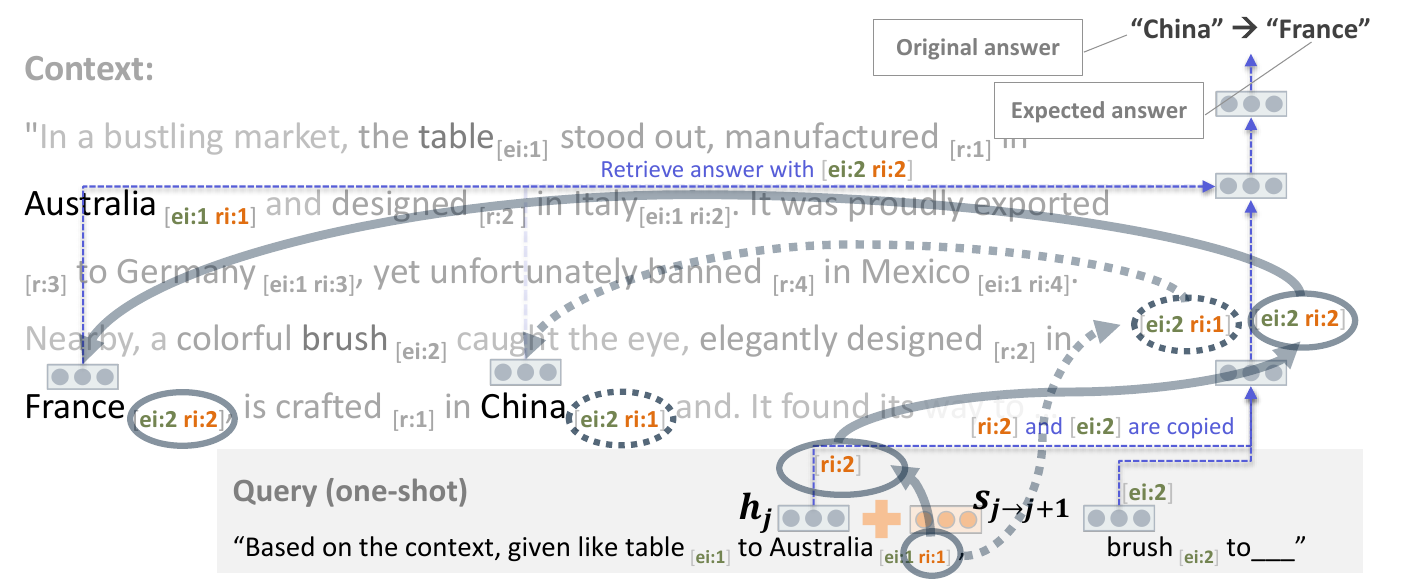}
\caption{Causal intervention on the CBR subspace reveals the CBR subspace based mechanism. Steering different components of the CBR subspace produces systematic changes in model behavior. For example, manipulating the relation index in the one-shot attribute activation (e.g., shifting from $ri:1$ to $ri:2$ in the activation of “Australia”) redirects the model toward predicting attributes associated with the intervened relation (e.g., changing the output from “China” to “France”).}
\label{fig:mechanism_att}
\end{figure*}

\subsection{Activation Steering on Qwen3-8B}
\label{sec:irs_steering_qwen}
\begin{figure*}[!htbp]
    \centering 
\begin{subfigure}{0.475\textwidth}
  \includegraphics[width=\linewidth]{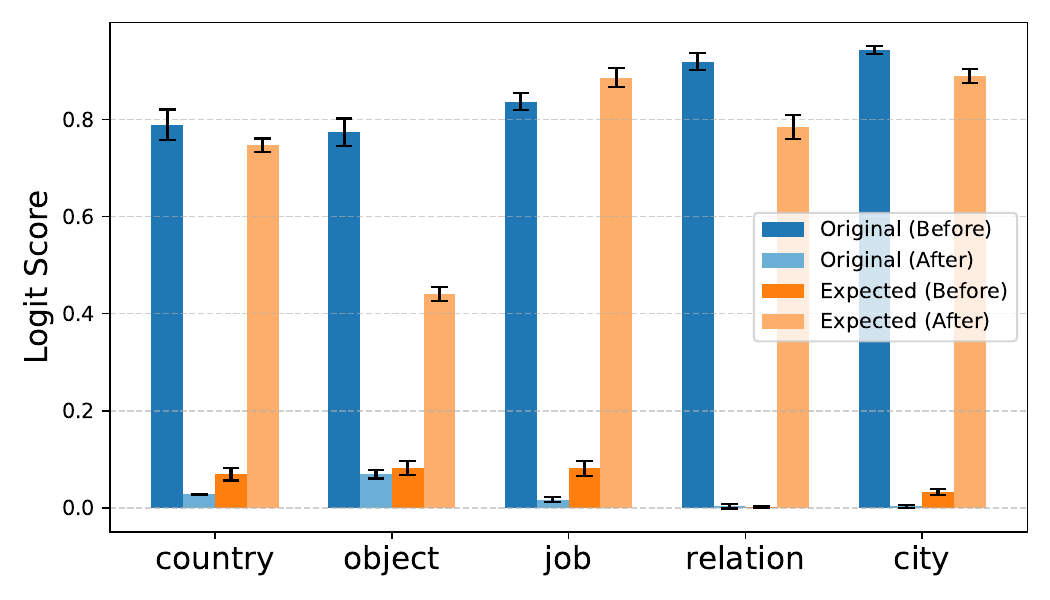}
  \caption{Relation-index (i.e., $ri$) steering on the attribute token.}
  \label{fig:steer_qwen_att}
\end{subfigure}\hfil 
\begin{subfigure}{0.475\textwidth}
  \includegraphics[width=\linewidth]{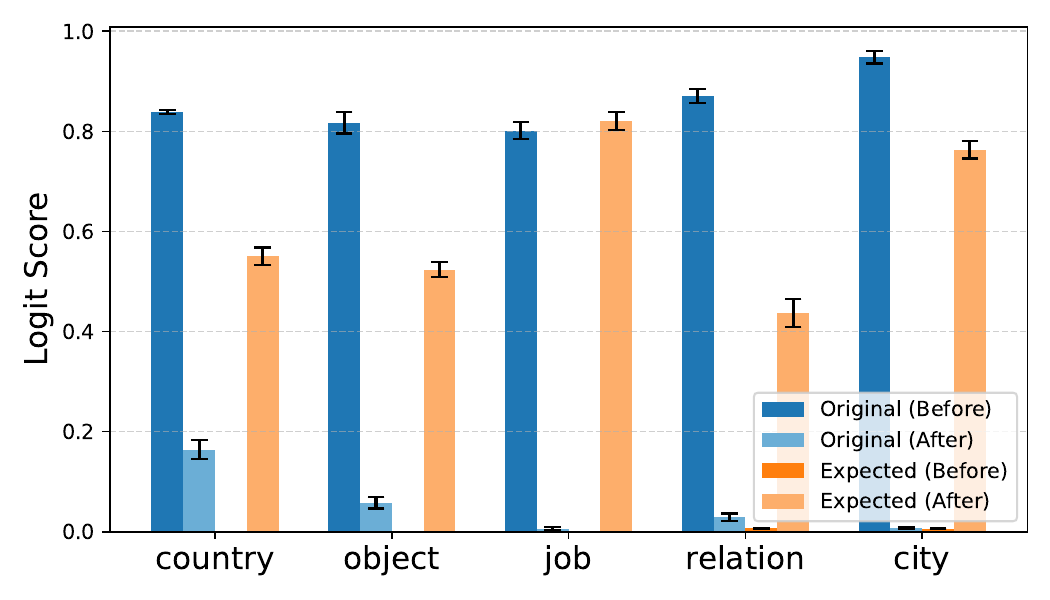}
  \caption{Entity-index (i.e., $ei$) steering on the entity token.}
  \label{fig:steer_qwen_ent}
\end{subfigure}\hfil 
\caption{Activation patching on corresponding token in query part (i.e., $\mathbf{h}$) across five contexts on Llama3-8B-Instruct.}
\label{fig:steer_qwen}
\end{figure*}

\begin{figure*}[!htbp]
    \centering 
\begin{subfigure}{0.475\textwidth}
  \includegraphics[width=\linewidth]{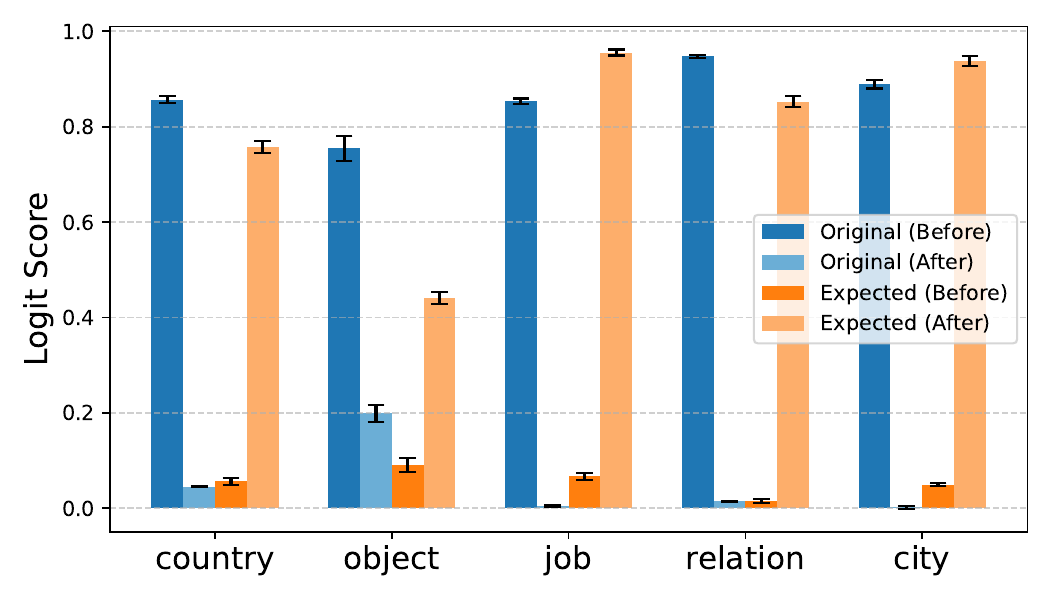}
  \caption{Relation-index (i.e., $ri$) steering.}
  \label{fig:steer_qwen_last_att}
\end{subfigure}\hfil 
\begin{subfigure}{0.475\textwidth}
  \includegraphics[width=\linewidth]{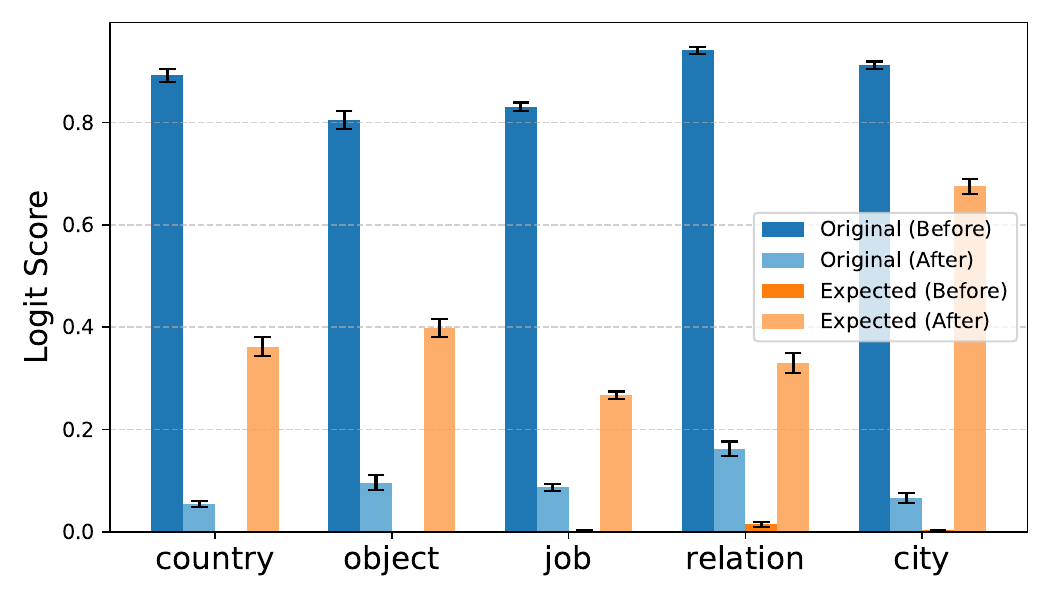}
  \caption{Entity-index (i.e., $ei$) steering.}
  \label{fig:steer_qwen_last_ent}
\end{subfigure}\hfil 
\caption{Activation patching on the activation of the last token (i.e., $\mathbf{h}$) across five contexts on Qwen3-8B.}
\label{fig:steer_qwen_last}
\end{figure*}

Moreover, the same behavior is consistently observed in Qwen3-8B, as show in Figure~\ref{fig:steer_qwen_att}, \ref{fig:steer_qwen_ent}, \ref{fig:steer_llama_last_att} and \ref{fig:steer_qwen_last_ent}, indicating that this mechanism is prevalent across different LLM families. Taken together, these results suggest that LLMs rely on an CBR subspace based mechanism to retrieve answers from context.


\subsection{Activation Steering on Shuffled and Ablated Datasets}
\label{sec:irs_steering_shuffle_ablate}

\begin{figure*}[!htbp]
    \centering 
\begin{subfigure}{0.475\textwidth}
  \includegraphics[width=\linewidth]{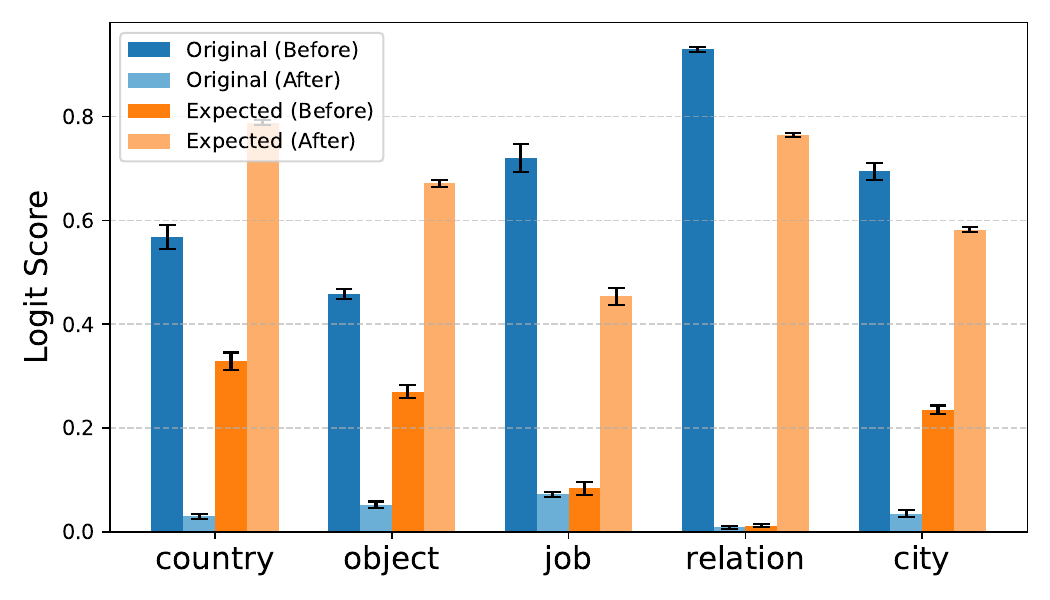}
  \caption{\textbf{ablated} dataset}
  \label{fig:steer_llama_ablate}
\end{subfigure}\hfil 
\begin{subfigure}{0.475\textwidth}
  \includegraphics[width=\linewidth]{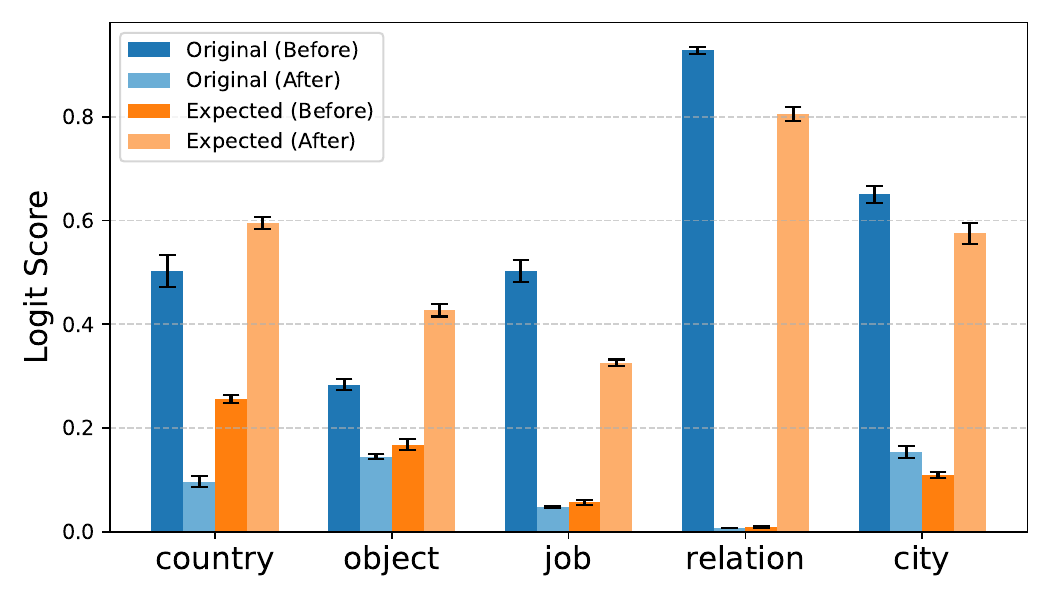}
  \caption{\textbf{shuffled} dataset}
  \label{fig:steer_llama_shuffle}
\end{subfigure}\hfil 
\caption{Activation patching via Relation-index (i.e., $ri$) steering on the last token (i.e., $\mathbf{h}$) on Llama3-8B-Instruct.}
\label{fig:steer_ablate_shuffle_llama}
\end{figure*}

\begin{figure*}[!htbp]
    \centering 
\begin{subfigure}{0.475\textwidth}
  \includegraphics[width=\linewidth]{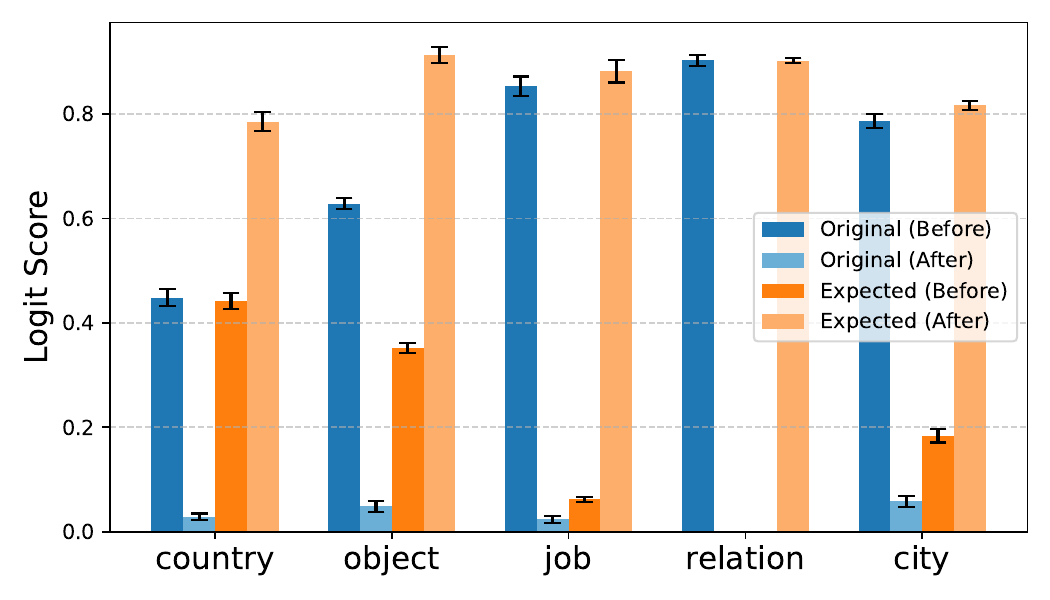}
  \caption{\textbf{ablated} dataset}
  \label{fig:steer_qwen_ablate}
\end{subfigure}\hfil 
\begin{subfigure}{0.475\textwidth}
  \includegraphics[width=\linewidth]{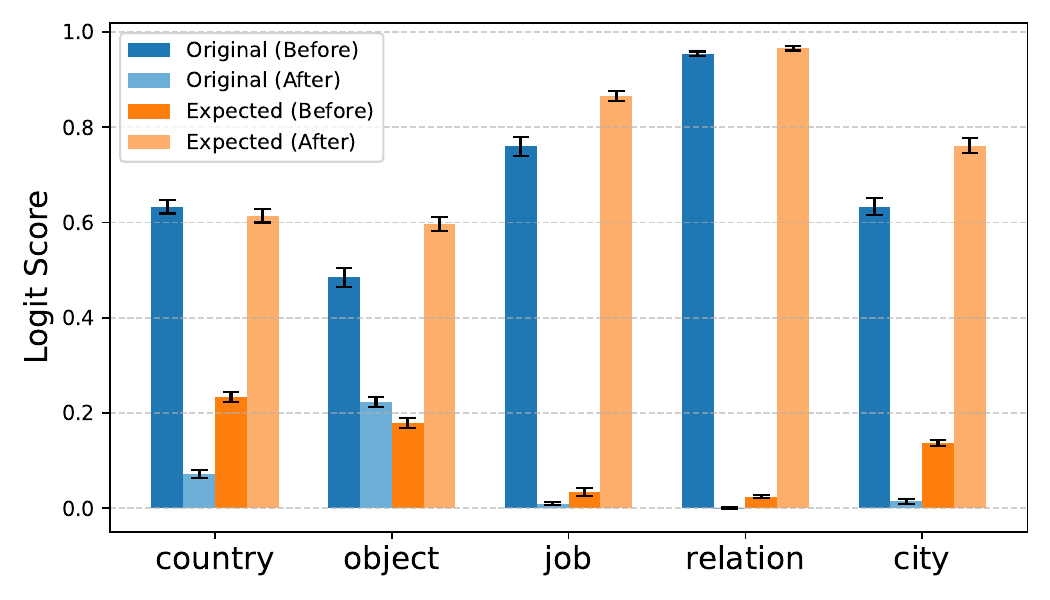}
  \caption{\textbf{shuffled} dataset}
  \label{fig:steer_qwen_shuffle}
\end{subfigure}\hfil 
\caption{Activation patching via Relation-index (i.e., $ri$) steering on the last token (i.e., $\mathbf{h}$) on Qwen3-8B.}
\label{fig:steer_ablate_shuffle_qwen}
\end{figure*}

Section~\secref{sec:irs_steering} shows that steering along the CBR subspace direction consistently suppresses the logit of the original answer and increases the logit of the expected answer, in precise alignment with the intended index manipulation. To further test the robustness of this mechanism, we conduct the same AP experiments under both shuffled and ablated settings mentioned in Section~\secref{sec:consistency_of_irs_subspace}, where surface-level relations are either reordered or partially removed without altering the underlying indices. As shown in Figure~\ref{fig:steer_llama_ablate}, \ref{fig:steer_llama_shuffle}, \ref{fig:steer_qwen_ablate} and \ref{fig:steer_qwen_shuffle}, across these modified settings on both Llama3-8B-Instruct and Qwen3-8B, we observe the same systematic pattern of logit suppression and promotion.

These results indicate that the CBR subspace based mechanism is stable under surface-level perturbations and does not depend on superficial modification. Instead, the behavior of the model is governed by the underlying relational structure. This finding strengthens our earlier conclusion that it is the CBR index information, rather than surface form modifications, that drives the organization of the CBR subspace and the relational retrieval behavior of LLMs.


\clearpage

\subsection{Visualization of CBR Subspace for Template Input}
\label{sec:irs_visualization_tabtemp}
\begin{figure*}[!htbp]
    \centering 
\begin{subfigure}{0.3\textwidth}
  \includegraphics[width=\linewidth]{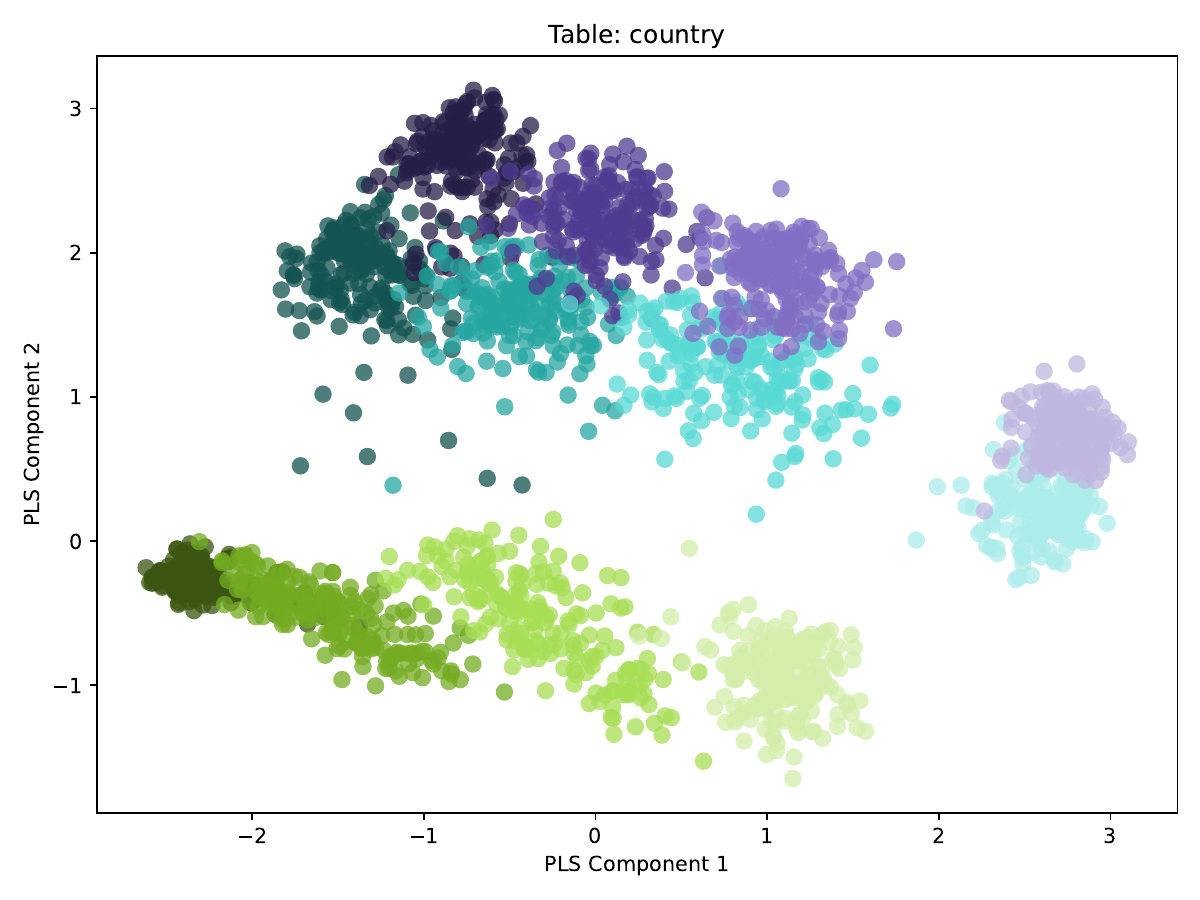}
  \caption{$C_{country}$}
\end{subfigure}\hfil 
\begin{subfigure}{0.3\textwidth}
  \includegraphics[width=\linewidth]{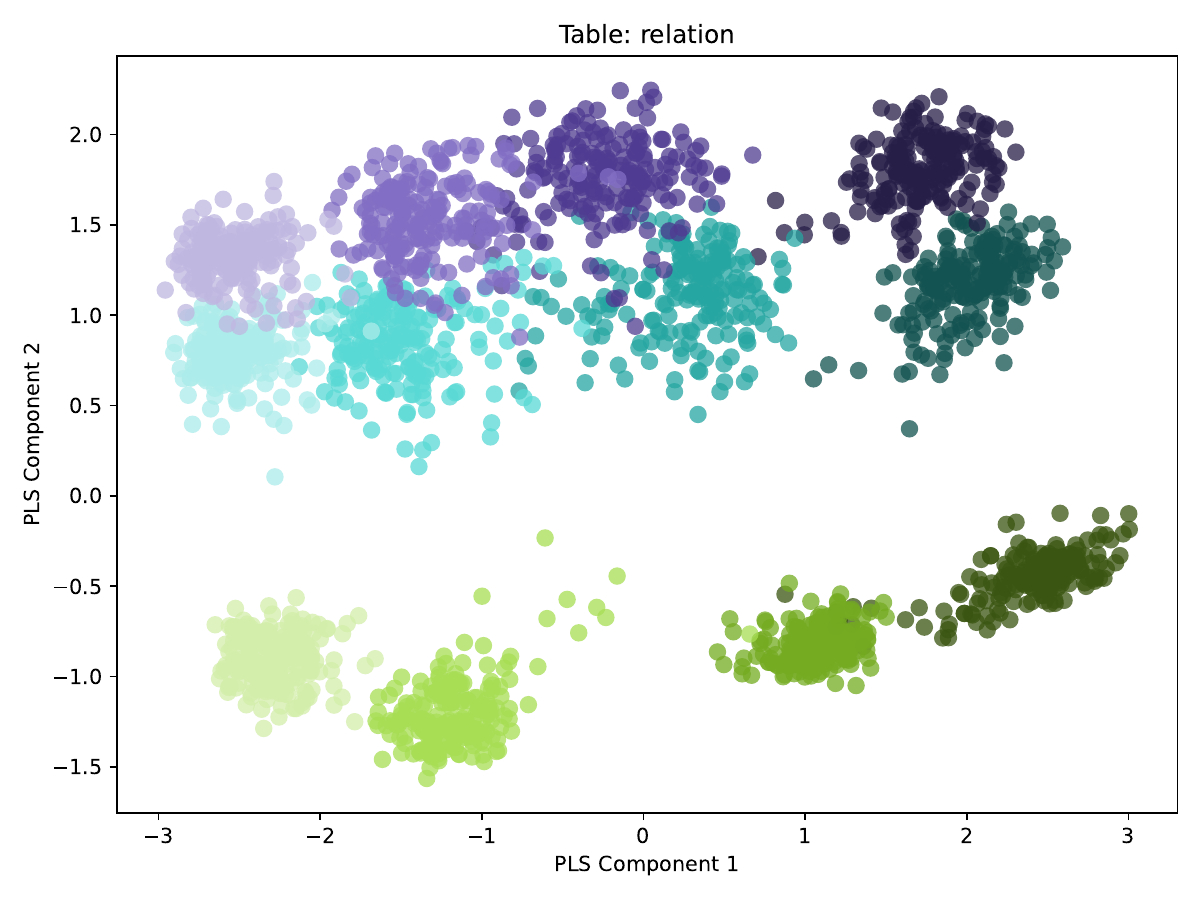}
  \caption{$C_{relation}$}
\end{subfigure}\hfil 
\begin{subfigure}{0.3\textwidth}
  \includegraphics[width=\linewidth]{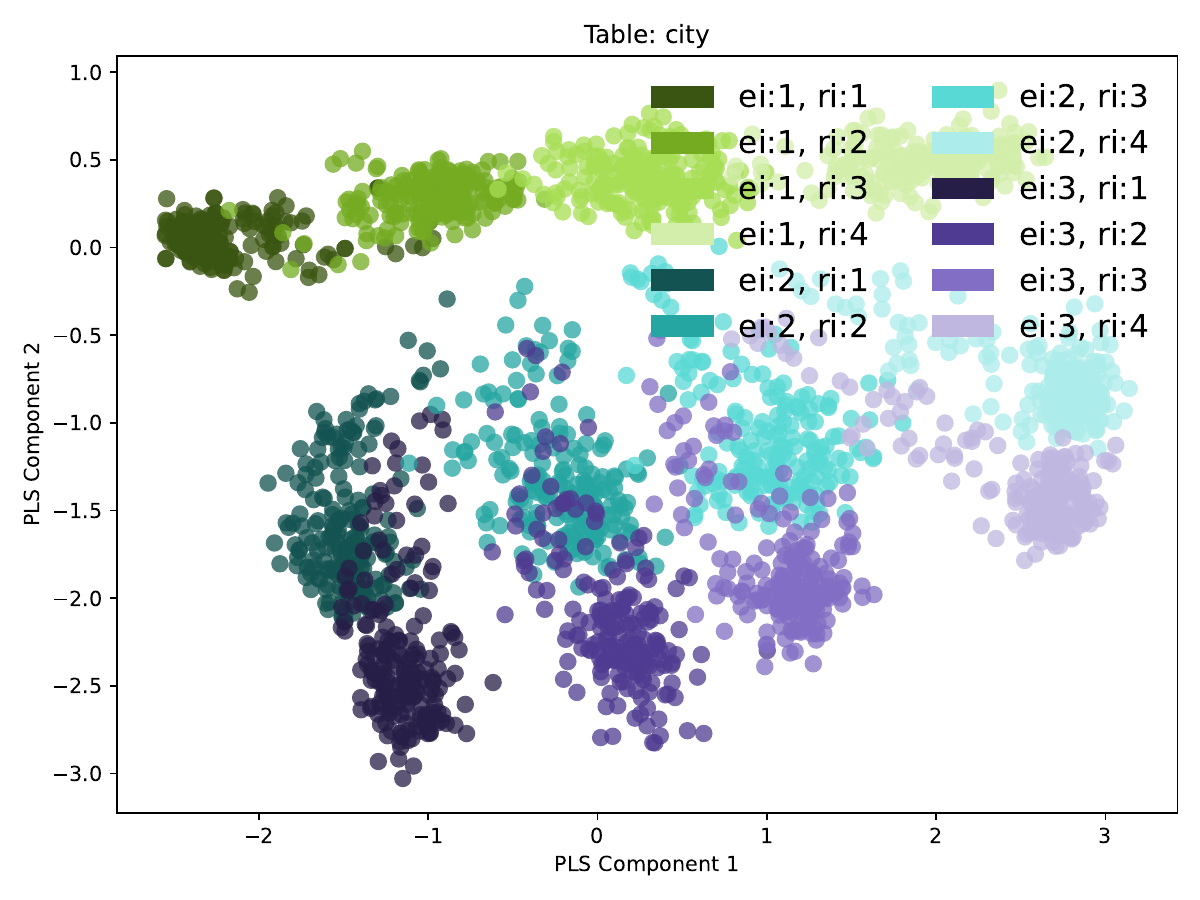}
  \caption{$C_{city}$}
\end{subfigure}\hfil
\begin{subfigure}{0.3\textwidth}
  \includegraphics[width=\linewidth]{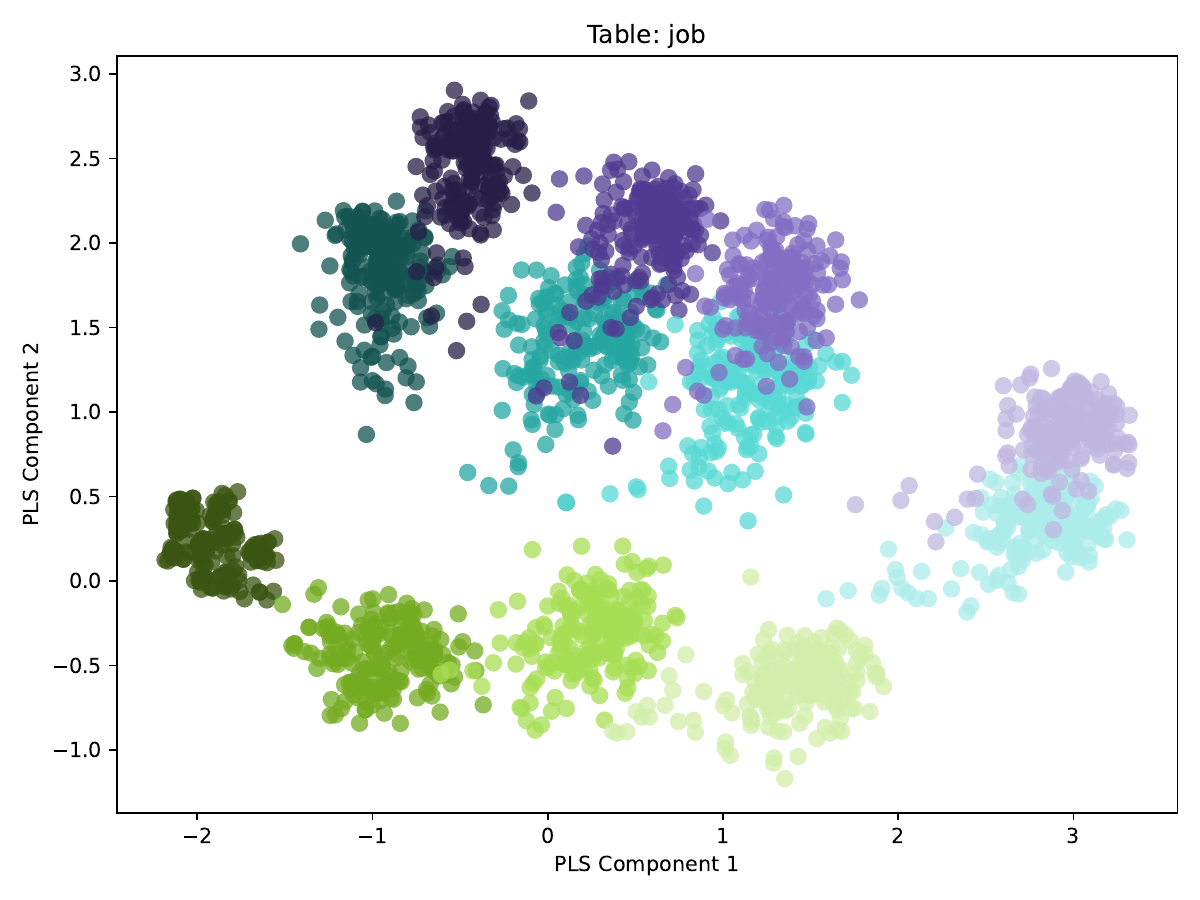}
  \caption{$C_{job}$}
\end{subfigure}\hfil 
\begin{subfigure}{0.3\textwidth}
  \includegraphics[width=\linewidth]{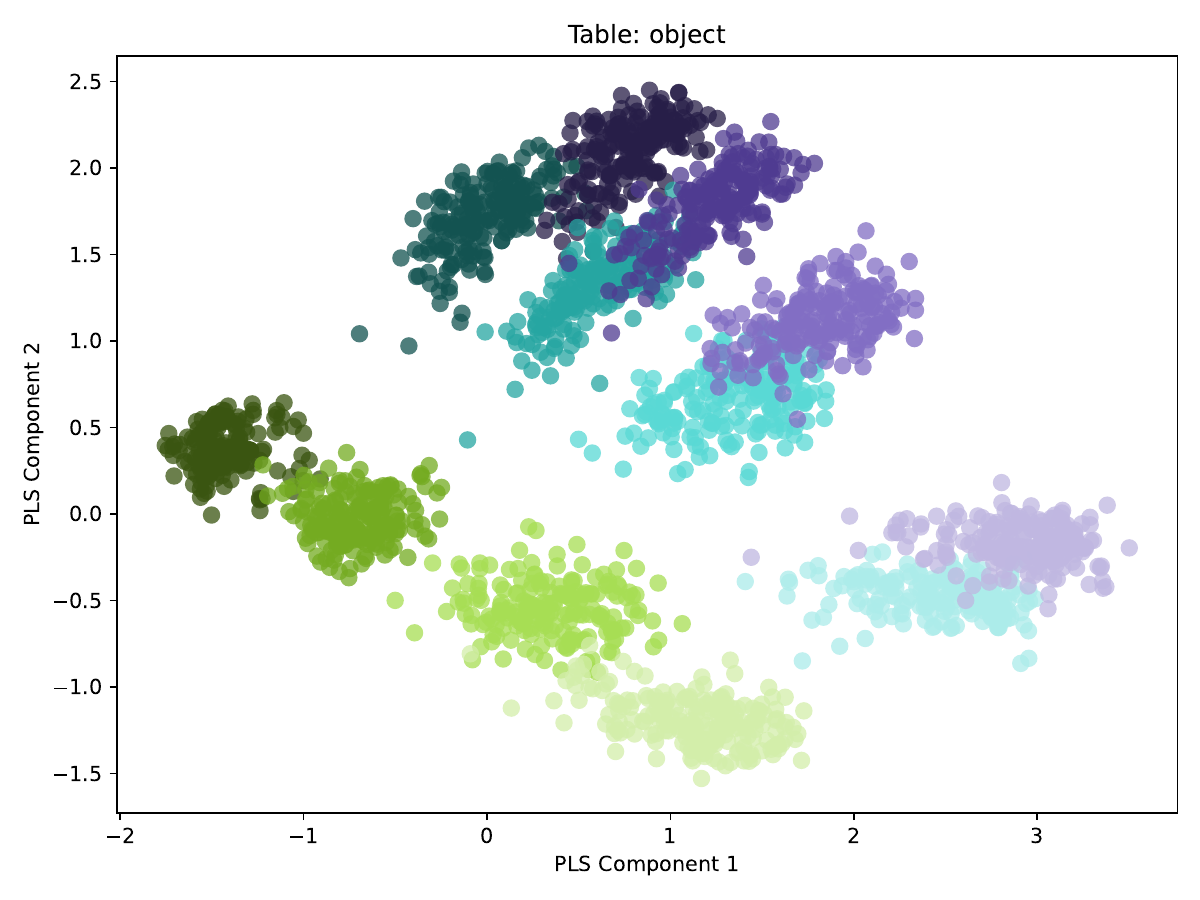}
  \caption{$C_{object}$}
\end{subfigure}\hfil 
\caption{Visualization of the CBR subspace for \textbf{Table Template Input} on Llama3-8B-Instruct.}
\label{fig:irs_visualization_tab_llama}
\end{figure*}

\begin{figure*}[!htbp]
    \centering 
\begin{subfigure}{0.3\textwidth}
  \includegraphics[width=\linewidth]{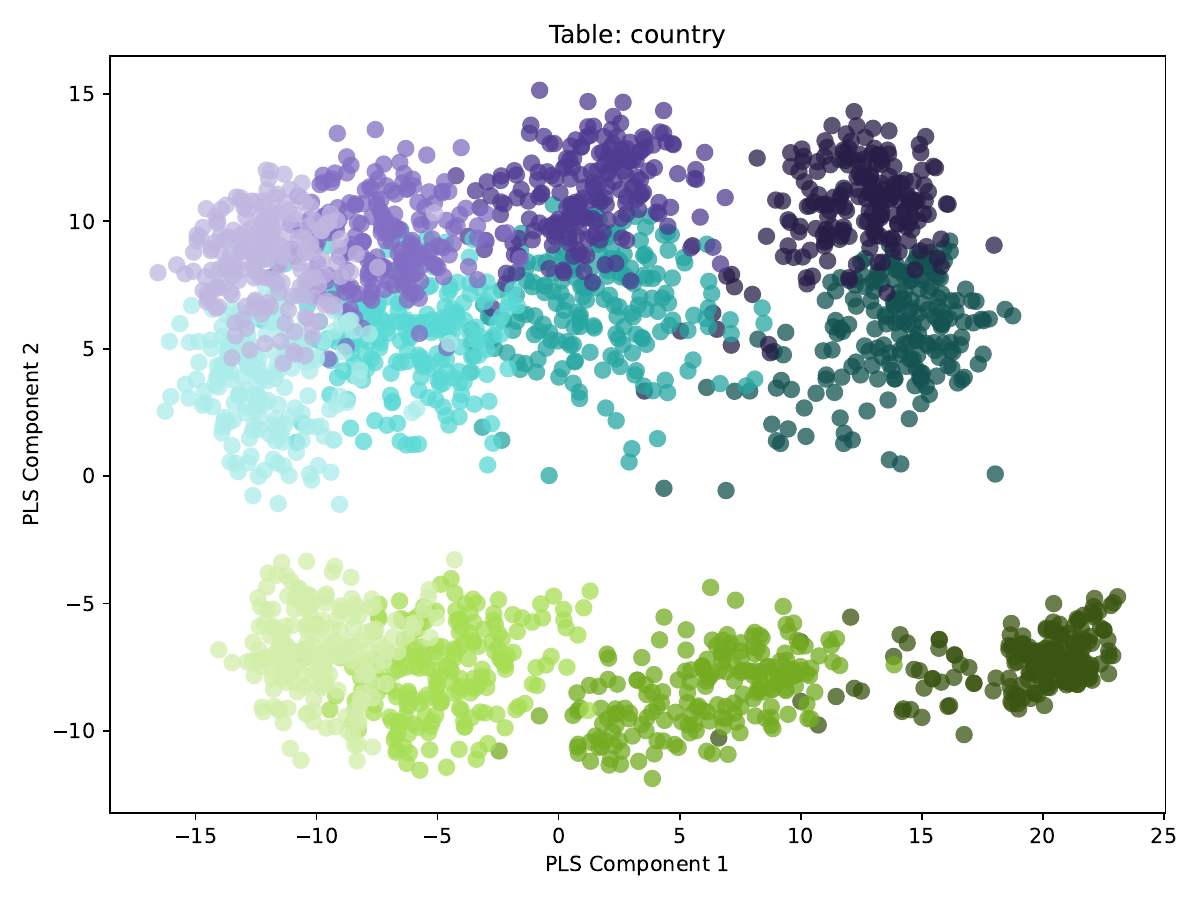}
  \caption{$C_{country}$}
\end{subfigure}\hfil 
\begin{subfigure}{0.3\textwidth}
  \includegraphics[width=\linewidth]{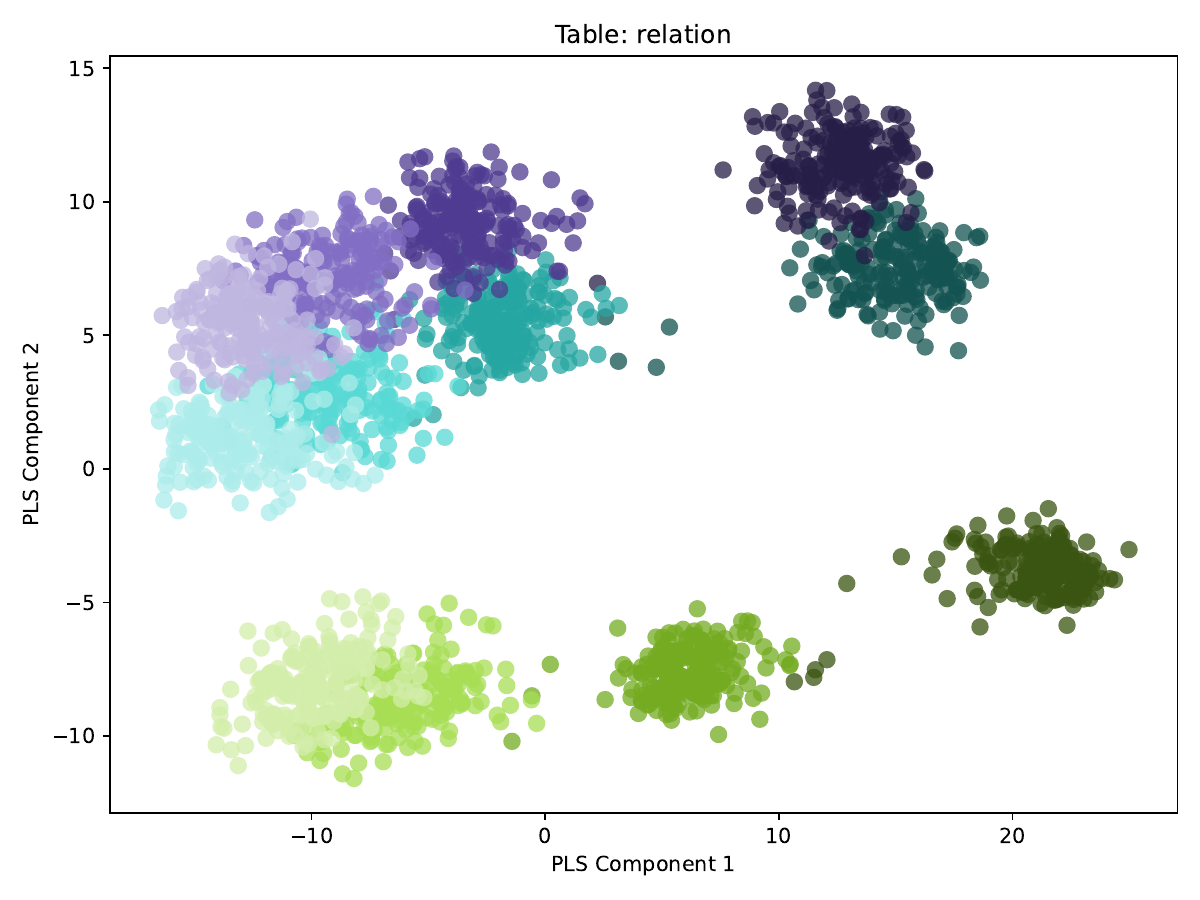}
  \caption{$C_{relation}$}
\end{subfigure}\hfil 
\begin{subfigure}{0.3\textwidth}
  \includegraphics[width=\linewidth]{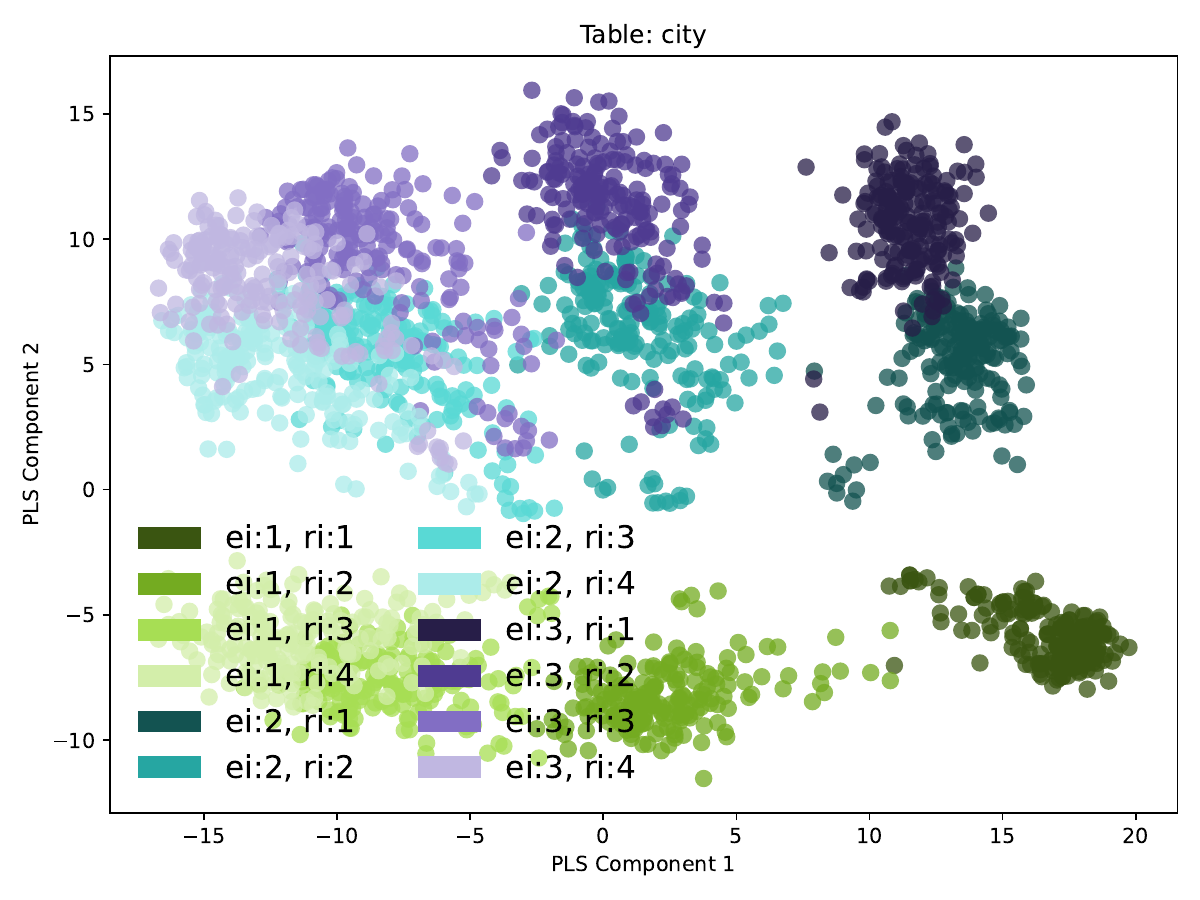}
  \caption{$C_{city}$}
\end{subfigure}\hfil
\begin{subfigure}{0.3\textwidth}
  \includegraphics[width=\linewidth]{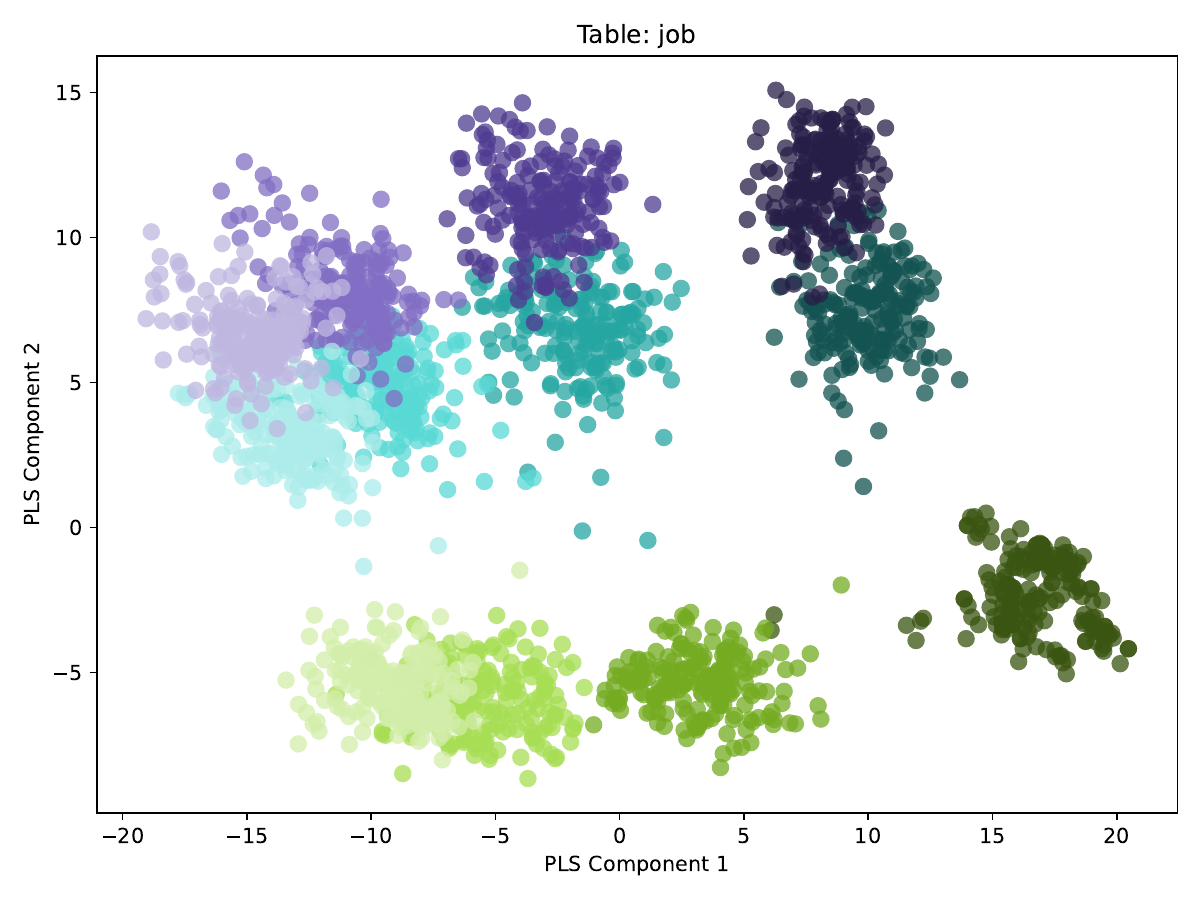}
  \caption{$C_{job}$}
\end{subfigure}\hfil 
\begin{subfigure}{0.3\textwidth}
  \includegraphics[width=\linewidth]{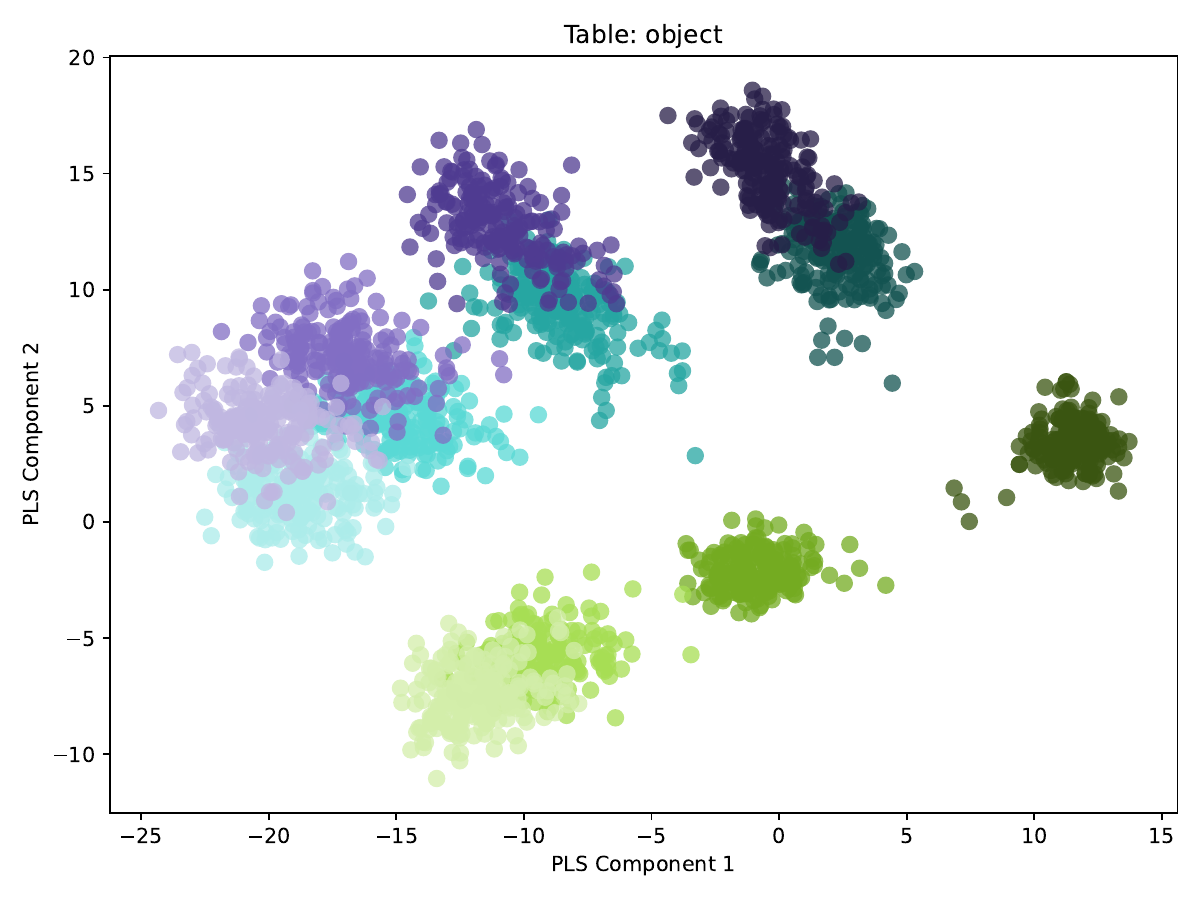}
  \caption{$C_{object}$}
\end{subfigure}\hfil 
\caption{Visualization of the CBR subspace for \textbf{Table Template Input} on Qwen3-8B.}
\label{fig:irs_visualization_tab_qwen}
\end{figure*}

\begin{figure*}[!htbp]
    \centering 
\begin{subfigure}{0.3\textwidth}
  \includegraphics[width=\linewidth]{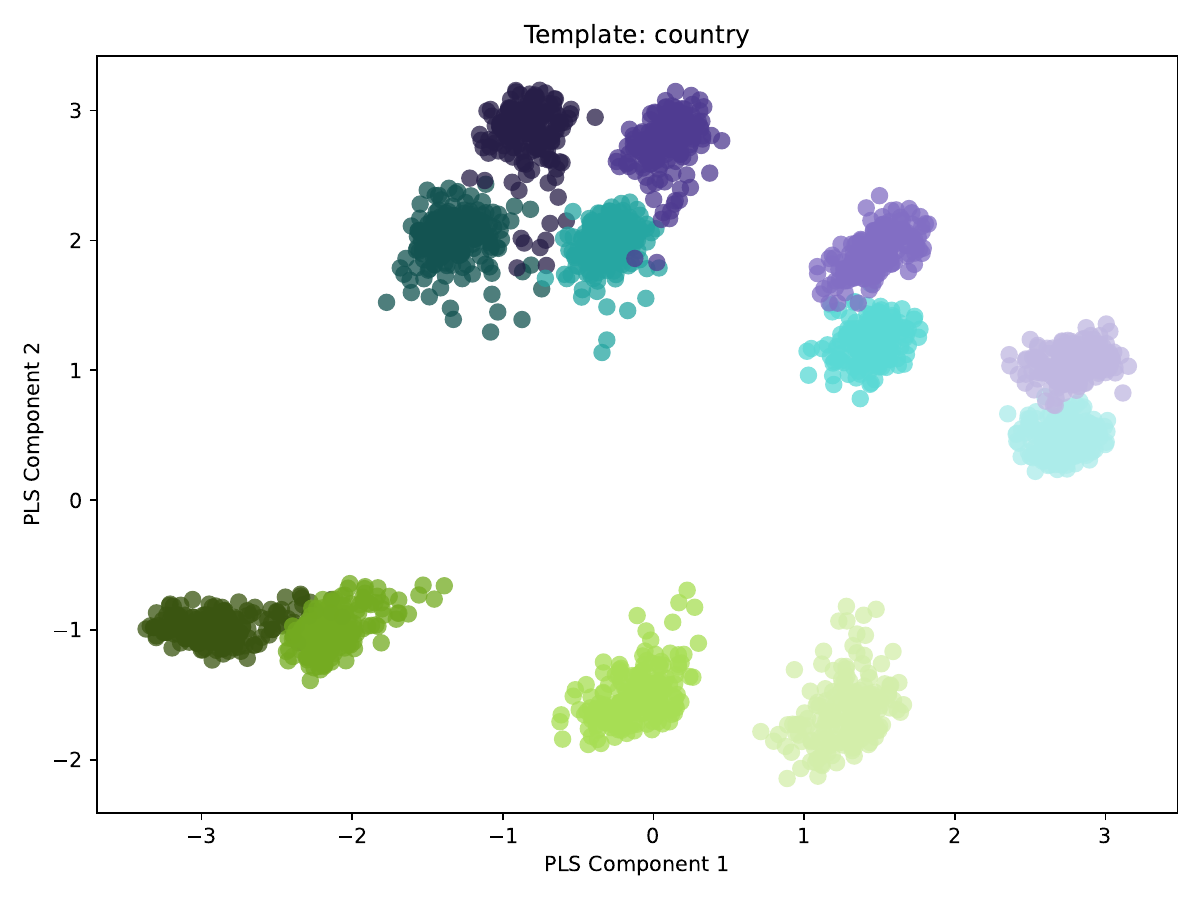}
  \caption{$C_{country}$}
\end{subfigure}\hfil 
\begin{subfigure}{0.3\textwidth}
  \includegraphics[width=\linewidth]{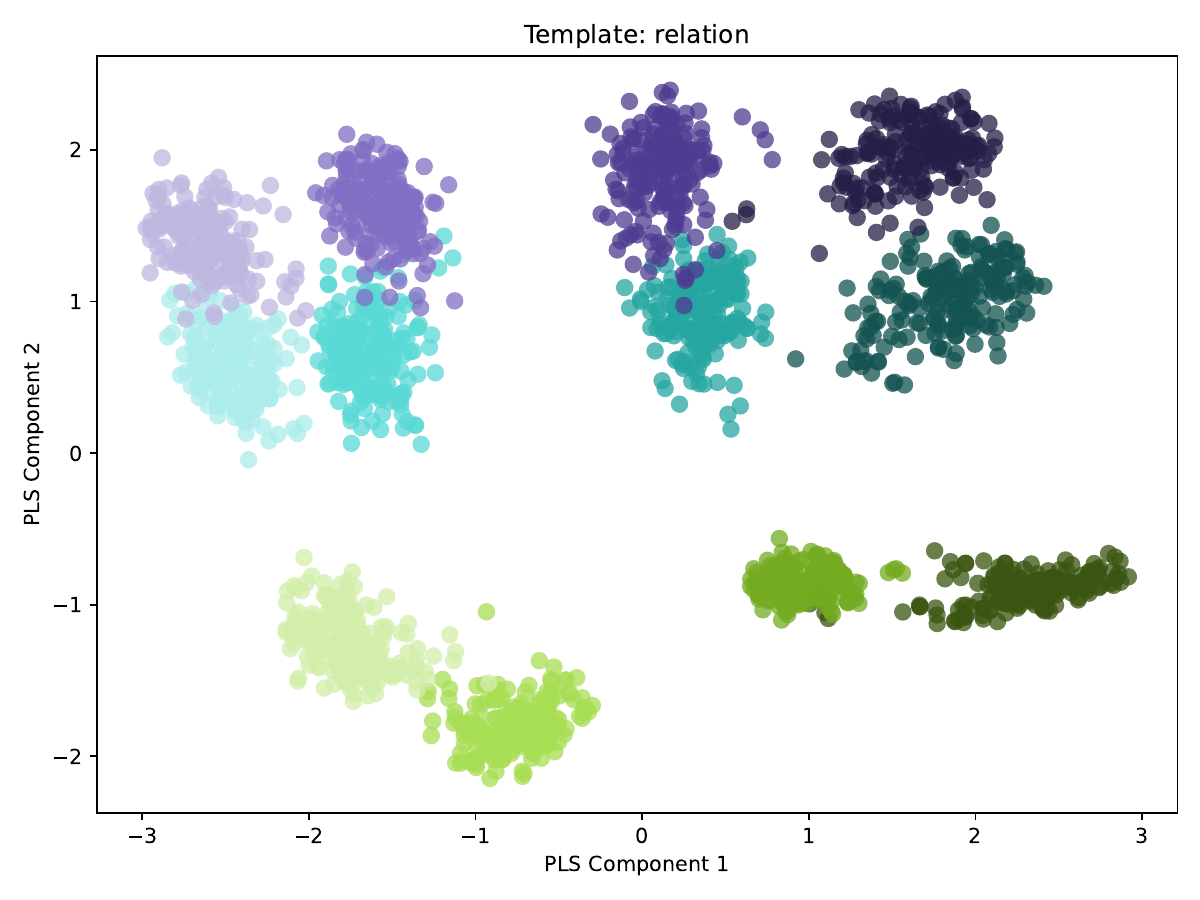}
  \caption{$C_{relation}$}
\end{subfigure}\hfil 
\begin{subfigure}{0.3\textwidth}
  \includegraphics[width=\linewidth]{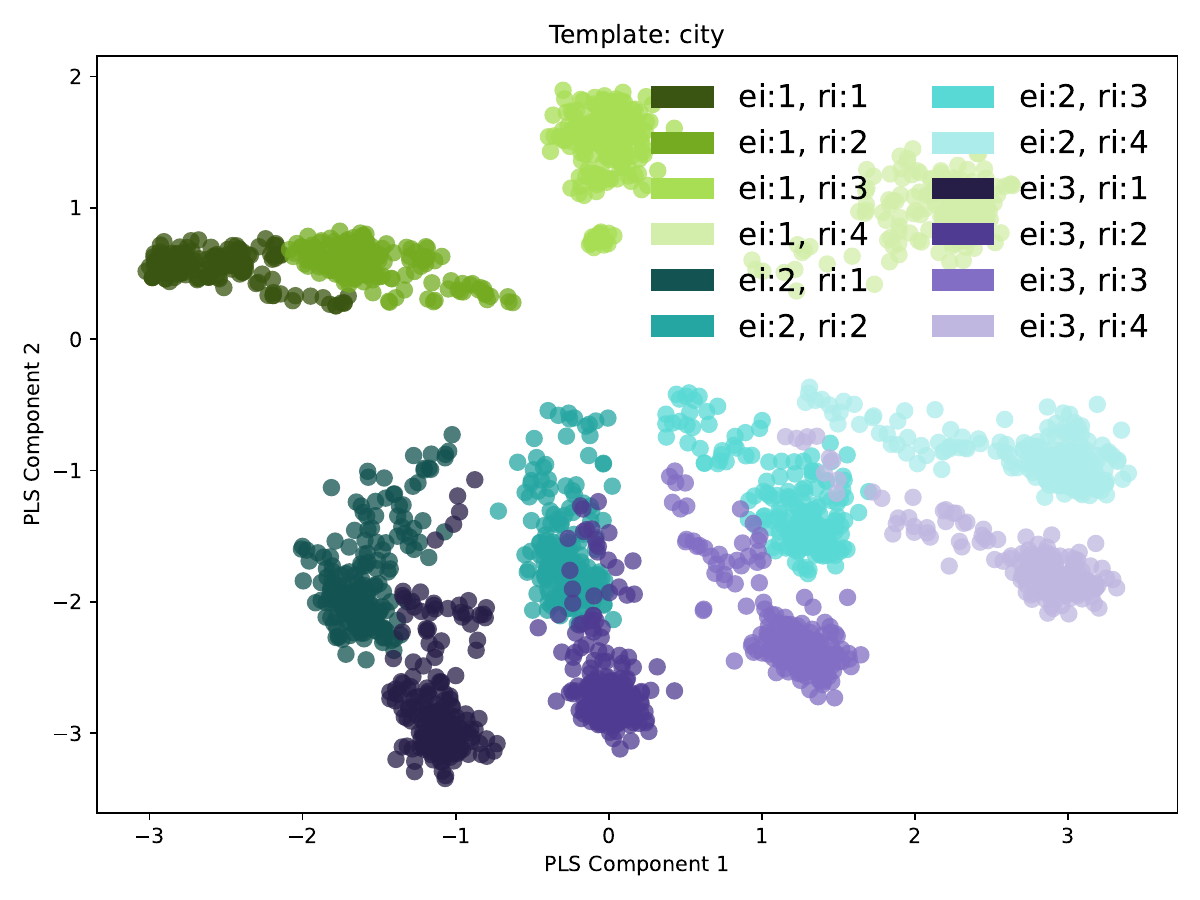}
  \caption{$C_{city}$}
\end{subfigure}\hfil
\begin{subfigure}{0.3\textwidth}
  \includegraphics[width=\linewidth]{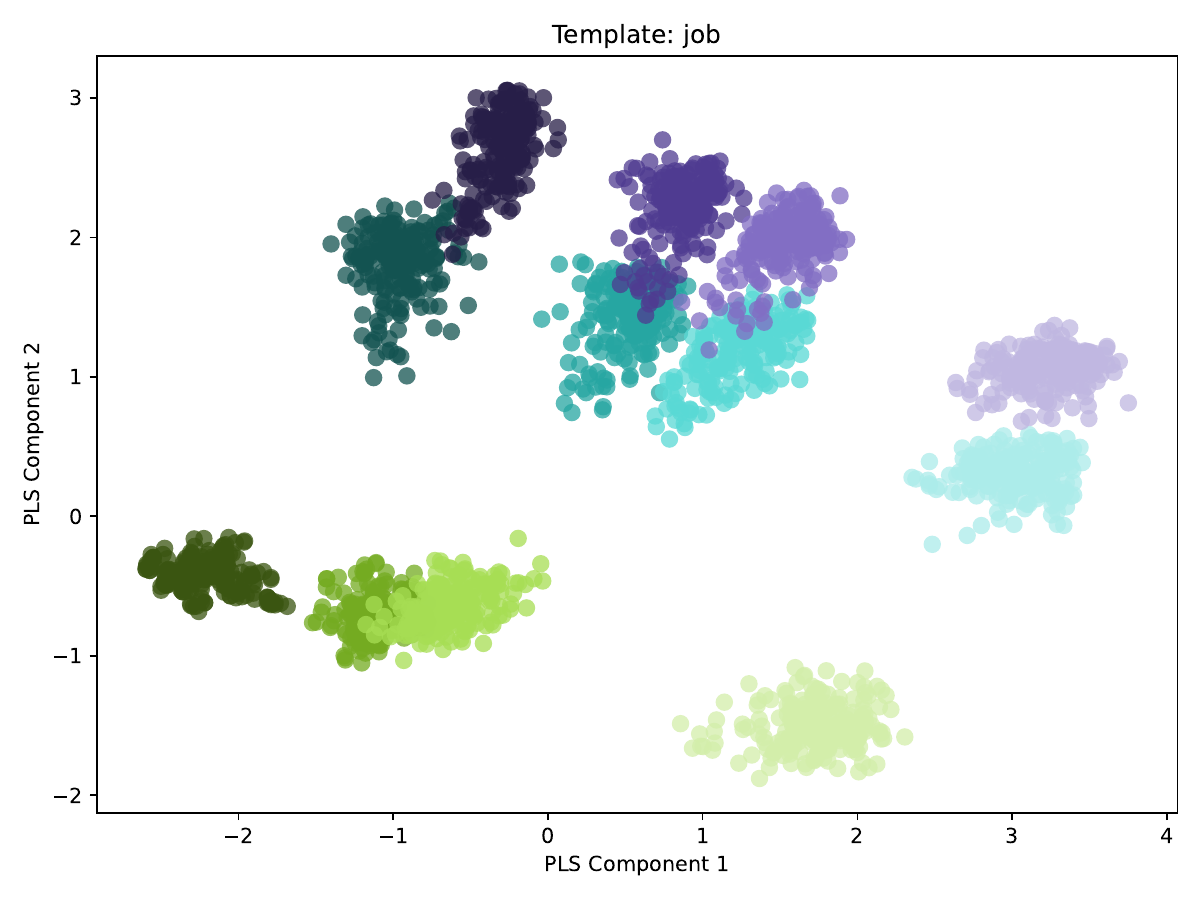}
  \caption{$C_{job}$}
\end{subfigure}\hfil 
\begin{subfigure}{0.3\textwidth}
  \includegraphics[width=\linewidth]{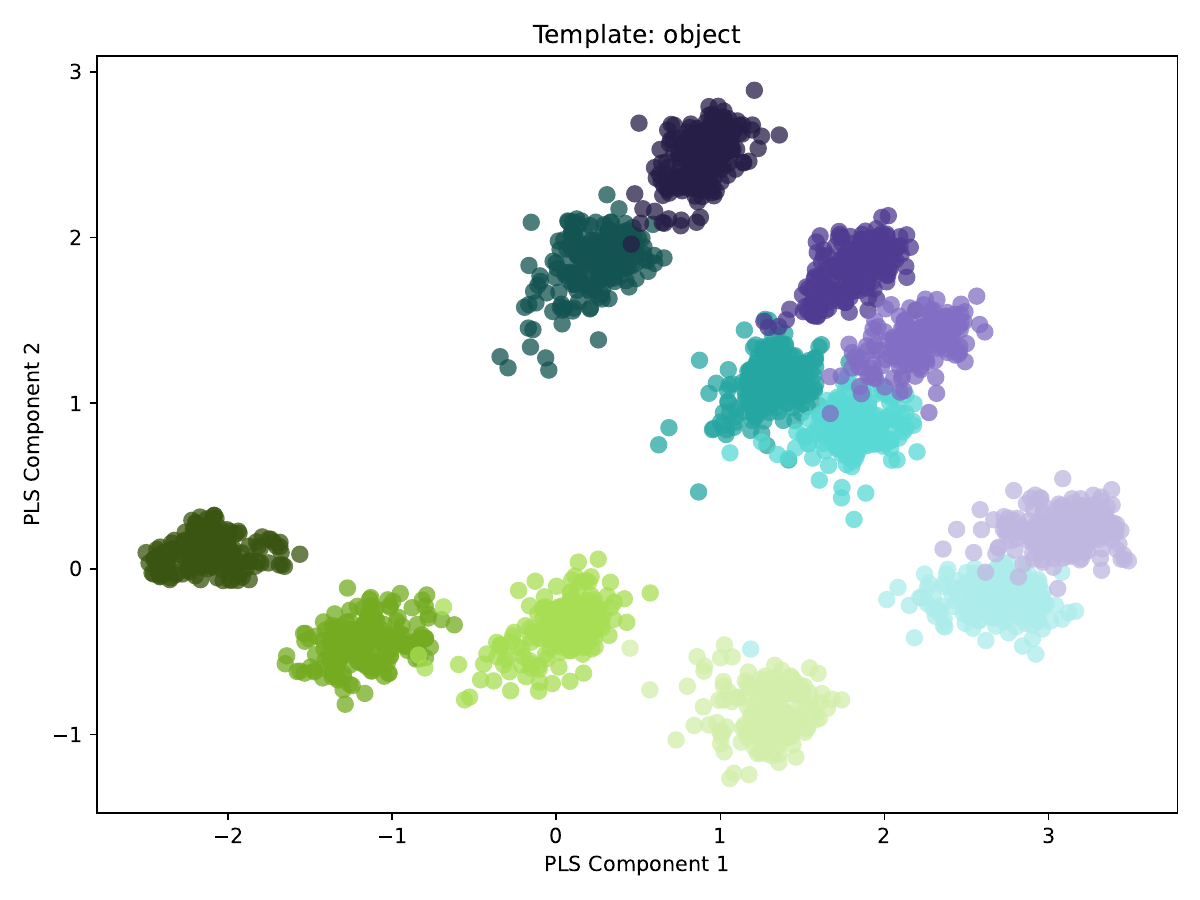}
  \caption{$C_{object}$}
\end{subfigure}\hfil 
\caption{Visualization of the CBR subspace for \textbf{Discourse Template Input} on Llama3-8B-Instruct.}
\label{fig:irs_visualization_temp_llama}
\end{figure*}

\begin{figure*}[!htbp]
    \centering 
\begin{subfigure}{0.3\textwidth}
  \includegraphics[width=\linewidth]{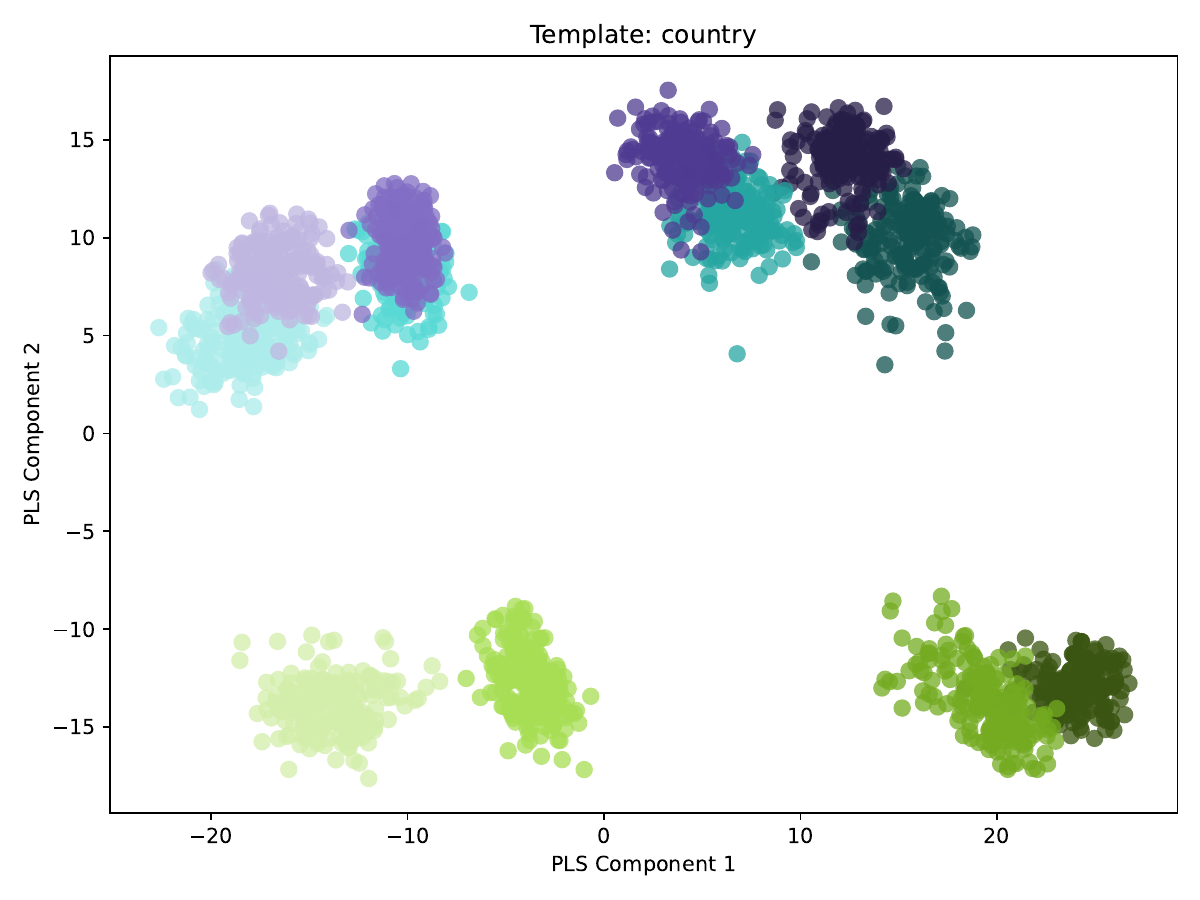}
  \caption{$C_{country}$}
\end{subfigure}\hfil 
\begin{subfigure}{0.3\textwidth}
  \includegraphics[width=\linewidth]{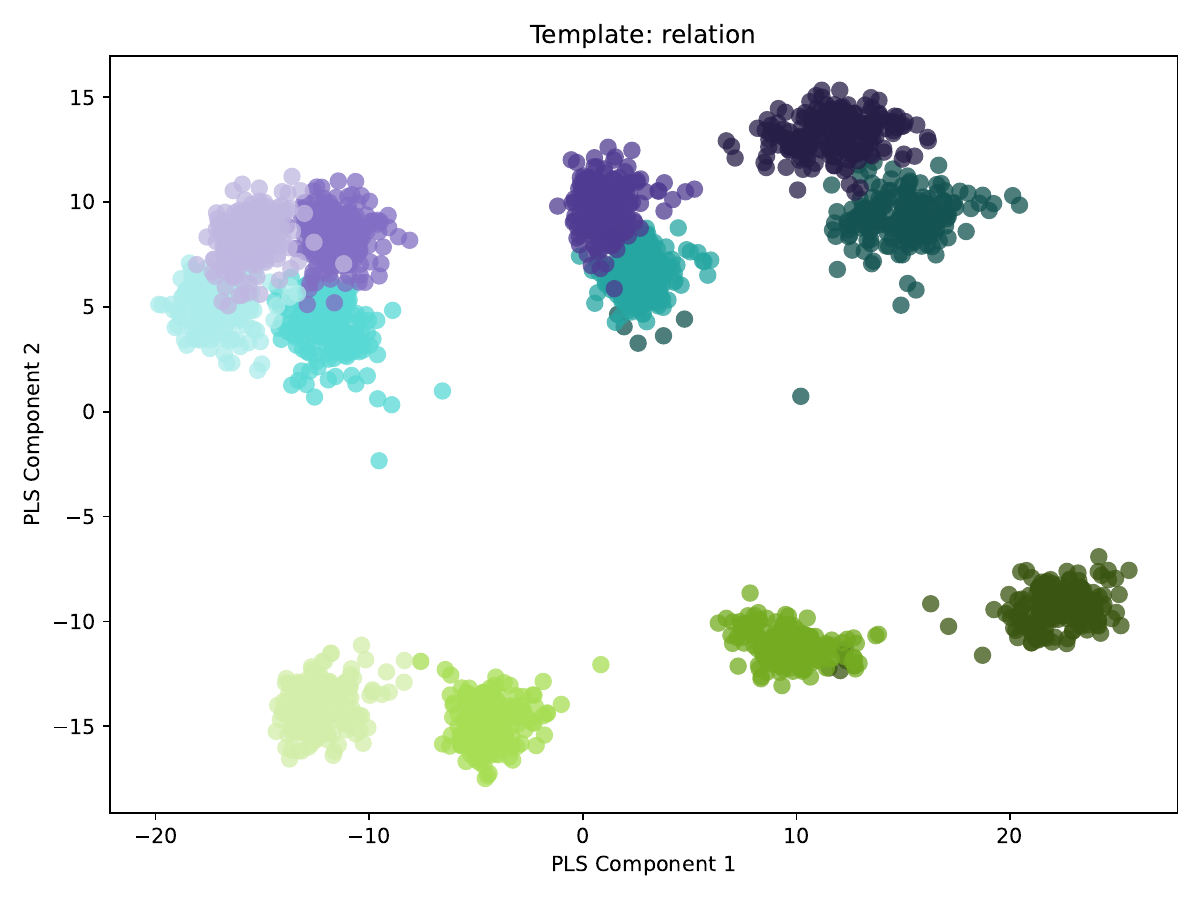}
  \caption{$C_{relation}$}
\end{subfigure}\hfil 
\begin{subfigure}{0.3\textwidth}
  \includegraphics[width=\linewidth]{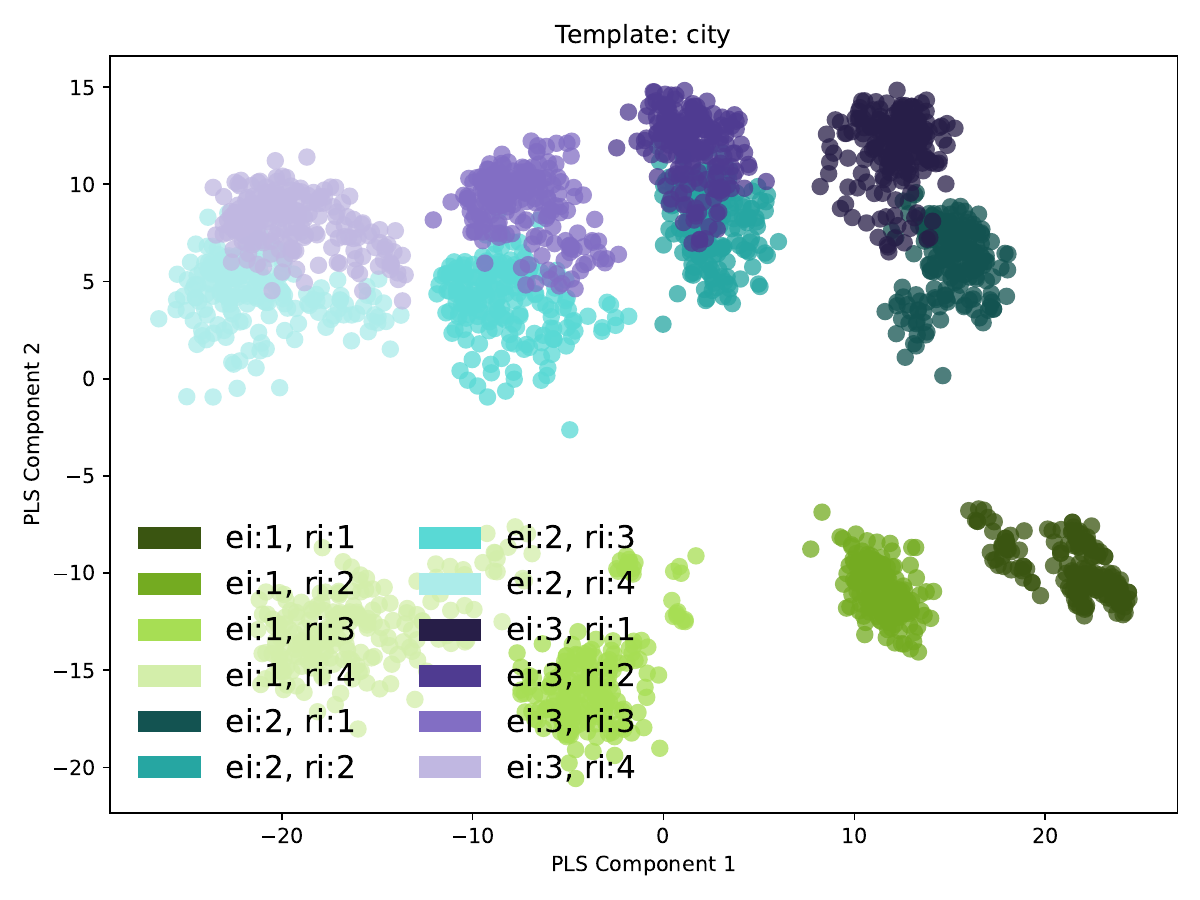}
  \caption{$C_{city}$}
\end{subfigure}\hfil
\begin{subfigure}{0.3\textwidth}
  \includegraphics[width=\linewidth]{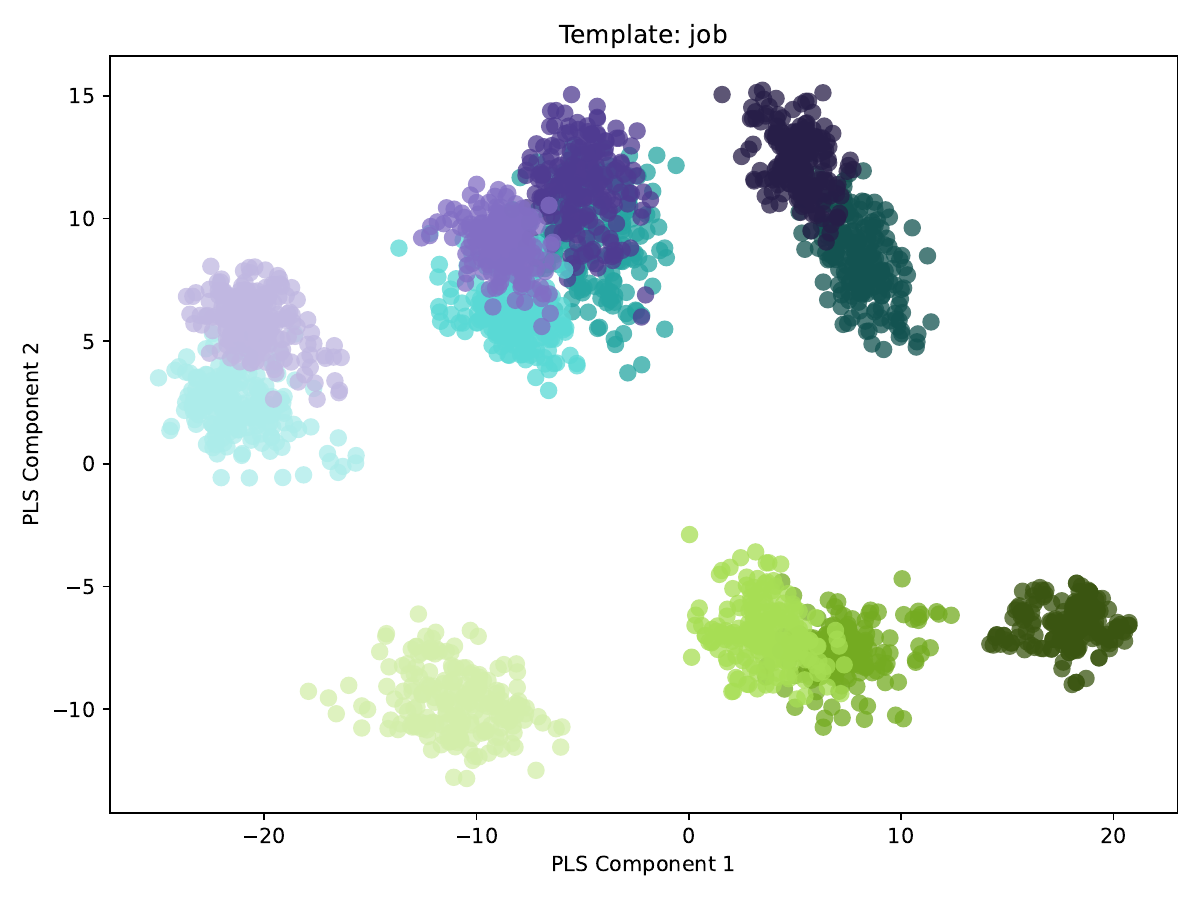}
  \caption{$C_{job}$}
\end{subfigure}\hfil 
\begin{subfigure}{0.3\textwidth}
  \includegraphics[width=\linewidth]{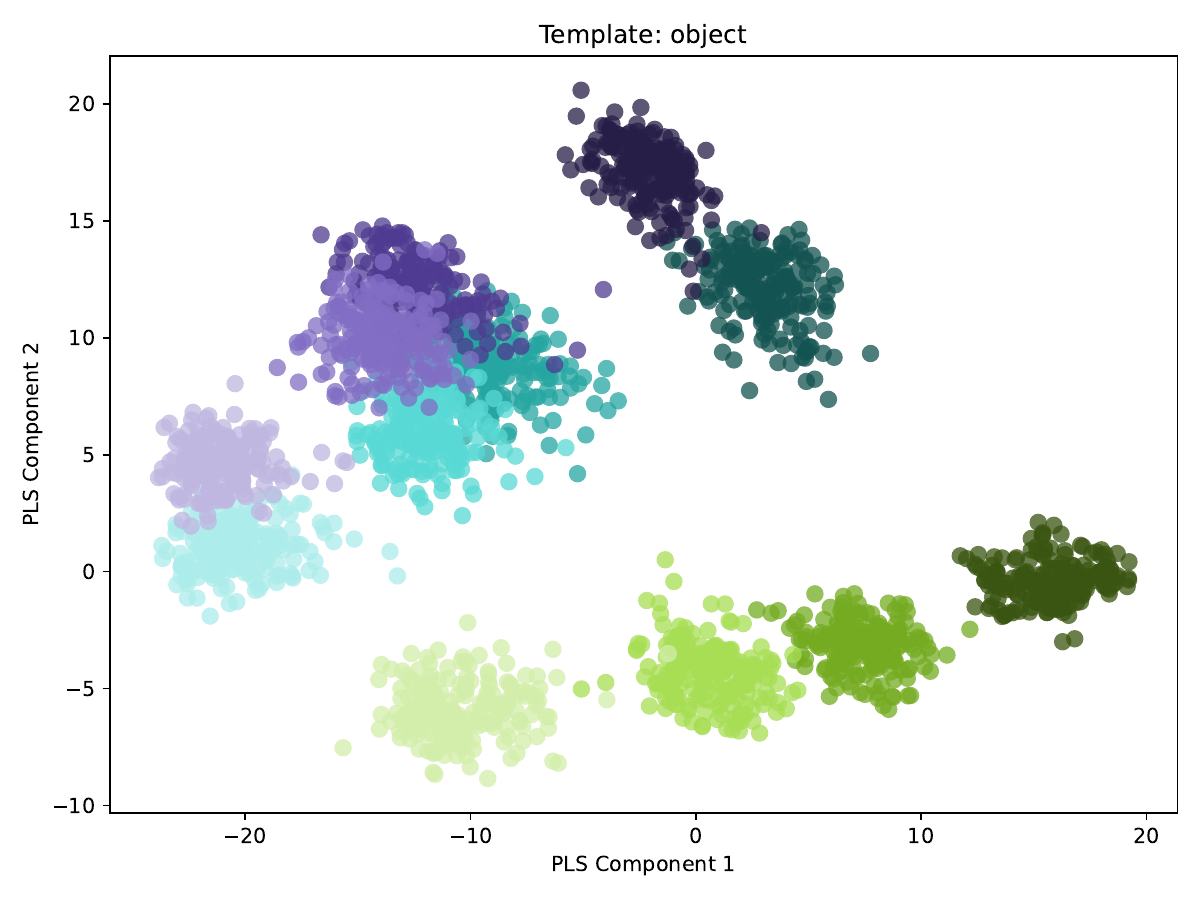}
  \caption{$C_{object}$}
\end{subfigure}\hfil 
\caption{Visualization of the CBR subspace for \textbf{Discourse Template Input} on Qwen3-8B.}
\label{fig:irs_visualization_temp_qwen}
\end{figure*}

The CBR subspace visualization for Template Input (i.e., Table Template Input as sampled in Table~\ref{tab:datat_temp_other_table}, and Discourse Template Input as sampled in Table~\ref{tab:datat_temp_other_temp}), shown in Figure~\ref{fig:irs_visualization_tab_llama}, \ref{fig:irs_visualization_tab_qwen}, \ref{fig:irs_visualization_temp_llama} and \ref{fig:irs_visualization_temp_qwen}, are similarly organized along the entity and relation index, indicating that the CBR subspace is a general geometric property common to Table Template Input and Discourse Template Input.


\clearpage

\subsection{Activation Steering on Template Input}
\label{sec:irs_steering_tabtemp}
\begin{figure*}[!htbp]
    \centering 
\begin{subfigure}{0.45\textwidth}
  \includegraphics[width=\linewidth]{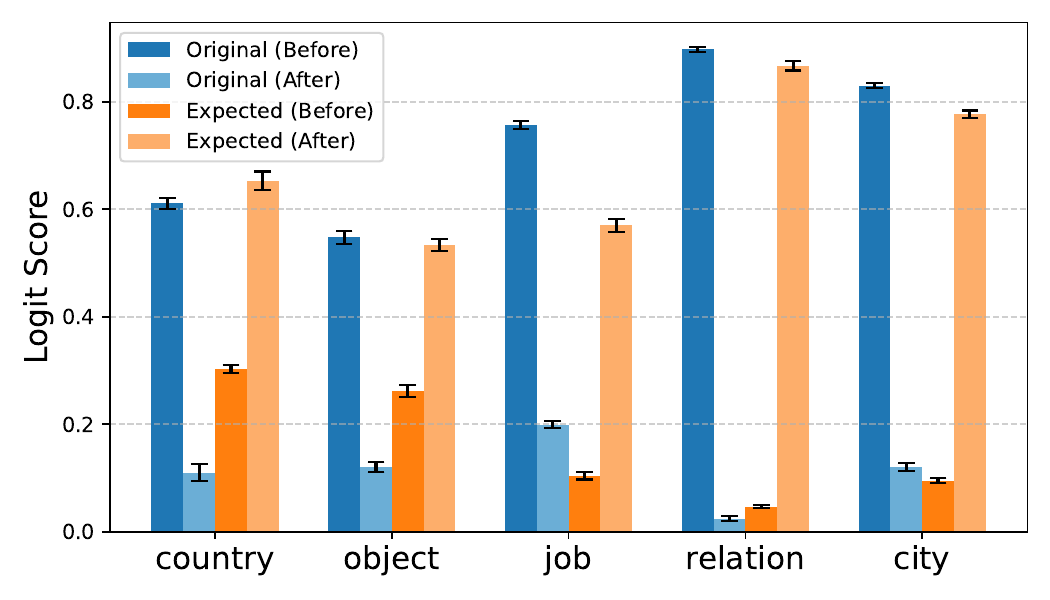}
  \caption{Relation-index (i.e., $ri$) steering on the attribute token.}
  \label{fig:steer_llama_tab_att}
\end{subfigure}\hfil 
\begin{subfigure}{0.45\textwidth}
  \includegraphics[width=\linewidth]{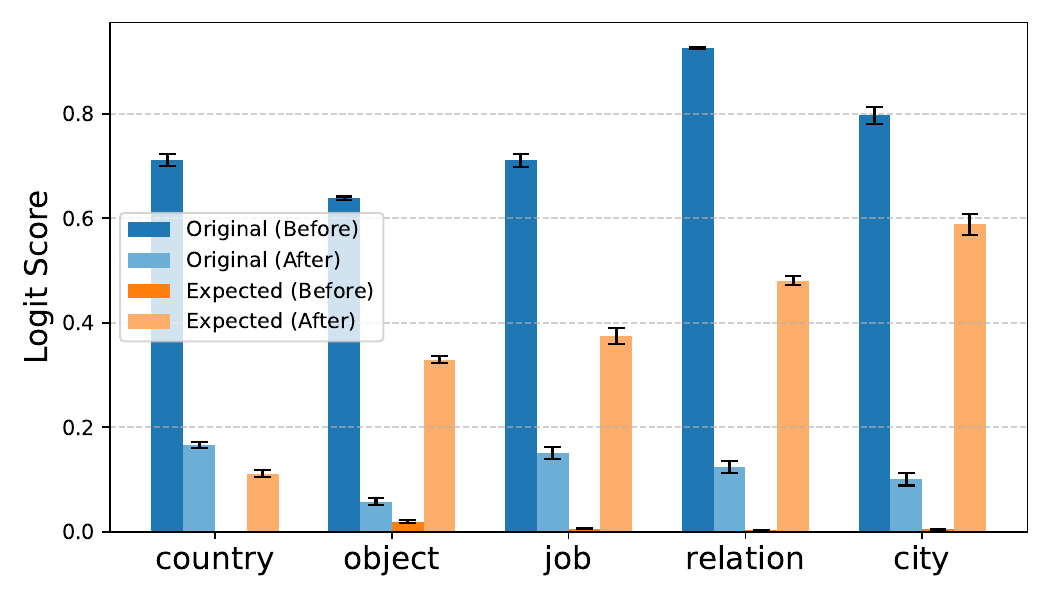}
  \caption{Entity-index (i.e., $ei$) steering on the entity token.}
  \label{fig:steer_llama_tab_ent}
\end{subfigure}\hfil 
\caption{Activation patching on \textbf{Table Template Input} in query part across five contexts on Llama3-8B-Instruct.}
\label{fig:steer_query_tab_llama}
\end{figure*}

\begin{figure*}[!htbp]
    \centering 
\begin{subfigure}{0.45\textwidth}
  \includegraphics[width=\linewidth]{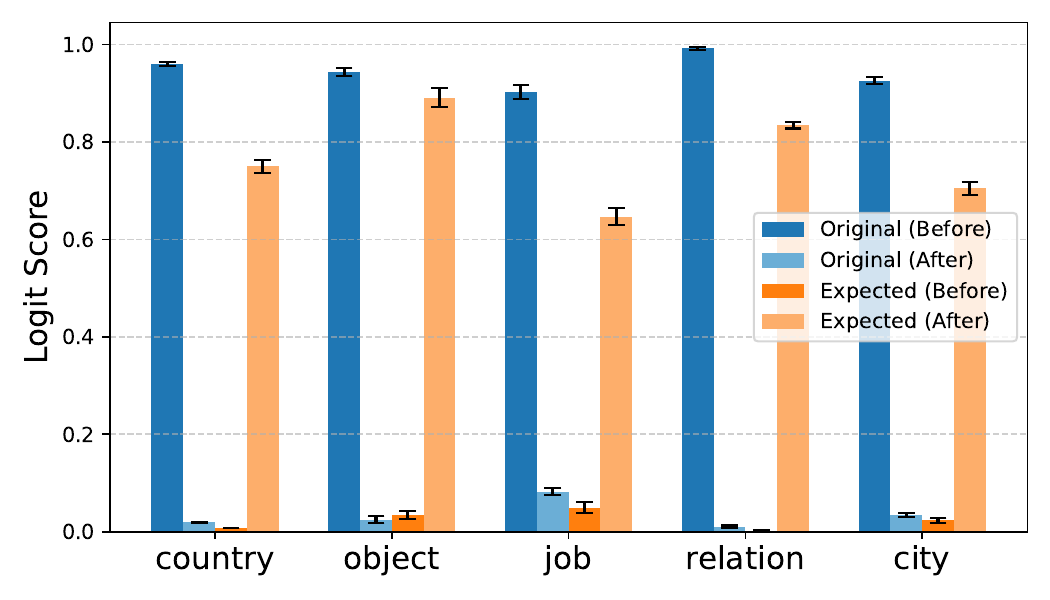}
  \caption{Relation-index (i.e., $ri$) steering on the attribute token.}
  \label{fig:steer_qwen_tab_att}
\end{subfigure}\hfil 
\begin{subfigure}{0.45\textwidth}
  \includegraphics[width=\linewidth]{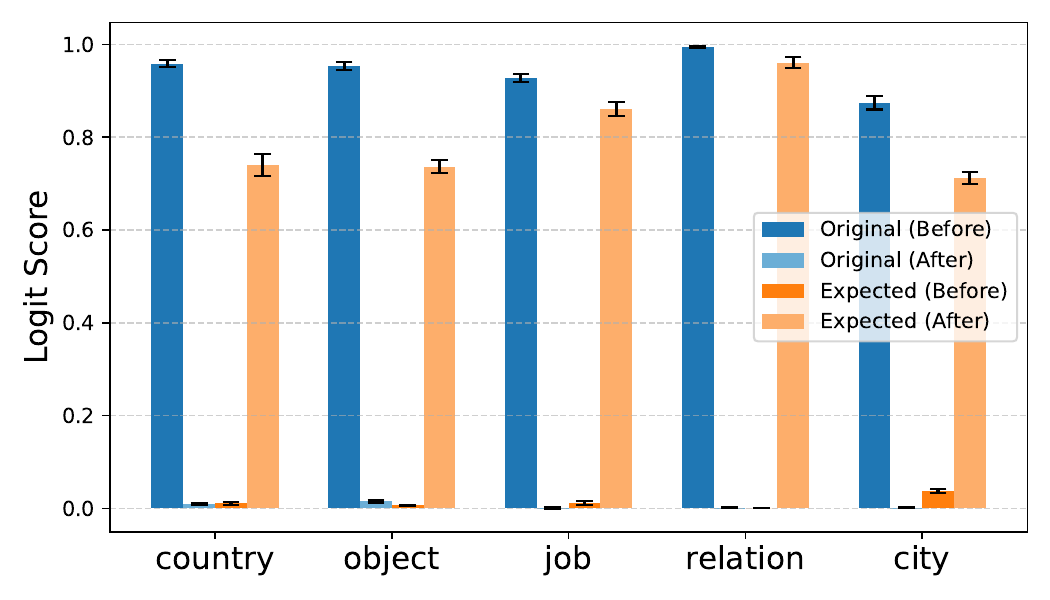}
  \caption{Entity-index (i.e., $ei$) steering on the entity token.}
  \label{fig:steer_qwen_tab_ent}
\end{subfigure}\hfil 
\caption{Activation patching on \textbf{Table Template Input} in query part across five contexts on Qwen3-8B.}
\label{fig:steer_query_tab_qwen}
\end{figure*}

\begin{figure*}[!htbp]
    \centering 
\begin{subfigure}{0.45\textwidth}
  \includegraphics[width=\linewidth]{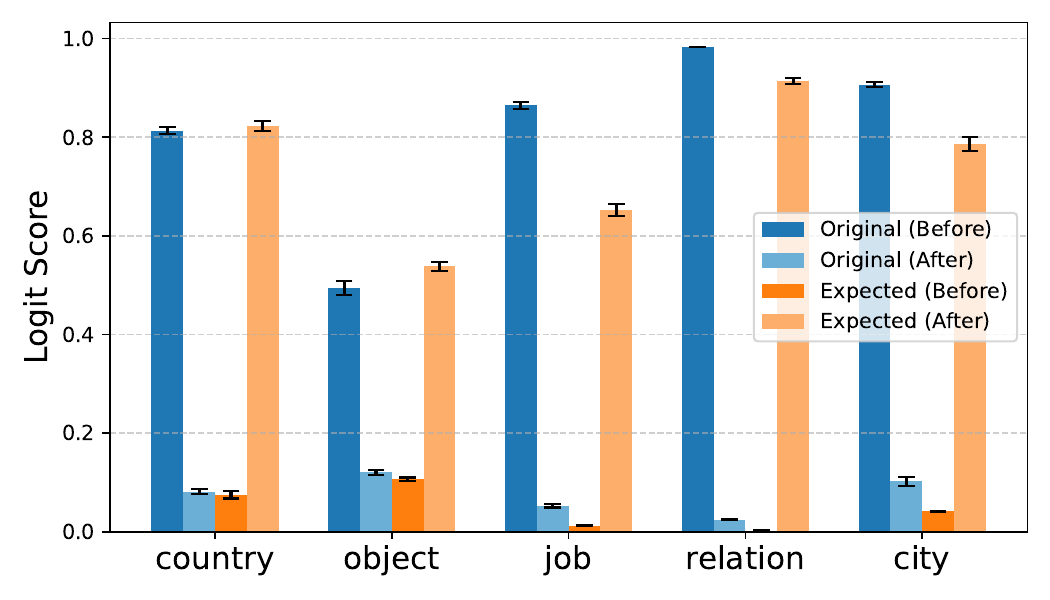}
  \caption{Relation-index (i.e., $ri$) steering on the attribute token.}
  \label{fig:steer_llama_temp_att}
\end{subfigure}\hfil 
\begin{subfigure}{0.45\textwidth}
  \includegraphics[width=\linewidth]{graph/steer_llama_tab_ent.pdf}
  \caption{Entity-index (i.e., $ei$) steering on the entity token.}
  \label{fig:steer_llama_temp_ent}
\end{subfigure}\hfil 
\caption{Activation patching on \textbf{Discourse Template Input} in query part across five contexts on Llama3-8B-Instruct.}
\label{fig:steer_query_temp_llama}
\end{figure*}

\begin{figure*}[!htbp]
    \centering 
\begin{subfigure}{0.45\textwidth}
  \includegraphics[width=\linewidth]{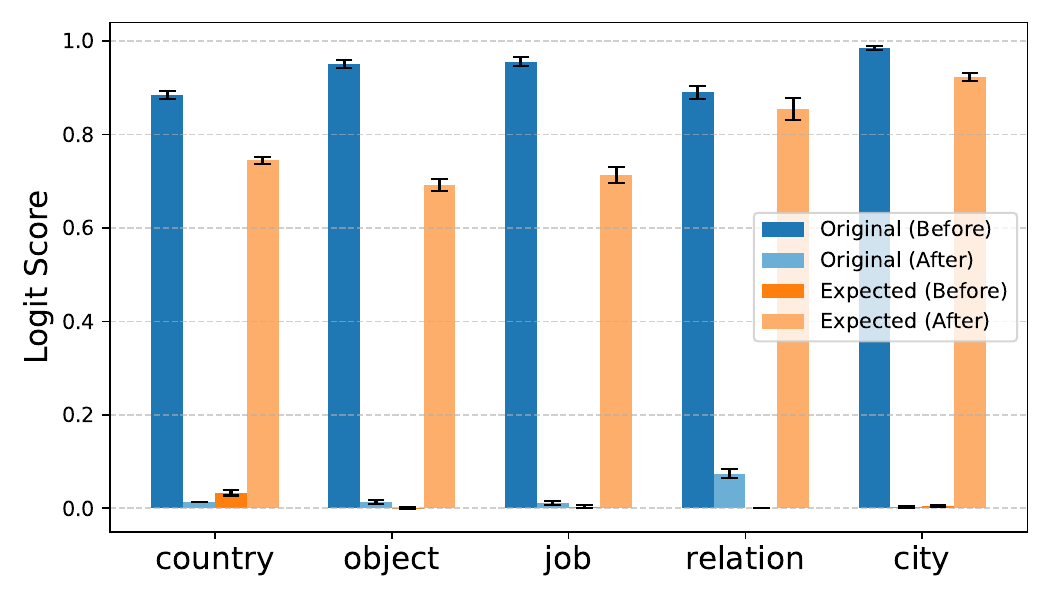}
  \caption{Relation-index (i.e., $ri$) steering on the attribute token.}
  \label{fig:steer_qwen_temp_att}
\end{subfigure}\hfil 
\begin{subfigure}{0.45\textwidth}
  \includegraphics[width=\linewidth]{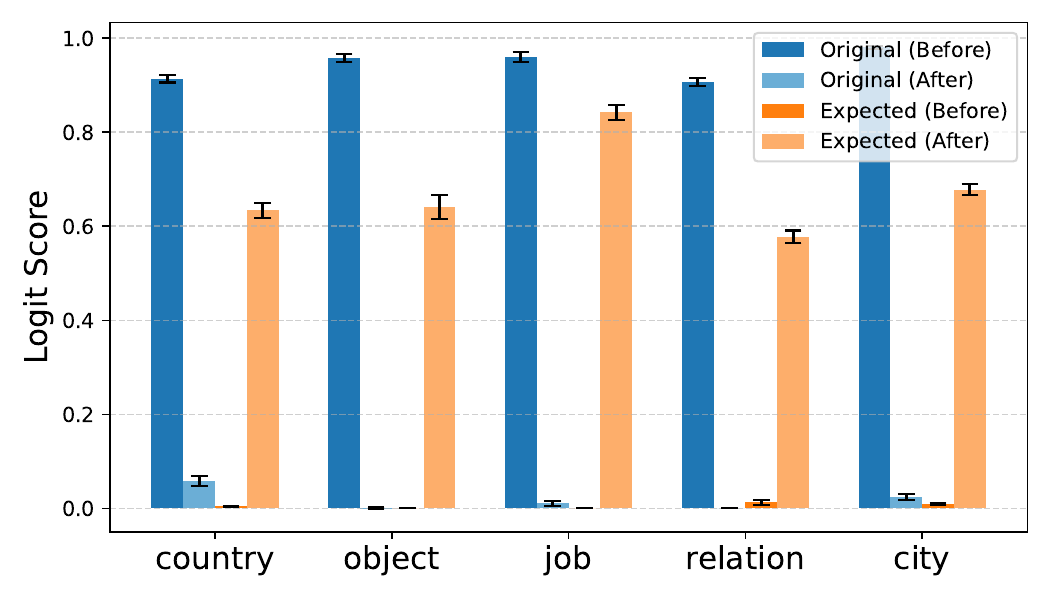}
  \caption{Entity-index (i.e., $ei$) steering on the entity token.}
  \label{fig:steer_qwen_temp_ent}
\end{subfigure}\hfil 
\caption{Activation patching on \textbf{Table Template Input} in query part across five contexts on Qwen3-8B.}
\label{fig:steer_query_temp_qwen}
\end{figure*}

The results of activation steering on Table Template Input and Discourse Template Input, as described in Section~\secref{sec:dataset}, under the one-shot query setting are shown in Figure~\ref{fig:steer_llama_tab_att}, \ref{fig:steer_llama_tab_ent},  \ref{fig:steer_qwen_tab_att}, \ref{fig:steer_qwen_tab_ent}, \ref{fig:steer_llama_temp_att}, \ref{fig:steer_llama_temp_ent},  \ref{fig:steer_qwen_temp_att}, and \ref{fig:steer_qwen_temp_ent}, These results demonstrate that the logit change patterns are consistent across different input formats, further confirming the prevalence of the CBR subspace based mechanism across input formats.

%
%

%

\subsection{Comparison with Hessian-Based Binding Analysis}
\label{sec:genderbias}

Identifying structured subspaces that encode specific functions (e.g., binding) enables more precise monitoring and intervention in model behavior than prompt-level manipulation alone. \citet{feng2024monitoring} propose a Hessian-based algorithm to monitor latent world states and identify binding tokens in LLMs. Their method extracts a binding subspace that encodes entity–attribute associations, enabling the recovery of correct bindings from internal representations. For example, in the sample: ``\textit{The nurse lives in Singapore. The \underline{CEO} lives in Canada. The person living in Singapore is male. The person living in Canada is \underline{female}.}'', the method correctly identifies the binding between ``\textit{CEO}'' and ``\textit{female}'', rather than the stereotypical association ``\textit{CEO--male}''. The authors report that this approach outperforms a prompt-based baseline that directly queries the model (e.g., ``\textit{The gender of the CEO is}'') in gender-bias evaluation settings.

We take the Hessian-based method as a baseline and compare it with our CBR-based framework under the same experimental setting. Specifically, we use the same LLM (Tulu-2-13B), identical templates and entity inventories. As described in Section~\secref{sec:irs_subspace}, our approach learns a projection matrix that maps hidden activations into a CBR index space, enabling the matrix to predict bound entity indices (e.g., $ei=1$ for “\textit{nurse}” and $ei=2$ for “\textit{CEO}”) from the activation of a target token (e.g., “\textit{female}”). To prevent data leakage, the projection matrix is learned using the dataset constructed from different entity inventories, while evaluation is performed on the dataset generated under the same template and settings, which contains $400$ samples.

The evaluation is conducted in two conditions: a \textit{pro} condition that follows stereotypical bias (e.g., “nurse–female”) and an \textit{anti} condition that counteracts the bias. Table~\ref{tab:hessian_comparison} reports the results. While the prompt-based baseline performs well in the stereotypical setting, its accuracy drops substantially in the anti-stereotypical case. The Hessian-based approach improves robustness, particularly under the anti condition. Our CBR-based framework achieves strong performance in both settings, with a notable improvement in the anti condition. These results indicate that the CBR-based framework more reliably captures entity--relation bindings in the model’s internal representations. In particular, its strong performance in the anti condition suggests that it can recover correct bindings even when they conflict with stereotypical associations encoded in the model. More broadly, CBR provides a structured way to analyze and monitor internal binding behavior in LLMs, which may support future interpretability and intervention techniques for improving reliability and mitigating bias.

\begin{table}[t]
\centering
\begin{tabular}{lccc}
\hline
Condition & Prompt & Hessian & CBR (ours) \\
\hline
pro  & 1.00$^{\dagger}$ & 0.93$^{\dagger}$ & 0.95 \\
anti & 0.56$^{\dagger}$ & 0.83$^{\dagger}$ & \textbf{0.94} \\
\hline
\end{tabular}
\caption{Comparison with the Hessian-based method. $^{\dagger}$ denotes results estimated from the bar plots in the original paper, as the exact numerical values are not explicitly reported.}
\label{tab:hessian_comparison}
\end{table}

\clearpage

\subsection{CBR Analysis on a Real-World Relation Extraction Dataset}
\label{sec:docred}

To examine whether the CBR representation emerges in real-world data, we analyze the CBR index using the Re-DocRED dataset~\cite{tan2022revisiting}, a widely used benchmark for document-level relation extraction. The CBR index is annotated according to the IRS schema described in Section~\secref{sec:irs_subspace}. Specifically, the indices are assigned based on the introduction order of relation triples provided in the dataset. An example annotated with the CBR index is shown below. We train the CBR projection matrix on the training set and evaluate it on the development set. If an attribute contains multiple tokens, we average their activations and then decode the corresponding CBR indices.

\begin{exe}
    \ex \label{ex:docred} \textit{William Paul “Bill” Cole III was born \underline{May 16, 1956}$_{[ei:1, ri:1]}$ and is an \underline{American}$_{[ei:1, ri:2]}$ businessman, politician and a former \underline{Republican}$_{[ei:1, ri:3]}$ member of the \underline{West Virginia Senate}$_{[ei:1, ri:4]}$, representing the 6th district from 2013 to 2017. ...}
\end{exe}

\begin{figure*}[htb]
    \centering 
\begin{subfigure}{0.485\textwidth}
  \includegraphics[width=\linewidth]{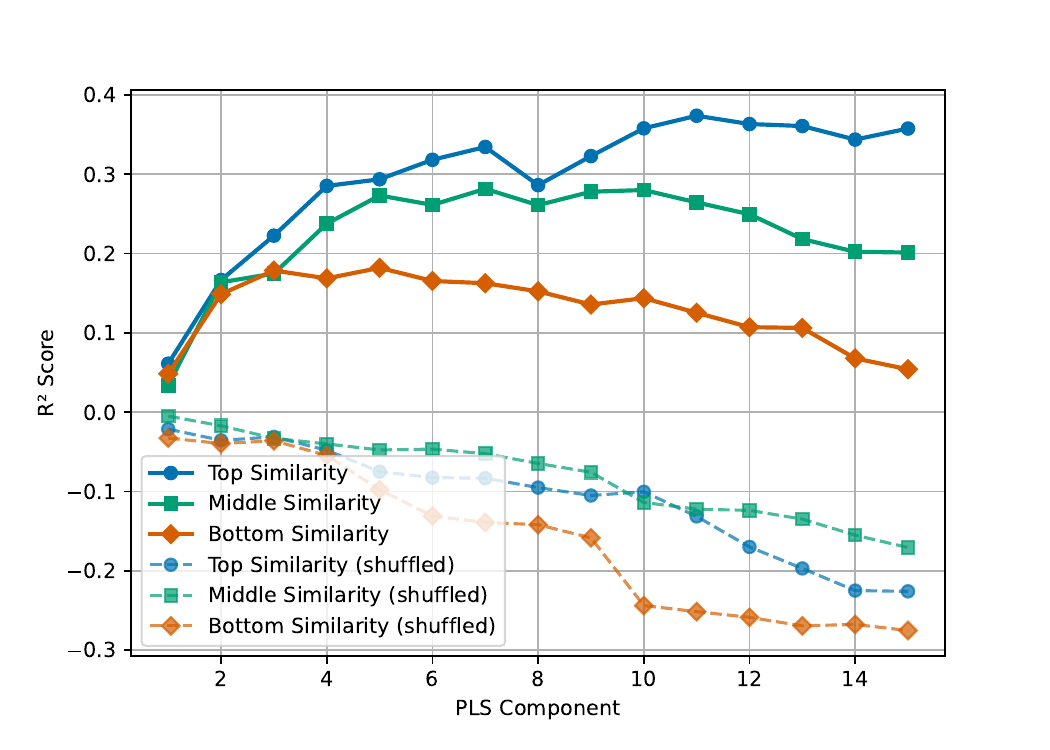}
  \caption{Decoding performance from Llama3-8B-Instruct}
\end{subfigure}\hfil 
\begin{subfigure}{0.485\textwidth}
  \includegraphics[width=\linewidth]{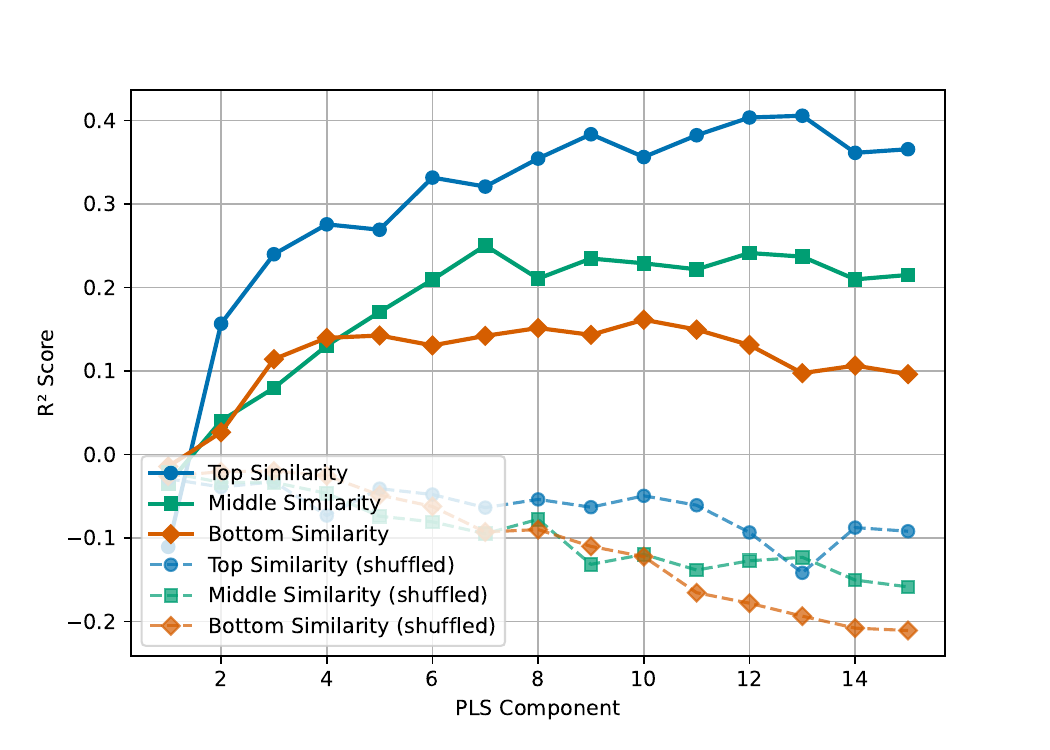}
  \caption{Decoding performance from Qwen3-8B}
\end{subfigure}\hfil 
\caption{Relationship between contextual similarity and CBR decoding performance ($R^2$) on Re-DocRED. “Top/Middle/Bottom Similarity” denote the groups of the Top $k$, Middle $k$, and Bottom $k$ samples. “Shuffled” indicates that the index assignments are randomly shuffled. Samples with higher contextual similarity to the training set exhibit higher $R^2$ scores, indicating that the CBR signal is more reliably recovered when similar contexts are present.}
\label{fig:docred}
\end{figure*}

To investigate how contextual similarity influences CBR recovery, we measure the embedding similarity between development samples and the training set using a sentence transformer. Based on this similarity score, the development samples are divided into three groups: the Top $k=50$ most similar samples, the Middle $k$ samples, and the Bottom $k$ least similar samples. We then compute the $R^2$ score of CBR index prediction for each group. The results are presented in Figure~\ref{fig:docred}. We observe that samples with higher contextual similarity achieve higher $R^2$ scores. This finding suggests that the CBR signal can be identified in the real-world dataset when the target sample shares strong contextual similarity with examples in the training set.

\clearpage

\subsection{Detection of CBR related Heads}
\label{sec:irs_head}

To identify CBR-related attention heads, we perform head-level activation patching on the final token (i.e., ``to'' in query part). Specifically, given an input~\footnote{we sampled $300$ instances for each context.} such as Sample~\ref{ex:context_b}, we create a corresponding counterfactual input (e.g., Sample~\ref{ex:context_c}) that differs in the CBR information associated with the final token. For example, the CBR information of ``to'' changes from $[ei:2:ri:2]$ to $[ei:2,ri:1]$ in Sample~\ref{ex:context_c}. We then patch the activation of each attention head individually from the last token of the counterfactual input into the original input and measure the resulting change in model behavior. Based on this procedure, we define a head-level patching score as $(\text{logit}_{\text{patch}} - \text{logit}_{\text{org}}) / \text{logit}_{\text{org}}$, where $\text{logit}_{\text{org}}$ and $\text{logit}_{\text{patch}}$ denote the target logit of answer (e.g., ``Berlin'') before and after patching, respectively. This score quantifies the contribution of each attention head to encoding CBR information.

\begin{exe}
    \ex\label{ex:context_b}\textbf{Input}: Sean, who hails from Phoenix, \colorbox{yellow}{currently resides in Perm}. ... Meanwhile, Jose was born in Austin and \colorbox{yellow}{is now living in Berlin}. ... Based on the context, given like Sean to Perm$_{[ei:1,ri:2]}$, Jose to$_{[ei:2,ri:2]}$ (Answer: Berlin)
    \ex\label{ex:context_c}\textbf{Counter Input}: Sean, who \colorbox{yellow}{hails from Perm}, currently resides in Phoenix. ... Meanwhile, Jose \colorbox{yellow}{was born in Berlin} and is now living in Austin. ... Based on the context, given like Sean to Perm$_{[ei:1,ri:1]}$, Jose to$_{[ei:2,ri:1]}$ (Answer: Berlin)
\end{exe}

\begin{figure*}[!htbp]
    \centering 
\begin{subfigure}{0.35\textwidth}
  \includegraphics[width=\linewidth]{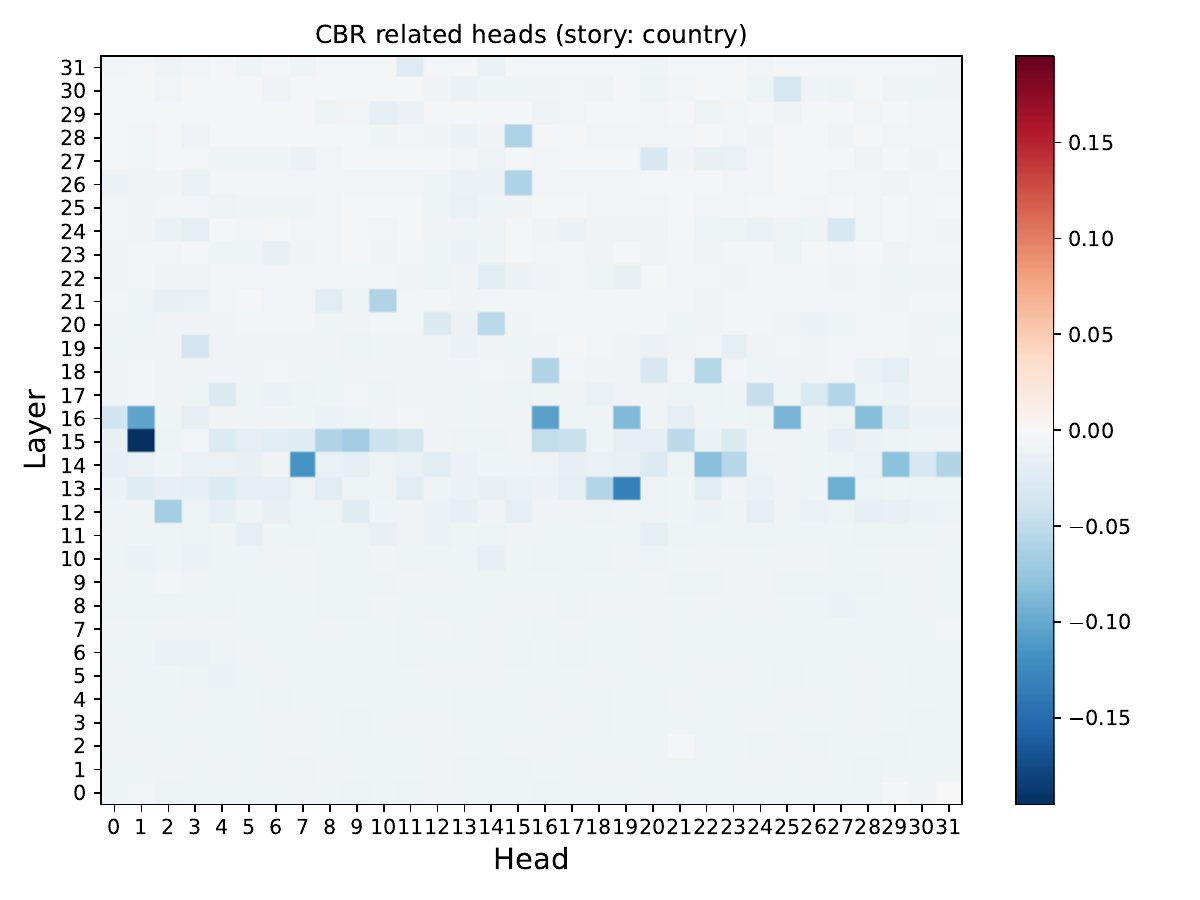}
  \caption{$C_{country}$}
\end{subfigure}\hfil 
\begin{subfigure}{0.35\textwidth}
  \includegraphics[width=\linewidth]{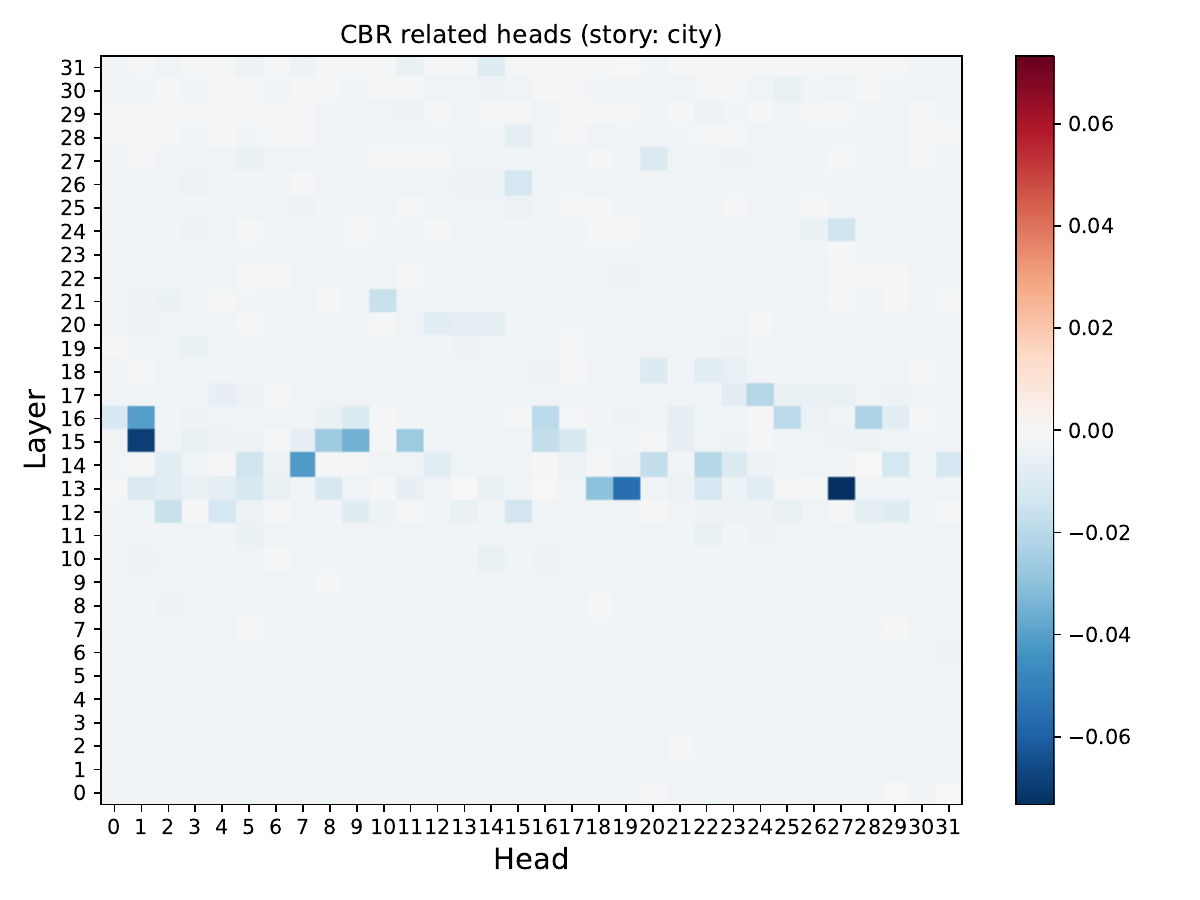}
  \caption{$C_{city}$}
\end{subfigure}\hfil 
\begin{subfigure}{0.3\textwidth}
  \includegraphics[width=\linewidth]{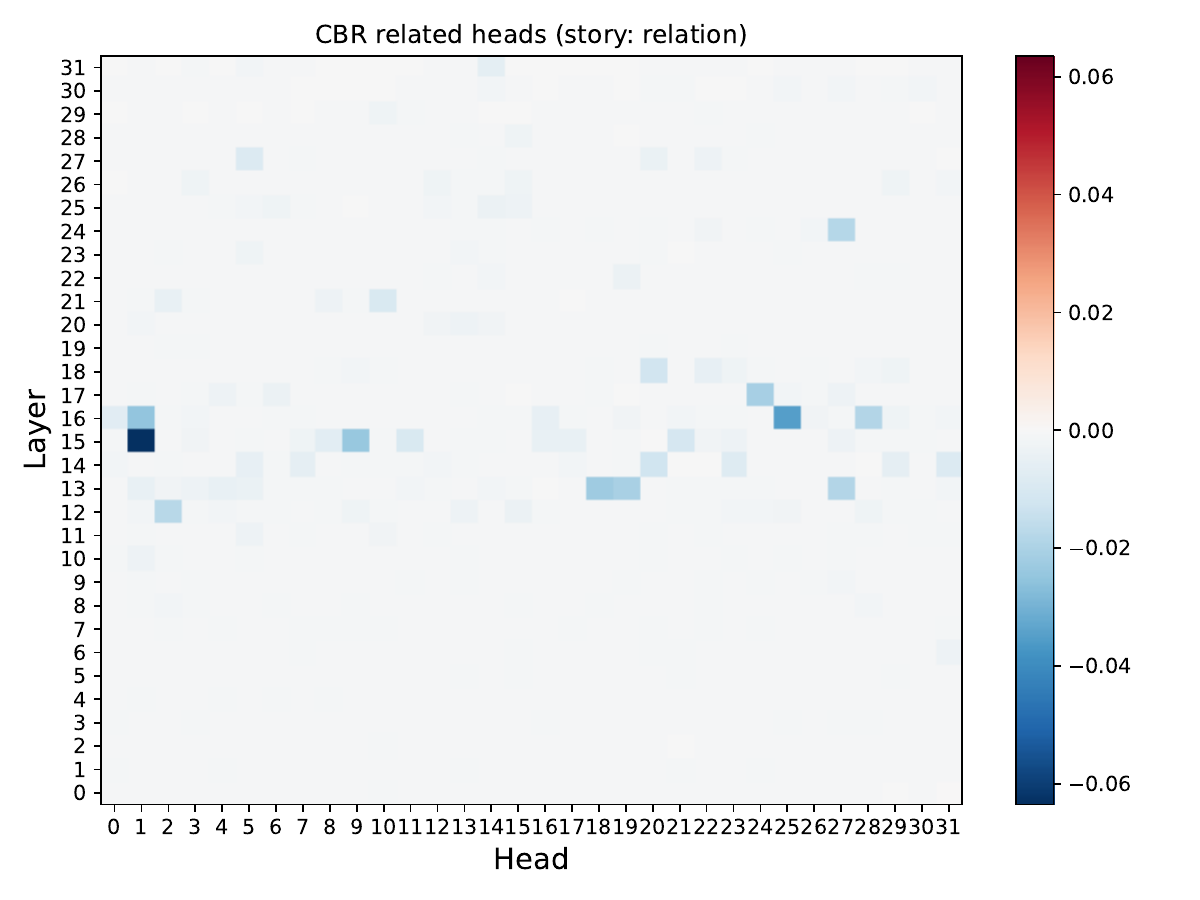}
  \caption{$C_{relation}$}
\end{subfigure}\hfil
\begin{subfigure}{0.3\textwidth}
  \includegraphics[width=\linewidth]{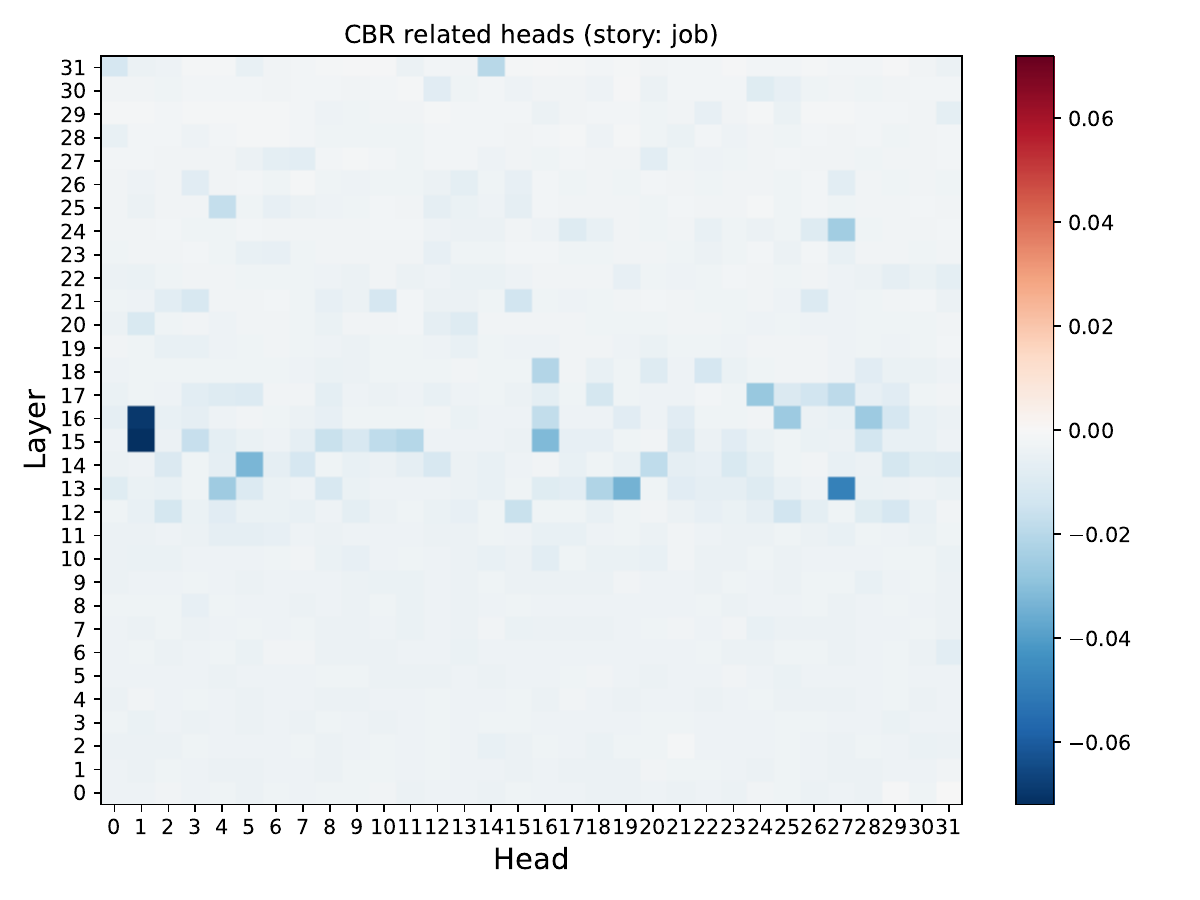}
  \caption{$C_{job}$}
\end{subfigure}\hfil 
\begin{subfigure}{0.3\textwidth}
  \includegraphics[width=\linewidth]{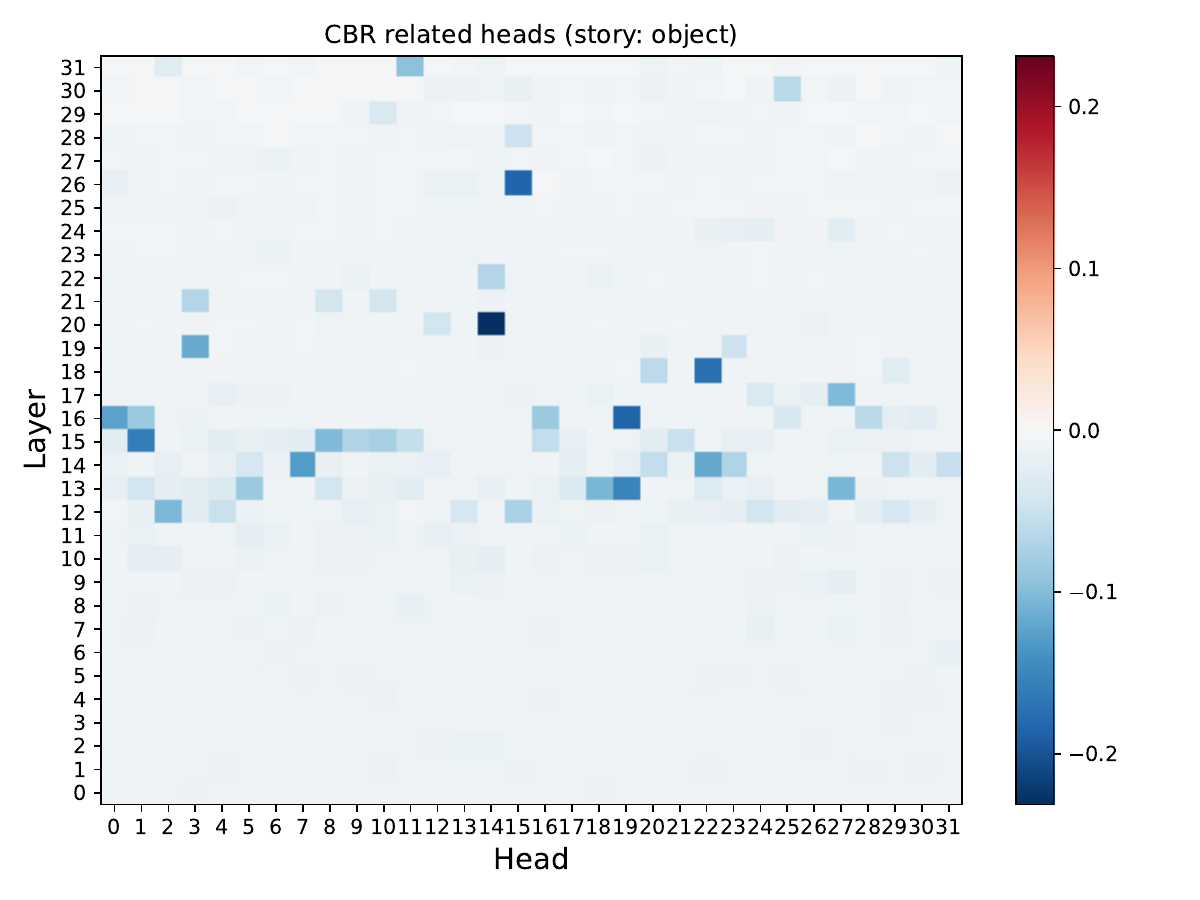}
  \caption{$C_{object}$}
\end{subfigure}\hfil 
\caption{Visualization of the CBR related heads on Llama3-8b-Instruct. Each cell shows the normalized logit change induced by patching a single attention head on the final token. Heads with high scores are primarily concentrated in middle layers, indicating their involvement in CBR-based attribute retrieval.}
\label{fig:irs_head_llama}
\end{figure*}

\begin{figure*}[!htbp]
    \centering 
\begin{subfigure}{0.35\textwidth}
  \includegraphics[width=\linewidth]{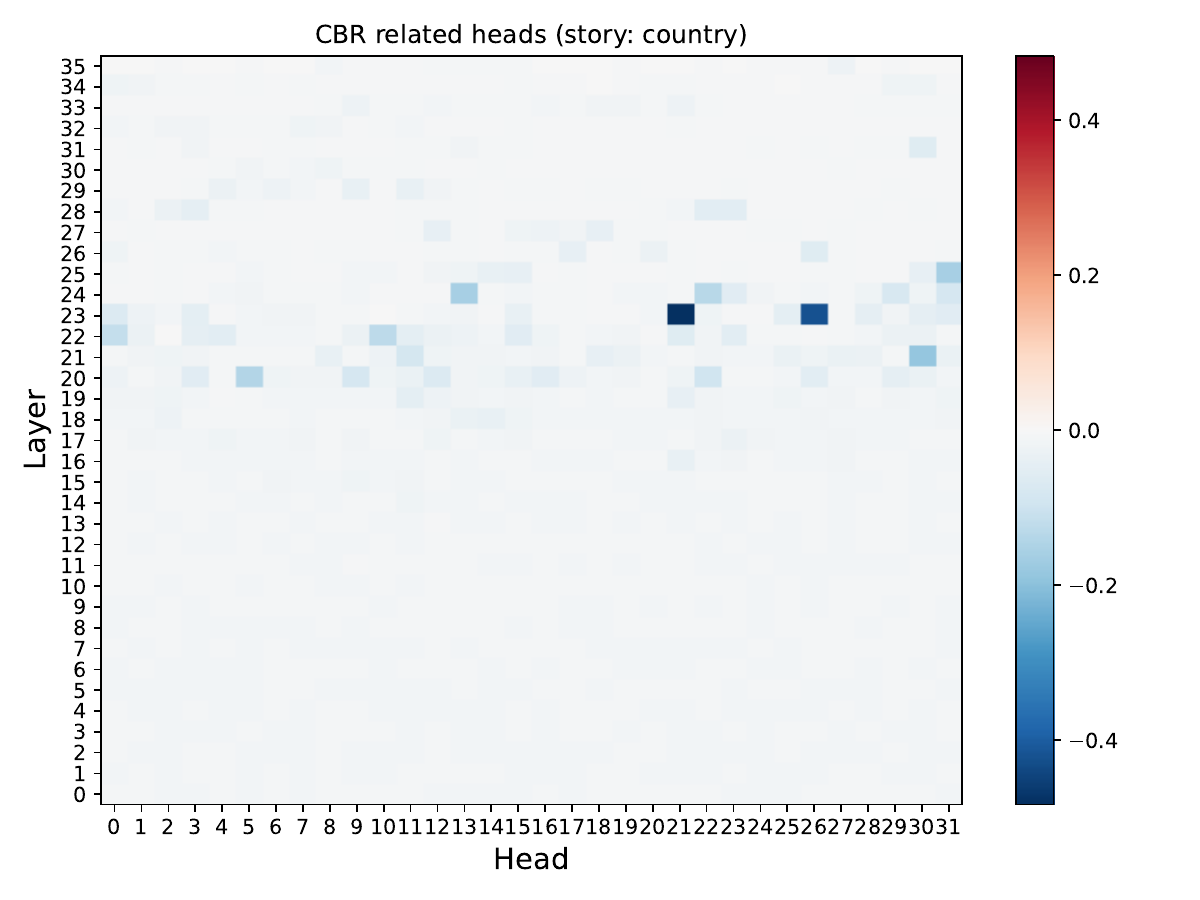}
  \caption{$C_{country}$}
\end{subfigure}\hfil 
\begin{subfigure}{0.35\textwidth}
  \includegraphics[width=\linewidth]{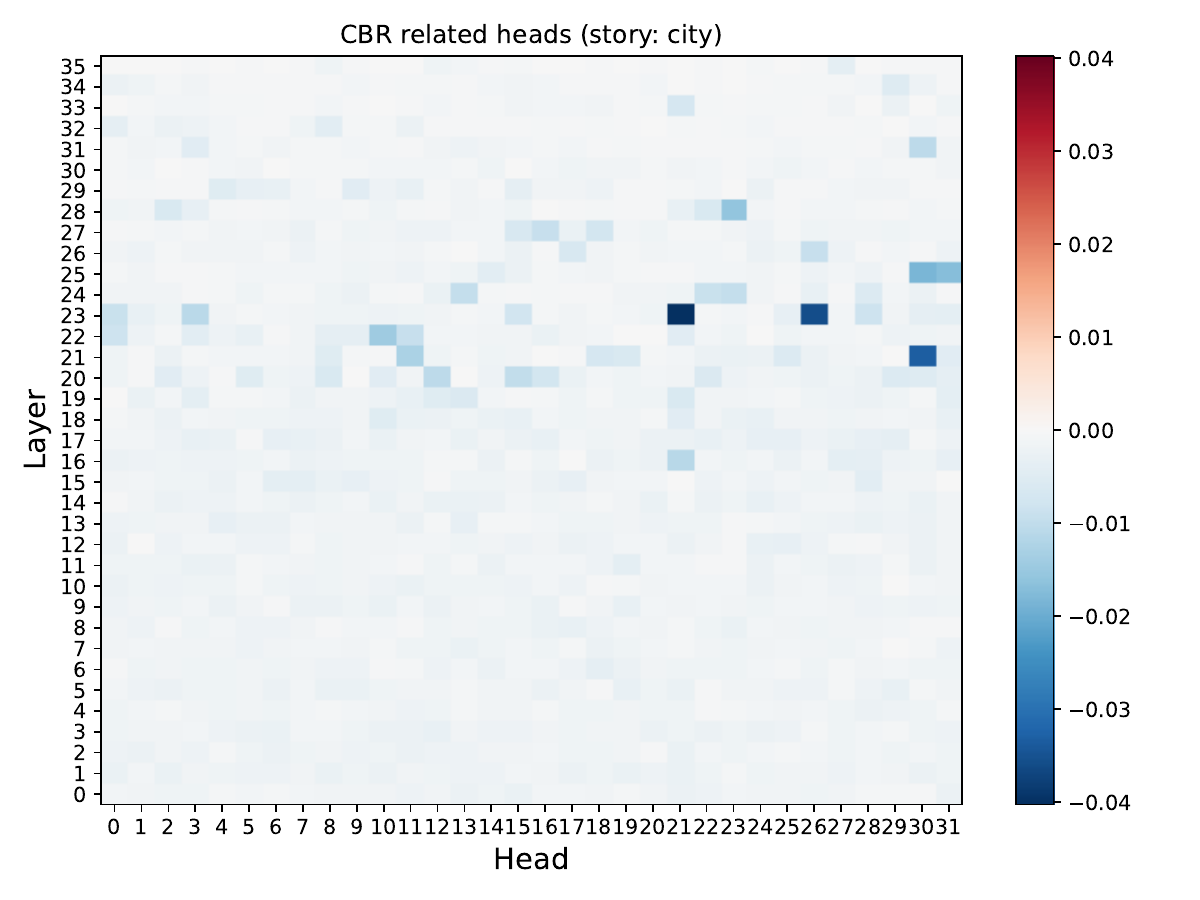}
  \caption{$C_{city}$}
\end{subfigure}\hfil 
\begin{subfigure}{0.3\textwidth}
  \includegraphics[width=\linewidth]{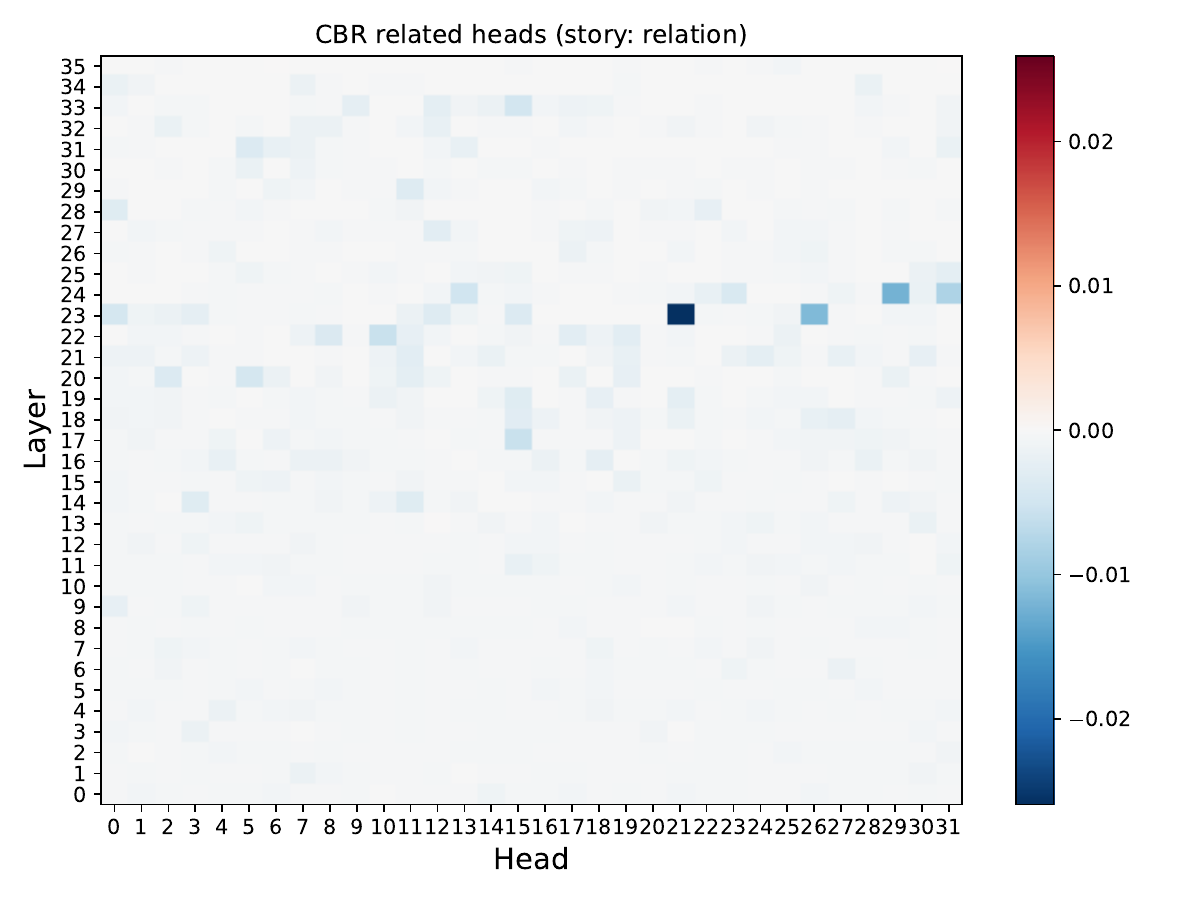}
  \caption{$C_{relation}$}
\end{subfigure}\hfil
\begin{subfigure}{0.3\textwidth}
  \includegraphics[width=\linewidth]{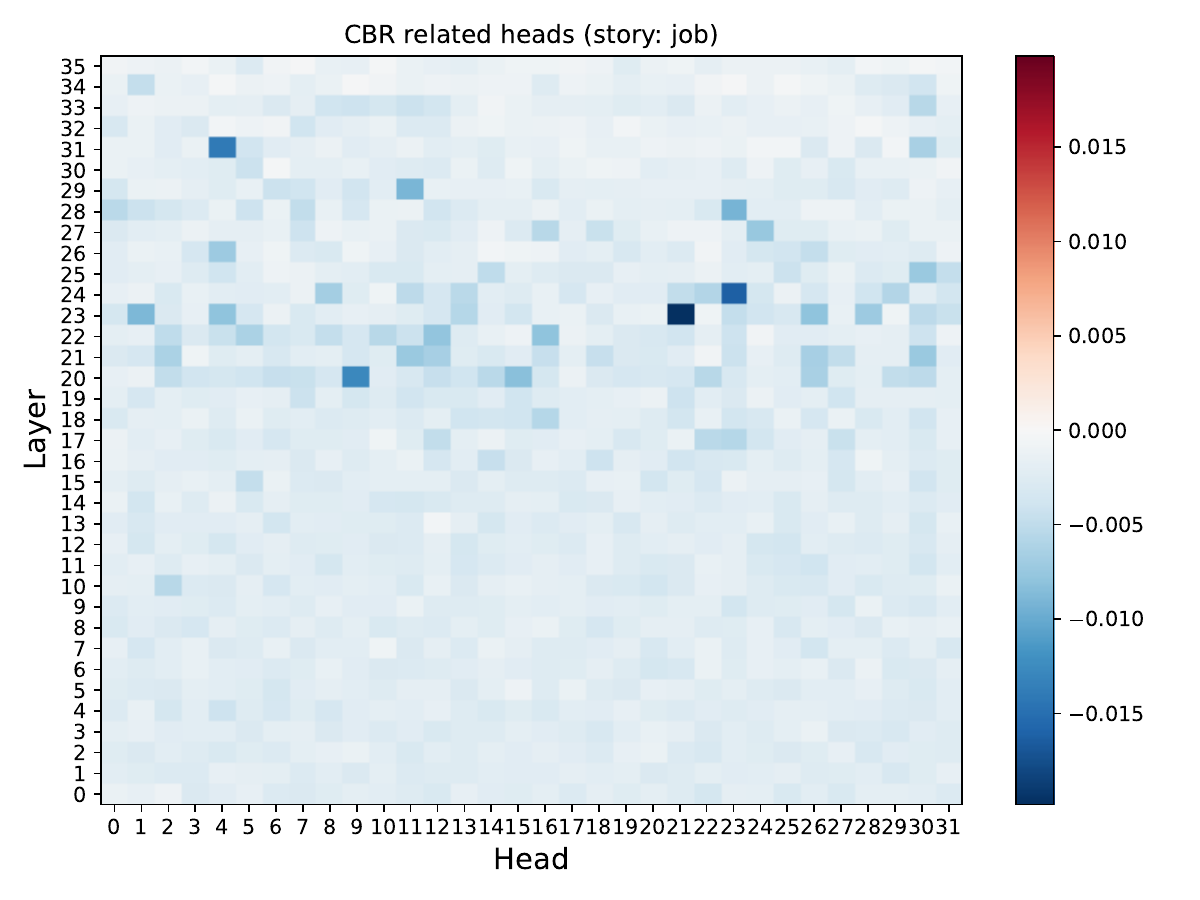}
  \caption{$C_{job}$}
\end{subfigure}\hfil 
\begin{subfigure}{0.3\textwidth}
  \includegraphics[width=\linewidth]{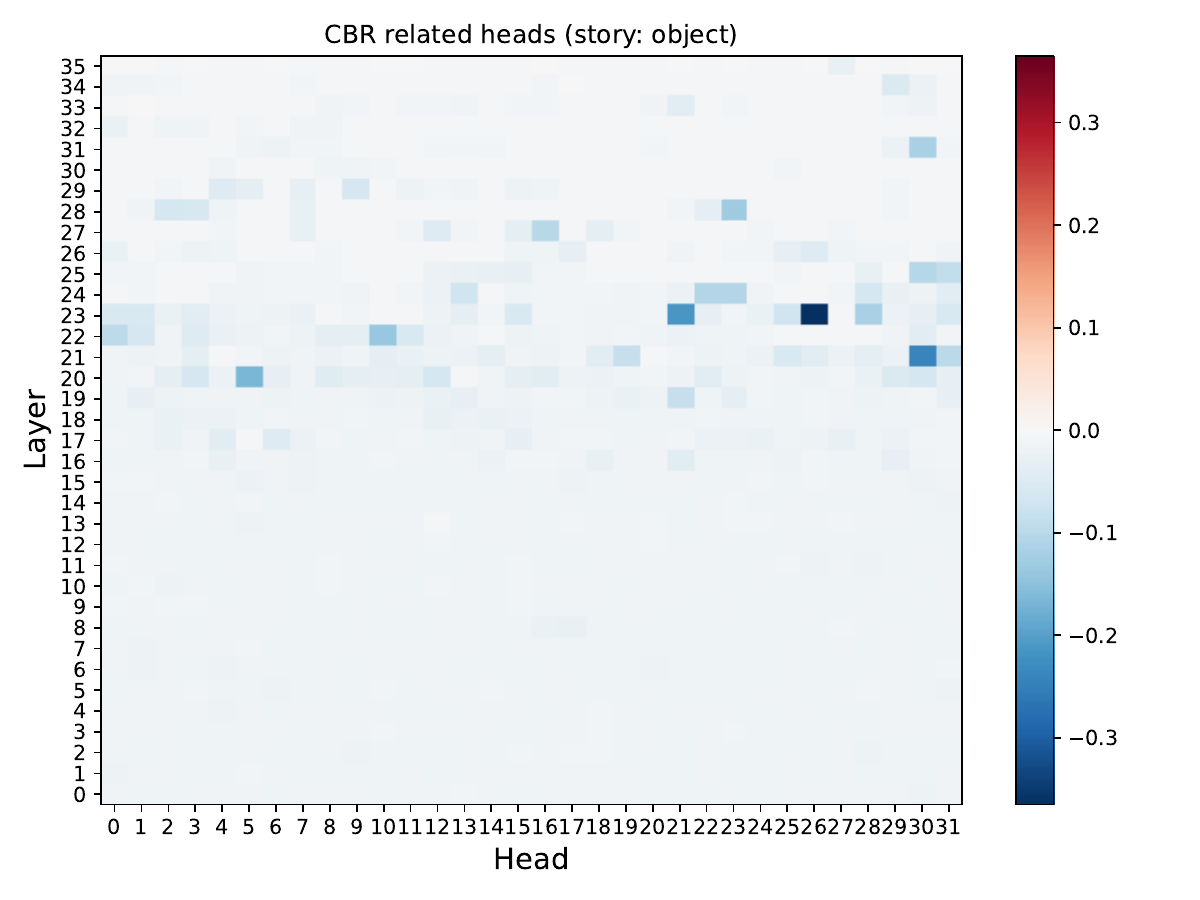}
  \caption{$C_{object}$}
\end{subfigure}\hfil 
\caption{Visualization of the CBR related heads on Qwen3-8b.}
\label{fig:irs_head_qwen}
\end{figure*}

\begin{figure*}[!htbp]
    \centering 
\begin{subfigure}{0.35\textwidth}
  \includegraphics[width=\linewidth]{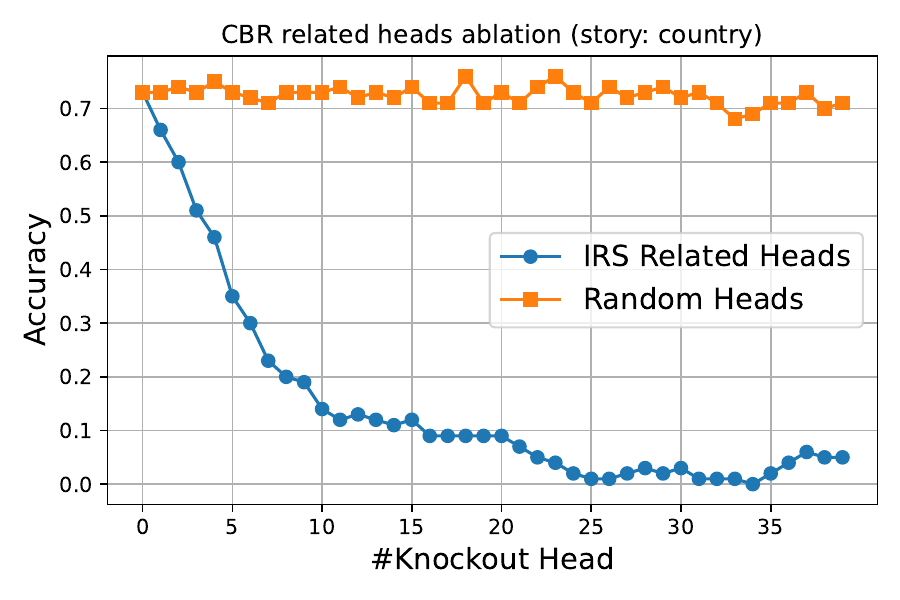}
  \caption{$C_{country}$}
\end{subfigure}\hfil 
\begin{subfigure}{0.35\textwidth}
  \includegraphics[width=\linewidth]{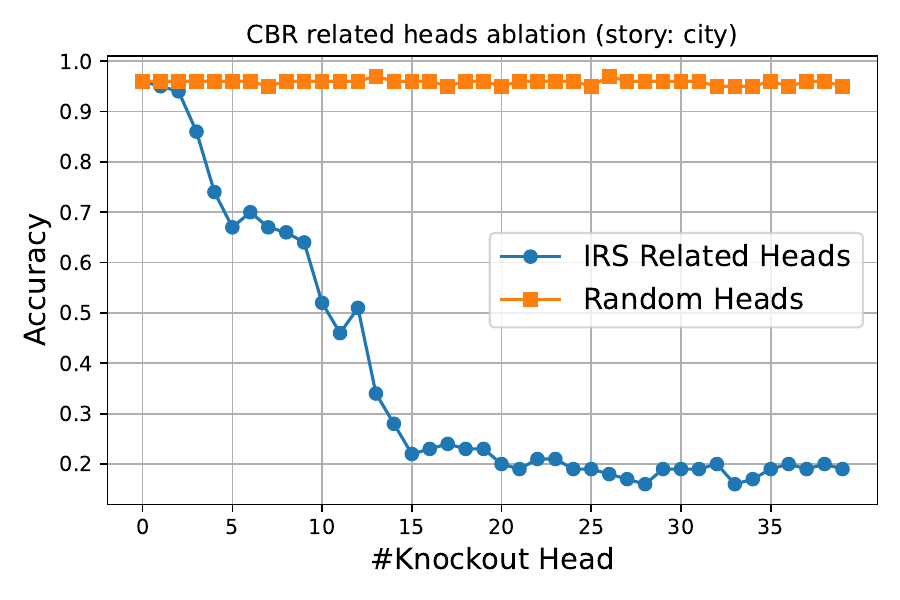}
  \caption{$C_{city}$}
\end{subfigure}\hfil 
\begin{subfigure}{0.3\textwidth}
  \includegraphics[width=\linewidth]{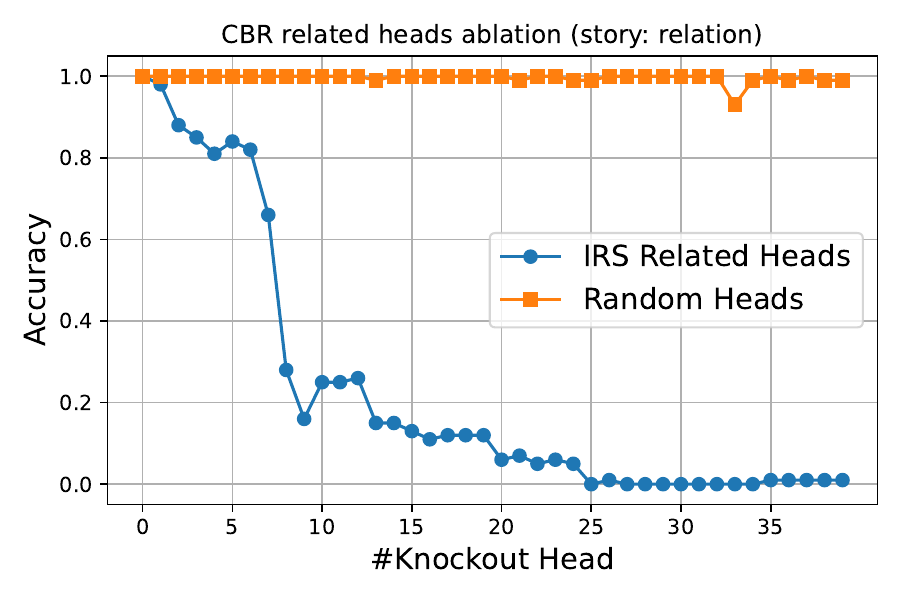}
  \caption{$C_{relation}$}
\end{subfigure}\hfil
\begin{subfigure}{0.3\textwidth}
  \includegraphics[width=\linewidth]{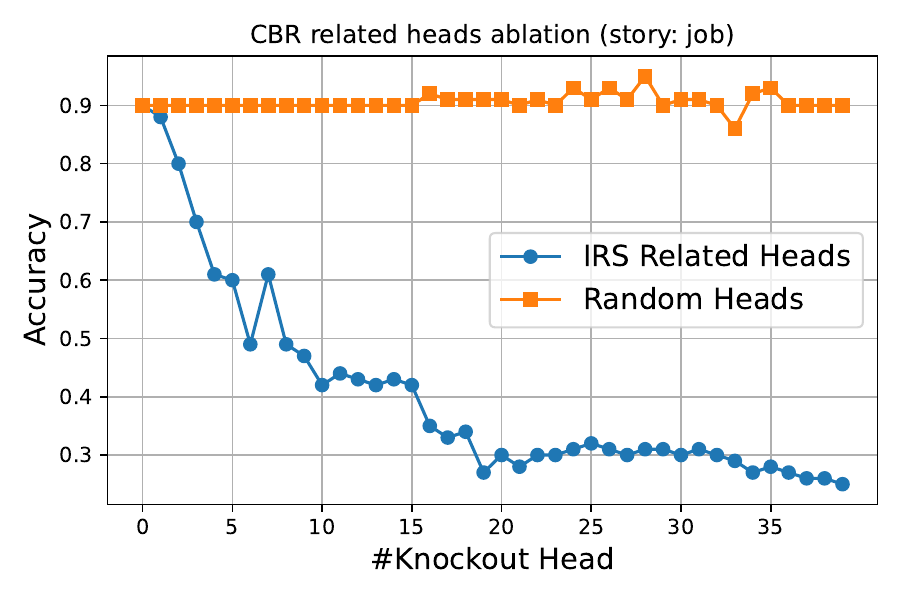}
  \caption{$C_{job}$}
\end{subfigure}\hfil 
\begin{subfigure}{0.3\textwidth}
  \includegraphics[width=\linewidth]{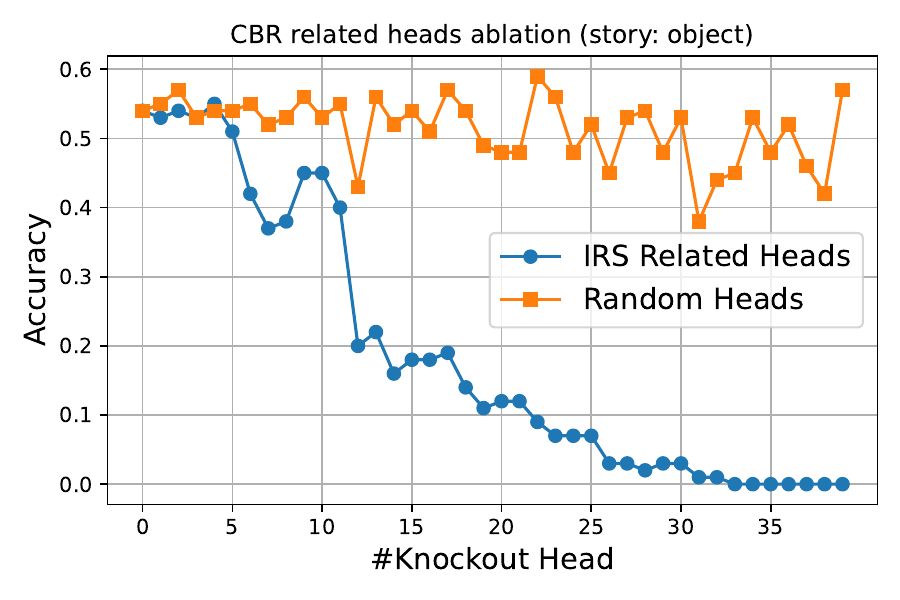}
  \caption{$C_{object}$}
\end{subfigure}\hfil 
\caption{CBR related head knockout on Llama3-8b-Instruct, where heads are ablated in descending order of patching score using mean ablation. Comparing with removing Random Heads, removing high-scoring heads causes a sharp drop in accuracy, indicating their critical role in CBR-based retrieval.}
\label{fig:head_ablate_llama}
\end{figure*}

\begin{figure*}[!htbp]
    \centering 
\begin{subfigure}{0.35\textwidth}
  \includegraphics[width=\linewidth]{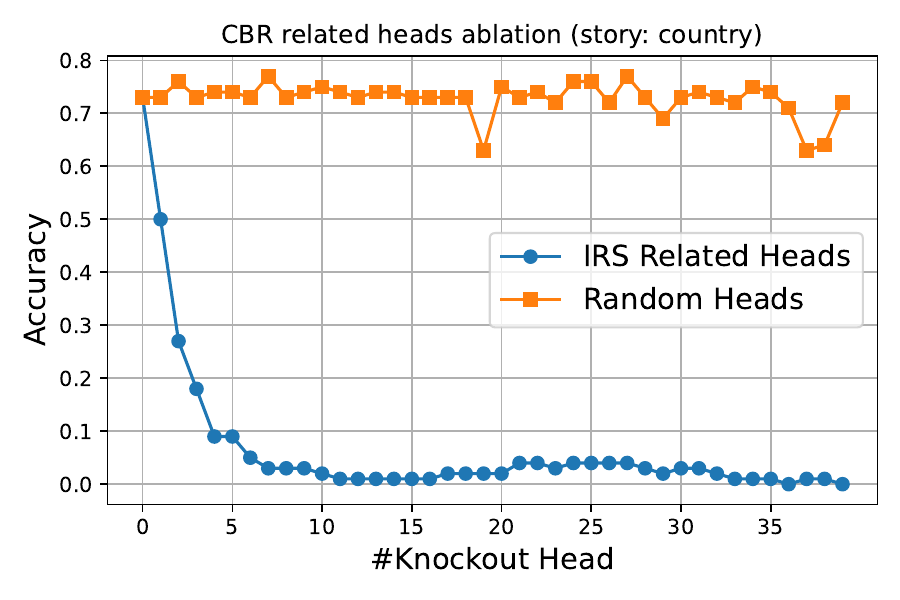}
  \caption{$C_{country}$}
\end{subfigure}\hfil 
\begin{subfigure}{0.35\textwidth}
  \includegraphics[width=\linewidth]{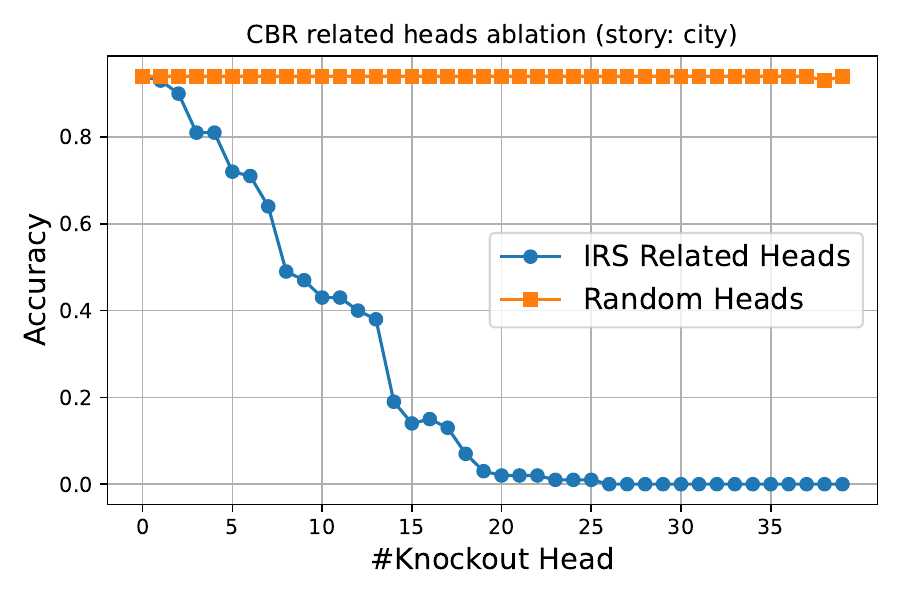}
  \caption{$C_{city}$}
\end{subfigure}\hfil 
\begin{subfigure}{0.3\textwidth}
  \includegraphics[width=\linewidth]{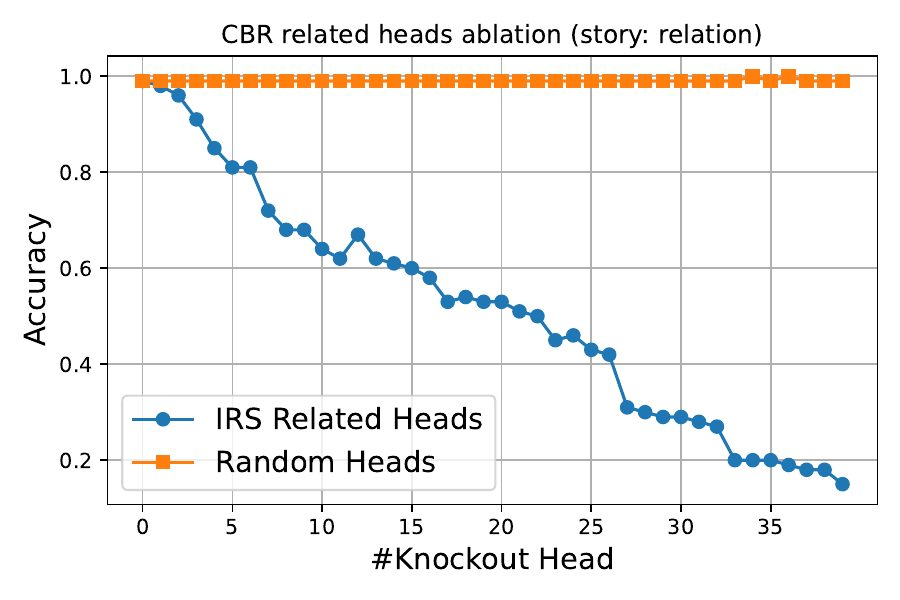}
  \caption{$C_{relation}$}
\end{subfigure}\hfil
\begin{subfigure}{0.3\textwidth}
  \includegraphics[width=\linewidth]{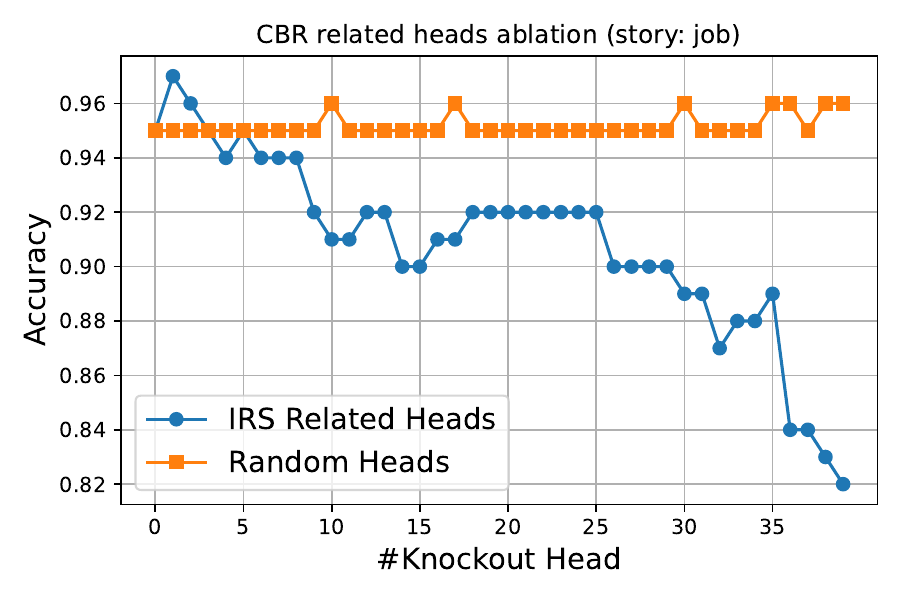}
  \caption{$C_{job}$}
\end{subfigure}\hfil 
\begin{subfigure}{0.3\textwidth}
  \includegraphics[width=\linewidth]{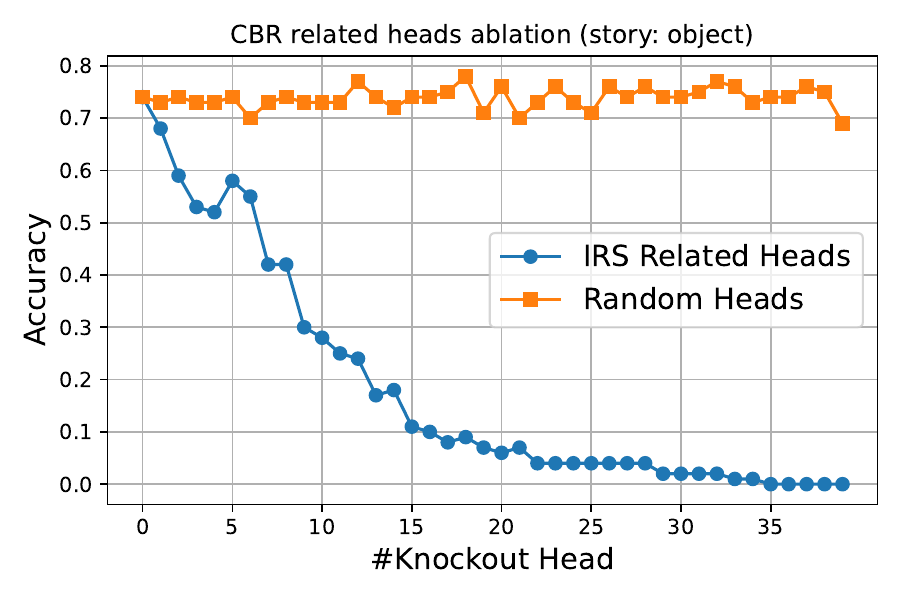}
  \caption{$C_{object}$}
\end{subfigure}\hfil 
\caption{CBR related head knockout on Qwen3-8b.}
\label{fig:head_ablate_qwen}
\end{figure*}

The patching scores are visualized in Figure~\ref{fig:irs_head_llama} and \ref{fig:irs_head_qwen}, revealing that CBR-related heads are primarily concentrated in the middle layers and are limited to a relatively small subset of heads. To further assess the functional importance of these heads, we ablate them according to their patching score ranking. Using mean ablation~\cite{wang2022interpretability} to selectively knock out the identified heads, we evaluate the resulting performance, with accuracies reported in Figure~\ref{fig:head_ablate_llama} and \ref{fig:head_ablate_qwen}. The results show that ablating CBR-related heads leads to a substantial degradation in accuracy, providing strong evidence that these heads play a critical role in CBR-based attribute retrieval.


\clearpage

\subsection{Additional Experimental Settings}
\paragraph{Machine} Causal intervention experiments on CBR subspace are conducted on 48 GB NVIDIA RTX 6000 Ada Generation GPUs. 

\paragraph{Other Hyperparameter} We set the dimensionality to 15 in Figures~\ref{fig:generality_llama} and \ref{fig:consistency_ablate_shuffle}, and to 4 in Figure~\ref{fig:semantic_pattern}. In Equation~\ref{eq:irs_sampling2}, we set $\alpha$ as $-0.4$. The value of $\alpha$ in Equation~\ref{eq:irs_steer2} is selected via beam search over the range $[0.4,1.6]$.

\paragraph{Software} The main libraries used in this work include Numpy~\cite{harris2020array}, Scikit-learn~\cite{pedregosa2011scikit}, Pytorch~\cite{paszke2019pytorch}, Transformers~\cite{wolf2020transformers} and Matplotlib~\cite{hunter2007matplotlib}.

\subsection{About AI Assistants for Writing}
In this paper, ChatGPT is used to assist with language polishing, grammar checking and minor simplification of visualization code. Its role is only limited to improving wording accuracy and representational clarity. It is important to note that all scientific ideas, methodologies, analyses and discussions are entirely derived from the authors’ own research and expertise.

\end{document}